\pdfoutput=1

\documentclass[11pt]{article}

\usepackage[final]{acl}

\usepackage{times}
\usepackage{latexsym}
\usepackage{booktabs}
\usepackage{amsmath}
\usepackage{multirow}
\usepackage{url}
\usepackage{amsfonts}
\usepackage{graphicx} 
\usepackage{pifont}
\usepackage{enumitem}

\usepackage{titletoc}

\usepackage[T1]{fontenc}

\usepackage[utf8]{inputenc}

\usepackage{microtype}

\usepackage{inconsolata}

\usepackage{graphicx}

\title{What Happened in LLM Layers when Trained for Fast vs. Slow Thinking: A Gradient Perspective}

\author{
  Ming Li \\
  University of Maryland \\
  \texttt{minglii@umd.edu}
  \And
  \hspace{-1cm}Yanhong Li \\
  \hspace{-1cm}University of Chicago \\
  \hspace{-1cm}\texttt{yanhongli@uchicago.edu}
  \And
  Tianyi Zhou \\
  University of Maryland \\
  \texttt{tianyi.david.zhou@gmail.com}
}

\begin{document}
\maketitle
\begin{abstract}
What makes a difference in the post-training of LLMs?
We investigate the training patterns of different layers in large language models (LLMs) through the lens of the gradient. 
We are specifically interested in how fast vs. slow thinking affects the layer-wise gradients, given the recent popularity of training LLMs on reasoning paths such as chain-of-thoughts (CoT) and process rewards. 
In our study, fast thinking without CoT leads to larger gradients and larger differences of gradients across layers than slow thinking (Detailed CoT), indicating the learning stability brought by the latter. 
Additionally, we study whether the gradient patterns can reflect the correctness of responses when training different LLMs using slow vs. fast thinking paths. 
The results show that the gradients of slow thinking can distinguish correct and irrelevant reasoning paths. As a comparison, we conduct similar gradient analyses on non-reasoning knowledge learning tasks, on which, however, trivially increasing the response length does not lead to similar behaviors of slow thinking. 
Our study strengthens fundamental understandings of LLM training and sheds novel insights on its efficiency and stability, which pave the way towards building a generalizable System-2 agent. 
Our code, data, and gradient statistics can be found in: \url{https://github.com/MingLiiii/Layer_Gradient}.
\end{abstract}

\section{Introduction}

Large language models (LLMs) excel at various complex tasks \cite{zhao2023survey, Xu2024ASO}. But their complexity notoriously makes them ``black-box'' whose inner mechanisms and training behaviors remain mysterious~\cite{zhao2023explainabilitylargelanguagemodels, singh2024rethinking}. 
How do they acquire reasoning capabilities or knowledge? 
When do they make mistakes, and why?
What change was made to each layer during training?
This lack of transparency extends to issues like unintentional generation of harmful or biased content \cite{huang2024trustllmtrustworthinesslargelanguage, Li2024CanLS} or hallucinations \cite{huang2023surveyhallucinationlargelanguage} and might hinder further understanding and mitigation of them. \looseness-1


Interpretable machine learning either develops models that are inherently interpretable~\cite{rudin2022interpretable} or adopts post-hoc interpretability methods~\cite{krishna2024post}, which do not alter the underlying model architecture but analyze models after training completes~\cite{gurnee2023finding, zou2023representationengineeringtopdownapproach, wang-etal-2023-label, wu2024interpretability}. 
Despite the broad advancements in interpreting static models, \textbf{gradients on dynamic training patterns of LLMs} remain underexplored, especially on how these gradients scale and distribute across different layers. 
Such an understanding is crucial as it directly reflects how the LLMs perceive training data and align with it.  
Recently, there has been a growing trend for layer-related methods for LLMs: \citet{gao2024higherlayersneedlora} propose that higher layers of LLM need more LoRA; \citet{li2024safetylayersalignedlarge} identify some of the layers in LLM related to safety, etc.~\cite {men2024shortgptlayerslargelanguage, chen2024streamliningredundantlayerscompress}.  
However, for the layer-wise analysis of LLMs, current research mainly employs probing methods~\cite{alain2017understanding, ju-etal-2024-large, jin2024exploringconceptdepthlarge, ye2024physicslanguagemodels21} that assess model behavior by observing changes in performance when certain layers are modified or removed~\cite{wang-etal-2023-label, fan2024layersllmsnecessaryinference}. 
These studies have been instrumental in illustrating how different layers capture and process various types of 
information, while they often do not provide direct insights into the gradients that drive the learning process.
Hence, we are motivated to move a step forward by directly investigating the layer-wise gradients inside LLMs. 
\looseness-1

\begin{table*}[!tbh]
\centering
\small
\scalebox{0.72}{
\begin{tabular}{l|l|cc|ccc|cc}
\toprule
\textbf{Task Type} & \textbf{Dataset} & \textbf{Correct} & \textbf{Irrelevant} & \textbf{None CoT} & \textbf{Simplified CoT} & \textbf{Detailed CoT (GPT4o)} & \textbf{Base LLMs} & \textbf{Instructed LLMs} \\
\midrule
\multirow{5}{*}{Math} 
 & AQuA & \ding{51} & \ding{51} & \ding{51} & \ding{51} & \ding{51} & \ding{51} & \ding{51} \\
 & GSM8K & \ding{51} & \ding{51} & \ding{51} & \ding{51} & \ding{51} & \ding{51} & \ding{51} \\
 & MATH-Algebra & \ding{51} & \ding{51} &  & \ding{51} & \ding{51} & \ding{51} & \ding{51} \\
 & MATH-Counting & \ding{51} & \ding{51} &  & \ding{51} & \ding{51} & \ding{51} & \ding{51} \\
 & MATH-Geometry & \ding{51} & \ding{51} &  & \ding{51} & \ding{51} & \ding{51} & \ding{51} \\
 \midrule
 \multirow{4}{*}{Commonsense} 
 & StrategyQA & \ding{51} & \ding{51} & \ding{51} & \ding{51} & \ding{51} & \ding{51} & \ding{51} \\
 & ECQA & \ding{51} & \ding{51} & \ding{51} & \ding{51} & \ding{51} & \ding{51} & \ding{51} \\
 & CREAK & \ding{51} & \ding{51} & \ding{51} & \ding{51} & \ding{51} & \ding{51} & \ding{51} \\
 & Sensemaking & \ding{51} & \ding{51} & \ding{51} & \ding{51} & \ding{51} & \ding{51} & \ding{51} \\
 \midrule
 \multirow{4}{*}{Wiki Knowledge} 
 & Popular (Length 100) & \ding{51} & \ding{51} &  \ding{51} &  &   & \ding{51} & \ding{51} \\
 & Popular (Length 500) & \ding{51} & \ding{51} &  \ding{51} &   &   & \ding{51} & \ding{51} \\
 & Popular (Length 1000) & \ding{51} & \ding{51} &  \ding{51} &   &   & \ding{51} & \ding{51} \\
 & Unpopular            & \ding{51} & \ding{51} &  \ding{51} &   &   & \ding{51} & \ding{51} \\
\bottomrule
\end{tabular}
}
\caption{
The scope of our study. \textbf{We compare the gradient patterns across different layers when training pretrained base LLMs vs. instruction-finetuned LLMs} using \textbf{correct vs. irrelevant responses, slow vs. fast thinking} (\textit{None CoT}, \textit{Simplified CoT}, and \textit{Detailed CoT} generated by GPT-4o) responses, on \textbf{three types of tasks: Math, Commonsense Reasoning, and Wiki Knowledge Learning}. The comparison of slow vs. fast thinking only applies to the first two types of tasks, and it is replaced by the comparison between different lengths of responses on the third type of task. Our study is conducted on 5 pretrained base LLMs and 5 instruction-finetuned LLMs. 
}
\label{tbl:scope}
\vspace{-4mm}
\end{table*}

Our study focuses on the post-training gradient of LLMs for instruction-tuning on instruction-response pairs \cite{mishra2021cross, wei2022finetuned, wang-etal-2023-self-instruct, alpaca, xu2023wizardlm, cherry, Li2024SuperfilteringWD, Li2024SelectiveRS, zhang2023instruction}. 
Instead of finetuning LLMs, we investigate the layer-wise gradients of 5 base LLMs and 5 instruction-finetuned LLMs on different data, including (1) three types of tasks, including Math, Commonsense Reasoning, and Knowledge Extraction, with several datasets per task type;  
(2) correct vs. irrelevant responses; and
(3) fast vs. slow thinking \cite{kahneman2011thinking, li202512surveyreasoning}, which corresponds to different levels of Chain of Thought (CoT) \cite{wei2023chainofthoughtpromptingelicitsreasoning} reasoning paths. 
Table \ref{tbl:scope} summarizes the scope of our study.
\looseness-1


Our study is based on comparisons of layer-wise gradients in terms of their spectral properties achieved by Singular Value Decomposition (SVD), 
focusing particularly on the projection layers for \textit{Query}, \textit{Key}, \textit{Value}, and \textit{Output} in transformer architectures \cite{NIPS2017_3f5ee243}. 
Specifically, we measure the gradient by its nuclear norm, compare the gradient norms across layers, and measure the sensitivity of gradients to different training data or initial models by the difference in gradient norm. 
These metrics serve as quantitative tools for examining the training behaviors and shed novel insights that could inform more efficient training strategies and analyses of model stability.
\citet{li2025instruction} pushes our analytical method further by introducing the analysis on effective ranks and gradient similarities, and providing a unified view on the effects of different data quality metrics for instruction vs. reasoning data.

\textbf{Main Contribution.}
\footnote{In addition to the observations and analysis included in this paper, all the gradient statistics (that cost thousands of GPU hours) within our experimental scope will be released in our GitHub repository. \textit{At the sample level}, the instruction-response pair, and the corresponding loss value are included. \textit{At the layer level}, the mean, maximum, and minimum values, the Frobenius norm, and Nuclear norm, and the maximum and minimum singular values of each gradient matrix are included. We hope these gradient statistics can contribute to the community's understanding of the gradient behaviors for different settings. } 
This paper investigates the behaviors of the gradient across different layers of LLMs through a spectral analysis of the layer-wise gradients. We compare the gradients of slow vs. fast thinking rationals when training different initial models using correct vs. irrelevant responses on different tasks. The difference in gradients reflects how these factors affect training dynamics, and, reversely, how sensitive LLM training is to these factors. Our observations reveal previously unrecognized patterns and shed novel insights for improving the stability and efficiency of LLM training. 
Our key findings are highlighted in the following:\looseness-1
\begin{enumerate}[ leftmargin=4mm]
    \item Training LLMs for slow thinking (Detailed CoT) leads to similar gradient norms of different layers, while fast thinking (Simplified/None CoT) results in larger gradients of earlier layers and drastic differences across layers. 
    \vspace{-2mm}
    \item The gradient of slow thinking (Detailed CoT) helps distinguish correct responses from irrelevant responses. Without CoT, the gradient patterns of the two types of responses are similar.      
    \vspace{-5mm}
    \item The instruction-finetuned LLMs do not show superior capability over pre-trained base LLMs in identifying incorrect reasoning paths.
    \vspace{-2mm}
    \item The above observations on reasoning tasks (math and commonsense) cannot be extended to knowledge learning tasks, where simply increasing response length does not show similar gradient patterns as slow thinking.     
\end{enumerate}


\section{Methodology}

\subsection{Preliminaries}

In our experiments, we utilize the most commonly used instruction tuning settings to investigate the gradients for LLM fine-tuning. Given an instruction-tuning dataset $D$, each data sample is represented by a tuple $(ins, res)$, where $ins$ represents the instruction and $res$ represents the corresponding response. Let $p_\theta$ denote the LLM with parameters $\theta$. In the instruction tuning setting, $p_\theta$ is typically fine-tuned by minimizing the following loss on each sample $(ins, res)$, in which $res_j$ represents the $j_{th}$ token of response $res$, $ res_{<j}$ represents the tokens before $res_j$, and $l$ represents the token length of $res$: 
\looseness-1

\vspace{-5mm}
\begin{align}
   L_\theta = \frac{1}{l} \sum_{j=1}^{l} - \log p_\theta\left(res_j | ins, res_{<j} \right),
\end{align}
\vspace{-5mm}


\subsection{Gradient Representation}

The attention mechanism \cite{NIPS2017_3f5ee243} is one of the most critical parts of modern LLMs, and it dominates the behavior of LLMs. 
Thus in this paper, we mainly focus on the gradients of the layers related to the attention mechanism, including the \textit{Query (Q)}, \textit{Key (K)}, \textit{Value (V)} projection layers and later output projection layer denoted as \textit{Output (O)}. 
\looseness-1
Considering the LLM contains $N$ attention layers in total, after the loss calculation and Backpropagation, the resulting gradient for each projection layer is a matrix with the same size as its weights, which can be notated as $G_{Q, i}$, $G_{K, i}$, $G_{V, i}$, and $G_{O, i}$ for the corresponding projection layers, where $i \in [0, N-1]$ represents the index of each layer in the LLM.  

Due to the dramatically large number of layers and parameters of the modern \textbf{Large} language models, it is unrealistic to directly investigate these large gradient matrices. 
Motivated by the advanced Singular Value Decomposition (SVD) technology used in this area \cite{biderman2024lora, carlini2024stealing}, we utilize Nuclear Norm (the $\ell_1$ norm of singular values) to represent the characteristics for the gradient of each layer, especially its strength. 
\looseness-1

SVD is a factorization of a real or complex matrix that generalizes the eigendecomposition of a square normal matrix to any \(m \times n\) matrix via an extension of the polar decomposition. Specifically, e.g., for our gradient matrix \(G_{X, i} \in \mathbb{R}^{m \times n}, X \in \{Q, K, V, O\}\), it can be decomposed as:

\vspace{-2mm}
\begin{equation}
G_{X, i} = U \Sigma V^T
\end{equation}
\vspace{-4mm}

where \( U \in \mathbb{R}^{m \times m} \) is an orthogonal matrix containing the left singular vectors;
\( \Sigma \in \mathbb{R}^{m \times n} \) is a diagonal matrix with singular values \( \sigma_1, \sigma_2, \ldots, \sigma_{\min\{m, n\}} \);
\( V \in \mathbb{R}^{n \times n} \) is an orthogonal matrix containing the right singular vectors. 
For simplicity, the subscript for these intermediate matrices is omitted.


\noindent
\textbf{Nuclear Norm}:
The nuclear norm of \(G\) is defined as the $\ell_1$ norm of the singular values, which reflects the sparsity of the spectrum and serves as a convex surrogate of the matrix rank. Hence, it not only quantifies the gradient magnitude but also the concentration of the spectrum on its top few singular values, which is vital to understand the gradient patterns in each layer, i.e.,
\looseness-1

\vspace{-2mm}
\begin{equation}
s_{X, i} = \| G_{X, i} \|_* = \sum_{j=1}^{\min\{m, n\}} |\sigma_j|
\end{equation}
\vspace{-4mm}

\subsection{Metrics of Gradient Difference}

In our experimental analysis, the nuclear norm $s_{X, i}$ of each layer will not be investigated individually, but to investigate the overall dynamic characteristics across every layer of the LLM. For simplicity, we notate the nuclear norm value, $s_{X, i}$, of a specific metric across all the layers as a curve notated as $s_X$. To analyze these gradient results, the visualization of the layer-wise curves is one of the most important tools to get a qualitative understanding of how the gradients change across the layers. However, quantitative analysis is still required for a better understanding of gradient representations. 

\noindent
\textbf{Gradient-Norm Difference between Layers. }
Considering that both the fluctuation and scale are important for a gradient curve, we utilize the Mean Absolute Difference (MAD) to represent a gradient curve. Specifically, the MAD of the curve $s_X$ is notated as MAD$_{s_X}$, which is calculated as:
\looseness-1

\vspace{-3mm}
\begin{equation}
\text{MAD}_{s_X} = \frac{1}{N-1} \sum_{i=1}^{N-1} \left| s_{X,i+1} - s_{X,i} \right|
\end{equation}
\vspace{-2mm}

where $N$ is the total number of layers of the target LLM. 
MAD is a measure of the average magnitude of change between consecutive points. Unlike standard deviation, MAD focuses on the direct differences between successive values without squaring or taking into account the direction (positive or negative change). It is useful for quantifying the overall variability of the data, especially when detecting fluctuations or local changes, which are more important than global trends. 

\noindent
\textbf{Gradient-Norm Difference. }
In addition to the value presentation for each individual curve, the pair-wise comparison between two curves, across different layers, is also important for our analysis. For this purpose, we use the layer-wise Relative Difference, RD, as our metric. At each layer, the RD between 2 values $s_{X, i}^{(1)}$ and $s_{X, i}^{(2)}$:

\begin{equation}
\text{RD}_{X,i} = \frac{s_{X, i}^{(2)} - s_{X, i}^{(1)}}{s_{X, i}^{(1)}}
\end{equation}
\vspace{-2mm}

where $s_{X, i}^{(1)}$ is utilized as the reference value. For this metric, the projection layer $X$ and the layer index $i$ should be kept the same. 
\looseness-1

\section{Experimental Setup}

\subsection{Models}

We investigate the gradients for 10 models including 5 base pre-trained models, Qwen2-1.5B \cite{yang2024qwen2technicalreport}, gemma-2-2b \cite{gemmateam2024gemma2improvingopen}, Llama-3.1-8B \cite{dubey2024llama3herdmodels}, gemma-2-9b \cite{gemmateam2024gemma2improvingopen}, Llama-2-7b-hf \cite{touvron2023llama2openfoundation} and their instruction-tuned version, Qwen2-1.5B-Instruct, gemma-2-2b-it, Llama-3.1-8B-Instruct, gemma-2-9b-it, Llama-2-7b-chat-hf. The main illustrations in our paper will be based on the results of Qwen2-1.5B models, results for other models will be in the appendix. 

\subsection{Datasets}

The datasets we use include three categories: \textbf{Math}, \textbf{Commonsense Reasoning}, and \textbf{Wiki Knowledge}: 
For the category of \textbf{Math}, AQuA \cite{ling-etal-2017-program}, GSM8K \cite{cobbe2021trainingverifierssolvemath}, and MATH \cite{hendrycks2021measuring} are utilized. 
For the category of \textbf{Commonsense Reasoning}, four datasets are utilized, including StrategyQA \cite{geva2021did}, ECQA \cite{aggarwal-etal-2021-explanations}, CREAK \cite{onoe2021creakdatasetcommonsensereasoning}, and Sensemaking, obtained from the FLAN collection \cite{longpre2023flancollectiondesigningdata}. 
For the category of \textbf{Wiki Knowledge}, we construct our own dataset where the wiki passages are grouped into two groups: popular and unpopular, based on their page views using the Pageviews Analysis tool\footnote{\url{https://pageviews.wmcloud.org}}. Detailed descriptions of these datasets can be found in Appendix \ref{sec:data_description}, and corresponding examples are provided in Appendix \ref{sec:data_examples}.
\looseness-1

The Math and Commonsense Reasoning datasets are utilized to explore the gradients when LLMs are fine-tuned to learn the reasoning process (slow or fast thinking), and the Wiki Knowledge datasets are utilized to explore the gradients when LLMs are fine-tuned to learn pure knowledge. Due to the slow process of calculating gradients, we randomly sample $500$ data instances for each task for our extensive experiments. 
The scope of our experiments is shown in Table \ref{tbl:scope}. 

\begin{figure*}[!ht]
\centering
\includegraphics[width=0.85\linewidth]{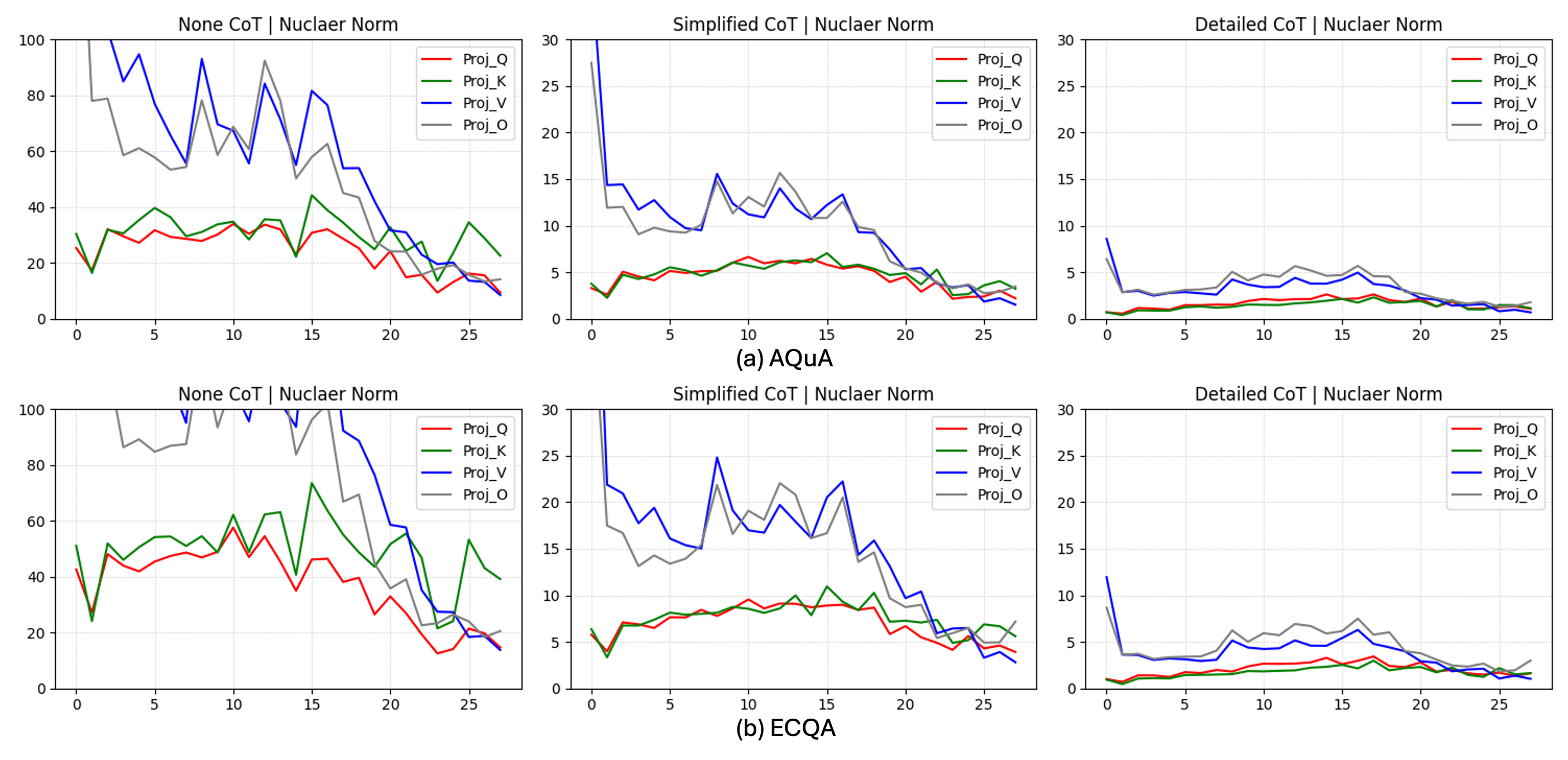}
\caption{The nuclear norm of gradients across different layers (x-axis) when trained with fast to slow reasoning paths (left to right columns), on (a) AQuA and (b) ECQA datasets. \textbf{When detailed CoT is utilized for training, the gradient norm tends to be similar across layers} (on both math and commonsense reasoning tasks). Note the y-axis scale for \textit{None CoT} is larger, and the scale for \textit{Simplified CoT} and \textit{Detailed CoT} are the same. \looseness-1}
\label{fig:fig_cot_cot}
\vspace{-2mm}
\end{figure*}
\begin{table}[t]
\centering
\small
\scalebox{0.75}{
\begin{tabular}{l|l|l|cccc}
\toprule
\multirow{2}{*}{\textbf{Dataset}} & \multirow{2}{*}{\textbf{Curve}} & \multirow{2}{*}{\textbf{Cot}} &  \multicolumn{4}{c}{\textbf{Mean Absolute Difference (MAD)}} \\
 & &  & Early & Middle & Last & All\\
\midrule
\multirow{12}{*}{AQuA} & \multirow{3}{*}{$s_Q$} 
  & None & 5.76 & 4.13 & 3.49 & 4.42 \\
& & Simplified & 0.89 & 0.52 & 0.77 & 0.69 \\
& & Detailed & \textbf{0.23} & \textbf{0.28} & \textbf{0.29} & \textbf{0.28} \\
\cmidrule{2-7}
& \multirow{3}{*}{$s_K$} 
  & None & 7.20 & 6.29 & 8.40 & 7.06 \\
& & Simplified & 1.01 & 0.56 & 1.11 & 0.81 \\
& & Detailed & \textbf{0.22} & \textbf{0.21} & \textbf{0.42} & \textbf{0.27} \\
\cmidrule{2-7}
& \multirow{3}{*}{$s_V$} 
  & None & 37.29 & 16.12 & 3.94 & 17.32 \\
& & Simplified & 5.08 & 2.14 & 0.86 & 2.36 \\
& & Detailed & \textbf{1.15} & \textbf{0.62} & \textbf{0.33} & \textbf{0.64} \\
\cmidrule{2-7}
& \multirow{3}{*}{$s_O$} 
  & None & 23.79 & 14.35 & 3.04 & 12.91 \\
& & Simplified & 3.31 & 2.18 & 0.63 & 1.97 \\
& & Detailed & \textbf{0.82} & \textbf{0.75} & \textbf{0.29} & \textbf{0.64} \\
\midrule
\multirow{12}{*}{ECQA} & \multirow{3}{*}{$s_Q$} 
  & None & 8.00 & 7.01 & 5.01 & 6.53 \\
& & Simplified & 1.11 & 0.70 & 0.86 & 0.85 \\
& & Detailed & \textbf{0.30} & \textbf{0.37} & \textbf{0.26} & \textbf{0.35} \\
\cmidrule{2-7}
& \multirow{3}{*}{$s_K$} 
  & None & 11.51 & 11.07 & 13.32 & 11.11 \\
& & Simplified & 1.34 & 1.24 & 1.01 & 1.13 \\
& & Detailed & \textbf{0.26} & \textbf{0.29} & \textbf{0.54} & \textbf{0.34} \\
\cmidrule{2-7}
& \multirow{3}{*}{$s_V$} 
  & None & 59.33 & 24.83 & 7.46 & 27.40 \\
& & Simplified & 8.53 & 3.55 & 1.66 & 4.01 \\
& & Detailed & \textbf{1.56} & \textbf{0.74} & \textbf{0.48} & \textbf{0.82} \\
\cmidrule{2-7}
& \multirow{3}{*}{$s_O$} 
  & None & 39.20 & 19.50 & 5.12 & 19.38 \\
& & Simplified & 5.56 & 3.33 & 1.41 & 3.22 \\
& & Detailed & \textbf{1.00} & \textbf{0.97} & \textbf{0.52} & \textbf{0.85} \\
\bottomrule
\end{tabular}
}
\caption{The mean absolute differences (MAD) of gradient's nuclear norm for $K, Q, V, O$ projection layers. Early, Middle, Last, and All represent the MAD scores calculated across the early, middle, last, and all the layers across the LLM. \textbf{A consistent decrease is observed in all layers when LLMs are trained to produce more detailed reasoning paths (slow thinking).} 
}
\label{tbl:table_cot_cot}
\vspace{-2mm}
\end{table}



\section{Empirical Analysis}

\begin{figure}[t]
\centering
\includegraphics[width=1\linewidth]{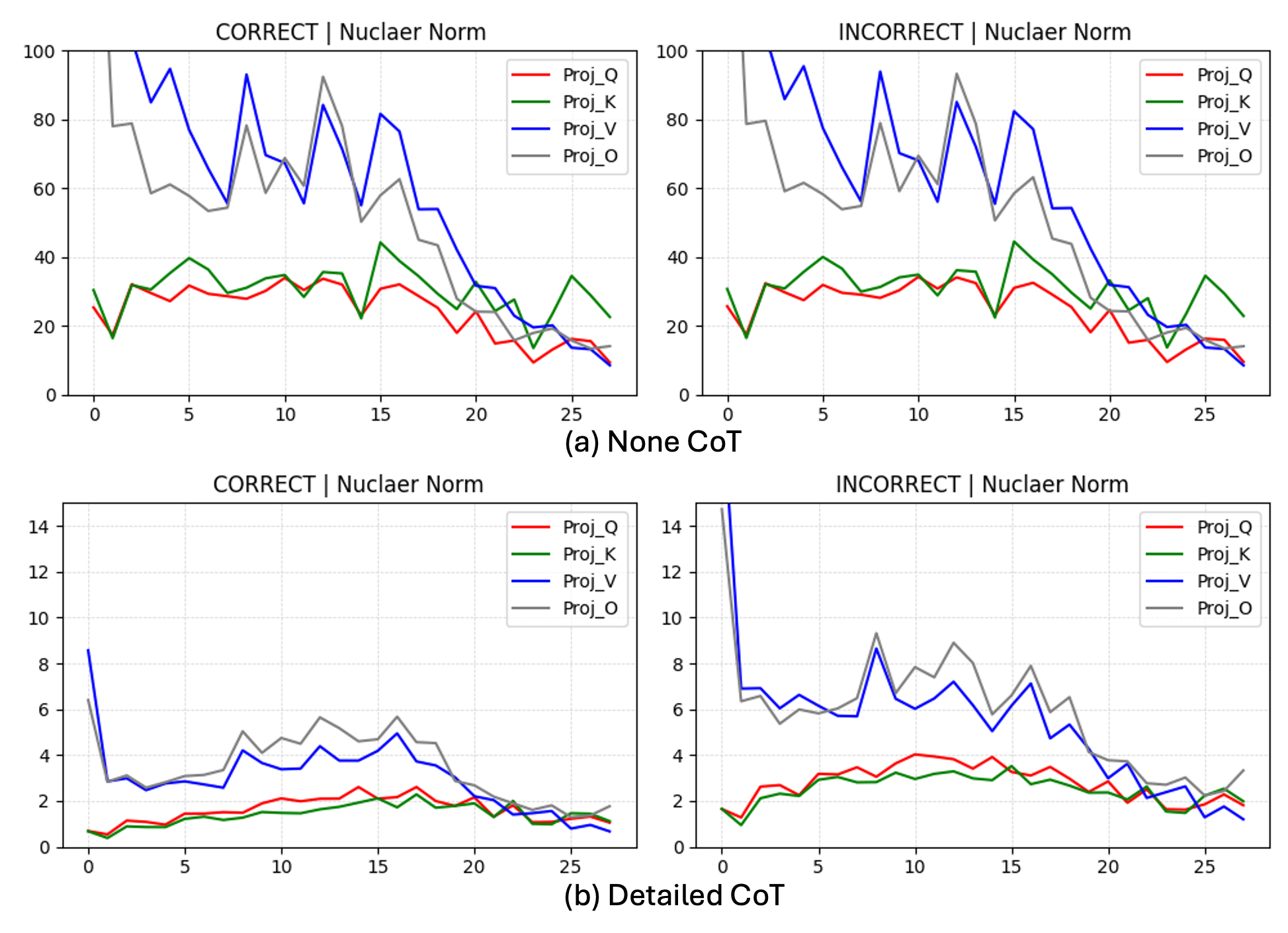}
\caption{The nuclear norm of gradients across different layers (x-axis) when trained with Correct vs. Irrelevant responses (a) without CoT (fast thinking); (b) with detailed CoT (slow thinking), on the AQuA dataset. \textbf{Gradient norm can help identify correct responses when provided Detailed CoT. But this does not extend to gradients without CoT.} 
}
\label{fig:fig_cot_correct}
\vspace{-2mm}
\end{figure}

\subsection{Math and Commonsense Reasoning Tasks}

This section focuses on tasks related to CoT reasoning, which includes the datasets within the Math and Commonsense Reasoning task type. 

\subsubsection{Slow vs. Fast Thinking}

We first investigate the gradient behaviors when LLMs learn the responses with the reasoning process. i.e., the CoT paths. For samples in the MATH dataset, the original responses already contain the necessary steps to solve the question, which we notate as \textit{Simplified CoT} setting. For the remaining datasets in these two task types, both the pure answer (resulting digits or options) and short CoT paths are provided, which we notate as \textit{None CoT} and \textit{Simplified CoT} settings. These configurations can help us to understand the gradients when LLMs learn responses with or without CoT, probably revealing the advantages of CoT training. 
Moreover, the provided CoT paths are all too simplified, 
thus we further prompt GPT4o to generate a more detailed version of reasoning paths to compare the effect of different CoT paths, i.e., slow or fast thinking, which is notated as \textit{Detailed CoT}. 
\looseness-1

The detailed statistical values for the gradient curves on AQuA and ECQA with different CoT settings are provided in Table \ref{tbl:table_cot_cot}, and the visualization of the gradient curves on AQuA and ECQA is shown in Figure \ref{fig:fig_cot_cot}, all the results in this table are based on the Qwen2-1.5B model. 
When no CoT reasoning paths are provided for LLMs to learn, fast thinking, the mean absolute differences (MADs) are the largest on all the curves of different projection layers, representing a severe fluctuation of the gradient scales across all the layers of the LLM, which might cause instability for training \cite{glorot2010understanding}. 
However, when CoT paths are given, the MADs drop accordingly, especially when the detailed CoT paths are given, slow thinking, as visualized in Figure \ref{fig:fig_cot_cot}. 
The large scale indicates that the response distributions that LLMs are going to learn have large discrepancies with what it has learned from the pretraining phase, which might harm the performances of the original pre-trained models \cite{ghosh2024closerlooklimitationsinstruction, biderman2024lora}. Our findings are aligned with the current success of utilizing more detailed CoT reasoning paths or responses for training \cite{mitra2023orca, Li2023ReflectionTuningDR} and provide another perspective on understanding the effectiveness of slow thinking. 
\looseness-1

\begin{table}[t]
\centering
\small
\scalebox{0.75}{
\begin{tabular}{l|l|l|c|ccccc}
\toprule
\multirow{2}{*}{\textbf{Dataset}} & \multirow{2}{*}{\textbf{CoT}} & \multirow{2}{*}{\textbf{Curve}} & \textbf{RD} & \multicolumn{5}{c}{\textbf{Top 5 Different Layer Idx}} \\
 & &  & Average &  1 &  2 &  3 & 4 & 5\\
\midrule
\multirow{4}{*}{AQuA} 
& \multirow{4}{*}{Detailed}  
  & $s_Q$ & 0.81 & 3 & 0 & 1 & 4 & 2 \\
& & $s_K$ & 0.90 & 3 & 4 & 1 & 0 & 7 \\
& & $s_V$ & 0.81 & 3 & 1 & 4 & 2 & 0 \\
& & $s_O$ & 0.72 & 0 & 1 & 4 & 2 & 3 \\
\midrule
\multirow{4}{*}{ECQA} 
& \multirow{4}{*}{Detailed}  
  & $s_Q$ & 0.46 & 0 & 3 & 2 & 1 & 4 \\
& & $s_K$ & 0.50 & 0 & 3 & 1 & 2 & 4 \\
& & $s_V$ & 0.47 & 3 & 1 & 2 & 4 & 0 \\
& & $s_O$ & 0.41 & 0 & 1 & 2 & 3 & 4 \\

\bottomrule
\end{tabular}
}
\caption{The average relative difference (RD) and the indexes of the top-5 layers that have the greatest gap between the curves of learning the correct and irrelevant responses. \textbf{It shows that the earlier layers change more sensitively to irrelevant responses.} 
}
\label{tbl:table_cot_correct}
\vspace{-2mm}
\end{table}

\subsubsection{Effect of Response Correctness}

\begin{figure*}[!ht]
\centering
\includegraphics[width=0.85\linewidth]{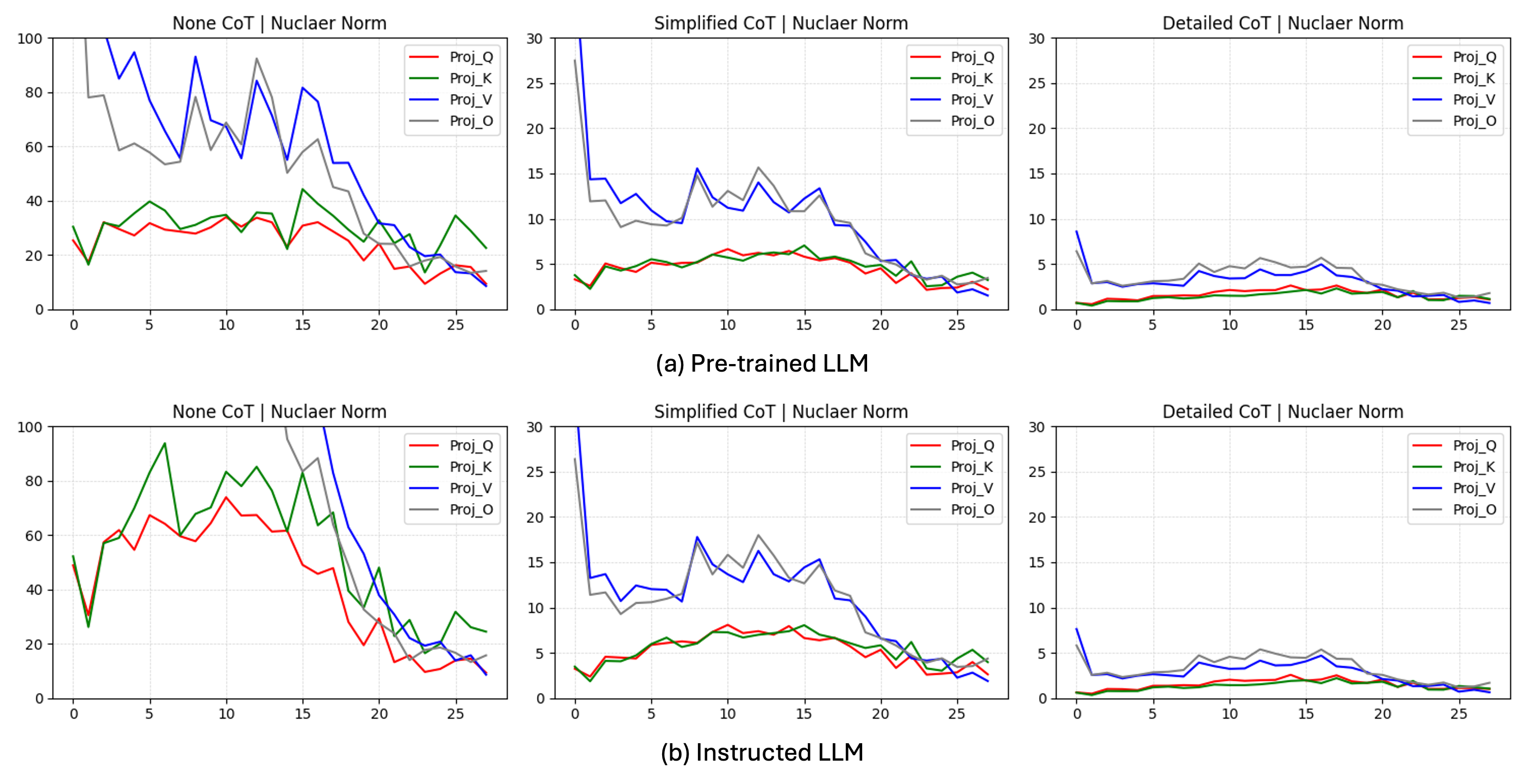}
\caption{The nuclear norm of gradients across different layers (x-axis) on \textbf{(a) pre-trained base LLM vs. (b) instruction-finetuned LLM}. On both models, training using detailed CoT (slow thinking) reduces the gradient norm and difference across layers. However, \textbf{the two models' gradient patterns differ when training with fast thinking} (Simplified/None CoT). 
The y-axis scale of \textit{None CoT} is greater than that of \textit{Simplified CoT} and \textit{Detailed CoT}. \looseness-1}
\label{fig:fig_cot_instruct}
\vspace{-4mm}
\end{figure*}
\begin{figure*}[t]
\centering
\includegraphics[width=1\linewidth]{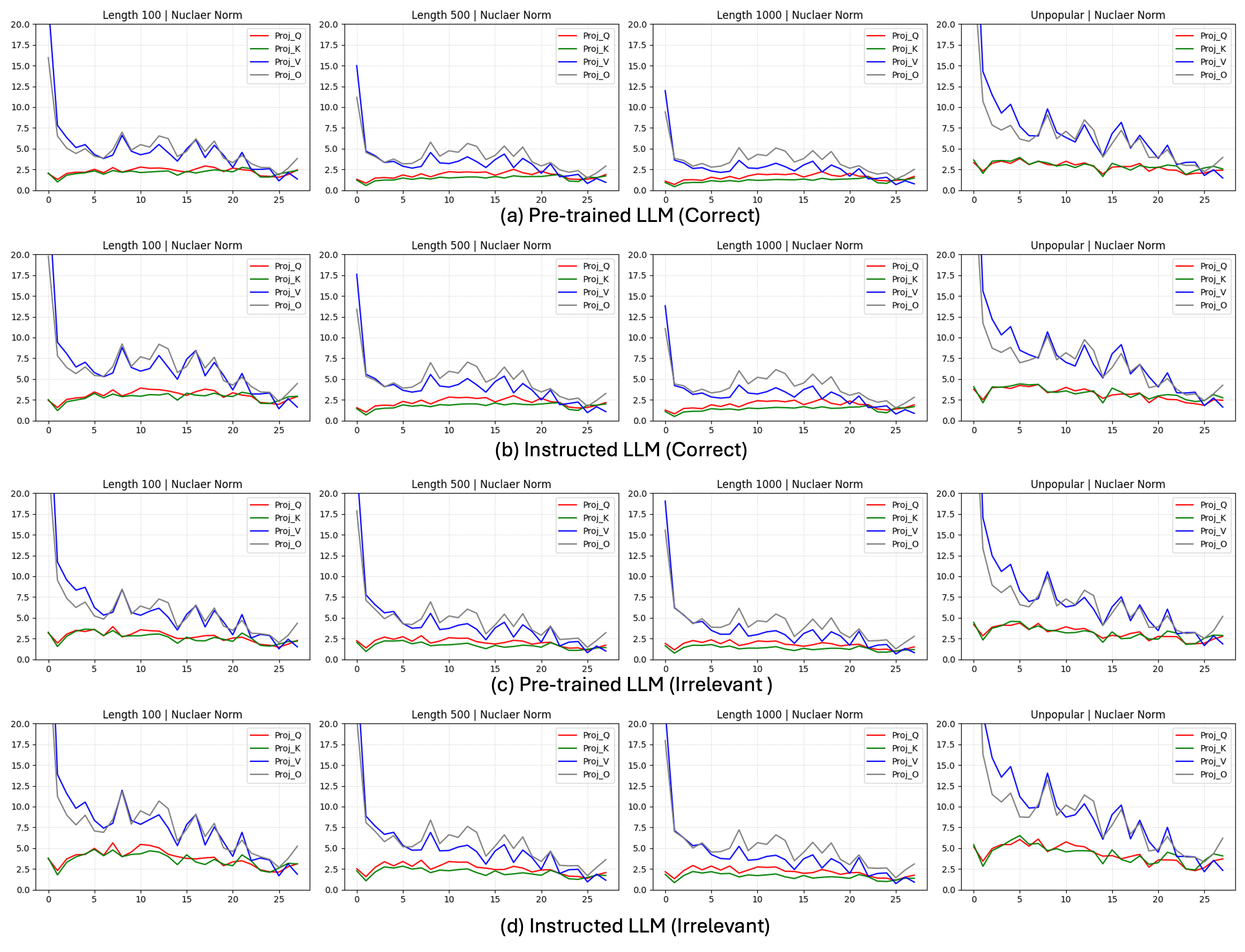}
\caption{The nuclear norm of gradients across different layers (x-axis) when trained with responses of different lengths (left 3 columns) and unpopular knowledge (rightmost column) on the Wiki knowledge learning (\textbf{knowledge-intensive}) task. 
Comparing ``Short vs. Long'', ``Popular vs. Unpopular'', and ``Correct vs. Irrelevant'' on the two types of models indicates: \textbf{(1) longer response$\neq$slower thinking}. Unlike Figure~\ref{fig:fig_cot_cot}, solely increasing the response length does not affect the gradient patterns; \textbf{(2) unpopular knowledge triggers larger gradients}; (3) Unlike Figure~\ref{fig:fig_cot_correct}, \textbf{gradient norm cannot help judge the response's correctness on knowledge-intensive tasks.}}
\label{fig:figure_wiki_all}
\vspace{-2mm}
\end{figure*}

In this section, we investigate the gradient behaviors when LLMs are learning the correct or irrelevant responses with different reasoning paths. Similarly, for datasets where pure ground truths are given, we investigate the gradient behaviors on three settings: \textit{None CoT}, \textit{Simplified Cot}, and \textit{Detailed CoT}, otherwise, only the last two settings can be investigated. In the \textit{None CoT} setting, we directly shuffle the answers across the dataset and make sure every question has the incorrect answer; In the \textit{Simplified Cot} and \textit{Detailed CoT} settings, we split every CoT path into individual sentences, then shuffle the sentences across the dataset. Under these circumstances, each sentence in the response is still complete, while the relation across sentences will be logically wrong, simulating the irrelevant CoT reasoning paths. 
\looseness-1

In these experiments, we try to investigate if the LLMs are able to identify the irrelevant responses during training with slow or fast thinking, reflected by gradient behaviors. 
The visualizations of LLMs learning on correct and irrelevant responses are presented in Figure \ref{fig:fig_cot_correct}, which contains 2 settings including (a) \textit{None CoT} and (b) \textit{Detailed CoT}. 
It is widely accepted that LLMs have learned all the knowledge in the pretraining phase \cite{zhou2023lima}, so when LLMs are forced to learn responses that conflict with their internal knowledge, more efforts might be needed (larger gradient) for this false alignment.
However, from the visualizations, it can be observed that when no CoT paths are given, the gradient behaviors between learning the correct and nonsense responses are almost identical, and their relative difference values on all the projection layers are less than $0.01$. 
Thus, this phenomenon indicates that LLMs can not build the necessary mapping relations from the question to the answer without explicit reasoning paths being given. 
On the contrary, when the detailed CoT reasoning paths are provided in the responses, the gradient behaviors will be different, mainly reflected by the larger scale of the gradient. This phenomenon indicates that LLMs can, to some extent, identify that the responses to be learned have potential conflicts with their internal knowledge, thus requiring more efforts (reflected by larger gradients) to adapt to the new nonsense responses. 
\looseness-1

We further investigate if there are specific layers that are directly related to LLMs' capability to perceive irrelevant knowledge, reflected by the larger differences between gradients of correct and irrelevant responses.  
As shown in Table \ref{tbl:table_cot_correct}, the relative difference values of the nuclear norm curves, and the indexes of the top 5 layers that have the greatest gap are presented. The results on the nuclear norm curves show that the earlier layers are more sensitive to nonsense responses, which might indicate the potential effects of earlier layers. 

\subsubsection{Effect of Initial Models}

In this section, we compare the gradient behaviors between pre-trained base LLMs and aligned instructed LLMs. For each instructed LLM, we utilize the conversation templates officially provided, avoiding the potential misalignment. In these experiments, we observe that the instructed LLMs do not have much better performance in identifying the irrelevant responses, evidenced by the minor relative differences between gradient curves obtained from base LLMs and aligned LLMs. 

However, as shown in Figure \ref{fig:fig_cot_instruct}, although the tendencies are consistent on both types of LLMs that detailed CoT reasoning paths make the scale and fluctuation of gradient smaller and smoother, the gradients on simplified CoT show a large discrepancy between the two types of LLMs. This discrepancy means that the instructed LLMs need more effort than the pre-trained LLMs to learn the simplified CoT paths. The phenomenon shows that (1) the distribution of the simplified CoT responses might have non-negligible discrepancies with the instruction datasets used for training this LLM; (2) the behaviors on gradient curves might be used as a measurement of how a specific data sample aligned with the internal knowledge of LLMs, which might be useful for the continued training settings. 

\subsection{Knowledge Learning Tasks}

\subsubsection{Response Length \& Popularity}

In this section, we investigate the effect on response length and popularity for the knowledge-intensive task. As shown in Figure \ref{fig:figure_wiki_all}, the left 3 columns represent the scenarios when LLMs are learning popular knowledge with different response lengths, and the right-most column represents the scenario when LLMs are learning unpopular knowledge. By comparing the left 3 columns of the figure, it is observed that for the knowledge-intensive task, the lengths of responses do not affect the gradient scales and fluctuations. This phenomenon is largely different from the findings observed in the reasoning tasks, where detailed CoTs can largely reduce the gradient scale and fluctuation. This comparison can further verify that the effects of the detailed CoT (slow thinking) in the responses are not caused by the increase in token length but by the detailed reasoning process. 
\looseness-1

On the contrary, as shown in the right-most figures, when LLMs are learning unpopular knowledge, the scales and fluctuation increase dramatically, indicating that LLMs need more effort to learn this unpopular knowledge. This phenomenon is reasonable as popular knowledge occurs frequently from diverse corpus sources, representing an augmentation of this knowledge.

\subsubsection{Response Correctness \& Initial Models}

Then we compare the curve differences when LLMs are learning the correct or nonsense responses as shown in Figure \ref{fig:figure_wiki_all}. The almost identical curves between (a) and (c) show that the pre-trained LLM is not able to identify the nonsense knowledge that it is learning, and the curves between (b) and (d) show that the instructed LLM also lacks this capability. This phenomenon is also different from the findings observed in the reasoning tasks, where LLMs are able to identify the nonsense reasoning paths reflected by the increase of gradient scales and fluctuation. 

As for the effect of the instructed LLMs, comparing between (a) and (b), and (c) and (d), a consistent increase in the gradient scales and fluctuation is observed, especially on the unpopular knowledge. This phenomenon indicates that it is harder for instructed LLMs to learn knowledge that is new or not well-learned during the pre-training phase.





\section{Further Discussion}

\subsection{Why is Slow Thinking more Stable?}

Why ``slow'' thinking (utilizing CoT) works much more effectively than ``fast'' thinking (no CoT) is still an open question. However, 
\citet{prystawski2023why} proposes that ``chain-of-thought reasoning becomes useful exactly when the training data is structured locally, in the sense that observations tend to occur in overlapping neighborhoods of concepts'', that is to say, detailed CoT paths can make the reasoning gaps between sentences in response much smaller, as the concept between steps of reasoning paths are more likely to co-occur frequently. Under these circumstances, the tokens can be inferred with a larger probability as their prior context has provided the necessary information to better align them with their corresponding pretraining data. Hence, during the finetuning phase, the losses of tokens become much smaller, and the alignment between the response and the pre-trained data further leads to a stable gradient. 
This can partially explain why irrelevant reasoning paths cause the larger and unstable gradient: they hinder the alignment between response and pretraining data, and their local context can no longer provide necessary information for the generation of correct responses. 
\looseness-1

\subsection{Behaviors of Different LLMs}

The generalization of our findings is promised after thoroughly examining the results across different LLMs. However, we further reveal some subtle differences in the gradient patterns across different models, e.g., there are typically two peaks in the curves of the gemma2 models; there is one peak in the curves of Qwen2 models; there is no obvious peak for Llama3 models. We hypothesize that these diverse behaviors are directly related to the specific characteristics of different models, potentially caused by the model structure, training data, and initialization settings, and are worth further exploration. 
\looseness-1

\subsection{Potential Applications}

Here are some possible applications regarding our findings:
(1) The gradient behaviors can be potentially used to evaluate the quality of the data and the compatibility between specific datasets and LLMs to be fine-tuned. 
(2) The gradient behaviors can be potentially used to identify the most critical layers for specific LLMs, which might motivate more efficient training strategies. 
(3) The differences in gradient behaviors between different LLMs can provide insights into the specific characteristics of LLMs, leading to a better understanding of these black-box models.

\section{Conclusion}

Our study reveals the significant differences in gradient behaviors between fast and slow thinking training methodologies in LLMs, offering insights into how training dynamics can influence these models. Specifically, we observe that slow thinking leads to stable gradient norms of different layers, while fast thinking results in larger gradients and fluctuation across layers. 
Moreover, the gradient of slow thinking helps distinguish correct responses from irrelevant responses, while without CoT, the gradient patterns of the two types of responses are similar. 
The above observations on reasoning tasks cannot be extended to knowledge-learning tasks, where simply increasing response length does not show similar gradient patterns as slow thinking. 

\section*{Limitations}

Due to the page limits, only a small proportion of results can be presented, which might weaken the findings of the paper. However, we try to include as many results as possible in the appendix through visualization and statistical results, hoping to provide further insights for the community. 
Moreover, the analysis in this paper focuses mainly on the strength of the layer-wise gradients, maybe more metrics can be used. 

\bibliography{custom}

\clearpage
\appendix


\startcontents[appendix]
\printcontents[appendix]{ }{0}{\section*{Table of Contents}}

\clearpage
\section{Dataset Description}
\label{sec:data_description}

\begin{table*}[h!]
\centering
\scalebox{0.7}{
\begin{tabular}{l|l|ccc|ccc}
\toprule
\multirow{2}{*}{\textbf{Task Type}} & \multirow{2}{*}{\textbf{Dataset}} & \multicolumn{3}{c|}{\textbf{Correct Response}} & \multicolumn{3}{c}{\textbf{Irrelevant Response}} \\ 
\cmidrule(lr){3-5}\cmidrule(lr){6-8}
 &  & \textbf{None CoT} & \textbf{Simplified CoT} & \textbf{Detailed CoT} & \textbf{None CoT} & \textbf{Simplified CoT} & \textbf{Detailed CoT} \\
\midrule
\multirow{5}{*}{Math}  
& AQuA          & 3  & 60  & 547  & 3  & 59  & 530 \\
& GSM8K         & 3  & 101 & 427  & 3  & 101 & 418 \\
& MATH-Algebra  &       & 225 & 698  &       & 223 & 691 \\
& MATH-Counting &       & 244 & 749  &       & 242 & 737 \\
& MATH-Geometry &       & 444 & 769  &       & 442 & 728 \\
\midrule
\multirow{4}{*}{Commonsense}  
& StrategyQA    & 1  & 53  & 663  & 1  & 53  & 645 \\
& ECQA          & 2  & 34  & 454  & 2  & 33  & 436 \\
& CREAK         & 1  & 24  & 420  & 1  & 23  & 401 \\
& Sensemaking   & 3  & 22  & 450  & 3  & 21  & 429 \\
\midrule
\multirow{4}{*}{Wiki Knowledge} 
& Popular (Length 100)  & 119 &        &         & 118 &        &        \\
& Popular (Length 500)  & 499 &        &         & 496 &        &        \\
& Popular (Length 1000) & 976 &        &         & 970 &        &        \\
& Unpopular             & 98  &        &         & 94  &        &        \\

\bottomrule
\end{tabular}
}
\caption{The average response lengths (token counts) of datasets with different configurations within the scope of our paper. }
\label{tbl:response_length}
\end{table*}

\subsection{Math Reasoning}

For the category of math, AQuA \cite{ling-etal-2017-program}, GSM8K \cite{cobbe2021trainingverifierssolvemath}, and MATH \cite{hendrycks2021measuring} are utilzied. The original ground truth format for AQuA is options from \textit{A} to \textit{E}, and for GSM8K is the \textit{resulting digits}, and additional CoT reasoning paths are provided as well. During our experiments, both the process of learning the original ground truth (fast thinking) and learning the CoT plus ground truth are investigated. The ground truth for MATH is the complete solution for the question, which can be regarded as answers with CoT. We select the question types Algebra, Counting, and Geometry in MATH for our experiments. 
Moreover, to further explore the effects of more detailed reasoning paths (slow thinking), GPT4o is utilized to obtain a detailed version of CoT paths. 

\subsection{Commonsense Reasoning}

For the category of commonsense reasoning, four datasets are utilized including StrategyQA \cite{geva2021did}, ECQA \cite{aggarwal-etal-2021-explanations}, CREAK \cite{onoe2021creakdatasetcommonsensereasoning}, and Sensemaking, obtained from the FLAN collection \cite{longpre2023flancollectiondesigningdata}. 
The original ground truth format for StrategyQA is options of \textit{yes} and \textit{no}, for ECQA is \textit{word options}, for CREAK is options of \textit{yes} and \textit{no}, and for Sensemaking is options of \textit{Sentence A} or \textit{Sentence B}. 
For all these datasets, the corresponding human-annotated CoT reasoning paths are provided, and both the process of learning the original ground truth and learning the CoT plus ground truth are investigated in our experiments. 
Similarly, further GPT4o-generated CoTs are also investigated. 

\subsection{Wiki Knowledge Learning}

This group of tasks represents LLMs learning on pure knowledge-intensive responses without any reasoning process. For Wikipedia knowledge, we categorize Wiki articles into two groups: popular and unpopular based on their total page views in 2021 using the Pageviews Analysis tool\footnote{\url{https://pageviews.wmcloud.org}} provided by Wikimedia's RESTBase API. After selection, we extract the first $100$, $500$, and $1000$ tokens (respecting paragraph boundaries) from each article for the controlled experiments on the effects of response lengths. 
For the unpopular wiki pages, we use the least viewed articles in 2021\footnote{\url{https://en.wikipedia.org/wiki/User:Colin_M/Least_viewed_articles_in_2021}}. As the length of unpopular wiki pages is generally short, we take the full articles; if the length exceeds 1024 tokens, we truncate them with respect to sentence boundaries. 
\looseness-1

We assume all these articles (both popular and unpopular) have been learned by the modern LLMs since we are using the popularity of 2021. Under this assumption: (1) The popular and unpopular ones still exhibit different gradient patterns under this assumption, where the unpopular knowledge triggers larger gradients; (2) The Gradient norm cannot help identify irrelevant responses even if the correct ones have been learned (because the responses are loosely related in knowledge-intensive tasks compared to reasoning chains). 

The average response lengths (token counts) of datasets with different configurations within the scope of our paper are presented in Table \ref{tbl:response_length}. 


\clearpage
\section{Data Examples}
\label{sec:data_examples}

In this section we provide examples for each of our datasets: AQuA (Figure \ref{fig: example_aqua}), CREAK (Figure \ref{fig: example_creak}), ECQA (Figure \ref{fig: example_ecqa}), GSM8K (Figure \ref{fig: example_gsm8k}), Sensemaking (Figure \ref{fig: example_sensemaking}), StrategyQA (Figure \ref{fig: example_strategyqa}), MATH-Algebra (Figure \ref{fig: example_algebra}), MATH-Counting (Figure \ref{fig: example_counting}), MATH-Geometry (Figure \ref{fig: example_geometry}), Wiki Popular (Figure \ref{fig: example_wiki_pop}), and Wiki Unpopular (Figure \ref{fig: example_wiki_unpop}).

\begin{figure*}[!h]

  \rule{\textwidth}{1.5pt} 
  \textbf{Question:} A 600 meter long train crosses a signal post in 40 seconds. How long will it take to cross a 3 kilometer long bridge, at the same speed? Options: (A) 4 min, (B) 2 min, (C) 8 min, (D) 9 min, (E) 5 min 
  
  \rule{\textwidth}{1.5pt}
  \textcolor{teal}{[None CoT]}
  
  (A) \\
    \rule{\textwidth}{0.8pt}
    \textcolor{teal}{[Simplified CoT]}

    S = 600/40 = 15 mps\\
    S = 3600/15 = 240 sec = 4 min\\
    The final answer is (A) \\
  \rule{\textwidth}{0.8pt}

    \textcolor{teal}{[Detailed CoT]} \\
    To find the time it will take for a 600-meter-long train to cross a 3-kilometer-long bridge, we need to follow these steps:

Step 1: Determine the speed of the train  - The train crosses a signal post (which can be treated as a point) in 40 seconds.
- This means the train travels its own length (600 meters) in 40 seconds.

Calculate the speed of the train:
\[ \text{Speed} = \frac{\text{Distance}}{\text{Time}} \]
\[ \text{Speed} = \frac{600 \text{ meters}}{40 \text{ seconds}} \]
\[ \text{Speed} = 15 \text{ meters/second} \]

Step 2: Determine the total distance to be covered when crossing the bridge
- The length of the bridge is 3 kilometers, which is 3000 meters.
- When crossing the bridge, the train needs to cover the length of the bridge plus its own length.
- Total distance to be covered = Length of the bridge + Length of the train
\[ \text{Total distance} = 3000 \text{ meters} + 600 \text{ meters} \]
\[ \text{Total distance} = 3600 \text{ meters} \]

Step 3: Calculate the time to cross the bridge
- Using the speed calculated earlier, we now find the time to cover 3600 meters.
\[ \text{Time} = \frac{\text{Distance}}{\text{Speed}} \]
\[ \text{Time} = \frac{3600 \text{ meters}}{15 \text{ meters/second}} \]
\[ \text{Time} = 240 \text{ seconds} \]

Step 4: Convert the time into minutes
- There are 60 seconds in a minute.
\[ \text{Time in minutes} = \frac{240 \text{ seconds}}{60 \text{ seconds/minute}} \]
\[ \text{Time in minutes} = 4 \text{ minutes} \]

Thus, it will take the train 4 minutes to cross the 3-kilometer-long bridge.

Conclusion:
The correct answer is (A) 4 minutes.

  \rule{\textwidth}{1.5pt} 
  \caption{An example from AQuA dataset.}
  \label{fig: example_aqua} 
\end{figure*}

\begin{figure*}[!h]

    \rule{\textwidth}{1.5pt} 
    \textbf{Question:} Which of the following is not a leap year ? Options: (A) 700; (B) 800; (C) 900; (D) 1000; (E) 1200
    
    \rule{\textwidth}{1.5pt}
    \textcolor{teal}{[None CoT]} \\
    (A)
    
        \rule{\textwidth}{0.8pt}
        \textcolor{teal}{[Simplified CoT]}
        
        The century divisible by 400 is
The final answer is (A)

        \rule{\textwidth}{0.8pt}
        \textcolor{teal}{[Detailed CoT]}

    To determine which of the given years is not a leap year, we need to understand the rules for leap years. Here’s a step-by-step breakdown:

Leap Year Rules:
1. A year is a leap year if it is divisible by 4.
2. However, if the year is also divisible by 100, it is not a leap year, unless:
3. The year is divisible by 400, in which case it is a leap year.

We will apply these rules to each of the given years.

Year 700
1. Check divisibility by 4: \( 700 \div 4 = 175 \) (no remainder, so divisible by 4).
2. Check divisibility by 100: \( 700 \div 100 = 7 \) (no remainder, so divisible by 100).
3. Check divisibility by 400: \( 700 \div 400 = 1.75 \) (remainder exists, so not divisible by 400).
   - Since 700 is divisible by 100 but not by 400, it is **not a leap year**.

Year 800
1. Check divisibility by 4: \( 800 \div 4 = 200 \) (no remainder, so divisible by 4).
2. Check divisibility by 100: \( 800 \div 100 = 8 \) (no remainder, so divisible by 100).
3. Check divisibility by 400: \( 800 \div 400 = 2 \) (no remainder, so divisible by 400).
   - Since 800 is divisible by 400, it is **a leap year**.

Year 900
1. Check divisibility by 4: \( 900 \div 4 = 225 \) (no remainder, so divisible by 4).
2. Check divisibility by 100: \( 900 \div 100 = 9 \) (no remainder, so divisible by 100).
3. Check divisibility by 400: \( 900 \div 400 = 2.25 \) (remainder exists, so not divisible by 400).
   - Since 900 is divisible by 100 but not by 400, it is **not a leap year**.

Year 1000
1. Check divisibility by 4: \( 1000 \div 4 = 250 \) (no remainder, so divisible by 4).
2. Check divisibility by 100: \( 1000 \div 100 = 10 \) (no remainder, so divisible by 100).
3. Check divisibility by 400: \( 1000 \div 400 = 2.5 \) (remainder exists, so not divisible by 400).
   - Since 1000 is divisible by 100 but not by 400, it is **not a leap year**.

Year 1200
1. Check divisibility by 4: \( 1200 \div 4 = 300 \) (no remainder, so divisible by 4).
2. Check divisibility by 100: \( 1200 \div 100 = 12 \) (no remainder, so divisible by 100).
3. Check divisibility by 400: \( 1200 \div 400 = 3 \) (no remainder, so divisible by 400).
   - Since 1200 is divisible by 400, it is **a leap year**.

Conclusion
From the analysis above, the years that are not leap years are:
- 700
- 900
- 1000

Among the options provided, the incorrect assumption was identifying which specific year is being asked about in question format that expects a singular correct answer.

However, strictly adhering to the problem formatting where typically one answer is sought:

- Since several options fit correctly and given the options stated as format; (A), (C), (D) are non-leap years.
Proceeding usually with first in order and options mentioned logically (A)700 provided correct inference.

Therefore:
**Option (A) 700 is not a leap year.**

    \rule{\textwidth}{1.5pt} 
    \caption{An example from CREAK dataset. }
    \label{fig: example_creak} 
    \end{figure*}

\begin{figure*}[!h]

    \rule{\textwidth}{1.5pt} 
    \textbf{Question:} The ability to remember things is a sign of what? Options: - neuronal network - memories - intelligence - recall - learning about
    
    \rule{\textwidth}{1.5pt}
    \textcolor{teal}{[None CoT]} 

    intelligence

        \rule{\textwidth}{0.8pt}
        \textcolor{teal}{[Simplified CoT]}
        
        Intelligence is an ability. The ability to remember things is a sign of intelligence.
The final answer is intelligence

        \rule{\textwidth}{0.8pt}
        \textcolor{teal}{[Detailed CoT]}

    To determine what the ability to remember things is most closely a sign of, it's helpful to analyze each option:

1. **Neuronal Network**:
   - The neuronal network refers to the structure and connections between neurons in the brain. While crucial for processing and storing information, the term itself does not directly signify the ability to remember. Rather, it describes the brain's architecture which facilitates functions including memory.

2. **Memories**:
   - Memories are the end result or the content of what is remembered. While they are directly linked to remembrance, the ability to have memories is an outcome of the memory process, not an indicator of the capacity to remember.

3. **Intelligence**:
   - Intelligence involves a broader range of cognitive abilities including problem-solving, reasoning, and understanding. Although memory is a component of intelligence, particularly in areas like working memory and long-term recall, intelligence encompasses far more than just the ability to remember.

4. **Recall**:
   - Recall is the process of retrieving information from memory. The ability to remember is a direct manifestation of recall, as it involves the capacity to retrieve stored information when needed.

5. **Learning About**:
   - Learning about refers to the process of acquiring new information or skills. While learning is related to memory, having the ability to remember indicates that the learning process has been successful, but it does not directly signify the capacity to remember.

Given this analysis, the most accurate answer is:

**Recall**.

The ability to remember things is a sign of recall because recall specifically refers to the cognitive process of retrieving information from memory.

    \rule{\textwidth}{1.5pt} 
    \caption{An example from ECQA dataset.}
    \label{fig: example_ecqa} 
    \end{figure*}

\begin{figure*}[!h]

    \rule{\textwidth}{1.5pt} 
    \textbf{Question:} Linda was going to pass out homemade cookies to each of her 24 classmates on the last day of school. She wanted to give each student 10 cookies and wanted to make chocolate chip cookies and oatmeal raisin cookies. Each cookie recipe made exactly 4 dozen cookies. She was able to make 2 batches of chocolate chip cookies before running out of chocolate chips. She made 1 batch of oatmeal raisin cookies. How many more batches of cookies does Linda need to bake?
    
    \rule{\textwidth}{1.5pt}
    \textcolor{teal}{[None CoT]} 

    2
    
        \rule{\textwidth}{0.8pt}
        \textcolor{teal}{[Simplified CoT]}
        
        She has 24 classmates and wants to give each 10 cookies, so she needs to bake 24 * 10 = 240 cookies. Each batch makes 4 dozen cookies so that means each batch has 4 * 12 = 48 cookies. She baked 2 batches of chocolate chip and one batch of oatmeal raisin so 2 + 1 = 3 batches. We know that each batch has 48 cookies so 3 * 48 = 144 cookies have been baked. She needs to have 240 cookies and has baked 144 so 240 - 144 = 96 cookies are needed. Again, each batch makes 48 cookies and we need 96 more cookies so 96 / 48 = 2 more batches are needed.
The final answer is 2

        \rule{\textwidth}{0.8pt}
        \textcolor{teal}{[Detailed CoT]}

    To determine how many more batches of cookies Linda needs to bake, let's proceed with the following steps:

1. **Calculate the total number of cookies Linda needs:**
   Linda wants to give each of her 24 classmates 10 cookies.
   \[
   \text{Total cookies needed} = 24 \text{ classmates} \times 10 \text{ cookies per classmate} = 240 \text{ cookies}
   \]

2. **Determine the number of cookies Linda has already made:**
   - Each batch of cookies makes exactly 4 dozen cookies.
   - 1 dozen = 12 cookies, so 4 dozen = \(4 \times 12 = 48\) cookies per batch.

   - Linda made 2 batches of chocolate chip cookies:
     \[
     \text{Cookies from chocolate chip cookies} = 2 \times 48 = 96 \text{ cookies}
     \]

   - Linda made 1 batch of oatmeal raisin cookies:
     \[
     \text{Cookies from oatmeal raisin cookies} = 1 \times 48 = 48 \text{ cookies}
     \]

   - Therefore, the total number of cookies she has made is:
     \[
     \text{Total cookies made} = 96 \text{ chocolate chip cookies} + 48 \text{ oatmeal raisin cookies} = 144 \text{ cookies}
     \]

3. **Determine the additional number of cookies needed:**
   - Total cookies needed are 240.
   - Cookies already made are 144.
   - Thus, the additional cookies needed are:
     \[
     \text{Additional cookies needed} = 240 - 144 = 96 \text{ cookies}
     \]

4. **Calculate the number of additional batches needed:**
   - Each batch makes 48 cookies.
   - To find out how many more batches are needed to produce 96 cookies:
     \[
     \text{Number of additional batches} = \frac{96 \text{ cookies}}{48 \text{ cookies per batch}} = 2 \text{ batches}
     \]

Therefore, Linda needs to bake **2 more batches** of cookies to have enough for her classmates.

    \rule{\textwidth}{1.5pt} 
    \caption{An example from GSM8K dataset.}
    \label{fig: example_gsm8k} 
    \end{figure*}

\begin{figure*}[!h]

    \rule{\textwidth}{1.5pt} 
    \textbf{Question:} Choose the sentence that goes against common sense. Options: - Sentence A: "Christopher Columbus sailed to North America with three ships: the Toyota, the Pontiac, and the Mazda." - Sentence B: "Christopher Columbus sailed to North America with three ships: the Santa Maria, the Pinta, and the Nina."
    
    \rule{\textwidth}{1.5pt}
    \textcolor{teal}{[None CoT]} 

    Sentence A
    
        \rule{\textwidth}{0.8pt}
        \textcolor{teal}{[Simplified CoT]}
        
        The Santa Maria, the Pinta, and the Nina are sailing ships, while the Toyota, the Pontiac, and the Mazda are cars.
The final answer is Sentence A

        \rule{\textwidth}{0.8pt}
        \textcolor{teal}{[Detailed CoT]}

    To determine which sentence goes against common sense, we'll evaluate the accuracy and historical correctness of the information provided in each option.

**Sentence A: "Christopher Columbus sailed to North America with three ships: the Toyota, the Pontiac, and the Mazda."**

1. **Names of the Ships**: The names mentioned are "Toyota," "Pontiac," and "Mazda."
   - **Analysis**: Toyota and Mazda are names of modern car manufacturing companies, and Pontiac is also associated with an automobile brand. These names have no historical connection to the period of Christopher Columbus or the names of ships used during his voyages.
2. **Historical Accuracy**: Christopher Columbus's voyages took place in the 15th century (1492 specifically for his first voyage) when such modern terms and companies did not exist.

**Sentence B: "Christopher Columbus sailed to North America with three ships: the Santa Maria, the Pinta, and the Nina."**

1. **Names of the Ships**: The names mentioned are "Santa Maria," "Pinta," and "Nina."
   - **Analysis**: These are historically accurate names of the three ships that Christopher Columbus used on his first voyage across the Atlantic Ocean in 1492.
2. **Historical Accuracy**: This information is consistent with the well-documented historical facts about Columbus's voyage.

Conclusion:
**Sentence A** goes against common sense because it lists modern car brands ("Toyota," "Pontiac," and "Mazda") as the names of the ships used by Christopher Columbus in the 15th century, which is factually incorrect and anachronistic.

**Reasoning Steps Summarized**:
1. Evaluate the historical context and time period relevant to Christopher Columbus’s voyages.
2. Identify the historically accurate ship names used by Columbus.
3. Compare the provided ship names in each sentence against the known historical facts.
4. Determine that modern car brand names in Sentence A are out of place and nonsensical in the given context.

    \rule{\textwidth}{1.5pt} 
    \caption{An example from Sensemaking dataset.}
    \label{fig: example_sensemaking} 
    \end{figure*}

\begin{figure*}[!h]

    \rule{\textwidth}{1.5pt} 
    \textbf{Question:} Could Durian cause someone's stomach to feel unwell?
    
    \rule{\textwidth}{1.5pt}
    \textcolor{teal}{[None CoT]} 

    yes
    
        \rule{\textwidth}{0.8pt}
        \textcolor{teal}{[Simplified CoT]}
        
        Durian has a pungent odor that many people describe as being similar to feet and onions. Unpleasant smells can make people feel nauseous.
The final answer is yes

        \rule{\textwidth}{0.8pt}
        \textcolor{teal}{[Detailed CoT]}

    Yes, durian can potentially cause someone's stomach to feel unwell. Here are some detailed reasoning steps to understand why this might happen:

1. **Digestive Sensitivity**
Different people have varying levels of sensitivity to foods. Durian, being a rich and complex fruit, might be harder to digest for some individuals.

**Reasoning:**
- Durian has a high fiber content which, while beneficial in moderate amounts, can cause gas, bloating, or uncomfortable bowel movements in those not accustomed to high-fiber foods.
- High-fat content can also be an issue for individuals with sensitivities or specific digestive conditions, as it might slow down the digestive process.

2. **High Sulfur Content**
Durian contains sulfur compounds, which contribute to its strong smell and unique flavor. 

**Reasoning:**
- Sulfur compounds can be challenging for some people to digest and might lead to stomach discomfort, including bloating and gas.
- People with sulfur sensitivities or related conditions may experience more pronounced gastrointestinal distress.

3. **Fermentation and Bloating**
Durian contains sugars that can ferment in the gut.

**Reasoning:**
- The process of fermentation can produce gas, which might lead to bloating and discomfort.
- If someone has an imbalance in their gut bacteria, the fermentation of these sugars could exacerbate feelings of unwellness in the stomach.

4. **Overconsumption**
Eating too much of any food, including durian, can cause digestive discomfort.

**Reasoning:**
- When consumed in large amounts, the high fiber and fat content in durian can overwhelm the digestive system, leading to symptoms such as stomachache, bloating, and gas.
- Overeating durian could also result in high caloric intake, leading to a feeling of fullness and discomfort.

5. **Preexisting Conditions**
Individuals with certain preexisting health conditions might be more susceptible to stomach discomfort from durian.

**Reasoning:**
- Conditions like Irritable Bowel Syndrome (IBS), gastritis, or other gastrointestinal disorders can make the stomach more sensitive to rich, complex foods like durian.
- Specific allergies or intolerances to compounds in durian can trigger adverse reactions leading to stomach distress.

6. **Interaction with Other Foods**
Combining durian with other foods, particularly certain types that don't mix well with rich, fatty, or high-fiber foods, might cause problems.

**Reasoning:**
- Consuming durian alongside alcohol, dairy, or other high-fat foods can stress the digestive system, leading to nausea or stomach pain.
- Certain food combinations might inhibit proper digestion, causing discomfort.

Conclusion
While durian is a nutritious and delicious fruit for many people, its complex composition can lead to stomach discomfort for others. Factors like individual digestive sensitivity, high sulfur and fiber content, fermentation processes, overeating, preexisting conditions, and interaction with other foods all contribute to this potential issue. To minimize discomfort, it is advisable for individuals to consume durian in moderation and pay attention to their body's response.

    \rule{\textwidth}{1.5pt} 
    \caption{An example from StrategyQA dataset.}
    \label{fig: example_strategyqa} 
    \end{figure*}

\begin{figure*}[!h]

    \rule{\textwidth}{1.5pt} 
    \textbf{Question:} Solve for $x>0$ in the following arithmetic sequence: $1^2, x^2, 3^2, \ldots$.
    
    \rule{\textwidth}{1.5pt}

        \textcolor{teal}{[Simplified CoT]}
        
        The term $x^2$ is simply the average of $1^2 = 1$ and $3^2 = 9$, so $x^2 = (1 + 9)/2 = 5$.  Because $x > 0$, $x = \boxed{\sqrt{5}}$.

        \rule{\textwidth}{0.8pt}
        \textcolor{teal}{[Detailed CoT]}

    To solve for \( x \) in the arithmetic sequence \( 1^2, x^2, 3^2, \ldots \) with \( x > 0 \), we will follow these detailed steps:

1. **Define the Terms of the Sequence**:
   The given sequence is \( 1^2, x^2, 3^2, \ldots \). These are the first three terms of the sequence, where \( 1^2 = 1 \), \( x^2 \), and \( 3^2 = 9 \).
   
2. **Arithmetic Sequence Property**:
   In an arithmetic sequence, the difference between consecutive terms is constant. Let this common difference be \( d \).

3. **Set Up Equations for the Common Difference**:
   From the sequence, we have:
   \[
   x^2 - 1^2 = d \quad \text{and} \quad 3^2 - x^2 = d
   \]
   
   Thus, the two equations based on the common difference \( d \) are:
   \[
   x^2 - 1 = d \quad \text{and} \quad 9 - x^2 = d
   \]
   
4. **Equating the Common Differences**:
   Since both expressions equal \( d \), we can set them equal to each other:
   \[
   x^2 - 1 = 9 - x^2
   \]
   
5. **Solve the Equation**:
   Combine like terms to solve for \( x^2 \):
   \[
   x^2 - 1 = 9 - x^2
   \]
   Add \( x^2 \) to both sides:
   \[
   2x^2 - 1 = 9
   \]
   Add 1 to both sides:
   \[
   2x^2 = 10
   \]
   Divide by 2:
   \[
   x^2 = 5
   \]
   Since \( x > 0 \), take the positive square root of both sides to find \( x \):
   \[
   x = \sqrt{5}
   \]

Therefore, the solution for \( x \) in the given arithmetic sequence is:
\[
x = \sqrt{5}
\]

    \rule{\textwidth}{1.5pt} 
    \caption{An example from MATH-Algebra dataset.}
    \label{fig: example_algebra} 
    \end{figure*}

\begin{figure*}[!h]

    \rule{\textwidth}{1.5pt} 
    \textbf{Question:} How many positive three-digit integers with each digit greater than 4 are divisible by 6?
    
    \rule{\textwidth}{1.5pt}

        \textcolor{teal}{[Simplified CoT]}
        
        To be divisible by 6, a number must have its digits add up to a multiple of 3, and be even.  Therefore, for the hundreds place, the possible digits are $\{5,6,7,8,9\}$, for the tens place the possible digits also are $\{5,6,7,8,9\}$, and for the ones digit, you can only choose from $\{6,8\}$.

First, let us choose 6 for the ones place.  The other two digits must add up to a multiple of 3, making a total of 8 pairs that satisfy that condition: $$\{5,7\}, \{6,6\}, \{6,9\}, \{7,5\}, \{7,8\}, \{8,7\}, \{9,6\}, \{9,9\}.$$

Next, let us choose 8 for the ones place.  The other two digits must be congruent to 1 mod 3, making a total of 8 pairs that satisfy that condition: $$\{5,5\}, \{5,8\}, \{6,7\}, \{7,6\}, \{7,9\}, \{8,5\}, \{8,8\}, \{9,7\}.$$

This makes a total of $\boxed{16}$ numbers.

        \rule{\textwidth}{0.8pt}
        \textcolor{teal}{[Detailed CoT]}

    To determine how many positive three-digit integers with each digit greater than 4 are divisible by 6, we need to consider the criteria that an integer must meet to be divisible by 6. Specifically, a number is divisible by 6 if it is divisible by both 2 and 3.

Step-by-step Reasoning

1. **Divisibility by 2:**
    - A number is divisible by 2 if its last digit is even.
    - The possible digits for a three-digit number where each digit is greater than 4 are 5, 6, 7, 8, 9.
    - Among these digits, the even digits are 6 and 8.

2. **Divisibility by 3:**
    - A number is divisible by 3 if the sum of its digits is divisible by 3.
    - We need to ensure the sum of the digits of the number fits this criterion.

Enumeration of Valid Numbers
Given the constraints (digits greater than 4 and the number must meet divisibility criteria for both 2 and 3), we can construct our solution:

**Step 1: Form digits**
The three-digit number is represented as \( \overline{abc} \), where \( a, b, \) and \( c \) are its digits.

**Step 2: Identify valid choices for \( c \) (last digit, must be even):**
Potential values for \( c \) can be 6 or 8.

Let's assume \( c = 6 \):  
Now we find valid pairs \((a, b)\) such that the sum \(a + b + 6\) is divisible by 3.

- Valid digits for \( a \) and \( b \) are 5, 6, 7, 8, and 9.

Let's check all pairs \((a, b)\):

For \( c = 6 \):

- \(6 \implies a + b \equiv 0 \ (\text{mod} \ 3) \)
- Possible pairs \((a, b)\) where \(a + b\) must be divisible by 3:
    - (5, 7), (5, 9)
    - (6, 6), (6, 9)
    - (7, 5), (7, 8)
    - (8, 7)
    - (9, 5), (9, 6)

For \( c = 8 \):
- \(8 \implies a + b \equiv 1 \ (\text{mod} \ 3)\)
- Possible pairs \((a, b)\) where \(a + b\) + 8  results in sum modulo 3 as:

- (a, b) needs to sum up to \( 1 \ (\text{mod} \ 3)\), this means:
   - (5, 5)
   - (5, 8)
   - (6, 7)
   - (7, 6), (7, 9)
   - (8 , 5), (8, 8)
   - (9, 9)

So combining both :
- Pairs with \( c = 6 \): 10 pairs.
- Pairs with \( c = 8 \): 8 pairs.

**Step 3: Combine and count the numbers of valid solutions:**
Hence, total such numbers = Total Pairs= \(10+ 8 = 18\).

Thus, there are \(18\ larger number\) possible solutions that satisfy the conditions: three-digit positive integers, each digit greater than 4, and divisible by 6.

    \rule{\textwidth}{1.5pt} 
    \caption{An example from MATH-Counting dataset.}
    \label{fig: example_counting} 
    \end{figure*}

\begin{figure*}[!h]

\small
    \rule{\textwidth}{1.5pt} 
    \textbf{Question:} Square $ABCD$ has side length $1$ unit.  Points $E$ and $F$ are on sides $AB$ and $CB$, respectively, with $AE = CF$.  When the square is folded along the lines $DE$ and $DF$, sides $AD$ and $CD$ coincide and lie on diagonal $BD$.  The length of segment $AE$ can be expressed in the form $\sqrt{k}-m$ units. What is the integer value of $k+m$?
    
    \rule{\textwidth}{1.5pt}

        \textcolor{teal}{[Simplified CoT]}
        
        We start by drawing a diagram.  When the paper is folded, sides $AD$ and $CD$ coincide on the longer dashed line, and points $A$ and $C$ meet at $G$, as you can see below.  [asy]
draw((0,0)--(1,0)--(1,1)--(0,1)--cycle);
draw((0,0)--(1,.4)); draw((0,0)--(.4,1));
draw((1,.4)--(.4,1),dashed);
draw((0,0)--(.7,.7),dashed);
label("$A$",(0,1), NW); label("$B$",(1,1), NE); label("$C$",(1,0), SE); label("$D$",(0,0), SW);
label("$F$",(1,.4), E); label("$E$",(.4,1), N);
label("$G$",(.7,.7), NE);
[/asy] Now, we assign variables.  We are looking for the length of $AE$, so let $AE=x$.  Then, $BE=1-x$.  Because of the symmetry of the square and the fold, everything to the left of line $BD$ is a mirror image of everything to the right of $BD$.  Thus, $\triangle BEF$ is an isosceles right triangle (45-45-90), so $EF=\sqrt{2}EB = \sqrt{2}(1-x)$.  Also, $\triangle EGB$ and $\triangle FGB$ are congruent 45-45-90 triangles, so $GB = \frac{EB}{\sqrt{2}} = \frac{(1-x)}{\sqrt{2}}$.

Also, notice that because the way the paper is folded (its original position versus its final position), we have more congruent triangles, $\triangle AED \cong \triangle GED$.  This means that $AD=GD=1$.

Lastly, notice that since $G$ is on $BD$, we have $BD=BG+GD$. $BD$ is a diagonal of the square, so it has side length $\sqrt{2}$, $GD=1$, and $GB = \frac{(1-x)}{\sqrt{2}}$.  Thus, our equation becomes \[\sqrt{2} = 1 + \frac{(1-x)}{\sqrt{2}}.\] Multiplying both sides by $\sqrt{2}$ yields $2=\sqrt{2}+1-x$; solving for $x$ yields $x=\sqrt{2}-1$.  Thus, $AE=\sqrt{2}-1=\sqrt{k}-m$, and we see that $k+m=2+1=\boxed{3}$.

        \rule{\textwidth}{0.8pt}
        \textcolor{teal}{[Detailed CoT]}

    First, we label the points on square \(ABCD\) as follows: with \(A(0, 1)\), \(B(0, 0)\), \(C(1, 0)\), and \(D(1, 1)\). Let \(E\) be a point on \(AB\) with coordinate \((0, y)\), where \(y = AE\), and let \(F\) be a point on \(CB\) with coordinate \((x, 0)\), where \(x = CF\). Since \(AE = CF\), it follows that \(y = x\).

When the square is folded along \(DE\) and \(DF\), the folded sides \(AD\) and \(CD\) align and coincide along diagonal \(BD\). We know \(AD\) will lie on \(BD\), so point \(A\) will overlap somewhere on line \(BD\).

The diagonal \(BD\) of the square can be parameterized as the line from \(B(0, 0)\) to \(D(1, 1)\). The equation of line \(BD\) is:
\[
y = x
\]

Considering points \(E\) on \(AB\), having coordinates \((0, y)\), and points \(F\) on \(CB\), having coordinates \((x, 0)\):

When folding \(AD\) onto \(BD\), point \(A(0, 1)\) will fold along some line segment \(DE\). To understand where \(A\) falls upon folding onto \(BD\), we calculate the perpendicular distance from \(A\) to \(BD\).

The distance \(d\) from a point \((x_1, y_1)\) to a line \(ax + by + c = 0\) is given by:
\[
d = \frac{|ax_1 + by_1 + c|}{\sqrt{a^2 + b^2}}
\]

In our case, the equation of the line \(BD\) is \(y = x\) and can also be written as:
\[
x - y = 0
\]
Thus, we have \(a = 1\), \(b = -1\), and \(c = 0\). The distance from point \(A(0, 1)\) to line \(x - y = 0\) is:
\[
d = \frac{|1 \cdot 0 + (-1) \cdot 1 + 0|}{\sqrt{1^2 + (-1)^2}} = \frac{|0 - 1|}{\sqrt{1 + 1}} = \frac{1}{\sqrt{2}} = \frac{\sqrt{2}}{2}
\]

Since \(E\) lies directly on the \(y\)-axis at \(y = AE\), and \(E\)’s distance should match the perpendicular distance from \(A\) to \(BD\), we solve for \(y = AE\):
\[
y = x = \frac{\sqrt{2}}{2}
\]

Thus, \(x = y = \frac{\sqrt{2}}{2}\).
However, the problem requires the expression in the form \(\sqrt{k} - m\):

Here \(AE = \frac{\sqrt{2}}{2} = \sqrt{2} \cdot \frac{1}{2} = \sqrt{2} \cdot \frac{\sqrt{4}}{2\sqrt{4}} = \sqrt{2} - 1\)

Setting \(\sqrt{k} - m = \sqrt{2} - 1\), we get \(k = 2\) and \(m = 1\). Thus:

\[
k + m = 2 + 1 = 3
\]

The integer value of \(k + m\) is:
\[
\boxed{3}
\]

    \rule{\textwidth}{1.5pt} 
    \caption{An example from MATH-Geometry dataset.}
    \label{fig: example_geometry} 
    \end{figure*}

\begin{figure*}[!h]
\small

\rule{\textwidth}{1.5pt} 
 \textbf{Question:} "Tell me about Charles, Prince of Wales."

\rule{\textwidth}{1.5pt}

    \textcolor{teal}{[Length 100]}
    
    Charles was born at 21:14 (GMT) on 14 November 1948, during the reign of his maternal grandfather, King George VI, as the first child of Princess Elizabeth, Duchess of Edinburgh (later Queen Elizabeth II), and Philip, Duke of Edinburgh. He was delivered by Caesarean section at Buckingham Palace. His parents had three more children, Anne (born 1950), Andrew (born 1960) and Edward (born 1964).

    \rule{\textwidth}{0.8pt}
    \textcolor{teal}{[Length 500]}

Charles was born at 21:14 (GMT) on 14 November 1948, during the reign of his maternal grandfather, King George VI, as the first child of Princess Elizabeth, Duchess of Edinburgh (later Queen Elizabeth II), and Philip, Duke of Edinburgh. He was delivered by Caesarean section at Buckingham Palace. His parents had three more children, Anne (born 1950), Andrew (born 1960) and Edward (born 1964). He was christened Charles Philip Arthur George on 15 December 1948 in the Music Room of Buckingham Palace by the Archbishop of Canterbury, Geoffrey Fisher.
George VI died on 6 February 1952 and Charles's mother acceded to the throne as Elizabeth II; Charles immediately became the heir apparent. Under a charter of Edward III in 1337, and as the monarch's eldest son, he automatically assumed the traditional titles of Duke of Cornwall and, in the Scottish peerage, the titles Duke of Rothesay, Earl of Carrick, Baron of Renfrew, Lord of the Isles, and Prince and Great Steward of Scotland. The following year, Charles attended his mother's coronation at Westminster Abbey.
When Charles turned five, Catherine Peebles was appointed as governess to oversee his education at Buckingham Palace. He then commenced classes at Hill House School in west London in November 1956. Charles was the first heir apparent to attend school, rather than be educated by a private tutor. He did not receive preferential treatment from the school's founder and headmaster, Stuart Townend, who advised the Queen to have Charles train in football, because the boys were never deferential to anyone on the football field. Charles subsequently attended two of his father's former schools: Cheam School in Hampshire, from 1958, followed by Gordonstoun, in the north-east of Scotland, beginning classes there in April 1962. He later became patron of Gordonstoun in May 2024. In his 1994 authorised biography by Jonathan Dimbleby, Charles's parents were described as physically and emotionally distant and Philip was blamed for his disregard of Charles's sensitive nature, including forcing him to attend Gordonstoun, where he was bullied.

    \rule{\textwidth}{0.8pt}
    \textcolor{teal}{[Length 1000]}

Charles was born at 21:14 (GMT) on 14 November 1948, during the reign of his maternal grandfather, King George VI, as the first child of Princess Elizabeth, Duchess of Edinburgh (later Queen Elizabeth II), and Philip, Duke of Edinburgh. He was delivered by Caesarean section at Buckingham Palace. His parents had three more children, Anne (born 1950), Andrew (born 1960) and Edward (born 1964). He was christened Charles Philip Arthur George on 15 December 1948 in the Music Room of Buckingham Palace by the Archbishop of Canterbury, Geoffrey Fisher.
George VI died on 6 February 1952 and Charles's mother acceded to the throne as Elizabeth II; Charles immediately became the heir apparent. Under a charter of Edward III in 1337, and as the monarch's eldest son, he automatically assumed the traditional titles of Duke of Cornwall and, in the Scottish peerage, the titles Duke of Rothesay, Earl of Carrick, Baron of Renfrew, Lord of the Isles, and Prince and Great Steward of Scotland. The following year, Charles attended his mother's coronation at Westminster Abbey.
When Charles turned five, Catherine Peebles was appointed as governess to oversee his education at Buckingham Palace. He then commenced classes at Hill House School in west London in November 1956. Charles was the first heir apparent to attend school, rather than be educated by a private tutor. He did not receive preferential treatment from the school's founder and headmaster, Stuart Townend, who advised the Queen to have Charles train in football, because the boys were never deferential to anyone on the football field. Charles subsequently attended two of his father's former schools: Cheam School in Hampshire, from 1958, followed by Gordonstoun, in the north-east of Scotland, beginning classes there in April 1962. He later became patron of Gordonstoun in May 2024. In his 1994 authorised biography by Jonathan Dimbleby, Charles's parents were described as physically and emotionally distant and Philip was blamed for his disregard of Charles's sensitive nature, including forcing him to attend Gordonstoun, where he was bullied. Though Charles reportedly described Gordonstoun, noted for its especially rigorous curriculum, as "Colditz in kilts", he later praised the school, stating it had taught him "a great deal about myself and my own abilities and disabilities". He said in a 1975 interview he was "glad" he had attended Gordonstoun and that the "toughness of the place" was "much exaggerated". In 1966 Charles spent two terms at the Timbertop campus of Geelong Grammar School in Victoria, Australia, during which time he visited Papua New Guinea on a school trip with his history tutor, Michael Collins Persse. In 1973 Charles described his time at Timbertop as the most enjoyable part of his whole education. Upon his return to Gordonstoun, he emulated his father in becoming head boy, and left in 1967 with six GCE O-levels and two A-levels in history and French, at grades B and C respectively. On his education, Charles later remarked, "I didn't enjoy school as much as I might have; but, that was only because I'm happier at home than anywhere else".
Charles broke royal tradition when he proceeded straight to university after his A-levels, rather than joining the British Armed Forces. In October 1967, he was admitted to Trinity College, Cambridge, where he studied archaeology and anthropology for the first part of the Tripos and then switched to history for the second part. During his second year, he attended the University College of Wales in Aberystwyth, studying Welsh history and the Welsh language for one term. Charles became the first British heir apparent to earn a university degree, graduating in June 1970 from the University of Cambridge with a 2:2 Bachelor of Arts (BA) degree. Following standard practice, in August 1975, his Bachelor of Arts was promoted to a Master of Arts (MA Cantab) degree. Charles served in the Royal Air Force (RAF) and the Royal Navy.

\rule{\textwidth}{1.5pt} 
\caption{Three examples from Wiki Knowledge dataset.}
\label{fig: example_wiki_pop} 
\end{figure*}

\begin{figure*}[!h]

\rule{\textwidth}{1.5pt} 
\textbf{Question:} "Tell me about Eucosmophora atlantis."

\rule{\textwidth}{1.5pt}
\textcolor{teal}{[Unpopular]} \\
    Eucosmophora atlantis is a moth of the family Gracillariidae. It is known from Costa Rica.
The length of the forewings is 3.6–4.5 mm for males and 4-4.8 mm. for females.
The larvae probably feed on a Sapotaceae species and probably mine the leaves of their host plant.

\rule{\textwidth}{1.5pt} 
\caption{An example from Wiki Knowledge dataset.}
\label{fig: example_wiki_unpop} 
\end{figure*}

\clearpage
\section{Results on gemma-2-2b}
\subsection{Pre-trained LLM on Correct Responses}

\subsubsection{Reasoning Tasks}

The visualizations and statistical results on MATH tasks: MATH-Algebra (Figure \ref{fig:math_correct_algebra}, Table \ref{tbl:math_lighteval_math_gemma2_2b_grads_algebra}), MATH-Counting (Figure \ref{fig:math_correct_counting}, Table \ref{tbl:math_lighteval_math_gemma2_2b_grads_counting_probability}), MATH-Geometry (Figure \ref{fig:math_correct_geometry}, Table \ref{tbl:math_lighteval_math_gemma2_2b_grads_geometry}).

\begin{table}[!tbh]
\centering
\small
\scalebox{0.75}{
\begin{tabular}{l|l|l|cccc}
\toprule
\multirow{2}{*}{\textbf{Dataset}} & \multirow{2}{*}{\textbf{Curve}} & \multirow{2}{*}{\textbf{Cot}} &  \multicolumn{4}{c}{\textbf{Mean Absolute Difference (MAD)}} \\
 & &  & Early & Middle & Last & All\\
\midrule
\multirow{16}{*}{Algebra} & \multirow{3}{*}{$s_Q$} 
  & Simplified & 0.61 & 0.53 & 0.49 & 0.53 \\
& & Detailed & 0.45 & 0.37 & 0.41 & 0.39 \\
\cmidrule{2-7}
& \multirow{3}{*}{$s_K$} 
  & Simplified & 0.51 & 0.45 & 0.48 & 0.48 \\
& & Detailed & 0.38 & 0.35 & 0.41 & 0.38 \\
\cmidrule{2-7}
& \multirow{3}{*}{$s_V$} 
  & Simplified & 2.11 & 2.05 & 0.70 & 1.82 \\
& & Detailed & 1.25 & 1.39 & 0.59 & 1.20 \\
\cmidrule{2-7}
& \multirow{3}{*}{$s_O$} 
  & Simplified & 1.91 & 1.48 & 0.80 & 1.42 \\
& & Detailed & 1.01 & 1.03 & 0.45 & 0.87 \\
\cmidrule{2-7}
 & \multirow{3}{*}{$r_Q$} 
  & Simplified & 0.02 & 0.02 & 0.02 & 0.02 \\
& & Detailed & 0.02 & 0.02 & 0.02 & 0.02 \\
\cmidrule{2-7}
& \multirow{3}{*}{$r_K$} 
  & Simplified & 0.03 & 0.01 & 0.03 & 0.02 \\
& & Detailed & 0.03 & 0.02 & 0.03 & 0.02 \\
\cmidrule{2-7}
 & \multirow{3}{*}{$r_V$} 
  & Simplified & 0.03 & 0.04 & 0.05 & 0.03 \\
& & Detailed & 0.02 & 0.03 & 0.03 & 0.03 \\
\cmidrule{2-7}
& \multirow{3}{*}{$r_O$} 
  & Simplified & 0.01 & 0.02 & 0.04 & 0.02 \\
& & Detailed & 0.01 & 0.02 & 0.03 & 0.02 \\
\bottomrule
\end{tabular}
}
\caption{
Statistical results for MATH-Algebra using gemma-2-2b on correct responses.
}
\label{tbl:math_lighteval_math_gemma2_2b_grads_algebra}
\end{table}

\begin{figure}[!h]
\centering
\includegraphics[width=\linewidth]{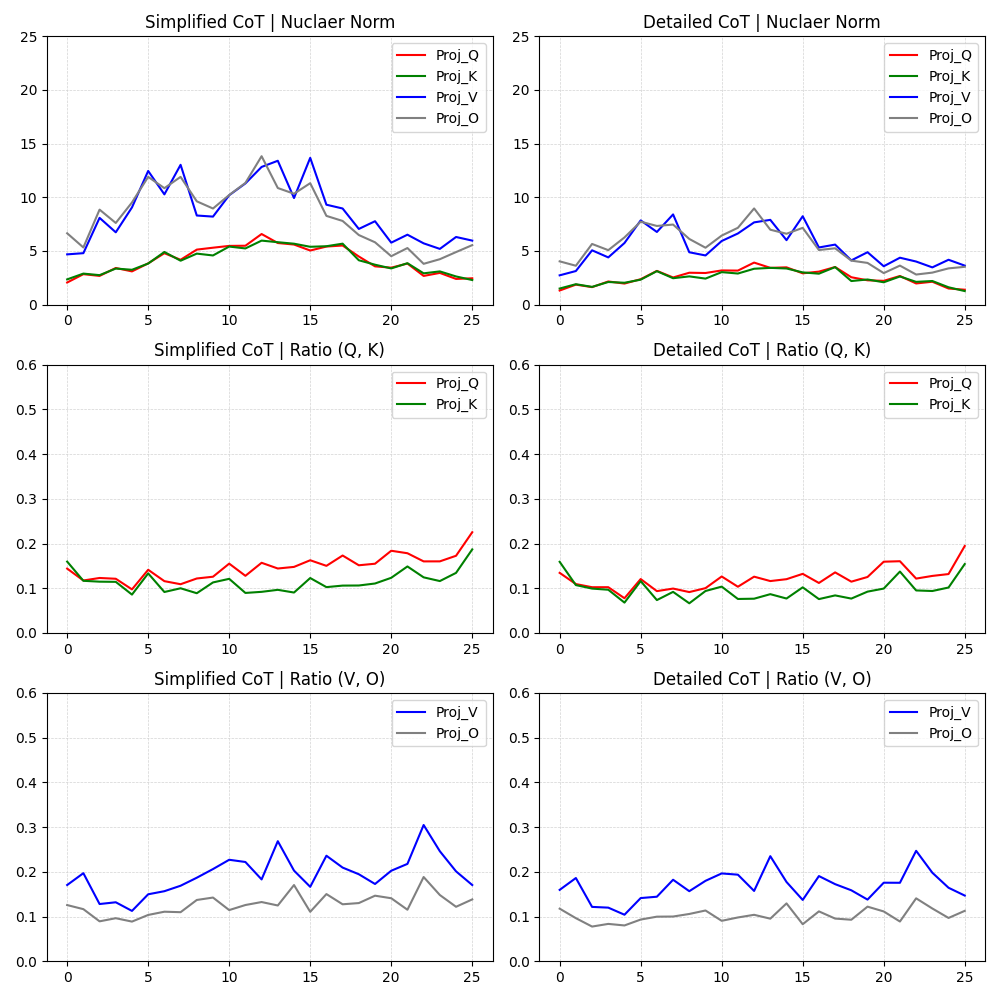}
\caption{Visualization for MATH-Algebra using gemma-2-2b on correct responses.}
\label{fig:math_correct_algebra}
\vspace{-2mm}
\end{figure}

\begin{table}[!tbh]
\centering
\small
\scalebox{0.75}{
\begin{tabular}{l|l|l|cccc}
\toprule
\multirow{2}{*}{\textbf{Dataset}} & \multirow{2}{*}{\textbf{Curve}} & \multirow{2}{*}{\textbf{Cot}} &  \multicolumn{4}{c}{\textbf{Mean Absolute Difference (MAD)}} \\
 & &  & Early & Middle & Last & All\\
\midrule
\multirow{16}{*}{Counting} & \multirow{3}{*}{$s_Q$} 
  & Simplified & 0.72 & 0.71 & 0.54 & 0.65 \\
& & Detailed & 0.56 & 0.52 & 0.43 & 0.50 \\
\cmidrule{2-7}
& \multirow{3}{*}{$s_K$} 
  & Simplified & 0.58 & 0.61 & 0.52 & 0.59 \\
& & Detailed & 0.46 & 0.47 & 0.47 & 0.47 \\
\cmidrule{2-7}
& \multirow{3}{*}{$s_V$} 
  & Simplified & 2.40 & 2.10 & 0.91 & 1.97 \\
& & Detailed & 1.57 & 1.60 & 0.84 & 1.45 \\
\cmidrule{2-7}
& \multirow{3}{*}{$s_O$} 
  & Simplified & 2.20 & 1.54 & 0.80 & 1.53 \\
& & Detailed & 1.27 & 1.23 & 0.48 & 1.04 \\
\cmidrule{2-7}
 & \multirow{3}{*}{$r_Q$} 
  & Simplified & 0.02 & 0.02 & 0.02 & 0.02 \\
& & Detailed & 0.02 & 0.02 & 0.02 & 0.02 \\
\cmidrule{2-7}
& \multirow{3}{*}{$r_K$} 
  & Simplified & 0.03 & 0.01 & 0.02 & 0.02 \\
& & Detailed & 0.03 & 0.02 & 0.03 & 0.02 \\
\cmidrule{2-7}
 & \multirow{3}{*}{$r_V$} 
  & Simplified & 0.03 & 0.03 & 0.04 & 0.03 \\
& & Detailed & 0.03 & 0.03 & 0.03 & 0.03 \\
\cmidrule{2-7}
& \multirow{3}{*}{$r_O$} 
  & Simplified & 0.01 & 0.03 & 0.03 & 0.02 \\
& & Detailed & 0.01 & 0.02 & 0.02 & 0.02 \\
\bottomrule
\end{tabular}
}
\caption{
Statistical results for MATH-Counting using gemma-2-2b on correct responses.
}
\label{tbl:math_lighteval_math_gemma2_2b_grads_counting_probability}
\end{table}

\begin{figure}[!h]
\centering
\includegraphics[width=\linewidth]{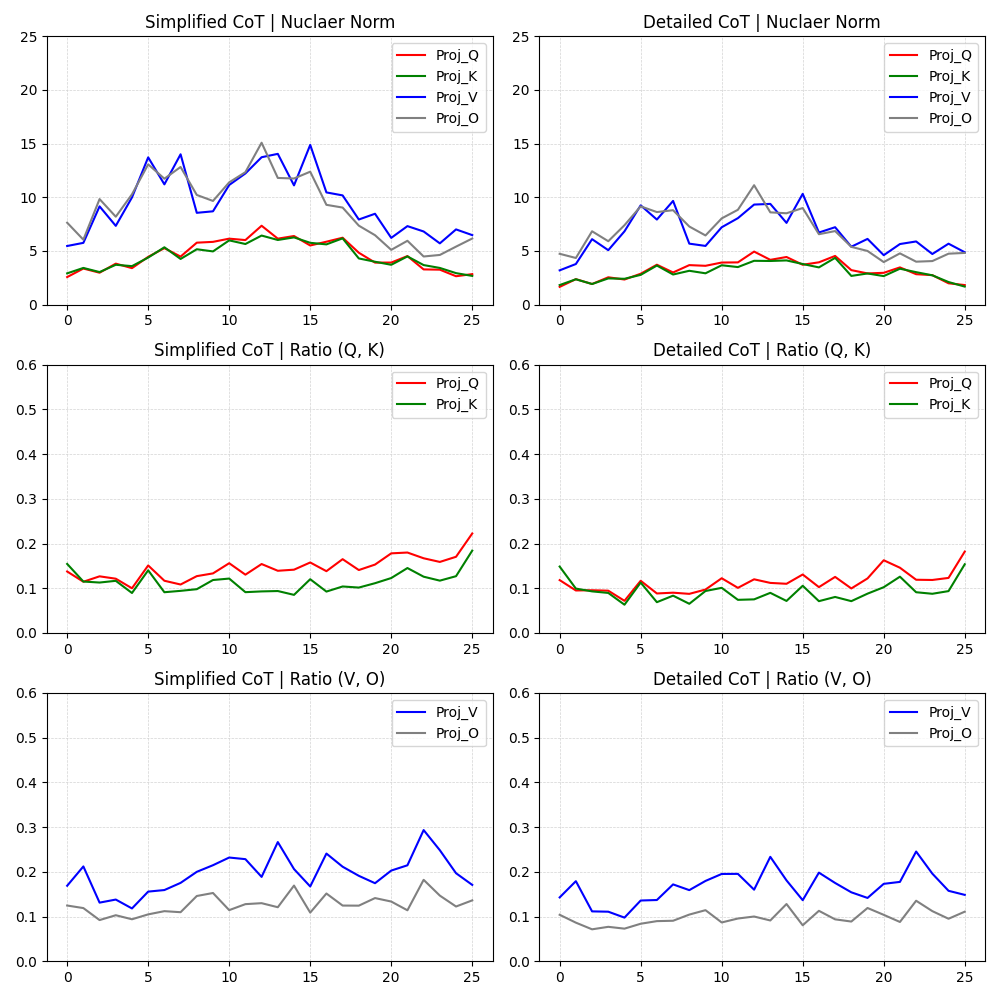}
\caption{Visualization for MATH-Counting using gemma-2-2b on correct responses.}
\label{fig:math_correct_counting}
\vspace{-2mm}
\end{figure}

\begin{table}[!tbh]
\centering
\small
\scalebox{0.75}{
\begin{tabular}{l|l|l|cccc}
\toprule
\multirow{2}{*}{\textbf{Dataset}} & \multirow{2}{*}{\textbf{Curve}} & \multirow{2}{*}{\textbf{Cot}} &  \multicolumn{4}{c}{\textbf{Mean Absolute Difference (MAD)}} \\
 & &  & Early & Middle & Last & All\\
\midrule
\multirow{16}{*}{Geometry} & \multirow{3}{*}{$s_Q$} 
  & Simplified & 0.70 & 0.64 & 0.49 & 0.62 \\
& & Detailed & 0.66 & 0.54 & 0.45 & 0.55 \\
\cmidrule{2-7}
& \multirow{3}{*}{$s_K$} 
  & Simplified & 0.59 & 0.54 & 0.47 & 0.54 \\
& & Detailed & 0.55 & 0.54 & 0.49 & 0.53 \\
\cmidrule{2-7}
& \multirow{3}{*}{$s_V$} 
  & Simplified & 2.12 & 1.92 & 0.51 & 1.71 \\
& & Detailed & 1.76 & 1.88 & 0.89 & 1.65 \\
\cmidrule{2-7}
& \multirow{3}{*}{$s_O$} 
  & Simplified & 1.87 & 1.44 & 0.68 & 1.36 \\
& & Detailed & 1.35 & 1.34 & 0.51 & 1.13 \\
\cmidrule{2-7}
 & \multirow{3}{*}{$r_Q$} 
  & Simplified & 0.02 & 0.02 & 0.01 & 0.02 \\
& & Detailed & 0.02 & 0.02 & 0.03 & 0.02 \\
\cmidrule{2-7}
& \multirow{3}{*}{$r_K$} 
  & Simplified & 0.03 & 0.01 & 0.03 & 0.02 \\
& & Detailed & 0.03 & 0.02 & 0.03 & 0.03 \\
\cmidrule{2-7}
 & \multirow{3}{*}{$r_V$} 
  & Simplified & 0.03 & 0.03 & 0.04 & 0.03 \\
& & Detailed & 0.02 & 0.03 & 0.03 & 0.03 \\
\cmidrule{2-7}
& \multirow{3}{*}{$r_O$} 
  & Simplified & 0.01 & 0.02 & 0.03 & 0.02 \\
& & Detailed & 0.01 & 0.02 & 0.03 & 0.02 \\
\bottomrule
\end{tabular}
}
\caption{
Statistical results for MATH-Geometry using gemma-2-2b on correct responses.
}
\label{tbl:math_lighteval_math_gemma2_2b_grads_geometry}
\end{table}

\begin{figure}[!h]
\centering
\includegraphics[width=\linewidth]{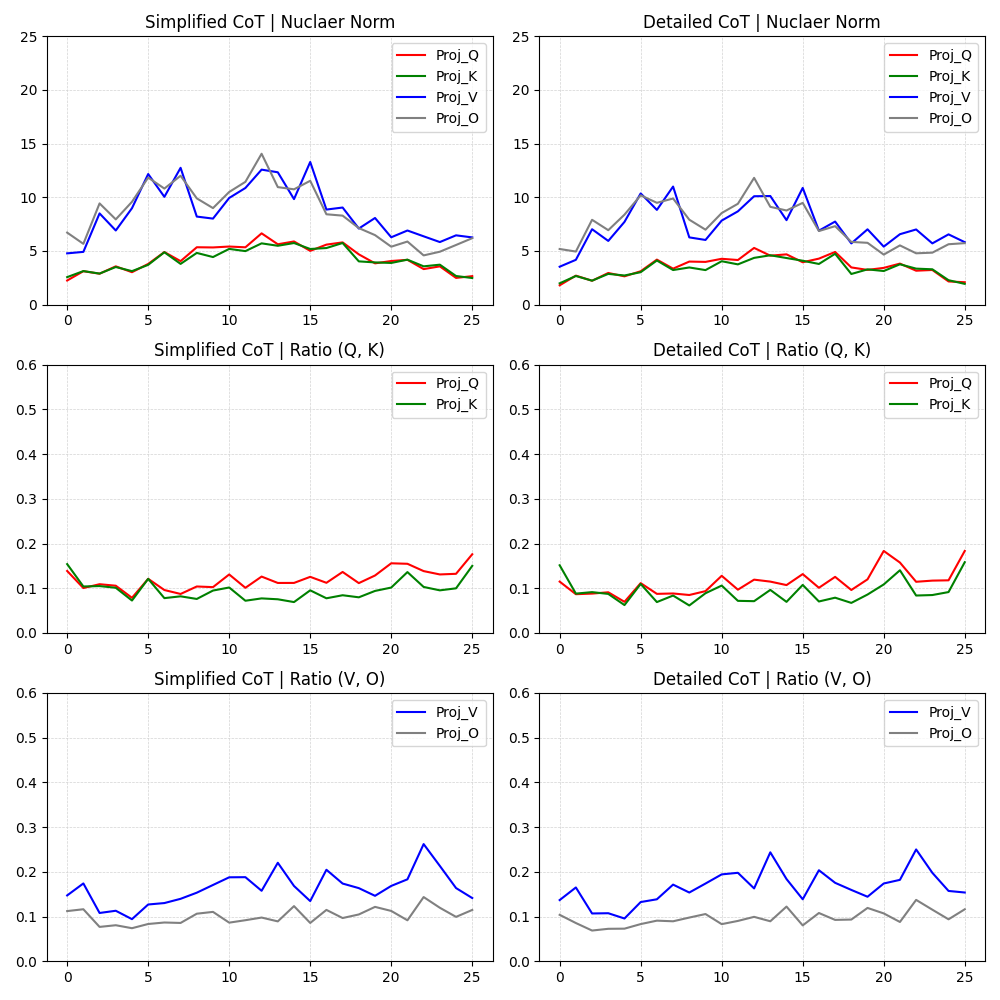}
\caption{Visualization for MATH-Geometry using gemma-2-2b on correct responses.}
\label{fig:math_correct_geometry}
\vspace{-2mm}
\end{figure}

The visualizations and statistical results on other reasoning tasks: AQuA (Figure \ref{fig:reasoning_correct_aqua}, Table \ref{tbl:reasoning_cot_500_gemma2_2b_grads_aqua_train}), GSM8K (Figure \ref{fig:reasoning_correct_gsm8k}, Table \ref{tbl:reasoning_cot_500_gemma2_2b_grads_gsm8k_train}), StrategyQA (Figure \ref{fig:reasoning_correct_strategyqa}, Table \ref{tbl:reasoning_cot_500_gemma2_2b_grads_strategyqa_train}), ECQA (Figure \ref{fig:reasoning_correct_ecqa}, Table \ref{tbl:reasoning_cot_500_gemma2_2b_grads_ecqa_train}), CREAK (Figure \ref{fig:reasoning_correct_creak}, Table \ref{tbl:reasoning_cot_500_gemma2_2b_grads_creak_train}), Sensemaking (Figure \ref{fig:reasoning_correct_sensemaking}, Table \ref{tbl:reasoning_cot_500_gemma2_2b_grads_sensemaking_train}).

\begin{table}[!tbh]
\centering
\small
\scalebox{0.75}{
\begin{tabular}{l|l|l|cccc}
\toprule
\multirow{2}{*}{\textbf{Dataset}} & \multirow{2}{*}{\textbf{Curve}} & \multirow{2}{*}{\textbf{Cot}} &  \multicolumn{4}{c}{\textbf{Mean Absolute Difference (MAD)}} \\
 & &  & Early & Middle & Last & All\\
\midrule
\multirow{24}{*}{AQuA} & \multirow{3}{*}{$s_Q$} 
  & None & 3.29 & 2.07 & 3.44 & 2.77 \\
& & Simplified & 1.65 & 1.06 & 0.90 & 1.17 \\
& & Detailed & 0.53 & 0.46 & 0.46 & 0.46 \\
\cmidrule{2-7}
& \multirow{3}{*}{$s_K$} 
  & None & 4.59 & 2.75 & 3.43 & 3.38 \\
& & Simplified & 1.77 & 1.11 & 0.88 & 1.22 \\
& & Detailed & 0.47 & 0.50 & 0.50 & 0.49 \\
\cmidrule{2-7}
& \multirow{3}{*}{$s_V$} 
  & None & 9.37 & 9.72 & 2.28 & 8.31 \\
& & Simplified & 4.19 & 3.19 & 1.00 & 3.07 \\
& & Detailed & 1.46 & 1.69 & 0.79 & 1.45 \\
\cmidrule{2-7}
& \multirow{3}{*}{$s_O$} 
  & None & 10.04 & 4.36 & 1.91 & 5.17 \\
& & Simplified & 3.87 & 2.19 & 0.94 & 2.26 \\
& & Detailed & 1.23 & 1.25 & 0.55 & 1.06 \\
\cmidrule{2-7}
 & \multirow{3}{*}{$r_Q$} 
  & None & 0.03 & 0.09 & 0.10 & 0.08 \\
& & Simplified & 0.02 & 0.03 & 0.02 & 0.02 \\
& & Detailed & 0.02 & 0.02 & 0.02 & 0.02 \\
\cmidrule{2-7}
& \multirow{3}{*}{$r_K$} 
  & None & 0.04 & 0.05 & 0.09 & 0.05 \\
& & Simplified & 0.03 & 0.02 & 0.03 & 0.02 \\
& & Detailed & 0.03 & 0.02 & 0.03 & 0.02 \\
\cmidrule{2-7}
 & \multirow{3}{*}{$r_V$} 
  & None & 0.04 & 0.07 & 0.03 & 0.05 \\
& & Simplified & 0.03 & 0.04 & 0.05 & 0.04 \\
& & Detailed & 0.03 & 0.03 & 0.03 & 0.03 \\
\cmidrule{2-7}
& \multirow{3}{*}{$r_O$} 
  & None & 0.04 & 0.05 & 0.07 & 0.05 \\
& & Simplified & 0.01 & 0.03 & 0.04 & 0.03 \\
& & Detailed & 0.01 & 0.02 & 0.03 & 0.02 \\
\bottomrule
\end{tabular}
}
\caption{
Statistical results for AQuA using gemma-2-2b on correct responses.
}
\label{tbl:reasoning_cot_500_gemma2_2b_grads_aqua_train}
\end{table}

\begin{figure*}[!h]
\centering
\includegraphics[width=\linewidth]{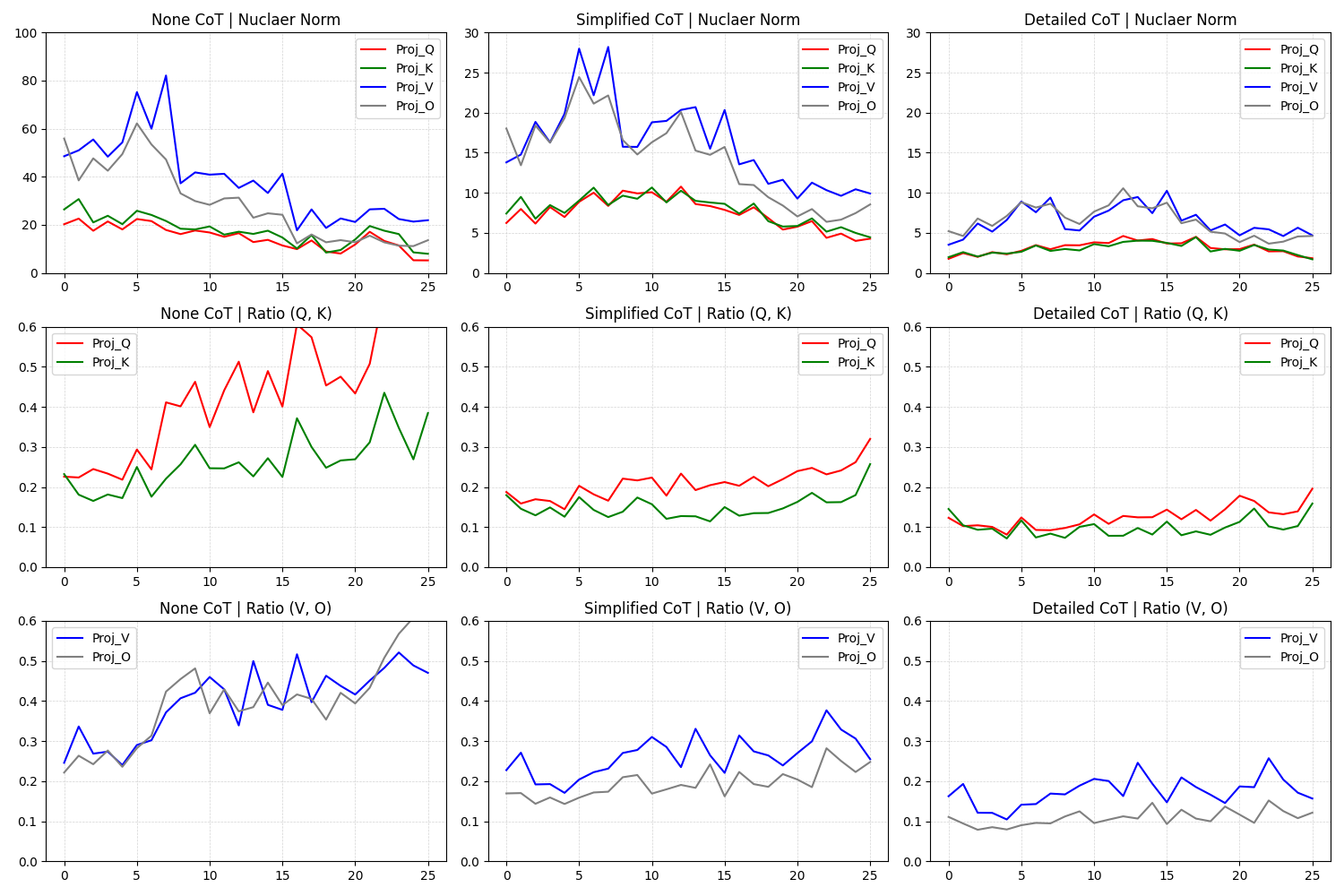}
\caption{Visualization for AQuA using gemma-2-2b on correct responses.}
\label{fig:reasoning_correct_aqua}
\vspace{-2mm}
\end{figure*}

\begin{table}[!tbh]
\centering
\small
\scalebox{0.75}{
\begin{tabular}{l|l|l|cccc}
\toprule
\multirow{2}{*}{\textbf{Dataset}} & \multirow{2}{*}{\textbf{Curve}} & \multirow{2}{*}{\textbf{Cot}} &  \multicolumn{4}{c}{\textbf{Mean Absolute Difference (MAD)}} \\
 & &  & Early & Middle & Last & All\\
\midrule
\multirow{24}{*}{GSM8K} & \multirow{3}{*}{$s_Q$} 
  & None & 2.55 & 1.15 & 3.91 & 1.97 \\
& & Simplified & 0.93 & 0.86 & 0.76 & 0.84 \\
& & Detailed & 0.48 & 0.47 & 0.45 & 0.46 \\
\cmidrule{2-7}
& \multirow{3}{*}{$s_K$} 
  & None & 2.37 & 2.30 & 4.75 & 2.65 \\
& & Simplified & 0.81 & 0.92 & 0.69 & 0.86 \\
& & Detailed & 0.42 & 0.55 & 0.49 & 0.50 \\
\cmidrule{2-7}
& \multirow{3}{*}{$s_V$} 
  & None & 9.17 & 9.09 & 3.56 & 8.21 \\
& & Simplified & 2.48 & 2.15 & 0.87 & 1.98 \\
& & Detailed & 1.40 & 1.62 & 0.82 & 1.39 \\
\cmidrule{2-7}
& \multirow{3}{*}{$s_O$} 
  & None & 9.14 & 3.86 & 2.05 & 4.61 \\
& & Simplified & 2.13 & 1.54 & 0.57 & 1.43 \\
& & Detailed & 1.20 & 1.19 & 0.45 & 1.00 \\
\cmidrule{2-7}
 & \multirow{3}{*}{$r_Q$} 
  & None & 0.03 & 0.03 & 0.08 & 0.05 \\
& & Simplified & 0.02 & 0.03 & 0.02 & 0.02 \\
& & Detailed & 0.02 & 0.02 & 0.02 & 0.02 \\
\cmidrule{2-7}
& \multirow{3}{*}{$r_K$} 
  & None & 0.04 & 0.04 & 0.07 & 0.04 \\
& & Simplified & 0.04 & 0.02 & 0.02 & 0.02 \\
& & Detailed & 0.03 & 0.02 & 0.02 & 0.02 \\
\cmidrule{2-7}
 & \multirow{3}{*}{$r_V$} 
  & None & 0.07 & 0.05 & 0.04 & 0.05 \\
& & Simplified & 0.05 & 0.04 & 0.04 & 0.04 \\
& & Detailed & 0.04 & 0.04 & 0.04 & 0.04 \\
\cmidrule{2-7}
& \multirow{3}{*}{$r_O$} 
  & None & 0.05 & 0.04 & 0.07 & 0.05 \\
& & Simplified & 0.02 & 0.04 & 0.04 & 0.03 \\
& & Detailed & 0.01 & 0.03 & 0.03 & 0.02 \\
\bottomrule
\end{tabular}
}
\caption{
Statistical results for GSM8K using gemma-2-2b on correct responses.
}
\label{tbl:reasoning_cot_500_gemma2_2b_grads_gsm8k_train}
\end{table}

\begin{figure*}[!h]
\centering
\includegraphics[width=\linewidth]{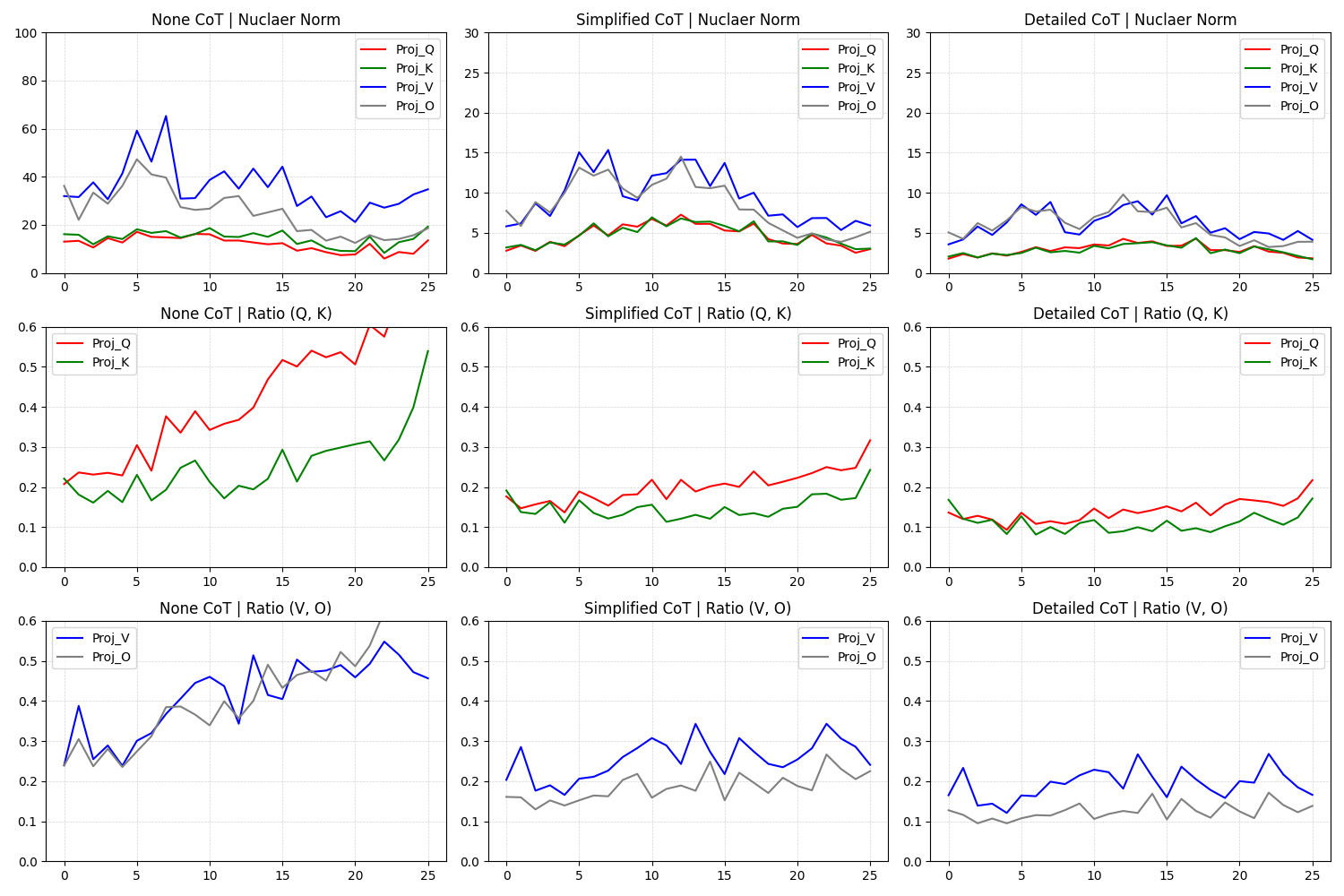}
\caption{Visualization for GSM8K using gemma-2-2b on correct responses.}
\label{fig:reasoning_correct_gsm8k}
\vspace{-2mm}
\end{figure*}

\begin{table}[!tbh]
\centering
\small
\scalebox{0.75}{
\begin{tabular}{l|l|l|cccc}
\toprule
\multirow{2}{*}{\textbf{Dataset}} & \multirow{2}{*}{\textbf{Curve}} & \multirow{2}{*}{\textbf{Cot}} &  \multicolumn{4}{c}{\textbf{Mean Absolute Difference (MAD)}} \\
 & &  & Early & Middle & Last & All\\
\midrule
\multirow{24}{*}{StrategyQA} & \multirow{3}{*}{$s_Q$} 
  & None & 10.57 & 6.09 & 4.57 & 7.07 \\
& & Simplified & 1.35 & 0.62 & 1.26 & 0.95 \\
& & Detailed & 0.61 & 0.66 & 0.53 & 0.60 \\
\cmidrule{2-7}
& \multirow{3}{*}{$s_K$} 
  & None & 10.24 & 9.84 & 4.23 & 9.08 \\
& & Simplified & 0.99 & 1.00 & 1.11 & 0.98 \\
& & Detailed & 0.47 & 0.63 & 0.47 & 0.53 \\
\cmidrule{2-7}
& \multirow{3}{*}{$s_V$} 
  & None & 34.59 & 42.47 & 20.23 & 36.18 \\
& & Simplified & 3.80 & 3.19 & 1.85 & 3.12 \\
& & Detailed & 2.06 & 2.02 & 0.90 & 1.80 \\
\cmidrule{2-7}
& \multirow{3}{*}{$s_O$} 
  & None & 27.75 & 12.40 & 4.63 & 13.80 \\
& & Simplified & 3.25 & 1.72 & 1.19 & 1.95 \\
& & Detailed & 1.70 & 1.63 & 0.84 & 1.44 \\
\cmidrule{2-7}
 & \multirow{3}{*}{$r_Q$} 
  & None & 0.03 & 0.06 & 0.08 & 0.06 \\
& & Simplified & 0.03 & 0.03 & 0.03 & 0.03 \\
& & Detailed & 0.03 & 0.02 & 0.02 & 0.02 \\
\cmidrule{2-7}
& \multirow{3}{*}{$r_K$} 
  & None & 0.02 & 0.03 & 0.04 & 0.03 \\
& & Simplified & 0.04 & 0.02 & 0.03 & 0.02 \\
& & Detailed & 0.03 & 0.02 & 0.02 & 0.02 \\
\cmidrule{2-7}
 & \multirow{3}{*}{$r_V$} 
  & None & 0.07 & 0.09 & 0.04 & 0.07 \\
& & Simplified & 0.04 & 0.06 & 0.05 & 0.05 \\
& & Detailed & 0.02 & 0.04 & 0.04 & 0.03 \\
\cmidrule{2-7}
& \multirow{3}{*}{$r_O$} 
  & None & 0.05 & 0.03 & 0.07 & 0.04 \\
& & Simplified & 0.02 & 0.04 & 0.07 & 0.04 \\
& & Detailed & 0.01 & 0.02 & 0.03 & 0.02 \\
\bottomrule
\end{tabular}
}
\caption{
Statistical results for StrategyQA using gemma-2-2b on correct responses.
}
\label{tbl:reasoning_cot_500_gemma2_2b_grads_strategyqa_train}
\end{table}

\begin{figure*}[!h]
\centering
\includegraphics[width=\linewidth]{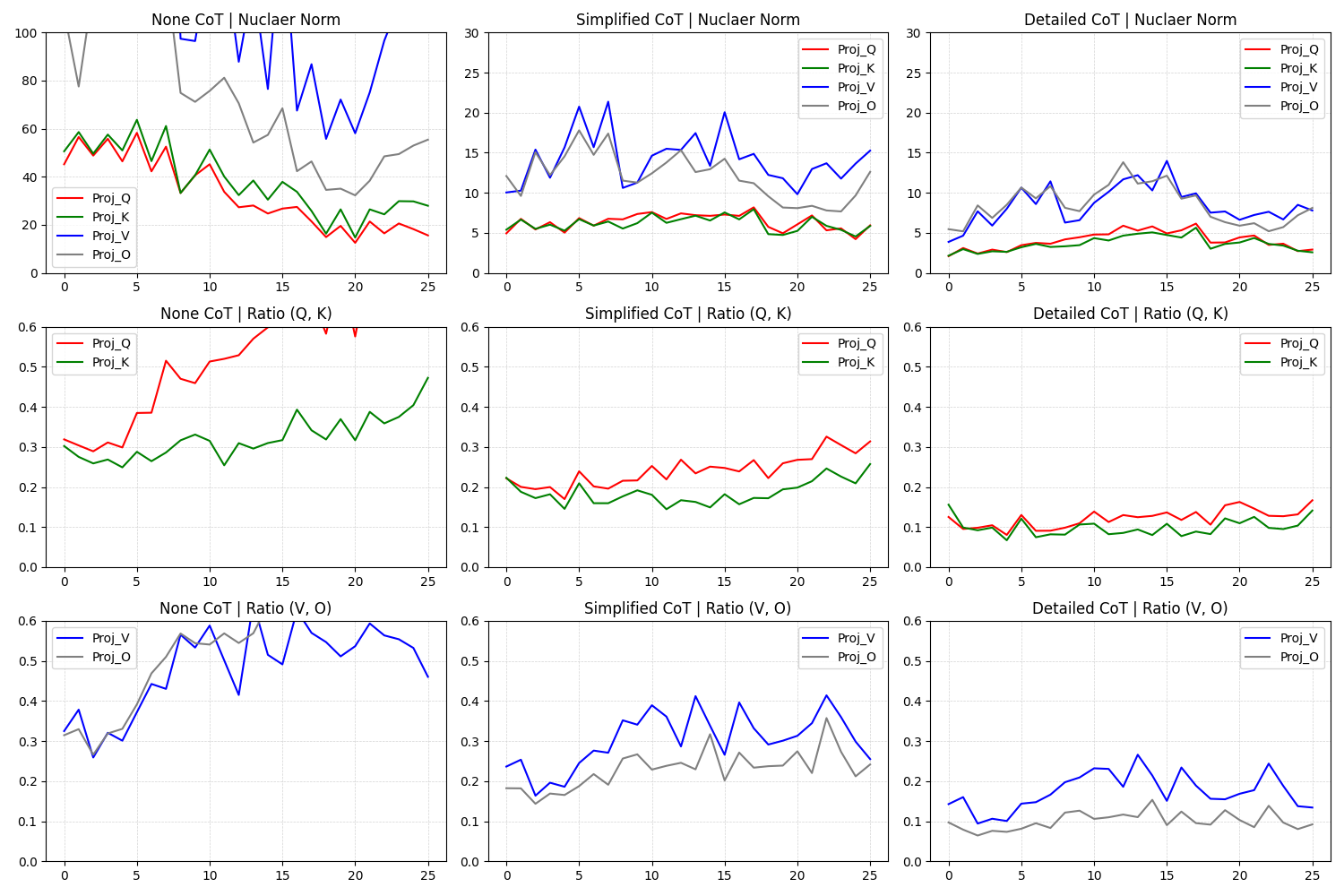}
\caption{Visualization for StrategyQA using gemma-2-2b on correct responses.}
\label{fig:reasoning_correct_strategyqa}
\vspace{-2mm}
\end{figure*}

\begin{table}[!tbh]
\centering
\small
\scalebox{0.75}{
\begin{tabular}{l|l|l|cccc}
\toprule
\multirow{2}{*}{\textbf{Dataset}} & \multirow{2}{*}{\textbf{Curve}} & \multirow{2}{*}{\textbf{Cot}} &  \multicolumn{4}{c}{\textbf{Mean Absolute Difference (MAD)}} \\
 & &  & Early & Middle & Last & All\\
\midrule
\multirow{24}{*}{ECQA} & \multirow{3}{*}{$s_Q$} 
  & None & 17.07 & 8.82 & 7.27 & 10.75 \\
& & Simplified & 1.90 & 1.34 & 1.48 & 1.47 \\
& & Detailed & 0.72 & 0.77 & 0.56 & 0.69 \\
\cmidrule{2-7}
& \multirow{3}{*}{$s_K$} 
  & None & 16.58 & 13.87 & 14.35 & 14.48 \\
& & Simplified & 1.52 & 2.08 & 2.22 & 1.82 \\
& & Detailed & 0.54 & 0.77 & 0.63 & 0.65 \\
\cmidrule{2-7}
& \multirow{3}{*}{$s_V$} 
  & None & 46.14 & 51.57 & 24.02 & 45.96 \\
& & Simplified & 6.01 & 5.32 & 3.23 & 5.19 \\
& & Detailed & 2.51 & 2.32 & 1.52 & 2.21 \\
\cmidrule{2-7}
& \multirow{3}{*}{$s_O$} 
  & None & 42.31 & 18.79 & 3.24 & 20.68 \\
& & Simplified & 5.41 & 2.82 & 1.33 & 3.04 \\
& & Detailed & 2.08 & 1.92 & 0.75 & 1.68 \\
\cmidrule{2-7}
 & \multirow{3}{*}{$r_Q$} 
  & None & 0.02 & 0.07 & 0.10 & 0.06 \\
& & Simplified & 0.02 & 0.04 & 0.03 & 0.03 \\
& & Detailed & 0.02 & 0.02 & 0.02 & 0.02 \\
\cmidrule{2-7}
& \multirow{3}{*}{$r_K$} 
  & None & 0.03 & 0.03 & 0.07 & 0.04 \\
& & Simplified & 0.03 & 0.02 & 0.03 & 0.03 \\
& & Detailed & 0.03 & 0.02 & 0.02 & 0.02 \\
\cmidrule{2-7}
 & \multirow{3}{*}{$r_V$} 
  & None & 0.05 & 0.07 & 0.06 & 0.06 \\
& & Simplified & 0.05 & 0.06 & 0.04 & 0.05 \\
& & Detailed & 0.03 & 0.04 & 0.03 & 0.03 \\
\cmidrule{2-7}
& \multirow{3}{*}{$r_O$} 
  & None & 0.03 & 0.04 & 0.09 & 0.05 \\
& & Simplified & 0.02 & 0.04 & 0.05 & 0.04 \\
& & Detailed & 0.01 & 0.03 & 0.02 & 0.02 \\
\bottomrule
\end{tabular}
}
\caption{
Statistical results for ECQA using gemma-2-2b on correct responses.
}
\label{tbl:reasoning_cot_500_gemma2_2b_grads_ecqa_train}
\end{table}

\begin{figure*}[!h]
\centering
\includegraphics[width=\linewidth]{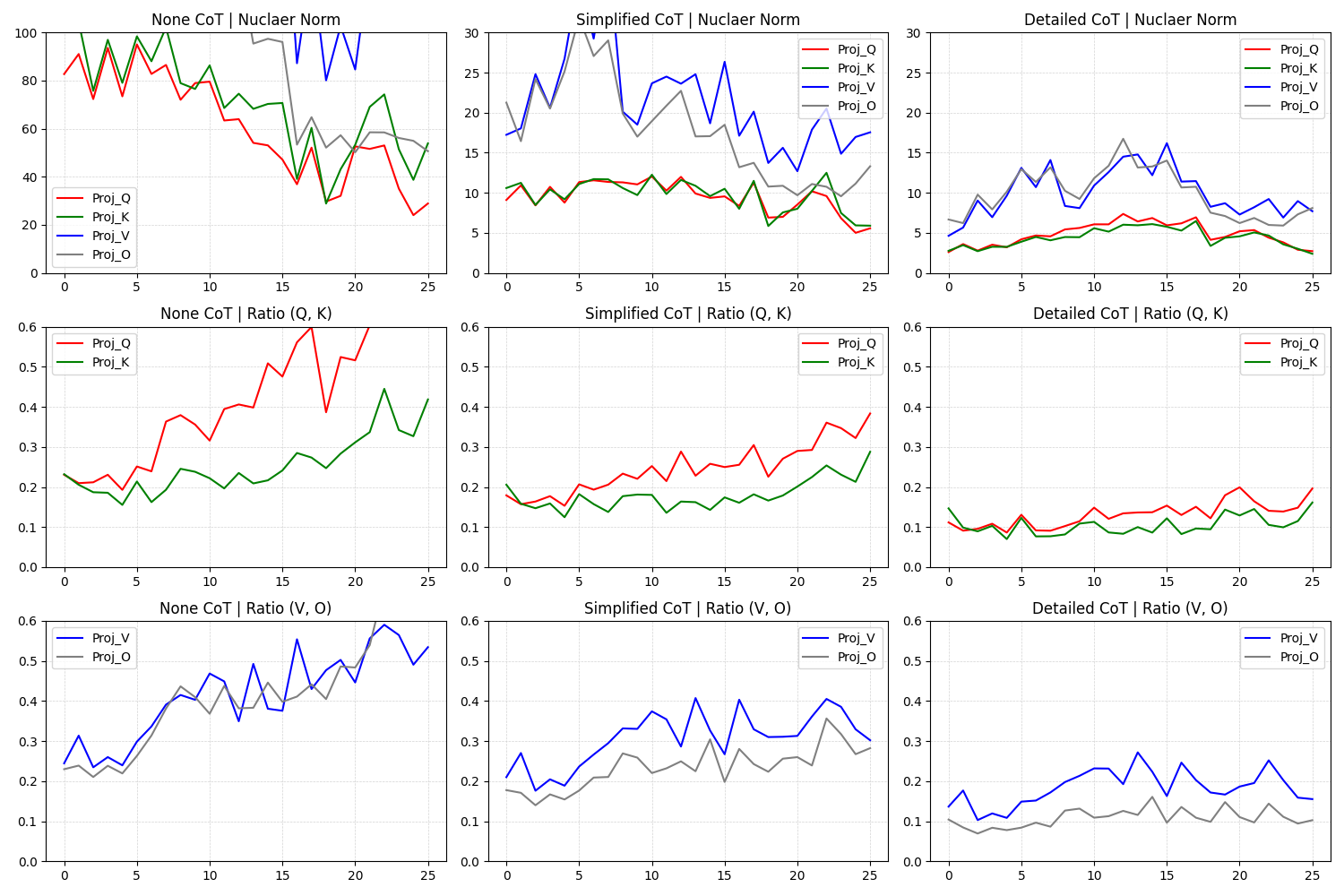}
\caption{Visualization for ECQA using gemma-2-2b on correct responses.}
\label{fig:reasoning_correct_ecqa}
\vspace{-2mm}
\end{figure*}

\begin{table}[!tbh]
\centering
\small
\scalebox{0.75}{
\begin{tabular}{l|l|l|cccc}
\toprule
\multirow{2}{*}{\textbf{Dataset}} & \multirow{2}{*}{\textbf{Curve}} & \multirow{2}{*}{\textbf{Cot}} &  \multicolumn{4}{c}{\textbf{Mean Absolute Difference (MAD)}} \\
 & &  & Early & Middle & Last & All\\
\midrule
\multirow{24}{*}{CREAK} & \multirow{3}{*}{$s_Q$} 
  & None & 15.91 & 7.63 & 9.89 & 10.00 \\
& & Simplified & 1.89 & 1.12 & 1.59 & 1.38 \\
& & Detailed & 0.75 & 0.71 & 0.56 & 0.67 \\
\cmidrule{2-7}
& \multirow{3}{*}{$s_K$} 
  & None & 18.49 & 11.02 & 12.53 & 13.18 \\
& & Simplified & 1.50 & 1.76 & 1.67 & 1.58 \\
& & Detailed & 0.60 & 0.73 & 0.56 & 0.63 \\
\cmidrule{2-7}
& \multirow{3}{*}{$s_V$} 
  & None & 41.16 & 41.56 & 15.30 & 37.29 \\
& & Simplified & 5.76 & 4.92 & 3.16 & 4.93 \\
& & Detailed & 2.45 & 2.32 & 1.09 & 2.10 \\
\cmidrule{2-7}
& \multirow{3}{*}{$s_O$} 
  & None & 40.93 & 15.01 & 4.05 & 18.38 \\
& & Simplified & 5.45 & 2.46 & 1.60 & 2.97 \\
& & Detailed & 2.03 & 1.81 & 0.91 & 1.64 \\
\cmidrule{2-7}
 & \multirow{3}{*}{$r_Q$} 
  & None & 0.02 & 0.08 & 0.11 & 0.07 \\
& & Simplified & 0.02 & 0.03 & 0.05 & 0.03 \\
& & Detailed & 0.03 & 0.02 & 0.02 & 0.02 \\
\cmidrule{2-7}
& \multirow{3}{*}{$r_K$} 
  & None & 0.03 & 0.03 & 0.07 & 0.04 \\
& & Simplified & 0.04 & 0.02 & 0.04 & 0.03 \\
& & Detailed & 0.03 & 0.02 & 0.02 & 0.02 \\
\cmidrule{2-7}
 & \multirow{3}{*}{$r_V$} 
  & None & 0.05 & 0.07 & 0.03 & 0.05 \\
& & Simplified & 0.04 & 0.06 & 0.03 & 0.05 \\
& & Detailed & 0.03 & 0.04 & 0.04 & 0.04 \\
\cmidrule{2-7}
& \multirow{3}{*}{$r_O$} 
  & None & 0.04 & 0.05 & 0.10 & 0.06 \\
& & Simplified & 0.02 & 0.04 & 0.06 & 0.04 \\
& & Detailed & 0.01 & 0.03 & 0.03 & 0.02 \\
\bottomrule
\end{tabular}
}
\caption{
Statistical results for CREAK using gemma-2-2b on correct responses.
}
\label{tbl:reasoning_cot_500_gemma2_2b_grads_creak_train}
\end{table}

\begin{figure*}[!h]
\centering
\includegraphics[width=\linewidth]{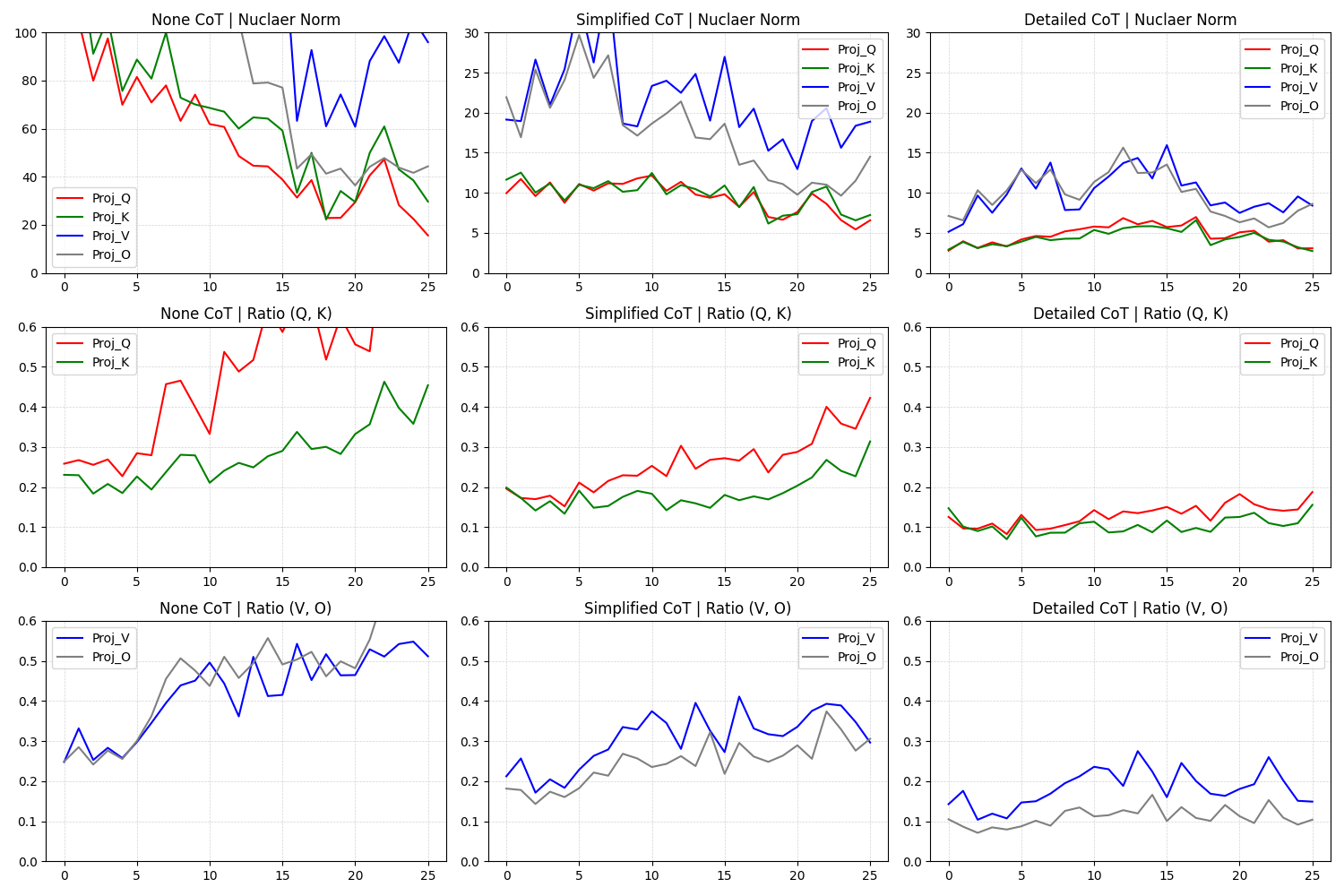}
\caption{Visualization for CREAK using gemma-2-2b on correct responses.}
\label{fig:reasoning_correct_creak}
\vspace{-2mm}
\end{figure*}

\begin{table}[!tbh]
\centering
\small
\scalebox{0.75}{
\begin{tabular}{l|l|l|cccc}
\toprule
\multirow{2}{*}{\textbf{Dataset}} & \multirow{2}{*}{\textbf{Curve}} & \multirow{2}{*}{\textbf{Cot}} &  \multicolumn{4}{c}{\textbf{Mean Absolute Difference (MAD)}} \\
 & &  & Early & Middle & Last & All\\
\midrule
\multirow{24}{*}{Sensemaking} & \multirow{3}{*}{$s_Q$} 
  & None & 9.01 & 4.68 & 6.31 & 6.48 \\
& & Simplified & 2.05 & 1.53 & 2.12 & 1.74 \\
& & Detailed & 0.69 & 0.69 & 0.53 & 0.64 \\
\cmidrule{2-7}
& \multirow{3}{*}{$s_K$} 
  & None & 7.59 & 8.43 & 10.70 & 8.79 \\
& & Simplified & 1.55 & 2.23 & 2.03 & 1.93 \\
& & Detailed & 0.54 & 0.73 & 0.62 & 0.63 \\
\cmidrule{2-7}
& \multirow{3}{*}{$s_V$} 
  & None & 23.08 & 25.63 & 12.47 & 23.21 \\
& & Simplified & 6.63 & 5.80 & 3.85 & 5.84 \\
& & Detailed & 2.05 & 2.02 & 1.21 & 1.86 \\
\cmidrule{2-7}
& \multirow{3}{*}{$s_O$} 
  & None & 24.96 & 9.49 & 4.32 & 11.66 \\
& & Simplified & 6.47 & 2.87 & 1.89 & 3.50 \\
& & Detailed & 1.73 & 1.74 & 0.67 & 1.47 \\
\cmidrule{2-7}
 & \multirow{3}{*}{$r_Q$} 
  & None & 0.02 & 0.06 & 0.07 & 0.06 \\
& & Simplified & 0.03 & 0.03 & 0.05 & 0.03 \\
& & Detailed & 0.03 & 0.02 & 0.02 & 0.02 \\
\cmidrule{2-7}
& \multirow{3}{*}{$r_K$} 
  & None & 0.01 & 0.04 & 0.09 & 0.04 \\
& & Simplified & 0.03 & 0.02 & 0.04 & 0.03 \\
& & Detailed & 0.04 & 0.02 & 0.03 & 0.02 \\
\cmidrule{2-7}
 & \multirow{3}{*}{$r_V$} 
  & None & 0.05 & 0.07 & 0.04 & 0.06 \\
& & Simplified & 0.05 & 0.05 & 0.03 & 0.04 \\
& & Detailed & 0.03 & 0.04 & 0.04 & 0.03 \\
\cmidrule{2-7}
& \multirow{3}{*}{$r_O$} 
  & None & 0.04 & 0.05 & 0.10 & 0.05 \\
& & Simplified & 0.03 & 0.04 & 0.05 & 0.03 \\
& & Detailed & 0.01 & 0.03 & 0.02 & 0.02 \\
\bottomrule
\end{tabular}
}
\caption{
Statistical results for Sensemaking using gemma-2-2b on correct responses.
}
\label{tbl:reasoning_cot_500_gemma2_2b_grads_sensemaking_train}
\end{table}

\begin{figure*}[!h]
\centering
\includegraphics[width=\linewidth]{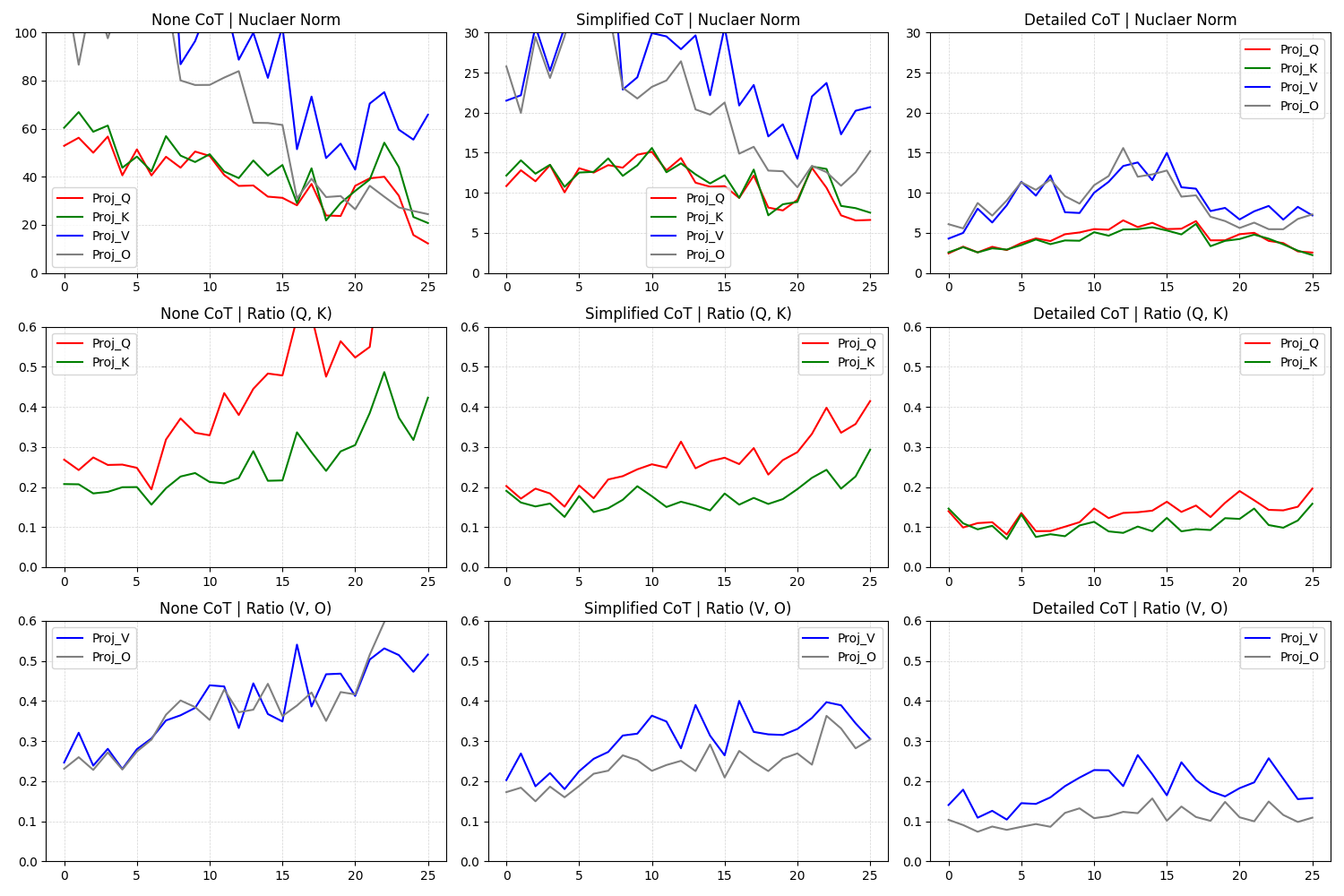}
\caption{Visualization for Sensemaking using gemma-2-2b on correct responses.}
\label{fig:reasoning_correct_sensemaking}
\vspace{-2mm}
\end{figure*}

\subsubsection{Wiki Tasks}

The visualizations and statistical results on Wiki tasks are shown in Figure \ref{fig:wiki_correct} and Table \ref{tbl:wiki_wiki_popularity_gemma2_2b_grads}.

\begin{table}[!tbh]
\centering
\small
\scalebox{0.75}{
\begin{tabular}{l|l|l|cccc}
\toprule
\multirow{2}{*}{\textbf{Dataset}} & \multirow{2}{*}{\textbf{Curve}} & \multirow{2}{*}{\textbf{Cot}} &  \multicolumn{4}{c}{\textbf{Mean Absolute Difference (MAD)}} \\
 & &  & Early & Middle & Last & All\\
\midrule
\multirow{32}{*}{Wiki} & \multirow{3}{*}{$s_Q$} 
  & Len 100 & 1.05 & 0.57 & 1.09 & 0.81 \\
& & Len 500 & 0.71 & 0.49 & 0.69 & 0.57 \\
& & Len 1000 & 0.65 & 0.48 & 0.54 & 0.52 \\
& & Unpopular & 1.84 & 0.86 & 1.20 & 1.20 \\
\cmidrule{2-7}
& \multirow{3}{*}{$s_K$} 
  & Len 100 & 0.76 & 0.77 & 0.94 & 0.77 \\
& & Len 500 & 0.53 & 0.46 & 0.56 & 0.49 \\
& & Len 1000 & 0.50 & 0.37 & 0.46 & 0.42 \\
& & Unpopular & 1.60 & 1.17 & 0.95 & 1.21 \\
\cmidrule{2-7}
& \multirow{3}{*}{$s_V$} 
  & Len 100 & 3.02 & 2.84 & 1.42 & 2.58 \\
& & Len 500 & 2.38 & 2.07 & 1.03 & 1.94 \\
& & Len 1000 & 2.05 & 1.73 & 0.89 & 1.65 \\
& & Unpopular & 3.96 & 3.47 & 1.89 & 3.27 \\
\cmidrule{2-7}
& \multirow{3}{*}{$s_O$} 
  & Len 100 & 2.63 & 1.76 & 1.87 & 1.94 \\
& & Len 500 & 1.97 & 1.50 & 1.45 & 1.57 \\
& & Len 1000 & 1.76 & 1.32 & 1.23 & 1.38 \\
& & Unpopular & 3.26 & 1.88 & 1.96 & 2.15 \\
\cmidrule{2-7}
 & \multirow{3}{*}{$r_Q$} 
  & Len 100 & 0.03 & 0.02 & 0.02 & 0.02 \\
& & Len 500 & 0.02 & 0.02 & 0.01 & 0.02 \\
& & Len 1000 & 0.02 & 0.02 & 0.01 & 0.02 \\
& & Unpopular & 0.03 & 0.01 & 0.03 & 0.02 \\
\cmidrule{2-7}
& \multirow{3}{*}{$r_K$} 
  & Len 100 & 0.04 & 0.02 & 0.02 & 0.02 \\
& & Len 500 & 0.03 & 0.01 & 0.01 & 0.02 \\
& & Len 1000 & 0.03 & 0.01 & 0.01 & 0.02 \\
& & Unpopular & 0.03 & 0.02 & 0.03 & 0.02 \\
\cmidrule{2-7}
 & \multirow{3}{*}{$r_V$} 
  & Len 100 & 0.03 & 0.05 & 0.08 & 0.05 \\
& & Len 500 & 0.02 & 0.04 & 0.05 & 0.03 \\
& & Len 1000 & 0.02 & 0.03 & 0.04 & 0.03 \\
& & Unpopular & 0.04 & 0.05 & 0.09 & 0.05 \\
\cmidrule{2-7}
& \multirow{3}{*}{$r_O$} 
  & Len 100 & 0.02 & 0.04 & 0.08 & 0.04 \\
& & Len 500 & 0.01 & 0.02 & 0.04 & 0.02 \\
& & Len 1000 & 0.01 & 0.02 & 0.03 & 0.02 \\
& & Unpopular & 0.02 & 0.04 & 0.08 & 0.04 \\
\bottomrule
\end{tabular}
}
\caption{
Statistical results for Wiki using gemma-2-2b on correct responses.
}
\label{tbl:wiki_wiki_popularity_gemma2_2b_grads}
\end{table}

\begin{figure*}[!h]
\centering
\includegraphics[width=\linewidth]{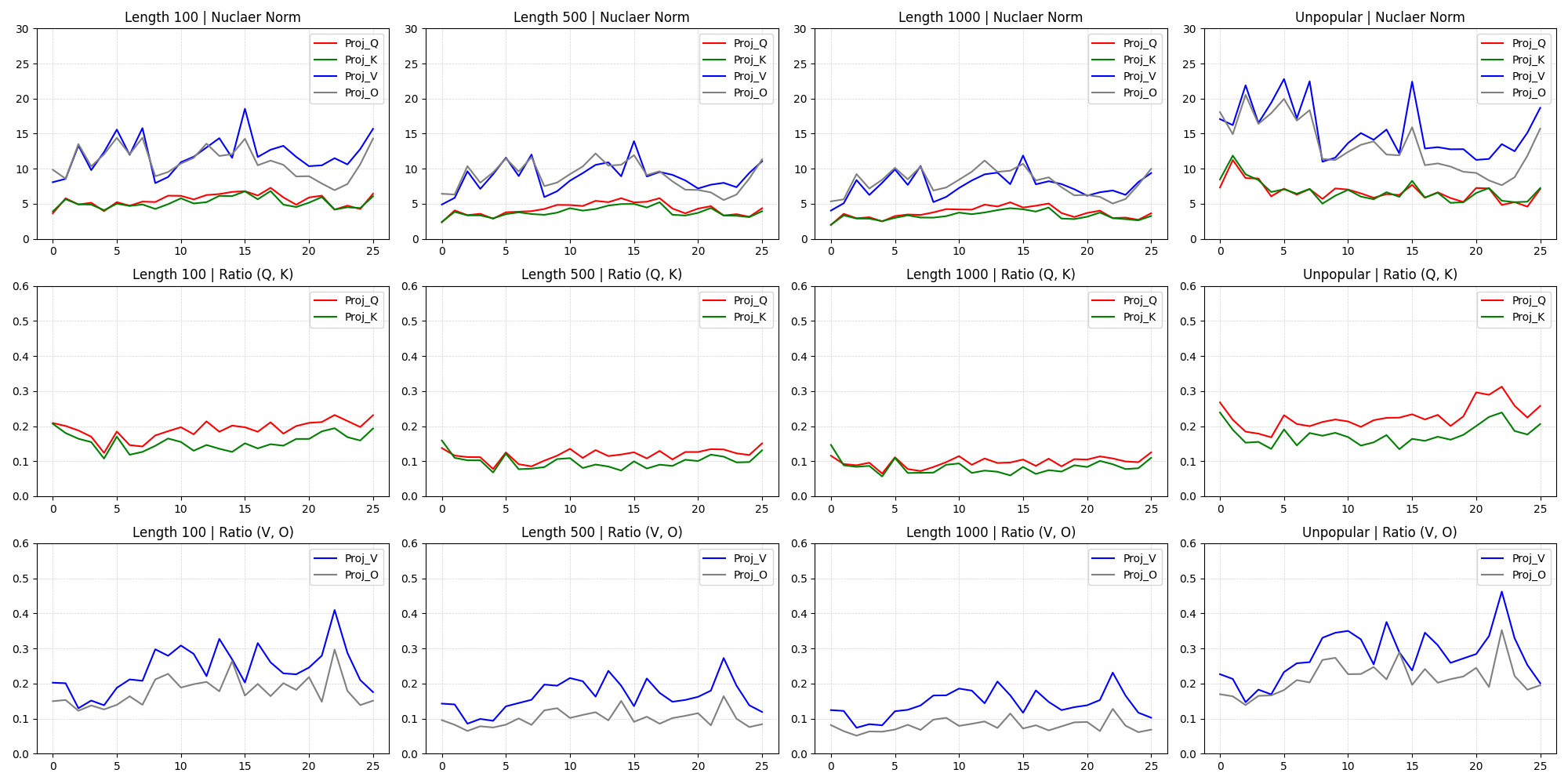}
\caption{Visualization for Wiki tasks using gemma-2-2b on correct responses.}
\label{fig:wiki_correct}
\vspace{-2mm}
\end{figure*}

\subsection{Pre-trained LLM on Irrelevant Responses}

\subsubsection{Reasoning Tasks}

The visualizations and statistical results on MATH tasks: MATH-Algebra (Figure \ref{fig:math_nonsense_algebra}, Table \ref{tbl:math_lighteval_math_wrong_answer_shuffle_gemma2_2b_grads_algebra}), MATH-Counting (Figure \ref{fig:math_nonsense_counting}, Table \ref{tbl:math_lighteval_math_wrong_answer_shuffle_gemma2_2b_grads_counting_probability}), MATH-Geometry (Figure \ref{fig:math_nonsense_geometry}, Table \ref{tbl:math_lighteval_math_wrong_answer_shuffle_gemma2_2b_grads_geometry}).

\begin{table}[!tbh]
\centering
\small
\scalebox{0.75}{
\begin{tabular}{l|l|l|cccc}
\toprule
\multirow{2}{*}{\textbf{Dataset}} & \multirow{2}{*}{\textbf{Curve}} & \multirow{2}{*}{\textbf{Cot}} &  \multicolumn{4}{c}{\textbf{Mean Absolute Difference (MAD)}} \\
 & &  & Early & Middle & Last & All\\
\midrule
\multirow{16}{*}{Algebra} & \multirow{3}{*}{$s_Q$} 
  & Simplified & 0.93 & 0.69 & 0.63 & 0.71 \\
& & Detailed & 0.68 & 0.42 & 0.53 & 0.50 \\
\cmidrule{2-7}
& \multirow{3}{*}{$s_K$} 
  & Simplified & 0.77 & 0.54 & 0.67 & 0.60 \\
& & Detailed & 0.56 & 0.38 & 0.56 & 0.46 \\
\cmidrule{2-7}
& \multirow{3}{*}{$s_V$} 
  & Simplified & 3.40 & 2.78 & 0.92 & 2.63 \\
& & Detailed & 2.18 & 1.84 & 0.76 & 1.75 \\
\cmidrule{2-7}
& \multirow{3}{*}{$s_O$} 
  & Simplified & 3.09 & 1.70 & 1.33 & 1.97 \\
& & Detailed & 1.74 & 1.25 & 0.84 & 1.26 \\
\cmidrule{2-7}
 & \multirow{3}{*}{$r_Q$} 
  & Simplified & 0.03 & 0.03 & 0.02 & 0.03 \\
& & Detailed & 0.03 & 0.02 & 0.02 & 0.02 \\
\cmidrule{2-7}
& \multirow{3}{*}{$r_K$} 
  & Simplified & 0.03 & 0.02 & 0.03 & 0.02 \\
& & Detailed & 0.04 & 0.02 & 0.03 & 0.02 \\
\cmidrule{2-7}
 & \multirow{3}{*}{$r_V$} 
  & Simplified & 0.03 & 0.04 & 0.05 & 0.04 \\
& & Detailed & 0.03 & 0.03 & 0.04 & 0.03 \\
\cmidrule{2-7}
& \multirow{3}{*}{$r_O$} 
  & Simplified & 0.01 & 0.02 & 0.04 & 0.02 \\
& & Detailed & 0.01 & 0.01 & 0.03 & 0.02 \\
\bottomrule
\end{tabular}
}
\caption{
Statistical results for MATH-Algebra using gemma-2-2b on irrelevant responses.
}
\label{tbl:math_lighteval_math_wrong_answer_shuffle_gemma2_2b_grads_algebra}
\end{table}

\begin{figure}[!h]
\centering
\includegraphics[width=\linewidth]{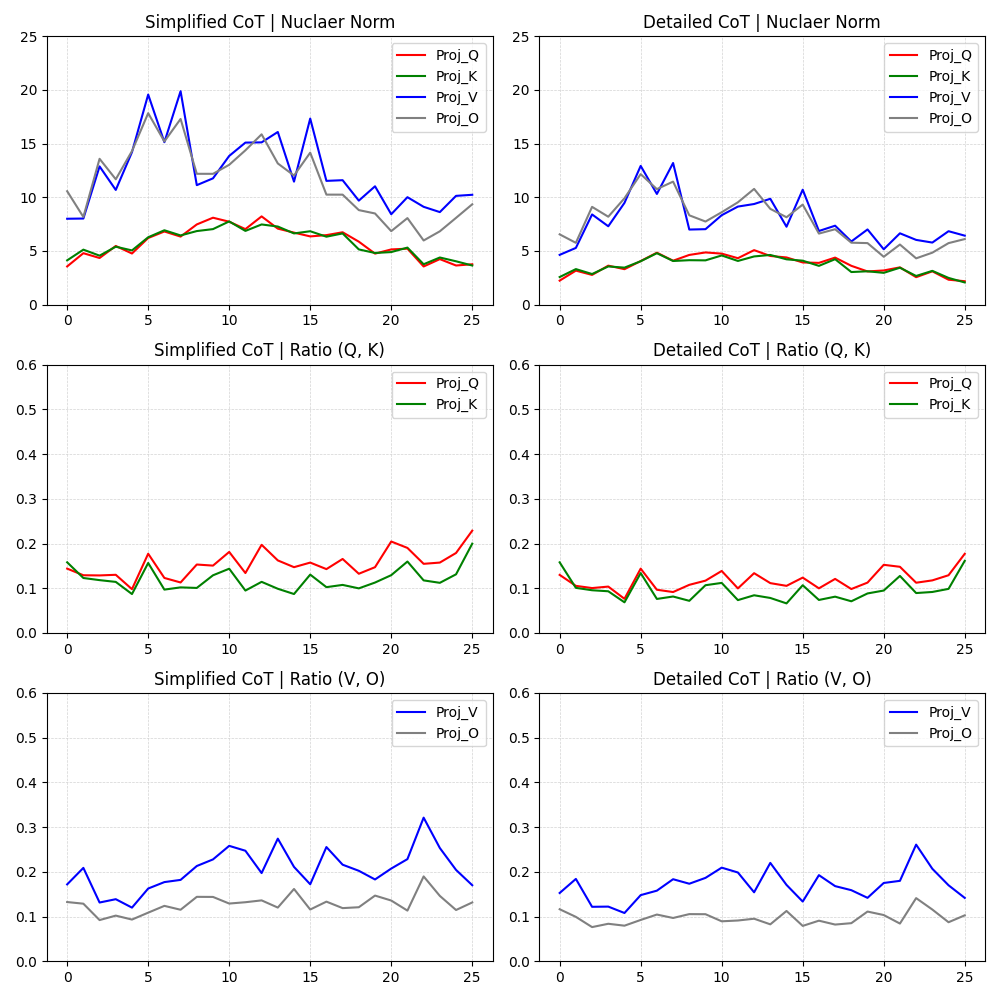}
\caption{Visualization for MATH-Algebra using gemma-2-2b on irrelevant responses.}
\label{fig:math_nonsense_algebra}
\vspace{-2mm}
\end{figure}

\begin{table}[!tbh]
\centering
\small
\scalebox{0.75}{
\begin{tabular}{l|l|l|cccc}
\toprule
\multirow{2}{*}{\textbf{Dataset}} & \multirow{2}{*}{\textbf{Curve}} & \multirow{2}{*}{\textbf{Cot}} &  \multicolumn{4}{c}{\textbf{Mean Absolute Difference (MAD)}} \\
 & &  & Early & Middle & Last & All\\
\midrule
\multirow{16}{*}{Counting} & \multirow{3}{*}{$s_Q$} 
  & Simplified & 0.96 & 0.77 & 0.78 & 0.80 \\
& & Detailed & 0.90 & 0.55 & 0.53 & 0.63 \\
\cmidrule{2-7}
& \multirow{3}{*}{$s_K$} 
  & Simplified & 0.74 & 0.63 & 0.58 & 0.63 \\
& & Detailed & 0.73 & 0.45 & 0.49 & 0.54 \\
\cmidrule{2-7}
& \multirow{3}{*}{$s_V$} 
  & Simplified & 3.60 & 2.74 & 1.05 & 2.67 \\
& & Detailed & 2.73 & 1.99 & 0.98 & 2.01 \\
\cmidrule{2-7}
& \multirow{3}{*}{$s_O$} 
  & Simplified & 3.25 & 1.75 & 1.34 & 2.02 \\
& & Detailed & 2.11 & 1.40 & 0.89 & 1.43 \\
\cmidrule{2-7}
 & \multirow{3}{*}{$r_Q$} 
  & Simplified & 0.03 & 0.03 & 0.02 & 0.03 \\
& & Detailed & 0.03 & 0.03 & 0.02 & 0.03 \\
\cmidrule{2-7}
& \multirow{3}{*}{$r_K$} 
  & Simplified & 0.04 & 0.02 & 0.03 & 0.03 \\
& & Detailed & 0.04 & 0.02 & 0.03 & 0.03 \\
\cmidrule{2-7}
 & \multirow{3}{*}{$r_V$} 
  & Simplified & 0.03 & 0.04 & 0.05 & 0.04 \\
& & Detailed & 0.03 & 0.03 & 0.04 & 0.03 \\
\cmidrule{2-7}
& \multirow{3}{*}{$r_O$} 
  & Simplified & 0.01 & 0.02 & 0.03 & 0.02 \\
& & Detailed & 0.01 & 0.01 & 0.02 & 0.02 \\
\bottomrule
\end{tabular}
}
\caption{
Statistical results for MATH-Counting using gemma-2-2b on irrelevant responses.
}
\label{tbl:math_lighteval_math_wrong_answer_shuffle_gemma2_2b_grads_counting_probability}
\end{table}

\begin{figure}[!h]
\centering
\includegraphics[width=\linewidth]{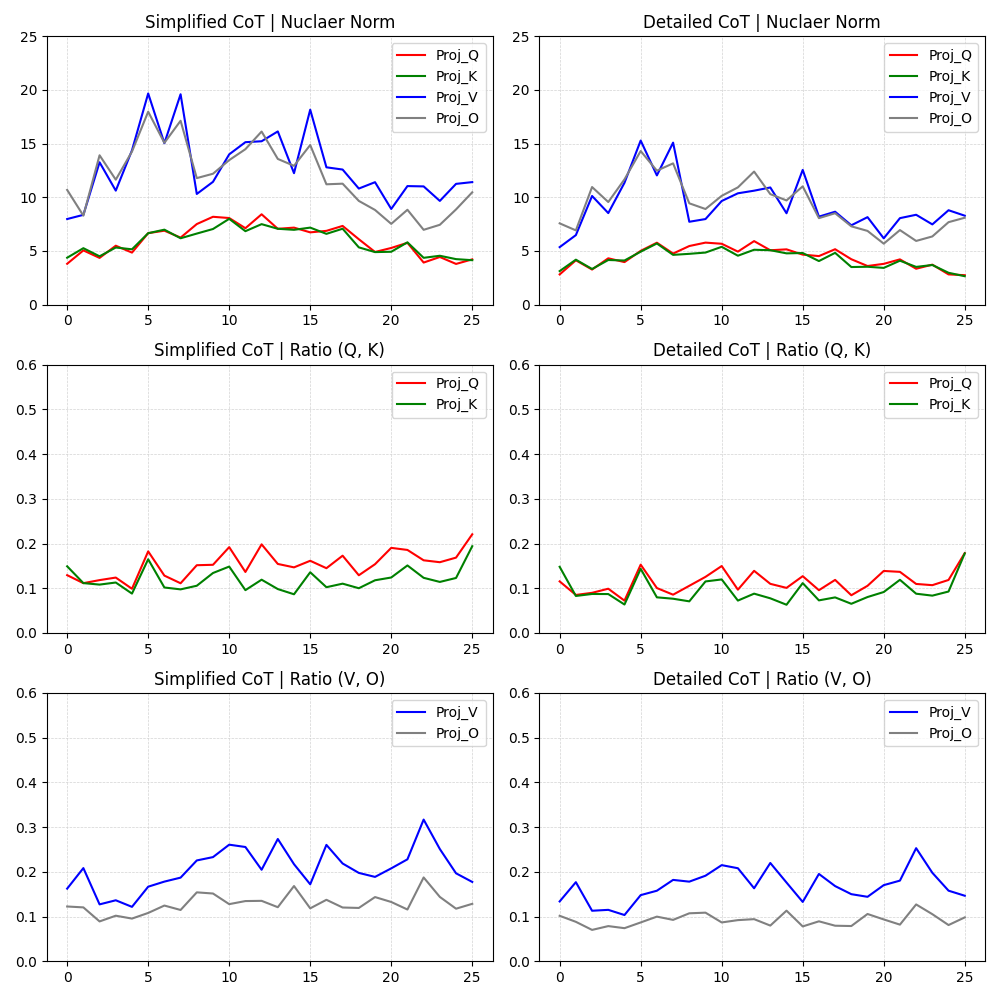}
\caption{Visualization for MATH-Counting using gemma-2-2b on irrelevant responses.}
\label{fig:math_nonsense_counting}
\vspace{-2mm}
\end{figure}

\begin{table}[!tbh]
\centering
\small
\scalebox{0.75}{
\begin{tabular}{l|l|l|cccc}
\toprule
\multirow{2}{*}{\textbf{Dataset}} & \multirow{2}{*}{\textbf{Curve}} & \multirow{2}{*}{\textbf{Cot}} &  \multicolumn{4}{c}{\textbf{Mean Absolute Difference (MAD)}} \\
 & &  & Early & Middle & Last & All\\
\midrule
\multirow{16}{*}{Geometry} & \multirow{3}{*}{$s_Q$} 
  & Simplified & 0.88 & 0.70 & 0.67 & 0.73 \\
& & Detailed & 0.90 & 0.39 & 0.75 & 0.61 \\
\cmidrule{2-7}
& \multirow{3}{*}{$s_K$} 
  & Simplified & 0.74 & 0.50 & 0.55 & 0.57 \\
& & Detailed & 0.76 & 0.52 & 0.75 & 0.64 \\
\cmidrule{2-7}
& \multirow{3}{*}{$s_V$} 
  & Simplified & 2.95 & 2.44 & 0.55 & 2.23 \\
& & Detailed & 2.78 & 2.32 & 0.84 & 2.14 \\
\cmidrule{2-7}
& \multirow{3}{*}{$s_O$} 
  & Simplified & 2.61 & 1.59 & 1.03 & 1.71 \\
& & Detailed & 2.10 & 1.54 & 1.08 & 1.53 \\
\cmidrule{2-7}
 & \multirow{3}{*}{$r_Q$} 
  & Simplified & 0.03 & 0.03 & 0.02 & 0.03 \\
& & Detailed & 0.03 & 0.02 & 0.04 & 0.03 \\
\cmidrule{2-7}
& \multirow{3}{*}{$r_K$} 
  & Simplified & 0.03 & 0.02 & 0.03 & 0.02 \\
& & Detailed & 0.04 & 0.02 & 0.04 & 0.03 \\
\cmidrule{2-7}
 & \multirow{3}{*}{$r_V$} 
  & Simplified & 0.03 & 0.04 & 0.05 & 0.03 \\
& & Detailed & 0.03 & 0.03 & 0.04 & 0.03 \\
\cmidrule{2-7}
& \multirow{3}{*}{$r_O$} 
  & Simplified & 0.01 & 0.02 & 0.03 & 0.02 \\
& & Detailed & 0.01 & 0.02 & 0.03 & 0.02 \\
\bottomrule
\end{tabular}
}
\caption{
Statistical results for MATH-Geometry using gemma-2-2b on irrelevant responses.
}
\label{tbl:math_lighteval_math_wrong_answer_shuffle_gemma2_2b_grads_geometry}
\end{table}

\begin{figure}[!h]
\centering
\includegraphics[width=\linewidth]{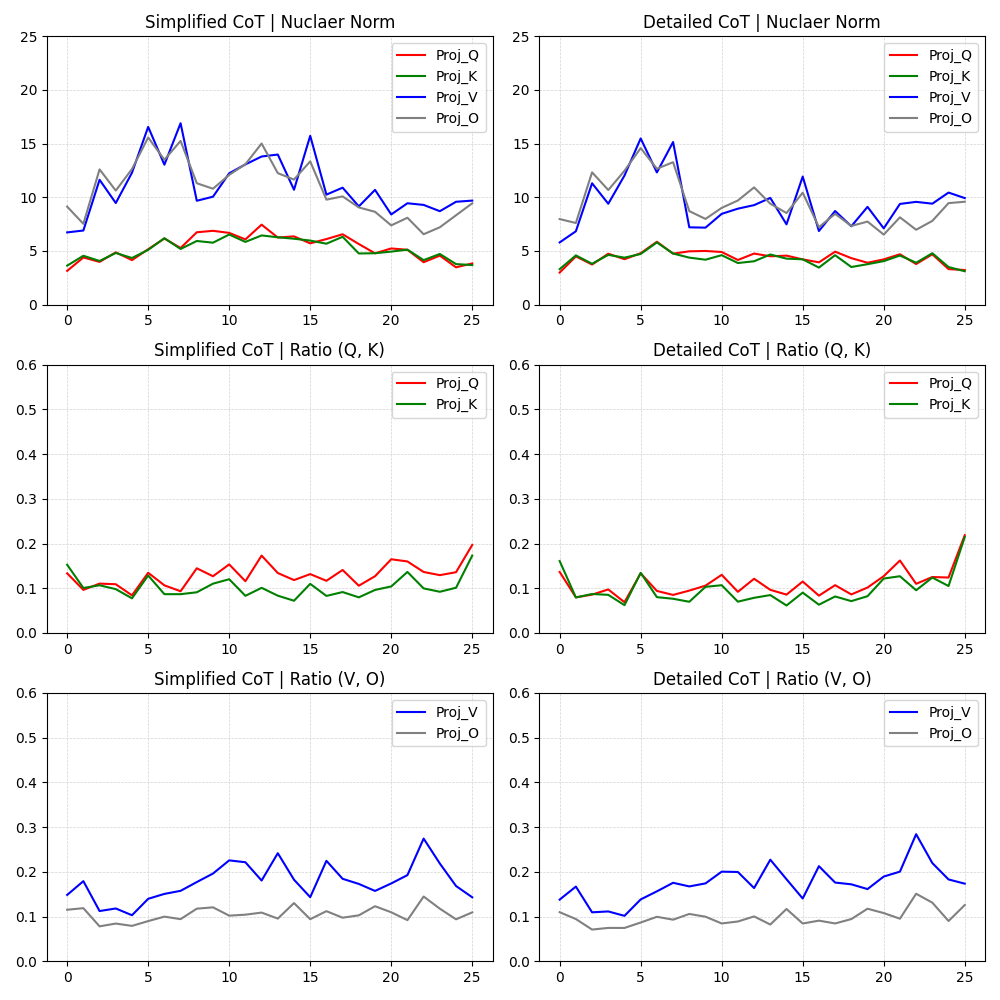}
\caption{Visualization for MATH-Geometry using gemma-2-2b on irrelevant responses.}
\label{fig:math_nonsense_geometry}
\vspace{-2mm}
\end{figure}

The visualizations and statistical results on other reasoning tasks: AQuA (Figure \ref{fig:reasoning_nonsense_aqua}, Table \ref{tbl:reasoning_cot_500_wrong_answer_shuffle_gemma2_2b_grads_aqua_train}), 
StrategyQA (Figure \ref{fig:reasoning_nonsense_strategyqa}, Table \ref{tbl:reasoning_cot_500_wrong_answer_shuffle_gemma2_2b_grads_strategyqa_train}), ECQA (Figure \ref{fig:reasoning_nonsense_ecqa}, Table \ref{tbl:reasoning_cot_500_wrong_answer_shuffle_gemma2_2b_grads_ecqa_train}), CREAK (Figure \ref{fig:reasoning_nonsense_creak}, Table \ref{tbl:reasoning_cot_500_wrong_answer_shuffle_gemma2_2b_grads_creak_train}), Sensemaking (Figure \ref{fig:reasoning_nonsense_sensemaking}, Table \ref{tbl:reasoning_cot_500_wrong_answer_shuffle_gemma2_2b_grads_sensemaking_train}).

\begin{table}[!tbh]
\centering
\small
\scalebox{0.75}{
\begin{tabular}{l|l|l|cccc}
\toprule
\multirow{2}{*}{\textbf{Dataset}} & \multirow{2}{*}{\textbf{Curve}} & \multirow{2}{*}{\textbf{Cot}} &  \multicolumn{4}{c}{\textbf{Mean Absolute Difference (MAD)}} \\
 & &  & Early & Middle & Last & All\\
\midrule
\multirow{24}{*}{AQuA} & \multirow{3}{*}{$s_Q$} 
  & None & 3.41 & 2.11 & 3.47 & 2.83 \\
& & Simplified & 1.65 & 1.06 & 0.90 & 1.17 \\
& & Detailed & 0.95 & 0.53 & 0.69 & 0.65 \\
\cmidrule{2-7}
& \multirow{3}{*}{$s_K$} 
  & None & 4.80 & 2.75 & 3.41 & 3.44 \\
& & Simplified & 1.77 & 1.11 & 0.88 & 1.22 \\
& & Detailed & 0.85 & 0.72 & 0.72 & 0.72 \\
\cmidrule{2-7}
& \multirow{3}{*}{$s_V$} 
  & None & 9.50 & 9.88 & 2.35 & 8.45 \\
& & Simplified & 4.19 & 3.19 & 1.00 & 3.07 \\
& & Detailed & 2.93 & 2.42 & 1.13 & 2.33 \\
\cmidrule{2-7}
& \multirow{3}{*}{$s_O$} 
  & None & 10.25 & 4.43 & 1.92 & 5.27 \\
& & Simplified & 3.87 & 2.19 & 0.94 & 2.26 \\
& & Detailed & 2.46 & 1.58 & 1.14 & 1.68 \\
\cmidrule{2-7}
 & \multirow{3}{*}{$r_Q$} 
  & None & 0.03 & 0.09 & 0.10 & 0.08 \\
& & Simplified & 0.02 & 0.03 & 0.02 & 0.02 \\
& & Detailed & 0.03 & 0.03 & 0.02 & 0.03 \\
\cmidrule{2-7}
& \multirow{3}{*}{$r_K$} 
  & None & 0.04 & 0.05 & 0.09 & 0.05 \\
& & Simplified & 0.03 & 0.02 & 0.03 & 0.02 \\
& & Detailed & 0.03 & 0.02 & 0.03 & 0.03 \\
\cmidrule{2-7}
 & \multirow{3}{*}{$r_V$} 
  & None & 0.04 & 0.07 & 0.03 & 0.05 \\
& & Simplified & 0.03 & 0.04 & 0.05 & 0.04 \\
& & Detailed & 0.03 & 0.03 & 0.04 & 0.03 \\
\cmidrule{2-7}
& \multirow{3}{*}{$r_O$} 
  & None & 0.04 & 0.05 & 0.07 & 0.05 \\
& & Simplified & 0.01 & 0.03 & 0.04 & 0.03 \\
& & Detailed & 0.01 & 0.02 & 0.03 & 0.02 \\
\bottomrule
\end{tabular}
}
\caption{
Statistical results for AQuA using gemma-2-2b on irrelevant responses.
}
\label{tbl:reasoning_cot_500_wrong_answer_shuffle_gemma2_2b_grads_aqua_train}
\end{table}

\begin{figure*}[!h]
\centering
\includegraphics[width=\linewidth]{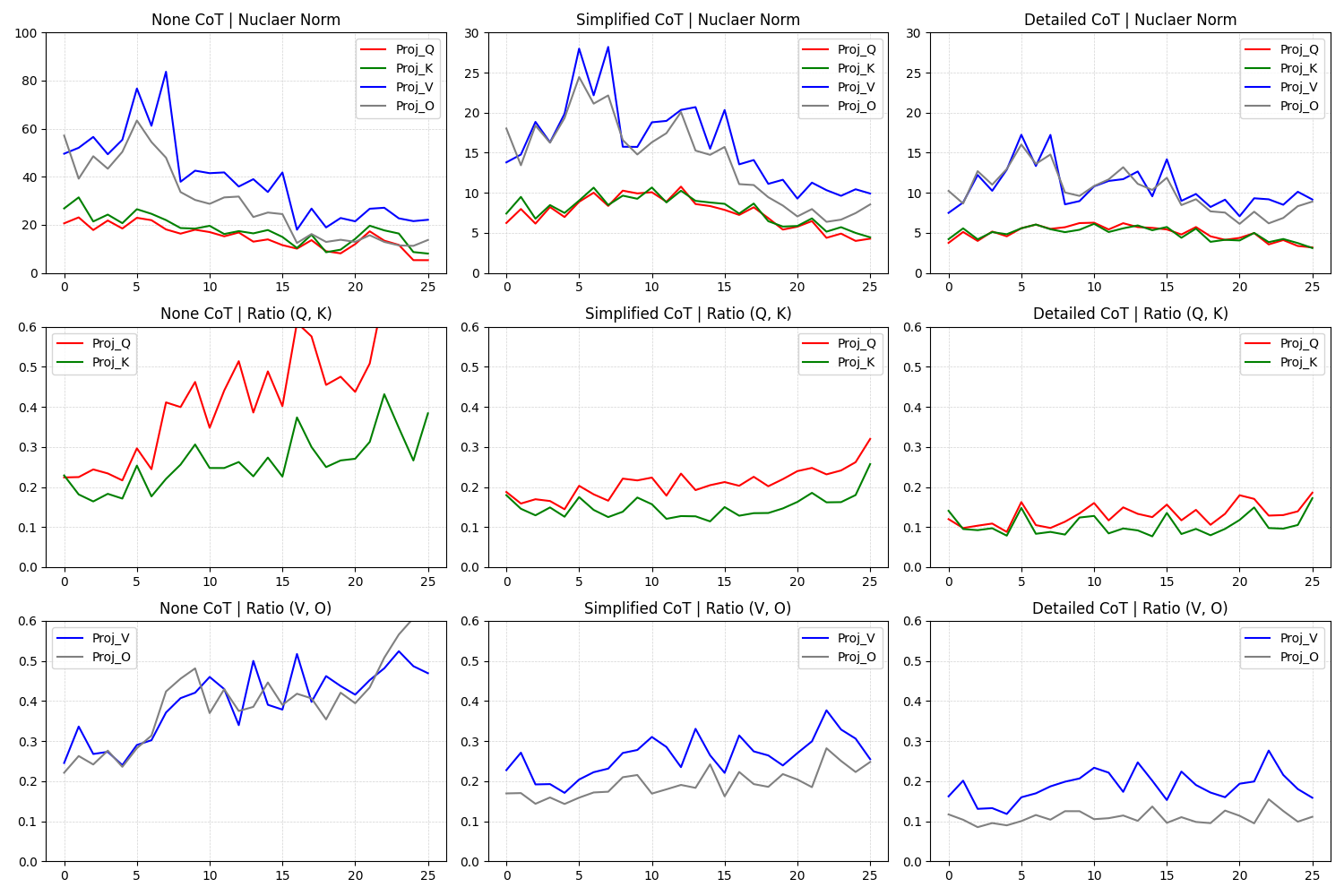}
\caption{Visualization for AQuA using gemma-2-2b on irrelevant responses.}
\label{fig:reasoning_nonsense_aqua}
\vspace{-2mm}
\end{figure*}



\begin{table}[!tbh]
\centering
\small
\scalebox{0.75}{
\begin{tabular}{l|l|l|cccc}
\toprule
\multirow{2}{*}{\textbf{Dataset}} & \multirow{2}{*}{\textbf{Curve}} & \multirow{2}{*}{\textbf{Cot}} &  \multicolumn{4}{c}{\textbf{Mean Absolute Difference (MAD)}} \\
 & &  & Early & Middle & Last & All\\
\midrule
\multirow{24}{*}{StrategyQA} & \multirow{3}{*}{$s_Q$} 
  & None & 111371.56 & 29719.68 & 9.10 & 48455.22 \\
& & Simplified & 2.75 & 1.19 & 1.52 & 1.67 \\
& & Detailed & 1.54 & 0.70 & 0.85 & 0.91 \\
\cmidrule{2-7}
& \multirow{3}{*}{$s_K$} 
  & None & 131198.87 & 34747.63 & 13.72 & 53897.70 \\
& & Simplified & 2.24 & 1.65 & 1.48 & 1.72 \\
& & Detailed & 1.18 & 0.83 & 0.68 & 0.84 \\
\cmidrule{2-7}
& \multirow{3}{*}{$s_V$} 
  & None & 683220.41 & 208522.69 & 69.15 & 289494.88 \\
& & Simplified & 6.23 & 4.76 & 5.15 & 5.23 \\
& & Detailed & 4.16 & 2.68 & 1.97 & 2.95 \\
\cmidrule{2-7}
& \multirow{3}{*}{$s_O$} 
  & None & 311685.32 & 92564.91 & 17.60 & 124625.79 \\
& & Simplified & 5.32 & 2.29 & 2.00 & 2.98 \\
& & Detailed & 3.14 & 1.84 & 1.54 & 2.07 \\
\cmidrule{2-7}
 & \multirow{3}{*}{$r_Q$} 
  & None & 0.02 & 0.06 & 0.08 & 0.06 \\
& & Simplified & 0.03 & 0.03 & 0.03 & 0.03 \\
& & Detailed & 0.04 & 0.02 & 0.03 & 0.03 \\
\cmidrule{2-7}
& \multirow{3}{*}{$r_K$} 
  & None & 0.02 & 0.03 & 0.05 & 0.03 \\
& & Simplified & 0.04 & 0.02 & 0.03 & 0.03 \\
& & Detailed & 0.05 & 0.02 & 0.03 & 0.03 \\
\cmidrule{2-7}
 & \multirow{3}{*}{$r_V$} 
  & None & 0.07 & 0.09 & 0.04 & 0.07 \\
& & Simplified & 0.04 & 0.06 & 0.06 & 0.05 \\
& & Detailed & 0.03 & 0.04 & 0.03 & 0.03 \\
\cmidrule{2-7}
& \multirow{3}{*}{$r_O$} 
  & None & 0.05 & 0.03 & 0.07 & 0.04 \\
& & Simplified & 0.02 & 0.03 & 0.06 & 0.03 \\
& & Detailed & 0.02 & 0.02 & 0.02 & 0.02 \\
\bottomrule
\end{tabular}
}
\caption{
Statistical results for StrategyQA using gemma-2-2b on irrelevant responses.
}
\label{tbl:reasoning_cot_500_wrong_answer_shuffle_gemma2_2b_grads_strategyqa_train}
\end{table}

\begin{figure*}[!h]
\centering
\includegraphics[width=\linewidth]{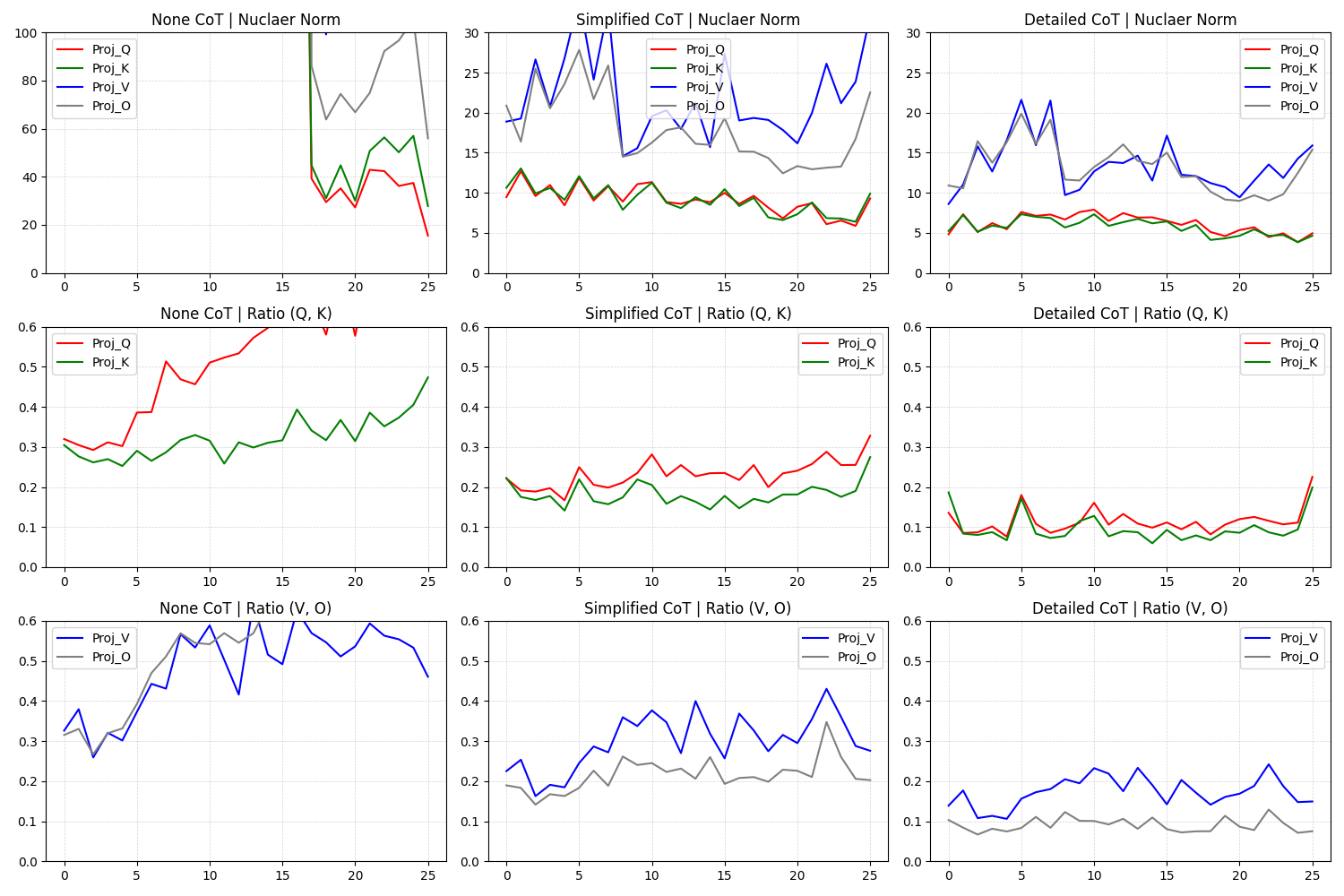}
\caption{Visualization for StrategyQA using gemma-2-2b on irrelevant responses.}
\label{fig:reasoning_nonsense_strategyqa}
\vspace{-2mm}
\end{figure*}

\begin{table}[!tbh]
\centering
\small
\scalebox{0.75}{
\begin{tabular}{l|l|l|cccc}
\toprule
\multirow{2}{*}{\textbf{Dataset}} & \multirow{2}{*}{\textbf{Curve}} & \multirow{2}{*}{\textbf{Cot}} &  \multicolumn{4}{c}{\textbf{Mean Absolute Difference (MAD)}} \\
 & &  & Early & Middle & Last & All\\
\midrule
\multirow{24}{*}{ECQA} & \multirow{3}{*}{$s_Q$} 
  & None & 14.09 & 5.80 & 9.27 & 8.31 \\
& & Simplified & 2.51 & 1.18 & 1.55 & 1.58 \\
& & Detailed & 1.29 & 0.81 & 0.71 & 0.89 \\
\cmidrule{2-7}
& \multirow{3}{*}{$s_K$} 
  & None & 13.60 & 10.84 & 12.96 & 11.64 \\
& & Simplified & 2.03 & 2.11 & 1.23 & 1.85 \\
& & Detailed & 1.02 & 0.98 & 0.67 & 0.89 \\
\cmidrule{2-7}
& \multirow{3}{*}{$s_V$} 
  & None & 40.86 & 44.16 & 41.30 & 43.71 \\
& & Simplified & 7.26 & 6.21 & 6.16 & 6.61 \\
& & Detailed & 4.19 & 2.79 & 2.39 & 3.11 \\
\cmidrule{2-7}
& \multirow{3}{*}{$s_O$} 
  & None & 38.03 & 16.51 & 4.59 & 18.87 \\
& & Simplified & 6.74 & 2.85 & 2.00 & 3.50 \\
& & Detailed & 3.32 & 1.98 & 1.17 & 2.12 \\
\cmidrule{2-7}
 & \multirow{3}{*}{$r_Q$} 
  & None & 0.02 & 0.05 & 0.08 & 0.05 \\
& & Simplified & 0.03 & 0.03 & 0.02 & 0.02 \\
& & Detailed & 0.03 & 0.03 & 0.03 & 0.03 \\
\cmidrule{2-7}
& \multirow{3}{*}{$r_K$} 
  & None & 0.03 & 0.03 & 0.07 & 0.04 \\
& & Simplified & 0.03 & 0.02 & 0.03 & 0.02 \\
& & Detailed & 0.04 & 0.03 & 0.03 & 0.03 \\
\cmidrule{2-7}
 & \multirow{3}{*}{$r_V$} 
  & None & 0.05 & 0.07 & 0.06 & 0.06 \\
& & Simplified & 0.05 & 0.06 & 0.04 & 0.05 \\
& & Detailed & 0.04 & 0.04 & 0.04 & 0.03 \\
\cmidrule{2-7}
& \multirow{3}{*}{$r_O$} 
  & None & 0.03 & 0.04 & 0.06 & 0.04 \\
& & Simplified & 0.02 & 0.03 & 0.05 & 0.03 \\
& & Detailed & 0.01 & 0.02 & 0.02 & 0.02 \\
\bottomrule
\end{tabular}
}
\caption{
Statistical results for ECQA using gemma-2-2b on irrelevant responses.
}
\label{tbl:reasoning_cot_500_wrong_answer_shuffle_gemma2_2b_grads_ecqa_train}
\end{table}

\begin{figure*}[!h]
\centering
\includegraphics[width=\linewidth]{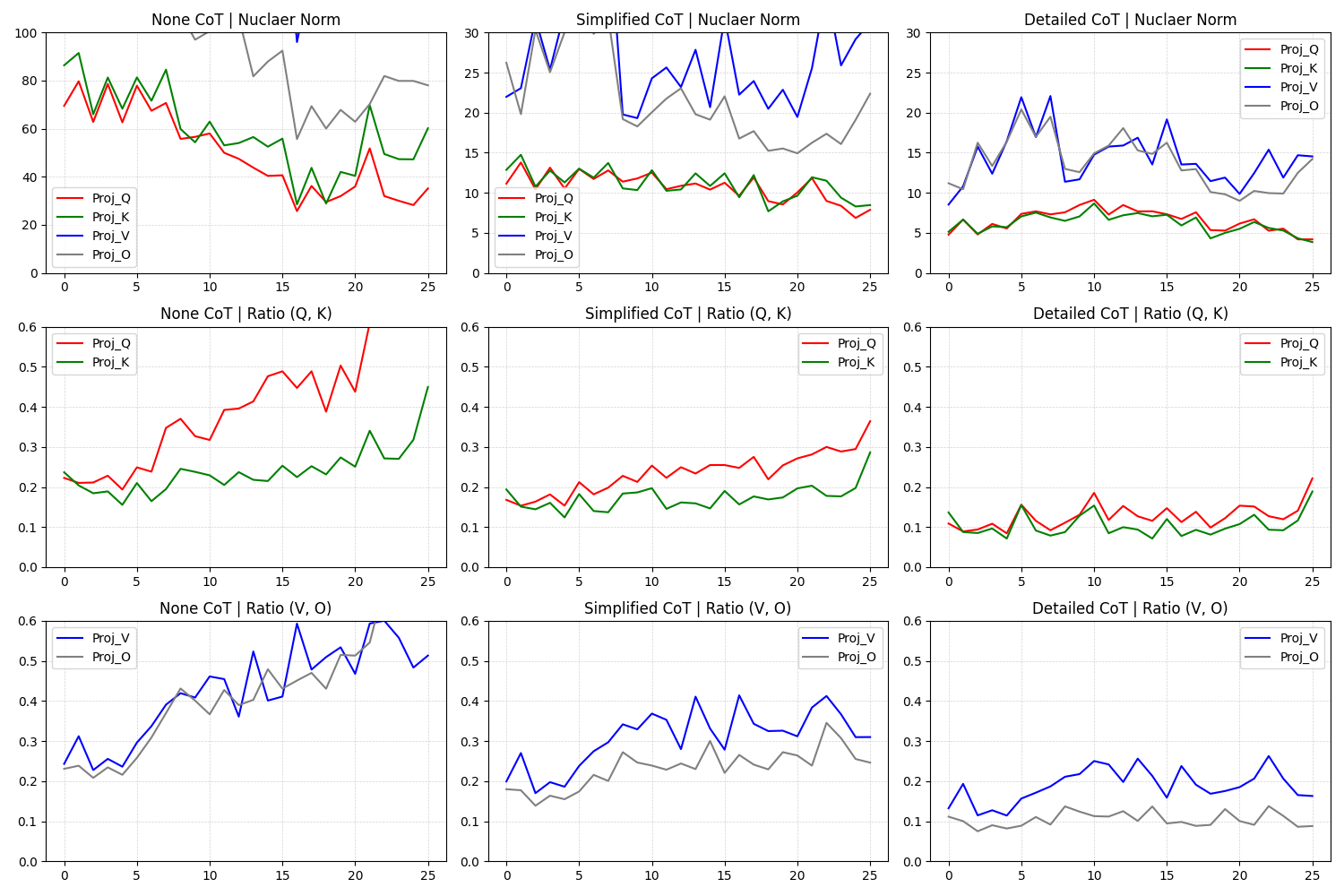}
\caption{Visualization for ECQA using gemma-2-2b on irrelevant responses.}
\label{fig:reasoning_nonsense_ecqa}
\vspace{-2mm}
\end{figure*}

\begin{table}[!tbh]
\centering
\small
\scalebox{0.75}{
\begin{tabular}{l|l|l|cccc}
\toprule
\multirow{2}{*}{\textbf{Dataset}} & \multirow{2}{*}{\textbf{Curve}} & \multirow{2}{*}{\textbf{Cot}} &  \multicolumn{4}{c}{\textbf{Mean Absolute Difference (MAD)}} \\
 & &  & Early & Middle & Last & All\\
\midrule
\multirow{24}{*}{CREAK} & \multirow{3}{*}{$s_Q$} 
  & None & 16.17 & 7.68 & 10.07 & 10.13 \\
& & Simplified & 2.55 & 1.20 & 1.40 & 1.59 \\
& & Detailed & 1.43 & 0.76 & 0.84 & 0.92 \\
\cmidrule{2-7}
& \multirow{3}{*}{$s_K$} 
  & None & 18.79 & 11.05 & 12.76 & 13.34 \\
& & Simplified & 2.19 & 1.97 & 1.18 & 1.81 \\
& & Detailed & 1.16 & 0.98 & 0.58 & 0.90 \\
\cmidrule{2-7}
& \multirow{3}{*}{$s_V$} 
  & None & 41.27 & 41.45 & 15.38 & 37.26 \\
& & Simplified & 7.35 & 6.05 & 5.39 & 6.39 \\
& & Detailed & 4.24 & 2.80 & 2.02 & 3.03 \\
\cmidrule{2-7}
& \multirow{3}{*}{$s_O$} 
  & None & 41.08 & 15.00 & 4.02 & 18.42 \\
& & Simplified & 6.97 & 2.68 & 2.17 & 3.55 \\
& & Detailed & 3.35 & 1.95 & 1.37 & 2.13 \\
\cmidrule{2-7}
 & \multirow{3}{*}{$r_Q$} 
  & None & 0.02 & 0.08 & 0.11 & 0.08 \\
& & Simplified & 0.02 & 0.03 & 0.03 & 0.03 \\
& & Detailed & 0.03 & 0.03 & 0.02 & 0.03 \\
\cmidrule{2-7}
& \multirow{3}{*}{$r_K$} 
  & None & 0.03 & 0.03 & 0.07 & 0.04 \\
& & Simplified & 0.03 & 0.02 & 0.03 & 0.02 \\
& & Detailed & 0.04 & 0.02 & 0.03 & 0.03 \\
\cmidrule{2-7}
 & \multirow{3}{*}{$r_V$} 
  & None & 0.05 & 0.07 & 0.03 & 0.05 \\
& & Simplified & 0.05 & 0.06 & 0.05 & 0.05 \\
& & Detailed & 0.03 & 0.04 & 0.04 & 0.03 \\
\cmidrule{2-7}
& \multirow{3}{*}{$r_O$} 
  & None & 0.04 & 0.05 & 0.10 & 0.06 \\
& & Simplified & 0.02 & 0.04 & 0.05 & 0.03 \\
& & Detailed & 0.02 & 0.02 & 0.03 & 0.02 \\
\bottomrule
\end{tabular}
}
\caption{
Statistical results for CREAK using gemma-2-2b on irrelevant responses.
}
\label{tbl:reasoning_cot_500_wrong_answer_shuffle_gemma2_2b_grads_creak_train}
\end{table}

\begin{figure*}[!h]
\centering
\includegraphics[width=\linewidth]{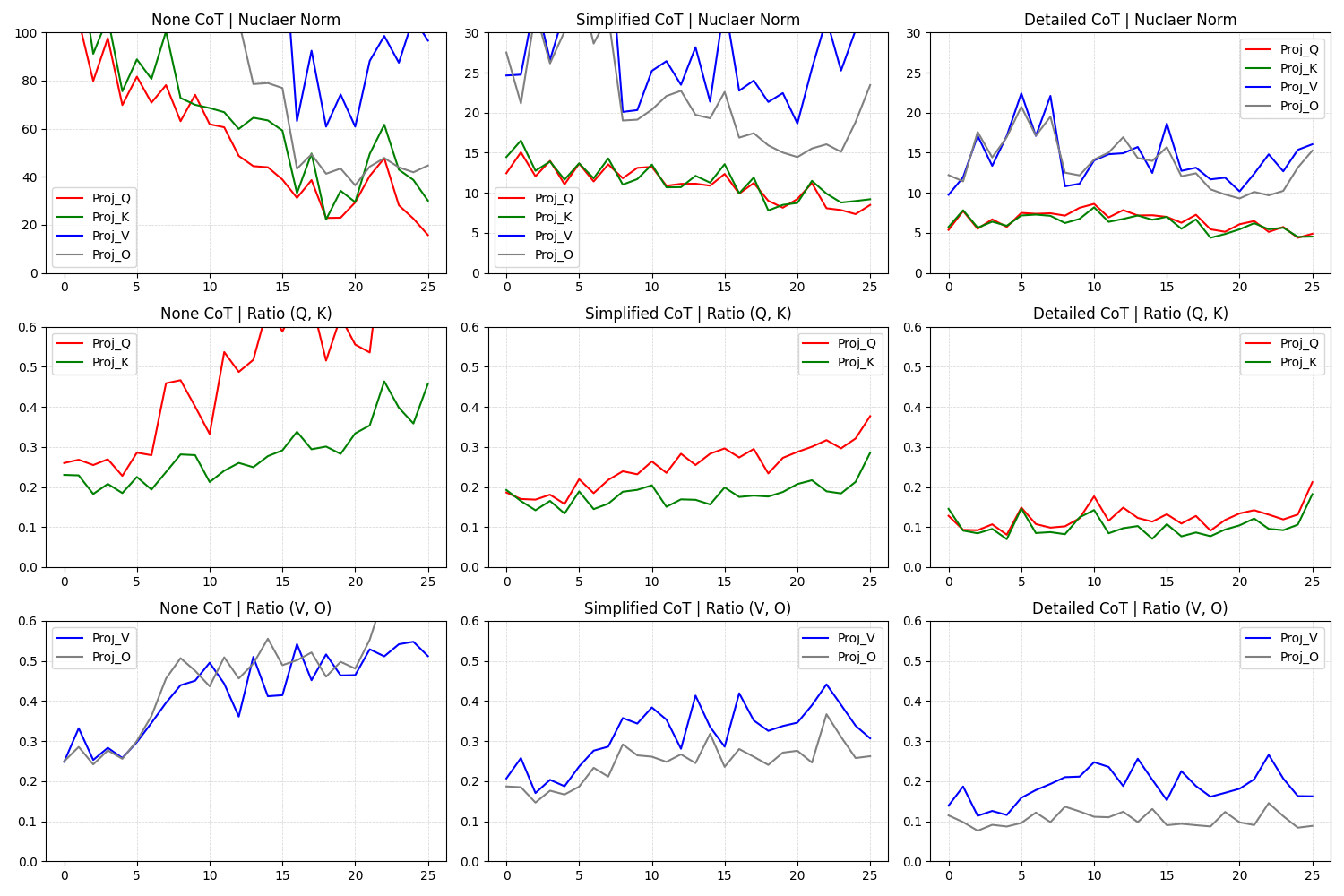}
\caption{Visualization for CREAK using gemma-2-2b on irrelevant responses.}
\label{fig:reasoning_nonsense_creak}
\vspace{-2mm}
\end{figure*}

\begin{table}[!tbh]
\centering
\small
\scalebox{0.75}{
\begin{tabular}{l|l|l|cccc}
\toprule
\multirow{2}{*}{\textbf{Dataset}} & \multirow{2}{*}{\textbf{Curve}} & \multirow{2}{*}{\textbf{Cot}} &  \multicolumn{4}{c}{\textbf{Mean Absolute Difference (MAD)}} \\
 & &  & Early & Middle & Last & All\\
\midrule
\multirow{24}{*}{Sensemaking} & \multirow{3}{*}{$s_Q$} 
  & None & 9.03 & 4.66 & 6.26 & 6.48 \\
& & Simplified & 2.64 & 1.72 & 2.34 & 2.06 \\
& & Detailed & 1.31 & 1.02 & 0.73 & 1.04 \\
\cmidrule{2-7}
& \multirow{3}{*}{$s_K$} 
  & None & 7.64 & 8.40 & 10.64 & 8.79 \\
& & Simplified & 1.96 & 2.21 & 2.12 & 2.09 \\
& & Detailed & 1.07 & 1.10 & 0.80 & 1.01 \\
\cmidrule{2-7}
& \multirow{3}{*}{$s_V$} 
  & None & 23.14 & 25.62 & 12.45 & 23.23 \\
& & Simplified & 8.12 & 7.29 & 5.94 & 7.49 \\
& & Detailed & 3.79 & 2.75 & 2.44 & 2.98 \\
\cmidrule{2-7}
& \multirow{3}{*}{$s_O$} 
  & None & 24.94 & 9.50 & 4.30 & 11.66 \\
& & Simplified & 7.80 & 3.16 & 2.30 & 4.04 \\
& & Detailed & 2.94 & 2.05 & 1.20 & 2.07 \\
\cmidrule{2-7}
 & \multirow{3}{*}{$r_Q$} 
  & None & 0.02 & 0.06 & 0.07 & 0.06 \\
& & Simplified & 0.03 & 0.04 & 0.08 & 0.05 \\
& & Detailed & 0.03 & 0.04 & 0.03 & 0.04 \\
\cmidrule{2-7}
& \multirow{3}{*}{$r_K$} 
  & None & 0.01 & 0.04 & 0.09 & 0.04 \\
& & Simplified & 0.03 & 0.02 & 0.06 & 0.03 \\
& & Detailed & 0.04 & 0.03 & 0.04 & 0.03 \\
\cmidrule{2-7}
 & \multirow{3}{*}{$r_V$} 
  & None & 0.05 & 0.07 & 0.04 & 0.06 \\
& & Simplified & 0.05 & 0.05 & 0.04 & 0.04 \\
& & Detailed & 0.03 & 0.04 & 0.03 & 0.03 \\
\cmidrule{2-7}
& \multirow{3}{*}{$r_O$} 
  & None & 0.04 & 0.05 & 0.10 & 0.05 \\
& & Simplified & 0.03 & 0.03 & 0.05 & 0.03 \\
& & Detailed & 0.01 & 0.02 & 0.02 & 0.02 \\
\bottomrule
\end{tabular}
}
\caption{
Statistical results for Sensemaking using gemma-2-2b on irrelevant responses.
}
\label{tbl:reasoning_cot_500_wrong_answer_shuffle_gemma2_2b_grads_sensemaking_train}
\end{table}

\begin{figure*}[!h]
\centering
\includegraphics[width=\linewidth]{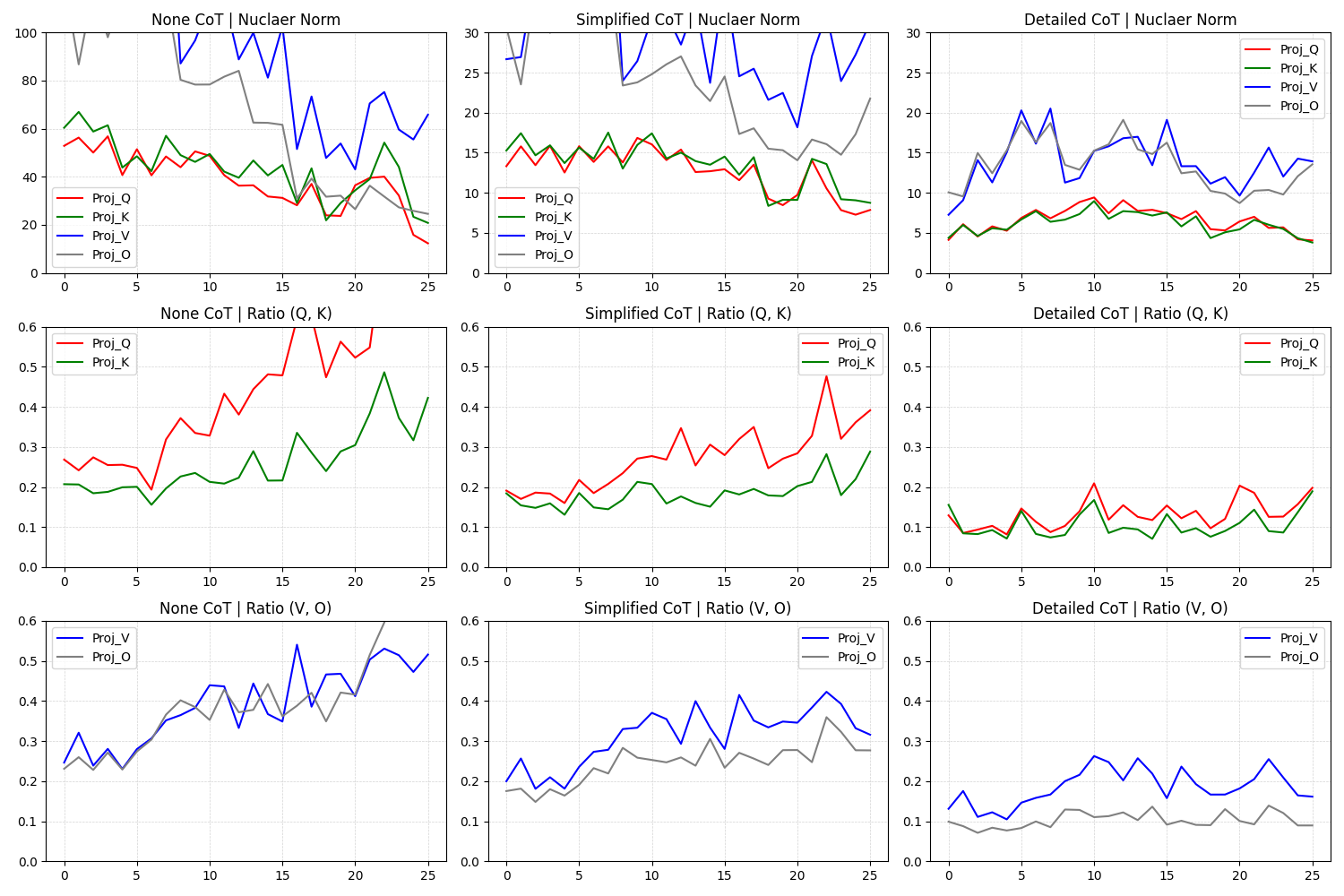}
\caption{Visualization for Sensemaking using gemma-2-2b on irrelevant responses.}
\label{fig:reasoning_nonsense_sensemaking}
\vspace{-2mm}
\end{figure*}

\subsubsection{Wiki Tasks}

The visualizations and statistical results on Wiki tasks are shown in Figure \ref{fig:wiki_nonsense} and Table \ref{tbl:wiki_wiki_popularity_wrong_answer_shuffle_gemma2_2b_grads}.

\begin{table}[!tbh]
\centering
\small
\scalebox{0.75}{
\begin{tabular}{l|l|l|cccc}
\toprule
\multirow{2}{*}{\textbf{Dataset}} & \multirow{2}{*}{\textbf{Curve}} & \multirow{2}{*}{\textbf{Cot}} &  \multicolumn{4}{c}{\textbf{Mean Absolute Difference (MAD)}} \\
 & &  & Early & Middle & Last & All\\
\midrule
\multirow{32}{*}{Wiki} & \multirow{3}{*}{$s_Q$} 
  & Len 100 & 2.37 & 1.03 & 1.23 & 1.42 \\
& & Len 500 & 1.63 & 0.66 & 0.91 & 0.96 \\
& & Len 1000 & 1.37 & 0.53 & 0.79 & 0.80 \\
& & Unpopular & 3.46 & 1.27 & 1.31 & 1.84 \\
\cmidrule{2-7}
& \multirow{3}{*}{$s_K$} 
  & Len 100 & 1.93 & 1.26 & 1.19 & 1.39 \\
& & Len 500 & 1.32 & 0.71 & 0.83 & 0.87 \\
& & Len 1000 & 1.11 & 0.55 & 0.73 & 0.72 \\
& & Unpopular & 2.92 & 1.75 & 1.20 & 1.89 \\
\cmidrule{2-7}
& \multirow{3}{*}{$s_V$} 
  & Len 100 & 4.86 & 4.01 & 2.83 & 3.95 \\
& & Len 500 & 3.76 & 2.93 & 1.94 & 2.93 \\
& & Len 1000 & 3.29 & 2.51 & 1.62 & 2.53 \\
& & Unpopular & 5.82 & 4.63 & 3.93 & 4.76 \\
\cmidrule{2-7}
& \multirow{3}{*}{$s_O$} 
  & Len 100 & 4.18 & 2.28 & 2.43 & 2.73 \\
& & Len 500 & 3.06 & 1.86 & 2.23 & 2.22 \\
& & Len 1000 & 2.70 & 1.59 & 2.07 & 1.97 \\
& & Unpopular & 5.14 & 2.49 & 3.05 & 3.15 \\
\cmidrule{2-7}
 & \multirow{3}{*}{$r_Q$} 
  & Len 100 & 0.04 & 0.03 & 0.02 & 0.03 \\
& & Len 500 & 0.04 & 0.02 & 0.03 & 0.03 \\
& & Len 1000 & 0.04 & 0.02 & 0.03 & 0.03 \\
& & Unpopular & 0.03 & 0.02 & 0.02 & 0.02 \\
\cmidrule{2-7}
& \multirow{3}{*}{$r_K$} 
  & Len 100 & 0.04 & 0.02 & 0.03 & 0.03 \\
& & Len 500 & 0.05 & 0.02 & 0.03 & 0.03 \\
& & Len 1000 & 0.05 & 0.01 & 0.03 & 0.03 \\
& & Unpopular & 0.04 & 0.02 & 0.02 & 0.02 \\
\cmidrule{2-7}
 & \multirow{3}{*}{$r_V$} 
  & Len 100 & 0.04 & 0.06 & 0.07 & 0.05 \\
& & Len 500 & 0.03 & 0.04 & 0.05 & 0.04 \\
& & Len 1000 & 0.02 & 0.03 & 0.04 & 0.03 \\
& & Unpopular & 0.04 & 0.06 & 0.08 & 0.05 \\
\cmidrule{2-7}
& \multirow{3}{*}{$r_O$} 
  & Len 100 & 0.02 & 0.03 & 0.07 & 0.03 \\
& & Len 500 & 0.02 & 0.02 & 0.04 & 0.02 \\
& & Len 1000 & 0.02 & 0.02 & 0.03 & 0.02 \\
& & Unpopular & 0.02 & 0.03 & 0.07 & 0.04 \\
\bottomrule
\end{tabular}
}
\caption{
Statistical results for Wiki using gemma-2-2b on irrelevant responses.
}
\label{tbl:wiki_wiki_popularity_wrong_answer_shuffle_gemma2_2b_grads}
\end{table}

\begin{figure*}[!h]
\centering
\includegraphics[width=\linewidth]{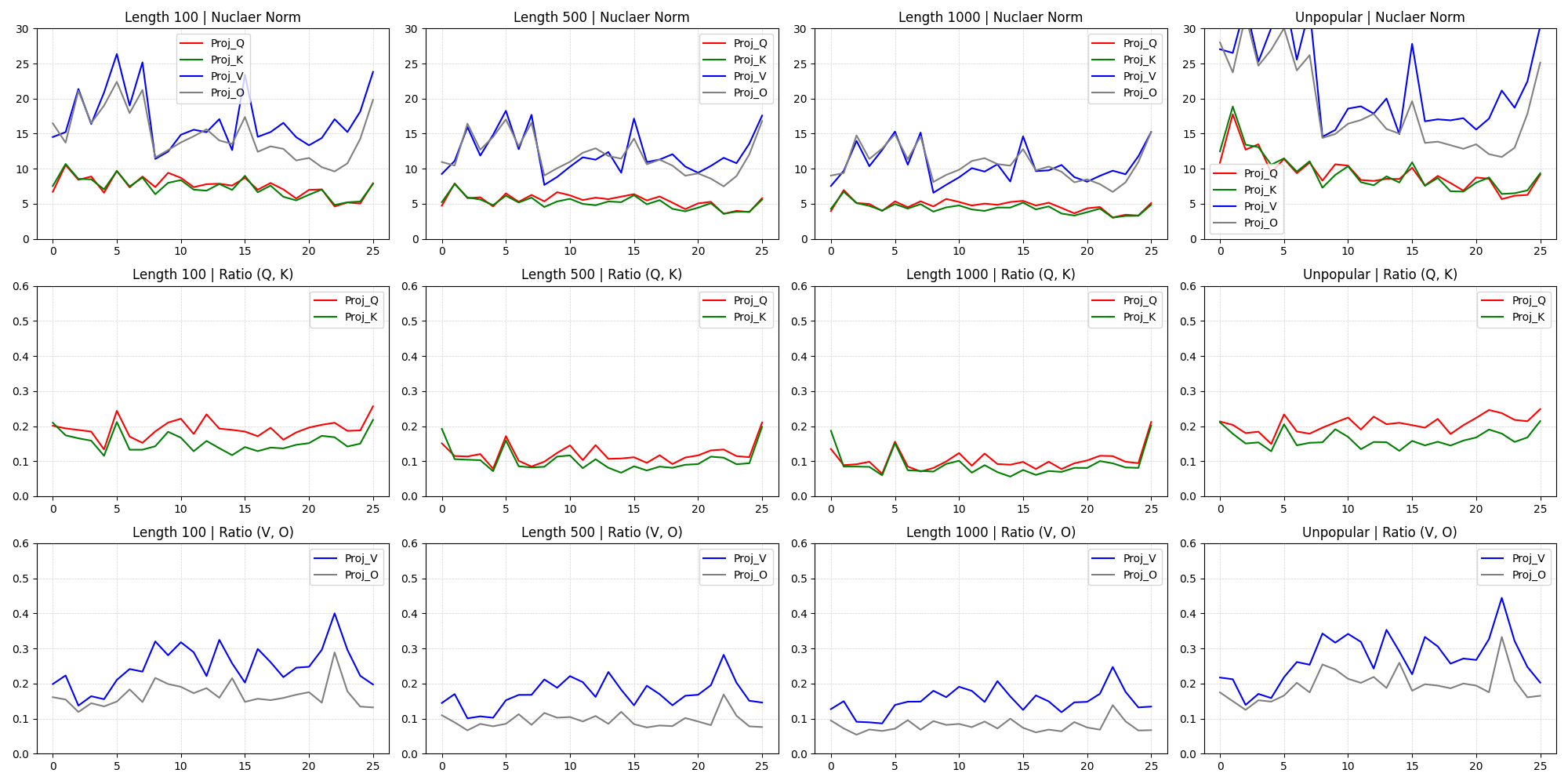}
\caption{Visualization for Wiki tasks using gemma-2-2b on irrelevant responses.}
\label{fig:wiki_nonsense}
\vspace{-2mm}
\end{figure*}

\clearpage

\subsection{Instructed LLM on Correct Responses}

\subsubsection{Reasoning Tasks}

The visualizations and statistical results on MATH tasks: MATH-Algebra (Figure \ref{fig:math_correct_algebra_it}, Table \ref{tbl:math_lighteval_math_gemma2_2b_it_grads_algebra}), MATH-Counting (Figure \ref{fig:math_correct_counting_it}, Table \ref{tbl:math_lighteval_math_gemma2_2b_it_grads_counting_probability}), MATH-Geometry (Figure \ref{fig:math_correct_geometry_it}, Table \ref{tbl:math_lighteval_math_gemma2_2b_it_grads_geometry}).

\begin{table}[!tbh]
\centering
\small
\scalebox{0.75}{
\begin{tabular}{l|l|l|cccc}
\toprule
\multirow{2}{*}{\textbf{Dataset}} & \multirow{2}{*}{\textbf{Curve}} & \multirow{2}{*}{\textbf{Cot}} &  \multicolumn{4}{c}{\textbf{Mean Absolute Difference (MAD)}} \\
 & &  & Early & Middle & Last & All\\
\midrule
\multirow{16}{*}{Algebra} & \multirow{3}{*}{$s_Q$} 
  & Simplified & 0.56 & 0.82 & 1.02 & 0.84 \\
& & Detailed & 0.36 & 0.53 & 0.73 & 0.57 \\
\cmidrule{2-7}
& \multirow{3}{*}{$s_K$} 
  & Simplified & 0.57 & 0.67 & 0.74 & 0.67 \\
& & Detailed & 0.35 & 0.52 & 0.59 & 0.51 \\
\cmidrule{2-7}
& \multirow{3}{*}{$s_V$} 
  & Simplified & 2.47 & 2.87 & 0.73 & 2.36 \\
& & Detailed & 1.35 & 1.70 & 0.65 & 1.40 \\
\cmidrule{2-7}
& \multirow{3}{*}{$s_O$} 
  & Simplified & 1.99 & 1.89 & 0.76 & 1.68 \\
& & Detailed & 1.01 & 1.17 & 0.43 & 0.97 \\
\cmidrule{2-7}
 & \multirow{3}{*}{$r_Q$} 
  & Simplified & 0.02 & 0.02 & 0.07 & 0.04 \\
& & Detailed & 0.02 & 0.03 & 0.08 & 0.05 \\
\cmidrule{2-7}
& \multirow{3}{*}{$r_K$} 
  & Simplified & 0.03 & 0.02 & 0.05 & 0.03 \\
& & Detailed & 0.04 & 0.03 & 0.05 & 0.04 \\
\cmidrule{2-7}
 & \multirow{3}{*}{$r_V$} 
  & Simplified & 0.03 & 0.05 & 0.05 & 0.04 \\
& & Detailed & 0.02 & 0.04 & 0.04 & 0.04 \\
\cmidrule{2-7}
& \multirow{3}{*}{$r_O$} 
  & Simplified & 0.01 & 0.02 & 0.04 & 0.02 \\
& & Detailed & 0.01 & 0.02 & 0.04 & 0.02 \\
\bottomrule
\end{tabular}
}
\caption{
Statistical results for MATH-Algebra using gemma-2-2b-it on correct responses.
}
\label{tbl:math_lighteval_math_gemma2_2b_it_grads_algebra}
\end{table}

\begin{figure}[!h]
\centering
\includegraphics[width=\linewidth]{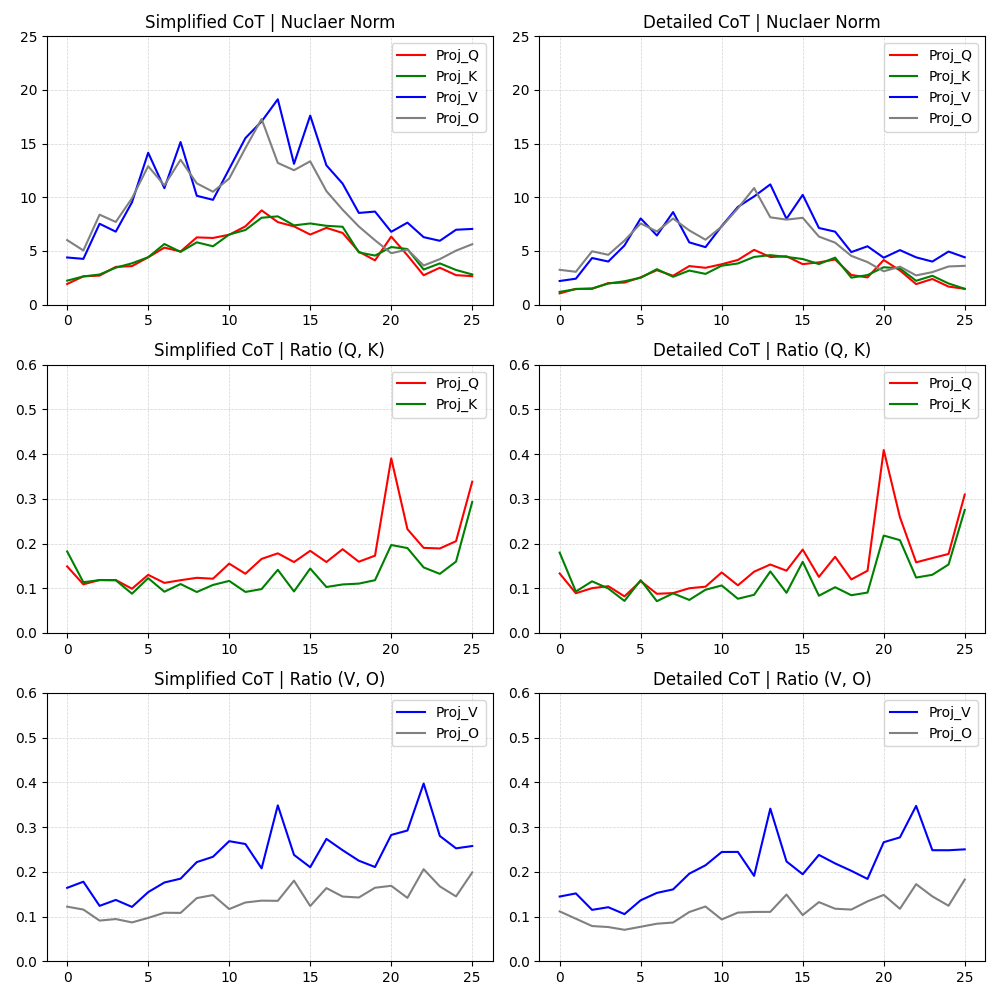}
\caption{Visualization for MATH-Algebra using gemma-2-2b-it on correct responses.}
\label{fig:math_correct_algebra_it}
\vspace{-2mm}
\end{figure}

\begin{table}[!tbh]
\centering
\small
\scalebox{0.75}{
\begin{tabular}{l|l|l|cccc}
\toprule
\multirow{2}{*}{\textbf{Dataset}} & \multirow{2}{*}{\textbf{Curve}} & \multirow{2}{*}{\textbf{Cot}} &  \multicolumn{4}{c}{\textbf{Mean Absolute Difference (MAD)}} \\
 & &  & Early & Middle & Last & All\\
\midrule
\multirow{16}{*}{Counting} & \multirow{3}{*}{$s_Q$} 
  & Simplified & 0.55 & 0.75 & 0.67 & 0.70 \\
& & Detailed & 0.40 & 0.62 & 0.55 & 0.57 \\
\cmidrule{2-7}
& \multirow{3}{*}{$s_K$} 
  & Simplified & 0.51 & 0.66 & 0.54 & 0.60 \\
& & Detailed & 0.36 & 0.55 & 0.42 & 0.49 \\
\cmidrule{2-7}
& \multirow{3}{*}{$s_V$} 
  & Simplified & 2.52 & 2.61 & 0.76 & 2.22 \\
& & Detailed & 1.52 & 1.72 & 0.67 & 1.45 \\
\cmidrule{2-7}
& \multirow{3}{*}{$s_O$} 
  & Simplified & 2.10 & 1.71 & 0.75 & 1.60 \\
& & Detailed & 1.17 & 1.20 & 0.44 & 1.02 \\
\cmidrule{2-7}
 & \multirow{3}{*}{$r_Q$} 
  & Simplified & 0.02 & 0.02 & 0.05 & 0.03 \\
& & Detailed & 0.02 & 0.03 & 0.07 & 0.04 \\
\cmidrule{2-7}
& \multirow{3}{*}{$r_K$} 
  & Simplified & 0.03 & 0.02 & 0.04 & 0.03 \\
& & Detailed & 0.04 & 0.03 & 0.05 & 0.04 \\
\cmidrule{2-7}
 & \multirow{3}{*}{$r_V$} 
  & Simplified & 0.02 & 0.05 & 0.05 & 0.04 \\
& & Detailed & 0.02 & 0.05 & 0.04 & 0.04 \\
\cmidrule{2-7}
& \multirow{3}{*}{$r_O$} 
  & Simplified & 0.01 & 0.03 & 0.04 & 0.02 \\
& & Detailed & 0.01 & 0.02 & 0.03 & 0.02 \\
\bottomrule
\end{tabular}
}
\caption{
Statistical results for MATH-Counting using gemma-2-2b-it on correct responses.
}
\label{tbl:math_lighteval_math_gemma2_2b_it_grads_counting_probability}
\end{table}

\begin{figure}[!h]
\centering
\includegraphics[width=\linewidth]{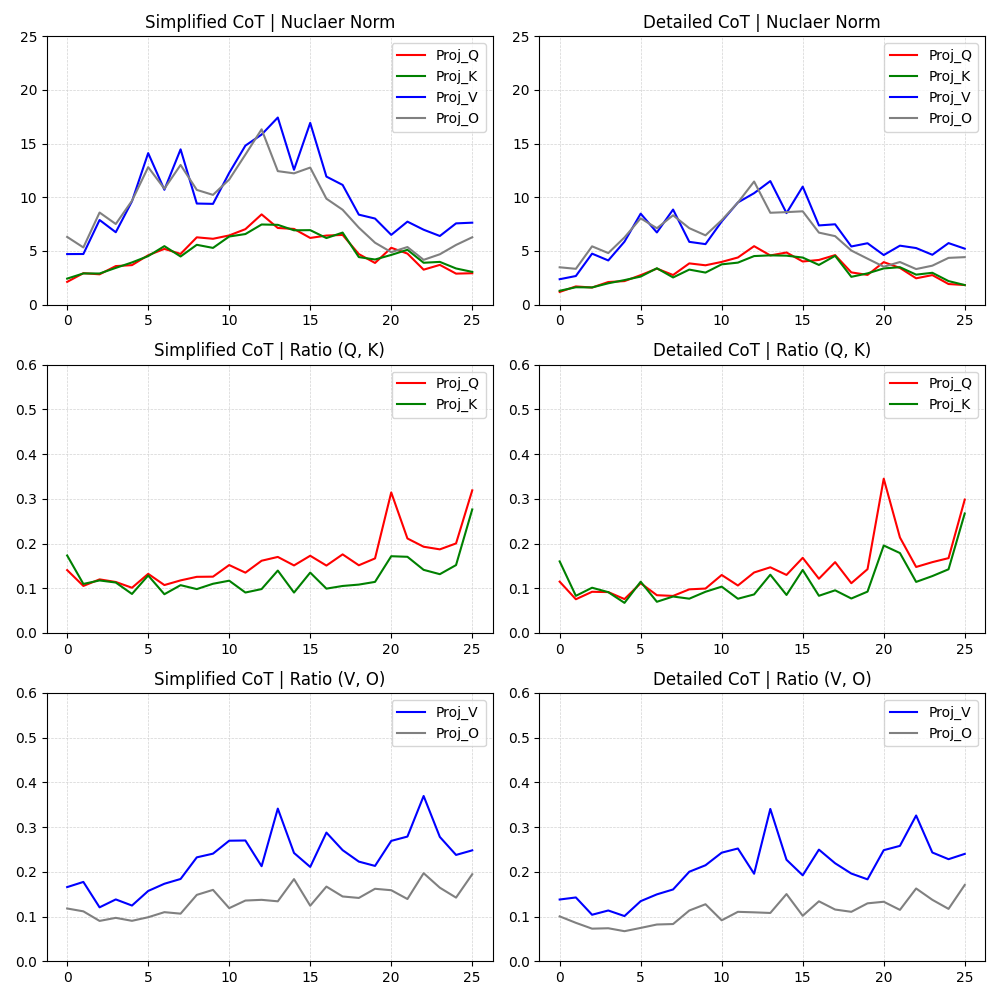}
\caption{Visualization for MATH-Counting using gemma-2-2b-it on correct responses.}
\label{fig:math_correct_counting_it}
\vspace{-2mm}
\end{figure}

\begin{table}[!tbh]
\centering
\small
\scalebox{0.75}{
\begin{tabular}{l|l|l|cccc}
\toprule
\multirow{2}{*}{\textbf{Dataset}} & \multirow{2}{*}{\textbf{Curve}} & \multirow{2}{*}{\textbf{Cot}} &  \multicolumn{4}{c}{\textbf{Mean Absolute Difference (MAD)}} \\
 & &  & Early & Middle & Last & All\\
\midrule
\multirow{16}{*}{Geometry} & \multirow{3}{*}{$s_Q$} 
  & Simplified & 0.55 & 0.76 & 0.94 & 0.78 \\
& & Detailed & 0.41 & 0.57 & 0.60 & 0.56 \\
\cmidrule{2-7}
& \multirow{3}{*}{$s_K$} 
  & Simplified & 0.47 & 0.53 & 0.72 & 0.58 \\
& & Detailed & 0.36 & 0.52 & 0.49 & 0.49 \\
\cmidrule{2-7}
& \multirow{3}{*}{$s_V$} 
  & Simplified & 2.11 & 2.00 & 0.48 & 1.74 \\
& & Detailed & 1.53 & 1.65 & 0.57 & 1.40 \\
\cmidrule{2-7}
& \multirow{3}{*}{$s_O$} 
  & Simplified & 1.72 & 1.53 & 0.66 & 1.39 \\
& & Detailed & 1.12 & 1.16 & 0.51 & 1.01 \\
\cmidrule{2-7}
 & \multirow{3}{*}{$r_Q$} 
  & Simplified & 0.02 & 0.02 & 0.05 & 0.03 \\
& & Detailed & 0.02 & 0.02 & 0.07 & 0.04 \\
\cmidrule{2-7}
& \multirow{3}{*}{$r_K$} 
  & Simplified & 0.03 & 0.02 & 0.04 & 0.03 \\
& & Detailed & 0.04 & 0.02 & 0.05 & 0.04 \\
\cmidrule{2-7}
 & \multirow{3}{*}{$r_V$} 
  & Simplified & 0.02 & 0.04 & 0.05 & 0.04 \\
& & Detailed & 0.02 & 0.04 & 0.04 & 0.04 \\
\cmidrule{2-7}
& \multirow{3}{*}{$r_O$} 
  & Simplified & 0.01 & 0.02 & 0.03 & 0.02 \\
& & Detailed & 0.01 & 0.02 & 0.04 & 0.02 \\
\bottomrule
\end{tabular}
}
\caption{
Statistical results for MATH-Geometry using gemma-2-2b-it on correct responses.
}
\label{tbl:math_lighteval_math_gemma2_2b_it_grads_geometry}
\end{table}

\begin{figure}[!h]
\centering
\includegraphics[width=\linewidth]{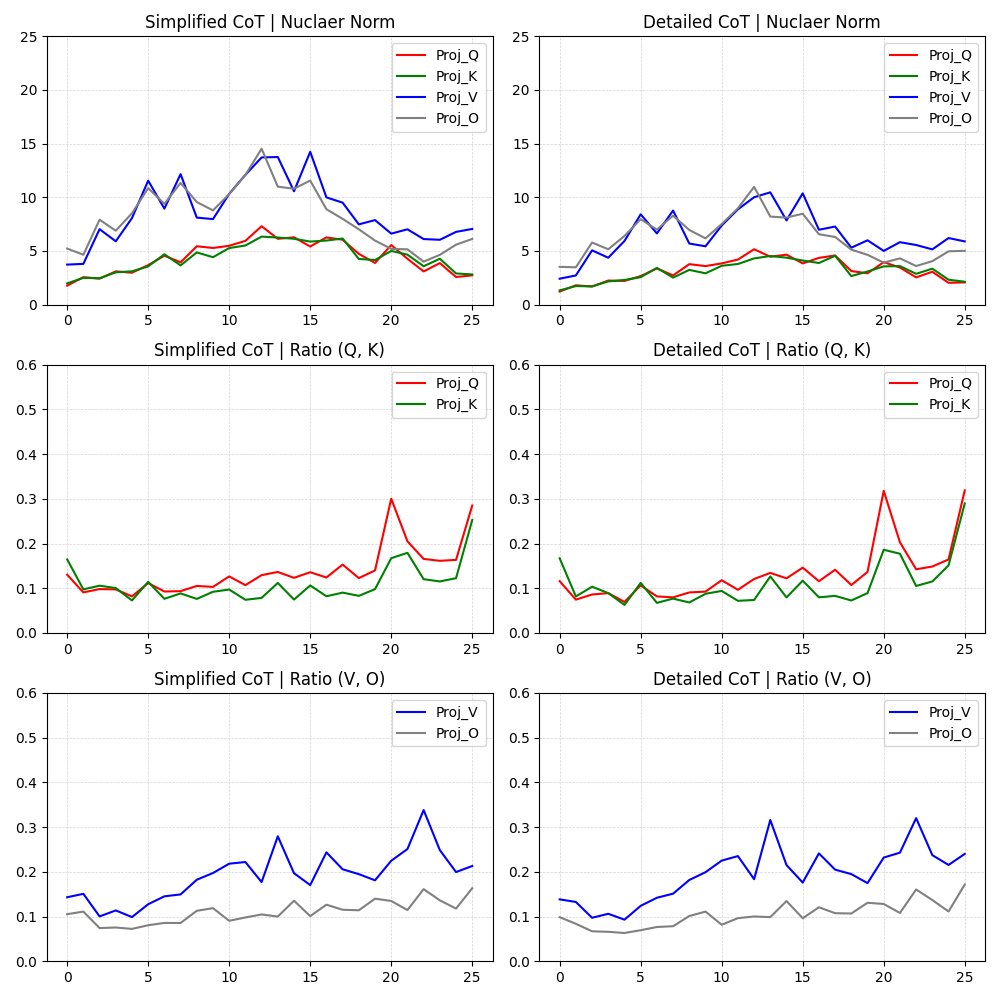}
\caption{Visualization for MATH-Geometry using gemma-2-2b-it on correct responses.}
\label{fig:math_correct_geometry_it}
\vspace{-2mm}
\end{figure}

The visualizations and statistical results on other reasoning tasks: AQuA (Figure \ref{fig:reasoning_correct_aqua_it}, Table \ref{tbl:reasoning_cot_500_gemma2_2b_it_grads_aqua_train}), GSM8K (Figure \ref{fig:reasoning_correct_gsm8k_it}, Table \ref{tbl:reasoning_cot_500_gemma2_2b_it_grads_gsm8k_train}), StrategyQA (Figure \ref{fig:reasoning_correct_strategyqa_it}, Table \ref{tbl:reasoning_cot_500_gemma2_2b_it_grads_strategyqa_train}), ECQA (Figure \ref{fig:reasoning_correct_ecqa_it}, Table \ref{tbl:reasoning_cot_500_gemma2_2b_it_grads_ecqa_train}), CREAK (Figure \ref{fig:reasoning_correct_creak_it}, Table \ref{tbl:reasoning_cot_500_gemma2_2b_it_grads_creak_train}), Sensemaking (Figure \ref{fig:reasoning_correct_sensemaking_it}, Table \ref{tbl:reasoning_cot_500_gemma2_2b_it_grads_sensemaking_train}).

\begin{table}[!tbh]
\centering
\small
\scalebox{0.75}{
\begin{tabular}{l|l|l|cccc}
\toprule
\multirow{2}{*}{\textbf{Dataset}} & \multirow{2}{*}{\textbf{Curve}} & \multirow{2}{*}{\textbf{Cot}} &  \multicolumn{4}{c}{\textbf{Mean Absolute Difference (MAD)}} \\
 & &  & Early & Middle & Last & All\\
\midrule
\multirow{24}{*}{AQuA} & \multirow{3}{*}{$s_Q$} 
  & None & 3.94 & 5.18 & 12.30 & 6.55 \\
& & Simplified & 1.18 & 1.42 & 2.39 & 1.68 \\
& & Detailed & 0.43 & 0.74 & 0.94 & 0.77 \\
\cmidrule{2-7}
& \multirow{3}{*}{$s_K$} 
  & None & 3.77 & 8.15 & 19.98 & 8.94 \\
& & Simplified & 1.18 & 1.94 & 2.05 & 1.82 \\
& & Detailed & 0.39 & 0.80 & 0.71 & 0.70 \\
\cmidrule{2-7}
& \multirow{3}{*}{$s_V$} 
  & None & 16.91 & 24.85 & 8.20 & 19.14 \\
& & Simplified & 4.70 & 5.12 & 1.32 & 4.20 \\
& & Detailed & 1.65 & 2.11 & 0.87 & 1.73 \\
\cmidrule{2-7}
& \multirow{3}{*}{$s_O$} 
  & None & 12.87 & 11.21 & 5.43 & 10.38 \\
& & Simplified & 3.52 & 3.00 & 1.04 & 2.72 \\
& & Detailed & 1.25 & 1.36 & 0.48 & 1.14 \\
\cmidrule{2-7}
 & \multirow{3}{*}{$r_Q$} 
  & None & 0.03 & 0.05 & 0.10 & 0.05 \\
& & Simplified & 0.02 & 0.03 & 0.08 & 0.04 \\
& & Detailed & 0.02 & 0.04 & 0.09 & 0.05 \\
\cmidrule{2-7}
& \multirow{3}{*}{$r_K$} 
  & None & 0.04 & 0.03 & 0.04 & 0.04 \\
& & Simplified & 0.04 & 0.02 & 0.04 & 0.03 \\
& & Detailed & 0.03 & 0.04 & 0.06 & 0.04 \\
\cmidrule{2-7}
 & \multirow{3}{*}{$r_V$} 
  & None & 0.02 & 0.05 & 0.16 & 0.07 \\
& & Simplified & 0.02 & 0.05 & 0.06 & 0.04 \\
& & Detailed & 0.02 & 0.05 & 0.04 & 0.04 \\
\cmidrule{2-7}
& \multirow{3}{*}{$r_O$} 
  & None & 0.01 & 0.05 & 0.08 & 0.04 \\
& & Simplified & 0.01 & 0.04 & 0.04 & 0.03 \\
& & Detailed & 0.01 & 0.03 & 0.04 & 0.02 \\
\bottomrule
\end{tabular}
}
\caption{
Statistical results for AQuA using gemma-2-2b-it on correct responses.
}
\label{tbl:reasoning_cot_500_gemma2_2b_it_grads_aqua_train}
\end{table}

\begin{figure*}[!h]
\centering
\includegraphics[width=\linewidth]{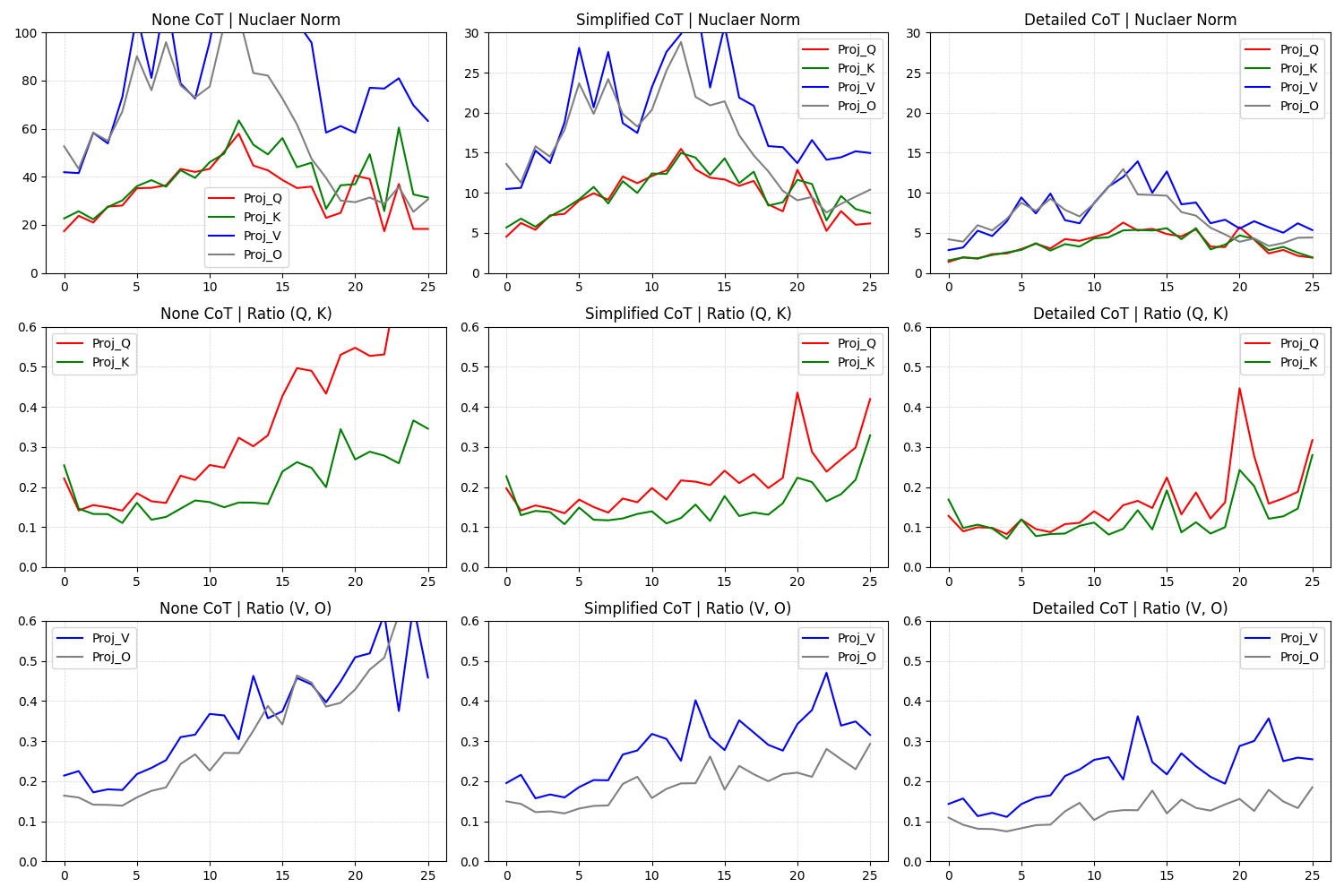}
\caption{Visualization for AQuA using gemma-2-2b-it on correct responses.}
\label{fig:reasoning_correct_aqua_it}
\vspace{-2mm}
\end{figure*}

\begin{table}[!tbh]
\centering
\small
\scalebox{0.75}{
\begin{tabular}{l|l|l|cccc}
\toprule
\multirow{2}{*}{\textbf{Dataset}} & \multirow{2}{*}{\textbf{Curve}} & \multirow{2}{*}{\textbf{Cot}} &  \multicolumn{4}{c}{\textbf{Mean Absolute Difference (MAD)}} \\
 & &  & Early & Middle & Last & All\\
\midrule
\multirow{24}{*}{GSM8K} & \multirow{3}{*}{$s_Q$} 
  & None & 4.64 & 6.94 & 11.69 & 7.45 \\
& & Simplified & 0.72 & 1.24 & 1.44 & 1.25 \\
& & Detailed & 0.41 & 0.89 & 1.11 & 0.91 \\
\cmidrule{2-7}
& \multirow{3}{*}{$s_K$} 
  & None & 4.28 & 10.24 & 15.36 & 9.16 \\
& & Simplified & 0.69 & 1.60 & 0.84 & 1.22 \\
& & Detailed & 0.36 & 1.04 & 0.71 & 0.81 \\
\cmidrule{2-7}
& \multirow{3}{*}{$s_V$} 
  & None & 25.54 & 33.15 & 13.66 & 26.66 \\
& & Simplified & 3.04 & 3.49 & 0.77 & 2.79 \\
& & Detailed & 1.79 & 2.38 & 0.96 & 1.91 \\
\cmidrule{2-7}
& \multirow{3}{*}{$s_O$} 
  & None & 19.49 & 14.00 & 10.25 & 14.57 \\
& & Simplified & 2.35 & 2.36 & 0.38 & 1.93 \\
& & Detailed & 1.39 & 1.59 & 0.35 & 1.27 \\
\cmidrule{2-7}
 & \multirow{3}{*}{$r_Q$} 
  & None & 0.03 & 0.04 & 0.05 & 0.04 \\
& & Simplified & 0.02 & 0.03 & 0.07 & 0.05 \\
& & Detailed & 0.02 & 0.05 & 0.09 & 0.06 \\
\cmidrule{2-7}
& \multirow{3}{*}{$r_K$} 
  & None & 0.04 & 0.03 & 0.05 & 0.04 \\
& & Simplified & 0.03 & 0.03 & 0.04 & 0.03 \\
& & Detailed & 0.04 & 0.04 & 0.06 & 0.05 \\
\cmidrule{2-7}
 & \multirow{3}{*}{$r_V$} 
  & None & 0.03 & 0.06 & 0.10 & 0.06 \\
& & Simplified & 0.03 & 0.05 & 0.05 & 0.04 \\
& & Detailed & 0.03 & 0.06 & 0.04 & 0.04 \\
\cmidrule{2-7}
& \multirow{3}{*}{$r_O$} 
  & None & 0.02 & 0.06 & 0.07 & 0.05 \\
& & Simplified & 0.01 & 0.04 & 0.04 & 0.03 \\
& & Detailed & 0.01 & 0.03 & 0.04 & 0.03 \\
\bottomrule
\end{tabular}
}
\caption{
Statistical results for GSM8K using gemma-2-2b-it on correct responses.
}
\label{tbl:reasoning_cot_500_gemma2_2b_it_grads_gsm8k_train}
\end{table}

\begin{figure*}[!h]
\centering
\includegraphics[width=\linewidth]{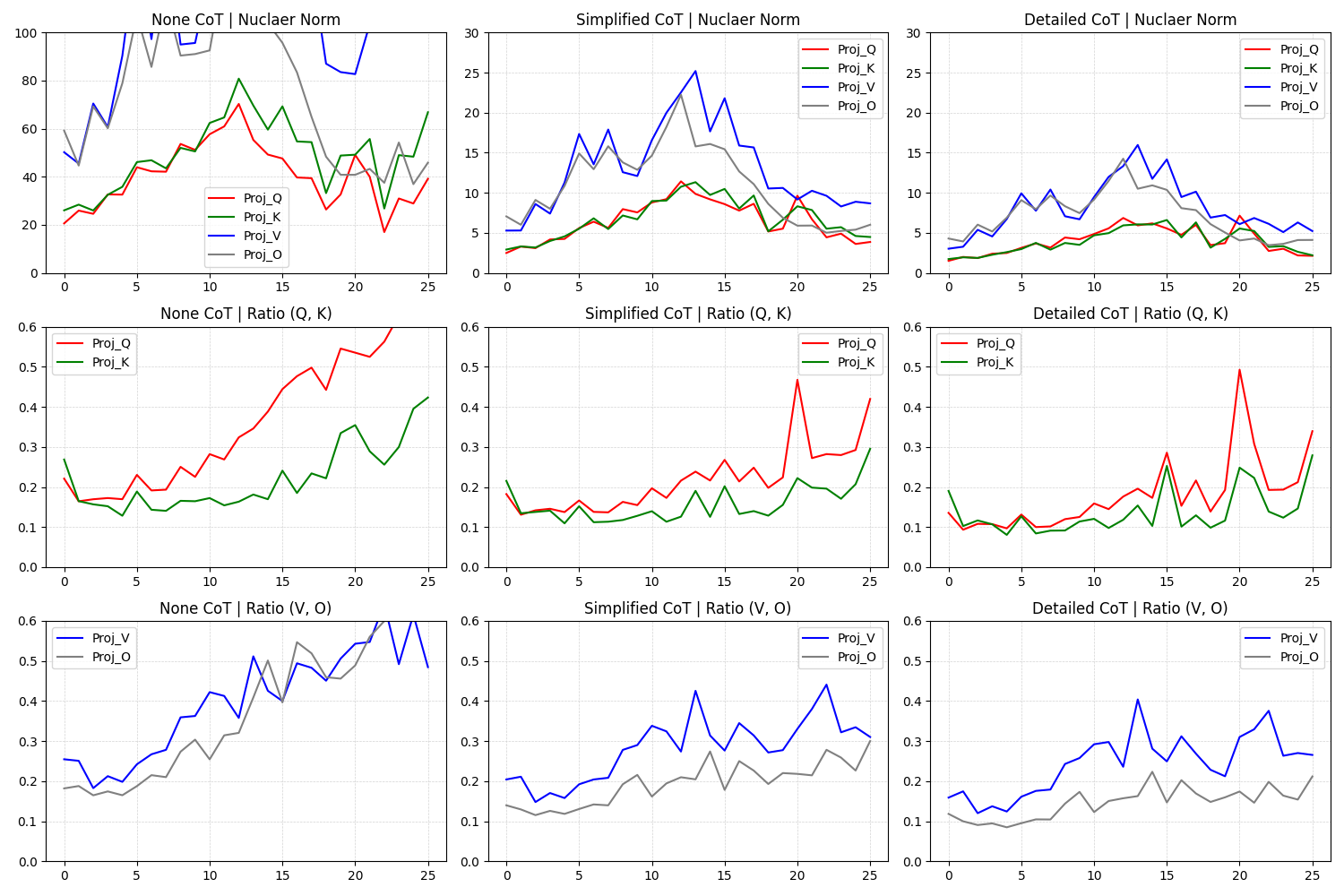}
\caption{Visualization for GSM8K using gemma-2-2b-it on correct responses.}
\label{fig:reasoning_correct_gsm8k_it}
\vspace{-2mm}
\end{figure*}

\begin{table}[!tbh]
\centering
\small
\scalebox{0.75}{
\begin{tabular}{l|l|l|cccc}
\toprule
\multirow{2}{*}{\textbf{Dataset}} & \multirow{2}{*}{\textbf{Curve}} & \multirow{2}{*}{\textbf{Cot}} &  \multicolumn{4}{c}{\textbf{Mean Absolute Difference (MAD)}} \\
 & &  & Early & Middle & Last & All\\
\midrule
\multirow{24}{*}{StrategyQA} & \multirow{3}{*}{$s_Q$} 
  & None & 11.01 & 9.56 & 4.80 & 9.15 \\
& & Simplified & 1.47 & 1.48 & 2.23 & 1.69 \\
& & Detailed & 0.44 & 0.78 & 0.72 & 0.68 \\
\cmidrule{2-7}
& \multirow{3}{*}{$s_K$} 
  & None & 8.98 & 10.89 & 9.19 & 10.78 \\
& & Simplified & 1.00 & 2.82 & 1.73 & 2.00 \\
& & Detailed & 0.36 & 0.64 & 0.44 & 0.52 \\
\cmidrule{2-7}
& \multirow{3}{*}{$s_V$} 
  & None & 48.02 & 63.95 & 29.91 & 52.33 \\
& & Simplified & 5.55 & 6.20 & 1.57 & 5.05 \\
& & Detailed & 2.19 & 2.25 & 0.44 & 1.84 \\
\cmidrule{2-7}
& \multirow{3}{*}{$s_O$} 
  & None & 33.59 & 22.10 & 11.38 & 23.64 \\
& & Simplified & 4.11 & 2.94 & 1.62 & 2.99 \\
& & Detailed & 1.68 & 1.59 & 0.78 & 1.42 \\
\cmidrule{2-7}
 & \multirow{3}{*}{$r_Q$} 
  & None & 0.02 & 0.06 & 0.07 & 0.05 \\
& & Simplified & 0.03 & 0.04 & 0.04 & 0.04 \\
& & Detailed & 0.02 & 0.02 & 0.04 & 0.03 \\
\cmidrule{2-7}
& \multirow{3}{*}{$r_K$} 
  & None & 0.03 & 0.02 & 0.05 & 0.03 \\
& & Simplified & 0.04 & 0.04 & 0.03 & 0.03 \\
& & Detailed & 0.03 & 0.02 & 0.02 & 0.03 \\
\cmidrule{2-7}
 & \multirow{3}{*}{$r_V$} 
  & None & 0.05 & 0.05 & 0.07 & 0.05 \\
& & Simplified & 0.04 & 0.07 & 0.07 & 0.06 \\
& & Detailed & 0.03 & 0.05 & 0.04 & 0.04 \\
\cmidrule{2-7}
& \multirow{3}{*}{$r_O$} 
  & None & 0.03 & 0.06 & 0.07 & 0.05 \\
& & Simplified & 0.02 & 0.04 & 0.06 & 0.04 \\
& & Detailed & 0.01 & 0.03 & 0.03 & 0.02 \\
\bottomrule
\end{tabular}
}
\caption{
Statistical results for StrategyQA using gemma-2-2b-it on correct responses.
}
\label{tbl:reasoning_cot_500_gemma2_2b_it_grads_strategyqa_train}
\end{table}

\begin{figure*}[!h]
\centering
\includegraphics[width=\linewidth]{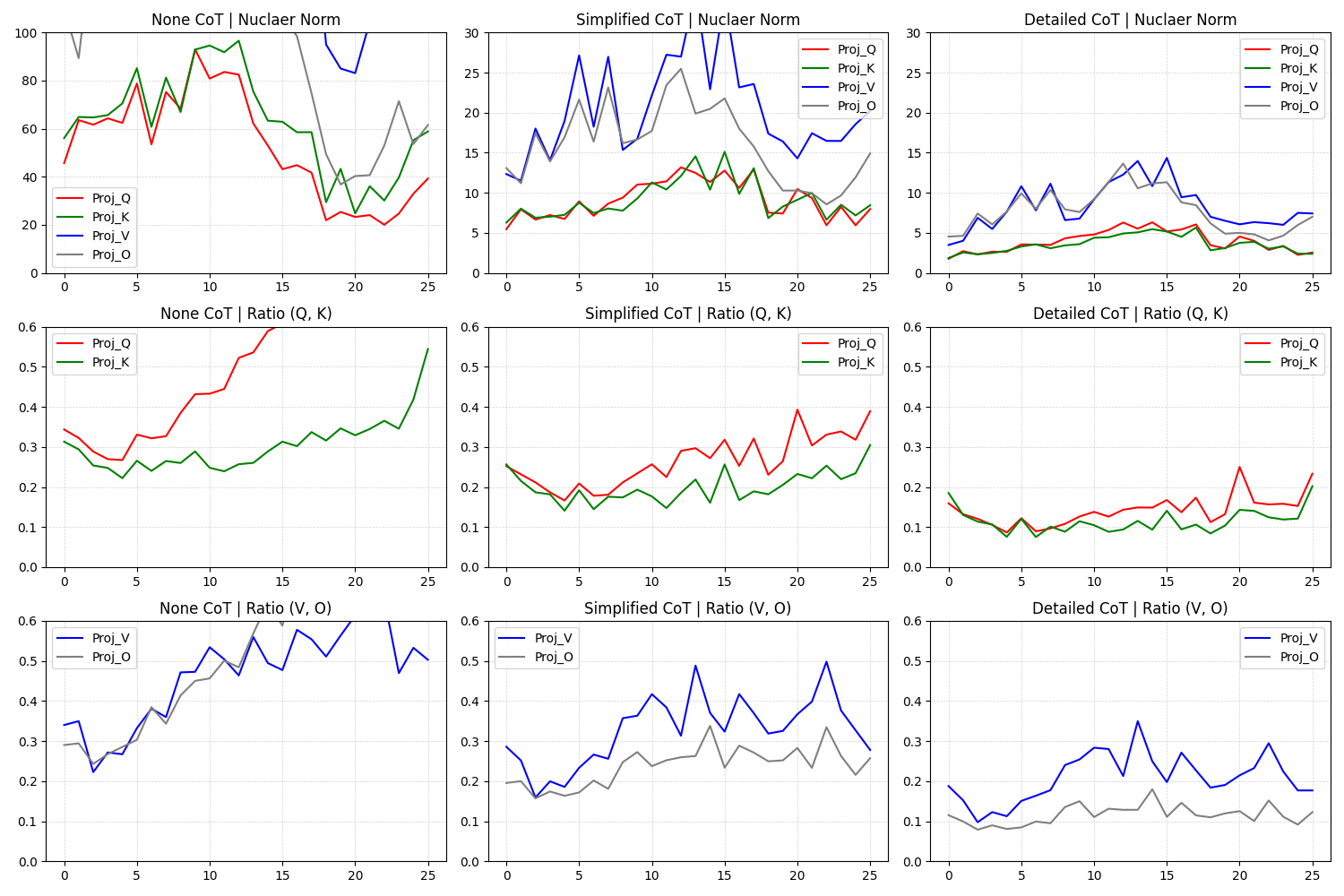}
\caption{Visualization for StrategyQA using gemma-2-2b-it on correct responses.}
\label{fig:reasoning_correct_strategyqa_it}
\vspace{-2mm}
\end{figure*}

\begin{table}[!tbh]
\centering
\small
\scalebox{0.75}{
\begin{tabular}{l|l|l|cccc}
\toprule
\multirow{2}{*}{\textbf{Dataset}} & \multirow{2}{*}{\textbf{Curve}} & \multirow{2}{*}{\textbf{Cot}} &  \multicolumn{4}{c}{\textbf{Mean Absolute Difference (MAD)}} \\
 & &  & Early & Middle & Last & All\\
\midrule
\multirow{24}{*}{ECQA} & \multirow{3}{*}{$s_Q$} 
  & None & 8.08 & 13.75 & 11.11 & 11.72 \\
& & Simplified & 1.85 & 3.70 & 2.91 & 3.27 \\
& & Detailed & 0.55 & 0.97 & 0.92 & 0.90 \\
\cmidrule{2-7}
& \multirow{3}{*}{$s_K$} 
  & None & 6.91 & 19.80 & 10.42 & 13.94 \\
& & Simplified & 1.39 & 5.91 & 1.71 & 3.69 \\
& & Detailed & 0.47 & 1.07 & 0.59 & 0.81 \\
\cmidrule{2-7}
& \multirow{3}{*}{$s_V$} 
  & None & 39.01 & 47.62 & 22.02 & 39.91 \\
& & Simplified & 8.46 & 9.71 & 2.91 & 7.90 \\
& & Detailed & 2.83 & 3.04 & 0.90 & 2.53 \\
\cmidrule{2-7}
& \multirow{3}{*}{$s_O$} 
  & None & 29.11 & 19.93 & 5.20 & 19.57 \\
& & Simplified & 6.28 & 5.20 & 1.27 & 4.71 \\
& & Detailed & 2.10 & 2.22 & 0.70 & 1.85 \\
\cmidrule{2-7}
 & \multirow{3}{*}{$r_Q$} 
  & None & 0.02 & 0.08 & 0.10 & 0.07 \\
& & Simplified & 0.02 & 0.05 & 0.07 & 0.05 \\
& & Detailed & 0.02 & 0.04 & 0.07 & 0.05 \\
\cmidrule{2-7}
& \multirow{3}{*}{$r_K$} 
  & None & 0.03 & 0.03 & 0.07 & 0.04 \\
& & Simplified & 0.03 & 0.05 & 0.05 & 0.05 \\
& & Detailed & 0.03 & 0.04 & 0.05 & 0.04 \\
\cmidrule{2-7}
 & \multirow{3}{*}{$r_V$} 
  & None & 0.03 & 0.06 & 0.07 & 0.05 \\
& & Simplified & 0.03 & 0.06 & 0.05 & 0.05 \\
& & Detailed & 0.02 & 0.05 & 0.03 & 0.04 \\
\cmidrule{2-7}
& \multirow{3}{*}{$r_O$} 
  & None & 0.02 & 0.07 & 0.11 & 0.06 \\
& & Simplified & 0.02 & 0.05 & 0.04 & 0.04 \\
& & Detailed & 0.01 & 0.03 & 0.03 & 0.02 \\
\bottomrule
\end{tabular}
}
\caption{
Statistical results for ECQA using gemma-2-2b-it on correct responses.
}
\label{tbl:reasoning_cot_500_gemma2_2b_it_grads_ecqa_train}
\end{table}

\begin{figure*}[!h]
\centering
\includegraphics[width=\linewidth]{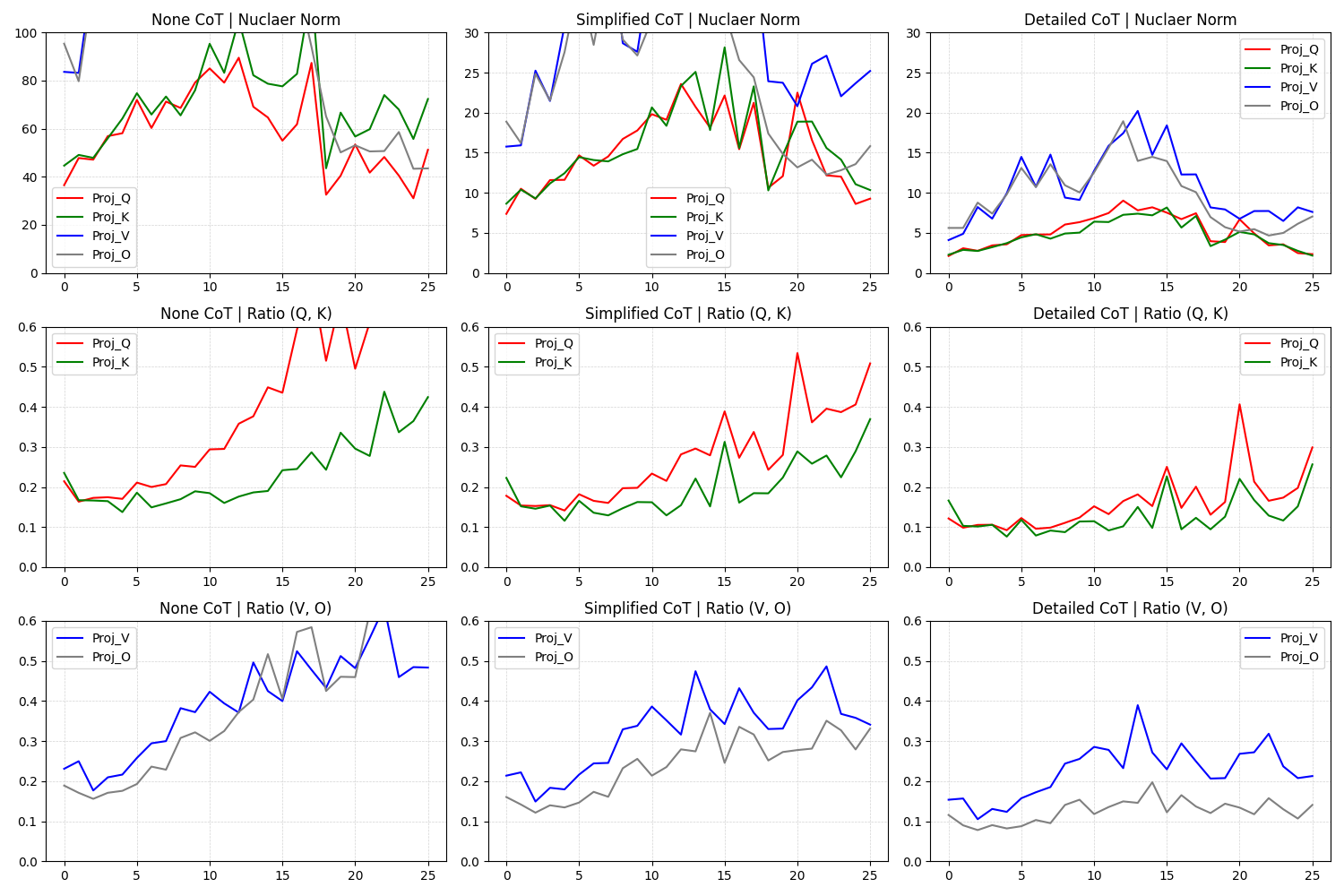}
\caption{Visualization for ECQA using gemma-2-2b-it on correct responses.}
\label{fig:reasoning_correct_ecqa_it}
\vspace{-2mm}
\end{figure*}

\begin{table}[!tbh]
\centering
\small
\scalebox{0.75}{
\begin{tabular}{l|l|l|cccc}
\toprule
\multirow{2}{*}{\textbf{Dataset}} & \multirow{2}{*}{\textbf{Curve}} & \multirow{2}{*}{\textbf{Cot}} &  \multicolumn{4}{c}{\textbf{Mean Absolute Difference (MAD)}} \\
 & &  & Early & Middle & Last & All\\
\midrule
\multirow{24}{*}{CREAK} & \multirow{3}{*}{$s_Q$} 
  & None & 19.40 & 21.48 & 18.64 & 20.16 \\
& & Simplified & 2.17 & 3.38 & 3.35 & 3.24 \\
& & Detailed & 0.52 & 0.96 & 1.08 & 0.92 \\
\cmidrule{2-7}
& \multirow{3}{*}{$s_K$} 
  & None & 15.51 & 28.99 & 28.47 & 26.59 \\
& & Simplified & 1.52 & 5.41 & 2.04 & 3.46 \\
& & Detailed & 0.42 & 0.96 & 0.70 & 0.77 \\
\cmidrule{2-7}
& \multirow{3}{*}{$s_V$} 
  & None & 97.03 & 109.60 & 26.16 & 90.55 \\
& & Simplified & 9.55 & 10.54 & 3.15 & 8.84 \\
& & Detailed & 2.64 & 2.84 & 0.70 & 2.34 \\
\cmidrule{2-7}
& \multirow{3}{*}{$s_O$} 
  & None & 67.77 & 37.98 & 9.63 & 41.16 \\
& & Simplified & 7.08 & 4.32 & 1.76 & 4.69 \\
& & Detailed & 1.96 & 1.90 & 0.81 & 1.67 \\
\cmidrule{2-7}
 & \multirow{3}{*}{$r_Q$} 
  & None & 0.02 & 0.07 & 0.09 & 0.07 \\
& & Simplified & 0.02 & 0.05 & 0.09 & 0.05 \\
& & Detailed & 0.02 & 0.03 & 0.06 & 0.04 \\
\cmidrule{2-7}
& \multirow{3}{*}{$r_K$} 
  & None & 0.03 & 0.04 & 0.08 & 0.04 \\
& & Simplified & 0.03 & 0.04 & 0.06 & 0.04 \\
& & Detailed & 0.03 & 0.03 & 0.04 & 0.03 \\
\cmidrule{2-7}
 & \multirow{3}{*}{$r_V$} 
  & None & 0.03 & 0.05 & 0.07 & 0.05 \\
& & Simplified & 0.04 & 0.06 & 0.05 & 0.05 \\
& & Detailed & 0.03 & 0.05 & 0.04 & 0.04 \\
\cmidrule{2-7}
& \multirow{3}{*}{$r_O$} 
  & None & 0.02 & 0.06 & 0.08 & 0.05 \\
& & Simplified & 0.02 & 0.04 & 0.04 & 0.03 \\
& & Detailed & 0.01 & 0.03 & 0.03 & 0.02 \\
\bottomrule
\end{tabular}
}
\caption{
Statistical results for CREAK using gemma-2-2b-it on correct responses.
}
\label{tbl:reasoning_cot_500_gemma2_2b_it_grads_creak_train}
\end{table}

\begin{figure*}[!h]
\centering
\includegraphics[width=\linewidth]{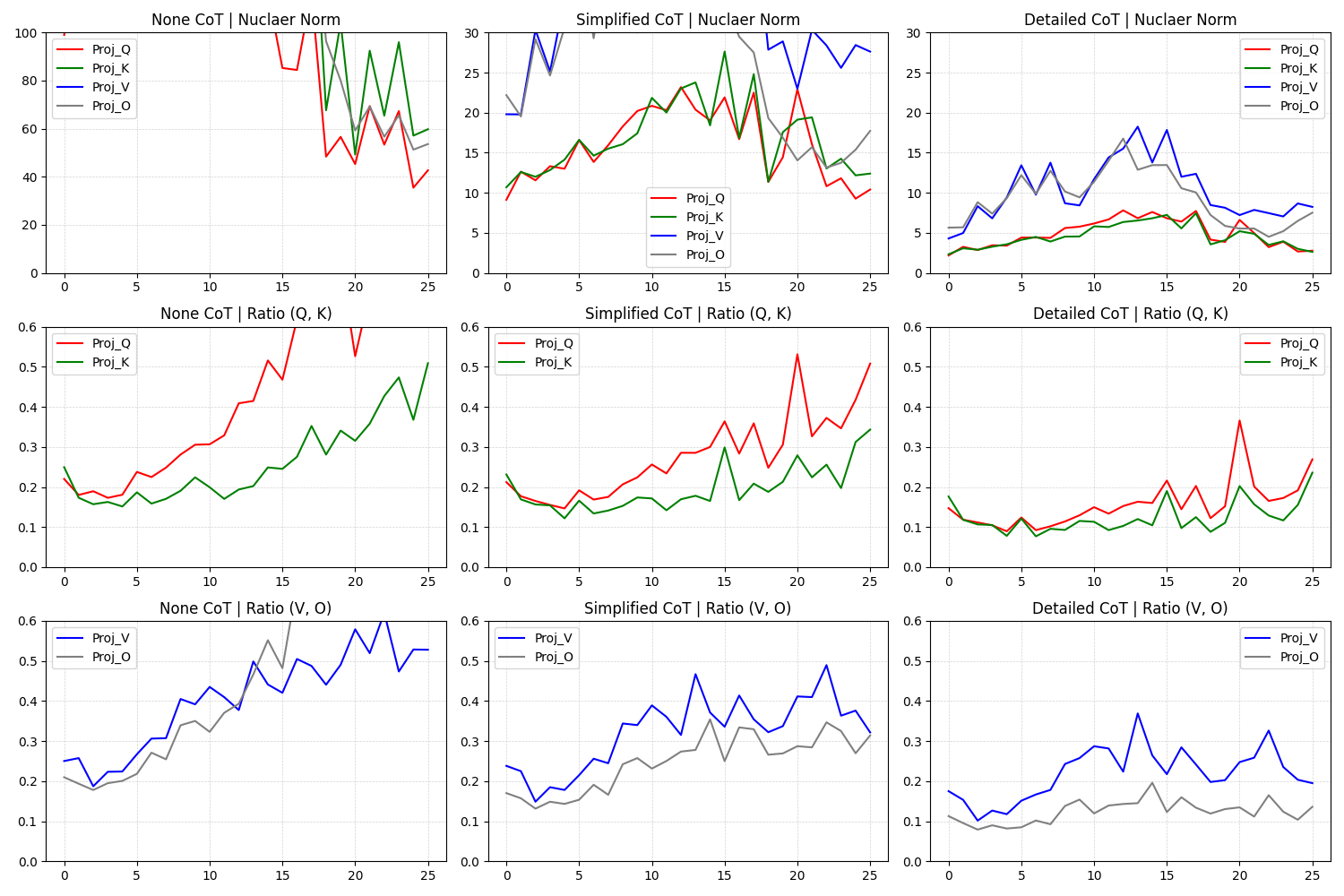}
\caption{Visualization for CREAK using gemma-2-2b-it on correct responses.}
\label{fig:reasoning_correct_creak_it}
\vspace{-2mm}
\end{figure*}

\begin{table}[!tbh]
\centering
\small
\scalebox{0.75}{
\begin{tabular}{l|l|l|cccc}
\toprule
\multirow{2}{*}{\textbf{Dataset}} & \multirow{2}{*}{\textbf{Curve}} & \multirow{2}{*}{\textbf{Cot}} &  \multicolumn{4}{c}{\textbf{Mean Absolute Difference (MAD)}} \\
 & &  & Early & Middle & Last & All\\
\midrule
\multirow{24}{*}{Sensemaking} & \multirow{3}{*}{$s_Q$} 
  & None & 6.07 & 12.93 & 10.41 & 10.52 \\
& & Simplified & 2.03 & 2.84 & 2.93 & 2.81 \\
& & Detailed & 0.51 & 0.99 & 1.00 & 0.92 \\
\cmidrule{2-7}
& \multirow{3}{*}{$s_K$} 
  & None & 6.14 & 17.32 & 13.03 & 12.55 \\
& & Simplified & 1.53 & 5.10 & 3.22 & 3.56 \\
& & Detailed & 0.43 & 0.97 & 0.67 & 0.78 \\
\cmidrule{2-7}
& \multirow{3}{*}{$s_V$} 
  & None & 34.29 & 44.41 & 9.37 & 35.20 \\
& & Simplified & 9.30 & 9.10 & 3.85 & 8.18 \\
& & Detailed & 2.29 & 2.79 & 0.84 & 2.24 \\
\cmidrule{2-7}
& \multirow{3}{*}{$s_O$} 
  & None & 24.30 & 19.65 & 3.97 & 18.15 \\
& & Simplified & 6.91 & 5.03 & 1.18 & 4.86 \\
& & Detailed & 1.68 & 2.06 & 0.58 & 1.63 \\
\cmidrule{2-7}
 & \multirow{3}{*}{$r_Q$} 
  & None & 0.02 & 0.08 & 0.07 & 0.06 \\
& & Simplified & 0.02 & 0.04 & 0.08 & 0.05 \\
& & Detailed & 0.02 & 0.04 & 0.07 & 0.05 \\
\cmidrule{2-7}
& \multirow{3}{*}{$r_K$} 
  & None & 0.04 & 0.02 & 0.05 & 0.03 \\
& & Simplified & 0.03 & 0.04 & 0.05 & 0.04 \\
& & Detailed & 0.04 & 0.04 & 0.05 & 0.04 \\
\cmidrule{2-7}
 & \multirow{3}{*}{$r_V$} 
  & None & 0.02 & 0.05 & 0.07 & 0.05 \\
& & Simplified & 0.03 & 0.05 & 0.05 & 0.05 \\
& & Detailed & 0.02 & 0.05 & 0.03 & 0.04 \\
\cmidrule{2-7}
& \multirow{3}{*}{$r_O$} 
  & None & 0.02 & 0.08 & 0.08 & 0.06 \\
& & Simplified & 0.02 & 0.04 & 0.04 & 0.03 \\
& & Detailed & 0.01 & 0.03 & 0.03 & 0.02 \\
\bottomrule
\end{tabular}
}
\caption{
Statistical results for Sensemaking using gemma-2-2b-it on correct responses.
}
\label{tbl:reasoning_cot_500_gemma2_2b_it_grads_sensemaking_train}
\end{table}

\begin{figure*}[!h]
\centering
\includegraphics[width=\linewidth]{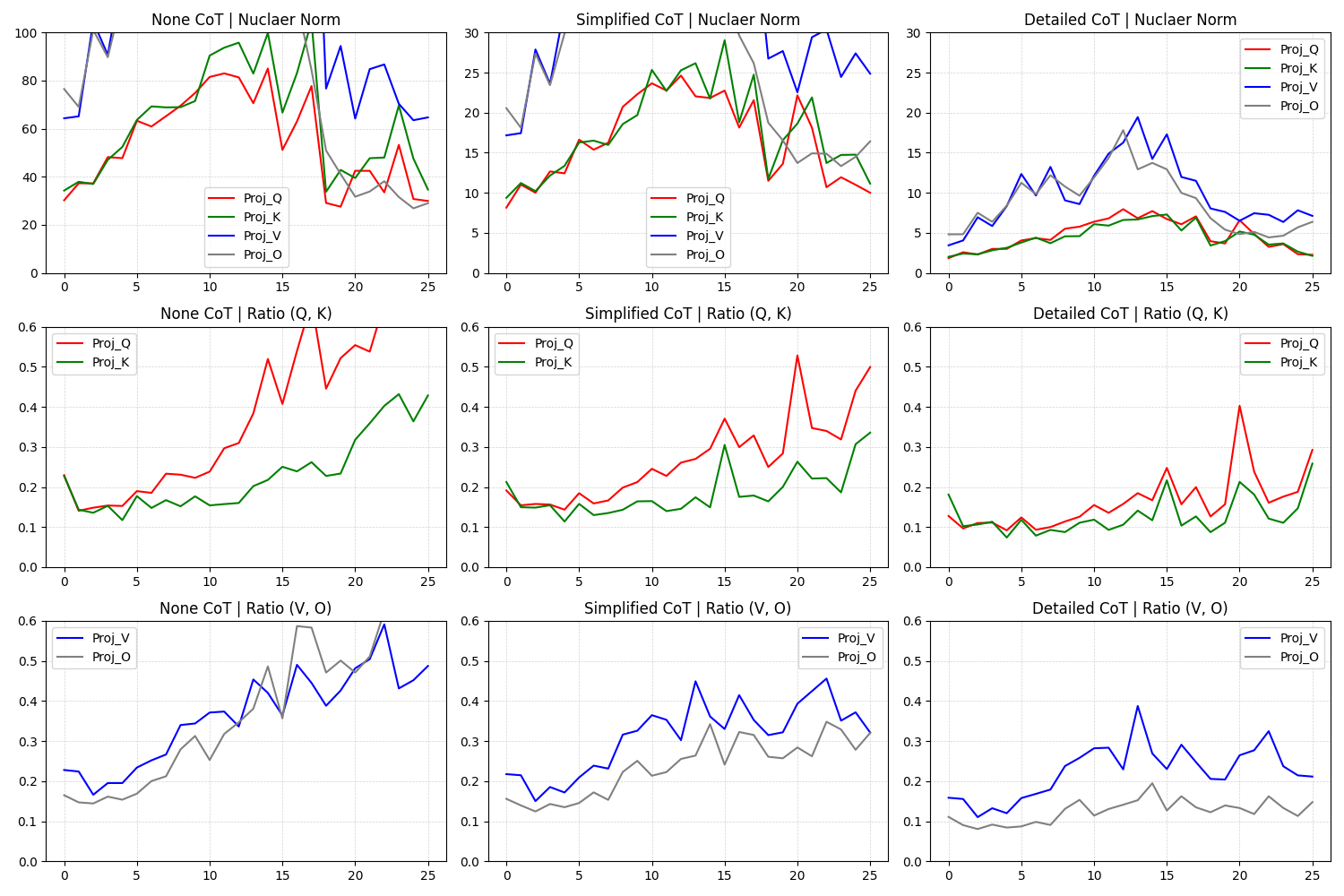}
\caption{Visualization for Sensemaking using gemma-2-2b-it on correct responses.}
\label{fig:reasoning_correct_sensemaking_it}
\vspace{-2mm}
\end{figure*}

\subsubsection{Wiki Tasks}

The visualizations and statistical results on Wiki tasks are shown in Figure \ref{fig:wiki_correct_it} and Table \ref{tbl:wiki_wiki_popularity_gemma2_2b_it_grads}.

\begin{table}[!tbh]
\centering
\small
\scalebox{0.75}{
\begin{tabular}{l|l|l|cccc}
\toprule
\multirow{2}{*}{\textbf{Dataset}} & \multirow{2}{*}{\textbf{Curve}} & \multirow{2}{*}{\textbf{Cot}} &  \multicolumn{4}{c}{\textbf{Mean Absolute Difference (MAD)}} \\
 & &  & Early & Middle & Last & All\\
\midrule
\multirow{32}{*}{Wiki} & \multirow{3}{*}{$s_Q$} 
  & Len 100 & 0.96 & 0.89 & 1.41 & 1.03 \\
& & Len 500 & 0.60 & 0.50 & 0.78 & 0.58 \\
& & Len 1000 & 0.55 & 0.46 & 0.64 & 0.52 \\
& & Unpopular & 1.58 & 1.09 & 1.08 & 1.19 \\
\cmidrule{2-7}
& \multirow{3}{*}{$s_K$} 
  & Len 100 & 0.71 & 1.39 & 1.11 & 1.10 \\
& & Len 500 & 0.42 & 0.78 & 0.66 & 0.65 \\
& & Len 1000 & 0.34 & 0.57 & 0.53 & 0.50 \\
& & Unpopular & 1.52 & 2.14 & 0.79 & 1.58 \\
\cmidrule{2-7}
& \multirow{3}{*}{$s_V$} 
  & Len 100 & 3.44 & 3.72 & 1.28 & 3.09 \\
& & Len 500 & 2.59 & 2.52 & 1.07 & 2.20 \\
& & Len 1000 & 2.20 & 2.06 & 0.86 & 1.82 \\
& & Unpopular & 4.56 & 5.46 & 1.83 & 4.42 \\
\cmidrule{2-7}
& \multirow{3}{*}{$s_O$} 
  & Len 100 & 2.80 & 2.03 & 2.23 & 2.28 \\
& & Len 500 & 2.09 & 1.58 & 1.57 & 1.69 \\
& & Len 1000 & 1.85 & 1.37 & 1.30 & 1.46 \\
& & Unpopular & 3.18 & 2.27 & 2.28 & 2.46 \\
\cmidrule{2-7}
 & \multirow{3}{*}{$r_Q$} 
  & Len 100 & 0.03 & 0.03 & 0.03 & 0.03 \\
& & Len 500 & 0.02 & 0.03 & 0.03 & 0.03 \\
& & Len 1000 & 0.02 & 0.03 & 0.03 & 0.03 \\
& & Unpopular & 0.03 & 0.03 & 0.02 & 0.03 \\
\cmidrule{2-7}
& \multirow{3}{*}{$r_K$} 
  & Len 100 & 0.04 & 0.03 & 0.03 & 0.03 \\
& & Len 500 & 0.03 & 0.04 & 0.03 & 0.03 \\
& & Len 1000 & 0.04 & 0.04 & 0.03 & 0.03 \\
& & Unpopular & 0.03 & 0.03 & 0.03 & 0.03 \\
\cmidrule{2-7}
 & \multirow{3}{*}{$r_V$} 
  & Len 100 & 0.04 & 0.07 & 0.09 & 0.06 \\
& & Len 500 & 0.03 & 0.05 & 0.06 & 0.05 \\
& & Len 1000 & 0.03 & 0.05 & 0.05 & 0.04 \\
& & Unpopular & 0.05 & 0.08 & 0.08 & 0.07 \\
\cmidrule{2-7}
& \multirow{3}{*}{$r_O$} 
  & Len 100 & 0.02 & 0.04 & 0.07 & 0.04 \\
& & Len 500 & 0.01 & 0.03 & 0.04 & 0.03 \\
& & Len 1000 & 0.01 & 0.02 & 0.03 & 0.02 \\
& & Unpopular & 0.02 & 0.03 & 0.06 & 0.03 \\
\bottomrule
\end{tabular}
}
\caption{
Statistical results for Wiki using gemma-2-2b-it on correct responses.
}
\label{tbl:wiki_wiki_popularity_gemma2_2b_it_new_grads}
\end{table}

\begin{figure*}[!h]
\centering
\includegraphics[width=\linewidth]{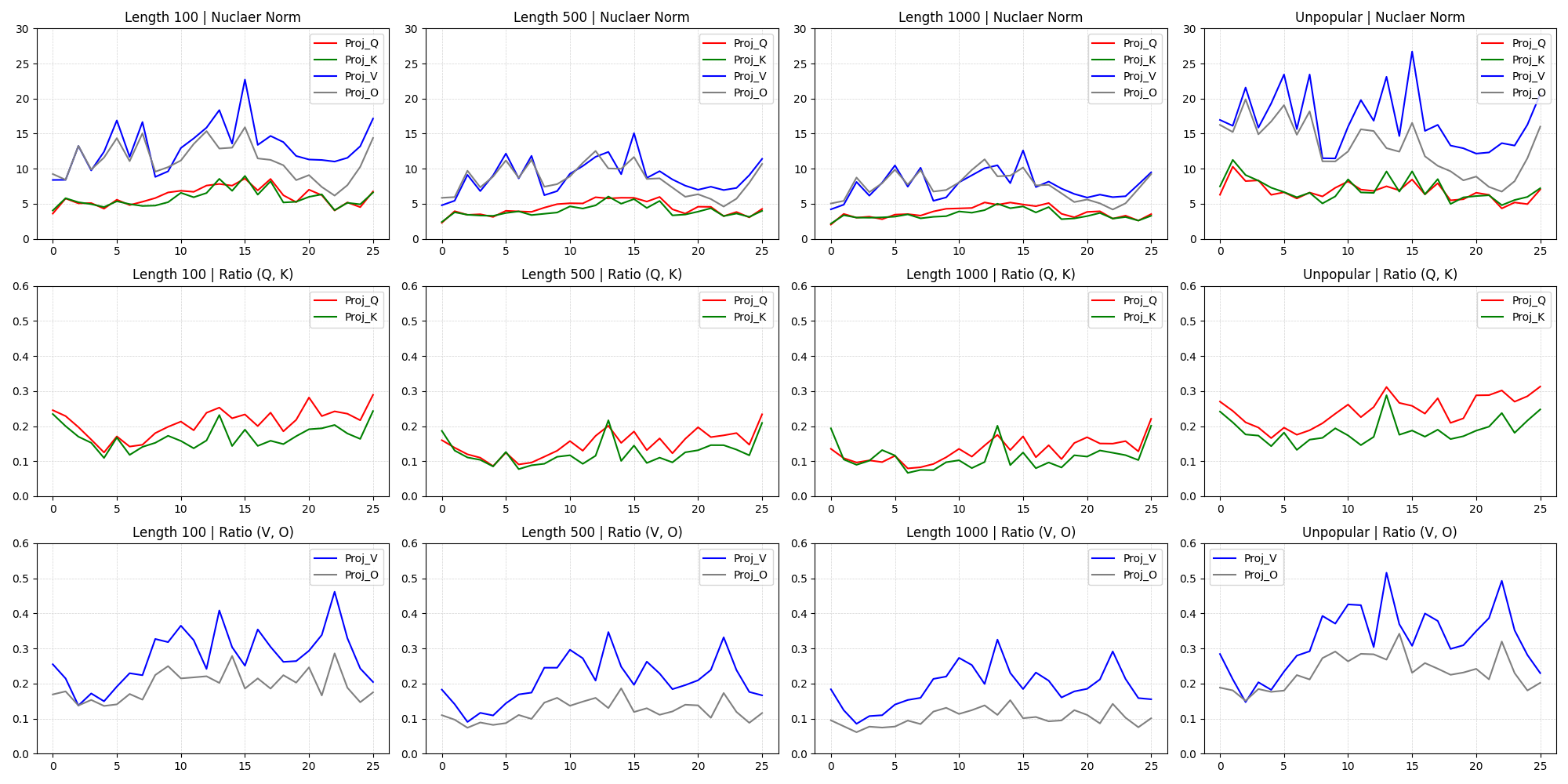}
\caption{Visualization for Wiki tasks using gemma-2-2b-it on correct responses.}
\label{fig:wiki_correct_it}
\vspace{-2mm}
\end{figure*}

\subsection{Instructed LLM on Irrelevant Responses}

\subsubsection{Reasoning Tasks}

The visualizations and statistical results on MATH tasks: MATH-Algebra (Figure \ref{fig:math_nonsense_algebra_it}, Table \ref{tbl:math_lighteval_math_wrong_answer_shuffle_gemma2_2b_it_grads_algebra}), MATH-Counting (Figure \ref{fig:math_nonsense_counting_it}, Table \ref{tbl:math_lighteval_math_wrong_answer_shuffle_gemma2_2b_it_grads_counting_probability}), MATH-Geometry (Figure \ref{fig:math_nonsense_geometry_it}, Table \ref{tbl:math_lighteval_math_wrong_answer_shuffle_gemma2_2b_it_grads_geometry}).

\begin{table}[!tbh]
\centering
\small
\scalebox{0.75}{
\begin{tabular}{l|l|l|cccc}
\toprule
\multirow{2}{*}{\textbf{Dataset}} & \multirow{2}{*}{\textbf{Curve}} & \multirow{2}{*}{\textbf{Cot}} &  \multicolumn{4}{c}{\textbf{Mean Absolute Difference (MAD)}} \\
 & &  & Early & Middle & Last & All\\
\midrule
\multirow{16}{*}{Algebra} & \multirow{3}{*}{$s_Q$} 
  & Simplified & 0.63 & 0.75 & 1.32 & 0.89 \\
& & Detailed & 0.47 & 0.47 & 0.83 & 0.59 \\
\cmidrule{2-7}
& \multirow{3}{*}{$s_K$} 
  & Simplified & 0.58 & 0.70 & 1.03 & 0.75 \\
& & Detailed & 0.44 & 0.48 & 0.63 & 0.52 \\
\cmidrule{2-7}
& \multirow{3}{*}{$s_V$} 
  & Simplified & 3.42 & 3.02 & 1.28 & 2.83 \\
& & Detailed & 2.10 & 1.81 & 0.84 & 1.72 \\
\cmidrule{2-7}
& \multirow{3}{*}{$s_O$} 
  & Simplified & 2.76 & 1.89 & 1.30 & 2.02 \\
& & Detailed & 1.55 & 1.24 & 0.83 & 1.25 \\
\cmidrule{2-7}
 & \multirow{3}{*}{$r_Q$} 
  & Simplified & 0.02 & 0.03 & 0.09 & 0.04 \\
& & Detailed & 0.03 & 0.03 & 0.09 & 0.05 \\
\cmidrule{2-7}
& \multirow{3}{*}{$r_K$} 
  & Simplified & 0.03 & 0.03 & 0.08 & 0.04 \\
& & Detailed & 0.04 & 0.03 & 0.08 & 0.04 \\
\cmidrule{2-7}
 & \multirow{3}{*}{$r_V$} 
  & Simplified & 0.03 & 0.05 & 0.06 & 0.05 \\
& & Detailed & 0.02 & 0.04 & 0.05 & 0.04 \\
\cmidrule{2-7}
& \multirow{3}{*}{$r_O$} 
  & Simplified & 0.01 & 0.02 & 0.04 & 0.02 \\
& & Detailed & 0.01 & 0.01 & 0.03 & 0.02 \\
\bottomrule
\end{tabular}
}
\caption{
Statistical results for MATH-Algebra using gemma-2-2b-it on irrelevant responses.
}
\label{tbl:math_lighteval_math_wrong_answer_shuffle_gemma2_2b_it_grads_algebra}
\end{table}

\begin{figure}[!h]
\centering
\includegraphics[width=\linewidth]{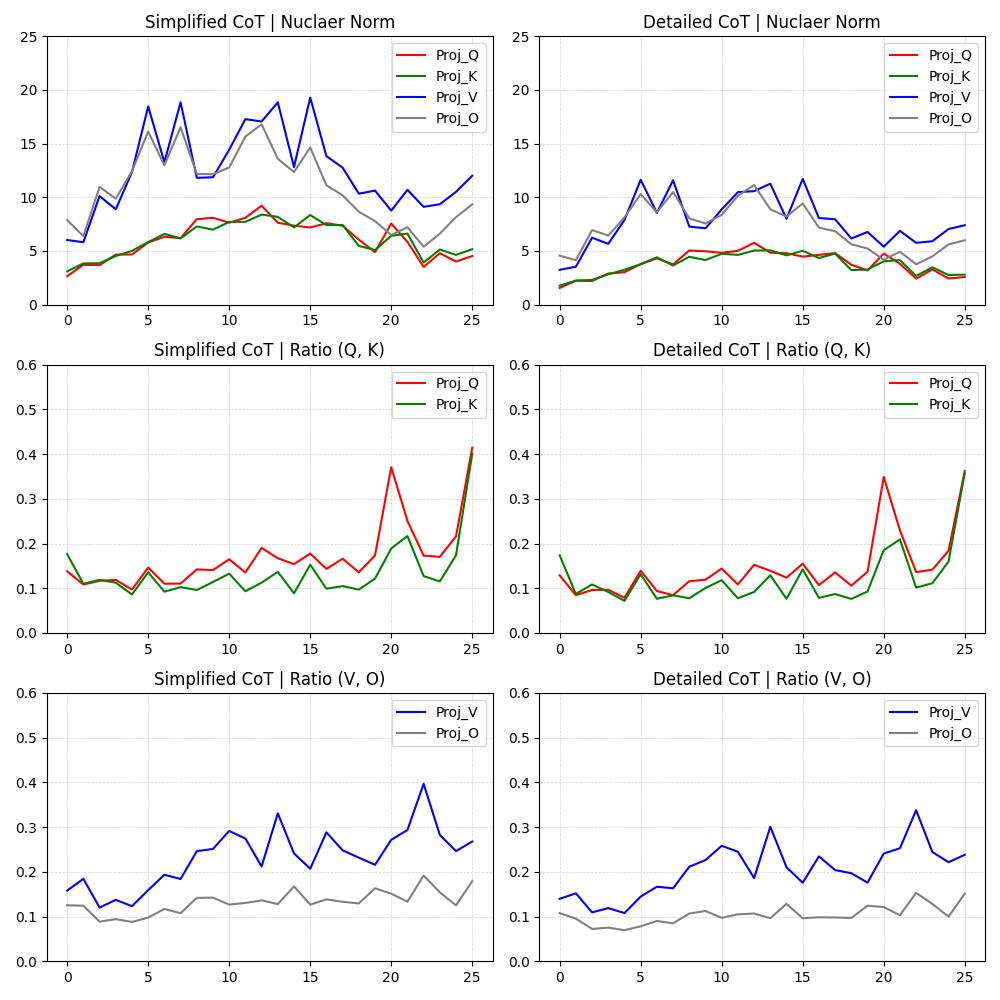}
\caption{Visualization for MATH-Algebra using gemma-2-2b-it on irrelevant responses.}
\label{fig:math_nonsense_algebra_it}
\vspace{-2mm}
\end{figure}

\begin{table}[!tbh]
\centering
\small
\scalebox{0.75}{
\begin{tabular}{l|l|l|cccc}
\toprule
\multirow{2}{*}{\textbf{Dataset}} & \multirow{2}{*}{\textbf{Curve}} & \multirow{2}{*}{\textbf{Cot}} &  \multicolumn{4}{c}{\textbf{Mean Absolute Difference (MAD)}} \\
 & &  & Early & Middle & Last & All\\
\midrule
\multirow{16}{*}{Counting} & \multirow{3}{*}{$s_Q$} 
  & Simplified & 0.63 & 0.64 & 1.22 & 0.80 \\
& & Detailed & 0.56 & 0.49 & 0.80 & 0.62 \\
\cmidrule{2-7}
& \multirow{3}{*}{$s_K$} 
  & Simplified & 0.58 & 0.69 & 0.99 & 0.74 \\
& & Detailed & 0.50 & 0.48 & 0.63 & 0.54 \\
\cmidrule{2-7}
& \multirow{3}{*}{$s_V$} 
  & Simplified & 3.47 & 3.00 & 1.20 & 2.76 \\
& & Detailed & 2.47 & 1.78 & 0.80 & 1.76 \\
\cmidrule{2-7}
& \multirow{3}{*}{$s_O$} 
  & Simplified & 2.87 & 1.78 & 1.22 & 1.94 \\
& & Detailed & 1.79 & 1.22 & 0.82 & 1.28 \\
\cmidrule{2-7}
 & \multirow{3}{*}{$r_Q$} 
  & Simplified & 0.03 & 0.03 & 0.08 & 0.04 \\
& & Detailed & 0.04 & 0.03 & 0.09 & 0.05 \\
\cmidrule{2-7}
& \multirow{3}{*}{$r_K$} 
  & Simplified & 0.04 & 0.03 & 0.08 & 0.04 \\
& & Detailed & 0.05 & 0.03 & 0.08 & 0.05 \\
\cmidrule{2-7}
 & \multirow{3}{*}{$r_V$} 
  & Simplified & 0.03 & 0.05 & 0.06 & 0.04 \\
& & Detailed & 0.02 & 0.04 & 0.05 & 0.04 \\
\cmidrule{2-7}
& \multirow{3}{*}{$r_O$} 
  & Simplified & 0.01 & 0.02 & 0.03 & 0.02 \\
& & Detailed & 0.01 & 0.02 & 0.03 & 0.02 \\
\bottomrule
\end{tabular}
}
\caption{
Statistical results for MATH-Counting using gemma-2-2b-it on irrelevant responses.
}
\label{tbl:math_lighteval_math_wrong_answer_shuffle_gemma2_2b_it_grads_counting_probability}
\end{table}

\begin{figure}[!h]
\centering
\includegraphics[width=\linewidth]{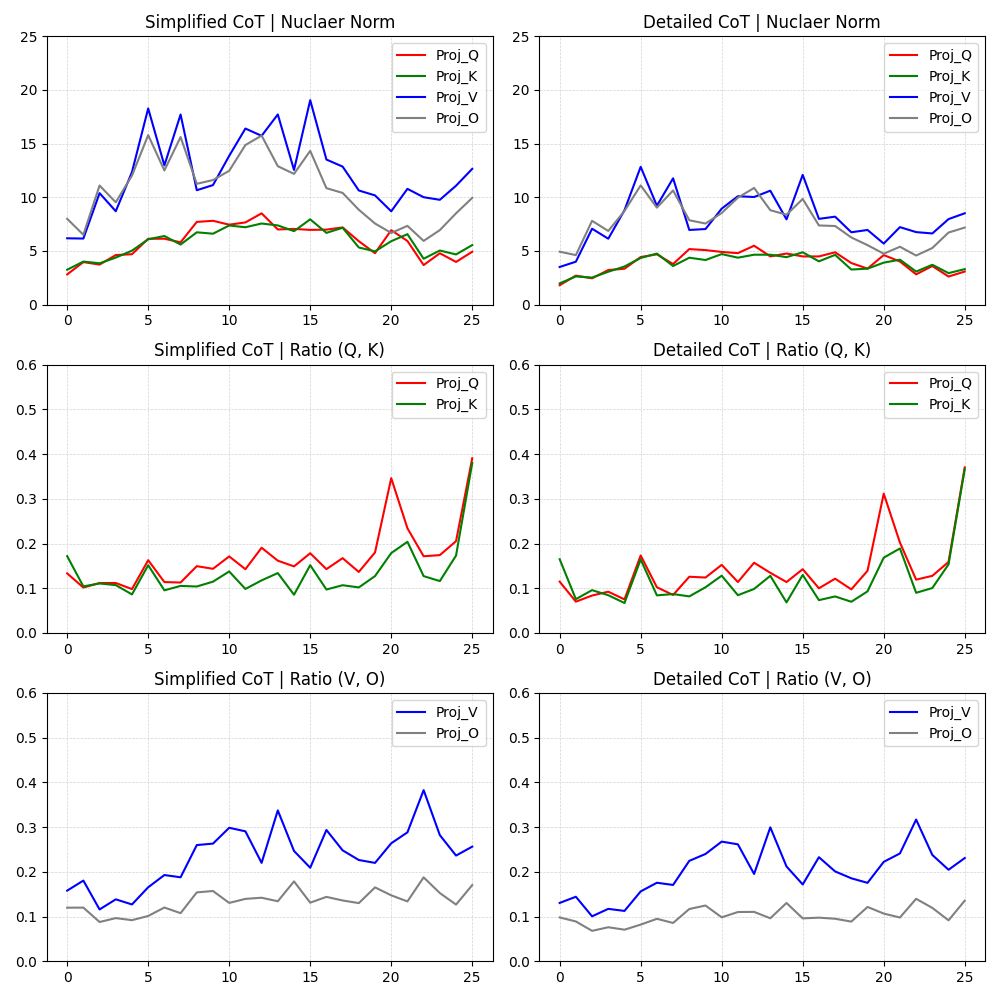}
\caption{Visualization for MATH-Counting using gemma-2-2b-it on irrelevant responses.}
\label{fig:math_nonsense_counting_it}
\vspace{-2mm}
\end{figure}

\begin{table}[!tbh]
\centering
\small
\scalebox{0.75}{
\begin{tabular}{l|l|l|cccc}
\toprule
\multirow{2}{*}{\textbf{Dataset}} & \multirow{2}{*}{\textbf{Curve}} & \multirow{2}{*}{\textbf{Cot}} &  \multicolumn{4}{c}{\textbf{Mean Absolute Difference (MAD)}} \\
 & &  & Early & Middle & Last & All\\
\midrule
\multirow{16}{*}{Geometry} & \multirow{3}{*}{$s_Q$} 
  & Simplified & 0.60 & 0.65 & 1.36 & 0.84 \\
& & Detailed & 0.57 & 0.44 & 1.23 & 0.70 \\
\cmidrule{2-7}
& \multirow{3}{*}{$s_K$} 
  & Simplified & 0.50 & 0.51 & 1.07 & 0.67 \\
& & Detailed & 0.51 & 0.55 & 1.26 & 0.75 \\
\cmidrule{2-7}
& \multirow{3}{*}{$s_V$} 
  & Simplified & 2.67 & 2.31 & 0.81 & 2.12 \\
& & Detailed & 2.60 & 2.25 & 1.13 & 2.06 \\
\cmidrule{2-7}
& \multirow{3}{*}{$s_O$} 
  & Simplified & 2.22 & 1.53 & 0.98 & 1.60 \\
& & Detailed & 1.81 & 1.34 & 1.13 & 1.40 \\
\cmidrule{2-7}
 & \multirow{3}{*}{$r_Q$} 
  & Simplified & 0.02 & 0.02 & 0.08 & 0.04 \\
& & Detailed & 0.04 & 0.03 & 0.10 & 0.05 \\
\cmidrule{2-7}
& \multirow{3}{*}{$r_K$} 
  & Simplified & 0.03 & 0.03 & 0.08 & 0.04 \\
& & Detailed & 0.05 & 0.03 & 0.10 & 0.05 \\
\cmidrule{2-7}
 & \multirow{3}{*}{$r_V$} 
  & Simplified & 0.03 & 0.04 & 0.06 & 0.04 \\
& & Detailed & 0.02 & 0.05 & 0.05 & 0.04 \\
\cmidrule{2-7}
& \multirow{3}{*}{$r_O$} 
  & Simplified & 0.01 & 0.02 & 0.03 & 0.02 \\
& & Detailed & 0.01 & 0.02 & 0.04 & 0.02 \\
\bottomrule
\end{tabular}
}
\caption{
Statistical results for MATH-Geometry using gemma-2-2b-it on irrelevant responses.
}
\label{tbl:math_lighteval_math_wrong_answer_shuffle_gemma2_2b_it_grads_geometry}
\end{table}

\begin{figure}[!h]
\centering
\includegraphics[width=\linewidth]{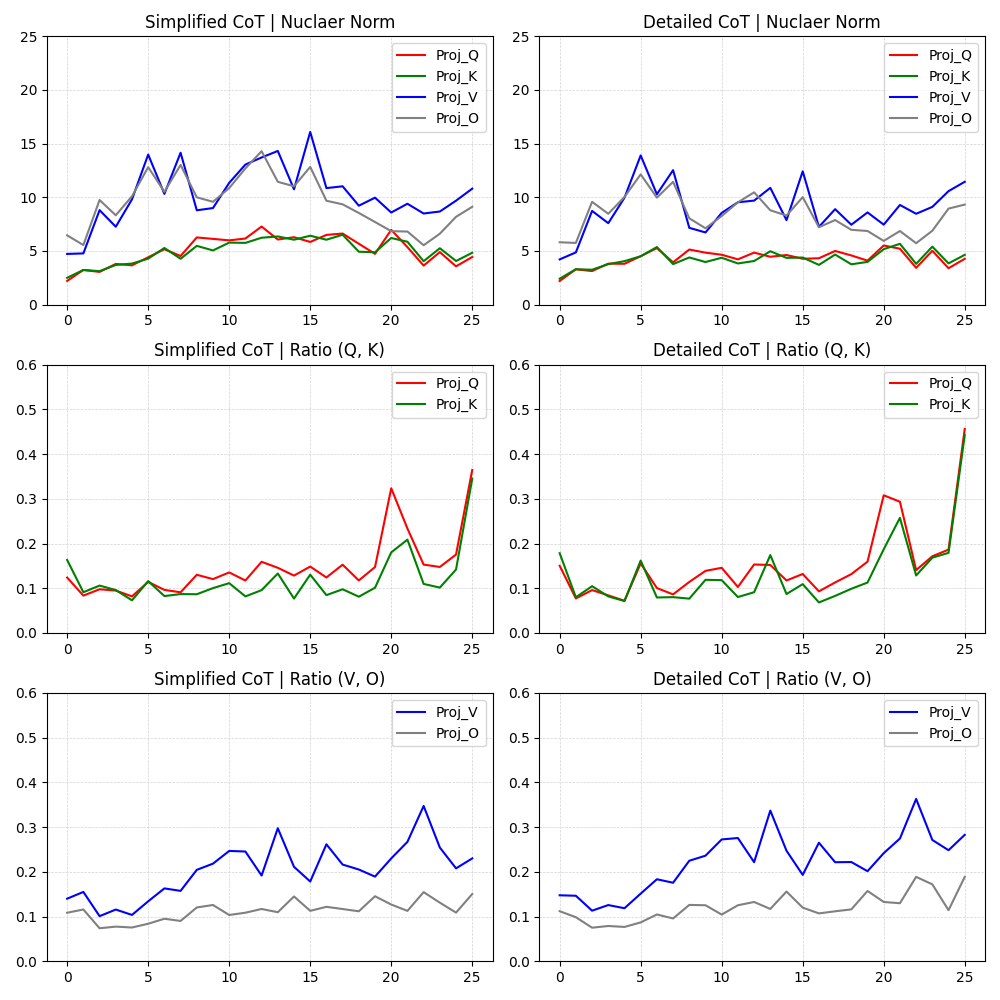}
\caption{Visualization for MATH-Geometry using gemma-2-2b-it on irrelevant responses.}
\label{fig:math_nonsense_geometry_it}
\vspace{-2mm}
\end{figure}

The visualizations and statistical results on other reasoning tasks: AQuA (Figure \ref{fig:reasoning_nonsense_aqua_it}, Table \ref{tbl:reasoning_cot_500_wrong_answer_shuffle_gemma2_2b_it_grads_aqua_train}), GSM8K (Figure \ref{fig:reasoning_nonsense_gsm8k_it}, Table \ref{tbl:reasoning_cot_500_wrong_answer_shuffle_gemma2_2b_it_grads_gsm8k_train}), StrategyQA (Figure \ref{fig:reasoning_nonsense_strategyqa_it}, Table \ref{tbl:reasoning_cot_500_wrong_answer_shuffle_gemma2_2b_it_grads_strategyqa_train}), ECQA (Figure \ref{fig:reasoning_nonsense_ecqa_it}, Table \ref{tbl:reasoning_cot_500_wrong_answer_shuffle_gemma2_2b_it_grads_ecqa_train}), CREAK (Figure \ref{fig:reasoning_nonsense_creak_it}, Table \ref{tbl:reasoning_cot_500_wrong_answer_shuffle_gemma2_2b_it_grads_creak_train}), Sensemaking (Figure \ref{fig:reasoning_nonsense_sensemaking_it}, Table \ref{tbl:reasoning_cot_500_wrong_answer_shuffle_gemma2_2b_it_grads_sensemaking_train}).

\begin{table}[!tbh]
\centering
\small
\scalebox{0.75}{
\begin{tabular}{l|l|l|cccc}
\toprule
\multirow{2}{*}{\textbf{Dataset}} & \multirow{2}{*}{\textbf{Curve}} & \multirow{2}{*}{\textbf{Cot}} &  \multicolumn{4}{c}{\textbf{Mean Absolute Difference (MAD)}} \\
 & &  & Early & Middle & Last & All\\
\midrule
\multirow{24}{*}{AQuA} & \multirow{3}{*}{$s_Q$} 
  & None & 3.97 & 5.15 & 12.37 & 6.55 \\
& & Simplified & 1.22 & 1.15 & 2.35 & 1.49 \\
& & Detailed & 0.69 & 0.58 & 1.15 & 0.80 \\
\cmidrule{2-7}
& \multirow{3}{*}{$s_K$} 
  & None & 3.84 & 8.17 & 20.16 & 9.03 \\
& & Simplified & 1.23 & 1.49 & 2.02 & 1.56 \\
& & Detailed & 0.62 & 0.86 & 0.76 & 0.81 \\
\cmidrule{2-7}
& \multirow{3}{*}{$s_V$} 
  & None & 17.02 & 24.77 & 8.19 & 19.12 \\
& & Simplified & 5.30 & 5.01 & 1.99 & 4.45 \\
& & Detailed & 3.11 & 2.46 & 1.02 & 2.34 \\
\cmidrule{2-7}
& \multirow{3}{*}{$s_O$} 
  & None & 12.88 & 11.19 & 5.43 & 10.37 \\
& & Simplified & 4.15 & 2.75 & 1.61 & 2.87 \\
& & Detailed & 2.21 & 1.55 & 1.03 & 1.62 \\
\cmidrule{2-7}
 & \multirow{3}{*}{$r_Q$} 
  & None & 0.03 & 0.05 & 0.10 & 0.05 \\
& & Simplified & 0.03 & 0.02 & 0.08 & 0.04 \\
& & Detailed & 0.04 & 0.03 & 0.09 & 0.05 \\
\cmidrule{2-7}
& \multirow{3}{*}{$r_K$} 
  & None & 0.04 & 0.03 & 0.04 & 0.04 \\
& & Simplified & 0.04 & 0.02 & 0.06 & 0.04 \\
& & Detailed & 0.05 & 0.03 & 0.07 & 0.05 \\
\cmidrule{2-7}
 & \multirow{3}{*}{$r_V$} 
  & None & 0.02 & 0.05 & 0.16 & 0.07 \\
& & Simplified & 0.03 & 0.05 & 0.06 & 0.05 \\
& & Detailed & 0.03 & 0.05 & 0.04 & 0.04 \\
\cmidrule{2-7}
& \multirow{3}{*}{$r_O$} 
  & None & 0.01 & 0.05 & 0.08 & 0.04 \\
& & Simplified & 0.01 & 0.03 & 0.04 & 0.02 \\
& & Detailed & 0.01 & 0.02 & 0.03 & 0.02 \\
\bottomrule
\end{tabular}
}
\caption{
Statistical results for AQuA using gemma-2-2b-it on irrelevant responses.
}
\label{tbl:reasoning_cot_500_wrong_answer_shuffle_gemma2_2b_it_grads_aqua_train}
\end{table}

\begin{figure*}[!h]
\centering
\includegraphics[width=\linewidth]{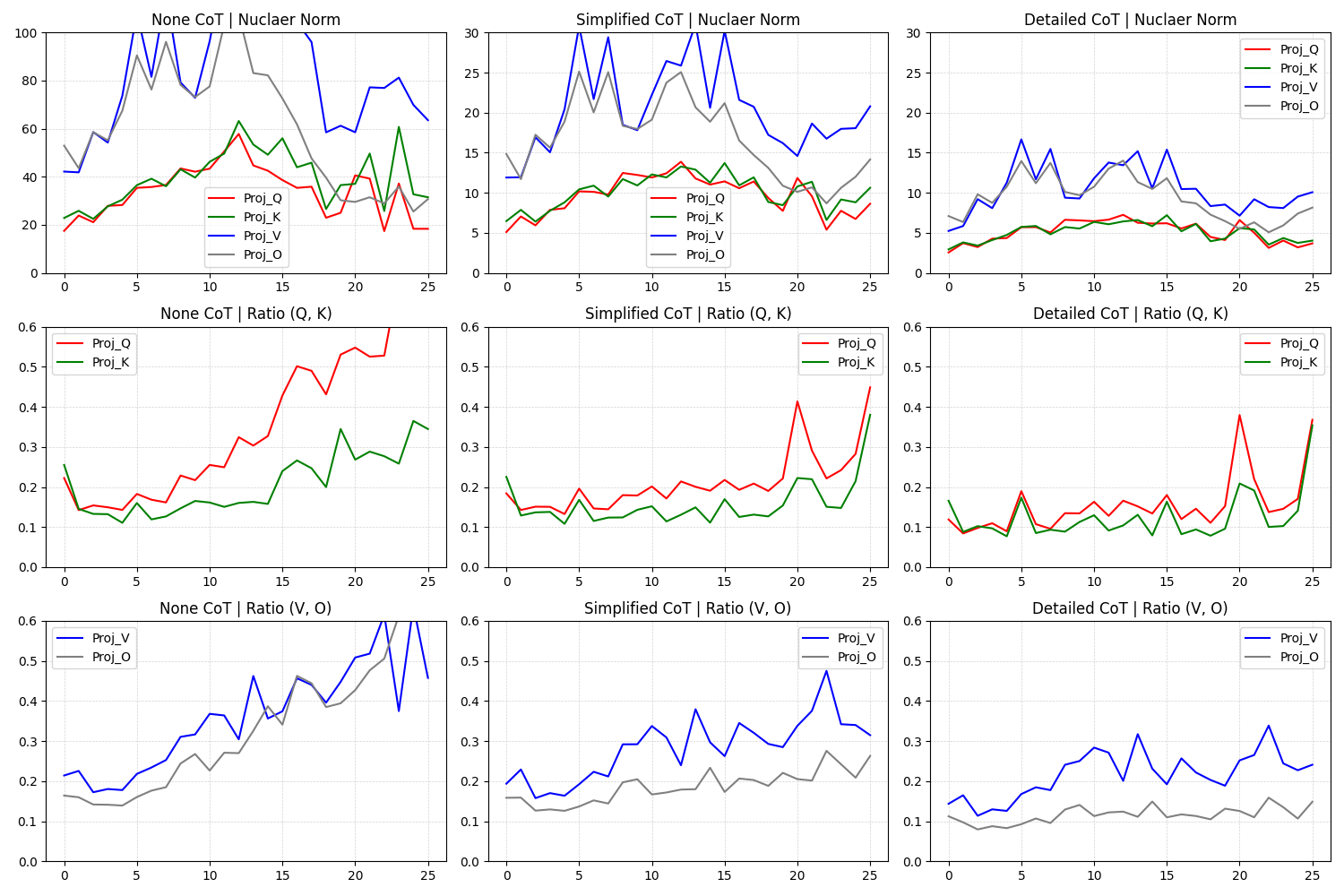}
\caption{Visualization for AQuA using gemma-2-2b-it on irrelevant responses.}
\label{fig:reasoning_nonsense_aqua_it}
\vspace{-2mm}
\end{figure*}

\begin{table}[!tbh]
\centering
\small
\scalebox{0.75}{
\begin{tabular}{l|l|l|cccc}
\toprule
\multirow{2}{*}{\textbf{Dataset}} & \multirow{2}{*}{\textbf{Curve}} & \multirow{2}{*}{\textbf{Cot}} &  \multicolumn{4}{c}{\textbf{Mean Absolute Difference (MAD)}} \\
 & &  & Early & Middle & Last & All\\
\midrule
\multirow{24}{*}{GSM8K} & \multirow{3}{*}{$s_Q$} 
  & None & 4.30 & 6.74 & 11.48 & 7.24 \\
& & Simplified & 0.96 & 0.88 & 1.83 & 1.17 \\
& & Detailed & 0.70 & 0.61 & 1.22 & 0.84 \\
\cmidrule{2-7}
& \multirow{3}{*}{$s_K$} 
  & None & 3.94 & 10.22 & 15.59 & 9.07 \\
& & Simplified & 0.95 & 1.10 & 1.60 & 1.19 \\
& & Detailed & 0.62 & 0.88 & 0.73 & 0.81 \\
\cmidrule{2-7}
& \multirow{3}{*}{$s_V$} 
  & None & 24.31 & 32.27 & 13.78 & 25.89 \\
& & Simplified & 4.61 & 3.78 & 1.43 & 3.47 \\
& & Detailed & 3.06 & 2.26 & 0.98 & 2.20 \\
\cmidrule{2-7}
& \multirow{3}{*}{$s_O$} 
  & None & 18.70 & 13.57 & 10.20 & 14.13 \\
& & Simplified & 3.54 & 2.31 & 1.04 & 2.34 \\
& & Detailed & 2.19 & 1.60 & 0.80 & 1.58 \\
\cmidrule{2-7}
 & \multirow{3}{*}{$r_Q$} 
  & None & 0.03 & 0.04 & 0.06 & 0.04 \\
& & Simplified & 0.03 & 0.02 & 0.07 & 0.04 \\
& & Detailed & 0.04 & 0.03 & 0.10 & 0.06 \\
\cmidrule{2-7}
& \multirow{3}{*}{$r_K$} 
  & None & 0.04 & 0.04 & 0.05 & 0.04 \\
& & Simplified & 0.05 & 0.02 & 0.06 & 0.04 \\
& & Detailed & 0.05 & 0.03 & 0.07 & 0.05 \\
\cmidrule{2-7}
 & \multirow{3}{*}{$r_V$} 
  & None & 0.03 & 0.06 & 0.10 & 0.06 \\
& & Simplified & 0.04 & 0.06 & 0.06 & 0.05 \\
& & Detailed & 0.03 & 0.05 & 0.04 & 0.04 \\
\cmidrule{2-7}
& \multirow{3}{*}{$r_O$} 
  & None & 0.02 & 0.06 & 0.07 & 0.05 \\
& & Simplified & 0.01 & 0.03 & 0.04 & 0.03 \\
& & Detailed & 0.01 & 0.02 & 0.03 & 0.02 \\
\bottomrule
\end{tabular}
}
\caption{
Statistical results for GSM8K using gemma-2-2b-it on irrelevant responses.
}
\label{tbl:reasoning_cot_500_wrong_answer_shuffle_gemma2_2b_it_grads_gsm8k_train}
\end{table}

\begin{figure*}[!h]
\centering
\includegraphics[width=\linewidth]{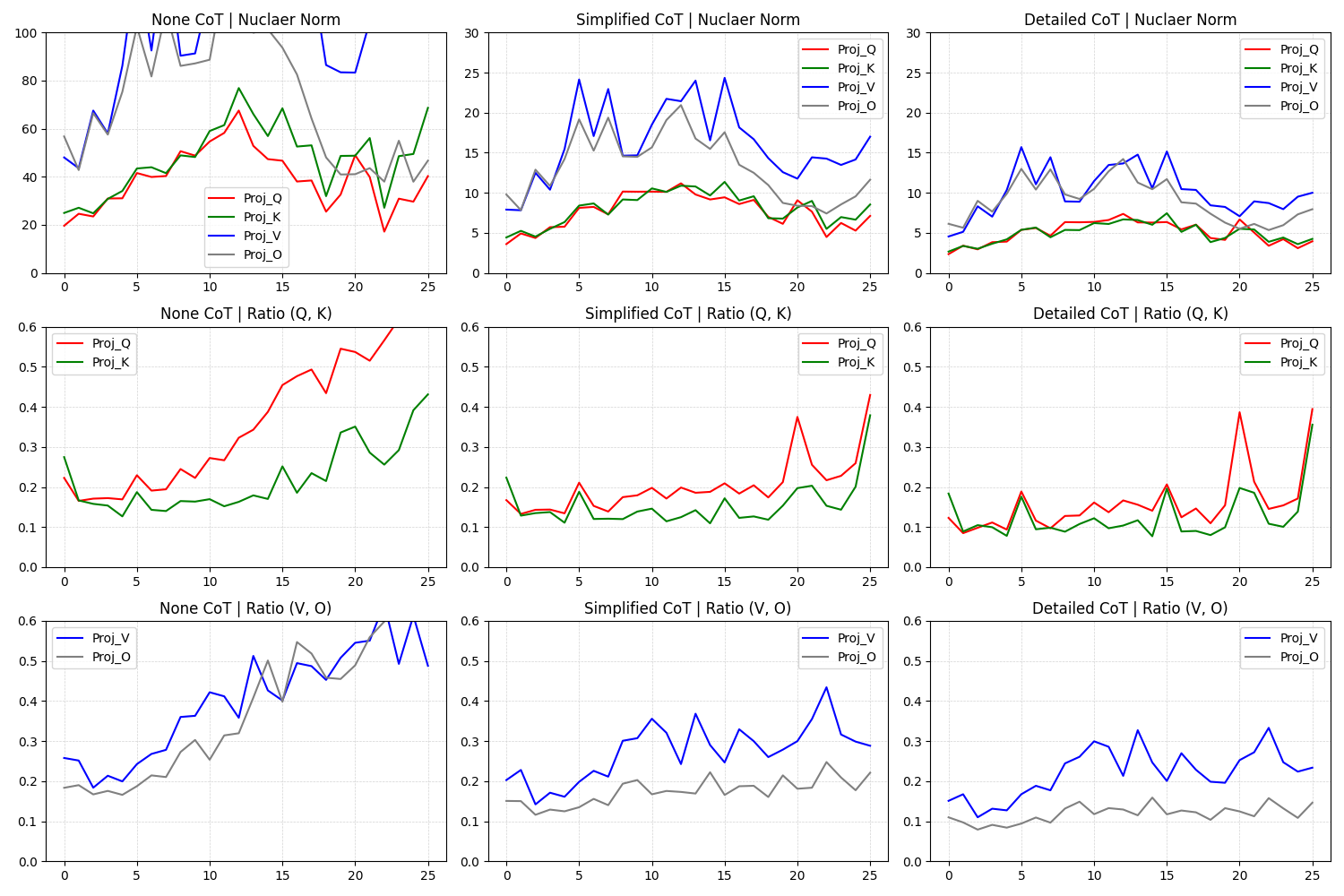}
\caption{Visualization for GSM8K using gemma-2-2b-it on irrelevant responses.}
\label{fig:reasoning_nonsense_gsm8k_it}
\vspace{-2mm}
\end{figure*}

\begin{table}[!tbh]
\centering
\small
\scalebox{0.75}{
\begin{tabular}{l|l|l|cccc}
\toprule
\multirow{2}{*}{\textbf{Dataset}} & \multirow{2}{*}{\textbf{Curve}} & \multirow{2}{*}{\textbf{Cot}} &  \multicolumn{4}{c}{\textbf{Mean Absolute Difference (MAD)}} \\
 & &  & Early & Middle & Last & All\\
\midrule
\multirow{24}{*}{StrategyQA} & \multirow{3}{*}{$s_Q$} 
  & None & 11.79 & 10.10 & 4.95 & 9.63 \\
& & Simplified & 2.20 & 0.96 & 1.81 & 1.51 \\
& & Detailed & 1.26 & 0.44 & 1.09 & 0.79 \\
\cmidrule{2-7}
& \multirow{3}{*}{$s_K$} 
  & None & 9.75 & 12.02 & 9.80 & 11.68 \\
& & Simplified & 1.74 & 1.62 & 1.69 & 1.61 \\
& & Detailed & 0.97 & 0.65 & 0.97 & 0.79 \\
\cmidrule{2-7}
& \multirow{3}{*}{$s_V$} 
  & None & 50.93 & 67.72 & 30.54 & 55.22 \\
& & Simplified & 6.11 & 4.74 & 3.01 & 4.66 \\
& & Detailed & 3.78 & 2.51 & 1.26 & 2.54 \\
\cmidrule{2-7}
& \multirow{3}{*}{$s_O$} 
  & None & 35.52 & 23.60 & 11.71 & 25.04 \\
& & Simplified & 4.47 & 2.37 & 2.34 & 2.99 \\
& & Detailed & 2.71 & 1.70 & 1.47 & 1.93 \\
\cmidrule{2-7}
 & \multirow{3}{*}{$r_Q$} 
  & None & 0.02 & 0.06 & 0.07 & 0.05 \\
& & Simplified & 0.04 & 0.03 & 0.04 & 0.03 \\
& & Detailed & 0.04 & 0.03 & 0.06 & 0.04 \\
\cmidrule{2-7}
& \multirow{3}{*}{$r_K$} 
  & None & 0.03 & 0.02 & 0.05 & 0.03 \\
& & Simplified & 0.04 & 0.03 & 0.04 & 0.03 \\
& & Detailed & 0.05 & 0.03 & 0.06 & 0.04 \\
\cmidrule{2-7}
 & \multirow{3}{*}{$r_V$} 
  & None & 0.05 & 0.05 & 0.07 & 0.05 \\
& & Simplified & 0.05 & 0.07 & 0.08 & 0.06 \\
& & Detailed & 0.03 & 0.05 & 0.04 & 0.04 \\
\cmidrule{2-7}
& \multirow{3}{*}{$r_O$} 
  & None & 0.03 & 0.06 & 0.07 & 0.05 \\
& & Simplified & 0.02 & 0.03 & 0.05 & 0.03 \\
& & Detailed & 0.02 & 0.02 & 0.02 & 0.02 \\
\bottomrule
\end{tabular}
}
\caption{
Statistical results for StrategyQA using gemma-2-2b-it on irrelevant responses.
}
\label{tbl:reasoning_cot_500_wrong_answer_shuffle_gemma2_2b_it_grads_strategyqa_train}
\end{table}

\begin{figure*}[!h]
\centering
\includegraphics[width=\linewidth]{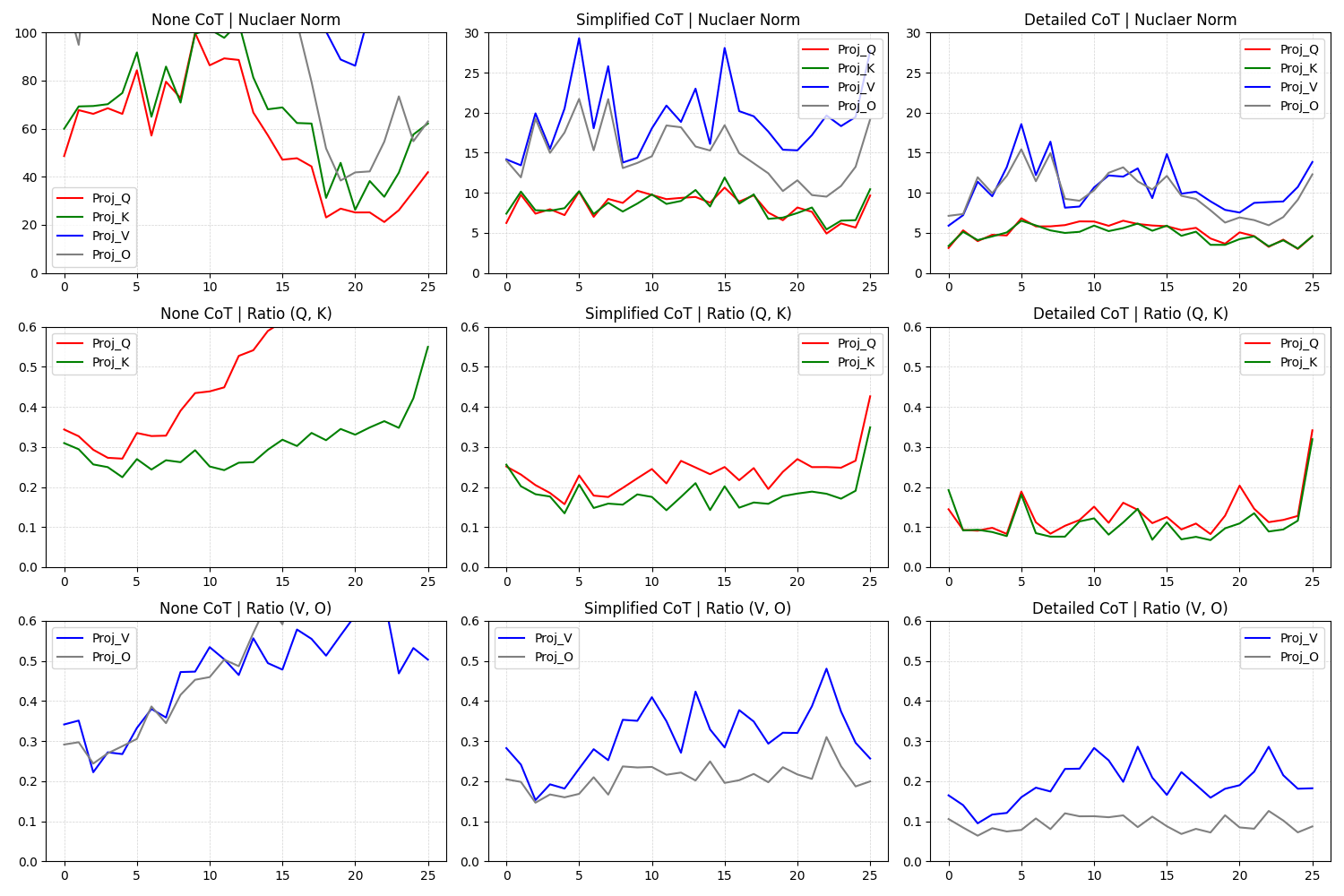}
\caption{Visualization for StrategyQA using gemma-2-2b-it on irrelevant responses.}
\label{fig:reasoning_nonsense_strategyqa_it}
\vspace{-2mm}
\end{figure*}

\begin{table}[!tbh]
\centering
\small
\scalebox{0.75}{
\begin{tabular}{l|l|l|cccc}
\toprule
\multirow{2}{*}{\textbf{Dataset}} & \multirow{2}{*}{\textbf{Curve}} & \multirow{2}{*}{\textbf{Cot}} &  \multicolumn{4}{c}{\textbf{Mean Absolute Difference (MAD)}} \\
 & &  & Early & Middle & Last & All\\
\midrule
\multirow{24}{*}{ECQA} & \multirow{3}{*}{$s_Q$} 
  & None & 6.22 & 8.86 & 12.56 & 8.62 \\
& & Simplified & 1.84 & 1.87 & 2.27 & 2.01 \\
& & Detailed & 1.07 & 0.83 & 1.28 & 1.03 \\
\cmidrule{2-7}
& \multirow{3}{*}{$s_K$} 
  & None & 5.62 & 14.70 & 12.56 & 11.71 \\
& & Simplified & 1.55 & 3.28 & 1.82 & 2.39 \\
& & Detailed & 0.77 & 1.06 & 0.97 & 0.99 \\
\cmidrule{2-7}
& \multirow{3}{*}{$s_V$} 
  & None & 32.47 & 32.21 & 31.67 & 31.90 \\
& & Simplified & 7.71 & 6.52 & 4.89 & 6.46 \\
& & Detailed & 4.24 & 2.95 & 1.67 & 3.00 \\
\cmidrule{2-7}
& \multirow{3}{*}{$s_O$} 
  & None & 24.35 & 13.34 & 7.35 & 15.24 \\
& & Simplified & 5.84 & 3.09 & 1.52 & 3.52 \\
& & Detailed & 3.00 & 1.92 & 1.12 & 2.02 \\
\cmidrule{2-7}
 & \multirow{3}{*}{$r_Q$} 
  & None & 0.02 & 0.05 & 0.09 & 0.05 \\
& & Simplified & 0.02 & 0.04 & 0.08 & 0.04 \\
& & Detailed & 0.04 & 0.04 & 0.09 & 0.05 \\
\cmidrule{2-7}
& \multirow{3}{*}{$r_K$} 
  & None & 0.03 & 0.04 & 0.06 & 0.04 \\
& & Simplified & 0.04 & 0.04 & 0.06 & 0.04 \\
& & Detailed & 0.05 & 0.04 & 0.08 & 0.05 \\
\cmidrule{2-7}
 & \multirow{3}{*}{$r_V$} 
  & None & 0.04 & 0.06 & 0.07 & 0.05 \\
& & Simplified & 0.04 & 0.07 & 0.05 & 0.05 \\
& & Detailed & 0.02 & 0.05 & 0.04 & 0.04 \\
\cmidrule{2-7}
& \multirow{3}{*}{$r_O$} 
  & None & 0.02 & 0.06 & 0.06 & 0.05 \\
& & Simplified & 0.02 & 0.03 & 0.05 & 0.03 \\
& & Detailed & 0.02 & 0.02 & 0.02 & 0.02 \\
\bottomrule
\end{tabular}
}
\caption{
Statistical results for ECQA using gemma-2-2b-it on irrelevant responses.
}
\label{tbl:reasoning_cot_500_wrong_answer_shuffle_gemma2_2b_it_grads_ecqa_train}
\end{table}

\begin{figure*}[!h]
\centering
\includegraphics[width=\linewidth]{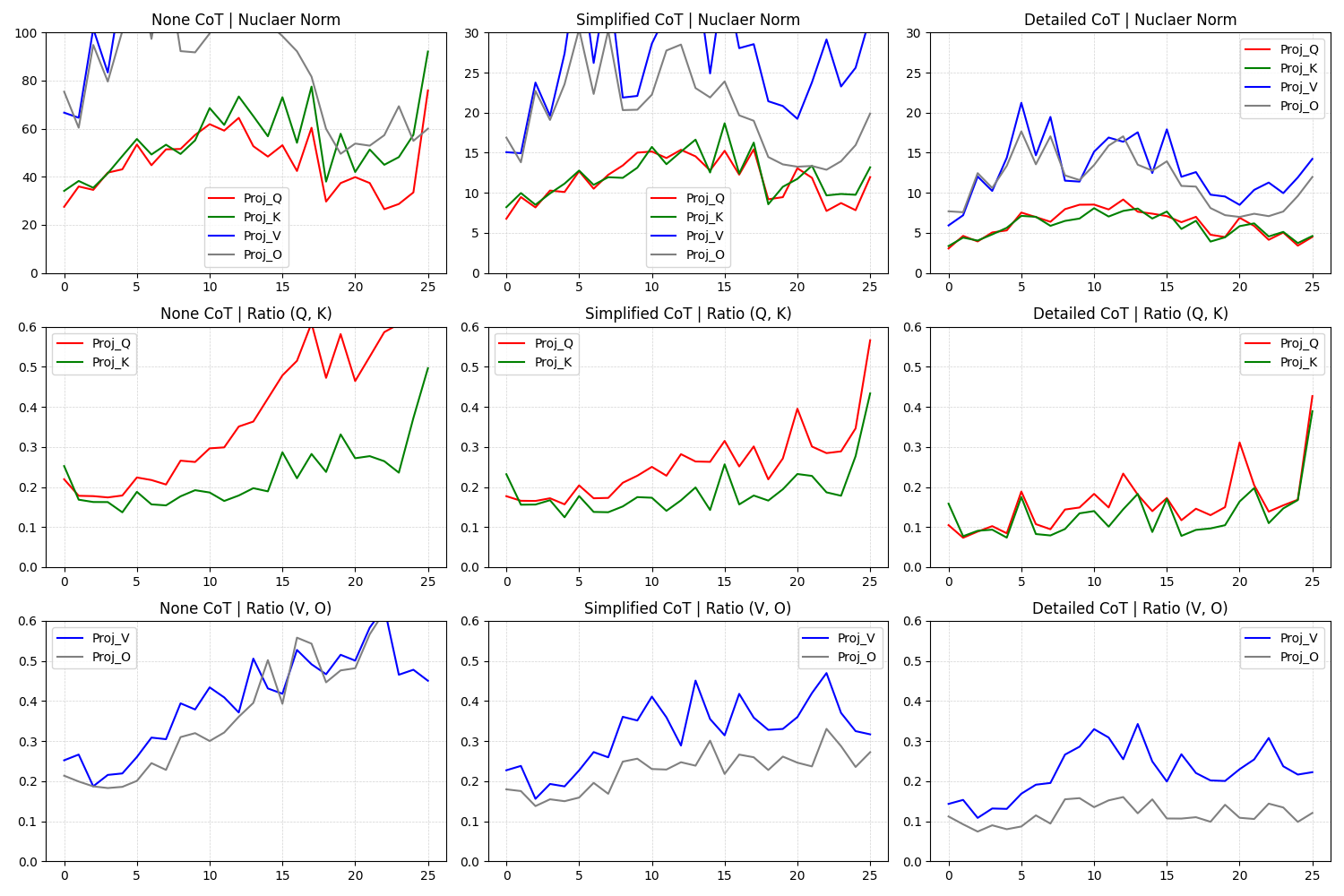}
\caption{Visualization for ECQA using gemma-2-2b-it on irrelevant responses.}
\label{fig:reasoning_nonsense_ecqa_it}
\vspace{-2mm}
\end{figure*}

\begin{table}[!tbh]
\centering
\small
\scalebox{0.75}{
\begin{tabular}{l|l|l|cccc}
\toprule
\multirow{2}{*}{\textbf{Dataset}} & \multirow{2}{*}{\textbf{Curve}} & \multirow{2}{*}{\textbf{Cot}} &  \multicolumn{4}{c}{\textbf{Mean Absolute Difference (MAD)}} \\
 & &  & Early & Middle & Last & All\\
\midrule
\multirow{24}{*}{CREAK} & \multirow{3}{*}{$s_Q$} 
  & None & 20.78 & 22.32 & 16.61 & 20.27 \\
& & Simplified & 2.42 & 2.21 & 2.19 & 2.30 \\
& & Detailed & 1.10 & 0.74 & 1.49 & 1.03 \\
\cmidrule{2-7}
& \multirow{3}{*}{$s_K$} 
  & None & 16.37 & 29.50 & 25.20 & 26.23 \\
& & Simplified & 1.90 & 3.55 & 2.29 & 2.70 \\
& & Detailed & 0.76 & 1.04 & 1.20 & 1.02 \\
\cmidrule{2-7}
& \multirow{3}{*}{$s_V$} 
  & None & 99.45 & 112.28 & 26.10 & 92.60 \\
& & Simplified & 8.94 & 8.00 & 4.34 & 7.52 \\
& & Detailed & 3.94 & 3.03 & 1.37 & 2.88 \\
\cmidrule{2-7}
& \multirow{3}{*}{$s_O$} 
  & None & 70.14 & 39.66 & 10.86 & 43.13 \\
& & Simplified & 6.66 & 3.32 & 1.95 & 4.05 \\
& & Detailed & 2.83 & 1.93 & 1.32 & 2.03 \\
\cmidrule{2-7}
 & \multirow{3}{*}{$r_Q$} 
  & None & 0.02 & 0.07 & 0.09 & 0.07 \\
& & Simplified & 0.03 & 0.04 & 0.08 & 0.05 \\
& & Detailed & 0.03 & 0.03 & 0.07 & 0.04 \\
\cmidrule{2-7}
& \multirow{3}{*}{$r_K$} 
  & None & 0.03 & 0.03 & 0.07 & 0.04 \\
& & Simplified & 0.04 & 0.04 & 0.06 & 0.04 \\
& & Detailed & 0.04 & 0.04 & 0.06 & 0.04 \\
\cmidrule{2-7}
 & \multirow{3}{*}{$r_V$} 
  & None & 0.03 & 0.05 & 0.07 & 0.05 \\
& & Simplified & 0.04 & 0.06 & 0.06 & 0.05 \\
& & Detailed & 0.02 & 0.05 & 0.04 & 0.04 \\
\cmidrule{2-7}
& \multirow{3}{*}{$r_O$} 
  & None & 0.02 & 0.06 & 0.08 & 0.05 \\
& & Simplified & 0.02 & 0.04 & 0.04 & 0.03 \\
& & Detailed & 0.02 & 0.02 & 0.02 & 0.02 \\
\bottomrule
\end{tabular}
}
\caption{
Statistical results for CREAK using gemma-2-2b-it on irrelevant responses.
}
\label{tbl:reasoning_cot_500_wrong_answer_shuffle_gemma2_2b_it_grads_creak_train}
\end{table}

\begin{figure*}[!h]
\centering
\includegraphics[width=\linewidth]{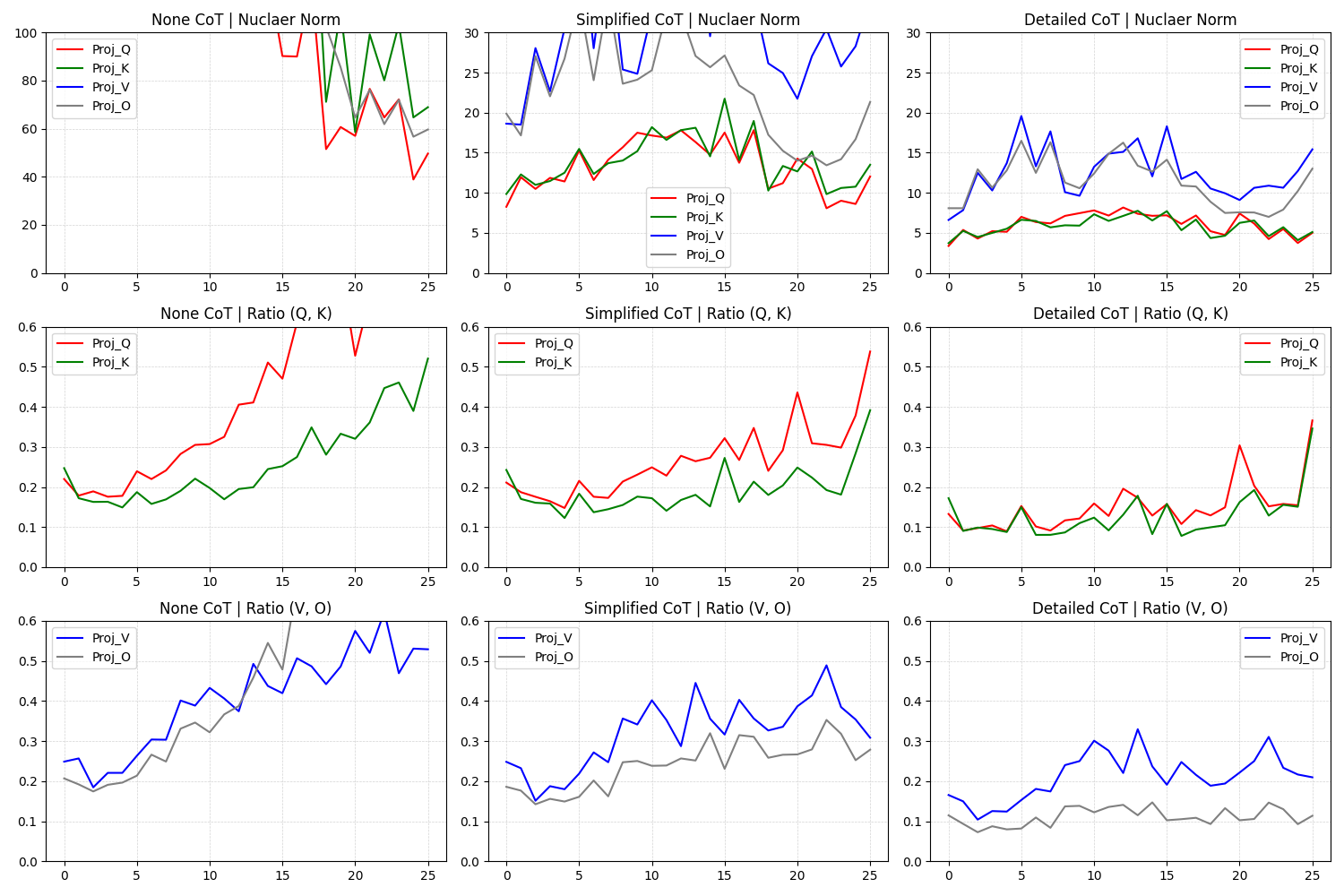}
\caption{Visualization for CREAK using gemma-2-2b-it on irrelevant responses.}
\label{fig:reasoning_nonsense_creak_it}
\vspace{-2mm}
\end{figure*}

\begin{table}[!tbh]
\centering
\small
\scalebox{0.75}{
\begin{tabular}{l|l|l|cccc}
\toprule
\multirow{2}{*}{\textbf{Dataset}} & \multirow{2}{*}{\textbf{Curve}} & \multirow{2}{*}{\textbf{Cot}} &  \multicolumn{4}{c}{\textbf{Mean Absolute Difference (MAD)}} \\
 & &  & Early & Middle & Last & All\\
\midrule
\multirow{24}{*}{Sensemaking} & \multirow{3}{*}{$s_Q$} 
  & None & 7.68 & 13.57 & 12.50 & 11.79 \\
& & Simplified & 2.20 & 1.91 & 1.91 & 2.01 \\
& & Detailed & 0.84 & 0.90 & 1.29 & 1.05 \\
\cmidrule{2-7}
& \multirow{3}{*}{$s_K$} 
  & None & 7.41 & 17.48 & 14.33 & 13.10 \\
& & Simplified & 1.79 & 3.61 & 2.98 & 2.79 \\
& & Detailed & 0.74 & 1.21 & 0.88 & 1.05 \\
\cmidrule{2-7}
& \multirow{3}{*}{$s_V$} 
  & None & 41.81 & 46.33 & 10.97 & 38.74 \\
& & Simplified & 9.20 & 7.94 & 5.11 & 7.75 \\
& & Detailed & 3.45 & 2.89 & 1.55 & 2.74 \\
\cmidrule{2-7}
& \multirow{3}{*}{$s_O$} 
  & None & 29.03 & 21.86 & 3.94 & 20.63 \\
& & Simplified & 6.87 & 3.71 & 1.61 & 4.22 \\
& & Detailed & 2.47 & 1.96 & 0.92 & 1.87 \\
\cmidrule{2-7}
 & \multirow{3}{*}{$r_Q$} 
  & None & 0.02 & 0.06 & 0.08 & 0.06 \\
& & Simplified & 0.02 & 0.03 & 0.07 & 0.04 \\
& & Detailed & 0.03 & 0.04 & 0.10 & 0.06 \\
\cmidrule{2-7}
& \multirow{3}{*}{$r_K$} 
  & None & 0.04 & 0.02 & 0.04 & 0.03 \\
& & Simplified & 0.04 & 0.03 & 0.06 & 0.04 \\
& & Detailed & 0.04 & 0.04 & 0.07 & 0.05 \\
\cmidrule{2-7}
 & \multirow{3}{*}{$r_V$} 
  & None & 0.03 & 0.05 & 0.06 & 0.05 \\
& & Simplified & 0.03 & 0.06 & 0.04 & 0.05 \\
& & Detailed & 0.02 & 0.05 & 0.03 & 0.04 \\
\cmidrule{2-7}
& \multirow{3}{*}{$r_O$} 
  & None & 0.02 & 0.07 & 0.08 & 0.06 \\
& & Simplified & 0.02 & 0.04 & 0.04 & 0.03 \\
& & Detailed & 0.01 & 0.02 & 0.02 & 0.02 \\
\bottomrule
\end{tabular}
}
\caption{
Statistical results for Sensemaking using gemma-2-2b-it on irrelevant responses.
}
\label{tbl:reasoning_cot_500_wrong_answer_shuffle_gemma2_2b_it_grads_sensemaking_train}
\end{table}

\begin{figure*}[!h]
\centering
\includegraphics[width=\linewidth]{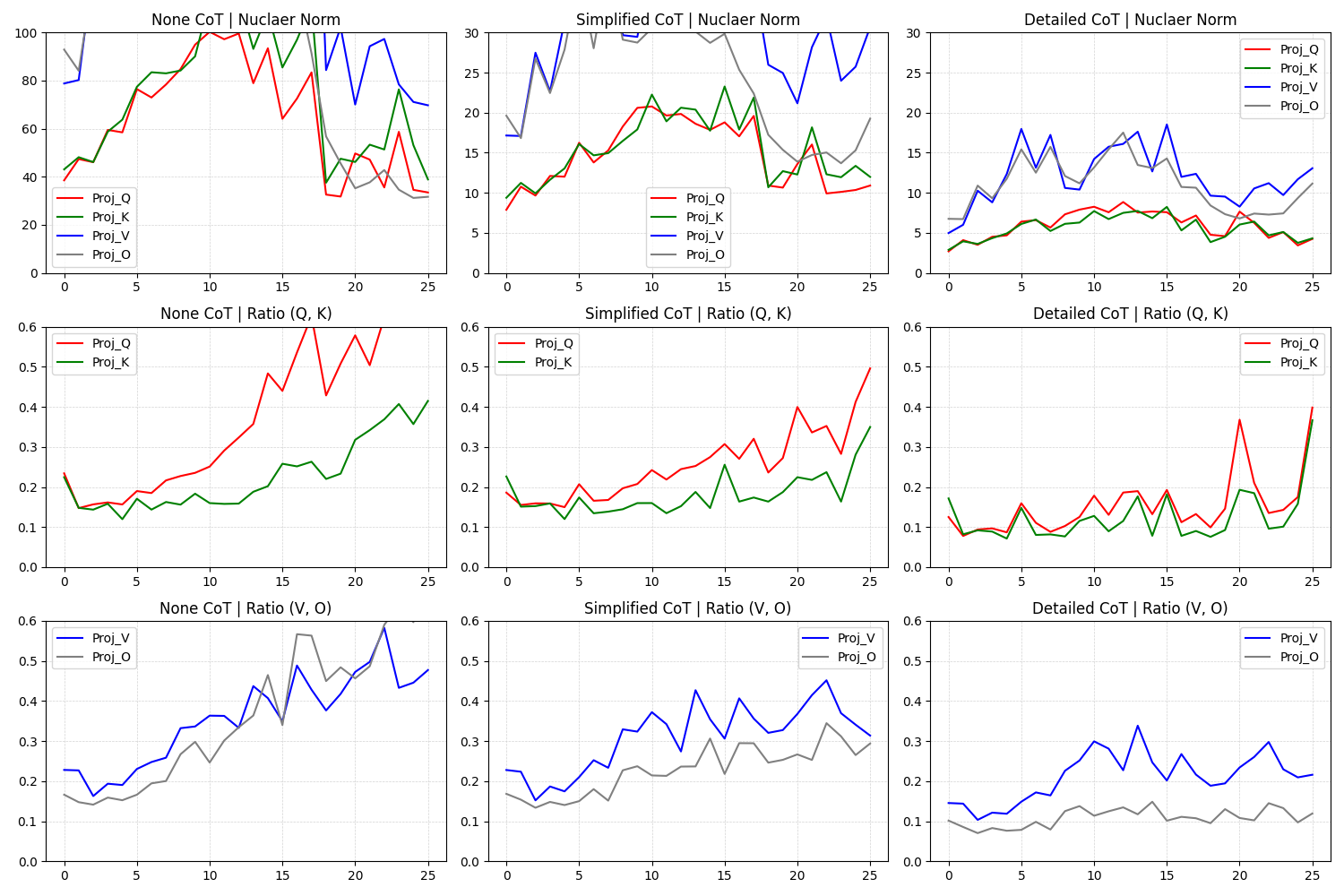}
\caption{Visualization for Sensemaking using gemma-2-2b-it on irrelevant responses.}
\label{fig:reasoning_nonsense_sensemaking_it}
\vspace{-2mm}
\end{figure*}

\subsubsection{Wiki Tasks}

The visualizations and statistical results on Wiki tasks are shown in Figure \ref{fig:wiki_nonsense_it} and Table \ref{tbl:wiki_wiki_popularity_wrong_answer_shuffle_gemma2_2b_it_grads}.

\begin{table}[!tbh]
\centering
\small
\scalebox{0.75}{
\begin{tabular}{l|l|l|cccc}
\toprule
\multirow{2}{*}{\textbf{Dataset}} & \multirow{2}{*}{\textbf{Curve}} & \multirow{2}{*}{\textbf{Cot}} &  \multicolumn{4}{c}{\textbf{Mean Absolute Difference (MAD)}} \\
 & &  & Early & Middle & Last & All\\
\midrule
\multirow{32}{*}{Wiki} & \multirow{3}{*}{$s_Q$} 
  & Len 100 & 1.97 & 0.93 & 1.44 & 1.32 \\
& & Len 500 & 1.36 & 0.52 & 0.91 & 0.81 \\
& & Len 1000 & 1.15 & 0.43 & 0.78 & 0.68 \\
& & Unpopular & 1.96 & 1.04 & 1.41 & 1.33 \\
\cmidrule{2-7}
& \multirow{3}{*}{$s_K$} 
  & Len 100 & 1.63 & 1.36 & 1.33 & 1.38 \\
& & Len 500 & 0.99 & 0.72 & 0.79 & 0.76 \\
& & Len 1000 & 0.80 & 0.55 & 0.67 & 0.60 \\
& & Unpopular & 1.56 & 1.84 & 1.29 & 1.54 \\
\cmidrule{2-7}
& \multirow{3}{*}{$s_V$} 
  & Len 100 & 4.96 & 3.75 & 1.90 & 3.62 \\
& & Len 500 & 3.73 & 2.66 & 1.56 & 2.64 \\
& & Len 1000 & 3.17 & 2.29 & 1.43 & 2.29 \\
& & Unpopular & 5.69 & 4.80 & 2.96 & 4.53 \\
\cmidrule{2-7}
& \multirow{3}{*}{$s_O$} 
  & Len 100 & 3.77 & 2.09 & 2.69 & 2.68 \\
& & Len 500 & 2.84 & 1.68 & 2.23 & 2.10 \\
& & Len 1000 & 2.57 & 1.46 & 2.00 & 1.87 \\
& & Unpopular & 4.17 & 2.31 & 3.15 & 2.93 \\
\cmidrule{2-7}
 & \multirow{3}{*}{$r_Q$} 
  & Len 100 & 0.04 & 0.03 & 0.03 & 0.03 \\
& & Len 500 & 0.04 & 0.03 & 0.04 & 0.03 \\
& & Len 1000 & 0.03 & 0.02 & 0.04 & 0.03 \\
& & Unpopular & 0.02 & 0.03 & 0.03 & 0.03 \\
\cmidrule{2-7}
& \multirow{3}{*}{$r_K$} 
  & Len 100 & 0.05 & 0.02 & 0.04 & 0.03 \\
& & Len 500 & 0.05 & 0.03 & 0.04 & 0.03 \\
& & Len 1000 & 0.05 & 0.03 & 0.05 & 0.04 \\
& & Unpopular & 0.03 & 0.03 & 0.04 & 0.03 \\
\cmidrule{2-7}
 & \multirow{3}{*}{$r_V$} 
  & Len 100 & 0.05 & 0.07 & 0.08 & 0.06 \\
& & Len 500 & 0.03 & 0.05 & 0.06 & 0.05 \\
& & Len 1000 & 0.03 & 0.05 & 0.06 & 0.04 \\
& & Unpopular & 0.05 & 0.07 & 0.09 & 0.06 \\
\cmidrule{2-7}
& \multirow{3}{*}{$r_O$} 
  & Len 100 & 0.02 & 0.02 & 0.06 & 0.03 \\
& & Len 500 & 0.02 & 0.02 & 0.03 & 0.02 \\
& & Len 1000 & 0.02 & 0.02 & 0.03 & 0.02 \\
& & Unpopular & 0.03 & 0.03 & 0.05 & 0.03 \\
\bottomrule
\end{tabular}
}
\caption{
Statistical results for Wiki using gemma-2-2b-it on irrelevant responses.
}
\label{tbl:wiki_wiki_popularity_wrong_answer_shuffle_gemma2_2b_it_grads}
\end{table}

\begin{figure*}[!h]
\centering
\includegraphics[width=\linewidth]{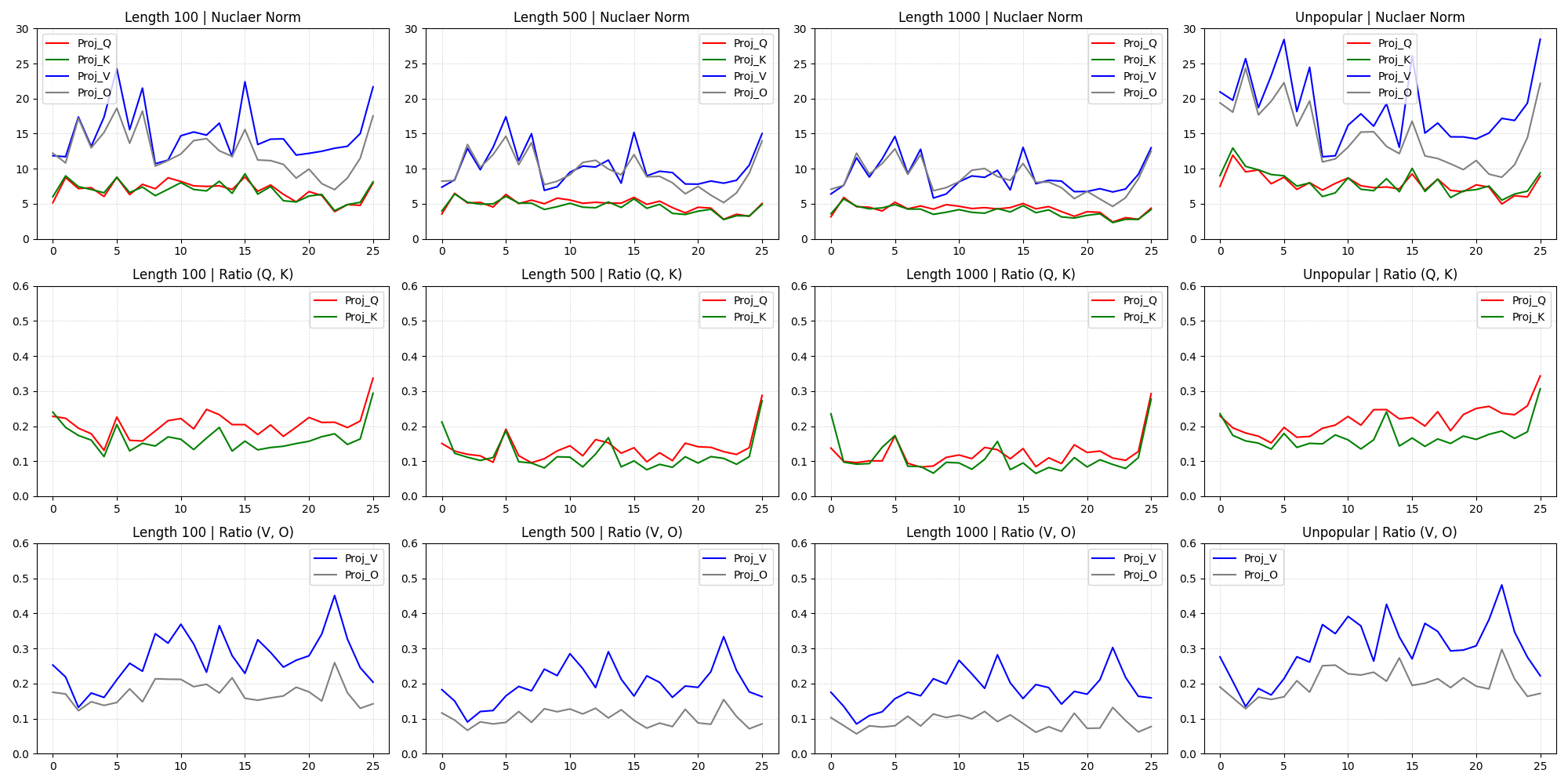}
\caption{Visualization for Wiki tasks using gemma-2-2b-it on irrelevant responses.}
\label{fig:wiki_nonsense_it}
\vspace{-2mm}
\end{figure*}

\clearpage
\section{Results on Llama-3.1-8B}
 \subsection{Pre-trained LLM on Correct Responses}

\subsubsection{Reasoning Tasks}

The visualizations and statistical results on MATH tasks: MATH-Algebra (Figure \ref{fig:math_correct_algebra}, Table \ref{tbl:math_lighteval_math_llama3_8b_grads_algebra}), MATH-Counting (Figure \ref{fig:math_correct_counting}, Table \ref{tbl:math_lighteval_math_llama3_8b_grads_counting_probability}), MATH-Geometry (Figure \ref{fig:math_correct_geometry}, Table \ref{tbl:math_lighteval_math_llama3_8b_grads_geometry}).

\begin{table}[!tbh]
\centering
\small
\scalebox{0.75}{
\begin{tabular}{l|l|l|cccc}
\toprule
\multirow{2}{*}{\textbf{Dataset}} & \multirow{2}{*}{\textbf{Curve}} & \multirow{2}{*}{\textbf{Cot}} &  \multicolumn{4}{c}{\textbf{Mean Absolute Difference (MAD)}} \\
 & &  & Early & Middle & Last & All\\
\midrule
\multirow{16}{*}{Algebra} & \multirow{3}{*}{$s_Q$} 
  & Simplified & 0.76 & 0.76 & 0.35 & 0.58 \\
& & Detailed & 0.42 & 0.37 & 0.17 & 0.29 \\
\cmidrule{2-7}
& \multirow{3}{*}{$s_K$} 
  & Simplified & 0.28 & 0.28 & 0.12 & 0.22 \\
& & Detailed & 0.17 & 0.10 & 0.07 & 0.11 \\
\cmidrule{2-7}
& \multirow{3}{*}{$s_V$} 
  & Simplified & 2.10 & 1.24 & 0.51 & 1.20 \\
& & Detailed & 0.87 & 0.56 & 0.27 & 0.53 \\
\cmidrule{2-7}
& \multirow{3}{*}{$s_O$} 
  & Simplified & 1.06 & 1.31 & 0.61 & 0.93 \\
& & Detailed & 0.61 & 0.70 & 0.33 & 0.51 \\
\cmidrule{2-7}
 & \multirow{3}{*}{$r_Q$} 
  & Simplified & 0.01 & 0.01 & 0.04 & 0.02 \\
& & Detailed & 0.01 & 0.01 & 0.04 & 0.02 \\
\cmidrule{2-7}
& \multirow{3}{*}{$r_K$} 
  & Simplified & 0.02 & 0.01 & 0.01 & 0.01 \\
& & Detailed & 0.02 & 0.01 & 0.02 & 0.01 \\
\cmidrule{2-7}
 & \multirow{3}{*}{$r_V$} 
  & Simplified & 0.02 & 0.01 & 0.02 & 0.02 \\
& & Detailed & 0.01 & 0.01 & 0.02 & 0.02 \\
\cmidrule{2-7}
& \multirow{3}{*}{$r_O$} 
  & Simplified & 0.01 & 0.01 & 0.01 & 0.01 \\
& & Detailed & 0.01 & 0.01 & 0.01 & 0.01 \\
\bottomrule
\end{tabular}
}
\caption{
Statistical results for MATH-Algebra using Llama-3.1-8B on correct responses.
}
\label{tbl:math_lighteval_math_llama3_8b_grads_algebra}
\end{table}

\begin{figure}[!h]
\centering
\includegraphics[width=\linewidth]{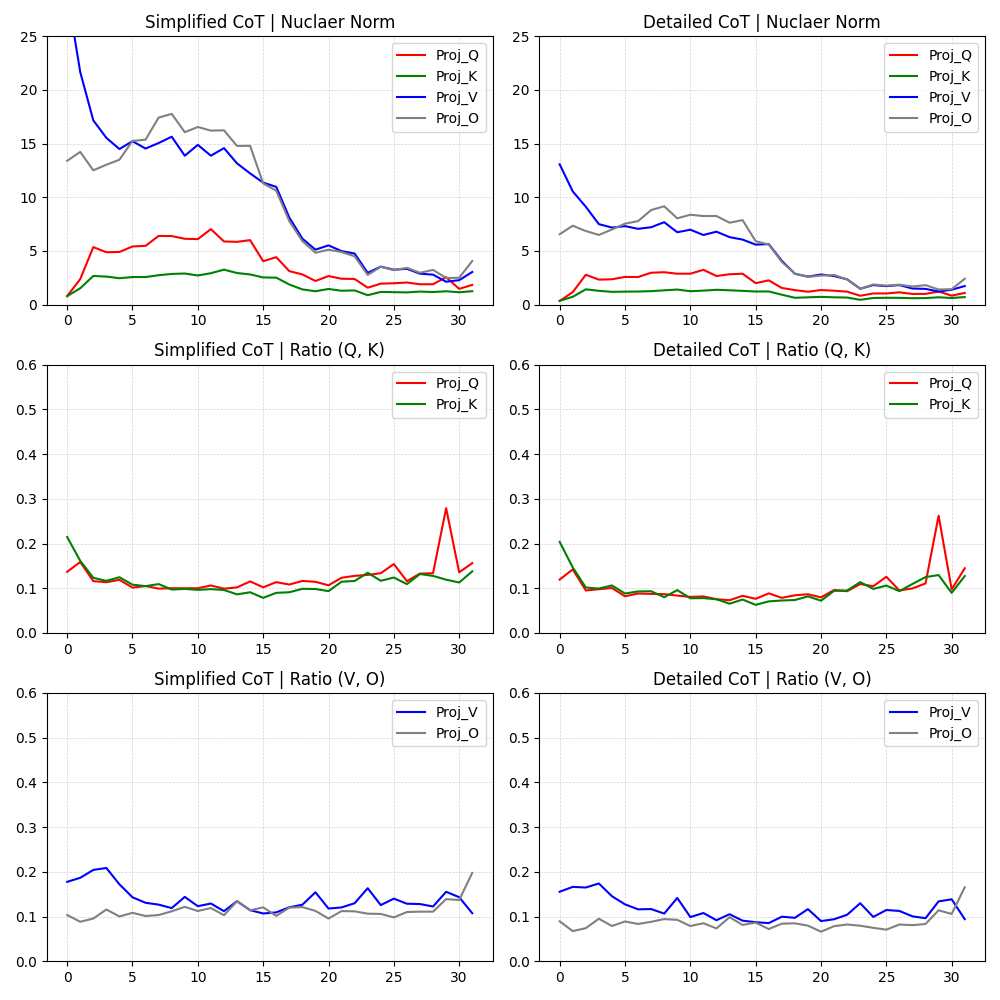}
\caption{Visualization for MATH-Algebra using Llama-3.1-8B on correct responses.}
\label{fig:math_correct_algebra}
\vspace{-2mm}
\end{figure}

\begin{table}[!tbh]
\centering
\small
\scalebox{0.75}{
\begin{tabular}{l|l|l|cccc}
\toprule
\multirow{2}{*}{\textbf{Dataset}} & \multirow{2}{*}{\textbf{Curve}} & \multirow{2}{*}{\textbf{Cot}} &  \multicolumn{4}{c}{\textbf{Mean Absolute Difference (MAD)}} \\
 & &  & Early & Middle & Last & All\\
\midrule
\multirow{16}{*}{Counting} & \multirow{3}{*}{$s_Q$} 
  & Simplified & 1.12 & 0.94 & 0.56 & 0.82 \\
& & Detailed & 0.54 & 0.42 & 0.24 & 0.37 \\
\cmidrule{2-7}
& \multirow{3}{*}{$s_K$} 
  & Simplified & 0.40 & 0.37 & 0.17 & 0.30 \\
& & Detailed & 0.22 & 0.12 & 0.08 & 0.14 \\
\cmidrule{2-7}
& \multirow{3}{*}{$s_V$} 
  & Simplified & 2.91 & 1.49 & 0.57 & 1.52 \\
& & Detailed & 1.11 & 0.62 & 0.35 & 0.63 \\
\cmidrule{2-7}
& \multirow{3}{*}{$s_O$} 
  & Simplified & 1.41 & 1.62 & 0.72 & 1.14 \\
& & Detailed & 0.84 & 0.84 & 0.44 & 0.66 \\
\cmidrule{2-7}
 & \multirow{3}{*}{$r_Q$} 
  & Simplified & 0.01 & 0.01 & 0.05 & 0.03 \\
& & Detailed & 0.01 & 0.01 & 0.06 & 0.03 \\
\cmidrule{2-7}
& \multirow{3}{*}{$r_K$} 
  & Simplified & 0.02 & 0.01 & 0.02 & 0.01 \\
& & Detailed & 0.02 & 0.00 & 0.02 & 0.01 \\
\cmidrule{2-7}
 & \multirow{3}{*}{$r_V$} 
  & Simplified & 0.02 & 0.01 & 0.01 & 0.02 \\
& & Detailed & 0.02 & 0.01 & 0.01 & 0.01 \\
\cmidrule{2-7}
& \multirow{3}{*}{$r_O$} 
  & Simplified & 0.01 & 0.01 & 0.01 & 0.01 \\
& & Detailed & 0.01 & 0.01 & 0.01 & 0.01 \\
\bottomrule
\end{tabular}
}
\caption{
Statistical results for MATH-Counting using Llama-3.1-8B on correct responses.
}
\label{tbl:math_lighteval_math_llama3_8b_grads_counting_probability}
\end{table}

\begin{figure}[!h]
\centering
\includegraphics[width=\linewidth]{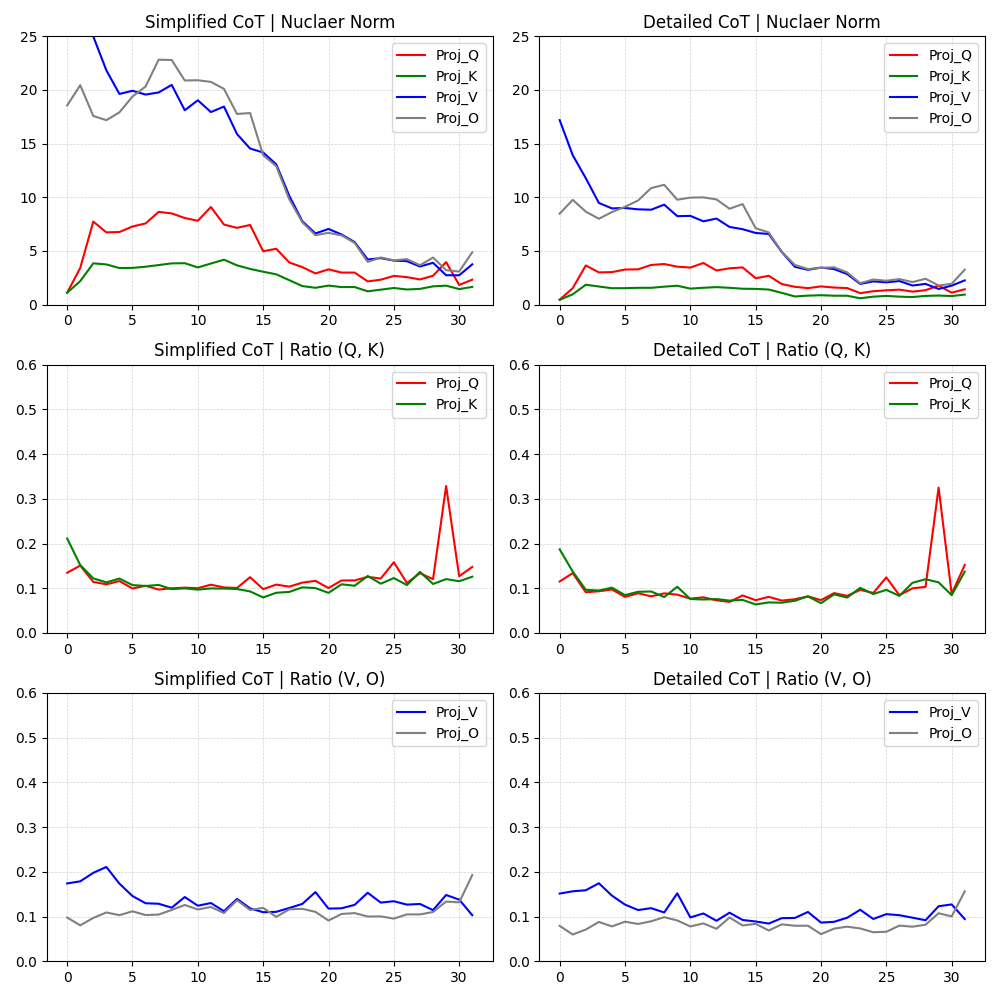}
\caption{Visualization for MATH-Counting using Llama-3.1-8B on correct responses.}
\label{fig:math_correct_counting}
\vspace{-2mm}
\end{figure}

\begin{table}[!tbh]
\centering
\small
\scalebox{0.75}{
\begin{tabular}{l|l|l|cccc}
\toprule
\multirow{2}{*}{\textbf{Dataset}} & \multirow{2}{*}{\textbf{Curve}} & \multirow{2}{*}{\textbf{Cot}} &  \multicolumn{4}{c}{\textbf{Mean Absolute Difference (MAD)}} \\
 & &  & Early & Middle & Last & All\\
\midrule
\multirow{16}{*}{Geometry} & \multirow{3}{*}{$s_Q$} 
  & Simplified & 1.14 & 1.10 & 0.61 & 0.88 \\
& & Detailed & 0.61 & 0.56 & 0.32 & 0.46 \\
\cmidrule{2-7}
& \multirow{3}{*}{$s_K$} 
  & Simplified & 0.44 & 0.41 & 0.22 & 0.33 \\
& & Detailed & 0.24 & 0.16 & 0.10 & 0.16 \\
\cmidrule{2-7}
& \multirow{3}{*}{$s_V$} 
  & Simplified & 3.17 & 1.85 & 0.70 & 1.77 \\
& & Detailed & 1.40 & 0.80 & 0.38 & 0.79 \\
\cmidrule{2-7}
& \multirow{3}{*}{$s_O$} 
  & Simplified & 1.70 & 1.93 & 0.89 & 1.40 \\
& & Detailed & 0.90 & 1.01 & 0.50 & 0.75 \\
\cmidrule{2-7}
 & \multirow{3}{*}{$r_Q$} 
  & Simplified & 0.01 & 0.01 & 0.05 & 0.03 \\
& & Detailed & 0.01 & 0.01 & 0.06 & 0.03 \\
\cmidrule{2-7}
& \multirow{3}{*}{$r_K$} 
  & Simplified & 0.02 & 0.00 & 0.01 & 0.01 \\
& & Detailed & 0.02 & 0.00 & 0.02 & 0.02 \\
\cmidrule{2-7}
 & \multirow{3}{*}{$r_V$} 
  & Simplified & 0.02 & 0.01 & 0.02 & 0.02 \\
& & Detailed & 0.01 & 0.01 & 0.01 & 0.01 \\
\cmidrule{2-7}
& \multirow{3}{*}{$r_O$} 
  & Simplified & 0.01 & 0.01 & 0.01 & 0.01 \\
& & Detailed & 0.01 & 0.01 & 0.01 & 0.01 \\
\bottomrule
\end{tabular}
}
\caption{
Statistical results for MATH-Geometry using Llama-3.1-8B on correct responses.
}
\label{tbl:math_lighteval_math_llama3_8b_grads_geometry}
\end{table}

\begin{figure}[!h]
\centering
\includegraphics[width=\linewidth]{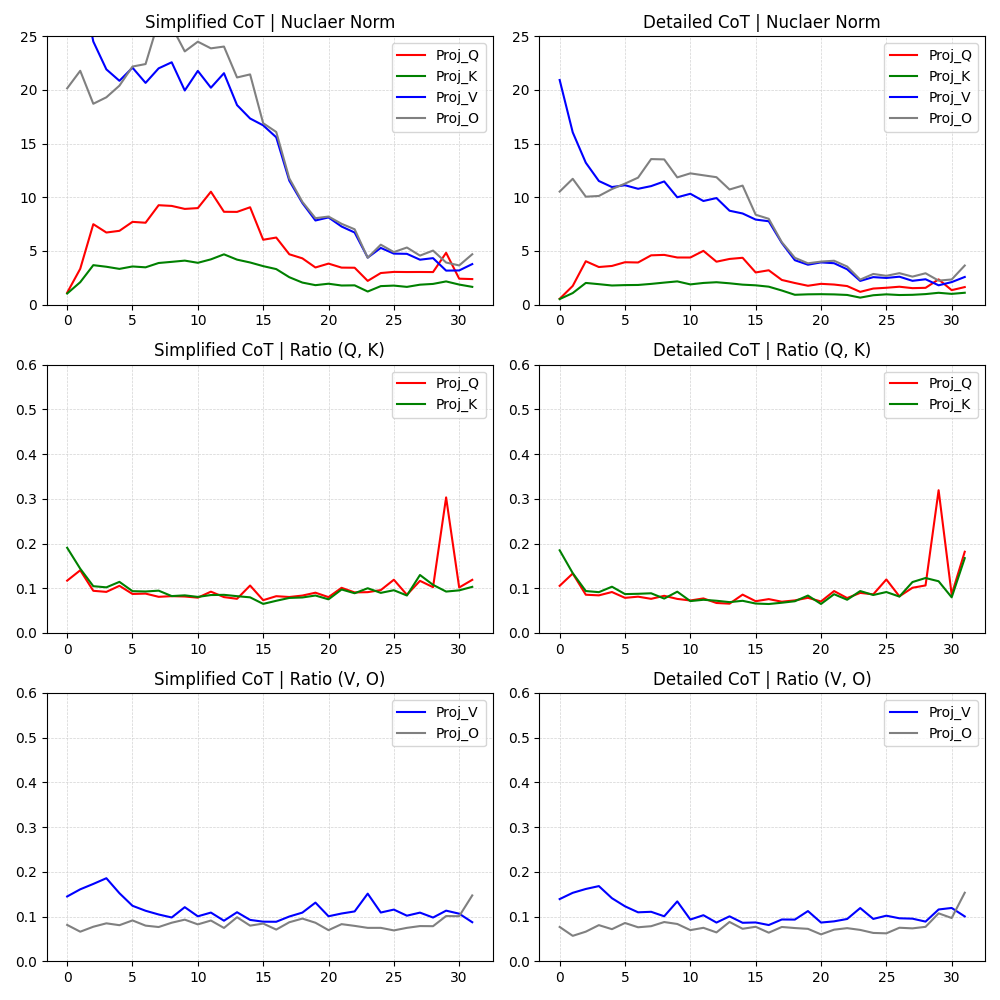}
\caption{Visualization for MATH-Geometry using Llama-3.1-8B on correct responses.}
\label{fig:math_correct_geometry}
\vspace{-2mm}
\end{figure}

The visualizations and statistical results on other reasoning tasks: AQuA (Figure \ref{fig:reasoning_correct_aqua}, Table \ref{tbl:reasoning_cot_500_llama3_8b_grads_aqua_train}), GSM8K (Figure \ref{fig:reasoning_correct_gsm8k}, Table \ref{tbl:reasoning_cot_500_llama3_8b_grads_gsm8k_train}), StrategyQA (Figure \ref{fig:reasoning_correct_strategyqa}, Table \ref{tbl:reasoning_cot_500_llama3_8b_grads_strategyqa_train}), ECQA (Figure \ref{fig:reasoning_correct_ecqa}, Table \ref{tbl:reasoning_cot_500_llama3_8b_grads_ecqa_train}), CREAK (Figure \ref{fig:reasoning_correct_creak}, Table \ref{tbl:reasoning_cot_500_llama3_8b_grads_creak_train}), Sensemaking (Figure \ref{fig:reasoning_correct_sensemaking}, Table \ref{tbl:reasoning_cot_500_llama3_8b_grads_sensemaking_train}).

\begin{table}[!tbh]
\centering
\small
\scalebox{0.75}{
\begin{tabular}{l|l|l|cccc}
\toprule
\multirow{2}{*}{\textbf{Dataset}} & \multirow{2}{*}{\textbf{Curve}} & \multirow{2}{*}{\textbf{Cot}} &  \multicolumn{4}{c}{\textbf{Mean Absolute Difference (MAD)}} \\
 & &  & Early & Middle & Last & All\\
\midrule
\multirow{24}{*}{AQuA} & \multirow{3}{*}{$s_Q$} 
  & None & 4.61 & 5.39 & 7.22 & 5.67 \\
& & Simplified & 1.42 & 1.03 & 0.57 & 0.93 \\
& & Detailed & 0.56 & 0.44 & 0.25 & 0.39 \\
\cmidrule{2-7}
& \multirow{3}{*}{$s_K$} 
  & None & 2.51 & 1.81 & 3.74 & 2.83 \\
& & Simplified & 0.53 & 0.35 & 0.26 & 0.36 \\
& & Detailed & 0.23 & 0.12 & 0.09 & 0.15 \\
\cmidrule{2-7}
& \multirow{3}{*}{$s_V$} 
  & None & 21.85 & 6.80 & 3.35 & 9.85 \\
& & Simplified & 4.59 & 1.80 & 0.83 & 2.24 \\
& & Detailed & 1.27 & 0.68 & 0.38 & 0.71 \\
\cmidrule{2-7}
& \multirow{3}{*}{$s_O$} 
  & None & 6.22 & 5.67 & 2.69 & 4.60 \\
& & Simplified & 1.95 & 2.09 & 0.78 & 1.48 \\
& & Detailed & 0.91 & 0.92 & 0.46 & 0.71 \\
\cmidrule{2-7}
 & \multirow{3}{*}{$r_Q$} 
  & None & 0.02 & 0.09 & 0.14 & 0.09 \\
& & Simplified & 0.01 & 0.01 & 0.05 & 0.03 \\
& & Detailed & 0.01 & 0.01 & 0.06 & 0.03 \\
\cmidrule{2-7}
& \multirow{3}{*}{$r_K$} 
  & None & 0.03 & 0.04 & 0.13 & 0.07 \\
& & Simplified & 0.02 & 0.01 & 0.03 & 0.02 \\
& & Detailed & 0.02 & 0.00 & 0.02 & 0.02 \\
\cmidrule{2-7}
 & \multirow{3}{*}{$r_V$} 
  & None & 0.03 & 0.05 & 0.04 & 0.04 \\
& & Simplified & 0.02 & 0.02 & 0.01 & 0.02 \\
& & Detailed & 0.02 & 0.01 & 0.02 & 0.02 \\
\cmidrule{2-7}
& \multirow{3}{*}{$r_O$} 
  & None & 0.02 & 0.04 & 0.09 & 0.05 \\
& & Simplified & 0.01 & 0.02 & 0.01 & 0.02 \\
& & Detailed & 0.01 & 0.01 & 0.01 & 0.01 \\
\bottomrule
\end{tabular}
}
\caption{
Statistical results for AQuA using Llama-3.1-8B on correct responses.
}
\label{tbl:reasoning_cot_500_llama3_8b_grads_aqua_train}
\end{table}

\begin{figure*}[!h]
\centering
\includegraphics[width=\linewidth]{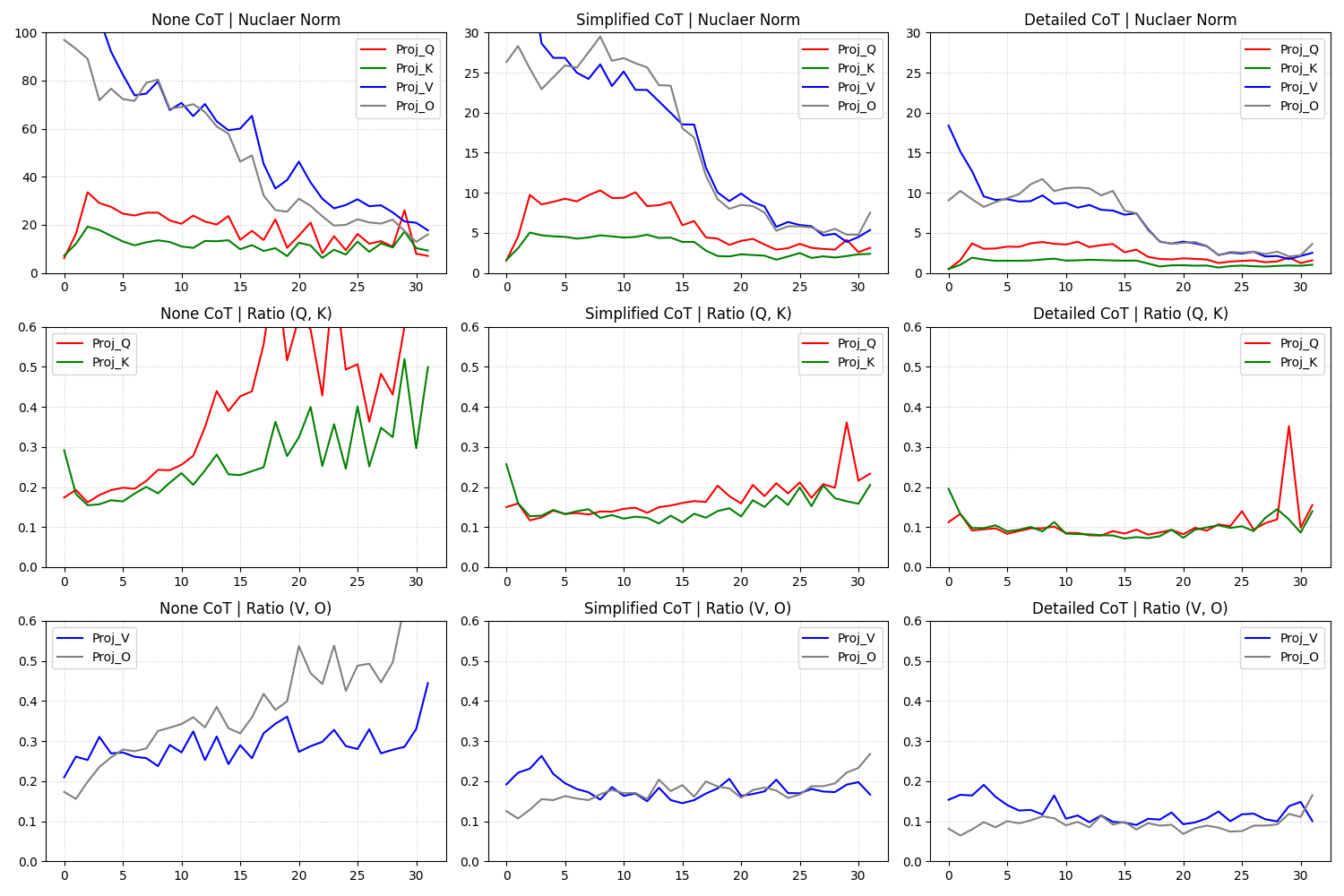}
\caption{Visualization for AQuA using Llama-3.1-8B on correct responses.}
\label{fig:reasoning_correct_aqua}
\vspace{-2mm}
\end{figure*}

\begin{table}[!tbh]
\centering
\small
\scalebox{0.75}{
\begin{tabular}{l|l|l|cccc}
\toprule
\multirow{2}{*}{\textbf{Dataset}} & \multirow{2}{*}{\textbf{Curve}} & \multirow{2}{*}{\textbf{Cot}} &  \multicolumn{4}{c}{\textbf{Mean Absolute Difference (MAD)}} \\
 & &  & Early & Middle & Last & All\\
\midrule
\multirow{24}{*}{GSM8K} & \multirow{3}{*}{$s_Q$} 
  & None & 4.91 & 6.38 & 2.62 & 4.37 \\
& & Simplified & 0.79 & 0.67 & 0.23 & 0.53 \\
& & Detailed & 0.47 & 0.37 & 0.18 & 0.31 \\
\cmidrule{2-7}
& \multirow{3}{*}{$s_K$} 
  & None & 2.38 & 1.98 & 1.67 & 2.02 \\
& & Simplified & 0.29 & 0.22 & 0.16 & 0.21 \\
& & Detailed & 0.21 & 0.11 & 0.08 & 0.13 \\
\cmidrule{2-7}
& \multirow{3}{*}{$s_V$} 
  & None & 19.12 & 6.14 & 5.86 & 9.85 \\
& & Simplified & 1.63 & 1.07 & 0.59 & 1.07 \\
& & Detailed & 1.13 & 0.60 & 0.35 & 0.64 \\
\cmidrule{2-7}
& \multirow{3}{*}{$s_O$} 
  & None & 5.29 & 5.43 & 2.82 & 4.56 \\
& & Simplified & 1.26 & 1.31 & 0.58 & 0.99 \\
& & Detailed & 0.73 & 0.79 & 0.43 & 0.61 \\
\cmidrule{2-7}
 & \multirow{3}{*}{$r_Q$} 
  & None & 0.02 & 0.08 & 0.11 & 0.07 \\
& & Simplified & 0.01 & 0.01 & 0.03 & 0.02 \\
& & Detailed & 0.01 & 0.01 & 0.05 & 0.02 \\
\cmidrule{2-7}
& \multirow{3}{*}{$r_K$} 
  & None & 0.02 & 0.05 & 0.03 & 0.03 \\
& & Simplified & 0.02 & 0.01 & 0.01 & 0.01 \\
& & Detailed & 0.02 & 0.00 & 0.02 & 0.02 \\
\cmidrule{2-7}
 & \multirow{3}{*}{$r_V$} 
  & None & 0.03 & 0.03 & 0.02 & 0.03 \\
& & Simplified & 0.02 & 0.01 & 0.02 & 0.02 \\
& & Detailed & 0.02 & 0.01 & 0.02 & 0.02 \\
\cmidrule{2-7}
& \multirow{3}{*}{$r_O$} 
  & None & 0.03 & 0.03 & 0.07 & 0.05 \\
& & Simplified & 0.01 & 0.02 & 0.02 & 0.02 \\
& & Detailed & 0.01 & 0.02 & 0.02 & 0.02 \\
\bottomrule
\end{tabular}
}
\caption{
Statistical results for GSM8K using Llama-3.1-8B on correct responses.
}
\label{tbl:reasoning_cot_500_llama3_8b_grads_gsm8k_train}
\end{table}

\begin{figure*}[!h]
\centering
\includegraphics[width=\linewidth]{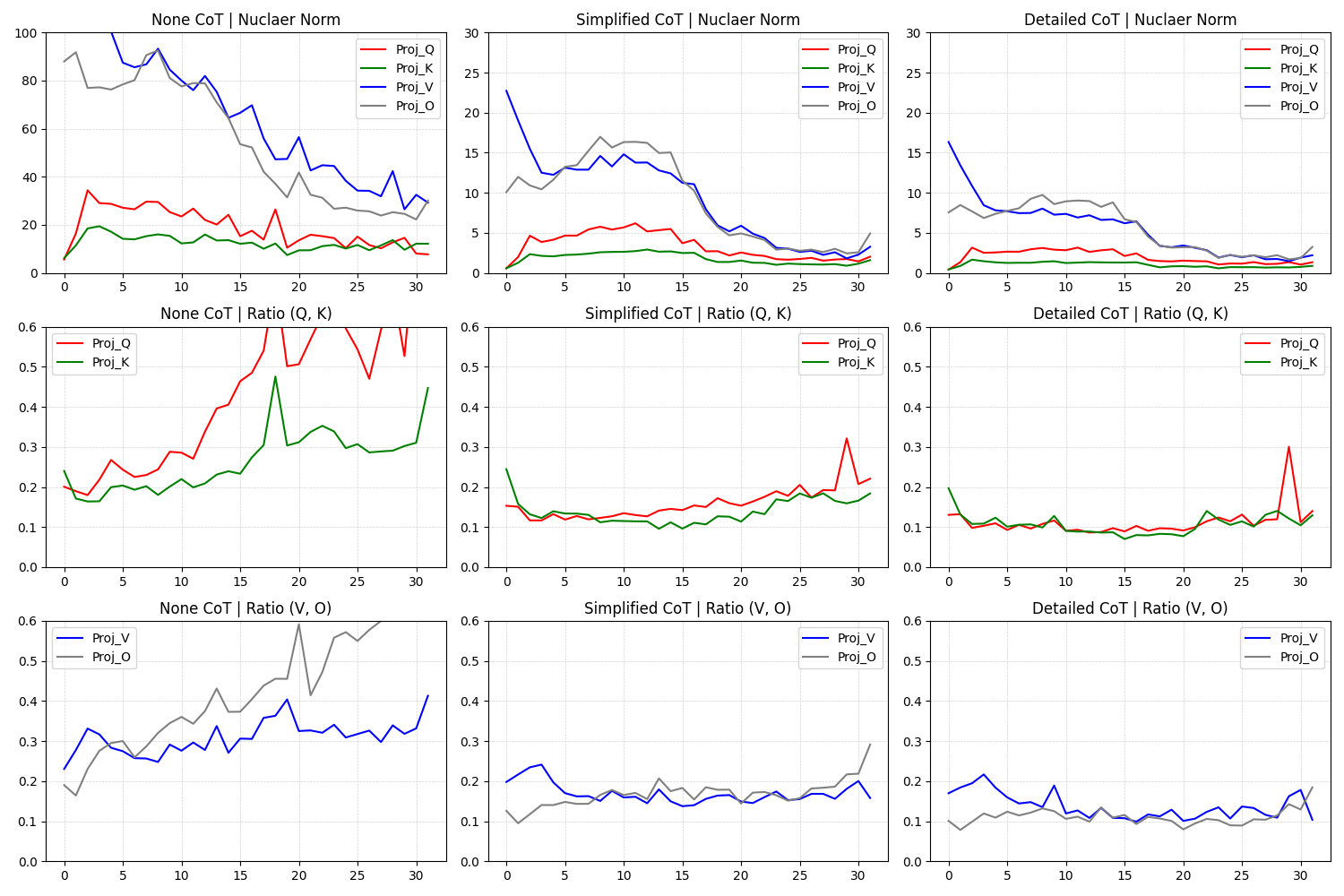}
\caption{Visualization for GSM8K using Llama-3.1-8B on correct responses.}
\label{fig:reasoning_correct_gsm8k}
\vspace{-2mm}
\end{figure*}

\begin{table}[!tbh]
\centering
\small
\scalebox{0.75}{
\begin{tabular}{l|l|l|cccc}
\toprule
\multirow{2}{*}{\textbf{Dataset}} & \multirow{2}{*}{\textbf{Curve}} & \multirow{2}{*}{\textbf{Cot}} &  \multicolumn{4}{c}{\textbf{Mean Absolute Difference (MAD)}} \\
 & &  & Early & Middle & Last & All\\
\midrule
\multirow{24}{*}{StrategyQA} & \multirow{3}{*}{$s_Q$} 
  & None & 4.71 & 2.43 & 0.93 & 2.65 \\
& & Simplified & 0.84 & 0.76 & 0.40 & 0.64 \\
& & Detailed & 0.57 & 0.50 & 0.22 & 0.40 \\
\cmidrule{2-7}
& \multirow{3}{*}{$s_K$} 
  & None & 2.80 & 1.25 & 1.17 & 1.77 \\
& & Simplified & 0.40 & 0.24 & 0.19 & 0.27 \\
& & Detailed & 0.23 & 0.13 & 0.11 & 0.15 \\
\cmidrule{2-7}
& \multirow{3}{*}{$s_V$} 
  & None & 26.10 & 6.82 & 2.59 & 10.73 \\
& & Simplified & 3.55 & 1.13 & 0.81 & 1.66 \\
& & Detailed & 1.25 & 0.77 & 0.55 & 0.80 \\
\cmidrule{2-7}
& \multirow{3}{*}{$s_O$} 
  & None & 10.88 & 5.17 & 3.06 & 5.94 \\
& & Simplified & 1.56 & 1.26 & 1.06 & 1.22 \\
& & Detailed & 0.95 & 1.10 & 0.79 & 0.90 \\
\cmidrule{2-7}
 & \multirow{3}{*}{$r_Q$} 
  & None & 0.02 & 0.05 & 0.08 & 0.05 \\
& & Simplified & 0.01 & 0.01 & 0.03 & 0.02 \\
& & Detailed & 0.01 & 0.01 & 0.06 & 0.03 \\
\cmidrule{2-7}
& \multirow{3}{*}{$r_K$} 
  & None & 0.02 & 0.03 & 0.06 & 0.04 \\
& & Simplified & 0.01 & 0.01 & 0.02 & 0.01 \\
& & Detailed & 0.02 & 0.01 & 0.02 & 0.01 \\
\cmidrule{2-7}
 & \multirow{3}{*}{$r_V$} 
  & None & 0.04 & 0.03 & 0.04 & 0.04 \\
& & Simplified & 0.04 & 0.02 & 0.02 & 0.02 \\
& & Detailed & 0.02 & 0.01 & 0.01 & 0.01 \\
\cmidrule{2-7}
& \multirow{3}{*}{$r_O$} 
  & None & 0.03 & 0.03 & 0.03 & 0.03 \\
& & Simplified & 0.01 & 0.03 & 0.02 & 0.02 \\
& & Detailed & 0.01 & 0.02 & 0.02 & 0.02 \\
\bottomrule
\end{tabular}
}
\caption{
Statistical results for StrategyQA using Llama-3.1-8B on correct responses.
}
\label{tbl:reasoning_cot_500_llama3_8b_grads_strategyqa_train}
\end{table}

\begin{figure*}[!h]
\centering
\includegraphics[width=\linewidth]{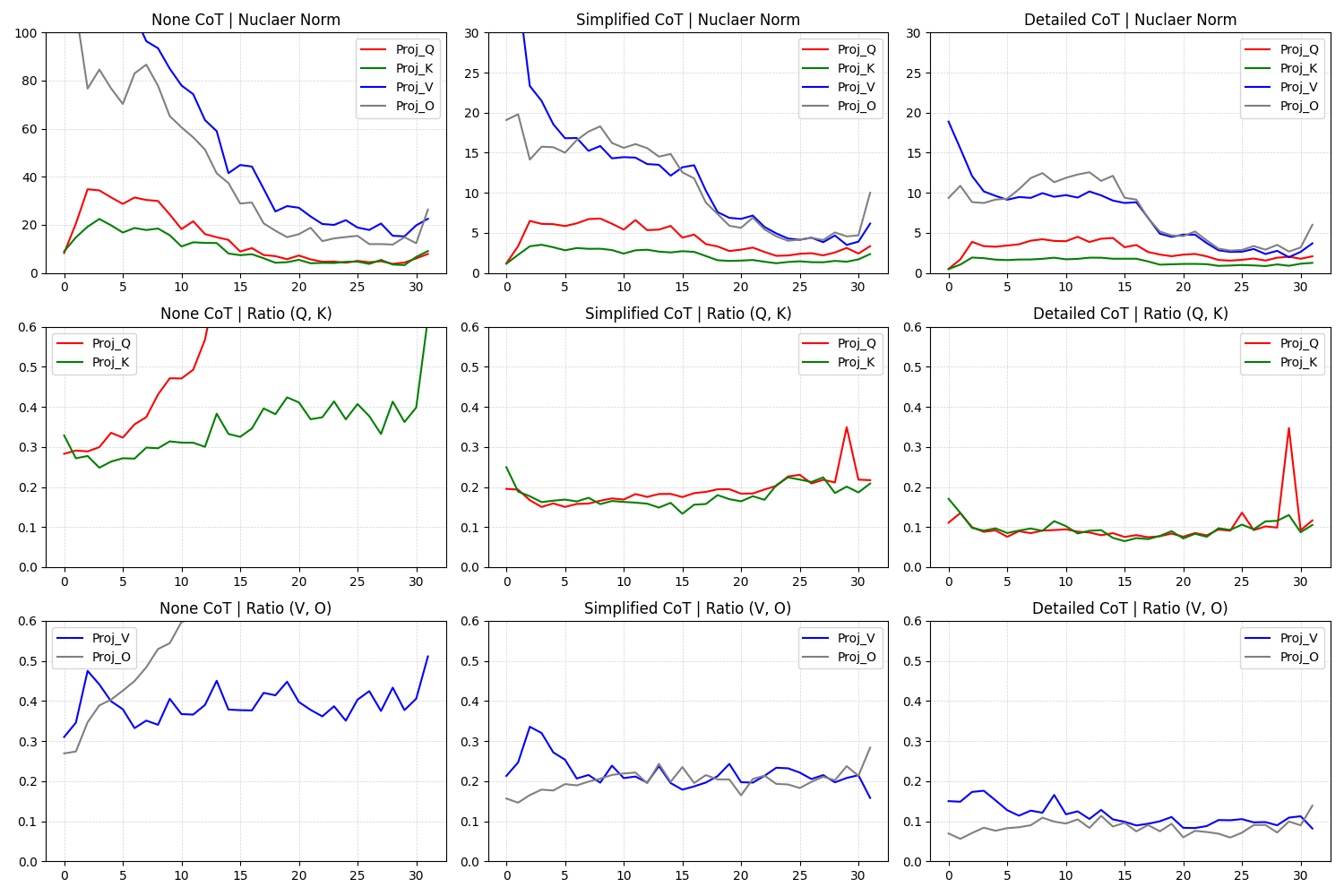}
\caption{Visualization for StrategyQA using Llama-3.1-8B on correct responses.}
\label{fig:reasoning_correct_strategyqa}
\vspace{-2mm}
\end{figure*}

\begin{table}[!tbh]
\centering
\small
\scalebox{0.75}{
\begin{tabular}{l|l|l|cccc}
\toprule
\multirow{2}{*}{\textbf{Dataset}} & \multirow{2}{*}{\textbf{Curve}} & \multirow{2}{*}{\textbf{Cot}} &  \multicolumn{4}{c}{\textbf{Mean Absolute Difference (MAD)}} \\
 & &  & Early & Middle & Last & All\\
\midrule
\multirow{24}{*}{ECQA} & \multirow{3}{*}{$s_Q$} 
  & None & 7.39 & 6.32 & 2.14 & 4.94 \\
& & Simplified & 1.46 & 1.12 & 0.37 & 0.92 \\
& & Detailed & 0.83 & 0.58 & 0.38 & 0.56 \\
\cmidrule{2-7}
& \multirow{3}{*}{$s_K$} 
  & None & 3.23 & 3.17 & 2.97 & 3.05 \\
& & Simplified & 0.65 & 0.30 & 0.33 & 0.42 \\
& & Detailed & 0.34 & 0.15 & 0.16 & 0.21 \\
\cmidrule{2-7}
& \multirow{3}{*}{$s_V$} 
  & None & 31.64 & 12.53 & 4.07 & 14.50 \\
& & Simplified & 5.53 & 1.53 & 0.97 & 2.45 \\
& & Detailed & 1.78 & 0.94 & 0.69 & 1.06 \\
\cmidrule{2-7}
& \multirow{3}{*}{$s_O$} 
  & None & 8.18 & 9.24 & 2.77 & 6.16 \\
& & Simplified & 1.42 & 1.78 & 0.99 & 1.30 \\
& & Detailed & 1.17 & 1.35 & 0.88 & 1.06 \\
\cmidrule{2-7}
 & \multirow{3}{*}{$r_Q$} 
  & None & 0.02 & 0.08 & 0.11 & 0.07 \\
& & Simplified & 0.01 & 0.01 & 0.03 & 0.02 \\
& & Detailed & 0.02 & 0.01 & 0.10 & 0.04 \\
\cmidrule{2-7}
& \multirow{3}{*}{$r_K$} 
  & None & 0.02 & 0.05 & 0.05 & 0.04 \\
& & Simplified & 0.02 & 0.01 & 0.02 & 0.02 \\
& & Detailed & 0.02 & 0.01 & 0.03 & 0.02 \\
\cmidrule{2-7}
 & \multirow{3}{*}{$r_V$} 
  & None & 0.02 & 0.03 & 0.03 & 0.03 \\
& & Simplified & 0.02 & 0.02 & 0.02 & 0.02 \\
& & Detailed & 0.02 & 0.01 & 0.01 & 0.02 \\
\cmidrule{2-7}
& \multirow{3}{*}{$r_O$} 
  & None & 0.02 & 0.04 & 0.07 & 0.05 \\
& & Simplified & 0.01 & 0.02 & 0.02 & 0.02 \\
& & Detailed & 0.01 & 0.02 & 0.02 & 0.02 \\
\bottomrule
\end{tabular}
}
\caption{
Statistical results for ECQA using Llama-3.1-8B on correct responses.
}
\label{tbl:reasoning_cot_500_llama3_8b_grads_ecqa_train}
\end{table}

\begin{figure*}[!h]
\centering
\includegraphics[width=\linewidth]{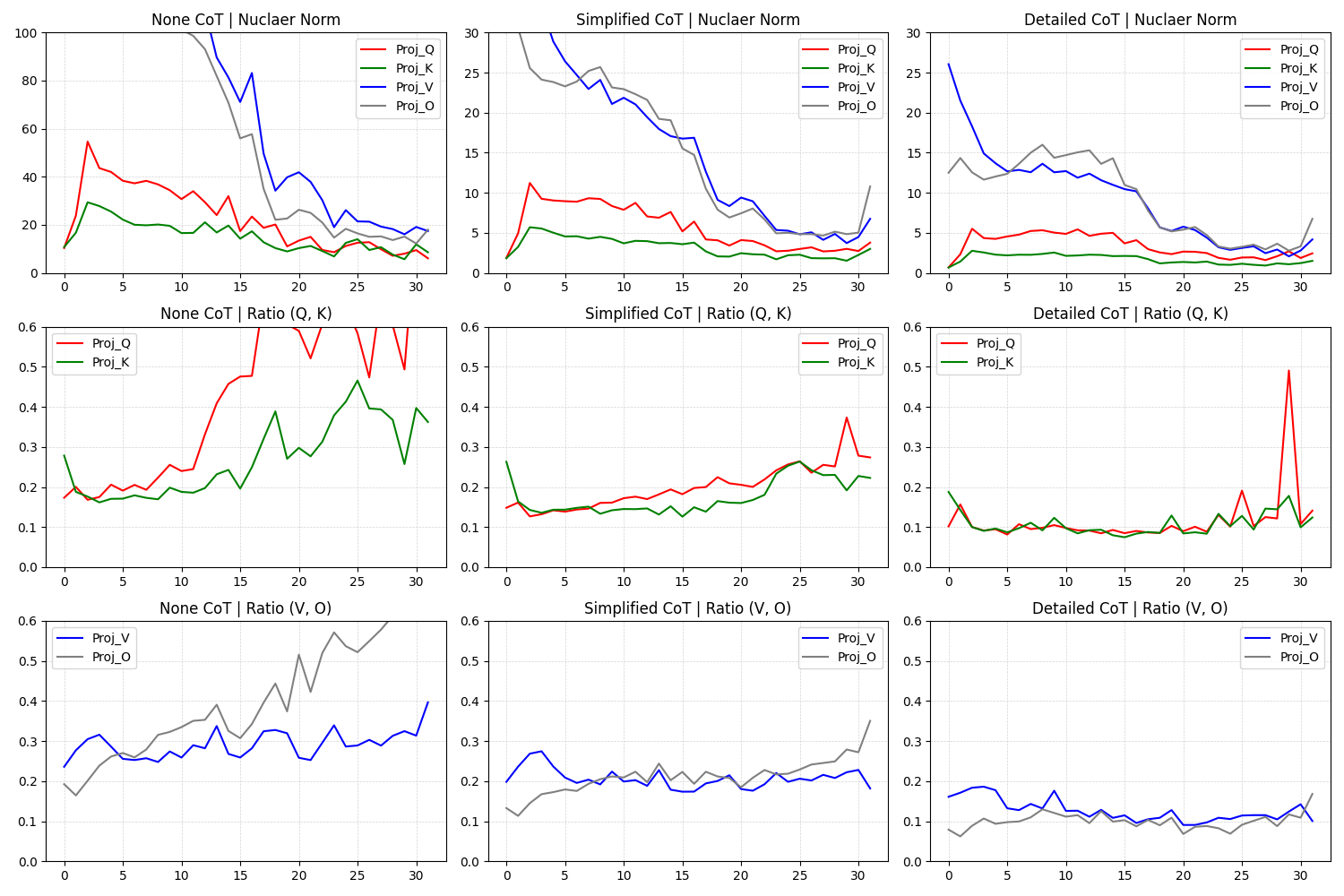}
\caption{Visualization for ECQA using Llama-3.1-8B on correct responses.}
\label{fig:reasoning_correct_ecqa}
\vspace{-2mm}
\end{figure*}

\begin{table}[!tbh]
\centering
\small
\scalebox{0.75}{
\begin{tabular}{l|l|l|cccc}
\toprule
\multirow{2}{*}{\textbf{Dataset}} & \multirow{2}{*}{\textbf{Curve}} & \multirow{2}{*}{\textbf{Cot}} &  \multicolumn{4}{c}{\textbf{Mean Absolute Difference (MAD)}} \\
 & &  & Early & Middle & Last & All\\
\midrule
\multirow{24}{*}{CREAK} & \multirow{3}{*}{$s_Q$} 
  & None & 5.05 & 5.06 & 3.79 & 4.40 \\
& & Simplified & 1.05 & 1.06 & 0.44 & 0.80 \\
& & Detailed & 0.65 & 0.56 & 0.31 & 0.47 \\
\cmidrule{2-7}
& \multirow{3}{*}{$s_K$} 
  & None & 2.74 & 2.39 & 4.00 & 2.97 \\
& & Simplified & 0.50 & 0.29 & 0.26 & 0.34 \\
& & Detailed & 0.30 & 0.16 & 0.12 & 0.19 \\
\cmidrule{2-7}
& \multirow{3}{*}{$s_V$} 
  & None & 26.92 & 10.35 & 4.35 & 12.72 \\
& & Simplified & 4.74 & 1.48 & 0.89 & 2.19 \\
& & Detailed & 1.83 & 0.84 & 0.65 & 1.02 \\
\cmidrule{2-7}
& \multirow{3}{*}{$s_O$} 
  & None & 9.03 & 8.36 & 3.81 & 6.62 \\
& & Simplified & 1.44 & 1.65 & 1.05 & 1.29 \\
& & Detailed & 1.08 & 1.28 & 0.87 & 1.01 \\
\cmidrule{2-7}
 & \multirow{3}{*}{$r_Q$} 
  & None & 0.02 & 0.06 & 0.14 & 0.07 \\
& & Simplified & 0.01 & 0.01 & 0.03 & 0.02 \\
& & Detailed & 0.01 & 0.01 & 0.07 & 0.03 \\
\cmidrule{2-7}
& \multirow{3}{*}{$r_K$} 
  & None & 0.02 & 0.03 & 0.08 & 0.04 \\
& & Simplified & 0.02 & 0.01 & 0.02 & 0.02 \\
& & Detailed & 0.02 & 0.01 & 0.02 & 0.02 \\
\cmidrule{2-7}
 & \multirow{3}{*}{$r_V$} 
  & None & 0.03 & 0.04 & 0.04 & 0.04 \\
& & Simplified & 0.03 & 0.02 & 0.02 & 0.02 \\
& & Detailed & 0.02 & 0.01 & 0.01 & 0.02 \\
\cmidrule{2-7}
& \multirow{3}{*}{$r_O$} 
  & None & 0.02 & 0.04 & 0.05 & 0.04 \\
& & Simplified & 0.01 & 0.03 & 0.02 & 0.02 \\
& & Detailed & 0.01 & 0.02 & 0.02 & 0.01 \\
\bottomrule
\end{tabular}
}
\caption{
Statistical results for CREAK using Llama-3.1-8B on correct responses.
}
\label{tbl:reasoning_cot_500_llama3_8b_grads_creak_train}
\end{table}

\begin{figure*}[!h]
\centering
\includegraphics[width=\linewidth]{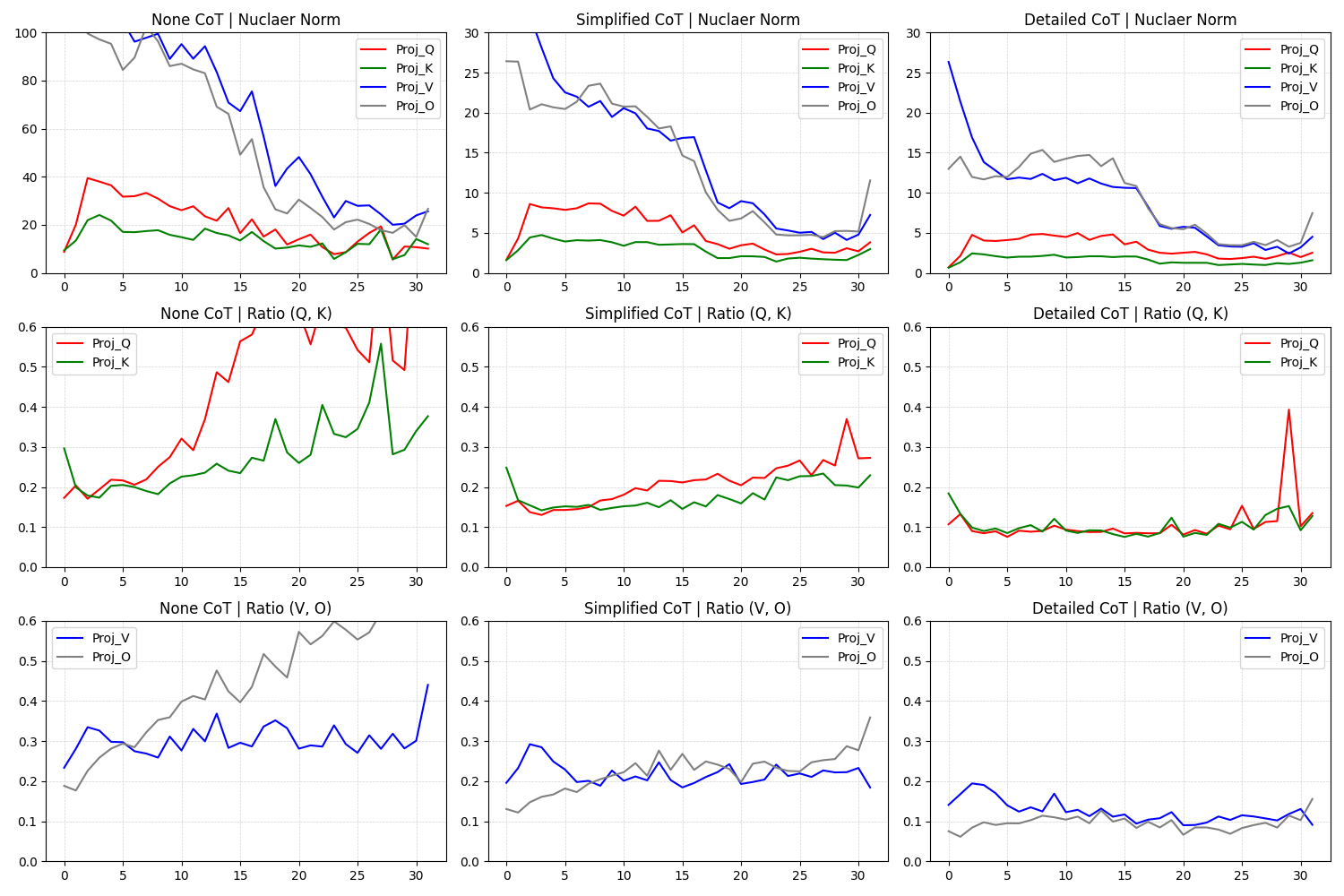}
\caption{Visualization for CREAK using Llama-3.1-8B on correct responses.}
\label{fig:reasoning_correct_creak}
\vspace{-2mm}
\end{figure*}

\begin{table}[!tbh]
\centering
\small
\scalebox{0.75}{
\begin{tabular}{l|l|l|cccc}
\toprule
\multirow{2}{*}{\textbf{Dataset}} & \multirow{2}{*}{\textbf{Curve}} & \multirow{2}{*}{\textbf{Cot}} &  \multicolumn{4}{c}{\textbf{Mean Absolute Difference (MAD)}} \\
 & &  & Early & Middle & Last & All\\
\midrule
\multirow{24}{*}{Sensemaking} & \multirow{3}{*}{$s_Q$} 
  & None & 4.61 & 6.54 & 3.41 & 4.65 \\
& & Simplified & 1.46 & 1.46 & 0.54 & 1.09 \\
& & Detailed & 0.68 & 0.53 & 0.26 & 0.46 \\
\cmidrule{2-7}
& \multirow{3}{*}{$s_K$} 
  & None & 2.22 & 2.63 & 2.83 & 2.55 \\
& & Simplified & 0.58 & 0.48 & 0.37 & 0.48 \\
& & Detailed & 0.31 & 0.14 & 0.13 & 0.19 \\
\cmidrule{2-7}
& \multirow{3}{*}{$s_V$} 
  & None & 25.94 & 8.37 & 3.03 & 11.13 \\
& & Simplified & 5.68 & 2.02 & 1.03 & 2.70 \\
& & Detailed & 1.56 & 0.77 & 0.57 & 0.89 \\
\cmidrule{2-7}
& \multirow{3}{*}{$s_O$} 
  & None & 8.32 & 7.22 & 2.38 & 5.61 \\
& & Simplified & 1.97 & 2.34 & 1.06 & 1.68 \\
& & Detailed & 1.10 & 1.20 & 0.72 & 0.94 \\
\cmidrule{2-7}
 & \multirow{3}{*}{$r_Q$} 
  & None & 0.02 & 0.08 & 0.13 & 0.08 \\
& & Simplified & 0.01 & 0.02 & 0.03 & 0.02 \\
& & Detailed & 0.01 & 0.00 & 0.06 & 0.03 \\
\cmidrule{2-7}
& \multirow{3}{*}{$r_K$} 
  & None & 0.03 & 0.03 & 0.09 & 0.05 \\
& & Simplified & 0.02 & 0.02 & 0.03 & 0.02 \\
& & Detailed & 0.02 & 0.01 & 0.02 & 0.02 \\
\cmidrule{2-7}
 & \multirow{3}{*}{$r_V$} 
  & None & 0.03 & 0.04 & 0.03 & 0.03 \\
& & Simplified & 0.03 & 0.02 & 0.02 & 0.02 \\
& & Detailed & 0.02 & 0.01 & 0.01 & 0.01 \\
\cmidrule{2-7}
& \multirow{3}{*}{$r_O$} 
  & None & 0.02 & 0.04 & 0.05 & 0.04 \\
& & Simplified & 0.02 & 0.03 & 0.02 & 0.02 \\
& & Detailed & 0.01 & 0.02 & 0.02 & 0.02 \\
\bottomrule
\end{tabular}
}
\caption{
Statistical results for Sensemaking using Llama-3.1-8B on correct responses.
}
\label{tbl:reasoning_cot_500_llama3_8b_grads_sensemaking_train}
\end{table}

\begin{figure*}[!h]
\centering
\includegraphics[width=\linewidth]{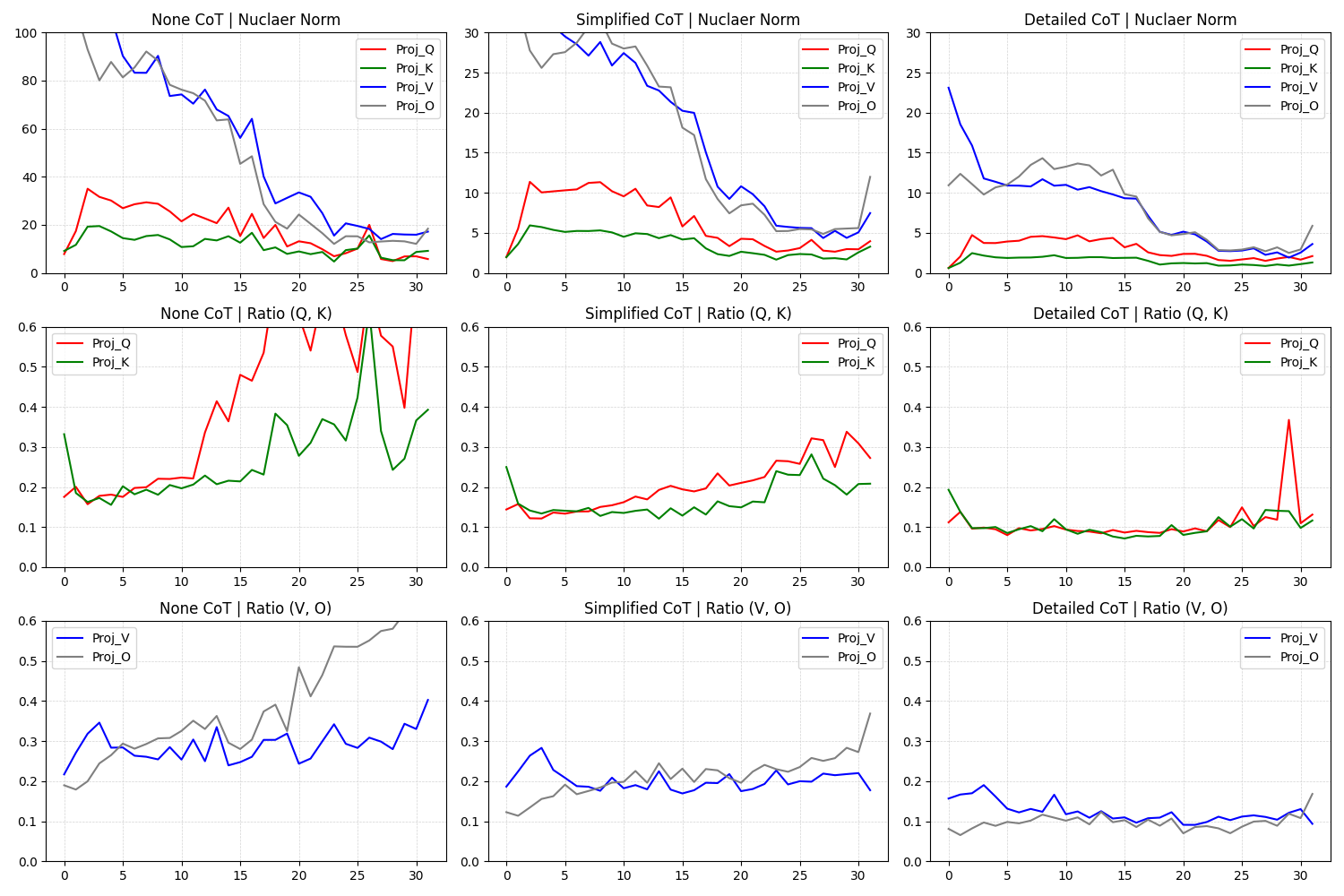}
\caption{Visualization for Sensemaking using Llama-3.1-8B on correct responses.}
\label{fig:reasoning_correct_sensemaking}
\vspace{-2mm}
\end{figure*}

\subsubsection{Wiki Tasks}

The visualizations and statistical results on Wiki tasks are shown in Figure \ref{fig:wiki_correct} and Table \ref{tbl:wiki_wiki_popularity_llama3_8b_grads}.

\begin{table}[!tbh]
\centering
\small
\scalebox{0.75}{
\begin{tabular}{l|l|l|cccc}
\toprule
\multirow{2}{*}{\textbf{Dataset}} & \multirow{2}{*}{\textbf{Curve}} & \multirow{2}{*}{\textbf{Cot}} &  \multicolumn{4}{c}{\textbf{Mean Absolute Difference (MAD)}} \\
 & &  & Early & Middle & Last & All\\
\midrule
\multirow{32}{*}{Wiki} & \multirow{3}{*}{$s_Q$} 
  & Len 100 & 0.88 & 0.76 & 1.12 & 0.92 \\
& & Len 500 & 0.71 & 0.71 & 0.78 & 0.71 \\
& & Len 1000 & 0.62 & 0.65 & 0.59 & 0.59 \\
& & Unpopular & 1.33 & 0.79 & 0.72 & 0.92 \\
\cmidrule{2-7}
& \multirow{3}{*}{$s_K$} 
  & Len 100 & 0.36 & 0.23 & 0.30 & 0.31 \\
& & Len 500 & 0.28 & 0.22 & 0.25 & 0.25 \\
& & Len 1000 & 0.24 & 0.18 & 0.20 & 0.20 \\
& & Unpopular & 0.45 & 0.25 & 0.21 & 0.32 \\
\cmidrule{2-7}
& \multirow{3}{*}{$s_V$} 
  & Len 100 & 4.19 & 0.97 & 0.89 & 1.89 \\
& & Len 500 & 2.55 & 1.09 & 0.87 & 1.41 \\
& & Len 1000 & 1.96 & 1.00 & 0.76 & 1.17 \\
& & Unpopular & 10.07 & 1.41 & 0.72 & 3.68 \\
\cmidrule{2-7}
& \multirow{3}{*}{$s_O$} 
  & Len 100 & 1.88 & 1.18 & 1.04 & 1.34 \\
& & Len 500 & 1.64 & 1.34 & 1.06 & 1.29 \\
& & Len 1000 & 1.53 & 1.27 & 0.98 & 1.19 \\
& & Unpopular & 3.14 & 1.12 & 0.85 & 1.63 \\
\cmidrule{2-7}
 & \multirow{3}{*}{$r_Q$} 
  & Len 100 & 0.02 & 0.01 & 0.06 & 0.03 \\
& & Len 500 & 0.01 & 0.01 & 0.05 & 0.03 \\
& & Len 1000 & 0.01 & 0.01 & 0.05 & 0.02 \\
& & Unpopular & 0.02 & 0.01 & 0.04 & 0.03 \\
\cmidrule{2-7}
& \multirow{3}{*}{$r_K$} 
  & Len 100 & 0.01 & 0.01 & 0.01 & 0.01 \\
& & Len 500 & 0.01 & 0.01 & 0.01 & 0.01 \\
& & Len 1000 & 0.01 & 0.00 & 0.01 & 0.01 \\
& & Unpopular & 0.02 & 0.02 & 0.03 & 0.02 \\
\cmidrule{2-7}
 & \multirow{3}{*}{$r_V$} 
  & Len 100 & 0.03 & 0.02 & 0.01 & 0.02 \\
& & Len 500 & 0.02 & 0.02 & 0.01 & 0.02 \\
& & Len 1000 & 0.02 & 0.01 & 0.01 & 0.01 \\
& & Unpopular & 0.04 & 0.02 & 0.03 & 0.03 \\
\cmidrule{2-7}
& \multirow{3}{*}{$r_O$} 
  & Len 100 & 0.01 & 0.03 & 0.02 & 0.02 \\
& & Len 500 & 0.01 & 0.02 & 0.02 & 0.01 \\
& & Len 1000 & 0.01 & 0.01 & 0.01 & 0.01 \\
& & Unpopular & 0.01 & 0.04 & 0.02 & 0.02 \\
\bottomrule
\end{tabular}
}
\caption{
Statistical results for Wiki using Llama-3.1-8B on correct responses.
}
\label{tbl:wiki_wiki_popularity_llama3_8b_grads}
\end{table}

\begin{figure*}[!h]
\centering
\includegraphics[width=\linewidth]{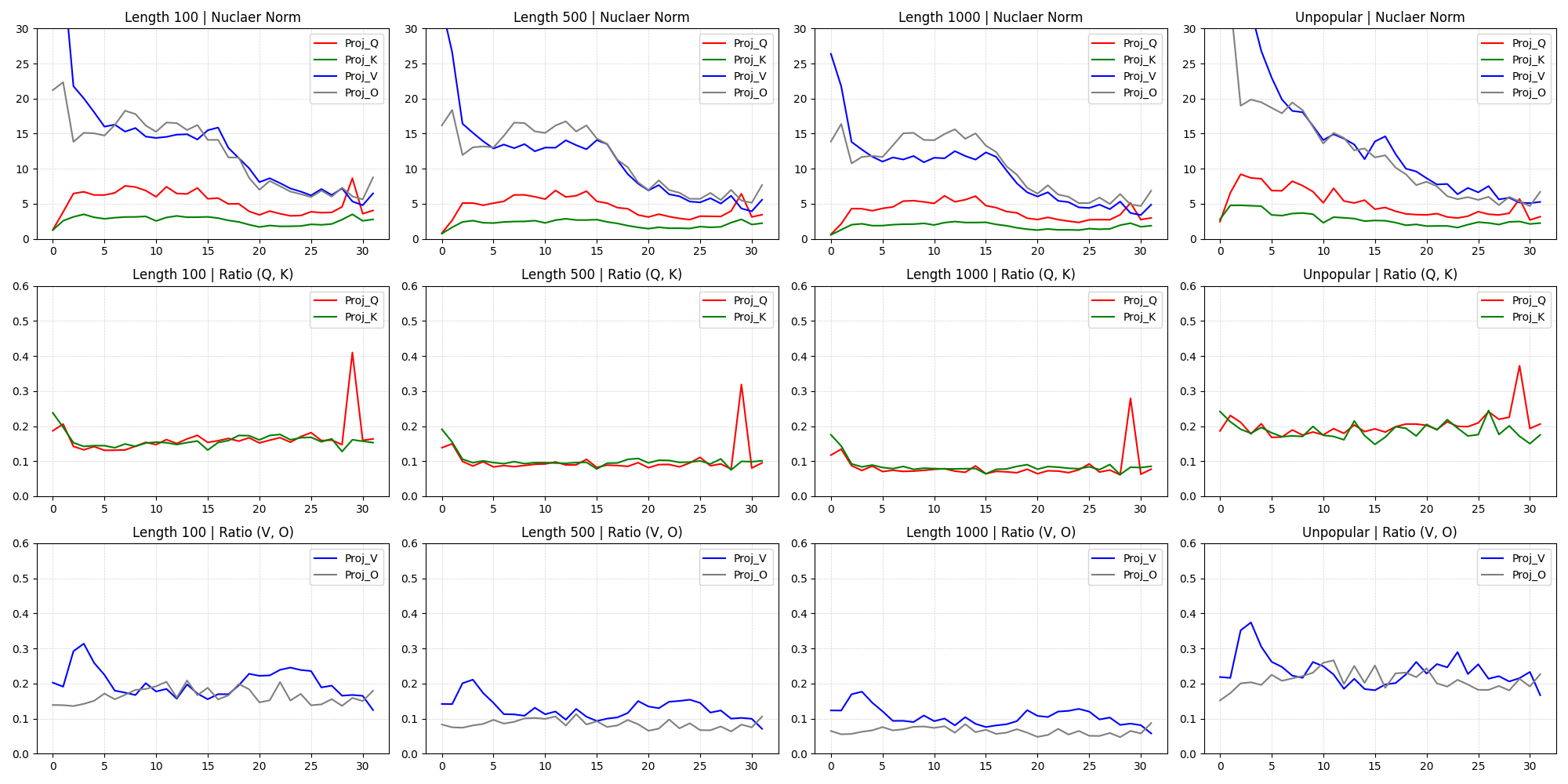}
\caption{Visualization for Wiki tasks using Llama-3.1-8B on correct responses.}
\label{fig:wiki_correct}
\vspace{-2mm}
\end{figure*}

\subsection{Pre-trained LLM on Irrelevant Responses}

\subsubsection{Reasoning Tasks}

The visualizations and statistical results on MATH tasks: MATH-Algebra (Figure \ref{fig:math_nonsense_algebra}, Table \ref{tbl:math_lighteval_math_wrong_answer_shuffle_llama3_8b_grads_algebra}), MATH-Counting (Figure \ref{fig:math_nonsense_counting}, Table \ref{tbl:math_lighteval_math_wrong_answer_shuffle_llama3_8b_grads_counting_probability}), MATH-Geometry (Figure \ref{fig:math_nonsense_geometry}, Table \ref{tbl:math_lighteval_math_wrong_answer_shuffle_llama3_8b_grads_geometry}).

\begin{table}[!tbh]
\centering
\small
\scalebox{0.75}{
\begin{tabular}{l|l|l|cccc}
\toprule
\multirow{2}{*}{\textbf{Dataset}} & \multirow{2}{*}{\textbf{Curve}} & \multirow{2}{*}{\textbf{Cot}} &  \multicolumn{4}{c}{\textbf{Mean Absolute Difference (MAD)}} \\
 & &  & Early & Middle & Last & All\\
\midrule
\multirow{16}{*}{Algebra} & \multirow{3}{*}{$s_Q$} 
  & Simplified & 1.31 & 1.14 & 0.82 & 1.03 \\
& & Detailed & 0.68 & 0.55 & 0.30 & 0.48 \\
\cmidrule{2-7}
& \multirow{3}{*}{$s_K$} 
  & Simplified & 0.47 & 0.39 & 0.24 & 0.36 \\
& & Detailed & 0.24 & 0.16 & 0.11 & 0.16 \\
\cmidrule{2-7}
& \multirow{3}{*}{$s_V$} 
  & Simplified & 3.38 & 1.55 & 0.71 & 1.70 \\
& & Detailed & 1.48 & 0.79 & 0.41 & 0.82 \\
\cmidrule{2-7}
& \multirow{3}{*}{$s_O$} 
  & Simplified & 1.75 & 1.74 & 0.89 & 1.36 \\
& & Detailed & 0.82 & 1.04 & 0.52 & 0.74 \\
\cmidrule{2-7}
 & \multirow{3}{*}{$r_Q$} 
  & Simplified & 0.01 & 0.02 & 0.08 & 0.04 \\
& & Detailed & 0.01 & 0.01 & 0.06 & 0.03 \\
\cmidrule{2-7}
& \multirow{3}{*}{$r_K$} 
  & Simplified & 0.01 & 0.01 & 0.03 & 0.02 \\
& & Detailed & 0.02 & 0.01 & 0.02 & 0.01 \\
\cmidrule{2-7}
 & \multirow{3}{*}{$r_V$} 
  & Simplified & 0.02 & 0.01 & 0.01 & 0.02 \\
& & Detailed & 0.01 & 0.01 & 0.01 & 0.01 \\
\cmidrule{2-7}
& \multirow{3}{*}{$r_O$} 
  & Simplified & 0.01 & 0.01 & 0.01 & 0.01 \\
& & Detailed & 0.01 & 0.01 & 0.01 & 0.01 \\
\bottomrule
\end{tabular}
}
\caption{
Statistical results for MATH-Algebra using Llama-3.1-8B on irrelevant responses.
}
\label{tbl:math_lighteval_math_wrong_answer_shuffle_llama3_8b_grads_algebra}
\end{table}

\begin{figure}[!h]
\centering
\includegraphics[width=\linewidth]{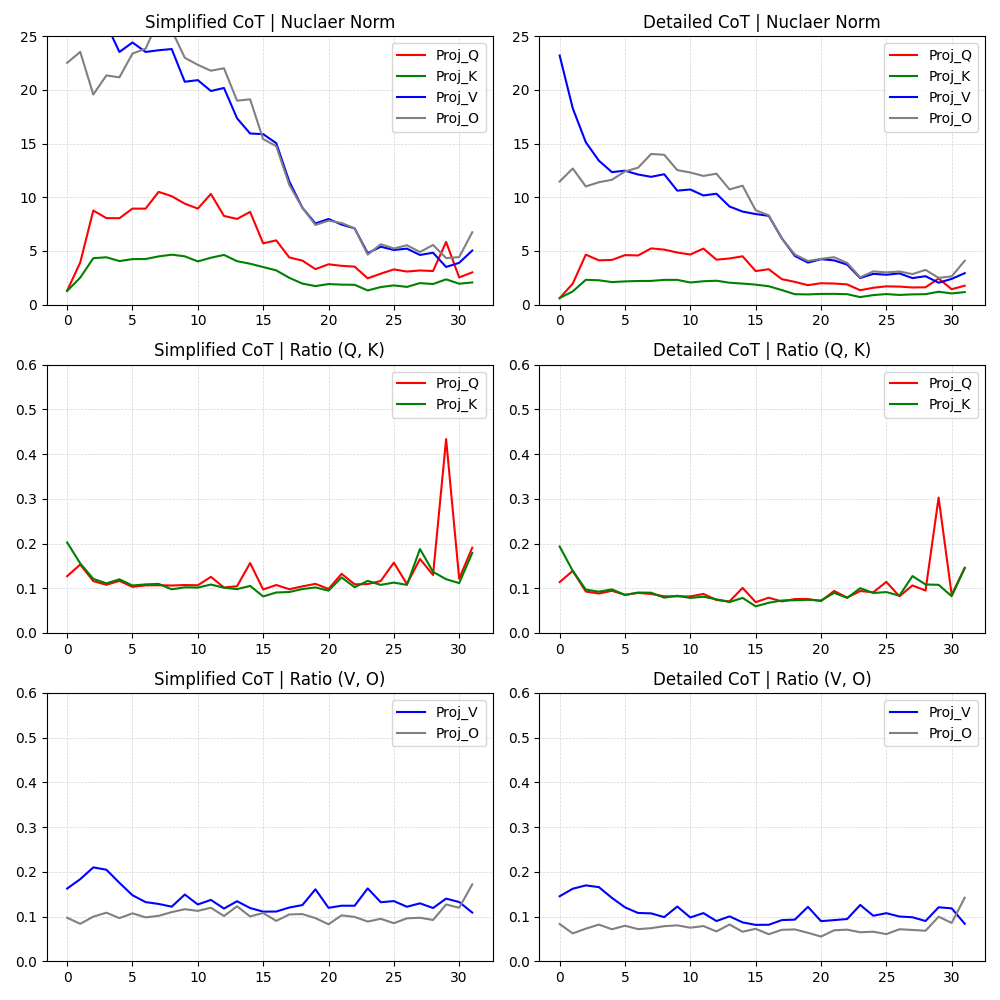}
\caption{Visualization for MATH-Algebra using Llama-3.1-8B on irrelevant responses.}
\label{fig:math_nonsense_algebra}
\vspace{-2mm}
\end{figure}

\begin{table}[!tbh]
\centering
\small
\scalebox{0.75}{
\begin{tabular}{l|l|l|cccc}
\toprule
\multirow{2}{*}{\textbf{Dataset}} & \multirow{2}{*}{\textbf{Curve}} & \multirow{2}{*}{\textbf{Cot}} &  \multicolumn{4}{c}{\textbf{Mean Absolute Difference (MAD)}} \\
 & &  & Early & Middle & Last & All\\
\midrule
\multirow{16}{*}{Counting} & \multirow{3}{*}{$s_Q$} 
  & Simplified & 1.67 & 1.26 & 1.12 & 1.28 \\
& & Detailed & 0.83 & 0.54 & 0.36 & 0.54 \\
\cmidrule{2-7}
& \multirow{3}{*}{$s_K$} 
  & Simplified & 0.56 & 0.44 & 0.33 & 0.43 \\
& & Detailed & 0.30 & 0.17 & 0.15 & 0.20 \\
\cmidrule{2-7}
& \multirow{3}{*}{$s_V$} 
  & Simplified & 4.09 & 1.77 & 0.69 & 1.97 \\
& & Detailed & 1.77 & 0.79 & 0.47 & 0.93 \\
\cmidrule{2-7}
& \multirow{3}{*}{$s_O$} 
  & Simplified & 1.99 & 1.81 & 0.94 & 1.48 \\
& & Detailed & 0.96 & 1.05 & 0.63 & 0.83 \\
\cmidrule{2-7}
 & \multirow{3}{*}{$r_Q$} 
  & Simplified & 0.01 & 0.03 & 0.10 & 0.05 \\
& & Detailed & 0.01 & 0.02 & 0.07 & 0.03 \\
\cmidrule{2-7}
& \multirow{3}{*}{$r_K$} 
  & Simplified & 0.01 & 0.01 & 0.03 & 0.02 \\
& & Detailed & 0.01 & 0.01 & 0.02 & 0.01 \\
\cmidrule{2-7}
 & \multirow{3}{*}{$r_V$} 
  & Simplified & 0.02 & 0.01 & 0.01 & 0.02 \\
& & Detailed & 0.01 & 0.01 & 0.01 & 0.01 \\
\cmidrule{2-7}
& \multirow{3}{*}{$r_O$} 
  & Simplified & 0.01 & 0.01 & 0.01 & 0.01 \\
& & Detailed & 0.01 & 0.01 & 0.01 & 0.01 \\
\bottomrule
\end{tabular}
}
\caption{
Statistical results for MATH-Counting using Llama-3.1-8B on irrelevant responses.
}
\label{tbl:math_lighteval_math_wrong_answer_shuffle_llama3_8b_grads_counting_probability}
\end{table}

\begin{figure}[!h]
\centering
\includegraphics[width=\linewidth]{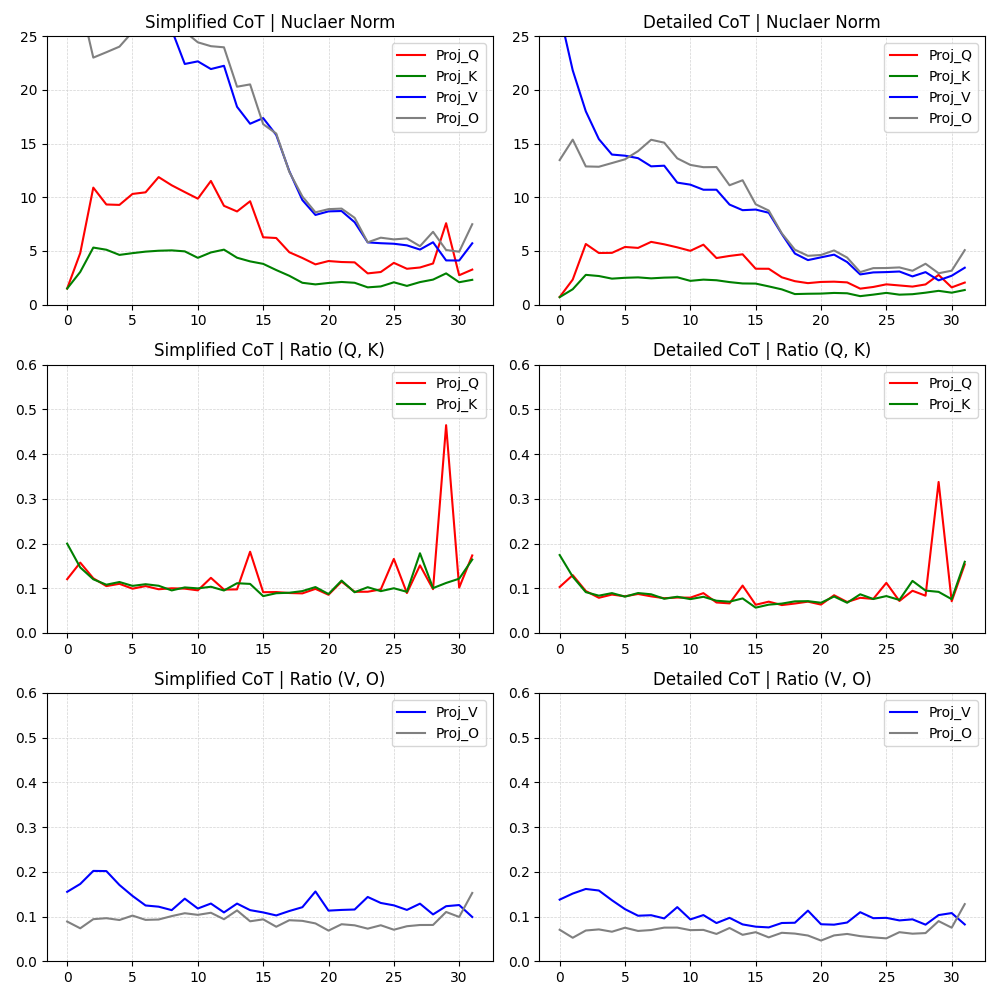}
\caption{Visualization for MATH-Counting using Llama-3.1-8B on irrelevant responses.}
\label{fig:math_nonsense_counting}
\vspace{-2mm}
\end{figure}

\begin{table}[!tbh]
\centering
\small
\scalebox{0.75}{
\begin{tabular}{l|l|l|cccc}
\toprule
\multirow{2}{*}{\textbf{Dataset}} & \multirow{2}{*}{\textbf{Curve}} & \multirow{2}{*}{\textbf{Cot}} &  \multicolumn{4}{c}{\textbf{Mean Absolute Difference (MAD)}} \\
 & &  & Early & Middle & Last & All\\
\midrule
\multirow{16}{*}{Geometry} & \multirow{3}{*}{$s_Q$} 
  & Simplified & 1.50 & 1.43 & 1.20 & 1.29 \\
& & Detailed & 0.86 & 0.52 & 0.47 & 0.57 \\
\cmidrule{2-7}
& \multirow{3}{*}{$s_K$} 
  & Simplified & 0.52 & 0.47 & 0.36 & 0.43 \\
& & Detailed & 0.27 & 0.17 & 0.23 & 0.22 \\
\cmidrule{2-7}
& \multirow{3}{*}{$s_V$} 
  & Simplified & 3.97 & 2.00 & 0.83 & 2.08 \\
& & Detailed & 2.19 & 0.85 & 0.54 & 1.08 \\
\cmidrule{2-7}
& \multirow{3}{*}{$s_O$} 
  & Simplified & 2.29 & 2.12 & 1.10 & 1.68 \\
& & Detailed & 1.21 & 1.16 & 0.72 & 0.96 \\
\cmidrule{2-7}
 & \multirow{3}{*}{$r_Q$} 
  & Simplified & 0.01 & 0.02 & 0.09 & 0.04 \\
& & Detailed & 0.01 & 0.01 & 0.06 & 0.03 \\
\cmidrule{2-7}
& \multirow{3}{*}{$r_K$} 
  & Simplified & 0.02 & 0.01 & 0.02 & 0.02 \\
& & Detailed & 0.01 & 0.01 & 0.03 & 0.02 \\
\cmidrule{2-7}
 & \multirow{3}{*}{$r_V$} 
  & Simplified & 0.02 & 0.01 & 0.01 & 0.02 \\
& & Detailed & 0.02 & 0.01 & 0.02 & 0.02 \\
\cmidrule{2-7}
& \multirow{3}{*}{$r_O$} 
  & Simplified & 0.01 & 0.01 & 0.01 & 0.01 \\
& & Detailed & 0.01 & 0.01 & 0.01 & 0.01 \\
\bottomrule
\end{tabular}
}
\caption{
Statistical results for MATH-Geometry using Llama-3.1-8B on irrelevant responses.
}
\label{tbl:math_lighteval_math_wrong_answer_shuffle_llama3_8b_grads_geometry}
\end{table}

\begin{figure}[!h]
\centering
\includegraphics[width=\linewidth]{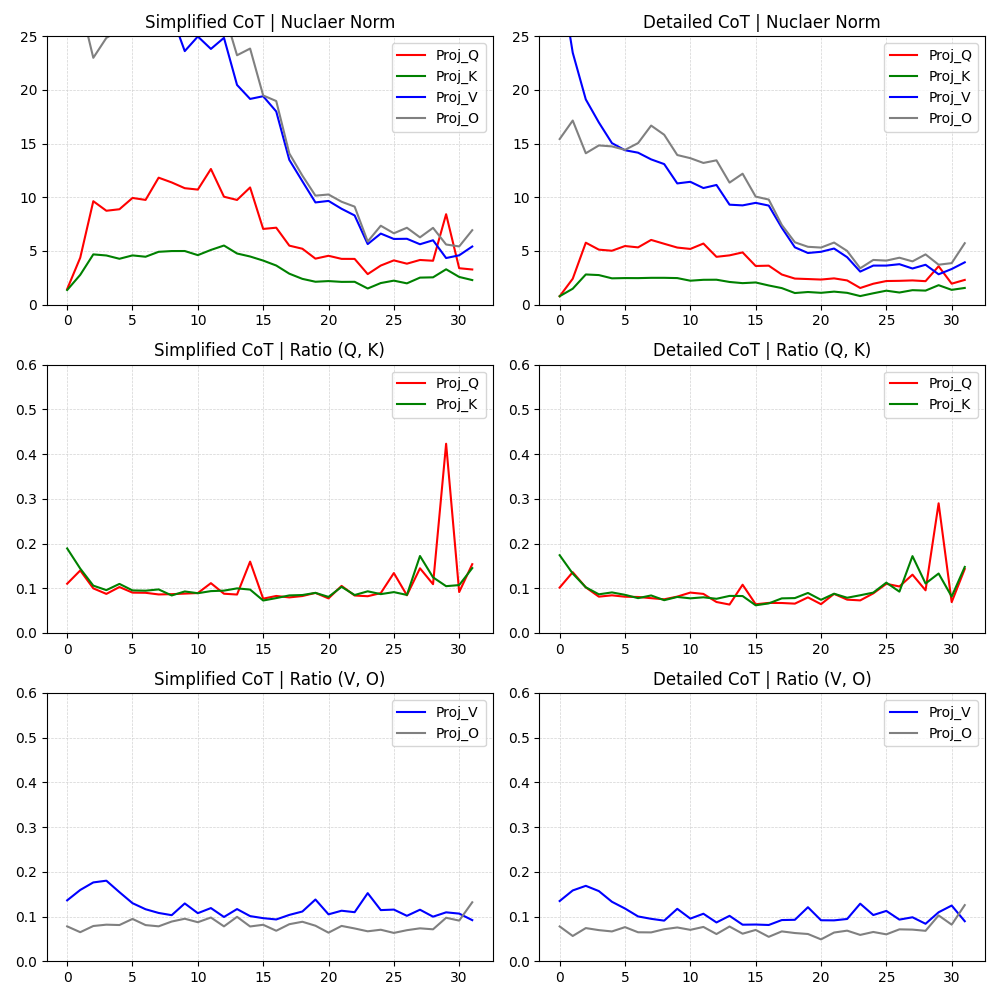}
\caption{Visualization for MATH-Geometry using Llama-3.1-8B on irrelevant responses.}
\label{fig:math_nonsense_geometry}
\vspace{-2mm}
\end{figure}

The visualizations and statistical results on other reasoning tasks: AQuA (Figure \ref{fig:reasoning_nonsense_aqua}, Table \ref{tbl:reasoning_cot_500_wrong_answer_shuffle_llama3_8b_grads_aqua_train}), GSM8K (Figure \ref{fig:reasoning_nonsense_gsm8k}, Table \ref{tbl:reasoning_cot_500_wrong_answer_shuffle_llama3_8b_grads_gsm8k_train}), StrategyQA (Figure \ref{fig:reasoning_nonsense_strategyqa}, Table \ref{tbl:reasoning_cot_500_wrong_answer_shuffle_llama3_8b_grads_strategyqa_train}), ECQA (Figure \ref{fig:reasoning_nonsense_ecqa}, Table \ref{tbl:reasoning_cot_500_wrong_answer_shuffle_llama3_8b_grads_ecqa_train}), CREAK (Figure \ref{fig:reasoning_nonsense_creak}, Table \ref{tbl:reasoning_cot_500_wrong_answer_shuffle_llama3_8b_grads_creak_train}), Sensemaking (Figure \ref{fig:reasoning_nonsense_sensemaking}, Table \ref{tbl:reasoning_cot_500_wrong_answer_shuffle_llama3_8b_grads_sensemaking_train}).

\begin{table}[!tbh]
\centering
\small
\scalebox{0.75}{
\begin{tabular}{l|l|l|cccc}
\toprule
\multirow{2}{*}{\textbf{Dataset}} & \multirow{2}{*}{\textbf{Curve}} & \multirow{2}{*}{\textbf{Cot}} &  \multicolumn{4}{c}{\textbf{Mean Absolute Difference (MAD)}} \\
 & &  & Early & Middle & Last & All\\
\midrule
\multirow{24}{*}{AQuA} & \multirow{3}{*}{$s_Q$} 
  & None & 4.98 & 4.67 & 5.91 & 5.20 \\
& & Simplified & 1.90 & 1.47 & 1.07 & 1.40 \\
& & Detailed & 0.95 & 0.61 & 0.52 & 0.66 \\
\cmidrule{2-7}
& \multirow{3}{*}{$s_K$} 
  & None & 2.50 & 1.54 & 3.45 & 2.72 \\
& & Simplified & 0.68 & 0.47 & 0.34 & 0.48 \\
& & Detailed & 0.36 & 0.19 & 0.18 & 0.24 \\
\cmidrule{2-7}
& \multirow{3}{*}{$s_V$} 
  & None & 23.13 & 5.89 & 3.67 & 10.02 \\
& & Simplified & 6.41 & 1.96 & 0.93 & 2.81 \\
& & Detailed & 2.42 & 0.86 & 0.54 & 1.17 \\
\cmidrule{2-7}
& \multirow{3}{*}{$s_O$} 
  & None & 5.71 & 5.32 & 2.75 & 4.42 \\
& & Simplified & 1.95 & 2.08 & 1.14 & 1.63 \\
& & Detailed & 1.04 & 1.16 & 0.72 & 0.92 \\
\cmidrule{2-7}
 & \multirow{3}{*}{$r_Q$} 
  & None & 0.02 & 0.08 & 0.12 & 0.07 \\
& & Simplified & 0.01 & 0.02 & 0.07 & 0.03 \\
& & Detailed & 0.01 & 0.01 & 0.07 & 0.03 \\
\cmidrule{2-7}
& \multirow{3}{*}{$r_K$} 
  & None & 0.02 & 0.04 & 0.13 & 0.07 \\
& & Simplified & 0.01 & 0.01 & 0.03 & 0.02 \\
& & Detailed & 0.01 & 0.01 & 0.02 & 0.01 \\
\cmidrule{2-7}
 & \multirow{3}{*}{$r_V$} 
  & None & 0.02 & 0.05 & 0.04 & 0.04 \\
& & Simplified & 0.02 & 0.02 & 0.01 & 0.02 \\
& & Detailed & 0.01 & 0.01 & 0.01 & 0.01 \\
\cmidrule{2-7}
& \multirow{3}{*}{$r_O$} 
  & None & 0.02 & 0.04 & 0.08 & 0.05 \\
& & Simplified & 0.01 & 0.02 & 0.01 & 0.01 \\
& & Detailed & 0.01 & 0.01 & 0.01 & 0.01 \\
\bottomrule
\end{tabular}
}
\caption{
Statistical results for AQuA using Llama-3.1-8B on irrelevant responses.
}
\label{tbl:reasoning_cot_500_wrong_answer_shuffle_llama3_8b_grads_aqua_train}
\end{table}

\begin{figure*}[!h]
\centering
\includegraphics[width=\linewidth]{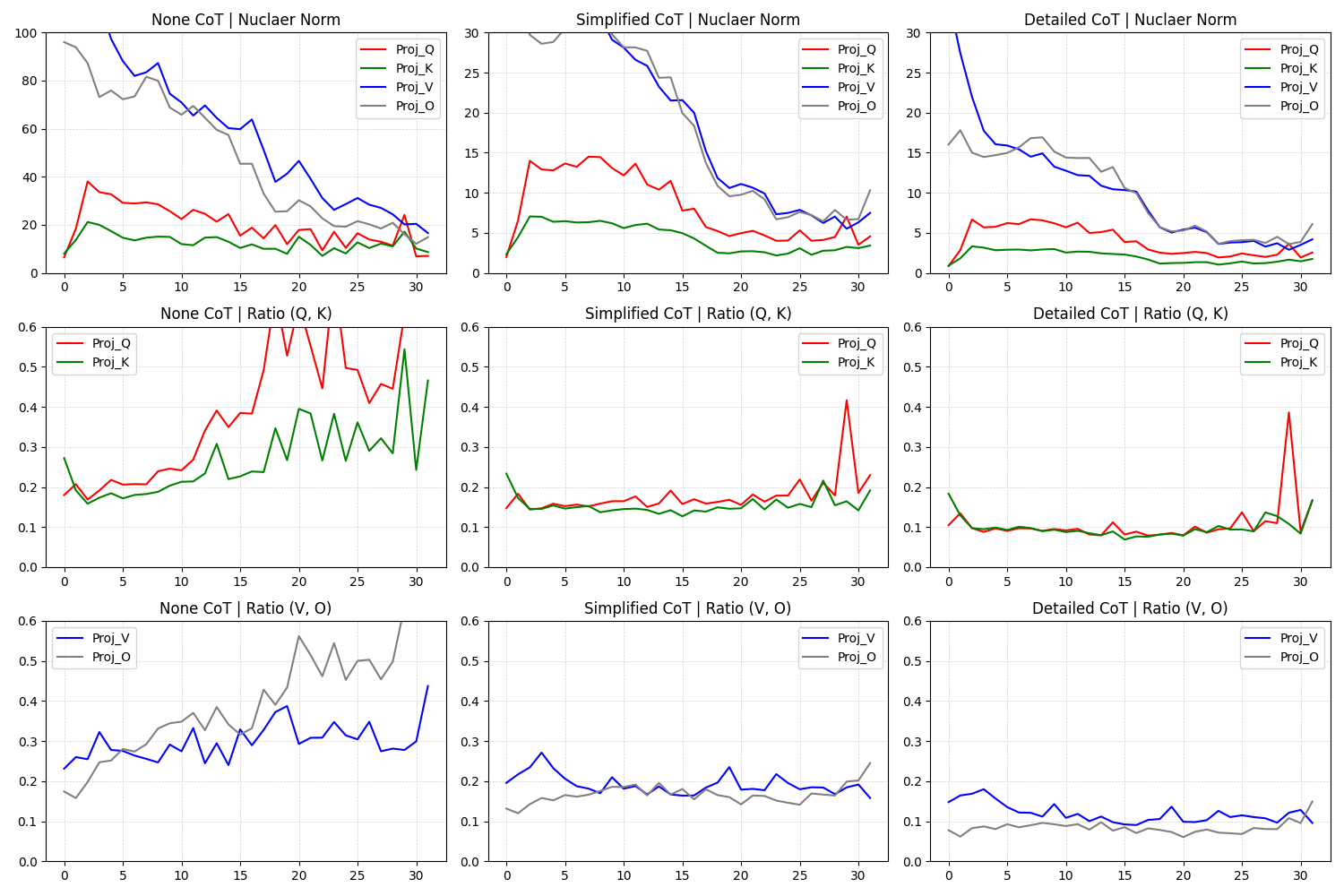}
\caption{Visualization for AQuA using Llama-3.1-8B on irrelevant responses.}
\label{fig:reasoning_nonsense_aqua}
\vspace{-2mm}
\end{figure*}

\begin{table}[!tbh]
\centering
\small
\scalebox{0.75}{
\begin{tabular}{l|l|l|cccc}
\toprule
\multirow{2}{*}{\textbf{Dataset}} & \multirow{2}{*}{\textbf{Curve}} & \multirow{2}{*}{\textbf{Cot}} &  \multicolumn{4}{c}{\textbf{Mean Absolute Difference (MAD)}} \\
 & &  & Early & Middle & Last & All\\
\midrule
\multirow{24}{*}{GSM8K} & \multirow{3}{*}{$s_Q$} 
  & None & 5.95 & 7.50 & 3.86 & 5.43 \\
& & Simplified & 1.50 & 1.00 & 0.72 & 0.99 \\
& & Detailed & 0.84 & 0.52 & 0.32 & 0.52 \\
\cmidrule{2-7}
& \multirow{3}{*}{$s_K$} 
  & None & 2.66 & 2.20 & 2.29 & 2.40 \\
& & Simplified & 0.45 & 0.35 & 0.23 & 0.33 \\
& & Detailed & 0.30 & 0.17 & 0.13 & 0.19 \\
\cmidrule{2-7}
& \multirow{3}{*}{$s_V$} 
  & None & 21.03 & 5.28 & 5.89 & 10.00 \\
& & Simplified & 3.16 & 1.43 & 0.68 & 1.60 \\
& & Detailed & 1.88 & 0.73 & 0.49 & 0.95 \\
\cmidrule{2-7}
& \multirow{3}{*}{$s_O$} 
  & None & 5.37 & 5.16 & 2.62 & 4.25 \\
& & Simplified & 1.50 & 1.52 & 0.89 & 1.22 \\
& & Detailed & 0.88 & 1.00 & 0.67 & 0.80 \\
\cmidrule{2-7}
 & \multirow{3}{*}{$r_Q$} 
  & None & 0.02 & 0.07 & 0.10 & 0.07 \\
& & Simplified & 0.01 & 0.02 & 0.08 & 0.04 \\
& & Detailed & 0.01 & 0.01 & 0.06 & 0.03 \\
\cmidrule{2-7}
& \multirow{3}{*}{$r_K$} 
  & None & 0.02 & 0.05 & 0.03 & 0.03 \\
& & Simplified & 0.02 & 0.01 & 0.02 & 0.02 \\
& & Detailed & 0.01 & 0.01 & 0.02 & 0.01 \\
\cmidrule{2-7}
 & \multirow{3}{*}{$r_V$} 
  & None & 0.02 & 0.03 & 0.02 & 0.03 \\
& & Simplified & 0.02 & 0.01 & 0.01 & 0.02 \\
& & Detailed & 0.01 & 0.01 & 0.01 & 0.01 \\
\cmidrule{2-7}
& \multirow{3}{*}{$r_O$} 
  & None & 0.03 & 0.03 & 0.07 & 0.04 \\
& & Simplified & 0.01 & 0.02 & 0.02 & 0.02 \\
& & Detailed & 0.01 & 0.01 & 0.01 & 0.01 \\
\bottomrule
\end{tabular}
}
\caption{
Statistical results for GSM8K using Llama-3.1-8B on irrelevant responses.
}
\label{tbl:reasoning_cot_500_wrong_answer_shuffle_llama3_8b_grads_gsm8k_train}
\end{table}

\begin{figure*}[!h]
\centering
\includegraphics[width=\linewidth]{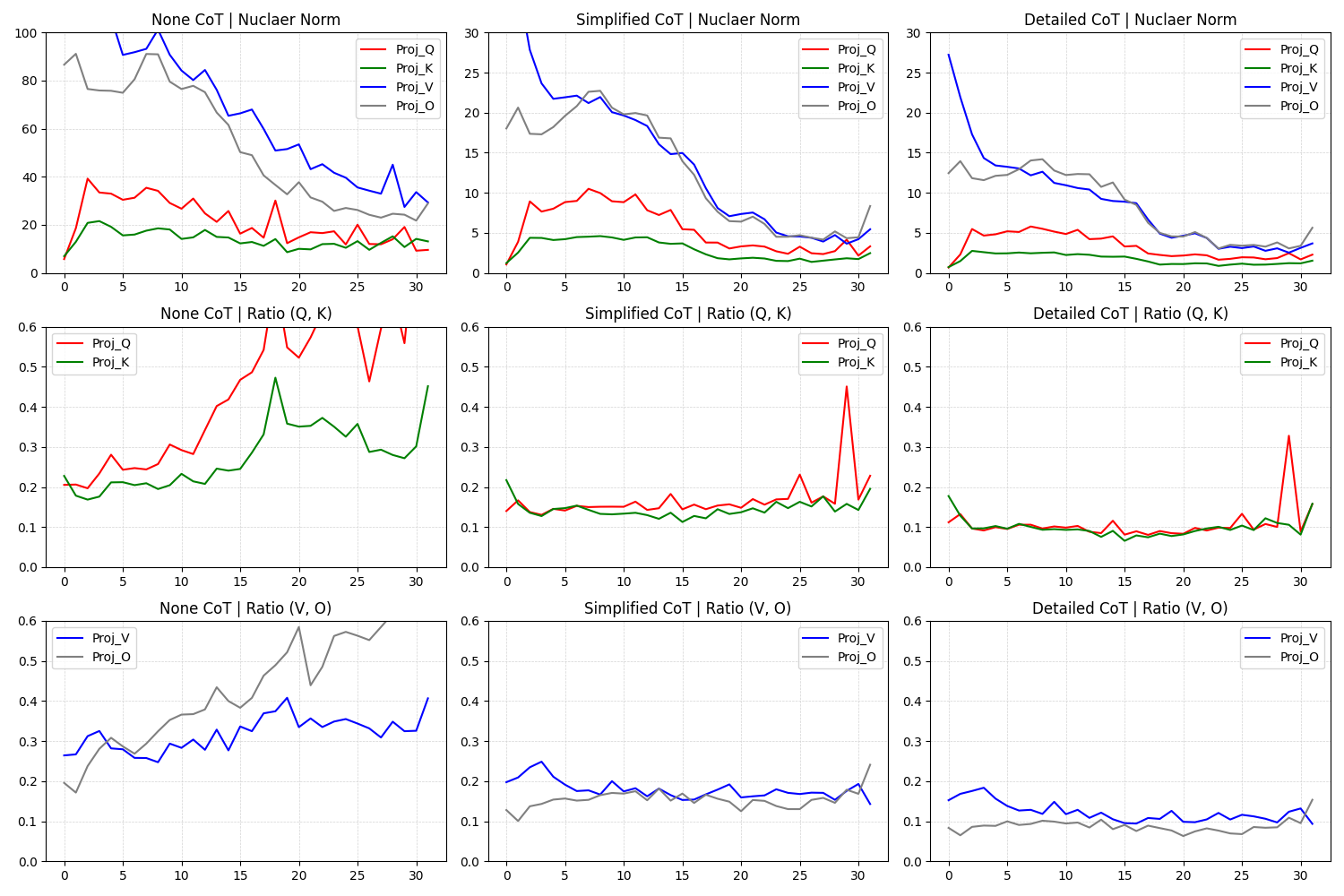}
\caption{Visualization for GSM8K using Llama-3.1-8B on irrelevant responses.}
\label{fig:reasoning_nonsense_gsm8k}
\vspace{-2mm}
\end{figure*}

\begin{table}[!tbh]
\centering
\small
\scalebox{0.75}{
\begin{tabular}{l|l|l|cccc}
\toprule
\multirow{2}{*}{\textbf{Dataset}} & \multirow{2}{*}{\textbf{Curve}} & \multirow{2}{*}{\textbf{Cot}} &  \multicolumn{4}{c}{\textbf{Mean Absolute Difference (MAD)}} \\
 & &  & Early & Middle & Last & All\\
\midrule
\multirow{24}{*}{StrategyQA} & \multirow{3}{*}{$s_Q$} 
  & None & 5.11 & 3.77 & 1.10 & 3.22 \\
& & Simplified & 1.35 & 0.92 & 0.74 & 0.96 \\
& & Detailed & 0.90 & 0.50 & 0.38 & 0.56 \\
\cmidrule{2-7}
& \multirow{3}{*}{$s_K$} 
  & None & 2.53 & 1.38 & 1.11 & 1.73 \\
& & Simplified & 0.68 & 0.37 & 0.29 & 0.43 \\
& & Detailed & 0.36 & 0.21 & 0.16 & 0.24 \\
\cmidrule{2-7}
& \multirow{3}{*}{$s_V$} 
  & None & 24.23 & 6.14 & 2.28 & 9.87 \\
& & Simplified & 5.56 & 1.39 & 0.94 & 2.41 \\
& & Detailed & 2.46 & 0.87 & 0.70 & 1.24 \\
\cmidrule{2-7}
& \multirow{3}{*}{$s_O$} 
  & None & 9.31 & 4.86 & 2.60 & 5.28 \\
& & Simplified & 2.32 & 1.22 & 1.22 & 1.53 \\
& & Detailed & 1.43 & 1.01 & 0.97 & 1.09 \\
\cmidrule{2-7}
 & \multirow{3}{*}{$r_Q$} 
  & None & 0.02 & 0.06 & 0.08 & 0.05 \\
& & Simplified & 0.01 & 0.02 & 0.04 & 0.03 \\
& & Detailed & 0.01 & 0.01 & 0.06 & 0.03 \\
\cmidrule{2-7}
& \multirow{3}{*}{$r_K$} 
  & None & 0.02 & 0.03 & 0.06 & 0.04 \\
& & Simplified & 0.01 & 0.02 & 0.02 & 0.01 \\
& & Detailed & 0.01 & 0.01 & 0.02 & 0.01 \\
\cmidrule{2-7}
 & \multirow{3}{*}{$r_V$} 
  & None & 0.04 & 0.03 & 0.04 & 0.04 \\
& & Simplified & 0.03 & 0.02 & 0.02 & 0.03 \\
& & Detailed & 0.02 & 0.01 & 0.01 & 0.01 \\
\cmidrule{2-7}
& \multirow{3}{*}{$r_O$} 
  & None & 0.03 & 0.03 & 0.04 & 0.03 \\
& & Simplified & 0.01 & 0.02 & 0.02 & 0.02 \\
& & Detailed & 0.01 & 0.01 & 0.01 & 0.01 \\
\bottomrule
\end{tabular}
}
\caption{
Statistical results for StrategyQA using Llama-3.1-8B on irrelevant responses.
}
\label{tbl:reasoning_cot_500_wrong_answer_shuffle_llama3_8b_grads_strategyqa_train}
\end{table}

\begin{figure*}[!h]
\centering
\includegraphics[width=\linewidth]{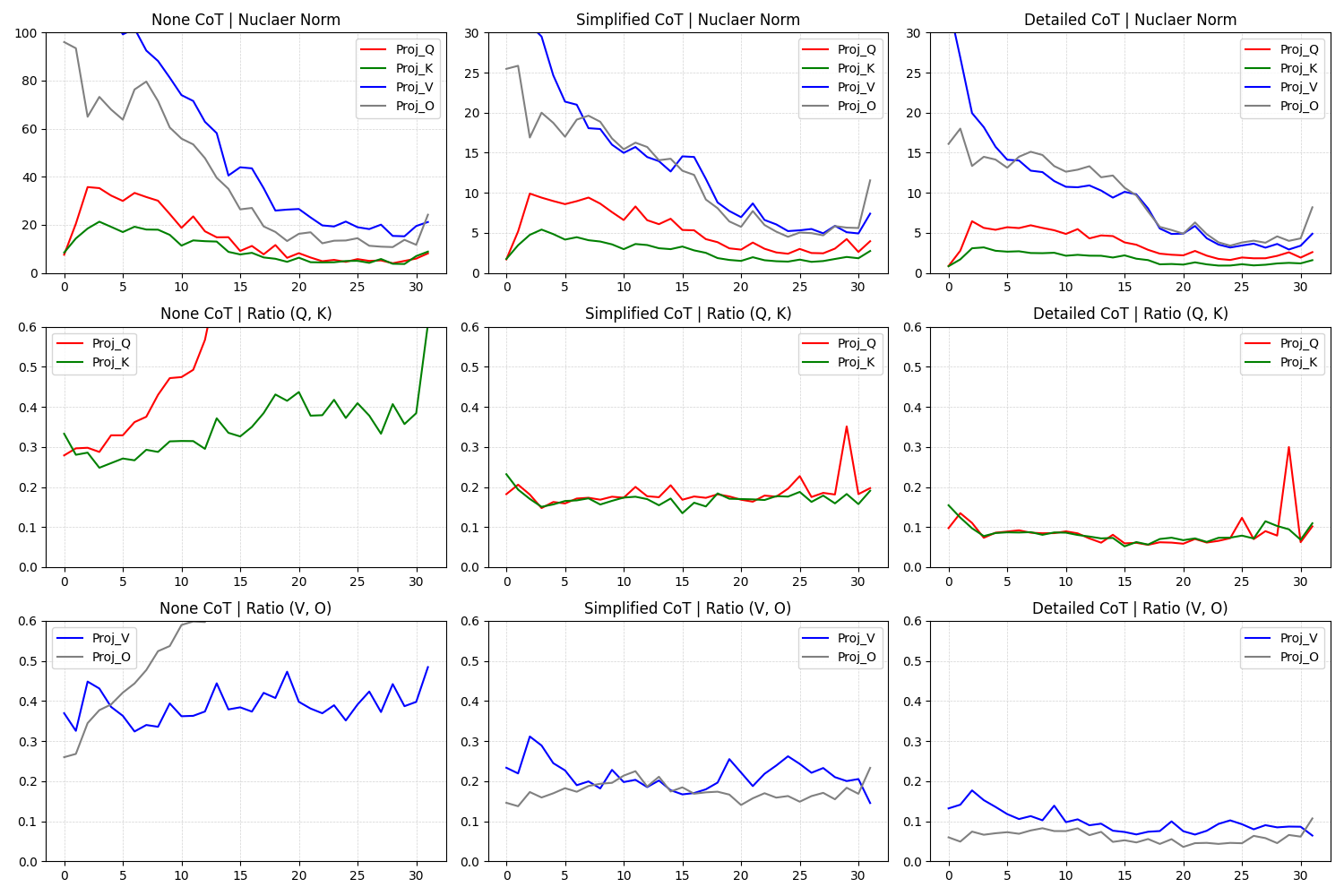}
\caption{Visualization for StrategyQA using Llama-3.1-8B on irrelevant responses.}
\label{fig:reasoning_nonsense_strategyqa}
\vspace{-2mm}
\end{figure*}

\begin{table}[!tbh]
\centering
\small
\scalebox{0.75}{
\begin{tabular}{l|l|l|cccc}
\toprule
\multirow{2}{*}{\textbf{Dataset}} & \multirow{2}{*}{\textbf{Curve}} & \multirow{2}{*}{\textbf{Cot}} &  \multicolumn{4}{c}{\textbf{Mean Absolute Difference (MAD)}} \\
 & &  & Early & Middle & Last & All\\
\midrule
\multirow{24}{*}{ECQA} & \multirow{3}{*}{$s_Q$} 
  & None & 6.06 & 3.84 & 3.92 & 4.40 \\
& & Simplified & 1.68 & 1.13 & 0.86 & 1.16 \\
& & Detailed & 1.09 & 0.52 & 0.53 & 0.68 \\
\cmidrule{2-7}
& \multirow{3}{*}{$s_K$} 
  & None & 2.57 & 0.95 & 1.93 & 1.82 \\
& & Simplified & 0.65 & 0.36 & 0.33 & 0.44 \\
& & Detailed & 0.38 & 0.25 & 0.23 & 0.28 \\
\cmidrule{2-7}
& \multirow{3}{*}{$s_V$} 
  & None & 24.55 & 4.13 & 2.53 & 9.50 \\
& & Simplified & 5.98 & 1.41 & 1.04 & 2.55 \\
& & Detailed & 2.33 & 0.92 & 0.79 & 1.27 \\
\cmidrule{2-7}
& \multirow{3}{*}{$s_O$} 
  & None & 5.68 & 4.69 & 2.78 & 4.29 \\
& & Simplified & 1.55 & 1.35 & 1.23 & 1.32 \\
& & Detailed & 1.13 & 1.13 & 1.02 & 1.05 \\
\cmidrule{2-7}
 & \multirow{3}{*}{$r_Q$} 
  & None & 0.02 & 0.04 & 0.08 & 0.05 \\
& & Simplified & 0.01 & 0.02 & 0.06 & 0.03 \\
& & Detailed & 0.01 & 0.01 & 0.09 & 0.04 \\
\cmidrule{2-7}
& \multirow{3}{*}{$r_K$} 
  & None & 0.02 & 0.03 & 0.06 & 0.04 \\
& & Simplified & 0.02 & 0.02 & 0.02 & 0.02 \\
& & Detailed & 0.01 & 0.01 & 0.03 & 0.02 \\
\cmidrule{2-7}
 & \multirow{3}{*}{$r_V$} 
  & None & 0.02 & 0.03 & 0.03 & 0.03 \\
& & Simplified & 0.02 & 0.02 & 0.01 & 0.02 \\
& & Detailed & 0.01 & 0.01 & 0.01 & 0.01 \\
\cmidrule{2-7}
& \multirow{3}{*}{$r_O$} 
  & None & 0.03 & 0.04 & 0.05 & 0.04 \\
& & Simplified & 0.01 & 0.02 & 0.02 & 0.02 \\
& & Detailed & 0.01 & 0.01 & 0.01 & 0.01 \\
\bottomrule
\end{tabular}
}
\caption{
Statistical results for ECQA using Llama-3.1-8B on irrelevant responses.
}
\label{tbl:reasoning_cot_500_wrong_answer_shuffle_llama3_8b_grads_ecqa_train}
\end{table}

\begin{figure*}[!h]
\centering
\includegraphics[width=\linewidth]{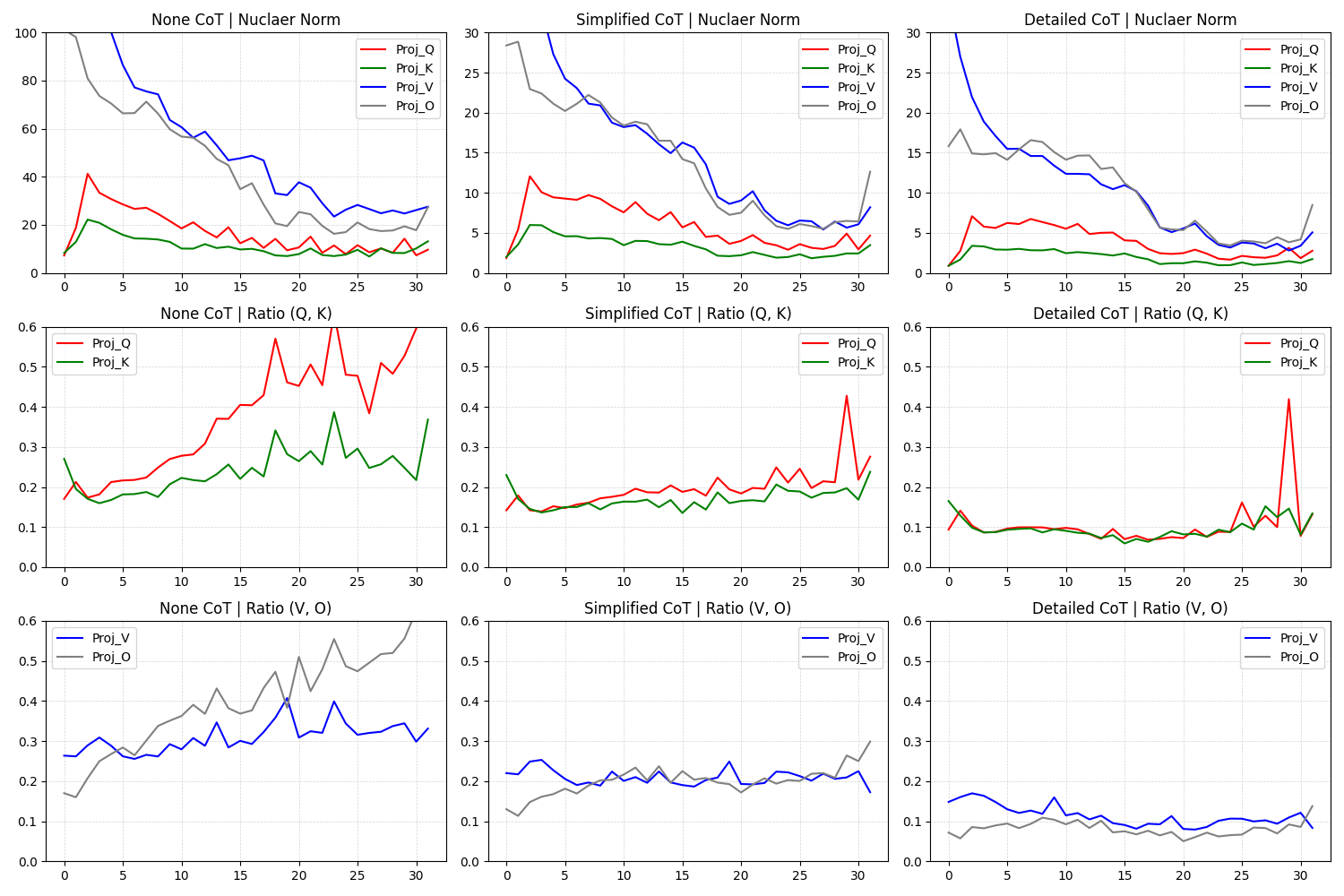}
\caption{Visualization for ECQA using Llama-3.1-8B on irrelevant responses.}
\label{fig:reasoning_nonsense_ecqa}
\vspace{-2mm}
\end{figure*}

\begin{table}[!tbh]
\centering
\small
\scalebox{0.75}{
\begin{tabular}{l|l|l|cccc}
\toprule
\multirow{2}{*}{\textbf{Dataset}} & \multirow{2}{*}{\textbf{Curve}} & \multirow{2}{*}{\textbf{Cot}} &  \multicolumn{4}{c}{\textbf{Mean Absolute Difference (MAD)}} \\
 & &  & Early & Middle & Last & All\\
\midrule
\multirow{24}{*}{CREAK} & \multirow{3}{*}{$s_Q$} 
  & None & 6.75 & 5.46 & 4.91 & 5.54 \\
& & Simplified & 1.52 & 1.27 & 0.80 & 1.15 \\
& & Detailed & 0.96 & 0.51 & 0.42 & 0.60 \\
\cmidrule{2-7}
& \multirow{3}{*}{$s_K$} 
  & None & 3.16 & 2.03 & 3.97 & 3.10 \\
& & Simplified & 0.67 & 0.52 & 0.32 & 0.49 \\
& & Detailed & 0.36 & 0.21 & 0.22 & 0.26 \\
\cmidrule{2-7}
& \multirow{3}{*}{$s_V$} 
  & None & 31.92 & 9.46 & 4.11 & 13.71 \\
& & Simplified & 6.64 & 1.63 & 0.94 & 2.76 \\
& & Detailed & 2.65 & 0.85 & 0.67 & 1.29 \\
\cmidrule{2-7}
& \multirow{3}{*}{$s_O$} 
  & None & 9.59 & 8.89 & 3.67 & 7.20 \\
& & Simplified & 2.17 & 1.58 & 1.29 & 1.60 \\
& & Detailed & 1.35 & 1.05 & 0.93 & 1.07 \\
\cmidrule{2-7}
 & \multirow{3}{*}{$r_Q$} 
  & None & 0.02 & 0.07 & 0.11 & 0.07 \\
& & Simplified & 0.02 & 0.01 & 0.05 & 0.03 \\
& & Detailed & 0.01 & 0.01 & 0.06 & 0.03 \\
\cmidrule{2-7}
& \multirow{3}{*}{$r_K$} 
  & None & 0.02 & 0.03 & 0.06 & 0.04 \\
& & Simplified & 0.02 & 0.02 & 0.02 & 0.02 \\
& & Detailed & 0.01 & 0.01 & 0.03 & 0.02 \\
\cmidrule{2-7}
 & \multirow{3}{*}{$r_V$} 
  & None & 0.02 & 0.04 & 0.04 & 0.04 \\
& & Simplified & 0.02 & 0.02 & 0.02 & 0.02 \\
& & Detailed & 0.02 & 0.01 & 0.01 & 0.01 \\
\cmidrule{2-7}
& \multirow{3}{*}{$r_O$} 
  & None & 0.02 & 0.04 & 0.06 & 0.05 \\
& & Simplified & 0.01 & 0.02 & 0.02 & 0.02 \\
& & Detailed & 0.01 & 0.01 & 0.01 & 0.01 \\
\bottomrule
\end{tabular}
}
\caption{
Statistical results for CREAK using Llama-3.1-8B on irrelevant responses.
}
\label{tbl:reasoning_cot_500_wrong_answer_shuffle_llama3_8b_grads_creak_train}
\end{table}

\begin{figure*}[!h]
\centering
\includegraphics[width=\linewidth]{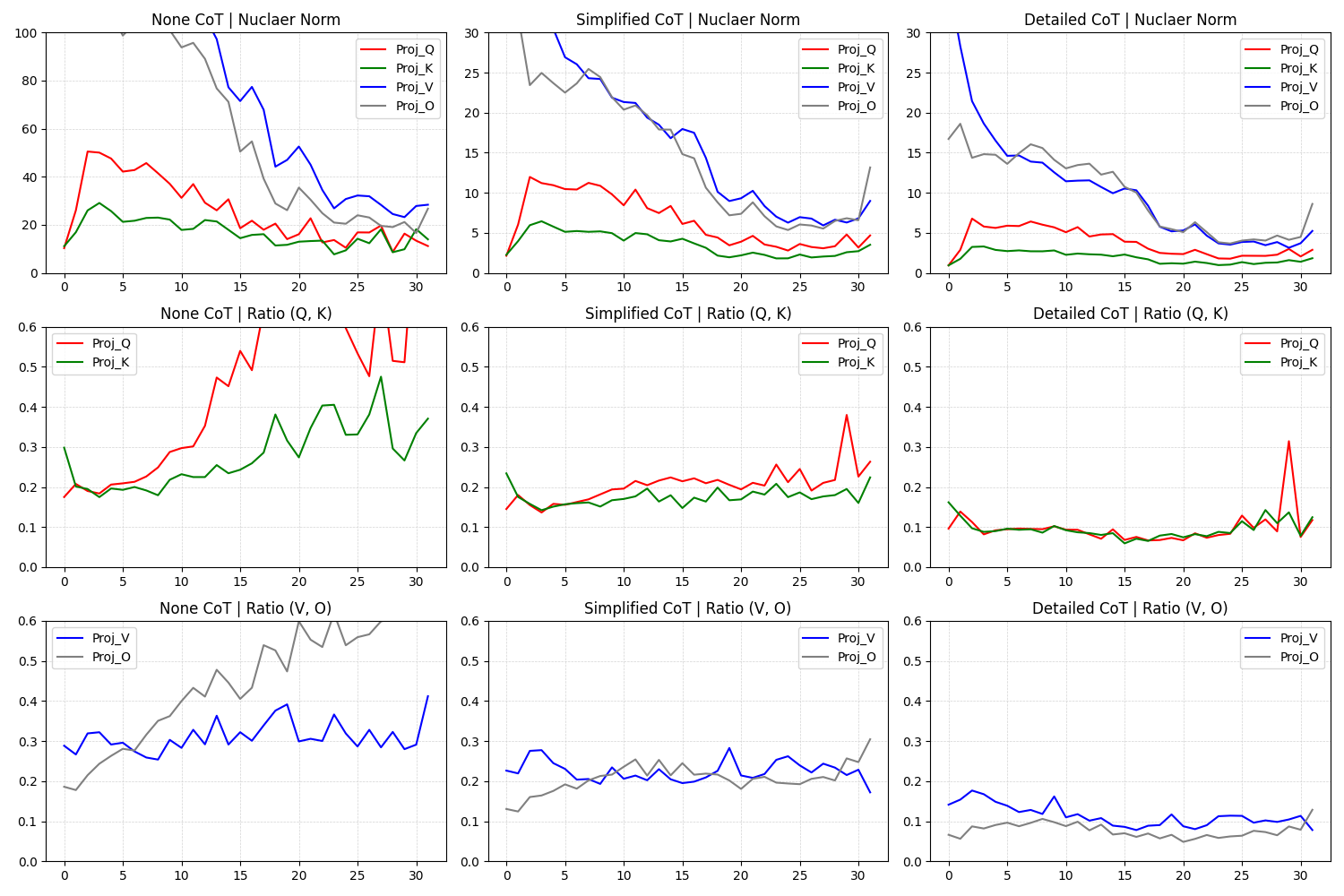}
\caption{Visualization for CREAK using Llama-3.1-8B on irrelevant responses.}
\label{fig:reasoning_nonsense_creak}
\vspace{-2mm}
\end{figure*}

\begin{table}[!tbh]
\centering
\small
\scalebox{0.75}{
\begin{tabular}{l|l|l|cccc}
\toprule
\multirow{2}{*}{\textbf{Dataset}} & \multirow{2}{*}{\textbf{Curve}} & \multirow{2}{*}{\textbf{Cot}} &  \multicolumn{4}{c}{\textbf{Mean Absolute Difference (MAD)}} \\
 & &  & Early & Middle & Last & All\\
\midrule
\multirow{24}{*}{Sensemaking} & \multirow{3}{*}{$s_Q$} 
  & None & 4.73 & 6.45 & 3.71 & 4.75 \\
& & Simplified & 2.06 & 1.95 & 0.76 & 1.52 \\
& & Detailed & 1.00 & 0.61 & 0.57 & 0.69 \\
\cmidrule{2-7}
& \multirow{3}{*}{$s_K$} 
  & None & 2.29 & 2.57 & 2.73 & 2.51 \\
& & Simplified & 0.84 & 0.76 & 0.32 & 0.64 \\
& & Detailed & 0.35 & 0.24 & 0.21 & 0.26 \\
\cmidrule{2-7}
& \multirow{3}{*}{$s_V$} 
  & None & 26.46 & 8.03 & 3.20 & 11.28 \\
& & Simplified & 7.69 & 2.40 & 1.11 & 3.41 \\
& & Detailed & 1.97 & 0.88 & 0.73 & 1.12 \\
\cmidrule{2-7}
& \multirow{3}{*}{$s_O$} 
  & None & 7.81 & 6.87 & 2.25 & 5.46 \\
& & Simplified & 2.37 & 2.30 & 1.35 & 1.94 \\
& & Detailed & 1.02 & 1.22 & 0.93 & 1.01 \\
\cmidrule{2-7}
 & \multirow{3}{*}{$r_Q$} 
  & None & 0.02 & 0.06 & 0.12 & 0.07 \\
& & Simplified & 0.02 & 0.02 & 0.05 & 0.03 \\
& & Detailed & 0.01 & 0.02 & 0.10 & 0.05 \\
\cmidrule{2-7}
& \multirow{3}{*}{$r_K$} 
  & None & 0.03 & 0.05 & 0.11 & 0.06 \\
& & Simplified & 0.02 & 0.03 & 0.04 & 0.03 \\
& & Detailed & 0.01 & 0.01 & 0.04 & 0.02 \\
\cmidrule{2-7}
 & \multirow{3}{*}{$r_V$} 
  & None & 0.03 & 0.04 & 0.03 & 0.04 \\
& & Simplified & 0.02 & 0.02 & 0.02 & 0.02 \\
& & Detailed & 0.01 & 0.01 & 0.01 & 0.01 \\
\cmidrule{2-7}
& \multirow{3}{*}{$r_O$} 
  & None & 0.02 & 0.04 & 0.06 & 0.04 \\
& & Simplified & 0.02 & 0.03 & 0.02 & 0.02 \\
& & Detailed & 0.01 & 0.01 & 0.01 & 0.01 \\
\bottomrule
\end{tabular}
}
\caption{
Statistical results for Sensemaking using Llama-3.1-8B on irrelevant responses.
}
\label{tbl:reasoning_cot_500_wrong_answer_shuffle_llama3_8b_grads_sensemaking_train}
\end{table}

\begin{figure*}[!h]
\centering
\includegraphics[width=\linewidth]{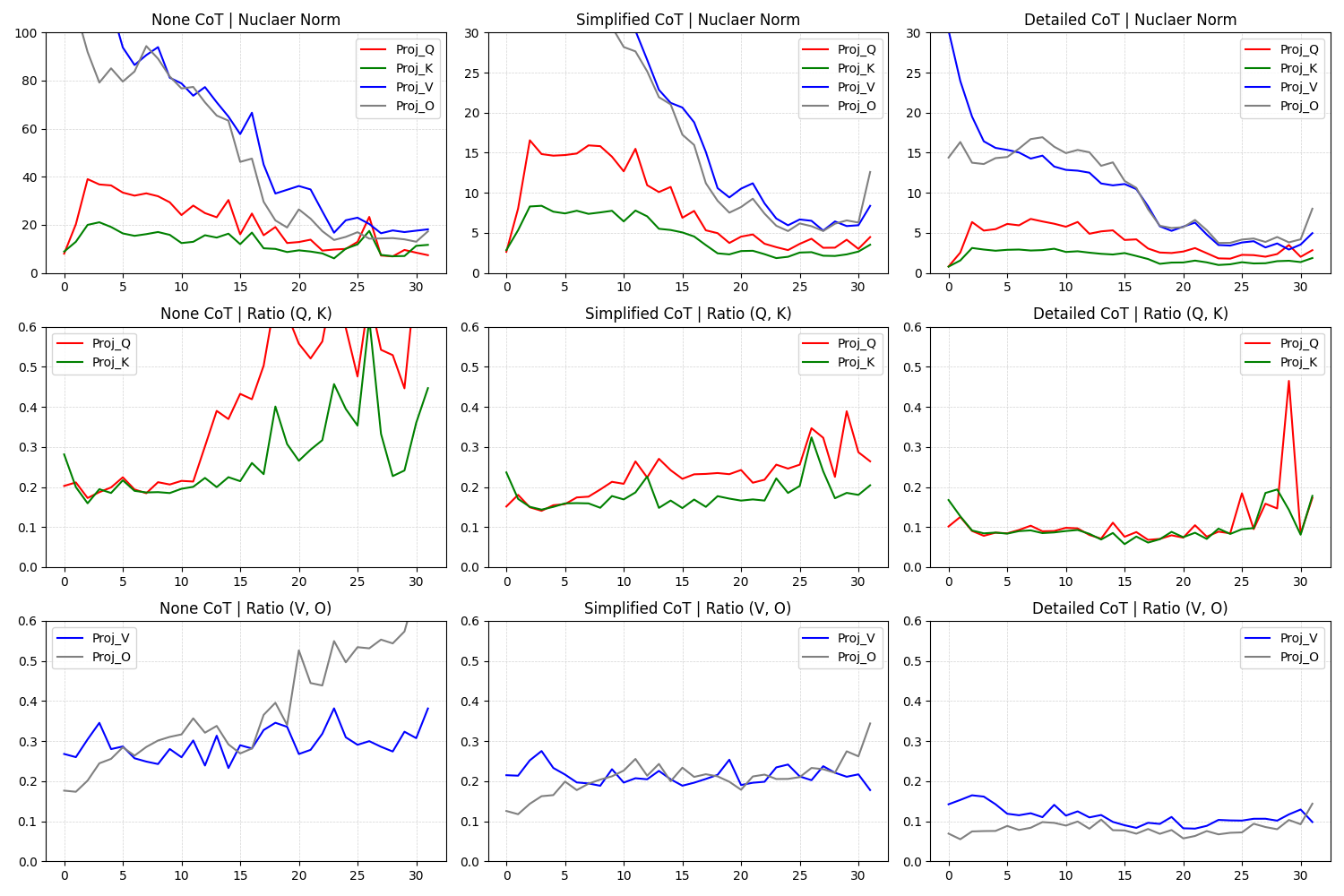}
\caption{Visualization for Sensemaking using Llama-3.1-8B on irrelevant responses.}
\label{fig:reasoning_nonsense_sensemaking}
\vspace{-2mm}
\end{figure*}

\subsubsection{Wiki Tasks}

The visualizations and statistical results on Wiki tasks are shown in Figure \ref{fig:wiki_nonsense} and Table \ref{tbl:wiki_wiki_popularity_wrong_answer_shuffle_llama3_8b_grads}.

\begin{table}[!tbh]
\centering
\small
\scalebox{0.75}{
\begin{tabular}{l|l|l|cccc}
\toprule
\multirow{2}{*}{\textbf{Dataset}} & \multirow{2}{*}{\textbf{Curve}} & \multirow{2}{*}{\textbf{Cot}} &  \multicolumn{4}{c}{\textbf{Mean Absolute Difference (MAD)}} \\
 & &  & Early & Middle & Last & All\\
\midrule
\multirow{32}{*}{Wiki} & \multirow{3}{*}{$s_Q$} 
  & Len 100 & 1.21 & 0.85 & 1.02 & 1.02 \\
& & Len 500 & 0.99 & 0.61 & 0.65 & 0.73 \\
& & Len 1000 & 0.81 & 0.53 & 0.48 & 0.59 \\
& & Unpopular & 1.53 & 1.00 & 1.08 & 1.18 \\
\cmidrule{2-7}
& \multirow{3}{*}{$s_K$} 
  & Len 100 & 0.57 & 0.27 & 0.32 & 0.39 \\
& & Len 500 & 0.44 & 0.20 & 0.24 & 0.29 \\
& & Len 1000 & 0.37 & 0.17 & 0.19 & 0.24 \\
& & Unpopular & 0.75 & 0.30 & 0.41 & 0.49 \\
\cmidrule{2-7}
& \multirow{3}{*}{$s_V$} 
  & Len 100 & 6.03 & 1.13 & 0.84 & 2.47 \\
& & Len 500 & 3.84 & 0.97 & 0.76 & 1.71 \\
& & Len 1000 & 3.06 & 0.86 & 0.67 & 1.41 \\
& & Unpopular & 11.89 & 1.57 & 0.88 & 4.36 \\
\cmidrule{2-7}
& \multirow{3}{*}{$s_O$} 
  & Len 100 & 2.61 & 1.20 & 1.03 & 1.58 \\
& & Len 500 & 2.25 & 1.11 & 0.97 & 1.40 \\
& & Len 1000 & 2.11 & 1.00 & 0.89 & 1.29 \\
& & Unpopular & 4.04 & 1.44 & 1.22 & 2.18 \\
\cmidrule{2-7}
 & \multirow{3}{*}{$r_Q$} 
  & Len 100 & 0.02 & 0.01 & 0.06 & 0.03 \\
& & Len 500 & 0.02 & 0.01 & 0.06 & 0.03 \\
& & Len 1000 & 0.02 & 0.01 & 0.05 & 0.03 \\
& & Unpopular & 0.02 & 0.01 & 0.05 & 0.03 \\
\cmidrule{2-7}
& \multirow{3}{*}{$r_K$} 
  & Len 100 & 0.01 & 0.01 & 0.01 & 0.01 \\
& & Len 500 & 0.01 & 0.01 & 0.01 & 0.01 \\
& & Len 1000 & 0.01 & 0.00 & 0.01 & 0.01 \\
& & Unpopular & 0.02 & 0.01 & 0.01 & 0.01 \\
\cmidrule{2-7}
 & \multirow{3}{*}{$r_V$} 
  & Len 100 & 0.03 & 0.02 & 0.01 & 0.02 \\
& & Len 500 & 0.02 & 0.01 & 0.01 & 0.02 \\
& & Len 1000 & 0.02 & 0.01 & 0.01 & 0.01 \\
& & Unpopular & 0.04 & 0.02 & 0.03 & 0.03 \\
\cmidrule{2-7}
& \multirow{3}{*}{$r_O$} 
  & Len 100 & 0.01 & 0.02 & 0.02 & 0.02 \\
& & Len 500 & 0.01 & 0.01 & 0.01 & 0.01 \\
& & Len 1000 & 0.01 & 0.01 & 0.01 & 0.01 \\
& & Unpopular & 0.01 & 0.03 & 0.01 & 0.02 \\
\bottomrule
\end{tabular}
}
\caption{
Statistical results for Wiki using Llama-3.1-8B on irrelevant responses.
}
\label{tbl:wiki_wiki_popularity_wrong_answer_shuffle_llama3_8b_grads}
\end{table}

\begin{figure*}[!h]
\centering
\includegraphics[width=\linewidth]{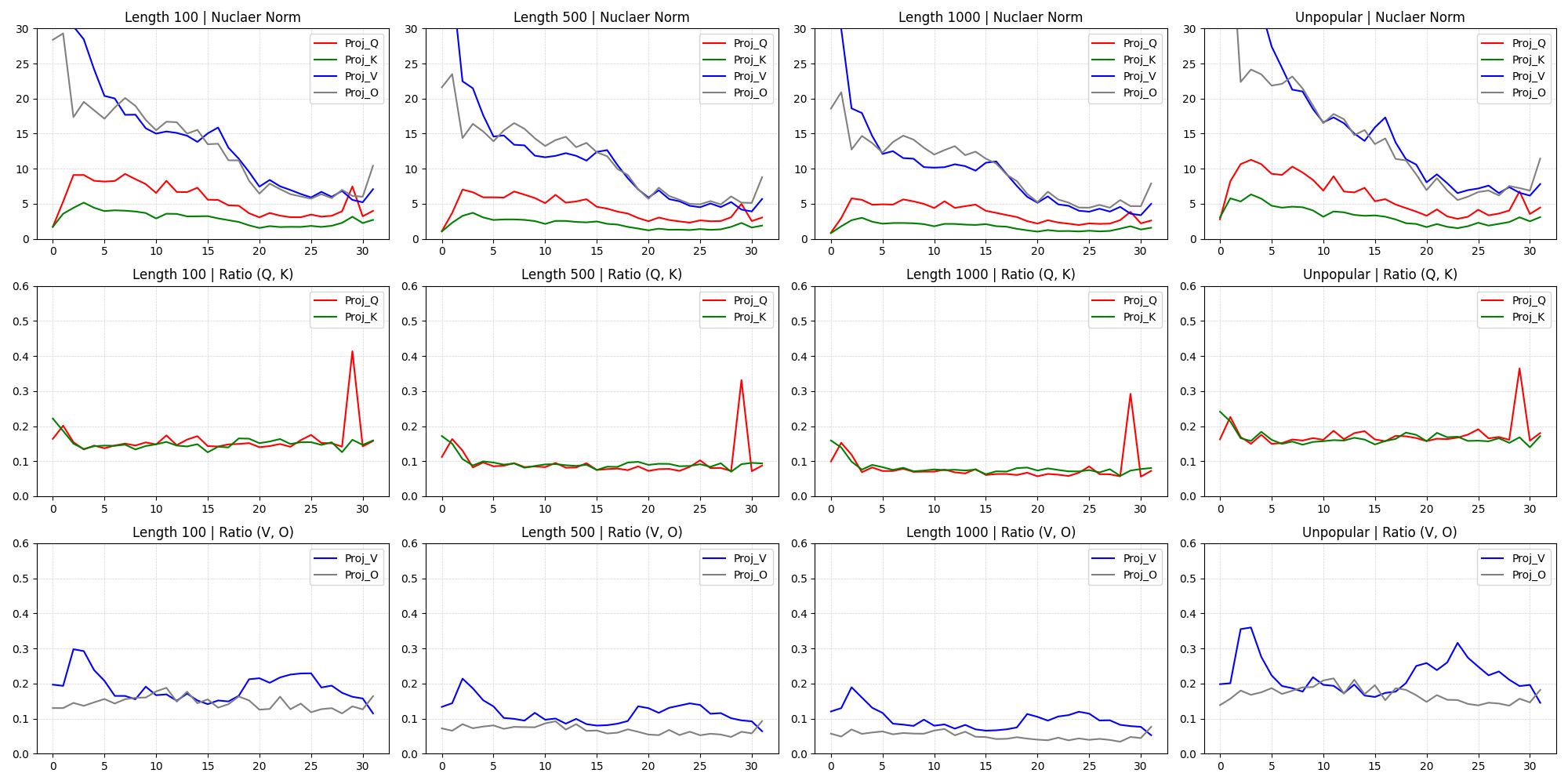}
\caption{Visualization for Wiki tasks using Llama-3.1-8B on irrelevant responses.}
\label{fig:wiki_nonsense}
\vspace{-2mm}
\end{figure*}

\clearpage

\subsection{Instructed LLM on Correct Responses}

\subsubsection{Reasoning Tasks}

The visualizations and statistical results on MATH tasks: MATH-Algebra (Figure \ref{fig:math_correct_algebra_it}, Table \ref{tbl:math_lighteval_math_llama3_8b_it_grads_algebra}), MATH-Counting (Figure \ref{fig:math_correct_counting_it}, Table \ref{tbl:math_lighteval_math_llama3_8b_it_grads_counting_probability}), MATH-Geometry (Figure \ref{fig:math_correct_geometry_it}, Table \ref{tbl:math_lighteval_math_llama3_8b_it_grads_geometry}).

\begin{table}[!tbh]
\centering
\small
\scalebox{0.75}{
\begin{tabular}{l|l|l|cccc}
\toprule
\multirow{2}{*}{\textbf{Dataset}} & \multirow{2}{*}{\textbf{Curve}} & \multirow{2}{*}{\textbf{Cot}} &  \multicolumn{4}{c}{\textbf{Mean Absolute Difference (MAD)}} \\
 & &  & Early & Middle & Last & All\\
\midrule
\multirow{16}{*}{Algebra} & \multirow{3}{*}{$s_Q$} 
  & Simplified & 0.74 & 0.89 & 0.33 & 0.61 \\
& & Detailed & 0.34 & 0.43 & 0.20 & 0.30 \\
\cmidrule{2-7}
& \multirow{3}{*}{$s_K$} 
  & Simplified & 0.36 & 0.33 & 0.11 & 0.27 \\
& & Detailed & 0.16 & 0.12 & 0.07 & 0.11 \\
\cmidrule{2-7}
& \multirow{3}{*}{$s_V$} 
  & Simplified & 2.20 & 1.30 & 0.49 & 1.22 \\
& & Detailed & 0.77 & 0.55 & 0.27 & 0.50 \\
\cmidrule{2-7}
& \multirow{3}{*}{$s_O$} 
  & Simplified & 1.15 & 1.50 & 0.55 & 0.98 \\
& & Detailed & 0.51 & 0.70 & 0.32 & 0.48 \\
\cmidrule{2-7}
 & \multirow{3}{*}{$r_Q$} 
  & Simplified & 0.01 & 0.01 & 0.05 & 0.03 \\
& & Detailed & 0.01 & 0.01 & 0.07 & 0.03 \\
\cmidrule{2-7}
& \multirow{3}{*}{$r_K$} 
  & Simplified & 0.02 & 0.01 & 0.02 & 0.02 \\
& & Detailed & 0.02 & 0.00 & 0.03 & 0.02 \\
\cmidrule{2-7}
 & \multirow{3}{*}{$r_V$} 
  & Simplified & 0.02 & 0.01 & 0.02 & 0.02 \\
& & Detailed & 0.02 & 0.01 & 0.02 & 0.02 \\
\cmidrule{2-7}
& \multirow{3}{*}{$r_O$} 
  & Simplified & 0.01 & 0.01 & 0.01 & 0.01 \\
& & Detailed & 0.01 & 0.01 & 0.01 & 0.01 \\
\bottomrule
\end{tabular}
}
\caption{
Statistical results for MATH-Algebra using Llama-3.1-8B-Instruct on correct responses.
}
\label{tbl:math_lighteval_math_llama3_8b_it_grads_algebra}
\end{table}

\begin{figure}[!h]
\centering
\includegraphics[width=\linewidth]{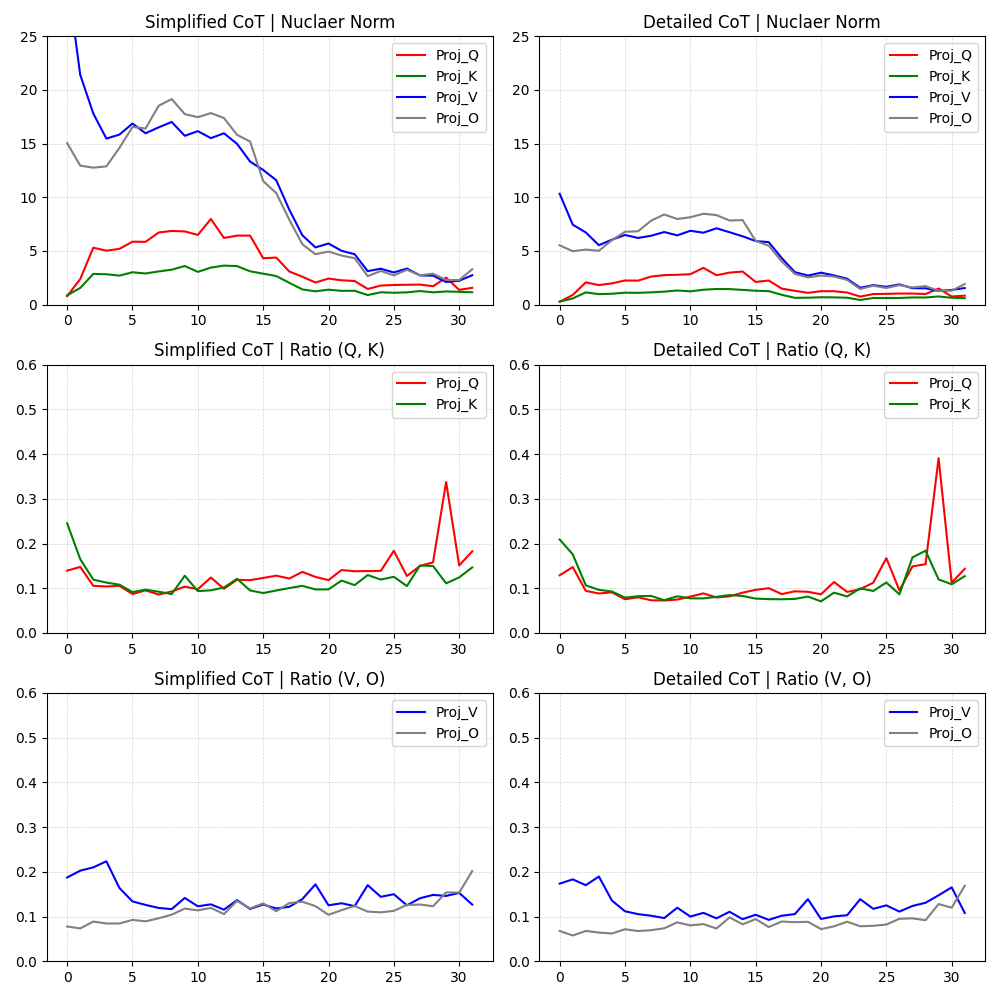}
\caption{Visualization for MATH-Algebra using Llama-3.1-8B-Instruct on correct responses.}
\label{fig:math_correct_algebra_it}
\vspace{-2mm}
\end{figure}

\begin{table}[!tbh]
\centering
\small
\scalebox{0.75}{
\begin{tabular}{l|l|l|cccc}
\toprule
\multirow{2}{*}{\textbf{Dataset}} & \multirow{2}{*}{\textbf{Curve}} & \multirow{2}{*}{\textbf{Cot}} &  \multicolumn{4}{c}{\textbf{Mean Absolute Difference (MAD)}} \\
 & &  & Early & Middle & Last & All\\
\midrule
\multirow{16}{*}{Counting} & \multirow{3}{*}{$s_Q$} 
  & Simplified & 0.91 & 1.00 & 0.40 & 0.72 \\
& & Detailed & 0.41 & 0.48 & 0.21 & 0.34 \\
\cmidrule{2-7}
& \multirow{3}{*}{$s_K$} 
  & Simplified & 0.42 & 0.36 & 0.10 & 0.29 \\
& & Detailed & 0.18 & 0.14 & 0.06 & 0.12 \\
\cmidrule{2-7}
& \multirow{3}{*}{$s_V$} 
  & Simplified & 2.32 & 1.43 & 0.54 & 1.30 \\
& & Detailed & 0.85 & 0.63 & 0.34 & 0.57 \\
\cmidrule{2-7}
& \multirow{3}{*}{$s_O$} 
  & Simplified & 1.21 & 1.73 & 0.58 & 1.08 \\
& & Detailed & 0.56 & 0.86 & 0.40 & 0.56 \\
\cmidrule{2-7}
 & \multirow{3}{*}{$r_Q$} 
  & Simplified & 0.01 & 0.01 & 0.06 & 0.03 \\
& & Detailed & 0.01 & 0.01 & 0.08 & 0.03 \\
\cmidrule{2-7}
& \multirow{3}{*}{$r_K$} 
  & Simplified & 0.02 & 0.01 & 0.02 & 0.02 \\
& & Detailed & 0.02 & 0.00 & 0.02 & 0.01 \\
\cmidrule{2-7}
 & \multirow{3}{*}{$r_V$} 
  & Simplified & 0.02 & 0.01 & 0.01 & 0.02 \\
& & Detailed & 0.02 & 0.01 & 0.02 & 0.02 \\
\cmidrule{2-7}
& \multirow{3}{*}{$r_O$} 
  & Simplified & 0.01 & 0.01 & 0.01 & 0.01 \\
& & Detailed & 0.01 & 0.01 & 0.01 & 0.01 \\
\bottomrule
\end{tabular}
}
\caption{
Statistical results for MATH-Counting using Llama-3.1-8B-Instruct on correct responses.
}
\label{tbl:math_lighteval_math_llama3_8b_it_grads_counting_probability}
\end{table}

\begin{figure}[!h]
\centering
\includegraphics[width=\linewidth]{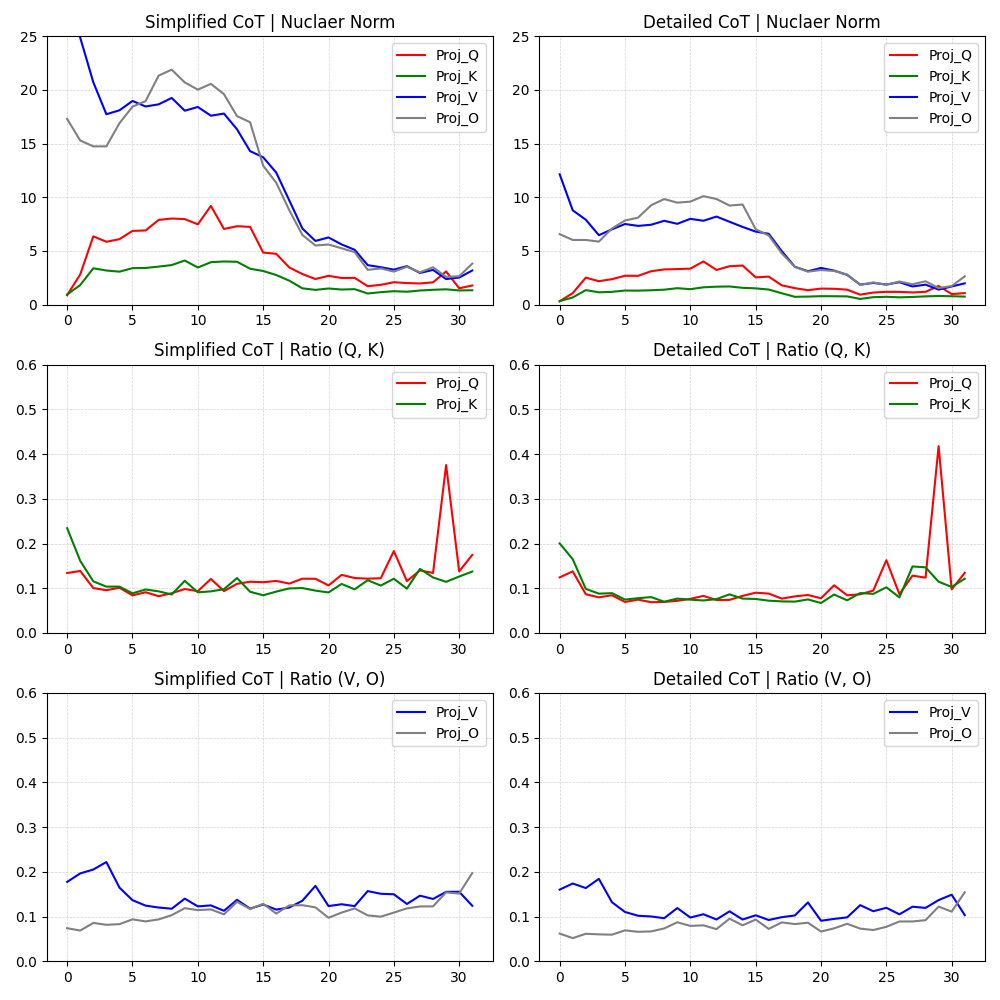}
\caption{Visualization for MATH-Counting using Llama-3.1-8B-Instruct on correct responses.}
\label{fig:math_correct_counting_it}
\vspace{-2mm}
\end{figure}

\begin{table}[!tbh]
\centering
\small
\scalebox{0.75}{
\begin{tabular}{l|l|l|cccc}
\toprule
\multirow{2}{*}{\textbf{Dataset}} & \multirow{2}{*}{\textbf{Curve}} & \multirow{2}{*}{\textbf{Cot}} &  \multicolumn{4}{c}{\textbf{Mean Absolute Difference (MAD)}} \\
 & &  & Early & Middle & Last & All\\
\midrule
\multirow{16}{*}{Geometry} & \multirow{3}{*}{$s_Q$} 
  & Simplified & 1.07 & 1.09 & 0.69 & 0.89 \\
& & Detailed & 0.45 & 0.55 & 0.20 & 0.37 \\
\cmidrule{2-7}
& \multirow{3}{*}{$s_K$} 
  & Simplified & 0.48 & 0.36 & 0.25 & 0.34 \\
& & Detailed & 0.20 & 0.15 & 0.07 & 0.13 \\
\cmidrule{2-7}
& \multirow{3}{*}{$s_V$} 
  & Simplified & 2.84 & 1.63 & 0.64 & 1.57 \\
& & Detailed & 0.97 & 0.70 & 0.34 & 0.63 \\
\cmidrule{2-7}
& \multirow{3}{*}{$s_O$} 
  & Simplified & 1.51 & 1.77 & 0.75 & 1.22 \\
& & Detailed & 0.64 & 0.88 & 0.43 & 0.61 \\
\cmidrule{2-7}
 & \multirow{3}{*}{$r_Q$} 
  & Simplified & 0.01 & 0.01 & 0.07 & 0.03 \\
& & Detailed & 0.01 & 0.01 & 0.06 & 0.03 \\
\cmidrule{2-7}
& \multirow{3}{*}{$r_K$} 
  & Simplified & 0.02 & 0.00 & 0.02 & 0.01 \\
& & Detailed & 0.01 & 0.00 & 0.02 & 0.01 \\
\cmidrule{2-7}
 & \multirow{3}{*}{$r_V$} 
  & Simplified & 0.02 & 0.01 & 0.02 & 0.02 \\
& & Detailed & 0.02 & 0.01 & 0.01 & 0.01 \\
\cmidrule{2-7}
& \multirow{3}{*}{$r_O$} 
  & Simplified & 0.01 & 0.01 & 0.01 & 0.01 \\
& & Detailed & 0.01 & 0.01 & 0.01 & 0.01 \\
\bottomrule
\end{tabular}
}
\caption{
Statistical results for MATH-Geometry using Llama-3.1-8B-Instruct on correct responses.
}
\label{tbl:math_lighteval_math_llama3_8b_it_grads_geometry}
\end{table}

\begin{figure}[!h]
\centering
\includegraphics[width=\linewidth]{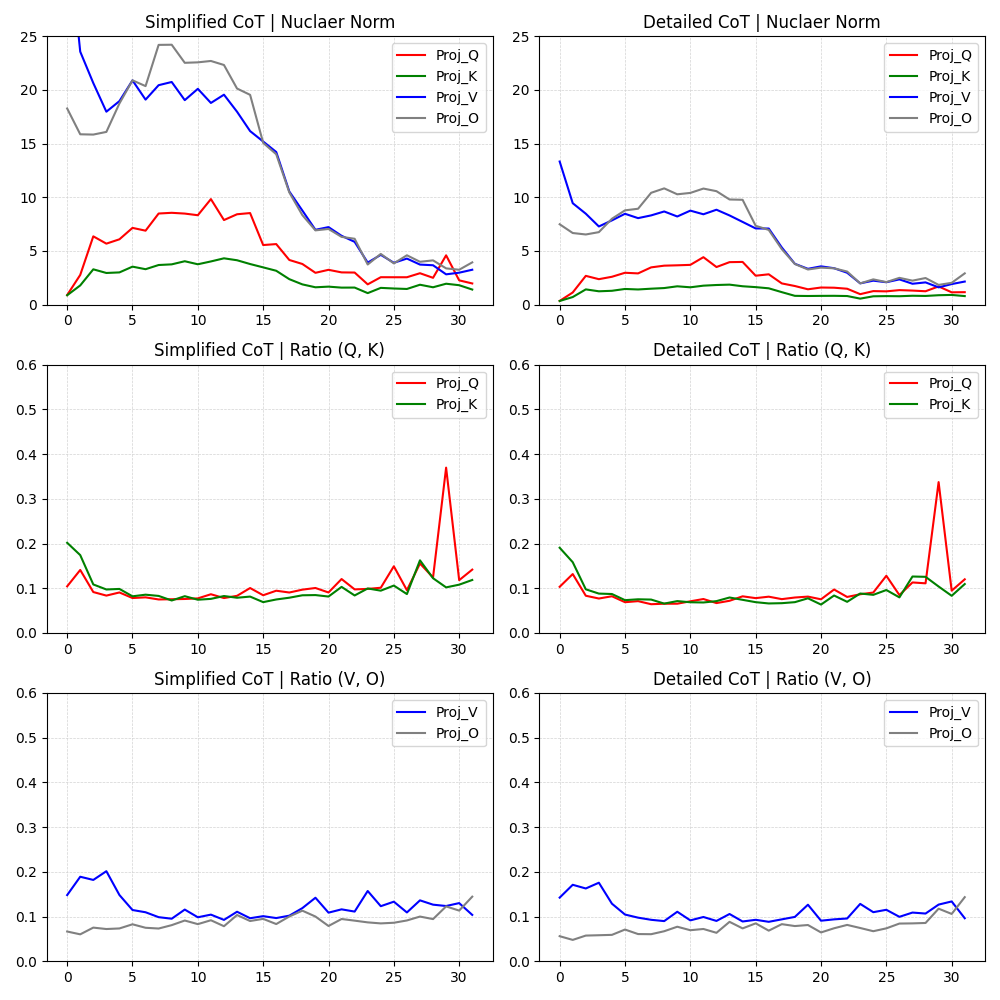}
\caption{Visualization for MATH-Geometry using Llama-3.1-8B-Instruct on correct responses.}
\label{fig:math_correct_geometry_it}
\vspace{-2mm}
\end{figure}

The visualizations and statistical results on other reasoning tasks: AQuA (Figure \ref{fig:reasoning_correct_aqua_it}, Table \ref{tbl:reasoning_cot_500_llama3_8b_it_grads_aqua_train}), GSM8K (Figure \ref{fig:reasoning_correct_gsm8k_it}, Table \ref{tbl:reasoning_cot_500_llama3_8b_it_grads_gsm8k_train}), StrategyQA (Figure \ref{fig:reasoning_correct_strategyqa_it}, Table \ref{tbl:reasoning_cot_500_llama3_8b_it_grads_strategyqa_train}), ECQA (Figure \ref{fig:reasoning_correct_ecqa_it}, Table \ref{tbl:reasoning_cot_500_llama3_8b_it_grads_ecqa_train}), CREAK (Figure \ref{fig:reasoning_correct_creak_it}, Table \ref{tbl:reasoning_cot_500_llama3_8b_it_grads_creak_train}), Sensemaking (Figure \ref{fig:reasoning_correct_sensemaking_it}, Table \ref{tbl:reasoning_cot_500_llama3_8b_it_grads_sensemaking_train}).

\begin{table}[!tbh]
\centering
\small
\scalebox{0.75}{
\begin{tabular}{l|l|l|cccc}
\toprule
\multirow{2}{*}{\textbf{Dataset}} & \multirow{2}{*}{\textbf{Curve}} & \multirow{2}{*}{\textbf{Cot}} &  \multicolumn{4}{c}{\textbf{Mean Absolute Difference (MAD)}} \\
 & &  & Early & Middle & Last & All\\
\midrule
\multirow{24}{*}{AQuA} & \multirow{3}{*}{$s_Q$} 
  & None & 19.05 & 15.54 & 10.80 & 14.54 \\
& & Simplified & 2.64 & 2.78 & 1.42 & 2.12 \\
& & Detailed & 0.43 & 0.52 & 0.29 & 0.38 \\
\cmidrule{2-7}
& \multirow{3}{*}{$s_K$} 
  & None & 10.39 & 11.00 & 3.13 & 7.80 \\
& & Simplified & 1.40 & 1.26 & 0.65 & 1.05 \\
& & Detailed & 0.20 & 0.16 & 0.09 & 0.14 \\
\cmidrule{2-7}
& \multirow{3}{*}{$s_V$} 
  & None & 63.48 & 30.54 & 8.67 & 30.72 \\
& & Simplified & 8.26 & 4.05 & 1.24 & 4.07 \\
& & Detailed & 1.04 & 0.65 & 0.36 & 0.65 \\
\cmidrule{2-7}
& \multirow{3}{*}{$s_O$} 
  & None & 25.90 & 35.39 & 7.55 & 21.70 \\
& & Simplified & 3.62 & 4.87 & 1.26 & 3.00 \\
& & Detailed & 0.63 & 0.94 & 0.42 & 0.62 \\
\cmidrule{2-7}
 & \multirow{3}{*}{$r_Q$} 
  & None & 0.02 & 0.07 & 0.10 & 0.06 \\
& & Simplified & 0.01 & 0.03 & 0.07 & 0.04 \\
& & Detailed & 0.01 & 0.01 & 0.09 & 0.04 \\
\cmidrule{2-7}
& \multirow{3}{*}{$r_K$} 
  & None & 0.03 & 0.04 & 0.06 & 0.04 \\
& & Simplified & 0.03 & 0.03 & 0.03 & 0.03 \\
& & Detailed & 0.02 & 0.01 & 0.03 & 0.02 \\
\cmidrule{2-7}
 & \multirow{3}{*}{$r_V$} 
  & None & 0.02 & 0.03 & 0.03 & 0.03 \\
& & Simplified & 0.02 & 0.02 & 0.02 & 0.02 \\
& & Detailed & 0.02 & 0.02 & 0.02 & 0.02 \\
\cmidrule{2-7}
& \multirow{3}{*}{$r_O$} 
  & None & 0.02 & 0.04 & 0.08 & 0.05 \\
& & Simplified & 0.01 & 0.02 & 0.02 & 0.02 \\
& & Detailed & 0.01 & 0.01 & 0.01 & 0.01 \\
\bottomrule
\end{tabular}
}
\caption{
Statistical results for AQuA using Llama-3.1-8B-Instruct on correct responses.
}
\label{tbl:reasoning_cot_500_llama3_8b_it_grads_aqua_train}
\end{table}

\begin{figure*}[!h]
\centering
\includegraphics[width=\linewidth]{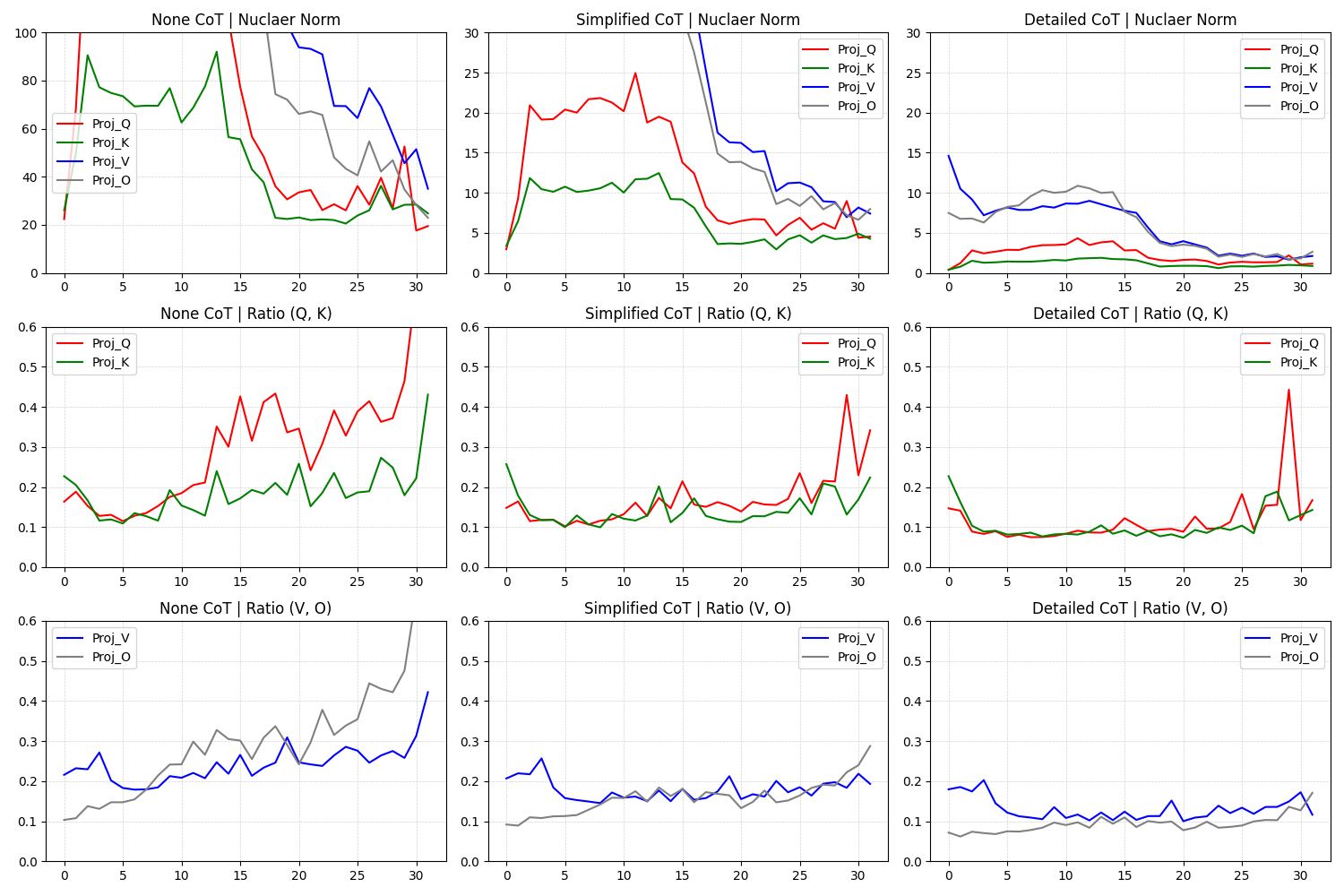}
\caption{Visualization for AQuA using Llama-3.1-8B-Instruct on correct responses.}
\label{fig:reasoning_correct_aqua_it}
\vspace{-2mm}
\end{figure*}

\begin{table}[!tbh]
\centering
\small
\scalebox{0.75}{
\begin{tabular}{l|l|l|cccc}
\toprule
\multirow{2}{*}{\textbf{Dataset}} & \multirow{2}{*}{\textbf{Curve}} & \multirow{2}{*}{\textbf{Cot}} &  \multicolumn{4}{c}{\textbf{Mean Absolute Difference (MAD)}} \\
 & &  & Early & Middle & Last & All\\
\midrule
\multirow{24}{*}{GSM8K} & \multirow{3}{*}{$s_Q$} 
  & None & 16.19 & 19.18 & 15.89 & 16.45 \\
& & Simplified & 1.30 & 1.53 & 1.04 & 1.20 \\
& & Detailed & 0.42 & 0.51 & 0.34 & 0.40 \\
\cmidrule{2-7}
& \multirow{3}{*}{$s_K$} 
  & None & 9.68 & 11.73 & 5.44 & 8.58 \\
& & Simplified & 0.68 & 0.65 & 0.30 & 0.51 \\
& & Detailed & 0.19 & 0.17 & 0.09 & 0.14 \\
\cmidrule{2-7}
& \multirow{3}{*}{$s_V$} 
  & None & 62.84 & 37.06 & 13.72 & 34.67 \\
& & Simplified & 3.26 & 2.12 & 0.73 & 1.89 \\
& & Detailed & 1.00 & 0.67 & 0.38 & 0.64 \\
\cmidrule{2-7}
& \multirow{3}{*}{$s_O$} 
  & None & 27.31 & 42.87 & 11.38 & 25.32 \\
& & Simplified & 1.75 & 2.78 & 0.78 & 1.61 \\
& & Detailed & 0.59 & 1.04 & 0.42 & 0.63 \\
\cmidrule{2-7}
 & \multirow{3}{*}{$r_Q$} 
  & None & 0.01 & 0.03 & 0.08 & 0.04 \\
& & Simplified & 0.01 & 0.03 & 0.10 & 0.05 \\
& & Detailed & 0.01 & 0.02 & 0.10 & 0.05 \\
\cmidrule{2-7}
& \multirow{3}{*}{$r_K$} 
  & None & 0.03 & 0.04 & 0.05 & 0.04 \\
& & Simplified & 0.03 & 0.04 & 0.03 & 0.03 \\
& & Detailed & 0.02 & 0.02 & 0.03 & 0.02 \\
\cmidrule{2-7}
 & \multirow{3}{*}{$r_V$} 
  & None & 0.02 & 0.03 & 0.03 & 0.03 \\
& & Simplified & 0.02 & 0.02 & 0.03 & 0.03 \\
& & Detailed & 0.02 & 0.02 & 0.02 & 0.02 \\
\cmidrule{2-7}
& \multirow{3}{*}{$r_O$} 
  & None & 0.02 & 0.04 & 0.08 & 0.05 \\
& & Simplified & 0.01 & 0.02 & 0.03 & 0.02 \\
& & Detailed & 0.01 & 0.02 & 0.02 & 0.01 \\
\bottomrule
\end{tabular}
}
\caption{
Statistical results for GSM8K using Llama-3.1-8B-Instruct on correct responses.
}
\label{tbl:reasoning_cot_500_llama3_8b_it_grads_gsm8k_train}
\end{table}

\begin{figure*}[!h]
\centering
\includegraphics[width=\linewidth]{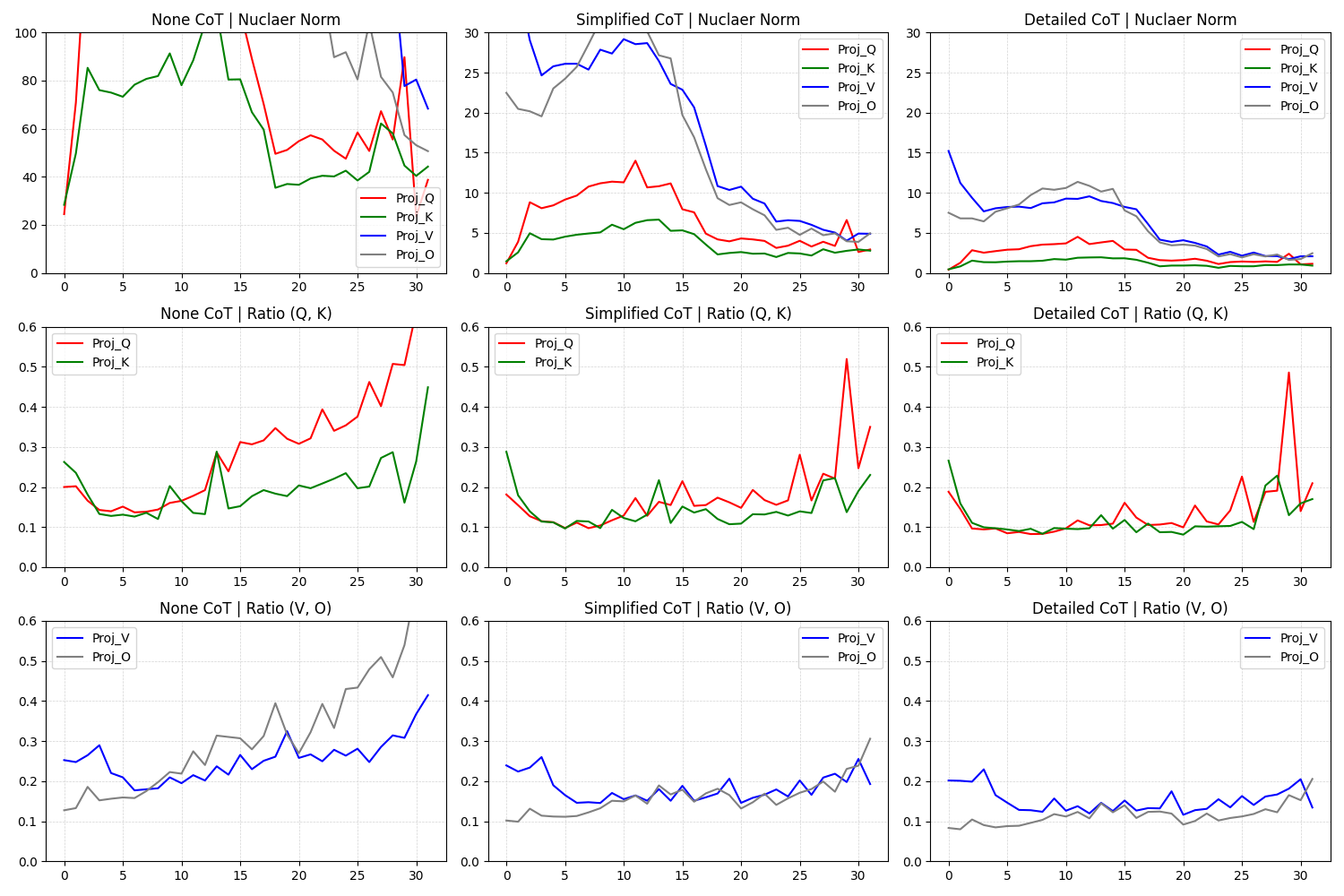}
\caption{Visualization for GSM8K using Llama-3.1-8B-Instruct on correct responses.}
\label{fig:reasoning_correct_gsm8k_it}
\vspace{-2mm}
\end{figure*}

\begin{table}[!tbh]
\centering
\small
\scalebox{0.75}{
\begin{tabular}{l|l|l|cccc}
\toprule
\multirow{2}{*}{\textbf{Dataset}} & \multirow{2}{*}{\textbf{Curve}} & \multirow{2}{*}{\textbf{Cot}} &  \multicolumn{4}{c}{\textbf{Mean Absolute Difference (MAD)}} \\
 & &  & Early & Middle & Last & All\\
\midrule
\multirow{24}{*}{StrategyQA} & \multirow{3}{*}{$s_Q$} 
  & None & 13.26 & 18.96 & 5.16 & 11.85 \\
& & Simplified & 1.85 & 3.19 & 1.54 & 2.05 \\
& & Detailed & 0.49 & 0.68 & 0.31 & 0.45 \\
\cmidrule{2-7}
& \multirow{3}{*}{$s_K$} 
  & None & 7.14 & 7.75 & 4.61 & 6.34 \\
& & Simplified & 0.91 & 1.33 & 0.39 & 0.82 \\
& & Detailed & 0.20 & 0.18 & 0.09 & 0.15 \\
\cmidrule{2-7}
& \multirow{3}{*}{$s_V$} 
  & None & 58.20 & 27.11 & 11.06 & 29.17 \\
& & Simplified & 6.78 & 3.32 & 0.95 & 3.36 \\
& & Detailed & 1.15 & 0.82 & 0.56 & 0.80 \\
\cmidrule{2-7}
& \multirow{3}{*}{$s_O$} 
  & None & 23.40 & 27.11 & 7.11 & 17.76 \\
& & Simplified & 3.26 & 3.89 & 1.32 & 2.60 \\
& & Detailed & 0.76 & 1.14 & 0.73 & 0.82 \\
\cmidrule{2-7}
 & \multirow{3}{*}{$r_Q$} 
  & None & 0.01 & 0.04 & 0.09 & 0.05 \\
& & Simplified & 0.01 & 0.04 & 0.08 & 0.05 \\
& & Detailed & 0.01 & 0.01 & 0.09 & 0.04 \\
\cmidrule{2-7}
& \multirow{3}{*}{$r_K$} 
  & None & 0.02 & 0.02 & 0.03 & 0.02 \\
& & Simplified & 0.02 & 0.04 & 0.02 & 0.03 \\
& & Detailed & 0.02 & 0.01 & 0.02 & 0.01 \\
\cmidrule{2-7}
 & \multirow{3}{*}{$r_V$} 
  & None & 0.03 & 0.03 & 0.03 & 0.03 \\
& & Simplified & 0.02 & 0.02 & 0.02 & 0.02 \\
& & Detailed & 0.02 & 0.01 & 0.02 & 0.02 \\
\cmidrule{2-7}
& \multirow{3}{*}{$r_O$} 
  & None & 0.02 & 0.04 & 0.08 & 0.05 \\
& & Simplified & 0.02 & 0.02 & 0.02 & 0.02 \\
& & Detailed & 0.01 & 0.01 & 0.01 & 0.01 \\
\bottomrule
\end{tabular}
}
\caption{
Statistical results for StrategyQA using Llama-3.1-8B-Instruct on correct responses.
}
\label{tbl:reasoning_cot_500_llama3_8b_it_grads_strategyqa_train}
\end{table}

\begin{figure*}[!h]
\centering
\includegraphics[width=\linewidth]{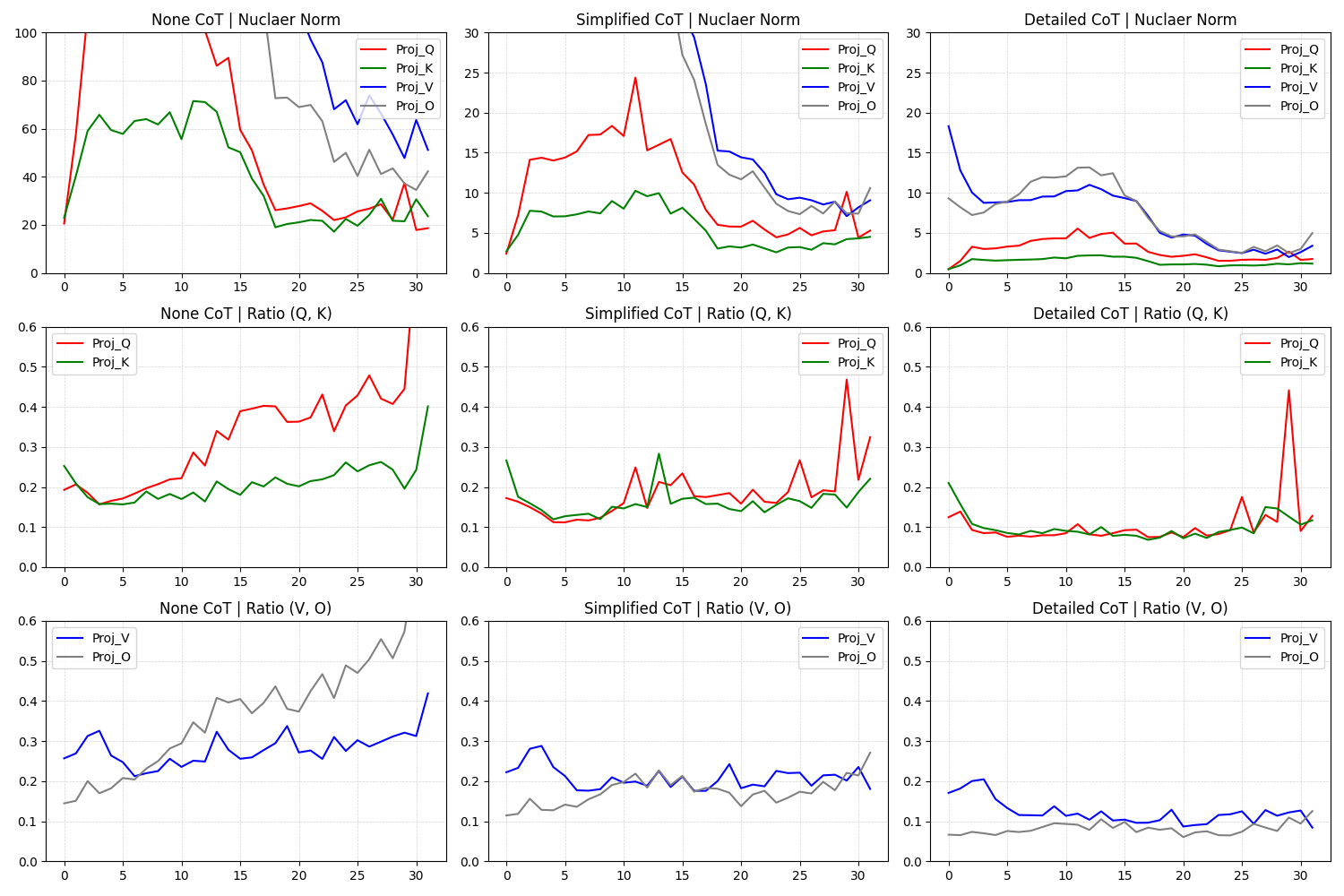}
\caption{Visualization for StrategyQA using Llama-3.1-8B-Instruct on correct responses.}
\label{fig:reasoning_correct_strategyqa_it}
\vspace{-2mm}
\end{figure*}

\begin{table}[!tbh]
\centering
\small
\scalebox{0.75}{
\begin{tabular}{l|l|l|cccc}
\toprule
\multirow{2}{*}{\textbf{Dataset}} & \multirow{2}{*}{\textbf{Curve}} & \multirow{2}{*}{\textbf{Cot}} &  \multicolumn{4}{c}{\textbf{Mean Absolute Difference (MAD)}} \\
 & &  & Early & Middle & Last & All\\
\midrule
\multirow{24}{*}{ECQA} & \multirow{3}{*}{$s_Q$} 
  & None & 16.08 & 16.53 & 5.36 & 11.91 \\
& & Simplified & 3.02 & 3.43 & 1.70 & 2.55 \\
& & Detailed & 0.67 & 0.81 & 0.46 & 0.60 \\
\cmidrule{2-7}
& \multirow{3}{*}{$s_K$} 
  & None & 7.86 & 8.50 & 6.50 & 7.53 \\
& & Simplified & 1.47 & 1.54 & 0.77 & 1.21 \\
& & Detailed & 0.28 & 0.22 & 0.11 & 0.20 \\
\cmidrule{2-7}
& \multirow{3}{*}{$s_V$} 
  & None & 48.82 & 32.86 & 11.34 & 28.05 \\
& & Simplified & 7.47 & 4.70 & 1.48 & 4.19 \\
& & Detailed & 1.40 & 1.00 & 0.62 & 0.96 \\
\cmidrule{2-7}
& \multirow{3}{*}{$s_O$} 
  & None & 20.81 & 34.23 & 6.08 & 19.08 \\
& & Simplified & 3.46 & 5.65 & 1.38 & 3.24 \\
& & Detailed & 0.85 & 1.40 & 0.74 & 0.93 \\
\cmidrule{2-7}
 & \multirow{3}{*}{$r_Q$} 
  & None & 0.01 & 0.05 & 0.11 & 0.06 \\
& & Simplified & 0.01 & 0.04 & 0.09 & 0.05 \\
& & Detailed & 0.01 & 0.01 & 0.12 & 0.05 \\
\cmidrule{2-7}
& \multirow{3}{*}{$r_K$} 
  & None & 0.02 & 0.02 & 0.05 & 0.03 \\
& & Simplified & 0.03 & 0.03 & 0.03 & 0.03 \\
& & Detailed & 0.02 & 0.01 & 0.02 & 0.02 \\
\cmidrule{2-7}
 & \multirow{3}{*}{$r_V$} 
  & None & 0.02 & 0.02 & 0.02 & 0.02 \\
& & Simplified & 0.02 & 0.02 & 0.02 & 0.02 \\
& & Detailed & 0.01 & 0.01 & 0.02 & 0.02 \\
\cmidrule{2-7}
& \multirow{3}{*}{$r_O$} 
  & None & 0.02 & 0.04 & 0.07 & 0.04 \\
& & Simplified & 0.01 & 0.02 & 0.03 & 0.02 \\
& & Detailed & 0.01 & 0.01 & 0.02 & 0.01 \\
\bottomrule
\end{tabular}
}
\caption{
Statistical results for ECQA using Llama-3.1-8B-Instruct on correct responses.
}
\label{tbl:reasoning_cot_500_llama3_8b_it_grads_ecqa_train}
\end{table}

\begin{figure*}[!h]
\centering
\includegraphics[width=\linewidth]{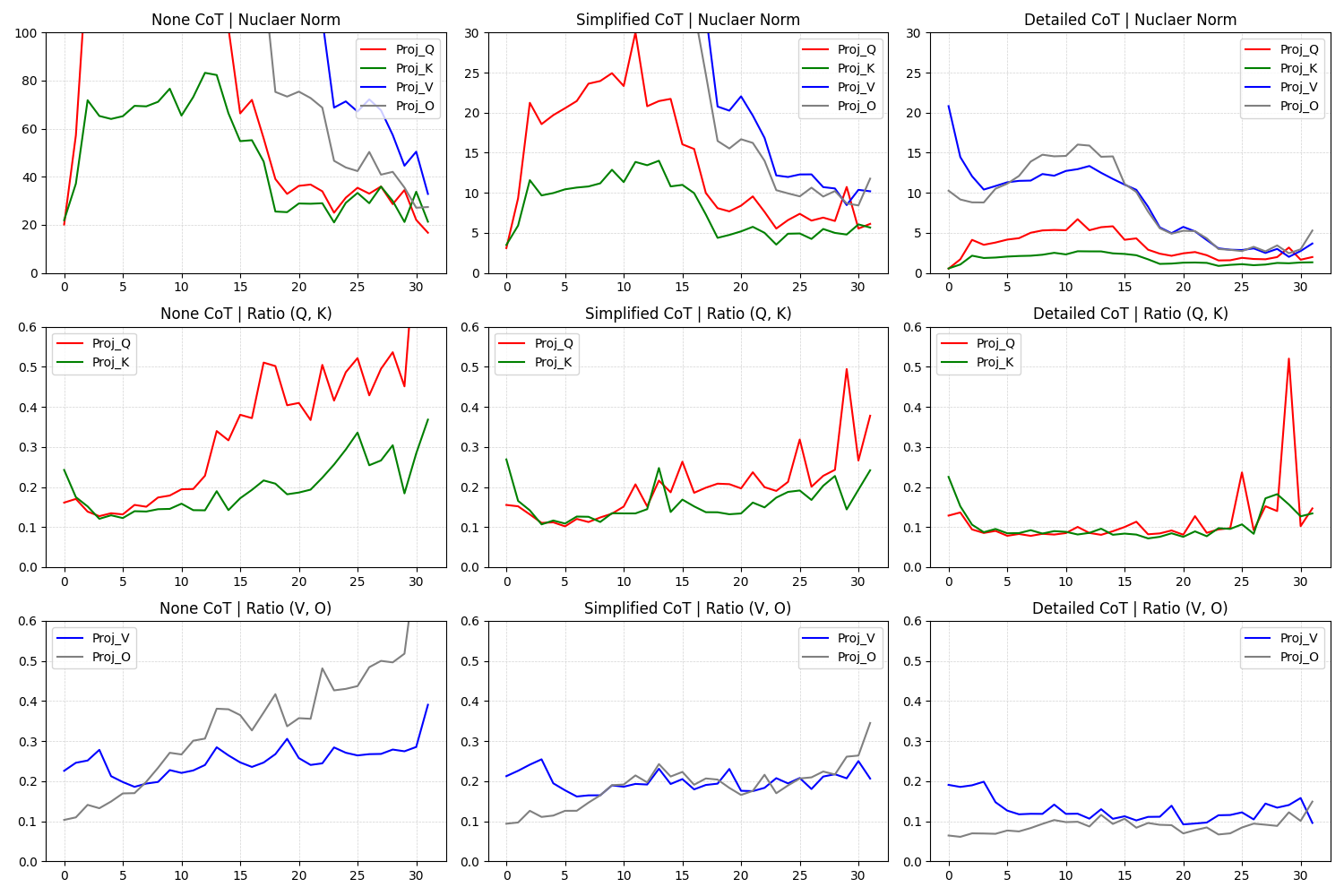}
\caption{Visualization for ECQA using Llama-3.1-8B-Instruct on correct responses.}
\label{fig:reasoning_correct_ecqa_it}
\vspace{-2mm}
\end{figure*}

\begin{table}[!tbh]
\centering
\small
\scalebox{0.75}{
\begin{tabular}{l|l|l|cccc}
\toprule
\multirow{2}{*}{\textbf{Dataset}} & \multirow{2}{*}{\textbf{Curve}} & \multirow{2}{*}{\textbf{Cot}} &  \multicolumn{4}{c}{\textbf{Mean Absolute Difference (MAD)}} \\
 & &  & Early & Middle & Last & All\\
\midrule
\multirow{24}{*}{CREAK} & \multirow{3}{*}{$s_Q$} 
  & None & 21.99 & 25.06 & 9.00 & 17.79 \\
& & Simplified & 3.10 & 4.30 & 2.72 & 3.21 \\
& & Detailed & 0.60 & 0.82 & 0.42 & 0.57 \\
\cmidrule{2-7}
& \multirow{3}{*}{$s_K$} 
  & None & 12.46 & 15.76 & 9.76 & 12.24 \\
& & Simplified & 1.48 & 1.73 & 1.17 & 1.41 \\
& & Detailed & 0.24 & 0.23 & 0.11 & 0.18 \\
\cmidrule{2-7}
& \multirow{3}{*}{$s_V$} 
  & None & 95.66 & 49.82 & 28.15 & 53.01 \\
& & Simplified & 9.71 & 4.98 & 1.85 & 5.06 \\
& & Detailed & 1.45 & 0.97 & 0.62 & 0.96 \\
\cmidrule{2-7}
& \multirow{3}{*}{$s_O$} 
  & None & 36.06 & 50.58 & 14.81 & 31.75 \\
& & Simplified & 4.36 & 6.26 & 1.66 & 3.79 \\
& & Detailed & 0.85 & 1.35 & 0.75 & 0.91 \\
\cmidrule{2-7}
 & \multirow{3}{*}{$r_Q$} 
  & None & 0.01 & 0.04 & 0.10 & 0.05 \\
& & Simplified & 0.01 & 0.04 & 0.10 & 0.05 \\
& & Detailed & 0.01 & 0.01 & 0.10 & 0.04 \\
\cmidrule{2-7}
& \multirow{3}{*}{$r_K$} 
  & None & 0.03 & 0.03 & 0.06 & 0.04 \\
& & Simplified & 0.02 & 0.03 & 0.03 & 0.02 \\
& & Detailed & 0.02 & 0.01 & 0.02 & 0.02 \\
\cmidrule{2-7}
 & \multirow{3}{*}{$r_V$} 
  & None & 0.02 & 0.02 & 0.03 & 0.02 \\
& & Simplified & 0.02 & 0.02 & 0.02 & 0.02 \\
& & Detailed & 0.02 & 0.01 & 0.02 & 0.02 \\
\cmidrule{2-7}
& \multirow{3}{*}{$r_O$} 
  & None & 0.02 & 0.04 & 0.08 & 0.05 \\
& & Simplified & 0.02 & 0.02 & 0.03 & 0.02 \\
& & Detailed & 0.01 & 0.01 & 0.01 & 0.01 \\
\bottomrule
\end{tabular}
}
\caption{
Statistical results for CREAK using Llama-3.1-8B-Instruct on correct responses.
}
\label{tbl:reasoning_cot_500_llama3_8b_it_grads_creak_train}
\end{table}

\begin{figure*}[!h]
\centering
\includegraphics[width=\linewidth]{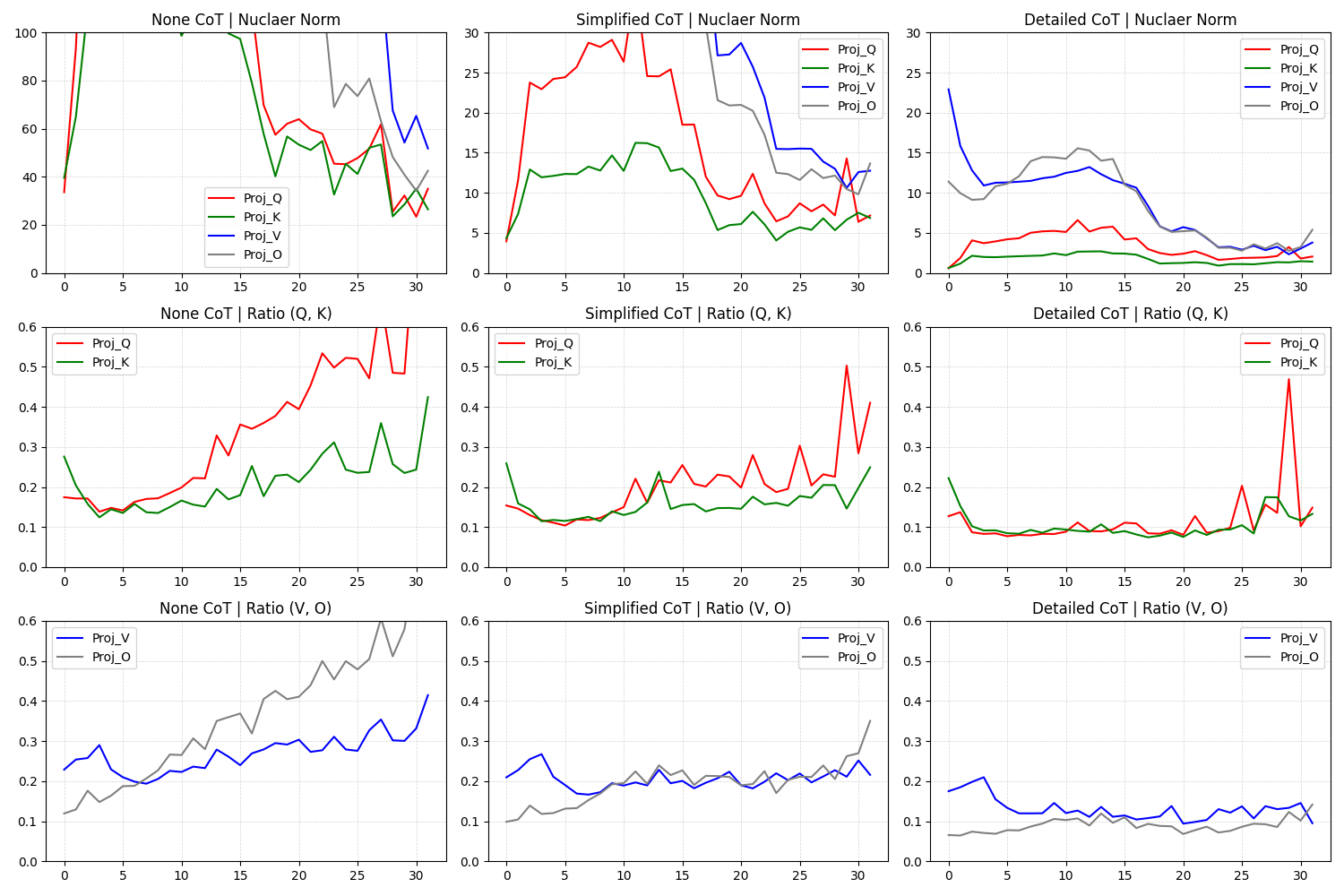}
\caption{Visualization for CREAK using Llama-3.1-8B-Instruct on correct responses.}
\label{fig:reasoning_correct_creak_it}
\vspace{-2mm}
\end{figure*}

\begin{table}[!tbh]
\centering
\small
\scalebox{0.75}{
\begin{tabular}{l|l|l|cccc}
\toprule
\multirow{2}{*}{\textbf{Dataset}} & \multirow{2}{*}{\textbf{Curve}} & \multirow{2}{*}{\textbf{Cot}} &  \multicolumn{4}{c}{\textbf{Mean Absolute Difference (MAD)}} \\
 & &  & Early & Middle & Last & All\\
\midrule
\multirow{24}{*}{Sensemaking} & \multirow{3}{*}{$s_Q$} 
  & None & 3.00 & 3.45 & 1.11 & 2.35 \\
& & Simplified & 3.50 & 3.96 & 2.41 & 3.14 \\
& & Detailed & 0.62 & 0.76 & 0.34 & 0.53 \\
\cmidrule{2-7}
& \multirow{3}{*}{$s_K$} 
  & None & 1.54 & 1.18 & 1.46 & 1.43 \\
& & Simplified & 1.78 & 1.69 & 1.10 & 1.48 \\
& & Detailed & 0.26 & 0.22 & 0.10 & 0.18 \\
\cmidrule{2-7}
& \multirow{3}{*}{$s_V$} 
  & None & 8.05 & 4.41 & 3.69 & 5.14 \\
& & Simplified & 8.79 & 5.79 & 2.31 & 5.28 \\
& & Detailed & 1.29 & 0.88 & 0.57 & 0.87 \\
\cmidrule{2-7}
& \multirow{3}{*}{$s_O$} 
  & None & 4.33 & 6.16 & 2.23 & 4.04 \\
& & Simplified & 4.32 & 7.21 & 1.83 & 4.20 \\
& & Detailed & 0.82 & 1.33 & 0.70 & 0.88 \\
\cmidrule{2-7}
 & \multirow{3}{*}{$r_Q$} 
  & None & 0.01 & 0.06 & 0.12 & 0.07 \\
& & Simplified & 0.01 & 0.03 & 0.09 & 0.04 \\
& & Detailed & 0.01 & 0.01 & 0.10 & 0.04 \\
\cmidrule{2-7}
& \multirow{3}{*}{$r_K$} 
  & None & 0.02 & 0.03 & 0.06 & 0.04 \\
& & Simplified & 0.02 & 0.04 & 0.04 & 0.03 \\
& & Detailed & 0.02 & 0.01 & 0.02 & 0.02 \\
\cmidrule{2-7}
 & \multirow{3}{*}{$r_V$} 
  & None & 0.02 & 0.03 & 0.04 & 0.03 \\
& & Simplified & 0.02 & 0.02 & 0.02 & 0.02 \\
& & Detailed & 0.02 & 0.01 & 0.02 & 0.02 \\
\cmidrule{2-7}
& \multirow{3}{*}{$r_O$} 
  & None & 0.01 & 0.06 & 0.09 & 0.05 \\
& & Simplified & 0.01 & 0.02 & 0.03 & 0.02 \\
& & Detailed & 0.01 & 0.01 & 0.02 & 0.01 \\
\bottomrule
\end{tabular}
}
\caption{
Statistical results for Sensemaking using Llama-3.1-8B-Instruct on correct responses.
}
\label{tbl:reasoning_cot_500_llama3_8b_it_grads_sensemaking_train}
\end{table}

\begin{figure*}[!h]
\centering
\includegraphics[width=\linewidth]{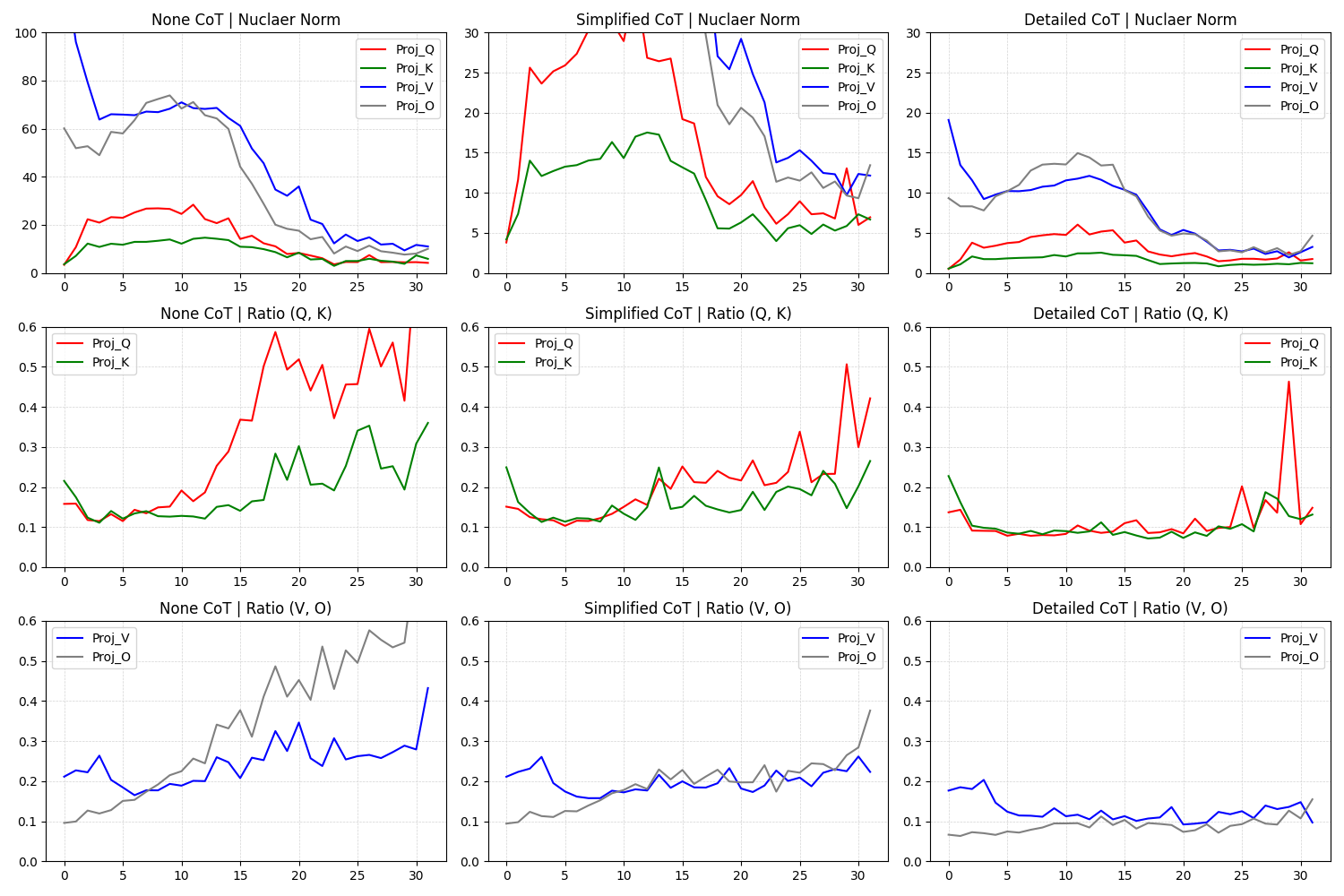}
\caption{Visualization for Sensemaking using Llama-3.1-8B-Instruct on correct responses.}
\label{fig:reasoning_correct_sensemaking_it}
\vspace{-2mm}
\end{figure*}

\subsubsection{Wiki Tasks}

The visualizations and statistical results on Wiki tasks are shown in Figure \ref{fig:wiki_correct_it} and Table \ref{tbl:wiki_wiki_popularity_llama3_8b_it_grads}.

\begin{table}[!tbh]
\centering
\small
\scalebox{0.75}{
\begin{tabular}{l|l|l|cccc}
\toprule
\multirow{2}{*}{\textbf{Dataset}} & \multirow{2}{*}{\textbf{Curve}} & \multirow{2}{*}{\textbf{Cot}} &  \multicolumn{4}{c}{\textbf{Mean Absolute Difference (MAD)}} \\
 & &  & Early & Middle & Last & All\\
\midrule
\multirow{32}{*}{Wiki} & \multirow{3}{*}{$s_Q$} 
  & Len 100 & 1.03 & 1.69 & 1.25 & 1.28 \\
& & Len 500 & 0.83 & 1.10 & 0.59 & 0.79 \\
& & Len 1000 & 0.72 & 0.90 & 0.38 & 0.62 \\
& & Unpopular & 1.36 & 1.65 & 0.99 & 1.29 \\
\cmidrule{2-7}
& \multirow{3}{*}{$s_K$} 
  & Len 100 & 0.47 & 0.51 & 0.29 & 0.43 \\
& & Len 500 & 0.37 & 0.32 & 0.15 & 0.28 \\
& & Len 1000 & 0.33 & 0.26 & 0.10 & 0.23 \\
& & Unpopular & 0.60 & 0.65 & 0.34 & 0.55 \\
\cmidrule{2-7}
& \multirow{3}{*}{$s_V$} 
  & Len 100 & 4.98 & 1.52 & 0.81 & 2.23 \\
& & Len 500 & 3.01 & 1.17 & 0.64 & 1.46 \\
& & Len 1000 & 2.38 & 0.98 & 0.55 & 1.19 \\
& & Unpopular & 12.54 & 2.64 & 0.75 & 4.80 \\
\cmidrule{2-7}
& \multirow{3}{*}{$s_O$} 
  & Len 100 & 2.44 & 2.12 & 1.12 & 1.82 \\
& & Len 500 & 1.80 & 1.49 & 0.87 & 1.31 \\
& & Len 1000 & 1.52 & 1.26 & 0.77 & 1.11 \\
& & Unpopular & 4.38 & 2.57 & 0.99 & 2.50 \\
\cmidrule{2-7}
 & \multirow{3}{*}{$r_Q$} 
  & Len 100 & 0.02 & 0.02 & 0.08 & 0.04 \\
& & Len 500 & 0.02 & 0.02 & 0.07 & 0.04 \\
& & Len 1000 & 0.02 & 0.02 & 0.06 & 0.03 \\
& & Unpopular & 0.03 & 0.02 & 0.05 & 0.03 \\
\cmidrule{2-7}
& \multirow{3}{*}{$r_K$} 
  & Len 100 & 0.02 & 0.02 & 0.01 & 0.02 \\
& & Len 500 & 0.02 & 0.02 & 0.01 & 0.02 \\
& & Len 1000 & 0.02 & 0.01 & 0.01 & 0.02 \\
& & Unpopular & 0.03 & 0.03 & 0.02 & 0.03 \\
\cmidrule{2-7}
 & \multirow{3}{*}{$r_V$} 
  & Len 100 & 0.03 & 0.02 & 0.02 & 0.02 \\
& & Len 500 & 0.03 & 0.01 & 0.01 & 0.02 \\
& & Len 1000 & 0.02 & 0.01 & 0.01 & 0.02 \\
& & Unpopular & 0.05 & 0.02 & 0.03 & 0.03 \\
\cmidrule{2-7}
& \multirow{3}{*}{$r_O$} 
  & Len 100 & 0.01 & 0.03 & 0.02 & 0.02 \\
& & Len 500 & 0.01 & 0.02 & 0.02 & 0.01 \\
& & Len 1000 & 0.01 & 0.01 & 0.01 & 0.01 \\
& & Unpopular & 0.01 & 0.03 & 0.02 & 0.02 \\
\bottomrule
\end{tabular}
}
\caption{
Statistical results for Wiki using Llama-3.1-8B-Instruct on correct responses.
}
\label{tbl:wiki_wiki_popularity_llama3_8b_it_grads}
\end{table}

\begin{figure*}[!h]
\centering
\includegraphics[width=\linewidth]{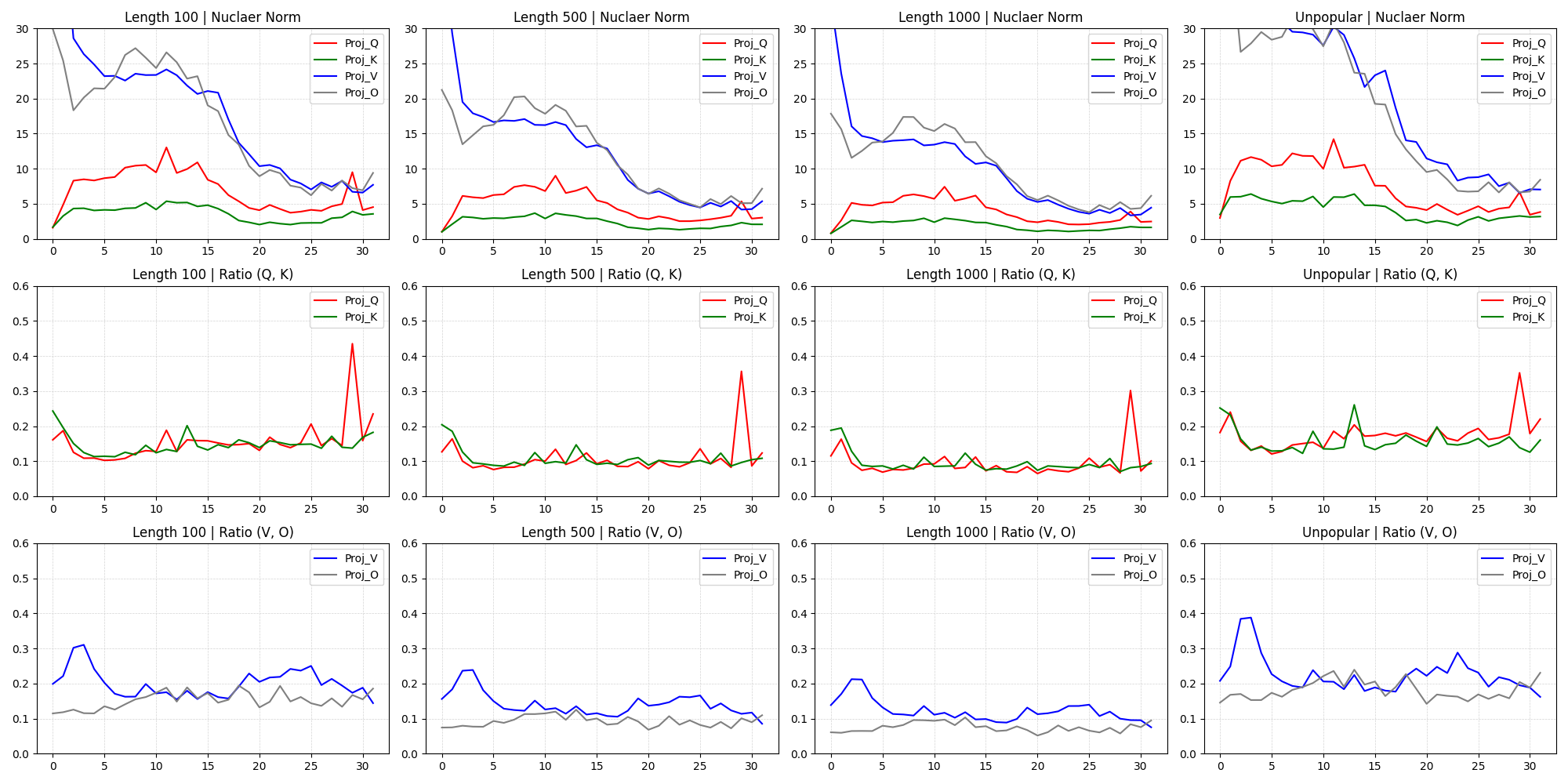}
\caption{Visualization for Wiki tasks using Llama-3.1-8B-Instruct on correct responses.}
\label{fig:wiki_correct_it}
\vspace{-2mm}
\end{figure*}

\subsection{Instructed LLM on Irrelevant Responses}

\subsubsection{Reasoning Tasks}

The visualizations and statistical results on MATH tasks: MATH-Algebra (Figure \ref{fig:math_nonsense_algebra_it}, Table \ref{tbl:math_lighteval_math_wrong_answer_shuffle_llama3_8b_it_grads_algebra}), MATH-Counting (Figure \ref{fig:math_nonsense_counting_it}, Table \ref{tbl:math_lighteval_math_wrong_answer_shuffle_llama3_8b_it_grads_counting_probability}), MATH-Geometry (Figure \ref{fig:math_nonsense_geometry_it}, Table \ref{tbl:math_lighteval_math_wrong_answer_shuffle_llama3_8b_it_grads_geometry}).

\begin{table}[!tbh]
\centering
\small
\scalebox{0.75}{
\begin{tabular}{l|l|l|cccc}
\toprule
\multirow{2}{*}{\textbf{Dataset}} & \multirow{2}{*}{\textbf{Curve}} & \multirow{2}{*}{\textbf{Cot}} &  \multicolumn{4}{c}{\textbf{Mean Absolute Difference (MAD)}} \\
 & &  & Early & Middle & Last & All\\
\midrule
\multirow{16}{*}{Algebra} & \multirow{3}{*}{$s_Q$} 
  & Simplified & 1.46 & 1.62 & 0.75 & 1.20 \\
& & Detailed & 0.83 & 0.85 & 0.30 & 0.61 \\
\cmidrule{2-7}
& \multirow{3}{*}{$s_K$} 
  & Simplified & 0.61 & 0.58 & 0.21 & 0.46 \\
& & Detailed & 0.34 & 0.29 & 0.12 & 0.24 \\
\cmidrule{2-7}
& \multirow{3}{*}{$s_V$} 
  & Simplified & 3.80 & 2.13 & 0.80 & 2.03 \\
& & Detailed & 1.67 & 1.10 & 0.45 & 0.98 \\
\cmidrule{2-7}
& \multirow{3}{*}{$s_O$} 
  & Simplified & 2.11 & 2.45 & 0.90 & 1.72 \\
& & Detailed & 0.98 & 1.31 & 0.53 & 0.89 \\
\cmidrule{2-7}
 & \multirow{3}{*}{$r_Q$} 
  & Simplified & 0.01 & 0.02 & 0.09 & 0.04 \\
& & Detailed & 0.01 & 0.01 & 0.07 & 0.03 \\
\cmidrule{2-7}
& \multirow{3}{*}{$r_K$} 
  & Simplified & 0.02 & 0.01 & 0.03 & 0.02 \\
& & Detailed & 0.02 & 0.01 & 0.02 & 0.02 \\
\cmidrule{2-7}
 & \multirow{3}{*}{$r_V$} 
  & Simplified & 0.02 & 0.01 & 0.02 & 0.02 \\
& & Detailed & 0.02 & 0.01 & 0.01 & 0.02 \\
\cmidrule{2-7}
& \multirow{3}{*}{$r_O$} 
  & Simplified & 0.01 & 0.01 & 0.02 & 0.01 \\
& & Detailed & 0.01 & 0.01 & 0.01 & 0.01 \\
\bottomrule
\end{tabular}
}
\caption{
Statistical results for MATH-Algebra using Llama-3.1-8B-Instruct on irrelevant responses.
}
\label{tbl:math_lighteval_math_wrong_answer_shuffle_llama3_8b_it_grads_algebra}
\end{table}

\begin{figure}[!h]
\centering
\includegraphics[width=\linewidth]{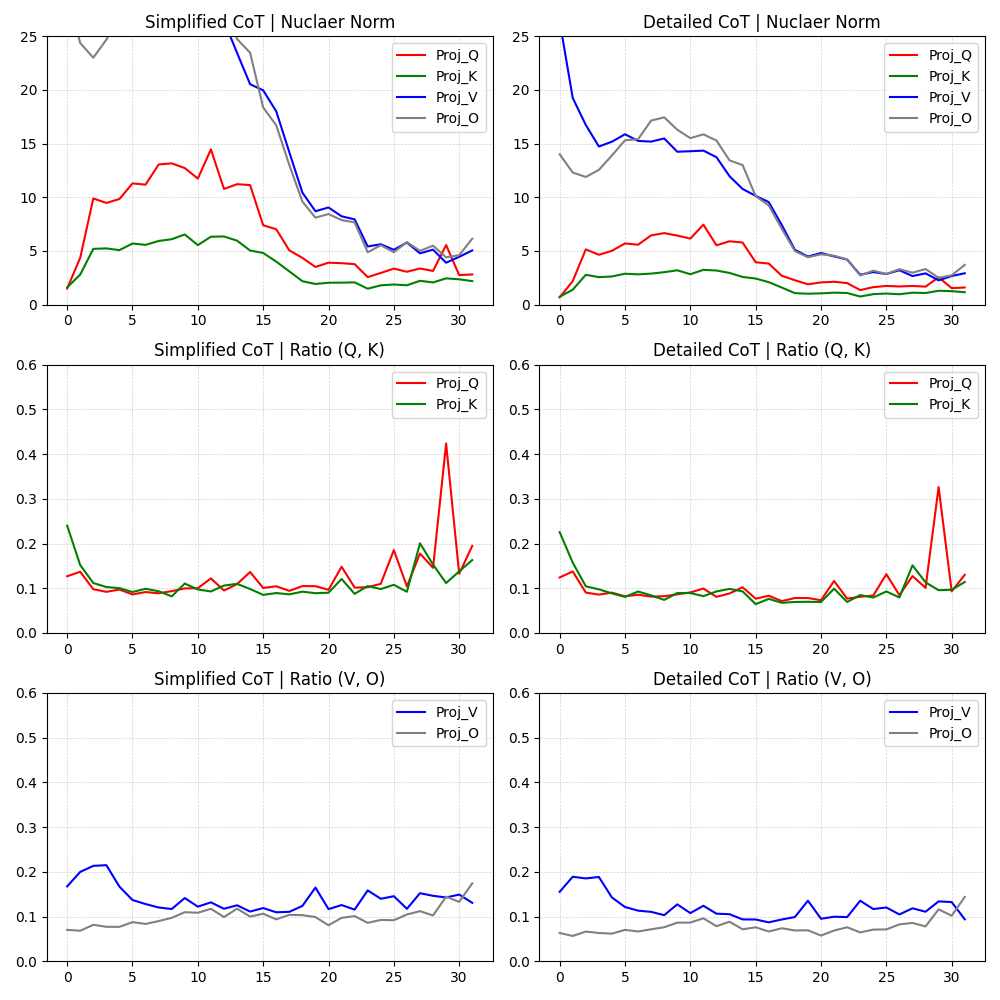}
\caption{Visualization for MATH-Algebra using Llama-3.1-8B-Instruct on irrelevant responses.}
\label{fig:math_nonsense_algebra_it}
\vspace{-2mm}
\end{figure}

\begin{table}[!tbh]
\centering
\small
\scalebox{0.75}{
\begin{tabular}{l|l|l|cccc}
\toprule
\multirow{2}{*}{\textbf{Dataset}} & \multirow{2}{*}{\textbf{Curve}} & \multirow{2}{*}{\textbf{Cot}} &  \multicolumn{4}{c}{\textbf{Mean Absolute Difference (MAD)}} \\
 & &  & Early & Middle & Last & All\\
\midrule
\multirow{16}{*}{Counting} & \multirow{3}{*}{$s_Q$} 
  & Simplified & 1.62 & 1.72 & 0.84 & 1.32 \\
& & Detailed & 0.88 & 0.85 & 0.30 & 0.62 \\
\cmidrule{2-7}
& \multirow{3}{*}{$s_K$} 
  & Simplified & 0.61 & 0.63 & 0.22 & 0.47 \\
& & Detailed & 0.34 & 0.28 & 0.12 & 0.24 \\
\cmidrule{2-7}
& \multirow{3}{*}{$s_V$} 
  & Simplified & 3.70 & 2.20 & 0.81 & 2.03 \\
& & Detailed & 1.61 & 1.04 & 0.45 & 0.94 \\
\cmidrule{2-7}
& \multirow{3}{*}{$s_O$} 
  & Simplified & 2.10 & 2.53 & 0.87 & 1.74 \\
& & Detailed & 0.91 & 1.26 & 0.59 & 0.88 \\
\cmidrule{2-7}
 & \multirow{3}{*}{$r_Q$} 
  & Simplified & 0.01 & 0.02 & 0.10 & 0.05 \\
& & Detailed & 0.01 & 0.01 & 0.08 & 0.04 \\
\cmidrule{2-7}
& \multirow{3}{*}{$r_K$} 
  & Simplified & 0.02 & 0.01 & 0.03 & 0.02 \\
& & Detailed & 0.02 & 0.01 & 0.02 & 0.01 \\
\cmidrule{2-7}
 & \multirow{3}{*}{$r_V$} 
  & Simplified & 0.02 & 0.01 & 0.02 & 0.02 \\
& & Detailed & 0.02 & 0.01 & 0.01 & 0.01 \\
\cmidrule{2-7}
& \multirow{3}{*}{$r_O$} 
  & Simplified & 0.01 & 0.01 & 0.01 & 0.01 \\
& & Detailed & 0.01 & 0.01 & 0.01 & 0.01 \\
\bottomrule
\end{tabular}
}
\caption{
Statistical results for MATH-Counting using Llama-3.1-8B-Instruct on irrelevant responses.
}
\label{tbl:math_lighteval_math_wrong_answer_shuffle_llama3_8b_it_grads_counting_probability}
\end{table}

\begin{figure}[!h]
\centering
\includegraphics[width=\linewidth]{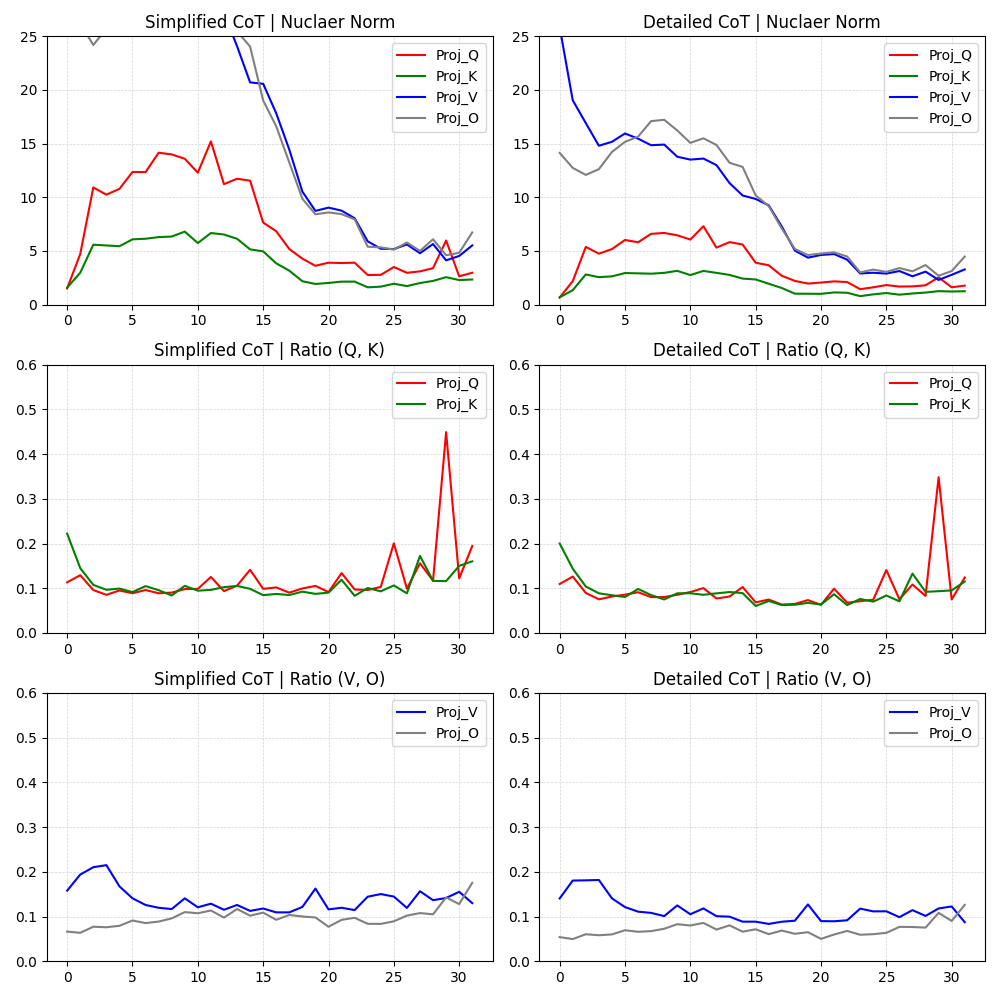}
\caption{Visualization for MATH-Counting using Llama-3.1-8B-Instruct on irrelevant responses.}
\label{fig:math_nonsense_counting_it}
\vspace{-2mm}
\end{figure}

\begin{table}[!tbh]
\centering
\small
\scalebox{0.75}{
\begin{tabular}{l|l|l|cccc}
\toprule
\multirow{2}{*}{\textbf{Dataset}} & \multirow{2}{*}{\textbf{Curve}} & \multirow{2}{*}{\textbf{Cot}} &  \multicolumn{4}{c}{\textbf{Mean Absolute Difference (MAD)}} \\
 & &  & Early & Middle & Last & All\\
\midrule
\multirow{16}{*}{Geometry} & \multirow{3}{*}{$s_Q$} 
  & Simplified & 1.48 & 1.52 & 1.04 & 1.26 \\
& & Detailed & 0.75 & 0.68 & 0.32 & 0.54 \\
\cmidrule{2-7}
& \multirow{3}{*}{$s_K$} 
  & Simplified & 0.60 & 0.47 & 0.33 & 0.45 \\
& & Detailed & 0.27 & 0.19 & 0.19 & 0.22 \\
\cmidrule{2-7}
& \multirow{3}{*}{$s_V$} 
  & Simplified & 3.59 & 1.88 & 0.79 & 1.89 \\
& & Detailed & 1.54 & 0.79 & 0.48 & 0.86 \\
\cmidrule{2-7}
& \multirow{3}{*}{$s_O$} 
  & Simplified & 2.20 & 2.19 & 0.94 & 1.64 \\
& & Detailed & 1.05 & 1.09 & 0.64 & 0.88 \\
\cmidrule{2-7}
 & \multirow{3}{*}{$r_Q$} 
  & Simplified & 0.01 & 0.02 & 0.09 & 0.04 \\
& & Detailed & 0.01 & 0.01 & 0.05 & 0.02 \\
\cmidrule{2-7}
& \multirow{3}{*}{$r_K$} 
  & Simplified & 0.02 & 0.01 & 0.03 & 0.02 \\
& & Detailed & 0.02 & 0.01 & 0.03 & 0.02 \\
\cmidrule{2-7}
 & \multirow{3}{*}{$r_V$} 
  & Simplified & 0.02 & 0.01 & 0.02 & 0.02 \\
& & Detailed & 0.02 & 0.01 & 0.02 & 0.02 \\
\cmidrule{2-7}
& \multirow{3}{*}{$r_O$} 
  & Simplified & 0.01 & 0.01 & 0.01 & 0.01 \\
& & Detailed & 0.01 & 0.01 & 0.01 & 0.01 \\
\bottomrule
\end{tabular}
}
\caption{
Statistical results for MATH-Geometry using Llama-3.1-8B-Instruct on irrelevant responses.
}
\label{tbl:math_lighteval_math_wrong_answer_shuffle_llama3_8b_it_grads_geometry}
\end{table}

\begin{figure}[!h]
\centering
\includegraphics[width=\linewidth]{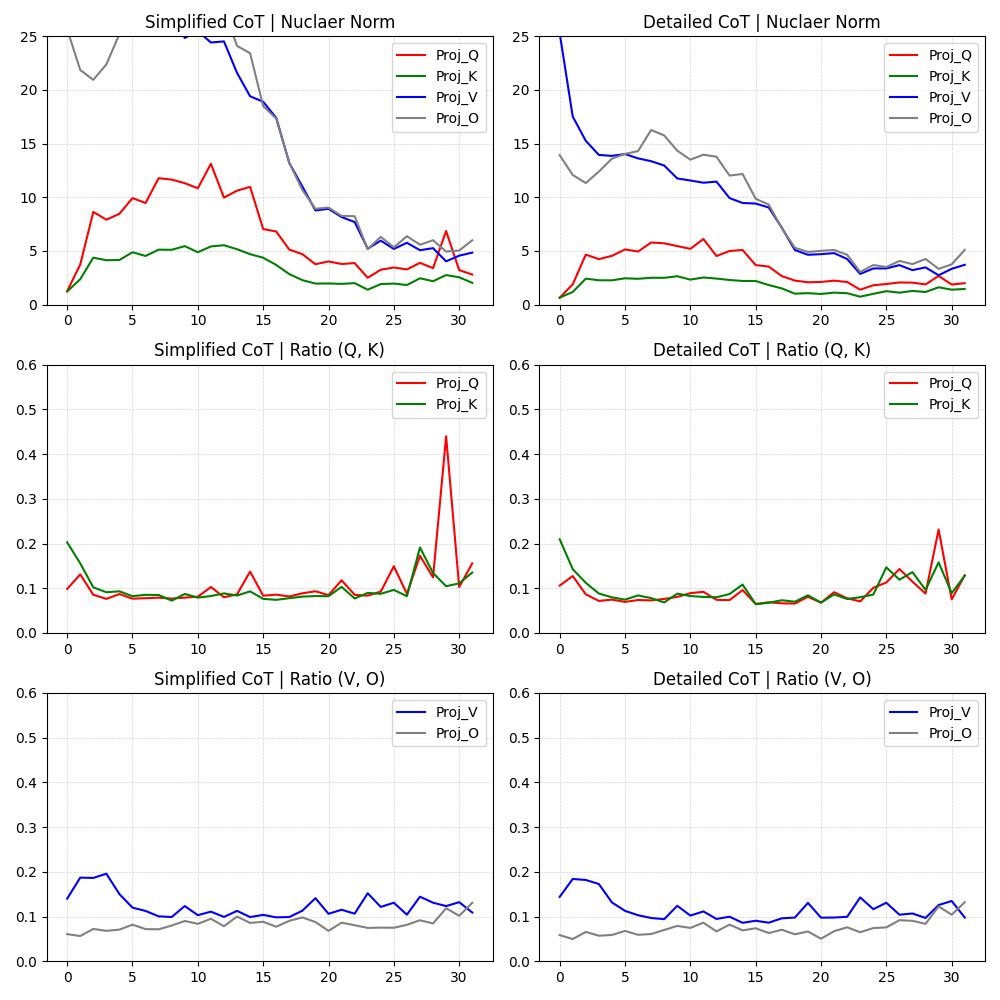}
\caption{Visualization for MATH-Geometry using Llama-3.1-8B-Instruct on irrelevant responses.}
\label{fig:math_nonsense_geometry_it}
\vspace{-2mm}
\end{figure}

The visualizations and statistical results on other reasoning tasks: AQuA (Figure \ref{fig:reasoning_nonsense_aqua_it}, Table \ref{tbl:reasoning_cot_500_wrong_answer_shuffle_llama3_8b_it_grads_aqua_train}), GSM8K (Figure \ref{fig:reasoning_nonsense_gsm8k_it}, Table \ref{tbl:reasoning_cot_500_wrong_answer_shuffle_llama3_8b_it_grads_gsm8k_train}), StrategyQA (Figure \ref{fig:reasoning_nonsense_strategyqa_it}, Table \ref{tbl:reasoning_cot_500_wrong_answer_shuffle_llama3_8b_it_grads_strategyqa_train}), ECQA (Figure \ref{fig:reasoning_nonsense_ecqa_it}, Table \ref{tbl:reasoning_cot_500_wrong_answer_shuffle_llama3_8b_it_grads_ecqa_train}), CREAK (Figure \ref{fig:reasoning_nonsense_creak_it}, Table \ref{tbl:reasoning_cot_500_wrong_answer_shuffle_llama3_8b_it_grads_creak_train}), Sensemaking (Figure \ref{fig:reasoning_nonsense_sensemaking_it}, Table \ref{tbl:reasoning_cot_500_wrong_answer_shuffle_llama3_8b_it_grads_sensemaking_train}).

\begin{table}[!tbh]
\centering
\small
\scalebox{0.75}{
\begin{tabular}{l|l|l|cccc}
\toprule
\multirow{2}{*}{\textbf{Dataset}} & \multirow{2}{*}{\textbf{Curve}} & \multirow{2}{*}{\textbf{Cot}} &  \multicolumn{4}{c}{\textbf{Mean Absolute Difference (MAD)}} \\
 & &  & Early & Middle & Last & All\\
\midrule
\multirow{24}{*}{AQuA} & \multirow{3}{*}{$s_Q$} 
  & None & 19.27 & 15.45 & 10.85 & 14.60 \\
& & Simplified & 3.21 & 3.18 & 1.62 & 2.50 \\
& & Detailed & 1.26 & 1.17 & 0.60 & 0.94 \\
\cmidrule{2-7}
& \multirow{3}{*}{$s_K$} 
  & None & 10.46 & 10.93 & 3.14 & 7.81 \\
& & Simplified & 1.35 & 1.33 & 0.52 & 1.01 \\
& & Detailed & 0.50 & 0.45 & 0.19 & 0.36 \\
\cmidrule{2-7}
& \multirow{3}{*}{$s_V$} 
  & None & 64.14 & 30.36 & 8.73 & 30.88 \\
& & Simplified & 9.00 & 4.50 & 1.13 & 4.35 \\
& & Detailed & 2.63 & 1.69 & 0.57 & 1.48 \\
\cmidrule{2-7}
& \multirow{3}{*}{$s_O$} 
  & None & 25.95 & 35.16 & 7.56 & 21.66 \\
& & Simplified & 3.41 & 5.21 & 1.29 & 3.14 \\
& & Detailed & 1.22 & 1.90 & 0.73 & 1.23 \\
\cmidrule{2-7}
 & \multirow{3}{*}{$r_Q$} 
  & None & 0.02 & 0.07 & 0.10 & 0.06 \\
& & Simplified & 0.01 & 0.01 & 0.09 & 0.04 \\
& & Detailed & 0.01 & 0.01 & 0.09 & 0.04 \\
\cmidrule{2-7}
& \multirow{3}{*}{$r_K$} 
  & None & 0.03 & 0.04 & 0.06 & 0.04 \\
& & Simplified & 0.02 & 0.02 & 0.03 & 0.02 \\
& & Detailed & 0.02 & 0.01 & 0.03 & 0.02 \\
\cmidrule{2-7}
 & \multirow{3}{*}{$r_V$} 
  & None & 0.02 & 0.03 & 0.03 & 0.03 \\
& & Simplified & 0.02 & 0.02 & 0.02 & 0.02 \\
& & Detailed & 0.02 & 0.01 & 0.01 & 0.02 \\
\cmidrule{2-7}
& \multirow{3}{*}{$r_O$} 
  & None & 0.02 & 0.04 & 0.08 & 0.05 \\
& & Simplified & 0.01 & 0.02 & 0.02 & 0.01 \\
& & Detailed & 0.01 & 0.01 & 0.01 & 0.01 \\
\bottomrule
\end{tabular}
}
\caption{
Statistical results for AQuA using Llama-3.1-8B-Instruct on irrelevant responses.
}
\label{tbl:reasoning_cot_500_wrong_answer_shuffle_llama3_8b_it_grads_aqua_train}
\end{table}

\begin{figure*}[!h]
\centering
\includegraphics[width=\linewidth]{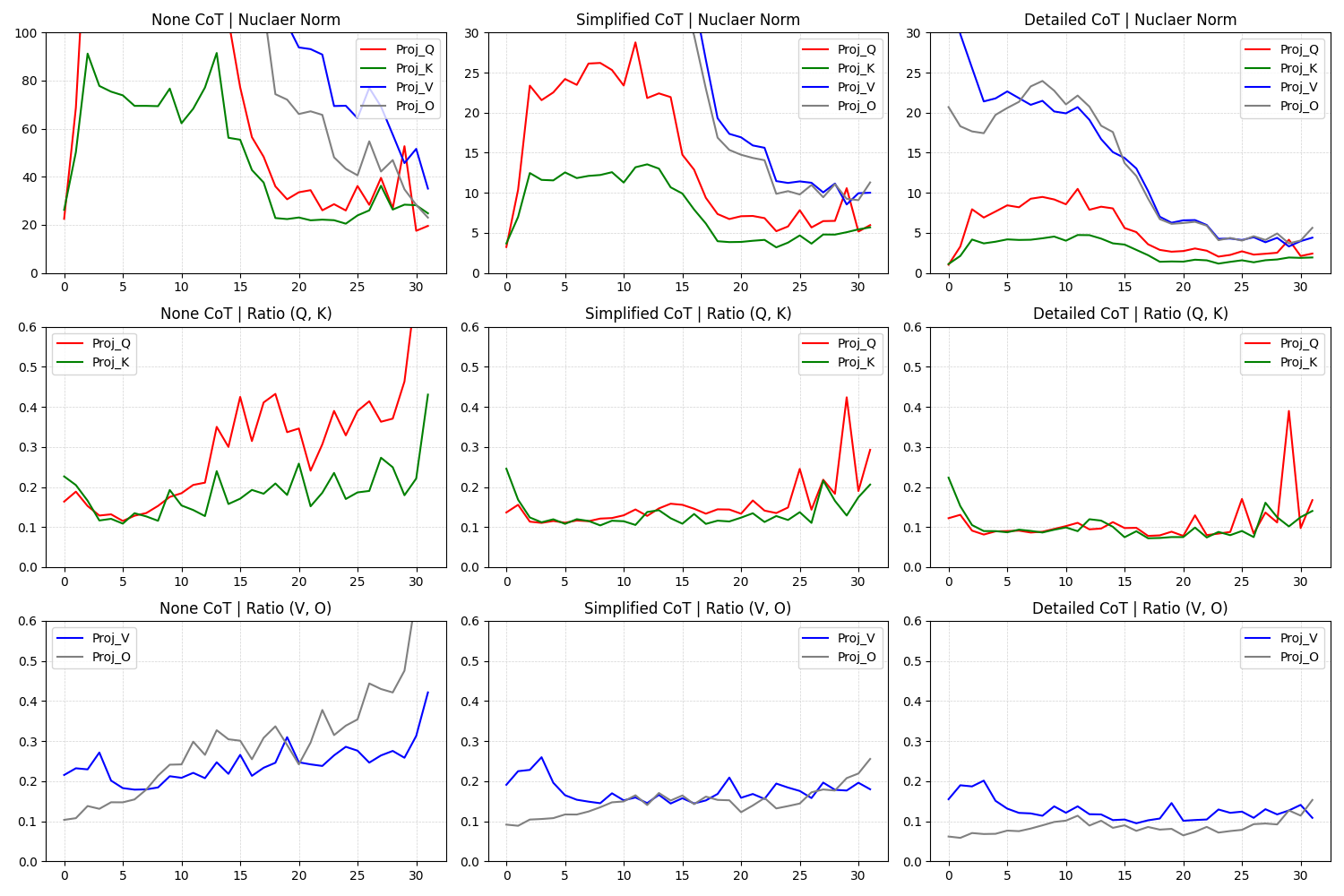}
\caption{Visualization for AQuA using Llama-3.1-8B-Instruct on irrelevant responses.}
\label{fig:reasoning_nonsense_aqua_it}
\vspace{-2mm}
\end{figure*}

\begin{table}[!tbh]
\centering
\small
\scalebox{0.75}{
\begin{tabular}{l|l|l|cccc}
\toprule
\multirow{2}{*}{\textbf{Dataset}} & \multirow{2}{*}{\textbf{Curve}} & \multirow{2}{*}{\textbf{Cot}} &  \multicolumn{4}{c}{\textbf{Mean Absolute Difference (MAD)}} \\
 & &  & Early & Middle & Last & All\\
\midrule
\multirow{24}{*}{GSM8K} & \multirow{3}{*}{$s_Q$} 
  & None & 15.97 & 18.81 & 16.46 & 16.48 \\
& & Simplified & 2.50 & 2.44 & 1.29 & 1.93 \\
& & Detailed & 1.14 & 1.14 & 0.65 & 0.90 \\
\cmidrule{2-7}
& \multirow{3}{*}{$s_K$} 
  & None & 9.54 & 11.64 & 5.46 & 8.52 \\
& & Simplified & 0.95 & 1.11 & 0.36 & 0.75 \\
& & Detailed & 0.47 & 0.45 & 0.20 & 0.35 \\
\cmidrule{2-7}
& \multirow{3}{*}{$s_V$} 
  & None & 61.59 & 36.45 & 13.56 & 34.03 \\
& & Simplified & 4.89 & 3.93 & 0.94 & 2.93 \\
& & Detailed & 2.19 & 1.62 & 0.61 & 1.33 \\
\cmidrule{2-7}
& \multirow{3}{*}{$s_O$} 
  & None & 26.92 & 42.07 & 11.22 & 24.89 \\
& & Simplified & 2.70 & 4.31 & 1.13 & 2.51 \\
& & Detailed & 1.21 & 1.87 & 0.79 & 1.22 \\
\cmidrule{2-7}
 & \multirow{3}{*}{$r_Q$} 
  & None & 0.01 & 0.03 & 0.08 & 0.04 \\
& & Simplified & 0.01 & 0.01 & 0.10 & 0.04 \\
& & Detailed & 0.01 & 0.01 & 0.11 & 0.05 \\
\cmidrule{2-7}
& \multirow{3}{*}{$r_K$} 
  & None & 0.03 & 0.04 & 0.05 & 0.04 \\
& & Simplified & 0.02 & 0.01 & 0.03 & 0.02 \\
& & Detailed & 0.02 & 0.02 & 0.03 & 0.02 \\
\cmidrule{2-7}
 & \multirow{3}{*}{$r_V$} 
  & None & 0.02 & 0.03 & 0.03 & 0.03 \\
& & Simplified & 0.02 & 0.01 & 0.01 & 0.02 \\
& & Detailed & 0.02 & 0.01 & 0.01 & 0.02 \\
\cmidrule{2-7}
& \multirow{3}{*}{$r_O$} 
  & None & 0.02 & 0.04 & 0.08 & 0.05 \\
& & Simplified & 0.01 & 0.01 & 0.02 & 0.01 \\
& & Detailed & 0.01 & 0.01 & 0.02 & 0.01 \\
\bottomrule
\end{tabular}
}
\caption{
Statistical results for GSM8K using Llama-3.1-8B-Instruct on irrelevant responses.
}
\label{tbl:reasoning_cot_500_wrong_answer_shuffle_llama3_8b_it_grads_gsm8k_train}
\end{table}

\begin{figure*}[!h]
\centering
\includegraphics[width=\linewidth]{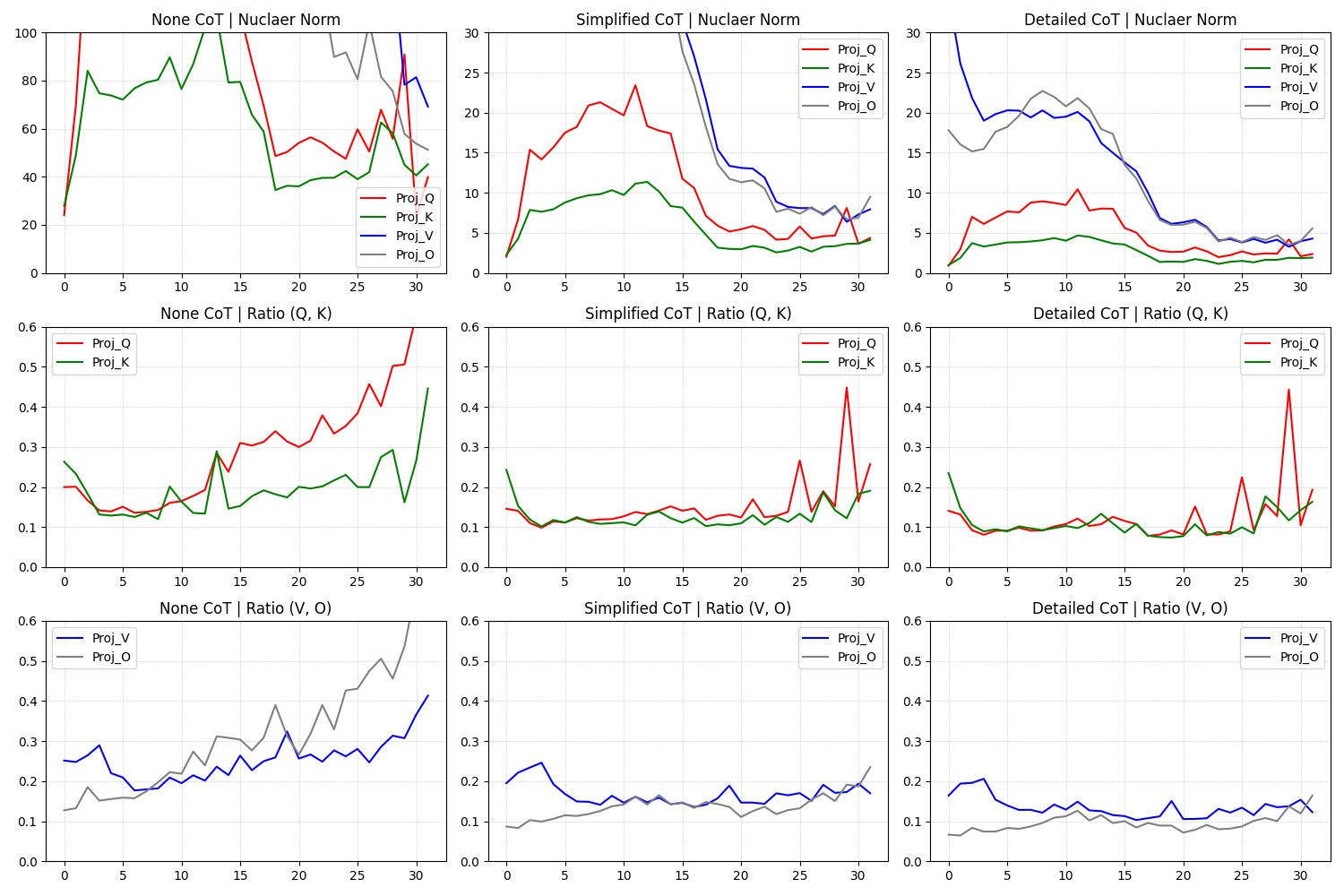}
\caption{Visualization for GSM8K using Llama-3.1-8B-Instruct on irrelevant responses.}
\label{fig:reasoning_nonsense_gsm8k_it}
\vspace{-2mm}
\end{figure*}

\begin{table}[!tbh]
\centering
\small
\scalebox{0.75}{
\begin{tabular}{l|l|l|cccc}
\toprule
\multirow{2}{*}{\textbf{Dataset}} & \multirow{2}{*}{\textbf{Curve}} & \multirow{2}{*}{\textbf{Cot}} &  \multicolumn{4}{c}{\textbf{Mean Absolute Difference (MAD)}} \\
 & &  & Early & Middle & Last & All\\
\midrule
\multirow{24}{*}{StrategyQA} & \multirow{3}{*}{$s_Q$} 
  & None & 14.08 & 20.47 & 5.98 & 12.84 \\
& & Simplified & 1.97 & 2.88 & 1.63 & 2.06 \\
& & Detailed & 0.93 & 0.92 & 0.38 & 0.69 \\
\cmidrule{2-7}
& \multirow{3}{*}{$s_K$} 
  & None & 7.59 & 8.34 & 4.67 & 6.68 \\
& & Simplified & 0.81 & 1.06 & 0.53 & 0.77 \\
& & Detailed & 0.34 & 0.36 & 0.19 & 0.28 \\
\cmidrule{2-7}
& \multirow{3}{*}{$s_V$} 
  & None & 61.77 & 28.52 & 11.64 & 30.85 \\
& & Simplified & 8.18 & 3.55 & 1.25 & 3.91 \\
& & Detailed & 2.21 & 1.28 & 0.64 & 1.25 \\
\cmidrule{2-7}
& \multirow{3}{*}{$s_O$} 
  & None & 24.75 & 28.78 & 7.41 & 18.74 \\
& & Simplified & 3.66 & 3.90 & 1.54 & 2.86 \\
& & Detailed & 1.36 & 1.40 & 0.88 & 1.16 \\
\cmidrule{2-7}
 & \multirow{3}{*}{$r_Q$} 
  & None & 0.01 & 0.04 & 0.09 & 0.05 \\
& & Simplified & 0.01 & 0.02 & 0.08 & 0.04 \\
& & Detailed & 0.02 & 0.01 & 0.06 & 0.03 \\
\cmidrule{2-7}
& \multirow{3}{*}{$r_K$} 
  & None & 0.02 & 0.02 & 0.03 & 0.02 \\
& & Simplified & 0.02 & 0.01 & 0.02 & 0.02 \\
& & Detailed & 0.01 & 0.01 & 0.02 & 0.02 \\
\cmidrule{2-7}
 & \multirow{3}{*}{$r_V$} 
  & None & 0.03 & 0.03 & 0.03 & 0.03 \\
& & Simplified & 0.03 & 0.02 & 0.02 & 0.02 \\
& & Detailed & 0.02 & 0.01 & 0.01 & 0.02 \\
\cmidrule{2-7}
& \multirow{3}{*}{$r_O$} 
  & None & 0.02 & 0.04 & 0.08 & 0.05 \\
& & Simplified & 0.01 & 0.02 & 0.02 & 0.02 \\
& & Detailed & 0.01 & 0.01 & 0.01 & 0.01 \\
\bottomrule
\end{tabular}
}
\caption{
Statistical results for StrategyQA using Llama-3.1-8B-Instruct on irrelevant responses.
}
\label{tbl:reasoning_cot_500_wrong_answer_shuffle_llama3_8b_it_grads_strategyqa_train}
\end{table}

\begin{figure*}[!h]
\centering
\includegraphics[width=\linewidth]{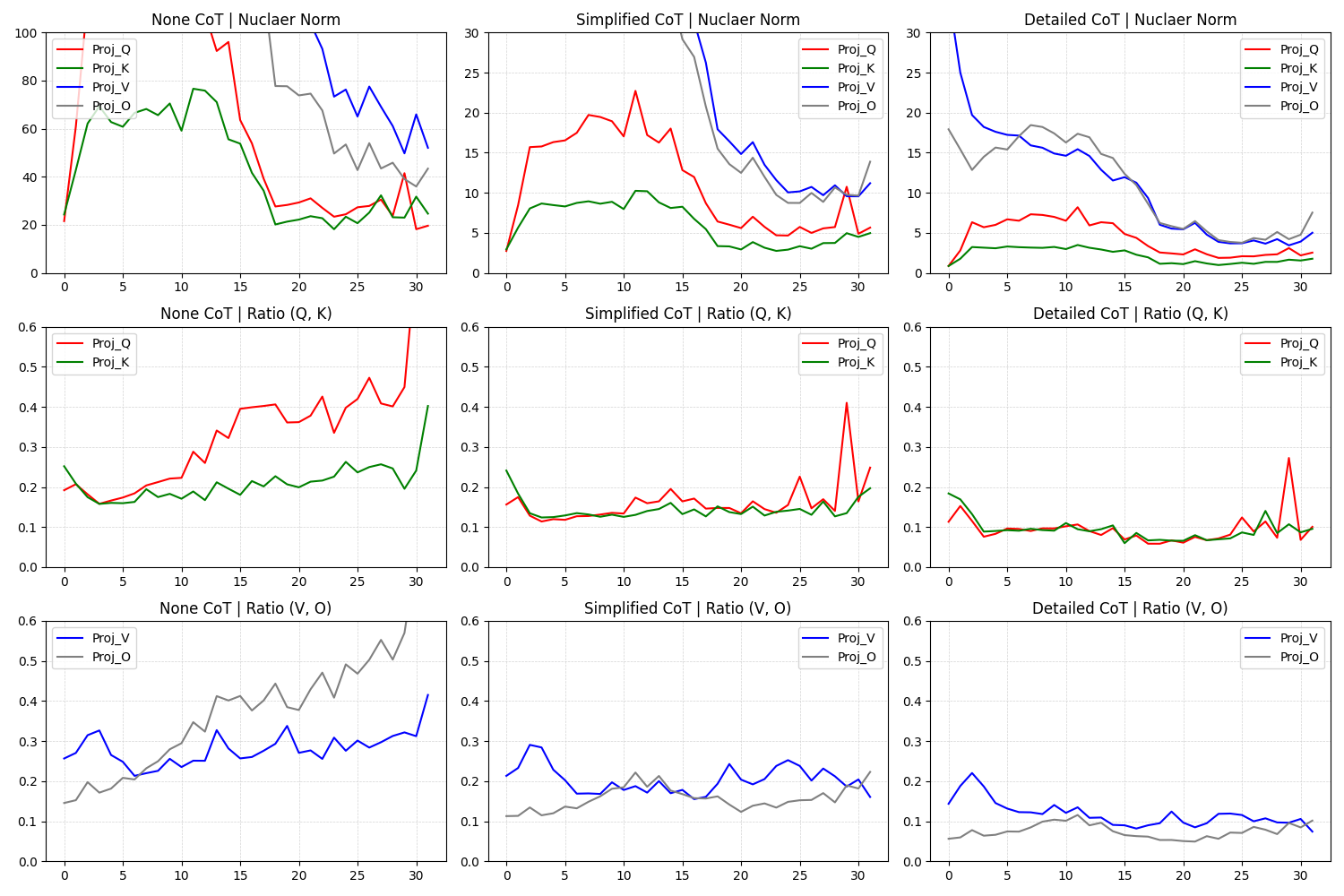}
\caption{Visualization for StrategyQA using Llama-3.1-8B-Instruct on irrelevant responses.}
\label{fig:reasoning_nonsense_strategyqa_it}
\vspace{-2mm}
\end{figure*}

\begin{table}[!tbh]
\centering
\small
\scalebox{0.75}{
\begin{tabular}{l|l|l|cccc}
\toprule
\multirow{2}{*}{\textbf{Dataset}} & \multirow{2}{*}{\textbf{Curve}} & \multirow{2}{*}{\textbf{Cot}} &  \multicolumn{4}{c}{\textbf{Mean Absolute Difference (MAD)}} \\
 & &  & Early & Middle & Last & All\\
\midrule
\multirow{24}{*}{ECQA} & \multirow{3}{*}{$s_Q$} 
  & None & 12.78 & 12.97 & 8.86 & 11.03 \\
& & Simplified & 2.73 & 3.31 & 2.05 & 2.57 \\
& & Detailed & 1.11 & 1.09 & 0.55 & 0.85 \\
\cmidrule{2-7}
& \multirow{3}{*}{$s_K$} 
  & None & 5.51 & 6.16 & 4.31 & 5.22 \\
& & Simplified & 1.08 & 1.33 & 0.83 & 1.04 \\
& & Detailed & 0.41 & 0.43 & 0.29 & 0.35 \\
\cmidrule{2-7}
& \multirow{3}{*}{$s_V$} 
  & None & 41.83 & 21.02 & 9.15 & 21.64 \\
& & Simplified & 7.75 & 4.50 & 1.56 & 4.14 \\
& & Detailed & 2.10 & 1.58 & 0.76 & 1.36 \\
\cmidrule{2-7}
& \multirow{3}{*}{$s_O$} 
  & None & 16.65 & 25.65 & 6.44 & 15.35 \\
& & Simplified & 3.12 & 5.13 & 1.64 & 3.11 \\
& & Detailed & 1.21 & 1.88 & 0.94 & 1.27 \\
\cmidrule{2-7}
 & \multirow{3}{*}{$r_Q$} 
  & None & 0.01 & 0.02 & 0.07 & 0.04 \\
& & Simplified & 0.01 & 0.02 & 0.10 & 0.05 \\
& & Detailed & 0.01 & 0.02 & 0.07 & 0.04 \\
\cmidrule{2-7}
& \multirow{3}{*}{$r_K$} 
  & None & 0.02 & 0.02 & 0.04 & 0.03 \\
& & Simplified & 0.02 & 0.01 & 0.03 & 0.02 \\
& & Detailed & 0.01 & 0.02 & 0.03 & 0.02 \\
\cmidrule{2-7}
 & \multirow{3}{*}{$r_V$} 
  & None & 0.02 & 0.02 & 0.02 & 0.02 \\
& & Simplified & 0.02 & 0.02 & 0.02 & 0.02 \\
& & Detailed & 0.01 & 0.01 & 0.01 & 0.01 \\
\cmidrule{2-7}
& \multirow{3}{*}{$r_O$} 
  & None & 0.02 & 0.03 & 0.05 & 0.03 \\
& & Simplified & 0.01 & 0.02 & 0.02 & 0.02 \\
& & Detailed & 0.01 & 0.01 & 0.01 & 0.01 \\
\bottomrule
\end{tabular}
}
\caption{
Statistical results for ECQA using Llama-3.1-8B-Instruct on irrelevant responses.
}
\label{tbl:reasoning_cot_500_wrong_answer_shuffle_llama3_8b_it_grads_ecqa_train}
\end{table}

\begin{figure*}[!h]
\centering
\includegraphics[width=\linewidth]{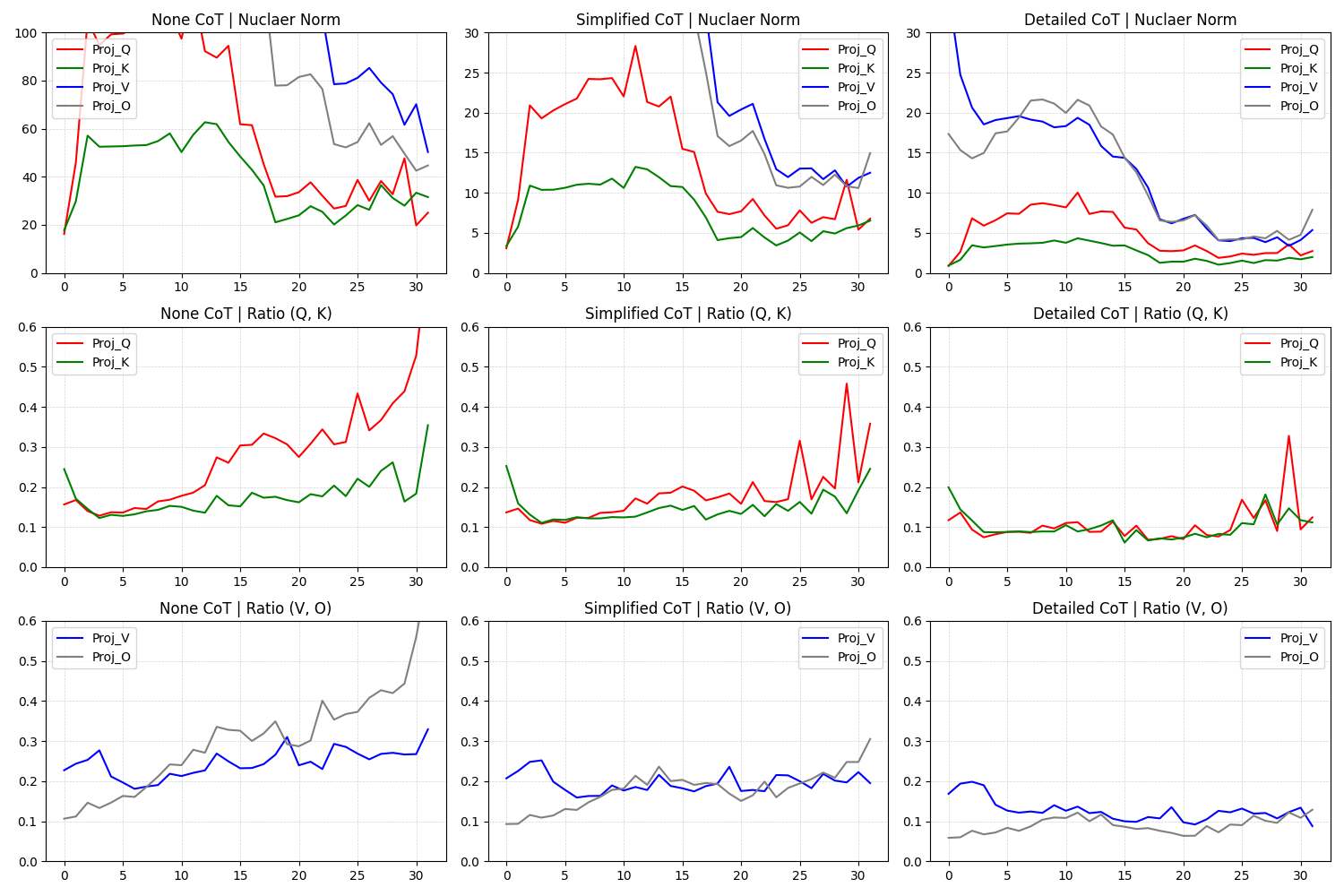}
\caption{Visualization for ECQA using Llama-3.1-8B-Instruct on irrelevant responses.}
\label{fig:reasoning_nonsense_ecqa_it}
\vspace{-2mm}
\end{figure*}

\begin{table}[!tbh]
\centering
\small
\scalebox{0.75}{
\begin{tabular}{l|l|l|cccc}
\toprule
\multirow{2}{*}{\textbf{Dataset}} & \multirow{2}{*}{\textbf{Curve}} & \multirow{2}{*}{\textbf{Cot}} &  \multicolumn{4}{c}{\textbf{Mean Absolute Difference (MAD)}} \\
 & &  & Early & Middle & Last & All\\
\midrule
\multirow{24}{*}{CREAK} & \multirow{3}{*}{$s_Q$} 
  & None & 23.94 & 27.74 & 9.32 & 19.33 \\
& & Simplified & 3.05 & 4.03 & 2.41 & 3.02 \\
& & Detailed & 1.01 & 1.03 & 0.59 & 0.82 \\
\cmidrule{2-7}
& \multirow{3}{*}{$s_K$} 
  & None & 12.65 & 16.29 & 10.48 & 12.68 \\
& & Simplified & 1.21 & 1.61 & 1.04 & 1.25 \\
& & Detailed & 0.35 & 0.39 & 0.32 & 0.34 \\
\cmidrule{2-7}
& \multirow{3}{*}{$s_V$} 
  & None & 99.38 & 52.11 & 28.59 & 54.88 \\
& & Simplified & 11.51 & 5.29 & 1.54 & 5.46 \\
& & Detailed & 2.59 & 1.47 & 0.66 & 1.42 \\
\cmidrule{2-7}
& \multirow{3}{*}{$s_O$} 
  & None & 38.19 & 54.19 & 15.59 & 33.78 \\
& & Simplified & 4.58 & 6.11 & 1.63 & 3.84 \\
& & Detailed & 1.50 & 1.76 & 0.91 & 1.31 \\
\cmidrule{2-7}
 & \multirow{3}{*}{$r_Q$} 
  & None & 0.01 & 0.03 & 0.11 & 0.05 \\
& & Simplified & 0.01 & 0.02 & 0.09 & 0.04 \\
& & Detailed & 0.01 & 0.02 & 0.07 & 0.03 \\
\cmidrule{2-7}
& \multirow{3}{*}{$r_K$} 
  & None & 0.03 & 0.03 & 0.05 & 0.04 \\
& & Simplified & 0.02 & 0.01 & 0.03 & 0.02 \\
& & Detailed & 0.02 & 0.02 & 0.03 & 0.02 \\
\cmidrule{2-7}
 & \multirow{3}{*}{$r_V$} 
  & None & 0.02 & 0.02 & 0.03 & 0.02 \\
& & Simplified & 0.02 & 0.02 & 0.02 & 0.02 \\
& & Detailed & 0.02 & 0.01 & 0.02 & 0.02 \\
\cmidrule{2-7}
& \multirow{3}{*}{$r_O$} 
  & None & 0.02 & 0.04 & 0.08 & 0.05 \\
& & Simplified & 0.01 & 0.02 & 0.03 & 0.02 \\
& & Detailed & 0.01 & 0.01 & 0.01 & 0.01 \\
\bottomrule
\end{tabular}
}
\caption{
Statistical results for CREAK using Llama-3.1-8B-Instruct on irrelevant responses.
}
\label{tbl:reasoning_cot_500_wrong_answer_shuffle_llama3_8b_it_grads_creak_train}
\end{table}

\begin{figure*}[!h]
\centering
\includegraphics[width=\linewidth]{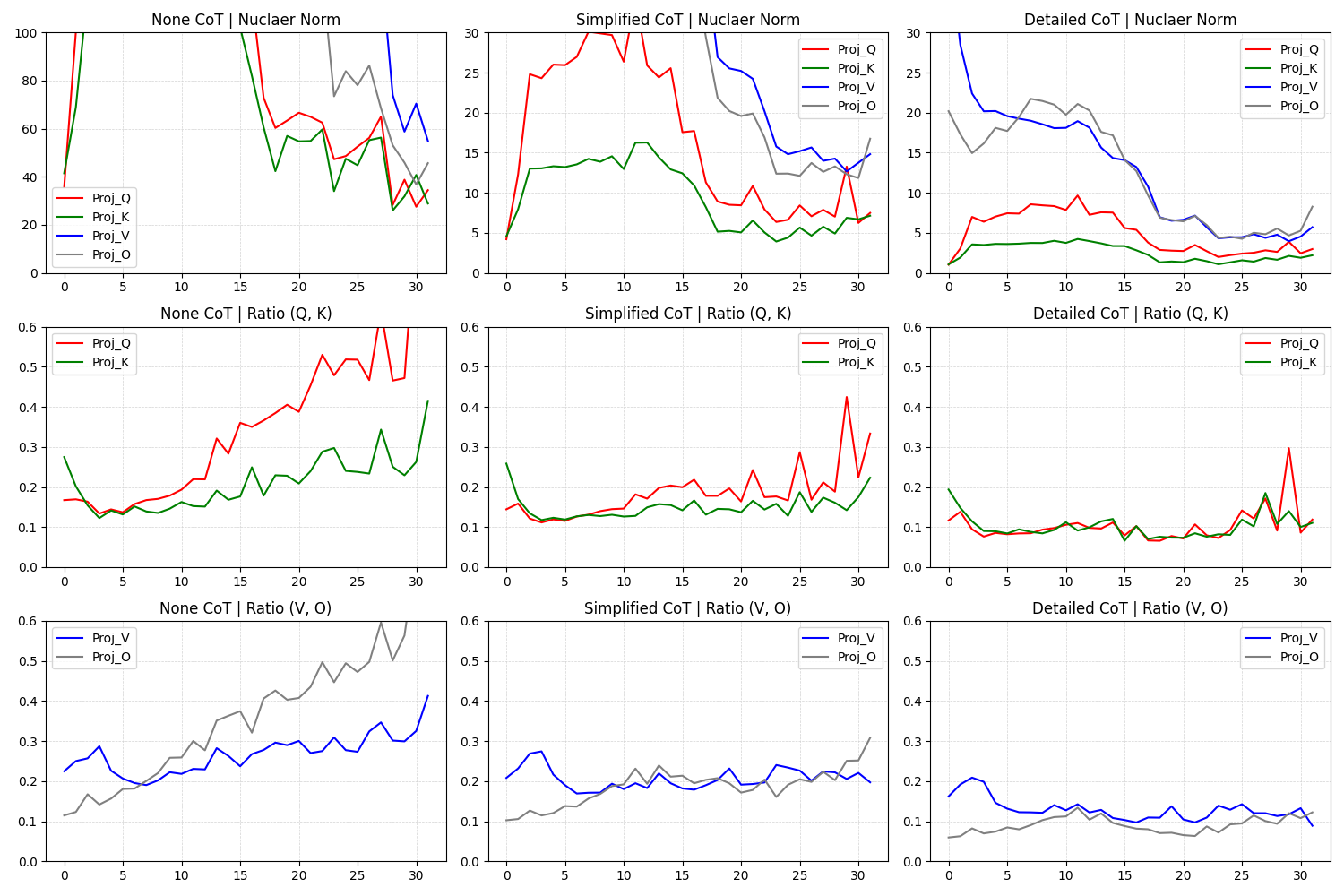}
\caption{Visualization for CREAK using Llama-3.1-8B-Instruct on irrelevant responses.}
\label{fig:reasoning_nonsense_creak_it}
\vspace{-2mm}
\end{figure*}

\begin{table}[!tbh]
\centering
\small
\scalebox{0.75}{
\begin{tabular}{l|l|l|cccc}
\toprule
\multirow{2}{*}{\textbf{Dataset}} & \multirow{2}{*}{\textbf{Curve}} & \multirow{2}{*}{\textbf{Cot}} &  \multicolumn{4}{c}{\textbf{Mean Absolute Difference (MAD)}} \\
 & &  & Early & Middle & Last & All\\
\midrule
\multirow{24}{*}{Sensemaking} & \multirow{3}{*}{$s_Q$} 
  & None & 8.17 & 9.72 & 2.41 & 6.35 \\
& & Simplified & 3.69 & 4.24 & 1.97 & 3.13 \\
& & Detailed & 1.04 & 1.09 & 0.59 & 0.84 \\
\cmidrule{2-7}
& \multirow{3}{*}{$s_K$} 
  & None & 3.67 & 4.53 & 2.74 & 3.56 \\
& & Simplified & 1.55 & 1.94 & 0.73 & 1.34 \\
& & Detailed & 0.45 & 0.41 & 0.24 & 0.35 \\
\cmidrule{2-7}
& \multirow{3}{*}{$s_V$} 
  & None & 19.95 & 13.96 & 5.37 & 12.18 \\
& & Simplified & 9.01 & 6.34 & 2.01 & 5.26 \\
& & Detailed & 1.93 & 1.52 & 0.74 & 1.29 \\
\cmidrule{2-7}
& \multirow{3}{*}{$s_O$} 
  & None & 10.11 & 14.51 & 3.31 & 8.78 \\
& & Simplified & 4.06 & 7.06 & 1.73 & 4.08 \\
& & Detailed & 1.10 & 1.84 & 0.93 & 1.24 \\
\cmidrule{2-7}
 & \multirow{3}{*}{$r_Q$} 
  & None & 0.02 & 0.06 & 0.09 & 0.05 \\
& & Simplified & 0.01 & 0.01 & 0.08 & 0.04 \\
& & Detailed & 0.01 & 0.01 & 0.10 & 0.04 \\
\cmidrule{2-7}
& \multirow{3}{*}{$r_K$} 
  & None & 0.02 & 0.02 & 0.04 & 0.03 \\
& & Simplified & 0.02 & 0.01 & 0.03 & 0.02 \\
& & Detailed & 0.02 & 0.02 & 0.02 & 0.02 \\
\cmidrule{2-7}
 & \multirow{3}{*}{$r_V$} 
  & None & 0.02 & 0.03 & 0.03 & 0.03 \\
& & Simplified & 0.02 & 0.02 & 0.02 & 0.02 \\
& & Detailed & 0.01 & 0.01 & 0.01 & 0.01 \\
\cmidrule{2-7}
& \multirow{3}{*}{$r_O$} 
  & None & 0.01 & 0.04 & 0.07 & 0.04 \\
& & Simplified & 0.01 & 0.02 & 0.03 & 0.02 \\
& & Detailed & 0.01 & 0.01 & 0.02 & 0.01 \\
\bottomrule
\end{tabular}
}
\caption{
Statistical results for Sensemaking using Llama-3.1-8B-Instruct on irrelevant responses.
}
\label{tbl:reasoning_cot_500_wrong_answer_shuffle_llama3_8b_it_grads_sensemaking_train}
\end{table}

\begin{figure*}[!h]
\centering
\includegraphics[width=\linewidth]{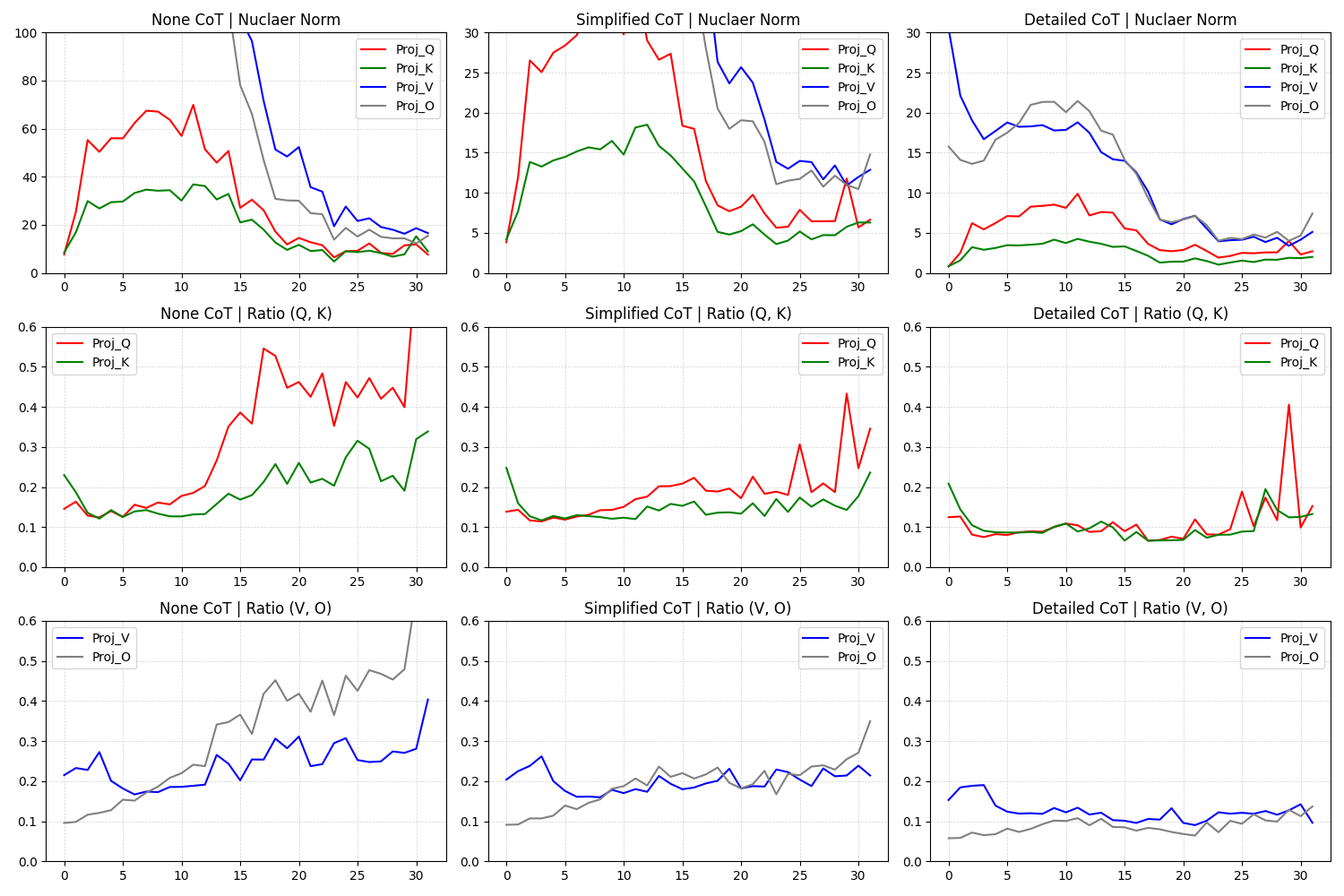}
\caption{Visualization for Sensemaking using Llama-3.1-8B-Instruct on irrelevant responses.}
\label{fig:reasoning_nonsense_sensemaking_it}
\vspace{-2mm}
\end{figure*}

\subsubsection{Wiki Tasks}

The visualizations and statistical results on Wiki tasks are shown in Figure \ref{fig:wiki_nonsense_it} and Table \ref{tbl:wiki_wiki_popularity_wrong_answer_shuffle_llama3_8b_it_grads}.

\begin{table}[!tbh]
\centering
\small
\scalebox{0.75}{
\begin{tabular}{l|l|l|cccc}
\toprule
\multirow{2}{*}{\textbf{Dataset}} & \multirow{2}{*}{\textbf{Curve}} & \multirow{2}{*}{\textbf{Cot}} &  \multicolumn{4}{c}{\textbf{Mean Absolute Difference (MAD)}} \\
 & &  & Early & Middle & Last & All\\
\midrule
\multirow{32}{*}{Wiki} & \multirow{3}{*}{$s_Q$} 
  & Len 100 & 1.61 & 2.07 & 1.37 & 1.63 \\
& & Len 500 & 1.04 & 1.01 & 0.53 & 0.82 \\
& & Len 1000 & 0.86 & 0.76 & 0.36 & 0.62 \\
& & Unpopular & 1.94 & 2.58 & 1.72 & 2.01 \\
\cmidrule{2-7}
& \multirow{3}{*}{$s_K$} 
  & Len 100 & 0.59 & 0.77 & 0.41 & 0.59 \\
& & Len 500 & 0.36 & 0.35 & 0.18 & 0.30 \\
& & Len 1000 & 0.32 & 0.26 & 0.13 & 0.23 \\
& & Unpopular & 0.82 & 0.89 & 0.65 & 0.79 \\
\cmidrule{2-7}
& \multirow{3}{*}{$s_V$} 
  & Len 100 & 7.75 & 2.36 & 0.96 & 3.37 \\
& & Len 500 & 4.06 & 1.22 & 0.67 & 1.82 \\
& & Len 1000 & 3.21 & 0.96 & 0.58 & 1.45 \\
& & Unpopular & 16.50 & 3.04 & 0.92 & 6.14 \\
\cmidrule{2-7}
& \multirow{3}{*}{$s_O$} 
  & Len 100 & 3.52 & 2.80 & 1.25 & 2.43 \\
& & Len 500 & 2.10 & 1.45 & 0.92 & 1.45 \\
& & Len 1000 & 1.74 & 1.15 & 0.82 & 1.21 \\
& & Unpopular & 6.20 & 3.38 & 1.46 & 3.51 \\
\cmidrule{2-7}
 & \multirow{3}{*}{$r_Q$} 
  & Len 100 & 0.02 & 0.02 & 0.08 & 0.04 \\
& & Len 500 & 0.02 & 0.02 & 0.06 & 0.03 \\
& & Len 1000 & 0.03 & 0.01 & 0.05 & 0.03 \\
& & Unpopular & 0.03 & 0.02 & 0.08 & 0.04 \\
\cmidrule{2-7}
& \multirow{3}{*}{$r_K$} 
  & Len 100 & 0.01 & 0.01 & 0.02 & 0.02 \\
& & Len 500 & 0.02 & 0.01 & 0.01 & 0.01 \\
& & Len 1000 & 0.03 & 0.01 & 0.01 & 0.02 \\
& & Unpopular & 0.02 & 0.01 & 0.02 & 0.02 \\
\cmidrule{2-7}
 & \multirow{3}{*}{$r_V$} 
  & Len 100 & 0.03 & 0.02 & 0.02 & 0.02 \\
& & Len 500 & 0.03 & 0.01 & 0.01 & 0.02 \\
& & Len 1000 & 0.03 & 0.01 & 0.01 & 0.02 \\
& & Unpopular & 0.05 & 0.02 & 0.02 & 0.03 \\
\cmidrule{2-7}
& \multirow{3}{*}{$r_O$} 
  & Len 100 & 0.01 & 0.02 & 0.02 & 0.02 \\
& & Len 500 & 0.01 & 0.01 & 0.01 & 0.01 \\
& & Len 1000 & 0.01 & 0.01 & 0.01 & 0.01 \\
& & Unpopular & 0.01 & 0.02 & 0.01 & 0.02 \\
\bottomrule
\end{tabular}
}
\caption{
Statistical results for Wiki using Llama-3.1-8B-Instruct on irrelevant responses.
}
\label{tbl:wiki_wiki_popularity_wrong_answer_shuffle_llama3_8b_it_grads}
\end{table}

\begin{figure*}[!h]
\centering
\includegraphics[width=\linewidth]{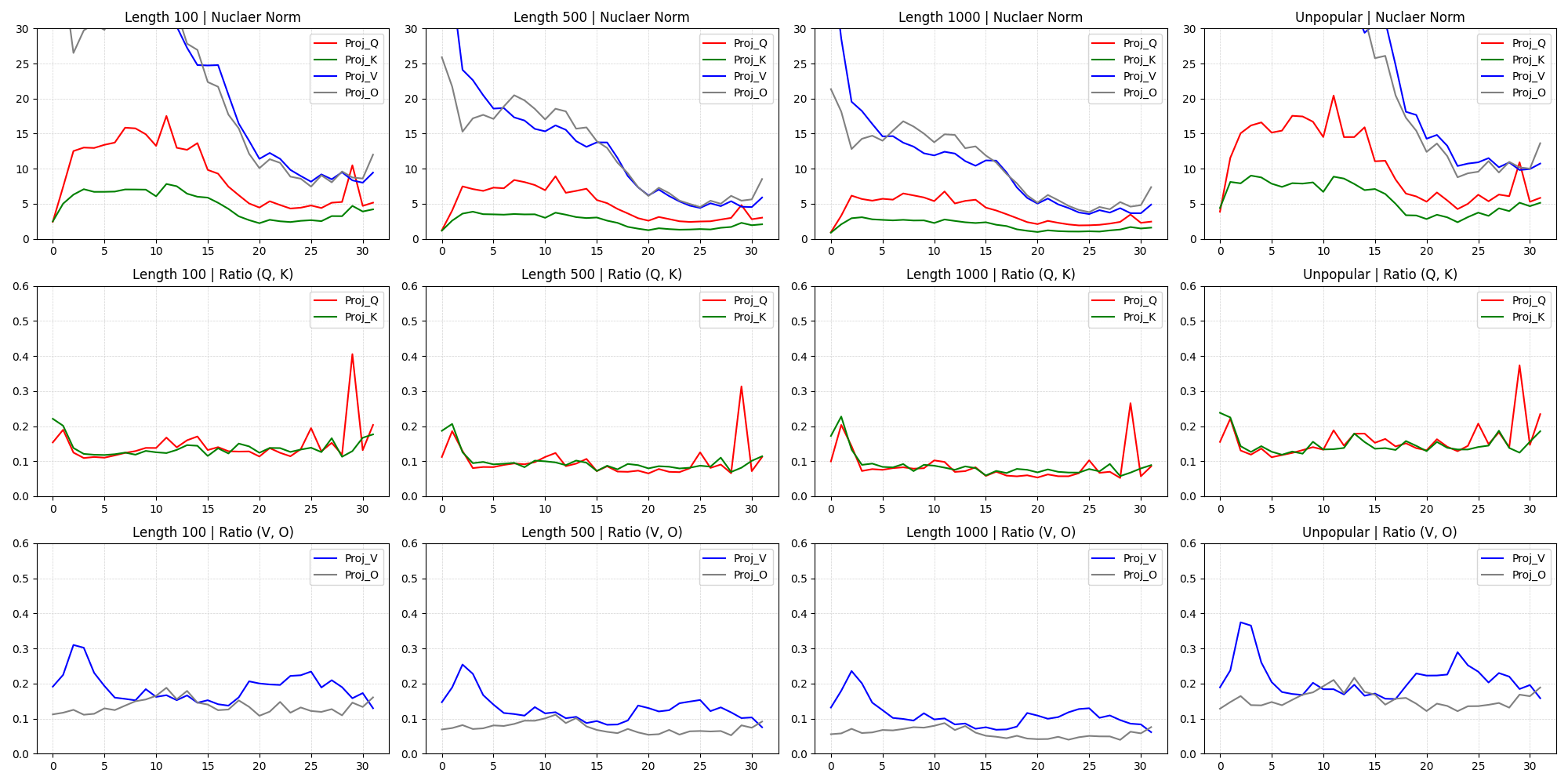}
\caption{Visualization for Wiki tasks using Llama-3.1-8B-Instruct on irrelevant responses.}
\label{fig:wiki_nonsense_it}
\vspace{-2mm}
\end{figure*}

\clearpage
\section{Results on Qwen2-1.5B }
\subsection{Pre-trained LLM on Correct Responses}

\subsubsection{Reasoning Tasks}

The visualizations and statistical results on MATH tasks: MATH-Algebra (Figure \ref{fig:math_correct_algebra}, Table \ref{tbl:math_lighteval_math_qwen2_1_5b_grads_algebra}), MATH-Counting (Figure \ref{fig:math_correct_counting}, Table \ref{tbl:math_lighteval_math_qwen2_1_5b_grads_counting_probability}), MATH-Geometry (Figure \ref{fig:math_correct_geometry}, Table \ref{tbl:math_lighteval_math_qwen2_1_5b_grads_geometry}).

\begin{table}[!tbh]
\centering
\small
\scalebox{0.75}{
\begin{tabular}{l|l|l|cccc}
\toprule
\multirow{2}{*}{\textbf{Dataset}} & \multirow{2}{*}{\textbf{Curve}} & \multirow{2}{*}{\textbf{Cot}} &  \multicolumn{4}{c}{\textbf{Mean Absolute Difference (MAD)}} \\
 & &  & Early & Middle & Last & All\\
\midrule
\multirow{16}{*}{Algebra} & \multirow{3}{*}{$s_Q$} 
  & Simplified & 0.32 & 0.32 & 0.51 & 0.38 \\
& & Detailed & 0.19 & 0.24 & 0.30 & 0.25 \\
\cmidrule{2-7}
& \multirow{3}{*}{$s_K$} 
  & Simplified & 0.32 & 0.34 & 0.54 & 0.39 \\
& & Detailed & 0.17 & 0.21 & 0.40 & 0.25 \\
\cmidrule{2-7}
& \multirow{3}{*}{$s_V$} 
  & Simplified & 1.61 & 1.15 & 0.42 & 1.02 \\
& & Detailed & 0.91 & 0.54 & 0.27 & 0.53 \\
\cmidrule{2-7}
& \multirow{3}{*}{$s_O$} 
  & Simplified & 1.15 & 1.24 & 0.37 & 0.97 \\
& & Detailed & 0.66 & 0.66 & 0.24 & 0.54 \\
\cmidrule{2-7}
 & \multirow{3}{*}{$r_Q$} 
  & Simplified & 0.01 & 0.01 & 0.03 & 0.01 \\
& & Detailed & 0.01 & 0.01 & 0.03 & 0.01 \\
\cmidrule{2-7}
& \multirow{3}{*}{$r_K$} 
  & Simplified & 0.02 & 0.03 & 0.01 & 0.02 \\
& & Detailed & 0.02 & 0.03 & 0.02 & 0.03 \\
\cmidrule{2-7}
 & \multirow{3}{*}{$r_V$} 
  & Simplified & 0.03 & 0.02 & 0.04 & 0.03 \\
& & Detailed & 0.03 & 0.02 & 0.03 & 0.02 \\
\cmidrule{2-7}
& \multirow{3}{*}{$r_O$} 
  & Simplified & 0.01 & 0.02 & 0.07 & 0.03 \\
& & Detailed & 0.01 & 0.01 & 0.06 & 0.03 \\
\bottomrule
\end{tabular}
}
\caption{
Statistical results for MATH-Algebra using Qwen2-1.5B on correct responses.
}
\label{tbl:math_lighteval_math_qwen2_1_5b_grads_algebra}
\end{table}

\begin{figure}[!h]
\centering
\includegraphics[width=\linewidth]{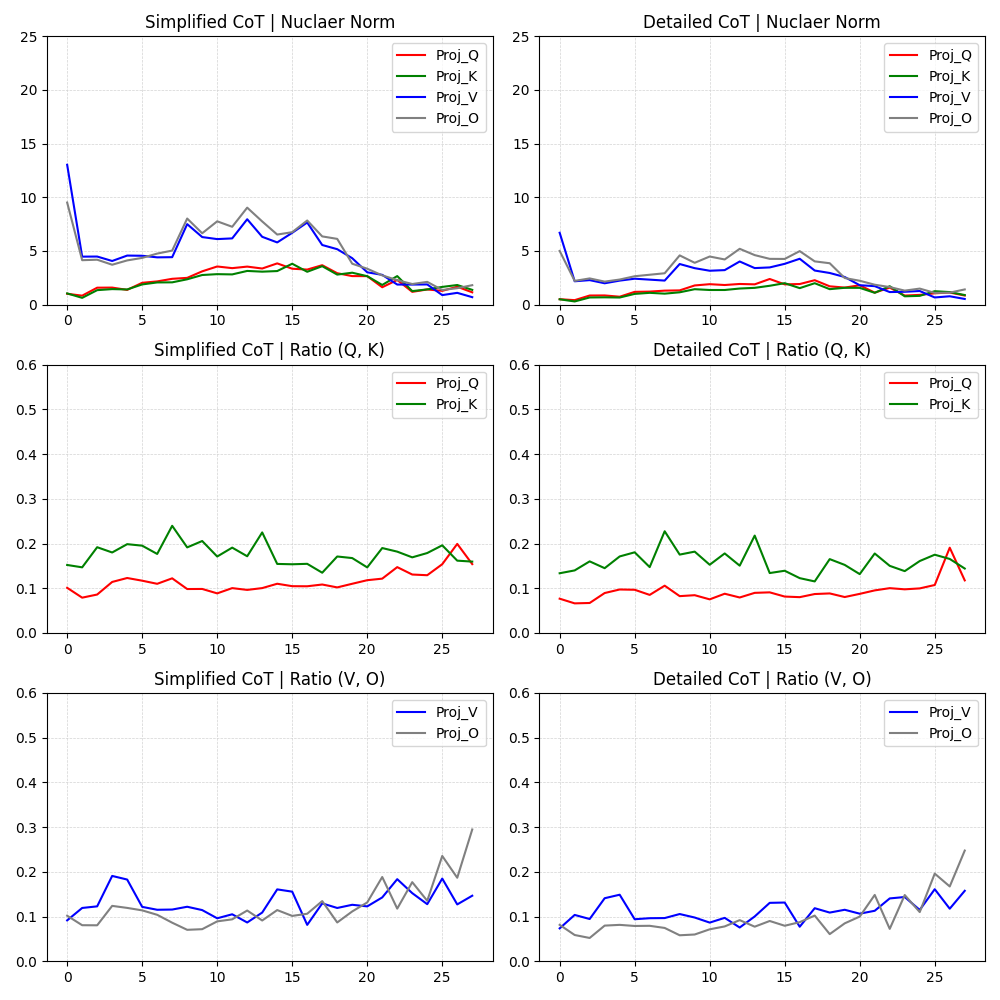}
\caption{Visualization for MATH-Algebra using Qwen2-1.5B on correct responses.}
\label{fig:math_correct_algebra}
\vspace{-2mm}
\end{figure}

\begin{table}[!tbh]
\centering
\small
\scalebox{0.75}{
\begin{tabular}{l|l|l|cccc}
\toprule
\multirow{2}{*}{\textbf{Dataset}} & \multirow{2}{*}{\textbf{Curve}} & \multirow{2}{*}{\textbf{Cot}} &  \multicolumn{4}{c}{\textbf{Mean Absolute Difference (MAD)}} \\
 & &  & Early & Middle & Last & All\\
\midrule
\multirow{16}{*}{Counting} & \multirow{3}{*}{$s_Q$} 
  & Simplified & 0.37 & 0.34 & 0.49 & 0.41 \\
& & Detailed & 0.23 & 0.29 & 0.32 & 0.30 \\
\cmidrule{2-7}
& \multirow{3}{*}{$s_K$} 
  & Simplified & 0.36 & 0.30 & 0.55 & 0.38 \\
& & Detailed & 0.20 & 0.22 & 0.47 & 0.27 \\
\cmidrule{2-7}
& \multirow{3}{*}{$s_V$} 
  & Simplified & 1.81 & 1.18 & 0.48 & 1.09 \\
& & Detailed & 1.06 & 0.56 & 0.30 & 0.58 \\
\cmidrule{2-7}
& \multirow{3}{*}{$s_O$} 
  & Simplified & 1.26 & 1.25 & 0.46 & 1.03 \\
& & Detailed & 0.77 & 0.73 & 0.31 & 0.62 \\
\cmidrule{2-7}
 & \multirow{3}{*}{$r_Q$} 
  & Simplified & 0.01 & 0.01 & 0.02 & 0.01 \\
& & Detailed & 0.01 & 0.01 & 0.02 & 0.01 \\
\cmidrule{2-7}
& \multirow{3}{*}{$r_K$} 
  & Simplified & 0.02 & 0.03 & 0.02 & 0.02 \\
& & Detailed & 0.02 & 0.03 & 0.02 & 0.03 \\
\cmidrule{2-7}
 & \multirow{3}{*}{$r_V$} 
  & Simplified & 0.03 & 0.02 & 0.04 & 0.03 \\
& & Detailed & 0.02 & 0.02 & 0.03 & 0.02 \\
\cmidrule{2-7}
& \multirow{3}{*}{$r_O$} 
  & Simplified & 0.02 & 0.02 & 0.07 & 0.03 \\
& & Detailed & 0.01 & 0.02 & 0.06 & 0.02 \\
\bottomrule
\end{tabular}
}
\caption{
Statistical results for MATH-Counting using Qwen2-1.5B on correct responses.
}
\label{tbl:math_lighteval_math_qwen2_1_5b_grads_counting_probability}
\end{table}

\begin{figure}[!h]
\centering
\includegraphics[width=\linewidth]{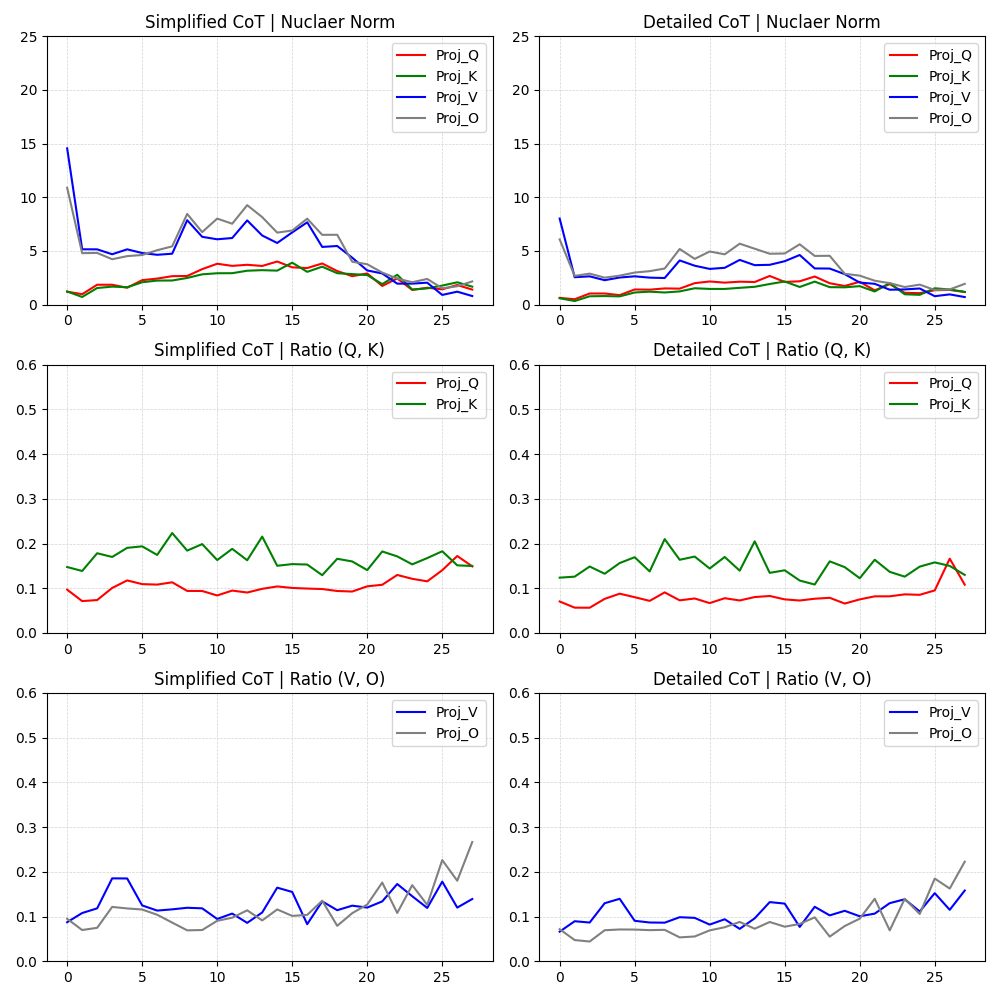}
\caption{Visualization for MATH-Counting using Qwen2-1.5B on correct responses.}
\label{fig:math_correct_counting}
\vspace{-2mm}
\end{figure}

\begin{table}[!tbh]
\centering
\small
\scalebox{0.75}{
\begin{tabular}{l|l|l|cccc}
\toprule
\multirow{2}{*}{\textbf{Dataset}} & \multirow{2}{*}{\textbf{Curve}} & \multirow{2}{*}{\textbf{Cot}} &  \multicolumn{4}{c}{\textbf{Mean Absolute Difference (MAD)}} \\
 & &  & Early & Middle & Last & All\\
\midrule
\multirow{16}{*}{Geometry} & \multirow{3}{*}{$s_Q$} 
  & Simplified & 0.36 & 0.40 & 0.61 & 0.46 \\
& & Detailed & 0.26 & 0.32 & 0.40 & 0.34 \\
\cmidrule{2-7}
& \multirow{3}{*}{$s_K$} 
  & Simplified & 0.32 & 0.30 & 0.74 & 0.41 \\
& & Detailed & 0.23 & 0.23 & 0.52 & 0.30 \\
\cmidrule{2-7}
& \multirow{3}{*}{$s_V$} 
  & Simplified & 1.78 & 1.06 & 0.53 & 1.03 \\
& & Detailed & 1.27 & 0.60 & 0.34 & 0.65 \\
\cmidrule{2-7}
& \multirow{3}{*}{$s_O$} 
  & Simplified & 1.42 & 1.35 & 0.40 & 1.08 \\
& & Detailed & 0.95 & 0.84 & 0.30 & 0.70 \\
\cmidrule{2-7}
 & \multirow{3}{*}{$r_Q$} 
  & Simplified & 0.01 & 0.01 & 0.02 & 0.01 \\
& & Detailed & 0.01 & 0.01 & 0.03 & 0.01 \\
\cmidrule{2-7}
& \multirow{3}{*}{$r_K$} 
  & Simplified & 0.01 & 0.03 & 0.01 & 0.02 \\
& & Detailed & 0.02 & 0.03 & 0.02 & 0.03 \\
\cmidrule{2-7}
 & \multirow{3}{*}{$r_V$} 
  & Simplified & 0.03 & 0.02 & 0.03 & 0.02 \\
& & Detailed & 0.02 & 0.02 & 0.03 & 0.02 \\
\cmidrule{2-7}
& \multirow{3}{*}{$r_O$} 
  & Simplified & 0.01 & 0.01 & 0.05 & 0.02 \\
& & Detailed & 0.01 & 0.01 & 0.05 & 0.02 \\
\bottomrule
\end{tabular}
}
\caption{
Statistical results for MATH-Geometry using Qwen2-1.5B on correct responses.
}
\label{tbl:math_lighteval_math_qwen2_1_5b_grads_geometry}
\end{table}

\begin{figure}[!h]
\centering
\includegraphics[width=\linewidth]{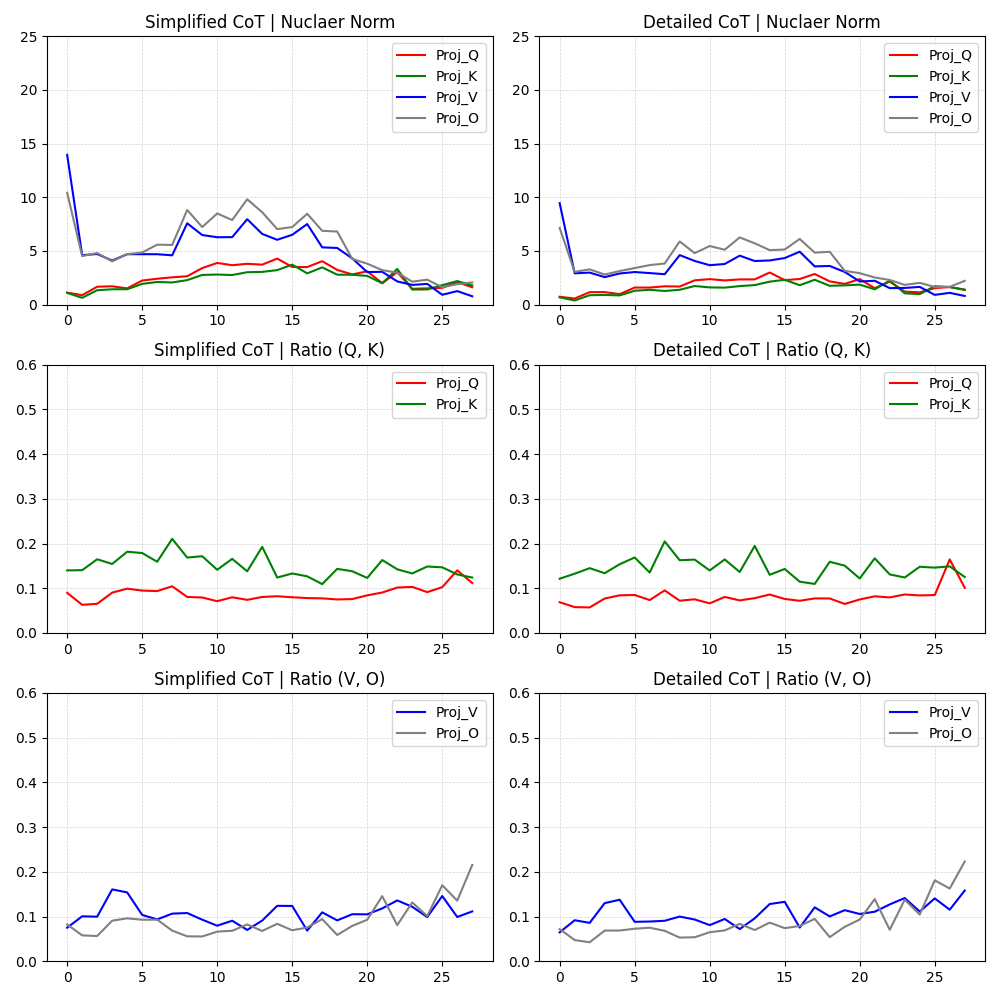}
\caption{Visualization for MATH-Geometry using Qwen2-1.5B on correct responses.}
\label{fig:math_correct_geometry}
\vspace{-2mm}
\end{figure}

The visualizations and statistical results on other reasoning tasks: AQuA (Figure \ref{fig:reasoning_correct_aqua}, Table \ref{tbl:reasoning_cot_500_qwen2_1_5b_grads_aqua_train}), GSM8K (Figure \ref{fig:reasoning_correct_gsm8k}, Table \ref{tbl:reasoning_cot_500_qwen2_1_5b_grads_gsm8k_train}), StrategyQA (Figure \ref{fig:reasoning_correct_strategyqa}, Table \ref{tbl:reasoning_cot_500_qwen2_1_5b_grads_strategyqa_train}), ECQA (Figure \ref{fig:reasoning_correct_ecqa}, Table \ref{tbl:reasoning_cot_500_qwen2_1_5b_grads_ecqa_train}), CREAK (Figure \ref{fig:reasoning_correct_creak}, Table \ref{tbl:reasoning_cot_500_qwen2_1_5b_grads_creak_train}), Sensemaking (Figure \ref{fig:reasoning_correct_sensemaking}, Table \ref{tbl:reasoning_cot_500_qwen2_1_5b_grads_sensemaking_train}).

\begin{table}[!tbh]
\centering
\small
\scalebox{0.75}{
\begin{tabular}{l|l|l|cccc}
\toprule
\multirow{2}{*}{\textbf{Dataset}} & \multirow{2}{*}{\textbf{Curve}} & \multirow{2}{*}{\textbf{Cot}} &  \multicolumn{4}{c}{\textbf{Mean Absolute Difference (MAD)}} \\
 & &  & Early & Middle & Last & All\\
\midrule
\multirow{24}{*}{AQuA} & \multirow{3}{*}{$s_Q$} 
  & None & 5.76 & 4.13 & 3.49 & 4.42 \\
& & Simplified & 0.89 & 0.52 & 0.77 & 0.69 \\
& & Detailed & 0.23 & 0.28 & 0.29 & 0.28 \\
\cmidrule{2-7}
& \multirow{3}{*}{$s_K$} 
  & None & 7.20 & 6.29 & 8.40 & 7.06 \\
& & Simplified & 1.01 & 0.56 & 1.11 & 0.81 \\
& & Detailed & 0.22 & 0.21 & 0.42 & 0.27 \\
\cmidrule{2-7}
& \multirow{3}{*}{$s_V$} 
  & None & 37.29 & 16.12 & 3.94 & 17.32 \\
& & Simplified & 5.08 & 2.14 & 0.86 & 2.36 \\
& & Detailed & 1.15 & 0.62 & 0.33 & 0.64 \\
\cmidrule{2-7}
& \multirow{3}{*}{$s_O$} 
  & None & 23.79 & 14.35 & 3.04 & 12.91 \\
& & Simplified & 3.31 & 2.18 & 0.63 & 1.97 \\
& & Detailed & 0.82 & 0.75 & 0.29 & 0.64 \\
\cmidrule{2-7}
 & \multirow{3}{*}{$r_Q$} 
  & None & 0.03 & 0.06 & 0.21 & 0.09 \\
& & Simplified & 0.02 & 0.01 & 0.02 & 0.02 \\
& & Detailed & 0.01 & 0.01 & 0.02 & 0.01 \\
\cmidrule{2-7}
& \multirow{3}{*}{$r_K$} 
  & None & 0.04 & 0.04 & 0.14 & 0.06 \\
& & Simplified & 0.03 & 0.03 & 0.02 & 0.03 \\
& & Detailed & 0.02 & 0.03 & 0.01 & 0.02 \\
\cmidrule{2-7}
 & \multirow{3}{*}{$r_V$} 
  & None & 0.04 & 0.06 & 0.03 & 0.05 \\
& & Simplified & 0.04 & 0.03 & 0.03 & 0.03 \\
& & Detailed & 0.03 & 0.02 & 0.03 & 0.02 \\
\cmidrule{2-7}
& \multirow{3}{*}{$r_O$} 
  & None & 0.02 & 0.04 & 0.09 & 0.05 \\
& & Simplified & 0.02 & 0.02 & 0.08 & 0.04 \\
& & Detailed & 0.01 & 0.02 & 0.06 & 0.03 \\
\bottomrule
\end{tabular}
}
\caption{
Statistical results for AQuA using Qwen2-1.5B on correct responses.
}
\label{tbl:reasoning_cot_500_qwen2_1_5b_grads_aqua_train}
\end{table}

\begin{figure*}[!h]
\centering
\includegraphics[width=\linewidth]{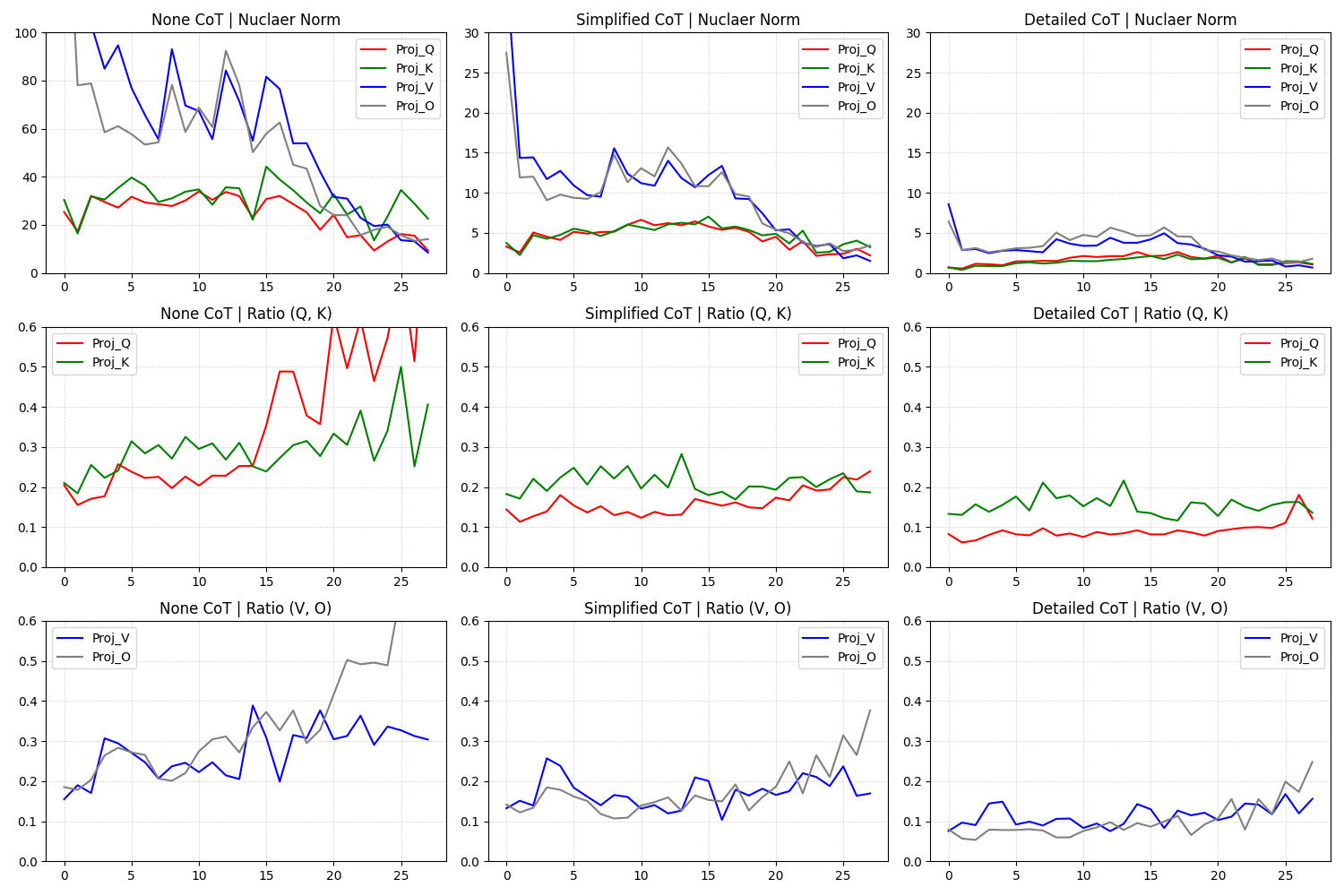}
\caption{Visualization for AQuA using Qwen2-1.5B on correct responses.}
\label{fig:reasoning_correct_aqua}
\vspace{-2mm}
\end{figure*}

\begin{table}[!tbh]
\centering
\small
\scalebox{0.75}{
\begin{tabular}{l|l|l|cccc}
\toprule
\multirow{2}{*}{\textbf{Dataset}} & \multirow{2}{*}{\textbf{Curve}} & \multirow{2}{*}{\textbf{Cot}} &  \multicolumn{4}{c}{\textbf{Mean Absolute Difference (MAD)}} \\
 & &  & Early & Middle & Last & All\\
\midrule
\multirow{24}{*}{GSM8K} & \multirow{3}{*}{$s_Q$} 
  & None & 3.39 & 2.78 & 6.73 & 3.95 \\
& & Simplified & 0.29 & 0.27 & 0.32 & 0.32 \\
& & Detailed & 0.22 & 0.23 & 0.26 & 0.25 \\
\cmidrule{2-7}
& \multirow{3}{*}{$s_K$} 
  & None & 4.47 & 5.53 & 9.31 & 6.02 \\
& & Simplified & 0.34 & 0.38 & 0.57 & 0.42 \\
& & Detailed & 0.22 & 0.25 & 0.45 & 0.30 \\
\cmidrule{2-7}
& \multirow{3}{*}{$s_V$} 
  & None & 23.57 & 14.68 & 5.20 & 13.66 \\
& & Simplified & 1.57 & 1.12 & 0.37 & 0.99 \\
& & Detailed & 1.14 & 0.69 & 0.28 & 0.67 \\
\cmidrule{2-7}
& \multirow{3}{*}{$s_O$} 
  & None & 14.85 & 14.41 & 3.40 & 11.45 \\
& & Simplified & 0.99 & 1.12 & 0.32 & 0.88 \\
& & Detailed & 0.77 & 0.79 & 0.26 & 0.64 \\
\cmidrule{2-7}
 & \multirow{3}{*}{$r_Q$} 
  & None & 0.02 & 0.04 & 0.13 & 0.06 \\
& & Simplified & 0.02 & 0.01 & 0.02 & 0.02 \\
& & Detailed & 0.02 & 0.01 & 0.03 & 0.01 \\
\cmidrule{2-7}
& \multirow{3}{*}{$r_K$} 
  & None & 0.04 & 0.03 & 0.04 & 0.04 \\
& & Simplified & 0.02 & 0.03 & 0.03 & 0.03 \\
& & Detailed & 0.02 & 0.03 & 0.02 & 0.03 \\
\cmidrule{2-7}
 & \multirow{3}{*}{$r_V$} 
  & None & 0.03 & 0.06 & 0.04 & 0.05 \\
& & Simplified & 0.03 & 0.03 & 0.03 & 0.03 \\
& & Detailed & 0.03 & 0.02 & 0.03 & 0.03 \\
\cmidrule{2-7}
& \multirow{3}{*}{$r_O$} 
  & None & 0.03 & 0.05 & 0.10 & 0.06 \\
& & Simplified & 0.02 & 0.02 & 0.08 & 0.04 \\
& & Detailed & 0.01 & 0.02 & 0.06 & 0.03 \\
\bottomrule
\end{tabular}
}
\caption{
Statistical results for GSM8K using Qwen2-1.5B on correct responses.
}
\label{tbl:reasoning_cot_500_qwen2_1_5b_grads_gsm8k_train}
\end{table}

\begin{figure*}[!h]
\centering
\includegraphics[width=\linewidth]{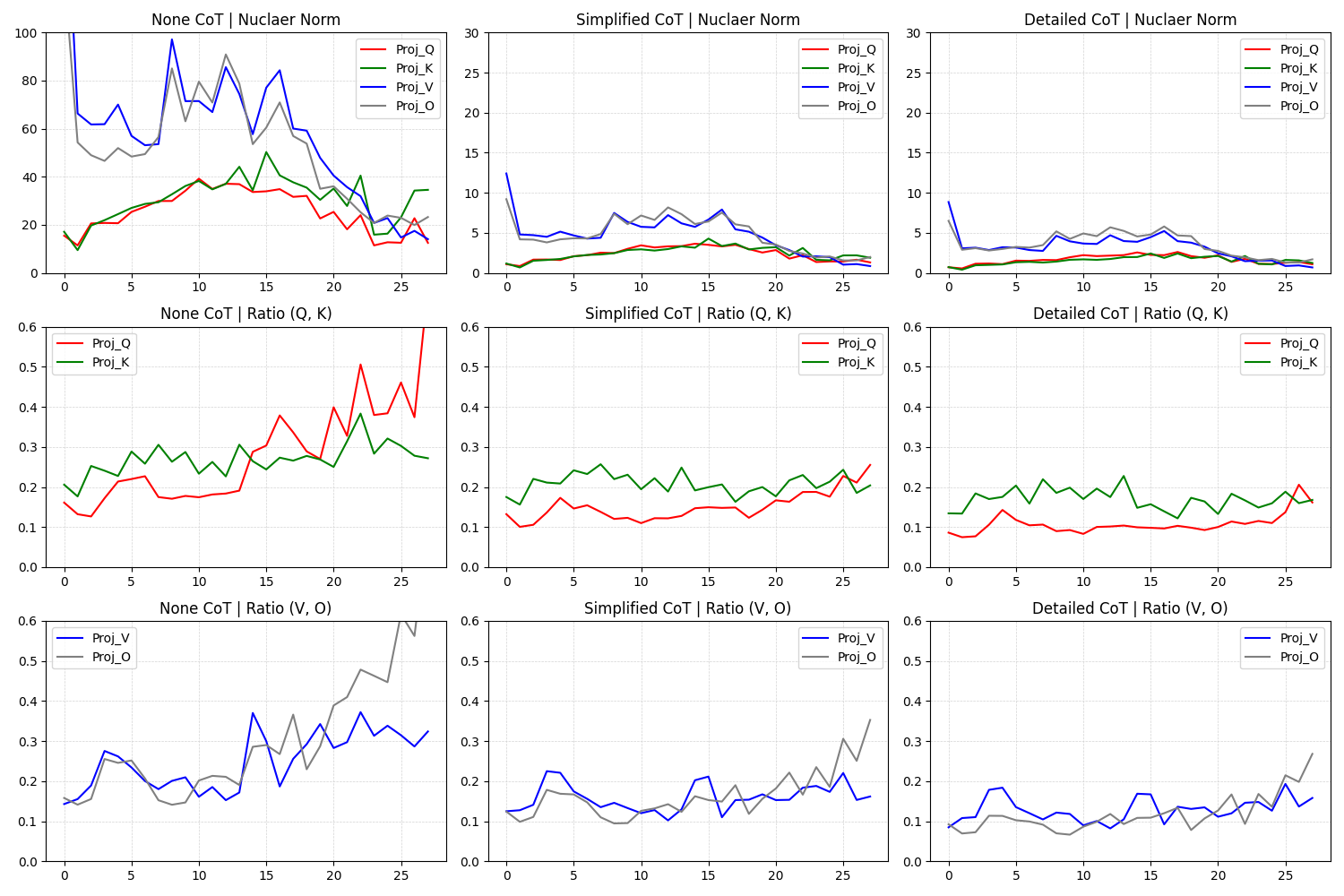}
\caption{Visualization for GSM8K using Qwen2-1.5B on correct responses.}
\label{fig:reasoning_correct_gsm8k}
\vspace{-2mm}
\end{figure*}

\begin{table}[!tbh]
\centering
\small
\scalebox{0.75}{
\begin{tabular}{l|l|l|cccc}
\toprule
\multirow{2}{*}{\textbf{Dataset}} & \multirow{2}{*}{\textbf{Curve}} & \multirow{2}{*}{\textbf{Cot}} &  \multicolumn{4}{c}{\textbf{Mean Absolute Difference (MAD)}} \\
 & &  & Early & Middle & Last & All\\
\midrule
\multirow{24}{*}{StrategyQA} & \multirow{3}{*}{$s_Q$} 
  & None & 3.03 & 1.74 & 0.90 & 1.87 \\
& & Simplified & 0.58 & 0.32 & 0.40 & 0.41 \\
& & Detailed & 0.27 & 0.33 & 0.27 & 0.32 \\
\cmidrule{2-7}
& \multirow{3}{*}{$s_K$} 
  & None & 5.11 & 3.58 & 1.88 & 3.57 \\
& & Simplified & 0.73 & 0.39 & 0.45 & 0.48 \\
& & Detailed & 0.23 & 0.22 & 0.49 & 0.28 \\
\cmidrule{2-7}
& \multirow{3}{*}{$s_V$} 
  & None & 28.52 & 10.02 & 3.49 & 12.07 \\
& & Simplified & 4.51 & 1.53 & 1.05 & 2.04 \\
& & Detailed & 1.57 & 0.63 & 0.47 & 0.77 \\
\cmidrule{2-7}
& \multirow{3}{*}{$s_O$} 
  & None & 20.65 & 6.22 & 4.33 & 8.83 \\
& & Simplified & 3.06 & 1.46 & 0.94 & 1.65 \\
& & Detailed & 1.00 & 0.90 & 0.52 & 0.81 \\
\cmidrule{2-7}
 & \multirow{3}{*}{$r_Q$} 
  & None & 0.04 & 0.07 & 0.07 & 0.06 \\
& & Simplified & 0.01 & 0.01 & 0.02 & 0.01 \\
& & Detailed & 0.01 & 0.01 & 0.01 & 0.01 \\
\cmidrule{2-7}
& \multirow{3}{*}{$r_K$} 
  & None & 0.05 & 0.04 & 0.09 & 0.05 \\
& & Simplified & 0.03 & 0.03 & 0.03 & 0.03 \\
& & Detailed & 0.02 & 0.03 & 0.01 & 0.02 \\
\cmidrule{2-7}
 & \multirow{3}{*}{$r_V$} 
  & None & 0.06 & 0.10 & 0.08 & 0.08 \\
& & Simplified & 0.04 & 0.06 & 0.10 & 0.06 \\
& & Detailed & 0.03 & 0.02 & 0.03 & 0.02 \\
\cmidrule{2-7}
& \multirow{3}{*}{$r_O$} 
  & None & 0.04 & 0.07 & 0.05 & 0.05 \\
& & Simplified & 0.01 & 0.04 & 0.08 & 0.04 \\
& & Detailed & 0.01 & 0.02 & 0.05 & 0.02 \\
\bottomrule
\end{tabular}
}
\caption{
Statistical results for StrategyQA using Qwen2-1.5B on correct responses.
}
\label{tbl:reasoning_cot_500_qwen2_1_5b_grads_strategyqa_train}
\end{table}

\begin{figure*}[!h]
\centering
\includegraphics[width=\linewidth]{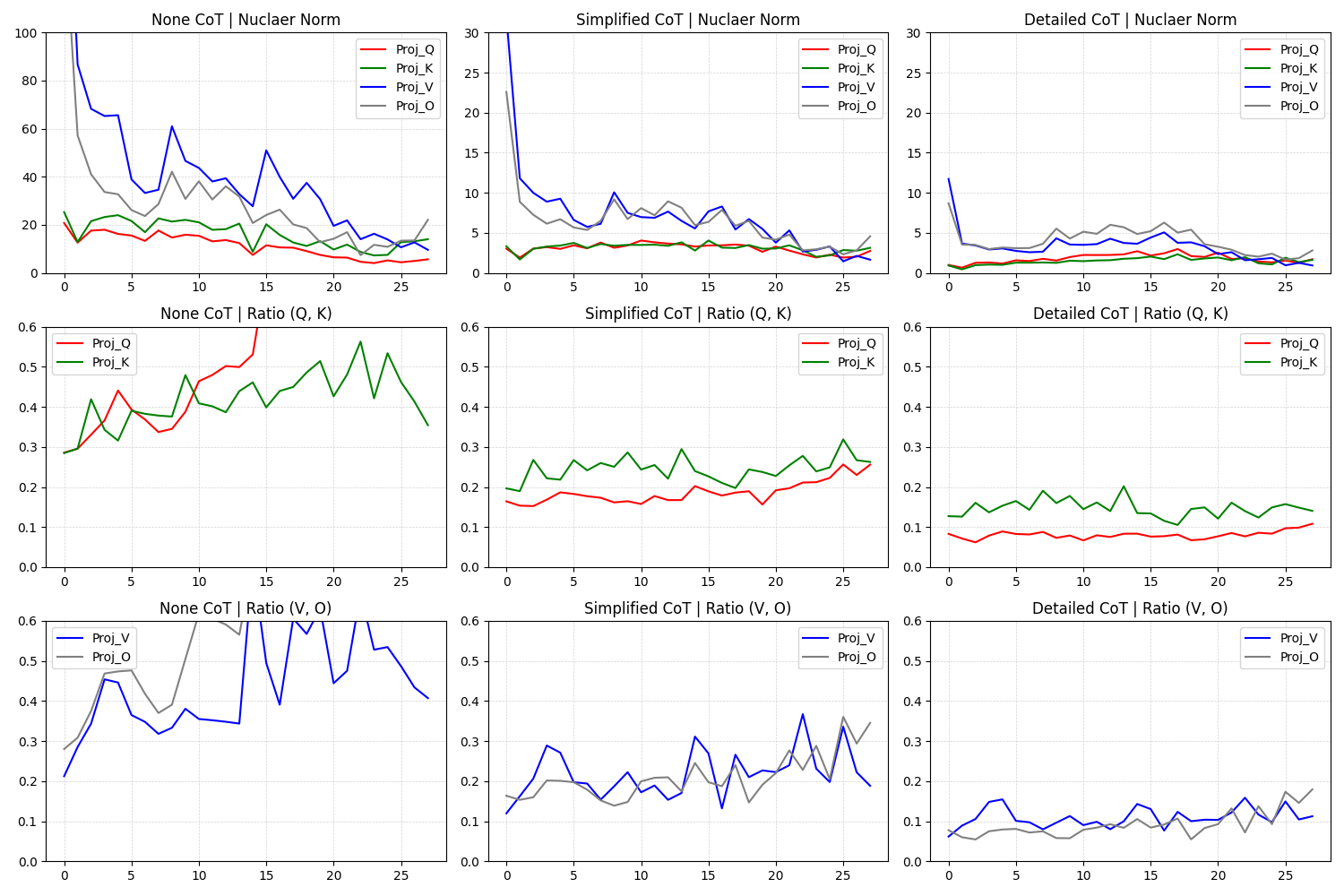}
\caption{Visualization for StrategyQA using Qwen2-1.5B on correct responses.}
\label{fig:reasoning_correct_strategyqa}
\vspace{-2mm}
\end{figure*}

\begin{table}[!tbh]
\centering
\small
\scalebox{0.75}{
\begin{tabular}{l|l|l|cccc}
\toprule
\multirow{2}{*}{\textbf{Dataset}} & \multirow{2}{*}{\textbf{Curve}} & \multirow{2}{*}{\textbf{Cot}} &  \multicolumn{4}{c}{\textbf{Mean Absolute Difference (MAD)}} \\
 & &  & Early & Middle & Last & All\\
\midrule
\multirow{24}{*}{ECQA} & \multirow{3}{*}{$s_Q$} 
  & None & 8.00 & 7.01 & 5.01 & 6.53 \\
& & Simplified & 1.11 & 0.70 & 0.86 & 0.85 \\
& & Detailed & 0.30 & 0.37 & 0.26 & 0.35 \\
\cmidrule{2-7}
& \multirow{3}{*}{$s_K$} 
  & None & 11.51 & 11.07 & 13.32 & 11.11 \\
& & Simplified & 1.34 & 1.24 & 1.01 & 1.13 \\
& & Detailed & 0.26 & 0.29 & 0.54 & 0.34 \\
\cmidrule{2-7}
& \multirow{3}{*}{$s_V$} 
  & None & 59.33 & 24.83 & 7.46 & 27.40 \\
& & Simplified & 8.53 & 3.55 & 1.66 & 4.01 \\
& & Detailed & 1.56 & 0.74 & 0.48 & 0.82 \\
\cmidrule{2-7}
& \multirow{3}{*}{$s_O$} 
  & None & 39.20 & 19.50 & 5.12 & 19.38 \\
& & Simplified & 5.56 & 3.33 & 1.41 & 3.22 \\
& & Detailed & 1.00 & 0.97 & 0.52 & 0.85 \\
\cmidrule{2-7}
 & \multirow{3}{*}{$r_Q$} 
  & None & 0.02 & 0.07 & 0.14 & 0.08 \\
& & Simplified & 0.02 & 0.02 & 0.05 & 0.03 \\
& & Detailed & 0.01 & 0.01 & 0.01 & 0.01 \\
\cmidrule{2-7}
& \multirow{3}{*}{$r_K$} 
  & None & 0.06 & 0.04 & 0.06 & 0.05 \\
& & Simplified & 0.03 & 0.04 & 0.04 & 0.04 \\
& & Detailed & 0.02 & 0.03 & 0.02 & 0.02 \\
\cmidrule{2-7}
 & \multirow{3}{*}{$r_V$} 
  & None & 0.04 & 0.07 & 0.05 & 0.05 \\
& & Simplified & 0.03 & 0.05 & 0.05 & 0.04 \\
& & Detailed & 0.02 & 0.02 & 0.03 & 0.02 \\
\cmidrule{2-7}
& \multirow{3}{*}{$r_O$} 
  & None & 0.02 & 0.05 & 0.11 & 0.06 \\
& & Simplified & 0.02 & 0.03 & 0.07 & 0.04 \\
& & Detailed & 0.01 & 0.02 & 0.05 & 0.02 \\
\bottomrule
\end{tabular}
}
\caption{
Statistical results for ECQA using Qwen2-1.5B on correct responses.
}
\label{tbl:reasoning_cot_500_qwen2_1_5b_grads_ecqa_train}
\end{table}

\begin{figure*}[!h]
\centering
\includegraphics[width=\linewidth]{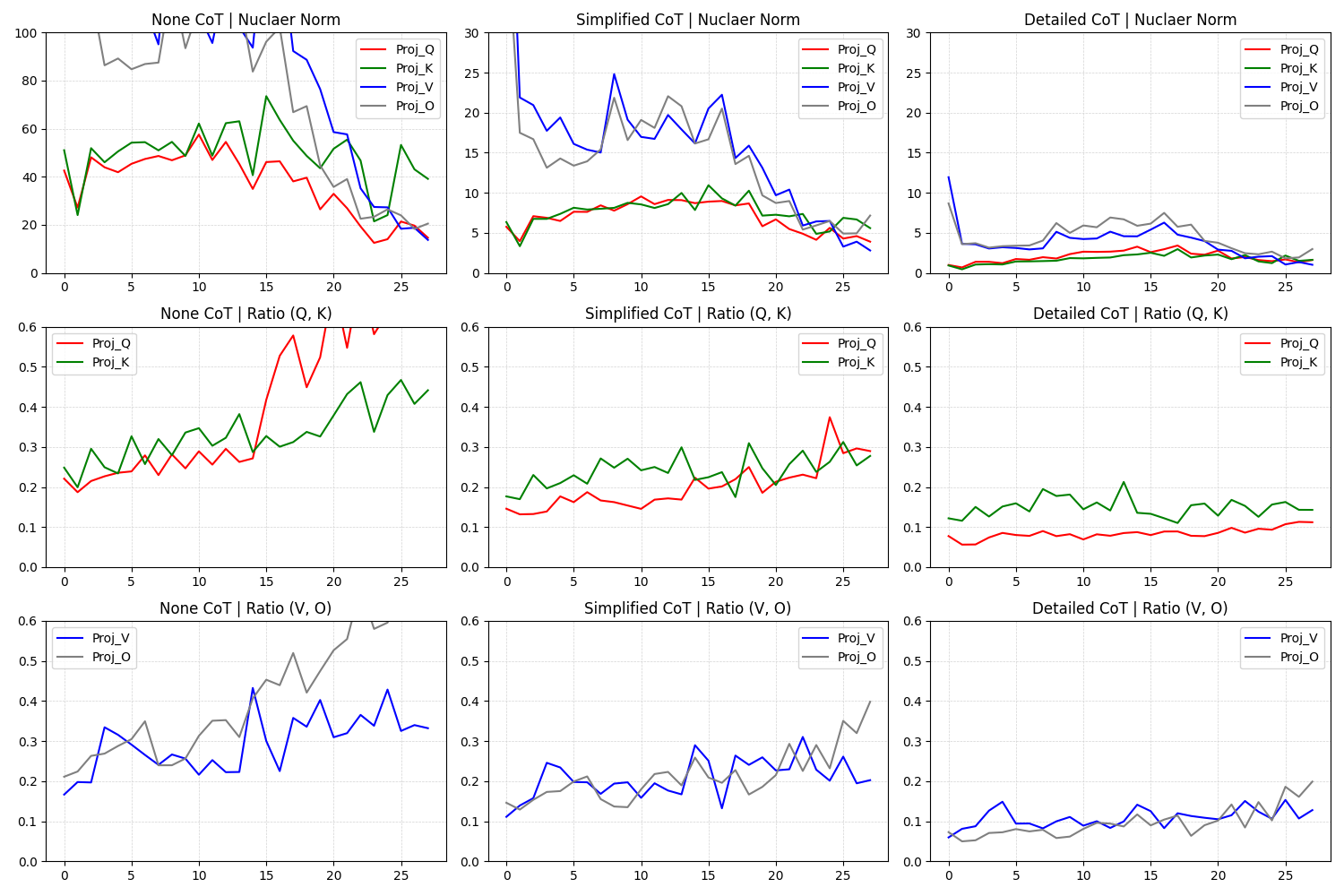}
\caption{Visualization for ECQA using Qwen2-1.5B on correct responses.}
\label{fig:reasoning_correct_ecqa}
\vspace{-2mm}
\end{figure*}

\begin{table}[!tbh]
\centering
\small
\scalebox{0.75}{
\begin{tabular}{l|l|l|cccc}
\toprule
\multirow{2}{*}{\textbf{Dataset}} & \multirow{2}{*}{\textbf{Curve}} & \multirow{2}{*}{\textbf{Cot}} &  \multicolumn{4}{c}{\textbf{Mean Absolute Difference (MAD)}} \\
 & &  & Early & Middle & Last & All\\
\midrule
\multirow{24}{*}{CREAK} & \multirow{3}{*}{$s_Q$} 
  & None & 13.91 & 9.73 & 8.08 & 9.73 \\
& & Simplified & 1.77 & 1.26 & 2.74 & 1.78 \\
& & Detailed & 0.31 & 0.33 & 0.25 & 0.33 \\
\cmidrule{2-7}
& \multirow{3}{*}{$s_K$} 
  & None & 20.31 & 15.99 & 17.98 & 16.92 \\
& & Simplified & 2.37 & 1.89 & 2.54 & 2.08 \\
& & Detailed & 0.28 & 0.26 & 0.51 & 0.32 \\
\cmidrule{2-7}
& \multirow{3}{*}{$s_V$} 
  & None & 111.18 & 40.86 & 12.03 & 47.98 \\
& & Simplified & 14.29 & 6.89 & 3.59 & 7.41 \\
& & Detailed & 1.80 & 0.81 & 0.54 & 0.92 \\
\cmidrule{2-7}
& \multirow{3}{*}{$s_O$} 
  & None & 71.70 & 36.08 & 4.27 & 35.01 \\
& & Simplified & 9.03 & 6.26 & 2.93 & 5.88 \\
& & Detailed & 1.14 & 1.07 & 0.55 & 0.93 \\
\cmidrule{2-7}
 & \multirow{3}{*}{$r_Q$} 
  & None & 0.02 & 0.06 & 0.12 & 0.07 \\
& & Simplified & 0.02 & 0.03 & 0.10 & 0.04 \\
& & Detailed & 0.01 & 0.01 & 0.01 & 0.01 \\
\cmidrule{2-7}
& \multirow{3}{*}{$r_K$} 
  & None & 0.04 & 0.05 & 0.05 & 0.05 \\
& & Simplified & 0.04 & 0.04 & 0.03 & 0.04 \\
& & Detailed & 0.02 & 0.03 & 0.01 & 0.03 \\
\cmidrule{2-7}
 & \multirow{3}{*}{$r_V$} 
  & None & 0.04 & 0.07 & 0.05 & 0.06 \\
& & Simplified & 0.03 & 0.06 & 0.09 & 0.06 \\
& & Detailed & 0.02 & 0.03 & 0.04 & 0.03 \\
\cmidrule{2-7}
& \multirow{3}{*}{$r_O$} 
  & None & 0.02 & 0.06 & 0.08 & 0.05 \\
& & Simplified & 0.02 & 0.03 & 0.08 & 0.04 \\
& & Detailed & 0.01 & 0.02 & 0.05 & 0.02 \\
\bottomrule
\end{tabular}
}
\caption{
Statistical results for CREAK using Qwen2-1.5B on correct responses.
}
\label{tbl:reasoning_cot_500_qwen2_1_5b_grads_creak_train}
\end{table}

\begin{figure*}[!h]
\centering
\includegraphics[width=\linewidth]{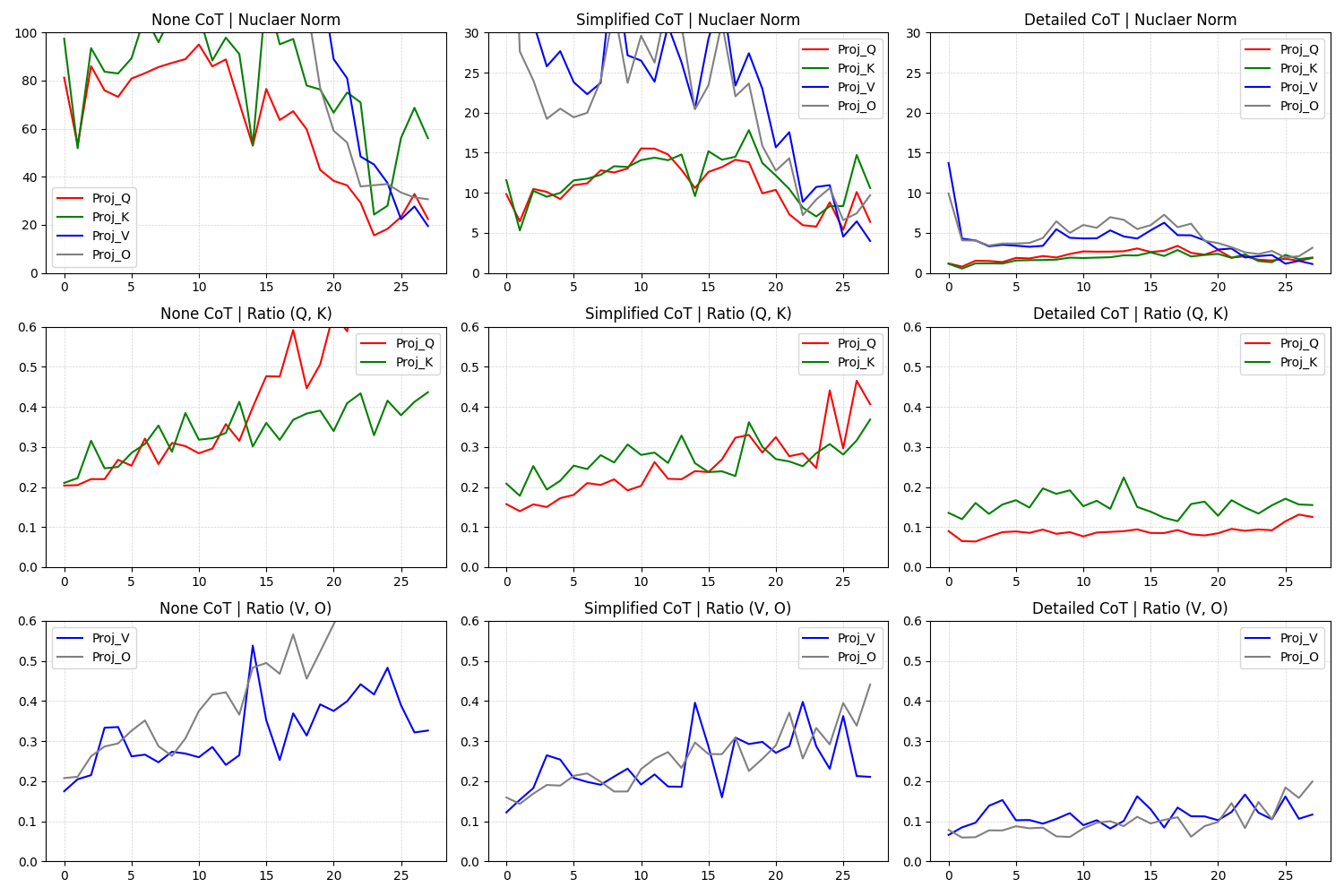}
\caption{Visualization for CREAK using Qwen2-1.5B on correct responses.}
\label{fig:reasoning_correct_creak}
\vspace{-2mm}
\end{figure*}

\begin{table}[!tbh]
\centering
\small
\scalebox{0.75}{
\begin{tabular}{l|l|l|cccc}
\toprule
\multirow{2}{*}{\textbf{Dataset}} & \multirow{2}{*}{\textbf{Curve}} & \multirow{2}{*}{\textbf{Cot}} &  \multicolumn{4}{c}{\textbf{Mean Absolute Difference (MAD)}} \\
 & &  & Early & Middle & Last & All\\
\midrule
\multirow{24}{*}{Sensemaking} & \multirow{3}{*}{$s_Q$} 
  & None & 4.76 & 3.42 & 5.92 & 4.28 \\
& & Simplified & 2.22 & 1.50 & 2.73 & 1.98 \\
& & Detailed & 0.31 & 0.35 & 0.38 & 0.37 \\
\cmidrule{2-7}
& \multirow{3}{*}{$s_K$} 
  & None & 6.84 & 4.71 & 11.16 & 6.59 \\
& & Simplified & 2.90 & 2.06 & 2.50 & 2.27 \\
& & Detailed & 0.26 & 0.35 & 0.71 & 0.41 \\
\cmidrule{2-7}
& \multirow{3}{*}{$s_V$} 
  & None & 38.73 & 13.42 & 5.15 & 16.68 \\
& & Simplified & 15.87 & 7.58 & 3.19 & 7.95 \\
& & Detailed & 1.52 & 0.75 & 0.40 & 0.80 \\
\cmidrule{2-7}
& \multirow{3}{*}{$s_O$} 
  & None & 25.32 & 12.11 & 2.65 & 12.44 \\
& & Simplified & 10.40 & 6.41 & 2.45 & 6.16 \\
& & Detailed & 1.06 & 0.88 & 0.43 & 0.80 \\
\cmidrule{2-7}
 & \multirow{3}{*}{$r_Q$} 
  & None & 0.04 & 0.06 & 0.20 & 0.09 \\
& & Simplified & 0.02 & 0.02 & 0.10 & 0.04 \\
& & Detailed & 0.01 & 0.01 & 0.02 & 0.01 \\
\cmidrule{2-7}
& \multirow{3}{*}{$r_K$} 
  & None & 0.03 & 0.07 & 0.12 & 0.07 \\
& & Simplified & 0.03 & 0.05 & 0.02 & 0.04 \\
& & Detailed & 0.02 & 0.04 & 0.03 & 0.03 \\
\cmidrule{2-7}
 & \multirow{3}{*}{$r_V$} 
  & None & 0.04 & 0.06 & 0.07 & 0.05 \\
& & Simplified & 0.04 & 0.05 & 0.08 & 0.05 \\
& & Detailed & 0.03 & 0.02 & 0.03 & 0.02 \\
\cmidrule{2-7}
& \multirow{3}{*}{$r_O$} 
  & None & 0.03 & 0.05 & 0.11 & 0.06 \\
& & Simplified & 0.01 & 0.03 & 0.08 & 0.04 \\
& & Detailed & 0.01 & 0.02 & 0.05 & 0.02 \\
\bottomrule
\end{tabular}
}
\caption{
Statistical results for Sensemaking using Qwen2-1.5B on correct responses.
}
\label{tbl:reasoning_cot_500_qwen2_1_5b_grads_sensemaking_train}
\end{table}

\begin{figure*}[!h]
\centering
\includegraphics[width=\linewidth]{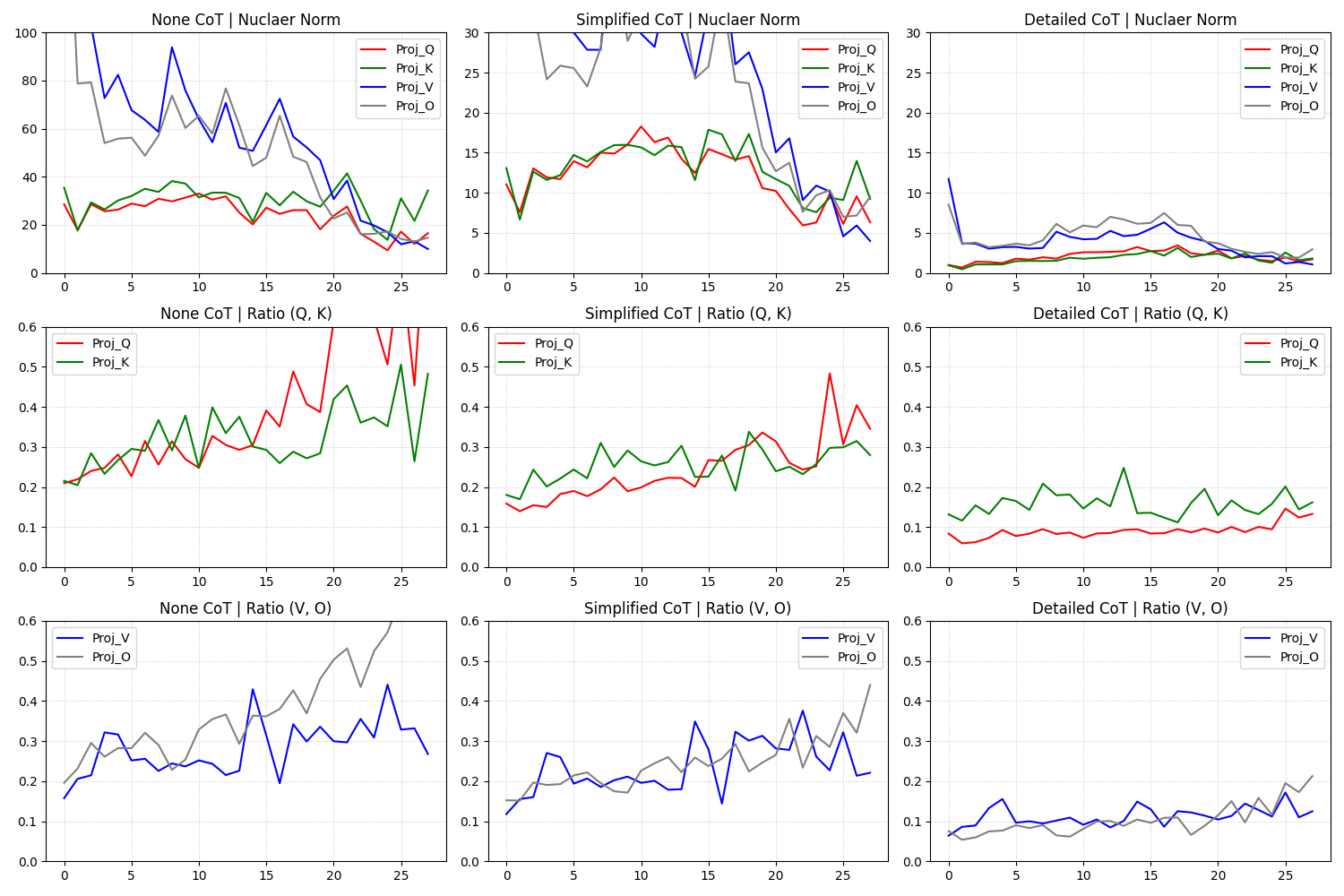}
\caption{Visualization for Sensemaking using Qwen2-1.5B on correct responses.}
\label{fig:reasoning_correct_sensemaking}
\vspace{-2mm}
\end{figure*}

\subsubsection{Wiki Tasks}

The visualizations and statistical results on Wiki tasks are shown in Figure \ref{fig:wiki_correct} and Table \ref{tbl:wiki_wiki_popularity_qwen2_1_5b_grads}.

\begin{table}[!tbh]
\centering
\small
\scalebox{0.75}{
\begin{tabular}{l|l|l|cccc}
\toprule
\multirow{2}{*}{\textbf{Dataset}} & \multirow{2}{*}{\textbf{Curve}} & \multirow{2}{*}{\textbf{Cot}} &  \multicolumn{4}{c}{\textbf{Mean Absolute Difference (MAD)}} \\
 & &  & Early & Middle & Last & All\\
\midrule
\multirow{32}{*}{Wiki} & \multirow{3}{*}{$s_Q$} 
  & Len 100 & 0.40 & 0.28 & 0.30 & 0.32 \\
& & Len 500 & 0.31 & 0.26 & 0.21 & 0.27 \\
& & Len 1000 & 0.28 & 0.25 & 0.18 & 0.25 \\
& & Unpopular & 0.63 & 0.44 & 0.19 & 0.43 \\
\cmidrule{2-7}
& \multirow{3}{*}{$s_K$} 
  & Len 100 & 0.47 & 0.20 & 0.33 & 0.31 \\
& & Len 500 & 0.29 & 0.11 & 0.29 & 0.19 \\
& & Len 1000 & 0.23 & 0.10 & 0.31 & 0.17 \\
& & Unpopular & 0.76 & 0.48 & 0.41 & 0.52 \\
\cmidrule{2-7}
& \multirow{3}{*}{$s_V$} 
  & Len 100 & 3.26 & 1.30 & 0.89 & 1.63 \\
& & Len 500 & 2.10 & 0.87 & 0.67 & 1.08 \\
& & Len 1000 & 1.67 & 0.65 & 0.52 & 0.84 \\
& & Unpopular & 5.49 & 1.90 & 0.99 & 2.41 \\
\cmidrule{2-7}
& \multirow{3}{*}{$s_O$} 
  & Len 100 & 2.21 & 1.28 & 0.73 & 1.34 \\
& & Len 500 & 1.48 & 1.06 & 0.61 & 1.01 \\
& & Len 1000 & 1.24 & 0.93 & 0.52 & 0.87 \\
& & Unpopular & 3.58 & 1.74 & 0.73 & 1.86 \\
\cmidrule{2-7}
 & \multirow{3}{*}{$r_Q$} 
  & Len 100 & 0.01 & 0.01 & 0.03 & 0.02 \\
& & Len 500 & 0.01 & 0.01 & 0.01 & 0.01 \\
& & Len 1000 & 0.01 & 0.01 & 0.01 & 0.01 \\
& & Unpopular & 0.03 & 0.01 & 0.06 & 0.03 \\
\cmidrule{2-7}
& \multirow{3}{*}{$r_K$} 
  & Len 100 & 0.03 & 0.03 & 0.03 & 0.03 \\
& & Len 500 & 0.02 & 0.02 & 0.02 & 0.02 \\
& & Len 1000 & 0.02 & 0.02 & 0.01 & 0.02 \\
& & Unpopular & 0.04 & 0.02 & 0.03 & 0.03 \\
\cmidrule{2-7}
 & \multirow{3}{*}{$r_V$} 
  & Len 100 & 0.03 & 0.05 & 0.09 & 0.06 \\
& & Len 500 & 0.02 & 0.03 & 0.06 & 0.03 \\
& & Len 1000 & 0.02 & 0.03 & 0.04 & 0.03 \\
& & Unpopular & 0.04 & 0.08 & 0.07 & 0.06 \\
\cmidrule{2-7}
& \multirow{3}{*}{$r_O$} 
  & Len 100 & 0.02 & 0.03 & 0.07 & 0.04 \\
& & Len 500 & 0.01 & 0.02 & 0.05 & 0.03 \\
& & Len 1000 & 0.01 & 0.02 & 0.05 & 0.02 \\
& & Unpopular & 0.02 & 0.05 & 0.08 & 0.05 \\
\bottomrule
\end{tabular}
}
\caption{
Statistical results for Wiki using Qwen2-1.5B on correct responses.
}
\label{tbl:wiki_wiki_popularity_qwen2_1_5b_grads}
\end{table}

\begin{figure*}[!h]
\centering
\includegraphics[width=\linewidth]{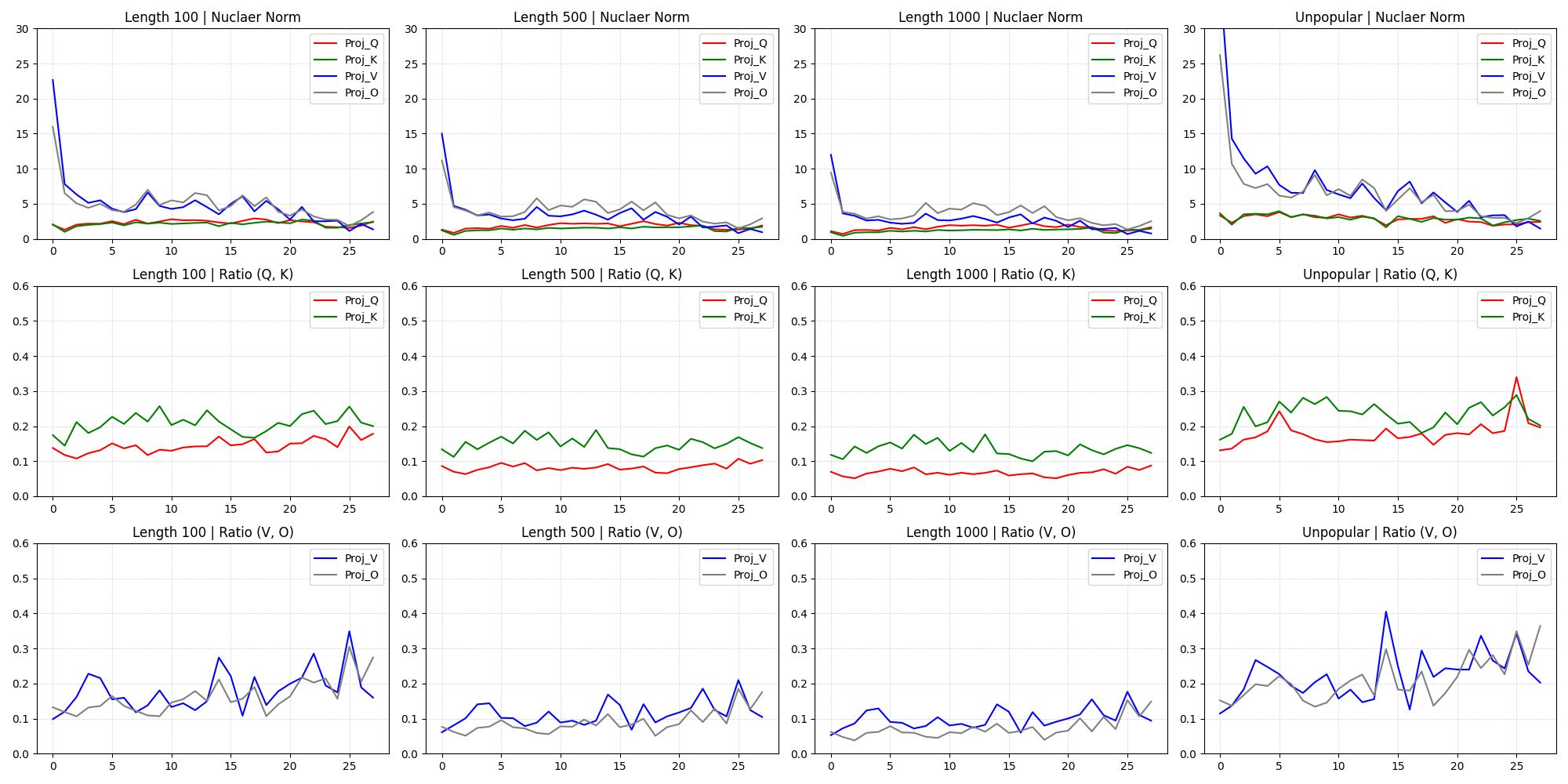}
\caption{Visualization for Wiki tasks using Qwen2-1.5B on correct responses.}
\label{fig:wiki_correct}
\vspace{-2mm}
\end{figure*}

\subsection{Pre-trained LLM on Irrelevant Responses}

\subsubsection{Reasoning Tasks}

The visualizations and statistical results on MATH tasks: MATH-Algebra (Figure \ref{fig:math_nonsense_algebra}, Table \ref{tbl:math_lighteval_math_wrong_answer_shuffle_qwen2_1_5b_grads_algebra}), MATH-Counting (Figure \ref{fig:math_nonsense_counting}, Table \ref{tbl:math_lighteval_math_wrong_answer_shuffle_qwen2_1_5b_grads_counting_probability}), MATH-Geometry (Figure \ref{fig:math_nonsense_geometry}, Table \ref{tbl:math_lighteval_math_wrong_answer_shuffle_qwen2_1_5b_grads_geometry}).

\begin{table}[!tbh]
\centering
\small
\scalebox{0.75}{
\begin{tabular}{l|l|l|cccc}
\toprule
\multirow{2}{*}{\textbf{Dataset}} & \multirow{2}{*}{\textbf{Curve}} & \multirow{2}{*}{\textbf{Cot}} &  \multicolumn{4}{c}{\textbf{Mean Absolute Difference (MAD)}} \\
 & &  & Early & Middle & Last & All\\
\midrule
\multirow{16}{*}{Algebra} & \multirow{3}{*}{$s_Q$} 
  & Simplified & 0.65 & 0.67 & 1.16 & 0.80 \\
& & Detailed & 0.39 & 0.39 & 0.48 & 0.42 \\
\cmidrule{2-7}
& \multirow{3}{*}{$s_K$} 
  & Simplified & 0.63 & 0.53 & 1.16 & 0.67 \\
& & Detailed & 0.36 & 0.25 & 0.54 & 0.34 \\
\cmidrule{2-7}
& \multirow{3}{*}{$s_V$} 
  & Simplified & 3.16 & 2.22 & 0.94 & 2.02 \\
& & Detailed & 1.66 & 0.98 & 0.49 & 0.96 \\
\cmidrule{2-7}
& \multirow{3}{*}{$s_O$} 
  & Simplified & 2.26 & 2.45 & 0.68 & 1.87 \\
& & Detailed & 1.24 & 1.21 & 0.37 & 0.96 \\
\cmidrule{2-7}
 & \multirow{3}{*}{$r_Q$} 
  & Simplified & 0.01 & 0.02 & 0.05 & 0.02 \\
& & Detailed & 0.01 & 0.02 & 0.03 & 0.02 \\
\cmidrule{2-7}
& \multirow{3}{*}{$r_K$} 
  & Simplified & 0.01 & 0.03 & 0.02 & 0.03 \\
& & Detailed & 0.02 & 0.03 & 0.03 & 0.03 \\
\cmidrule{2-7}
 & \multirow{3}{*}{$r_V$} 
  & Simplified & 0.03 & 0.04 & 0.04 & 0.03 \\
& & Detailed & 0.03 & 0.03 & 0.04 & 0.03 \\
\cmidrule{2-7}
& \multirow{3}{*}{$r_O$} 
  & Simplified & 0.01 & 0.02 & 0.06 & 0.03 \\
& & Detailed & 0.01 & 0.01 & 0.06 & 0.02 \\
\bottomrule
\end{tabular}
}
\caption{
Statistical results for MATH-Algebra using Qwen2-1.5B on irrelevant responses.
}
\label{tbl:math_lighteval_math_wrong_answer_shuffle_qwen2_1_5b_grads_algebra}
\end{table}

\begin{figure}[!h]
\centering
\includegraphics[width=\linewidth]{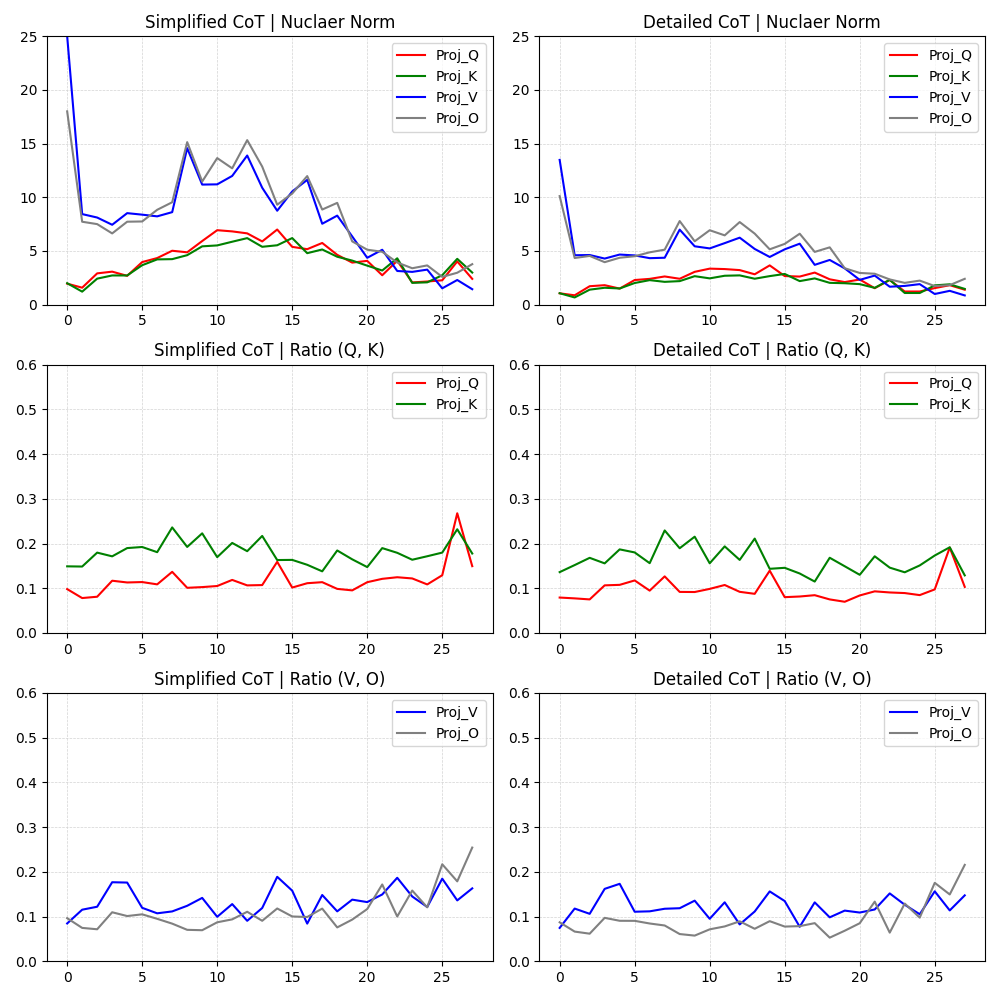}
\caption{Visualization for MATH-Algebra using Qwen2-1.5B on irrelevant responses.}
\label{fig:math_nonsense_algebra}
\vspace{-2mm}
\end{figure}

\begin{table}[!tbh]
\centering
\small
\scalebox{0.75}{
\begin{tabular}{l|l|l|cccc}
\toprule
\multirow{2}{*}{\textbf{Dataset}} & \multirow{2}{*}{\textbf{Curve}} & \multirow{2}{*}{\textbf{Cot}} &  \multicolumn{4}{c}{\textbf{Mean Absolute Difference (MAD)}} \\
 & &  & Early & Middle & Last & All\\
\midrule
\multirow{16}{*}{Counting} & \multirow{3}{*}{$s_Q$} 
  & Simplified & 0.67 & 0.66 & 0.89 & 0.73 \\
& & Detailed & 0.46 & 0.41 & 0.38 & 0.42 \\
\cmidrule{2-7}
& \multirow{3}{*}{$s_K$} 
  & Simplified & 0.64 & 0.42 & 0.92 & 0.56 \\
& & Detailed & 0.41 & 0.25 & 0.50 & 0.34 \\
\cmidrule{2-7}
& \multirow{3}{*}{$s_V$} 
  & Simplified & 3.20 & 2.08 & 0.98 & 1.98 \\
& & Detailed & 1.94 & 0.93 & 0.52 & 1.02 \\
\cmidrule{2-7}
& \multirow{3}{*}{$s_O$} 
  & Simplified & 2.36 & 2.25 & 0.77 & 1.81 \\
& & Detailed & 1.43 & 1.21 & 0.45 & 1.01 \\
\cmidrule{2-7}
 & \multirow{3}{*}{$r_Q$} 
  & Simplified & 0.01 & 0.02 & 0.04 & 0.02 \\
& & Detailed & 0.01 & 0.02 & 0.03 & 0.02 \\
\cmidrule{2-7}
& \multirow{3}{*}{$r_K$} 
  & Simplified & 0.01 & 0.03 & 0.02 & 0.02 \\
& & Detailed & 0.02 & 0.03 & 0.02 & 0.03 \\
\cmidrule{2-7}
 & \multirow{3}{*}{$r_V$} 
  & Simplified & 0.03 & 0.04 & 0.04 & 0.04 \\
& & Detailed & 0.03 & 0.03 & 0.03 & 0.03 \\
\cmidrule{2-7}
& \multirow{3}{*}{$r_O$} 
  & Simplified & 0.02 & 0.02 & 0.06 & 0.03 \\
& & Detailed & 0.01 & 0.01 & 0.05 & 0.02 \\
\bottomrule
\end{tabular}
}
\caption{
Statistical results for MATH-Counting using Qwen2-1.5B on irrelevant responses.
}
\label{tbl:math_lighteval_math_wrong_answer_shuffle_qwen2_1_5b_grads_counting_probability}
\end{table}

\begin{figure}[!h]
\centering
\includegraphics[width=\linewidth]{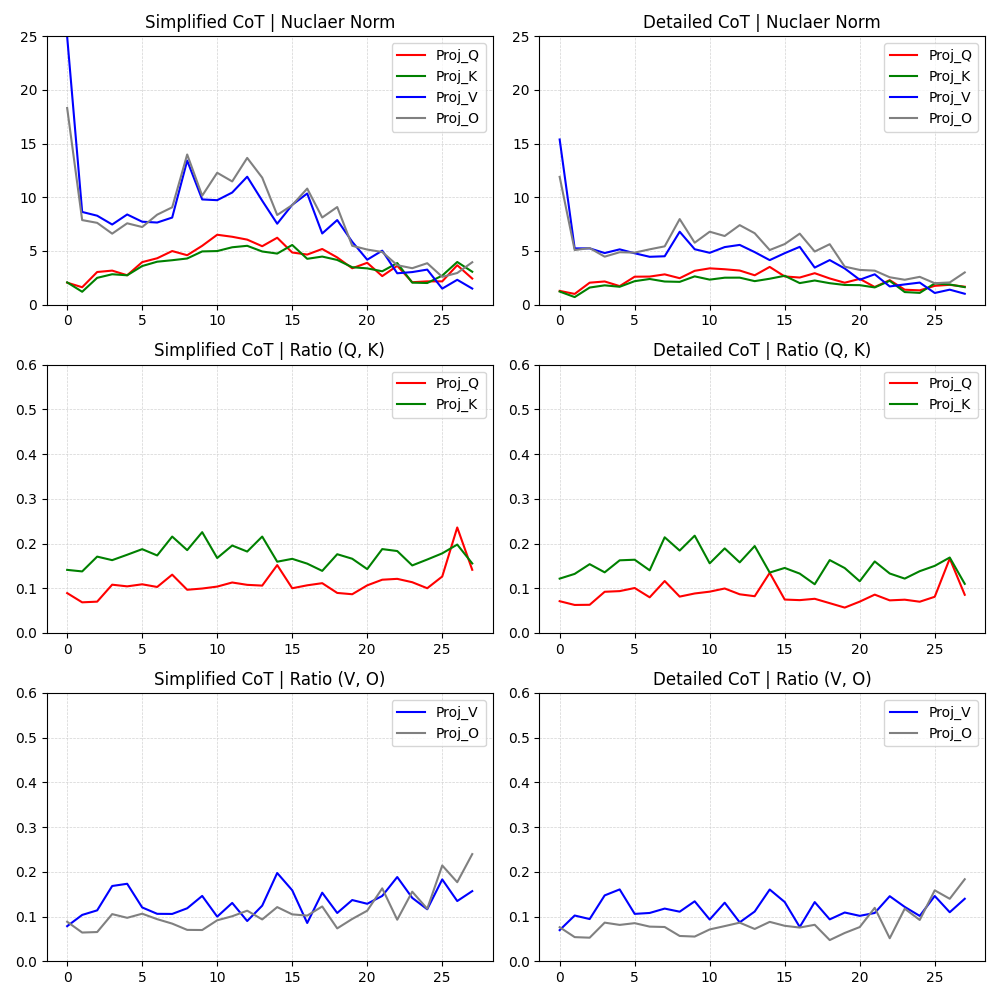}
\caption{Visualization for MATH-Counting using Qwen2-1.5B on irrelevant responses.}
\label{fig:math_nonsense_counting}
\vspace{-2mm}
\end{figure}

\begin{table}[!tbh]
\centering
\small
\scalebox{0.75}{
\begin{tabular}{l|l|l|cccc}
\toprule
\multirow{2}{*}{\textbf{Dataset}} & \multirow{2}{*}{\textbf{Curve}} & \multirow{2}{*}{\textbf{Cot}} &  \multicolumn{4}{c}{\textbf{Mean Absolute Difference (MAD)}} \\
 & &  & Early & Middle & Last & All\\
\midrule
\multirow{16}{*}{Geometry} & \multirow{3}{*}{$s_Q$} 
  & Simplified & 0.64 & 0.64 & 1.28 & 0.80 \\
& & Detailed & 0.54 & 0.32 & 0.57 & 0.43 \\
\cmidrule{2-7}
& \multirow{3}{*}{$s_K$} 
  & Simplified & 0.58 & 0.45 & 1.32 & 0.65 \\
& & Detailed & 0.46 & 0.18 & 0.70 & 0.37 \\
\cmidrule{2-7}
& \multirow{3}{*}{$s_V$} 
  & Simplified & 3.22 & 1.91 & 0.97 & 1.88 \\
& & Detailed & 2.43 & 1.00 & 0.54 & 1.18 \\
\cmidrule{2-7}
& \multirow{3}{*}{$s_O$} 
  & Simplified & 2.53 & 2.40 & 0.65 & 1.87 \\
& & Detailed & 1.80 & 1.40 & 0.47 & 1.19 \\
\cmidrule{2-7}
 & \multirow{3}{*}{$r_Q$} 
  & Simplified & 0.01 & 0.01 & 0.04 & 0.02 \\
& & Detailed & 0.01 & 0.02 & 0.02 & 0.02 \\
\cmidrule{2-7}
& \multirow{3}{*}{$r_K$} 
  & Simplified & 0.01 & 0.03 & 0.02 & 0.02 \\
& & Detailed & 0.02 & 0.03 & 0.03 & 0.03 \\
\cmidrule{2-7}
 & \multirow{3}{*}{$r_V$} 
  & Simplified & 0.02 & 0.03 & 0.03 & 0.03 \\
& & Detailed & 0.03 & 0.03 & 0.04 & 0.03 \\
\cmidrule{2-7}
& \multirow{3}{*}{$r_O$} 
  & Simplified & 0.01 & 0.01 & 0.05 & 0.02 \\
& & Detailed & 0.01 & 0.01 & 0.04 & 0.02 \\
\bottomrule
\end{tabular}
}
\caption{
Statistical results for MATH-Geometry using Qwen2-1.5B on irrelevant responses.
}
\label{tbl:math_lighteval_math_wrong_answer_shuffle_qwen2_1_5b_grads_geometry}
\end{table}

\begin{figure}[!h]
\centering
\includegraphics[width=\linewidth]{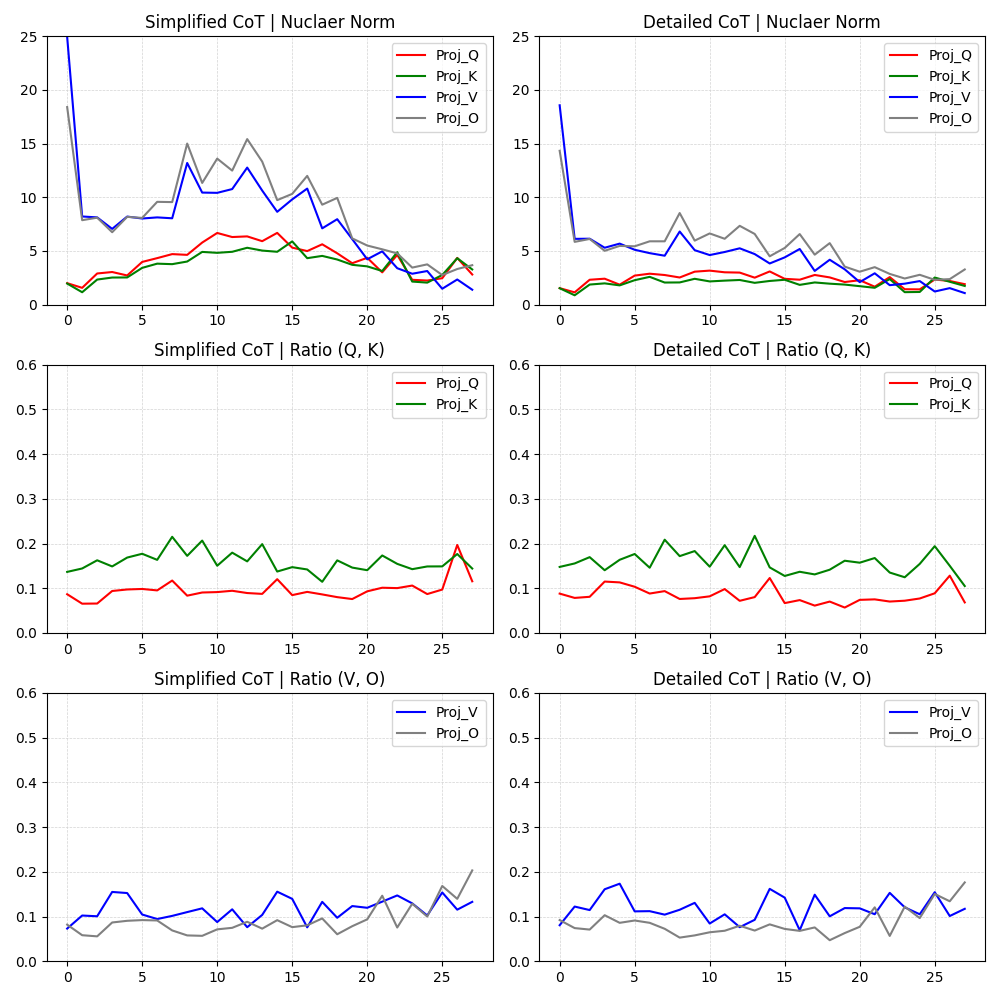}
\caption{Visualization for MATH-Geometry using Qwen2-1.5B on irrelevant responses.}
\label{fig:math_nonsense_geometry}
\vspace{-2mm}
\end{figure}

The visualizations and statistical results on other reasoning tasks: AQuA (Figure \ref{fig:reasoning_nonsense_aqua}, Table \ref{tbl:reasoning_cot_500_wrong_answer_shuffle_qwen2_1_5b_grads_aqua_train}), GSM8K (Figure \ref{fig:reasoning_nonsense_gsm8k}, Table \ref{tbl:reasoning_cot_500_wrong_answer_shuffle_qwen2_1_5b_grads_gsm8k_train}), StrategyQA (Figure \ref{fig:reasoning_nonsense_strategyqa}, Table \ref{tbl:reasoning_cot_500_wrong_answer_shuffle_qwen2_1_5b_grads_strategyqa_train}), ECQA (Figure \ref{fig:reasoning_nonsense_ecqa}, Table \ref{tbl:reasoning_cot_500_wrong_answer_shuffle_qwen2_1_5b_grads_ecqa_train}), CREAK (Figure \ref{fig:reasoning_nonsense_creak}, Table \ref{tbl:reasoning_cot_500_wrong_answer_shuffle_qwen2_1_5b_grads_creak_train}), Sensemaking (Figure \ref{fig:reasoning_nonsense_sensemaking}, Table \ref{tbl:reasoning_cot_500_wrong_answer_shuffle_qwen2_1_5b_grads_sensemaking_train}).

\begin{table}[!tbh]
\centering
\small
\scalebox{0.75}{
\begin{tabular}{l|l|l|cccc}
\toprule
\multirow{2}{*}{\textbf{Dataset}} & \multirow{2}{*}{\textbf{Curve}} & \multirow{2}{*}{\textbf{Cot}} &  \multicolumn{4}{c}{\textbf{Mean Absolute Difference (MAD)}} \\
 & &  & Early & Middle & Last & All\\
\midrule
\multirow{24}{*}{AQuA} & \multirow{3}{*}{$s_Q$} 
  & None & 5.79 & 4.22 & 3.47 & 4.46 \\
& & Simplified & 1.09 & 0.67 & 1.33 & 0.95 \\
& & Detailed & 0.53 & 0.40 & 0.43 & 0.45 \\
\cmidrule{2-7}
& \multirow{3}{*}{$s_K$} 
  & None & 7.28 & 6.31 & 8.41 & 7.10 \\
& & Simplified & 1.12 & 0.69 & 1.58 & 0.97 \\
& & Detailed & 0.50 & 0.28 & 0.55 & 0.39 \\
\cmidrule{2-7}
& \multirow{3}{*}{$s_V$} 
  & None & 37.62 & 16.30 & 4.00 & 17.49 \\
& & Simplified & 6.56 & 3.14 & 1.47 & 3.36 \\
& & Detailed & 2.53 & 1.26 & 0.73 & 1.35 \\
\cmidrule{2-7}
& \multirow{3}{*}{$s_O$} 
  & None & 23.99 & 14.51 & 3.06 & 13.04 \\
& & Simplified & 4.42 & 3.22 & 1.07 & 2.83 \\
& & Detailed & 1.81 & 1.48 & 0.54 & 1.25 \\
\cmidrule{2-7}
 & \multirow{3}{*}{$r_Q$} 
  & None & 0.03 & 0.06 & 0.21 & 0.09 \\
& & Simplified & 0.02 & 0.02 & 0.03 & 0.02 \\
& & Detailed & 0.01 & 0.01 & 0.03 & 0.02 \\
\cmidrule{2-7}
& \multirow{3}{*}{$r_K$} 
  & None & 0.04 & 0.04 & 0.14 & 0.06 \\
& & Simplified & 0.03 & 0.03 & 0.02 & 0.03 \\
& & Detailed & 0.02 & 0.03 & 0.02 & 0.03 \\
\cmidrule{2-7}
 & \multirow{3}{*}{$r_V$} 
  & None & 0.04 & 0.06 & 0.03 & 0.05 \\
& & Simplified & 0.03 & 0.05 & 0.05 & 0.04 \\
& & Detailed & 0.03 & 0.04 & 0.04 & 0.03 \\
\cmidrule{2-7}
& \multirow{3}{*}{$r_O$} 
  & None & 0.02 & 0.04 & 0.09 & 0.05 \\
& & Simplified & 0.02 & 0.02 & 0.08 & 0.04 \\
& & Detailed & 0.01 & 0.01 & 0.06 & 0.02 \\
\bottomrule
\end{tabular}
}
\caption{
Statistical results for AQuA using Qwen2-1.5B on irrelevant responses.
}
\label{tbl:reasoning_cot_500_wrong_answer_shuffle_qwen2_1_5b_grads_aqua_train}
\end{table}

\begin{figure*}[!h]
\centering
\includegraphics[width=\linewidth]{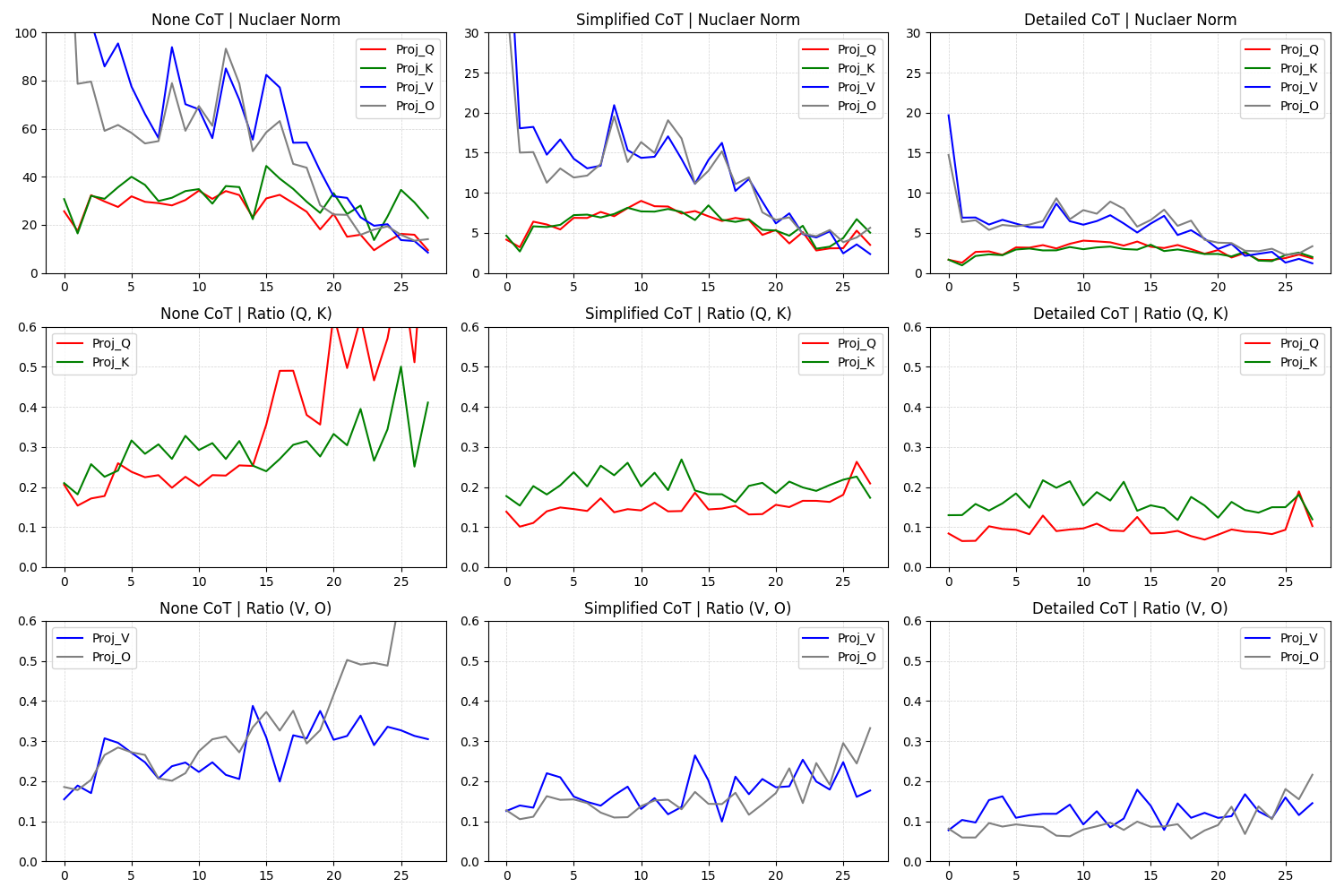}
\caption{Visualization for AQuA using Qwen2-1.5B on irrelevant responses.}
\label{fig:reasoning_nonsense_aqua}
\vspace{-2mm}
\end{figure*}

\begin{table}[!tbh]
\centering
\small
\scalebox{0.75}{
\begin{tabular}{l|l|l|cccc}
\toprule
\multirow{2}{*}{\textbf{Dataset}} & \multirow{2}{*}{\textbf{Curve}} & \multirow{2}{*}{\textbf{Cot}} &  \multicolumn{4}{c}{\textbf{Mean Absolute Difference (MAD)}} \\
 & &  & Early & Middle & Last & All\\
\midrule
\multirow{24}{*}{GSM8K} & \multirow{3}{*}{$s_Q$} 
  & None & 3.34 & 2.96 & 7.10 & 4.14 \\
& & Simplified & 0.85 & 0.62 & 1.12 & 0.85 \\
& & Detailed & 0.47 & 0.37 & 0.40 & 0.42 \\
\cmidrule{2-7}
& \multirow{3}{*}{$s_K$} 
  & None & 4.41 & 5.64 & 9.89 & 6.21 \\
& & Simplified & 0.91 & 0.84 & 1.29 & 0.94 \\
& & Detailed & 0.43 & 0.30 & 0.54 & 0.38 \\
\cmidrule{2-7}
& \multirow{3}{*}{$s_V$} 
  & None & 23.08 & 15.01 & 5.51 & 13.78 \\
& & Simplified & 4.56 & 3.05 & 1.30 & 2.87 \\
& & Detailed & 2.31 & 1.26 & 0.64 & 1.30 \\
\cmidrule{2-7}
& \multirow{3}{*}{$s_O$} 
  & None & 14.64 & 14.57 & 3.60 & 11.53 \\
& & Simplified & 3.12 & 3.09 & 0.96 & 2.47 \\
& & Detailed & 1.54 & 1.43 & 0.47 & 1.16 \\
\cmidrule{2-7}
 & \multirow{3}{*}{$r_Q$} 
  & None & 0.02 & 0.04 & 0.13 & 0.05 \\
& & Simplified & 0.02 & 0.02 & 0.04 & 0.02 \\
& & Detailed & 0.01 & 0.02 & 0.04 & 0.02 \\
\cmidrule{2-7}
& \multirow{3}{*}{$r_K$} 
  & None & 0.04 & 0.04 & 0.04 & 0.04 \\
& & Simplified & 0.03 & 0.03 & 0.02 & 0.03 \\
& & Detailed & 0.02 & 0.04 & 0.02 & 0.03 \\
\cmidrule{2-7}
 & \multirow{3}{*}{$r_V$} 
  & None & 0.03 & 0.06 & 0.04 & 0.05 \\
& & Simplified & 0.03 & 0.05 & 0.05 & 0.04 \\
& & Detailed & 0.02 & 0.04 & 0.03 & 0.03 \\
\cmidrule{2-7}
& \multirow{3}{*}{$r_O$} 
  & None & 0.03 & 0.05 & 0.09 & 0.05 \\
& & Simplified & 0.02 & 0.02 & 0.07 & 0.03 \\
& & Detailed & 0.01 & 0.01 & 0.05 & 0.02 \\
\bottomrule
\end{tabular}
}
\caption{
Statistical results for GSM8K using Qwen2-1.5B on irrelevant responses.
}
\label{tbl:reasoning_cot_500_wrong_answer_shuffle_qwen2_1_5b_grads_gsm8k_train}
\end{table}

\begin{figure*}[!h]
\centering
\includegraphics[width=\linewidth]{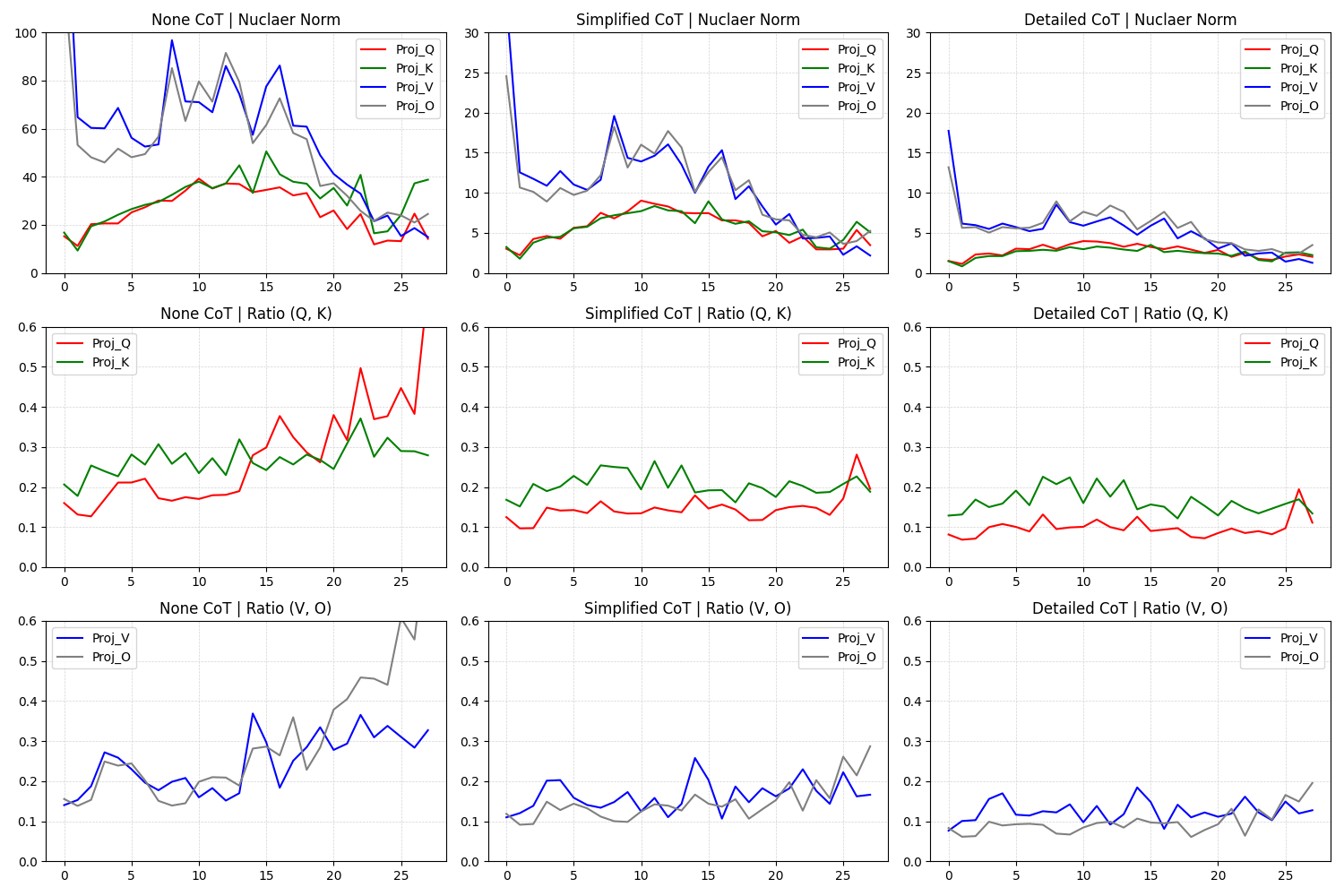}
\caption{Visualization for GSM8K using Qwen2-1.5B on irrelevant responses.}
\label{fig:reasoning_nonsense_gsm8k}
\vspace{-2mm}
\end{figure*}

\begin{table}[!tbh]
\centering
\small
\scalebox{0.75}{
\begin{tabular}{l|l|l|cccc}
\toprule
\multirow{2}{*}{\textbf{Dataset}} & \multirow{2}{*}{\textbf{Curve}} & \multirow{2}{*}{\textbf{Cot}} &  \multicolumn{4}{c}{\textbf{Mean Absolute Difference (MAD)}} \\
 & &  & Early & Middle & Last & All\\
\midrule
\multirow{24}{*}{StrategyQA} & \multirow{3}{*}{$s_Q$} 
  & None & 3.15 & 1.80 & 0.89 & 1.93 \\
& & Simplified & 0.81 & 0.49 & 0.42 & 0.56 \\
& & Detailed & 0.69 & 0.35 & 0.25 & 0.40 \\
\cmidrule{2-7}
& \multirow{3}{*}{$s_K$} 
  & None & 5.22 & 3.66 & 1.87 & 3.64 \\
& & Simplified & 1.04 & 0.52 & 0.44 & 0.66 \\
& & Detailed & 0.55 & 0.30 & 0.40 & 0.37 \\
\cmidrule{2-7}
& \multirow{3}{*}{$s_V$} 
  & None & 28.90 & 10.12 & 3.51 & 12.21 \\
& & Simplified & 6.18 & 1.80 & 1.39 & 2.67 \\
& & Detailed & 3.17 & 0.95 & 0.80 & 1.40 \\
\cmidrule{2-7}
& \multirow{3}{*}{$s_O$} 
  & None & 21.00 & 6.26 & 4.35 & 8.94 \\
& & Simplified & 4.29 & 1.92 & 1.19 & 2.24 \\
& & Detailed & 2.29 & 1.41 & 0.75 & 1.40 \\
\cmidrule{2-7}
 & \multirow{3}{*}{$r_Q$} 
  & None & 0.04 & 0.07 & 0.06 & 0.06 \\
& & Simplified & 0.01 & 0.02 & 0.02 & 0.02 \\
& & Detailed & 0.02 & 0.02 & 0.01 & 0.01 \\
\cmidrule{2-7}
& \multirow{3}{*}{$r_K$} 
  & None & 0.05 & 0.04 & 0.09 & 0.05 \\
& & Simplified & 0.04 & 0.03 & 0.02 & 0.03 \\
& & Detailed & 0.02 & 0.03 & 0.02 & 0.03 \\
\cmidrule{2-7}
 & \multirow{3}{*}{$r_V$} 
  & None & 0.06 & 0.10 & 0.08 & 0.08 \\
& & Simplified & 0.04 & 0.08 & 0.13 & 0.08 \\
& & Detailed & 0.03 & 0.04 & 0.06 & 0.04 \\
\cmidrule{2-7}
& \multirow{3}{*}{$r_O$} 
  & None & 0.04 & 0.07 & 0.05 & 0.05 \\
& & Simplified & 0.02 & 0.03 & 0.08 & 0.04 \\
& & Detailed & 0.01 & 0.01 & 0.05 & 0.02 \\
\bottomrule
\end{tabular}
}
\caption{
Statistical results for StrategyQA using Qwen2-1.5B on irrelevant responses.
}
\label{tbl:reasoning_cot_500_wrong_answer_shuffle_qwen2_1_5b_grads_strategyqa_train}
\end{table}

\begin{figure*}[!h]
\centering
\includegraphics[width=\linewidth]{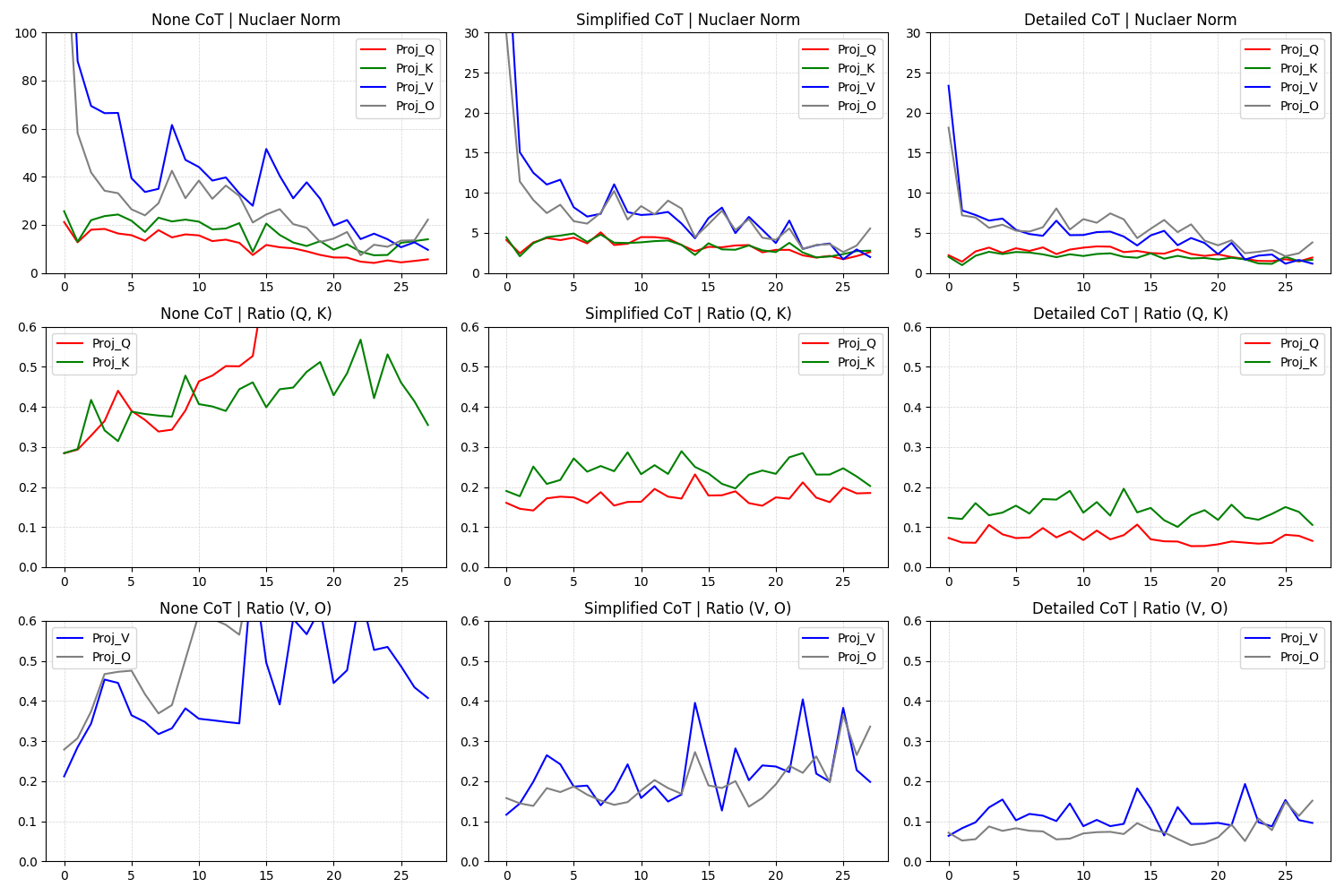}
\caption{Visualization for StrategyQA using Qwen2-1.5B on irrelevant responses.}
\label{fig:reasoning_nonsense_strategyqa}
\vspace{-2mm}
\end{figure*}

\begin{table}[!tbh]
\centering
\small
\scalebox{0.75}{
\begin{tabular}{l|l|l|cccc}
\toprule
\multirow{2}{*}{\textbf{Dataset}} & \multirow{2}{*}{\textbf{Curve}} & \multirow{2}{*}{\textbf{Cot}} &  \multicolumn{4}{c}{\textbf{Mean Absolute Difference (MAD)}} \\
 & &  & Early & Middle & Last & All\\
\midrule
\multirow{24}{*}{ECQA} & \multirow{3}{*}{$s_Q$} 
  & None & 5.61 & 4.74 & 3.44 & 4.48 \\
& & Simplified & 1.34 & 0.90 & 1.70 & 1.20 \\
& & Detailed & 0.64 & 0.42 & 0.29 & 0.45 \\
\cmidrule{2-7}
& \multirow{3}{*}{$s_K$} 
  & None & 8.65 & 6.92 & 6.53 & 6.98 \\
& & Simplified & 1.58 & 1.25 & 1.92 & 1.41 \\
& & Detailed & 0.56 & 0.32 & 0.56 & 0.43 \\
\cmidrule{2-7}
& \multirow{3}{*}{$s_V$} 
  & None & 45.35 & 16.32 & 6.82 & 20.02 \\
& & Simplified & 10.01 & 4.28 & 2.30 & 4.90 \\
& & Detailed & 3.07 & 1.13 & 0.79 & 1.44 \\
\cmidrule{2-7}
& \multirow{3}{*}{$s_O$} 
  & None & 29.13 & 14.21 & 5.63 & 14.91 \\
& & Simplified & 6.62 & 4.35 & 2.41 & 4.21 \\
& & Detailed & 2.11 & 1.51 & 0.80 & 1.41 \\
\cmidrule{2-7}
 & \multirow{3}{*}{$r_Q$} 
  & None & 0.02 & 0.04 & 0.07 & 0.04 \\
& & Simplified & 0.01 & 0.01 & 0.12 & 0.04 \\
& & Detailed & 0.01 & 0.02 & 0.02 & 0.02 \\
\cmidrule{2-7}
& \multirow{3}{*}{$r_K$} 
  & None & 0.04 & 0.04 & 0.05 & 0.04 \\
& & Simplified & 0.03 & 0.04 & 0.04 & 0.04 \\
& & Detailed & 0.02 & 0.03 & 0.03 & 0.03 \\
\cmidrule{2-7}
 & \multirow{3}{*}{$r_V$} 
  & None & 0.04 & 0.07 & 0.04 & 0.05 \\
& & Simplified & 0.03 & 0.07 & 0.08 & 0.06 \\
& & Detailed & 0.02 & 0.03 & 0.04 & 0.03 \\
\cmidrule{2-7}
& \multirow{3}{*}{$r_O$} 
  & None & 0.02 & 0.05 & 0.07 & 0.05 \\
& & Simplified & 0.02 & 0.03 & 0.07 & 0.04 \\
& & Detailed & 0.02 & 0.02 & 0.04 & 0.02 \\
\bottomrule
\end{tabular}
}
\caption{
Statistical results for ECQA using Qwen2-1.5B on irrelevant responses.
}
\label{tbl:reasoning_cot_500_wrong_answer_shuffle_qwen2_1_5b_grads_ecqa_train}
\end{table}

\begin{figure*}[!h]
\centering
\includegraphics[width=\linewidth]{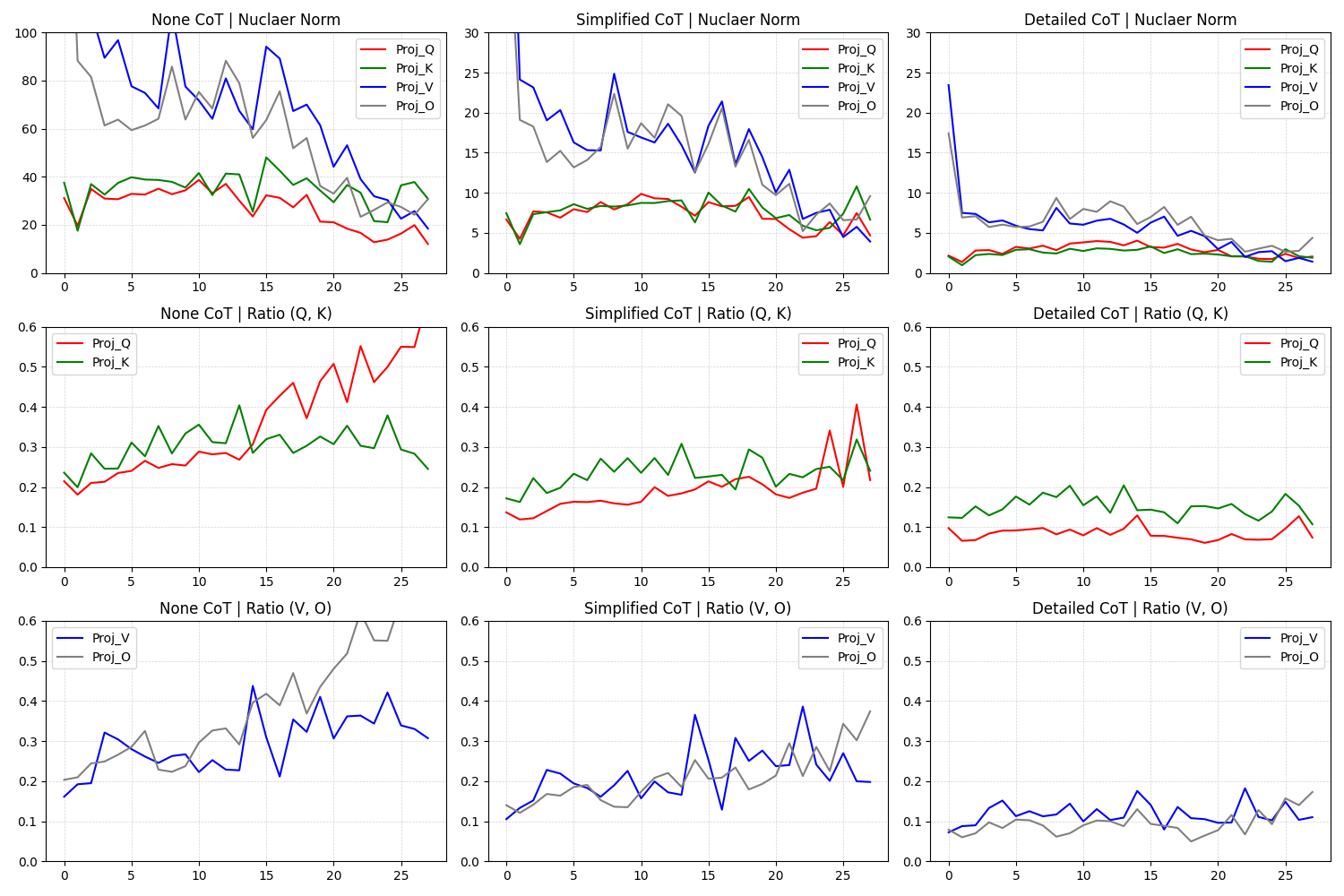}
\caption{Visualization for ECQA using Qwen2-1.5B on irrelevant responses.}
\label{fig:reasoning_nonsense_ecqa}
\vspace{-2mm}
\end{figure*}

\begin{table}[!tbh]
\centering
\small
\scalebox{0.75}{
\begin{tabular}{l|l|l|cccc}
\toprule
\multirow{2}{*}{\textbf{Dataset}} & \multirow{2}{*}{\textbf{Curve}} & \multirow{2}{*}{\textbf{Cot}} &  \multicolumn{4}{c}{\textbf{Mean Absolute Difference (MAD)}} \\
 & &  & Early & Middle & Last & All\\
\midrule
\multirow{24}{*}{CREAK} & \multirow{3}{*}{$s_Q$} 
  & None & 13.97 & 9.47 & 8.58 & 9.75 \\
& & Simplified & 1.97 & 1.63 & 4.16 & 2.30 \\
& & Detailed & 0.70 & 0.37 & 0.20 & 0.42 \\
\cmidrule{2-7}
& \multirow{3}{*}{$s_K$} 
  & None & 20.17 & 15.47 & 18.95 & 16.82 \\
& & Simplified & 2.36 & 2.12 & 4.41 & 2.56 \\
& & Detailed & 0.61 & 0.33 & 0.50 & 0.42 \\
\cmidrule{2-7}
& \multirow{3}{*}{$s_V$} 
  & None & 110.47 & 41.19 & 12.41 & 47.98 \\
& & Simplified & 15.38 & 6.33 & 4.18 & 7.61 \\
& & Detailed & 3.76 & 1.38 & 0.91 & 1.75 \\
\cmidrule{2-7}
& \multirow{3}{*}{$s_O$} 
  & None & 70.90 & 36.52 & 4.43 & 35.11 \\
& & Simplified & 9.95 & 6.47 & 3.95 & 6.43 \\
& & Detailed & 2.53 & 1.76 & 0.82 & 1.64 \\
\cmidrule{2-7}
 & \multirow{3}{*}{$r_Q$} 
  & None & 0.02 & 0.06 & 0.12 & 0.07 \\
& & Simplified & 0.01 & 0.01 & 0.11 & 0.03 \\
& & Detailed & 0.01 & 0.01 & 0.03 & 0.02 \\
\cmidrule{2-7}
& \multirow{3}{*}{$r_K$} 
  & None & 0.04 & 0.05 & 0.05 & 0.05 \\
& & Simplified & 0.04 & 0.04 & 0.05 & 0.04 \\
& & Detailed & 0.02 & 0.03 & 0.03 & 0.03 \\
\cmidrule{2-7}
 & \multirow{3}{*}{$r_V$} 
  & None & 0.04 & 0.07 & 0.05 & 0.06 \\
& & Simplified & 0.03 & 0.08 & 0.12 & 0.07 \\
& & Detailed & 0.02 & 0.04 & 0.05 & 0.04 \\
\cmidrule{2-7}
& \multirow{3}{*}{$r_O$} 
  & None & 0.02 & 0.06 & 0.08 & 0.05 \\
& & Simplified & 0.02 & 0.05 & 0.09 & 0.05 \\
& & Detailed & 0.02 & 0.02 & 0.05 & 0.02 \\
\bottomrule
\end{tabular}
}
\caption{
Statistical results for CREAK using Qwen2-1.5B on irrelevant responses.
}
\label{tbl:reasoning_cot_500_wrong_answer_shuffle_qwen2_1_5b_grads_creak_train}
\end{table}

\begin{figure*}[!h]
\centering
\includegraphics[width=\linewidth]{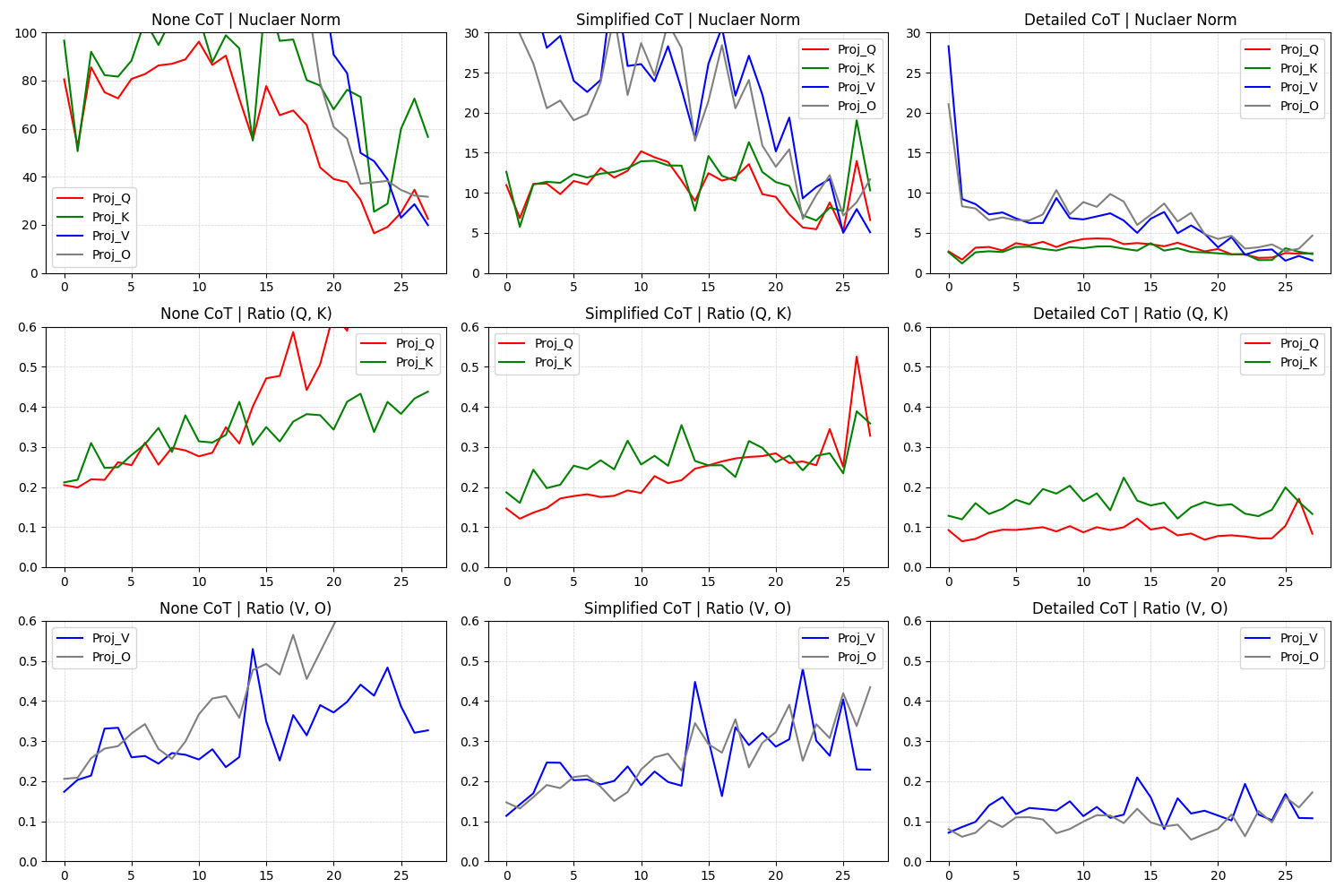}
\caption{Visualization for CREAK using Qwen2-1.5B on irrelevant responses.}
\label{fig:reasoning_nonsense_creak}
\vspace{-2mm}
\end{figure*}

\begin{table}[!tbh]
\centering
\small
\scalebox{0.75}{
\begin{tabular}{l|l|l|cccc}
\toprule
\multirow{2}{*}{\textbf{Dataset}} & \multirow{2}{*}{\textbf{Curve}} & \multirow{2}{*}{\textbf{Cot}} &  \multicolumn{4}{c}{\textbf{Mean Absolute Difference (MAD)}} \\
 & &  & Early & Middle & Last & All\\
\midrule
\multirow{24}{*}{Sensemaking} & \multirow{3}{*}{$s_Q$} 
  & None & 5.14 & 3.36 & 5.88 & 4.39 \\
& & Simplified & 1.86 & 1.29 & 2.41 & 1.73 \\
& & Detailed & 0.67 & 0.58 & 0.41 & 0.57 \\
\cmidrule{2-7}
& \multirow{3}{*}{$s_K$} 
  & None & 7.37 & 4.08 & 10.96 & 6.43 \\
& & Simplified & 2.35 & 1.62 & 2.64 & 1.93 \\
& & Detailed & 0.55 & 0.50 & 0.78 & 0.54 \\
\cmidrule{2-7}
& \multirow{3}{*}{$s_V$} 
  & None & 39.58 & 14.36 & 5.81 & 17.47 \\
& & Simplified & 13.80 & 5.90 & 2.85 & 6.67 \\
& & Detailed & 2.90 & 1.12 & 0.68 & 1.37 \\
\cmidrule{2-7}
& \multirow{3}{*}{$s_O$} 
  & None & 26.23 & 12.84 & 2.81 & 13.07 \\
& & Simplified & 9.13 & 5.58 & 2.46 & 5.46 \\
& & Detailed & 2.14 & 1.48 & 0.60 & 1.36 \\
\cmidrule{2-7}
 & \multirow{3}{*}{$r_Q$} 
  & None & 0.03 & 0.05 & 0.19 & 0.07 \\
& & Simplified & 0.02 & 0.02 & 0.10 & 0.04 \\
& & Detailed & 0.01 & 0.02 & 0.03 & 0.02 \\
\cmidrule{2-7}
& \multirow{3}{*}{$r_K$} 
  & None & 0.03 & 0.06 & 0.11 & 0.06 \\
& & Simplified & 0.03 & 0.03 & 0.07 & 0.04 \\
& & Detailed & 0.02 & 0.05 & 0.02 & 0.04 \\
\cmidrule{2-7}
 & \multirow{3}{*}{$r_V$} 
  & None & 0.04 & 0.05 & 0.06 & 0.05 \\
& & Simplified & 0.03 & 0.06 & 0.08 & 0.06 \\
& & Detailed & 0.02 & 0.03 & 0.03 & 0.03 \\
\cmidrule{2-7}
& \multirow{3}{*}{$r_O$} 
  & None & 0.03 & 0.04 & 0.10 & 0.05 \\
& & Simplified & 0.01 & 0.03 & 0.08 & 0.04 \\
& & Detailed & 0.01 & 0.02 & 0.04 & 0.02 \\
\bottomrule
\end{tabular}
}
\caption{
Statistical results for Sensemaking using Qwen2-1.5B on irrelevant responses.
}
\label{tbl:reasoning_cot_500_wrong_answer_shuffle_qwen2_1_5b_grads_sensemaking_train}
\end{table}

\begin{figure*}[!h]
\centering
\includegraphics[width=\linewidth]{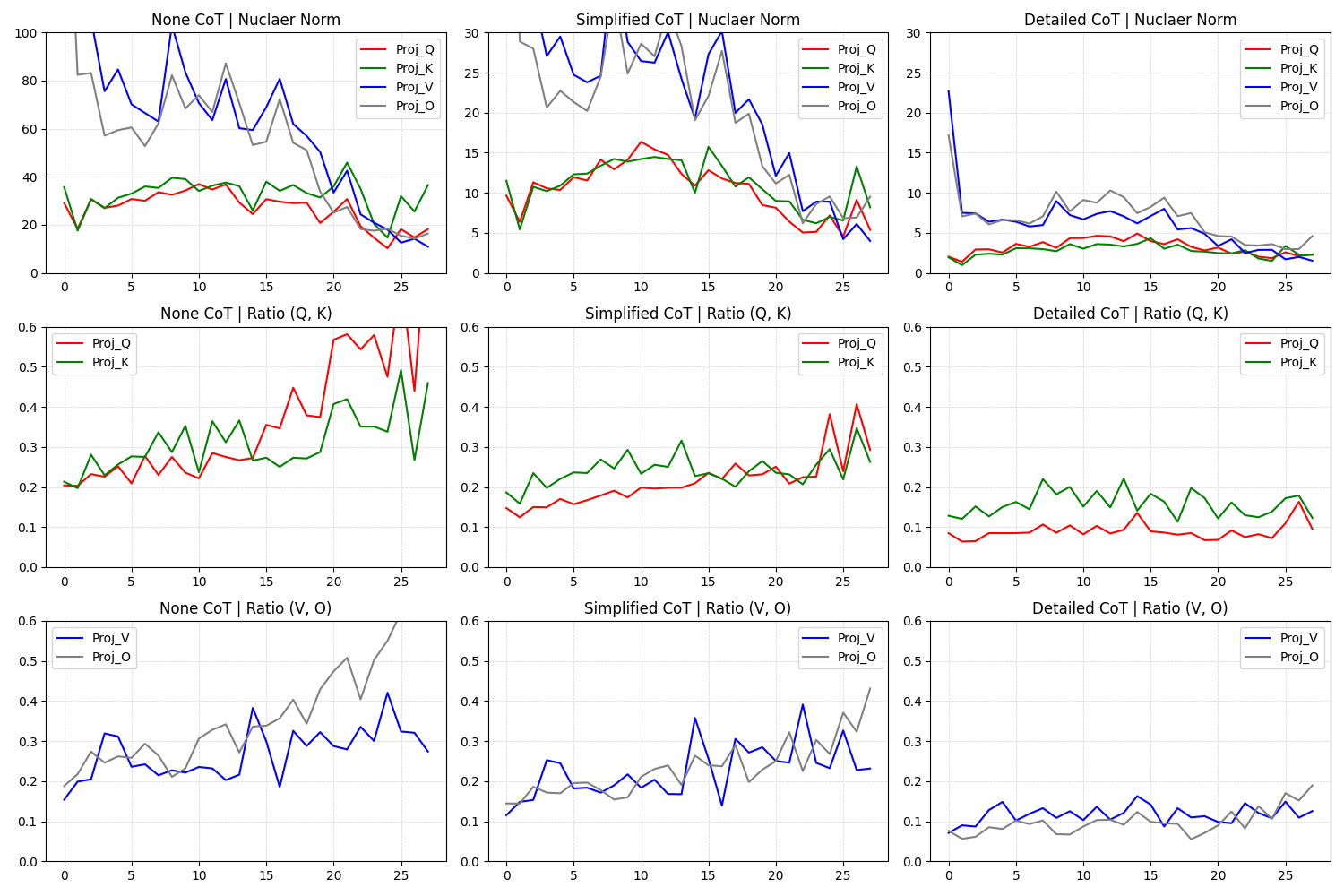}
\caption{Visualization for Sensemaking using Qwen2-1.5B on irrelevant responses.}
\label{fig:reasoning_nonsense_sensemaking}
\vspace{-2mm}
\end{figure*}

\subsubsection{Wiki Tasks}

The visualizations and statistical results on Wiki tasks are shown in Figure \ref{fig:wiki_nonsense} and Table \ref{tbl:wiki_wiki_popularity_wrong_answer_shuffle_qwen2_1_5b_grads}.

\begin{table}[!tbh]
\centering
\small
\scalebox{0.75}{
\begin{tabular}{l|l|l|cccc}
\toprule
\multirow{2}{*}{\textbf{Dataset}} & \multirow{2}{*}{\textbf{Curve}} & \multirow{2}{*}{\textbf{Cot}} &  \multicolumn{4}{c}{\textbf{Mean Absolute Difference (MAD)}} \\
 & &  & Early & Middle & Last & All\\
\midrule
\multirow{32}{*}{Wiki} & \multirow{3}{*}{$s_Q$} 
  & Len 100 & 0.65 & 0.34 & 0.33 & 0.43 \\
& & Len 500 & 0.56 & 0.26 & 0.27 & 0.34 \\
& & Len 1000 & 0.49 & 0.23 & 0.25 & 0.29 \\
& & Unpopular & 0.63 & 0.46 & 0.30 & 0.45 \\
\cmidrule{2-7}
& \multirow{3}{*}{$s_K$} 
  & Len 100 & 0.76 & 0.32 & 0.36 & 0.46 \\
& & Len 500 & 0.48 & 0.19 & 0.21 & 0.28 \\
& & Len 1000 & 0.38 & 0.14 & 0.17 & 0.21 \\
& & Unpopular & 0.88 & 0.45 & 0.47 & 0.56 \\
\cmidrule{2-7}
& \multirow{3}{*}{$s_V$} 
  & Len 100 & 4.83 & 1.52 & 1.18 & 2.17 \\
& & Len 500 & 3.35 & 1.06 & 0.94 & 1.54 \\
& & Len 1000 & 2.75 & 0.83 & 0.79 & 1.25 \\
& & Unpopular & 6.99 & 1.80 & 1.14 & 2.78 \\
\cmidrule{2-7}
& \multirow{3}{*}{$s_O$} 
  & Len 100 & 3.36 & 1.57 & 0.84 & 1.78 \\
& & Len 500 & 2.49 & 1.36 & 0.77 & 1.44 \\
& & Len 1000 & 2.16 & 1.22 & 0.70 & 1.28 \\
& & Unpopular & 4.71 & 1.71 & 0.92 & 2.18 \\
\cmidrule{2-7}
 & \multirow{3}{*}{$r_Q$} 
  & Len 100 & 0.02 & 0.02 & 0.03 & 0.02 \\
& & Len 500 & 0.02 & 0.01 & 0.01 & 0.02 \\
& & Len 1000 & 0.01 & 0.01 & 0.01 & 0.01 \\
& & Unpopular & 0.01 & 0.02 & 0.03 & 0.02 \\
\cmidrule{2-7}
& \multirow{3}{*}{$r_K$} 
  & Len 100 & 0.03 & 0.03 & 0.02 & 0.03 \\
& & Len 500 & 0.03 & 0.02 & 0.01 & 0.02 \\
& & Len 1000 & 0.02 & 0.02 & 0.01 & 0.02 \\
& & Unpopular & 0.04 & 0.03 & 0.02 & 0.03 \\
\cmidrule{2-7}
 & \multirow{3}{*}{$r_V$} 
  & Len 100 & 0.03 & 0.07 & 0.11 & 0.07 \\
& & Len 500 & 0.02 & 0.05 & 0.09 & 0.05 \\
& & Len 1000 & 0.02 & 0.05 & 0.08 & 0.04 \\
& & Unpopular & 0.04 & 0.08 & 0.11 & 0.07 \\
\cmidrule{2-7}
& \multirow{3}{*}{$r_O$} 
  & Len 100 & 0.02 & 0.03 & 0.07 & 0.03 \\
& & Len 500 & 0.02 & 0.01 & 0.06 & 0.02 \\
& & Len 1000 & 0.01 & 0.01 & 0.05 & 0.02 \\
& & Unpopular & 0.02 & 0.04 & 0.08 & 0.04 \\
\bottomrule
\end{tabular}
}
\caption{
Statistical results for Wiki using Qwen2-1.5B on irrelevant responses.
}
\label{tbl:wiki_wiki_popularity_wrong_answer_shuffle_qwen2_1_5b_grads}
\end{table}

\begin{figure*}[!h]
\centering
\includegraphics[width=\linewidth]{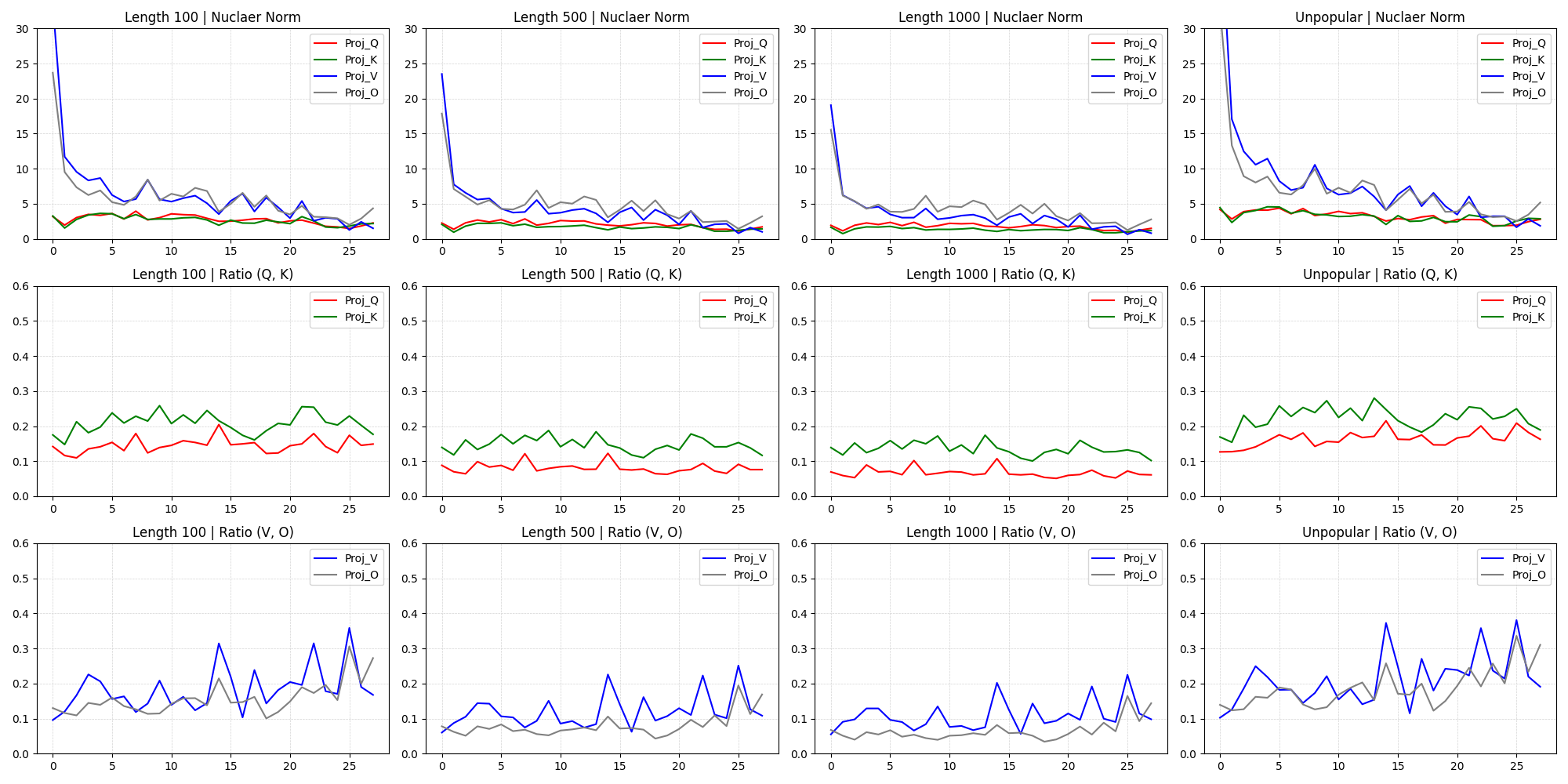}
\caption{Visualization for Wiki tasks using Qwen2-1.5B on irrelevant responses.}
\label{fig:wiki_nonsense}
\vspace{-2mm}
\end{figure*}

\clearpage

\subsection{Instructed LLM on Correct Responses}

\subsubsection{Reasoning Tasks}

The visualizations and statistical results on MATH tasks: MATH-Algebra (Figure \ref{fig:math_correct_algebra_it}, Table \ref{tbl:math_lighteval_math_qwen2_1_5b_it_grads_algebra}), MATH-Counting (Figure \ref{fig:math_correct_counting_it}, Table \ref{tbl:math_lighteval_math_qwen2_1_5b_it_grads_counting_probability}), MATH-Geometry (Figure \ref{fig:math_correct_geometry_it}, Table \ref{tbl:math_lighteval_math_qwen2_1_5b_it_grads_geometry}).

\begin{table}[!tbh]
\centering
\small
\scalebox{0.75}{
\begin{tabular}{l|l|l|cccc}
\toprule
\multirow{2}{*}{\textbf{Dataset}} & \multirow{2}{*}{\textbf{Curve}} & \multirow{2}{*}{\textbf{Cot}} &  \multicolumn{4}{c}{\textbf{Mean Absolute Difference (MAD)}} \\
 & &  & Early & Middle & Last & All\\
\midrule
\multirow{16}{*}{Algebra} & \multirow{3}{*}{$s_Q$} 
  & Simplified & 0.40 & 0.38 & 0.52 & 0.43 \\
& & Detailed & 0.18 & 0.27 & 0.27 & 0.26 \\
\cmidrule{2-7}
& \multirow{3}{*}{$s_K$} 
  & Simplified & 0.40 & 0.35 & 0.59 & 0.42 \\
& & Detailed & 0.17 & 0.20 & 0.39 & 0.24 \\
\cmidrule{2-7}
& \multirow{3}{*}{$s_V$} 
  & Simplified & 1.93 & 1.25 & 0.48 & 1.16 \\
& & Detailed & 0.88 & 0.54 & 0.30 & 0.53 \\
\cmidrule{2-7}
& \multirow{3}{*}{$s_O$} 
  & Simplified & 1.40 & 1.34 & 0.43 & 1.09 \\
& & Detailed & 0.66 & 0.67 & 0.28 & 0.55 \\
\cmidrule{2-7}
 & \multirow{3}{*}{$r_Q$} 
  & Simplified & 0.02 & 0.01 & 0.03 & 0.02 \\
& & Detailed & 0.01 & 0.01 & 0.03 & 0.01 \\
\cmidrule{2-7}
& \multirow{3}{*}{$r_K$} 
  & Simplified & 0.03 & 0.03 & 0.02 & 0.03 \\
& & Detailed & 0.02 & 0.03 & 0.02 & 0.03 \\
\cmidrule{2-7}
 & \multirow{3}{*}{$r_V$} 
  & Simplified & 0.03 & 0.03 & 0.04 & 0.03 \\
& & Detailed & 0.02 & 0.02 & 0.03 & 0.02 \\
\cmidrule{2-7}
& \multirow{3}{*}{$r_O$} 
  & Simplified & 0.02 & 0.02 & 0.07 & 0.03 \\
& & Detailed & 0.01 & 0.01 & 0.07 & 0.03 \\
\bottomrule
\end{tabular}
}
\caption{
Statistical results for MATH-Algebra using Qwen2-1.5B-Instruct on correct responses.
}
\label{tbl:math_lighteval_math_qwen2_1_5b_it_grads_algebra}
\end{table}

\begin{figure}[!h]
\centering
\includegraphics[width=\linewidth]{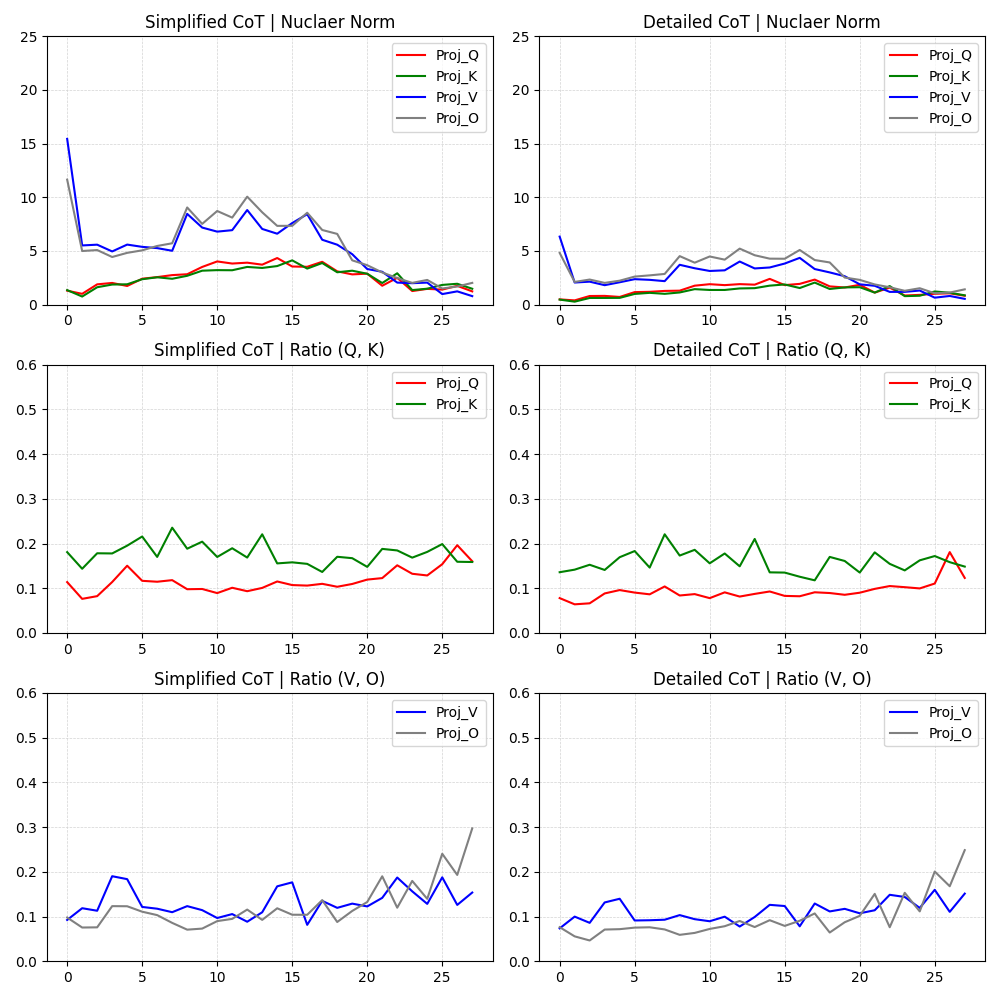}
\caption{Visualization for MATH-Algebra using Qwen2-1.5B-Instruct on correct responses.}
\label{fig:math_correct_algebra_it}
\vspace{-2mm}
\end{figure}

\begin{table}[!tbh]
\centering
\small
\scalebox{0.75}{
\begin{tabular}{l|l|l|cccc}
\toprule
\multirow{2}{*}{\textbf{Dataset}} & \multirow{2}{*}{\textbf{Curve}} & \multirow{2}{*}{\textbf{Cot}} &  \multicolumn{4}{c}{\textbf{Mean Absolute Difference (MAD)}} \\
 & &  & Early & Middle & Last & All\\
\midrule
\multirow{16}{*}{Counting} & \multirow{3}{*}{$s_Q$} 
  & Simplified & 0.44 & 0.40 & 0.51 & 0.46 \\
& & Detailed & 0.22 & 0.32 & 0.30 & 0.31 \\
\cmidrule{2-7}
& \multirow{3}{*}{$s_K$} 
  & Simplified & 0.42 & 0.30 & 0.59 & 0.40 \\
& & Detailed & 0.20 & 0.21 & 0.45 & 0.27 \\
\cmidrule{2-7}
& \multirow{3}{*}{$s_V$} 
  & Simplified & 2.05 & 1.28 & 0.55 & 1.21 \\
& & Detailed & 1.07 & 0.57 & 0.35 & 0.60 \\
\cmidrule{2-7}
& \multirow{3}{*}{$s_O$} 
  & Simplified & 1.49 & 1.36 & 0.52 & 1.14 \\
& & Detailed & 0.79 & 0.73 & 0.37 & 0.64 \\
\cmidrule{2-7}
 & \multirow{3}{*}{$r_Q$} 
  & Simplified & 0.02 & 0.01 & 0.02 & 0.01 \\
& & Detailed & 0.01 & 0.01 & 0.02 & 0.01 \\
\cmidrule{2-7}
& \multirow{3}{*}{$r_K$} 
  & Simplified & 0.02 & 0.03 & 0.02 & 0.02 \\
& & Detailed & 0.02 & 0.03 & 0.01 & 0.03 \\
\cmidrule{2-7}
 & \multirow{3}{*}{$r_V$} 
  & Simplified & 0.03 & 0.03 & 0.04 & 0.03 \\
& & Detailed & 0.02 & 0.02 & 0.03 & 0.02 \\
\cmidrule{2-7}
& \multirow{3}{*}{$r_O$} 
  & Simplified & 0.02 & 0.02 & 0.07 & 0.03 \\
& & Detailed & 0.01 & 0.02 & 0.06 & 0.02 \\
\bottomrule
\end{tabular}
}
\caption{
Statistical results for MATH-Counting using Qwen2-1.5B-Instruct on correct responses.
}
\label{tbl:math_lighteval_math_qwen2_1_5b_it_grads_counting_probability}
\end{table}

\begin{figure}[!h]
\centering
\includegraphics[width=\linewidth]{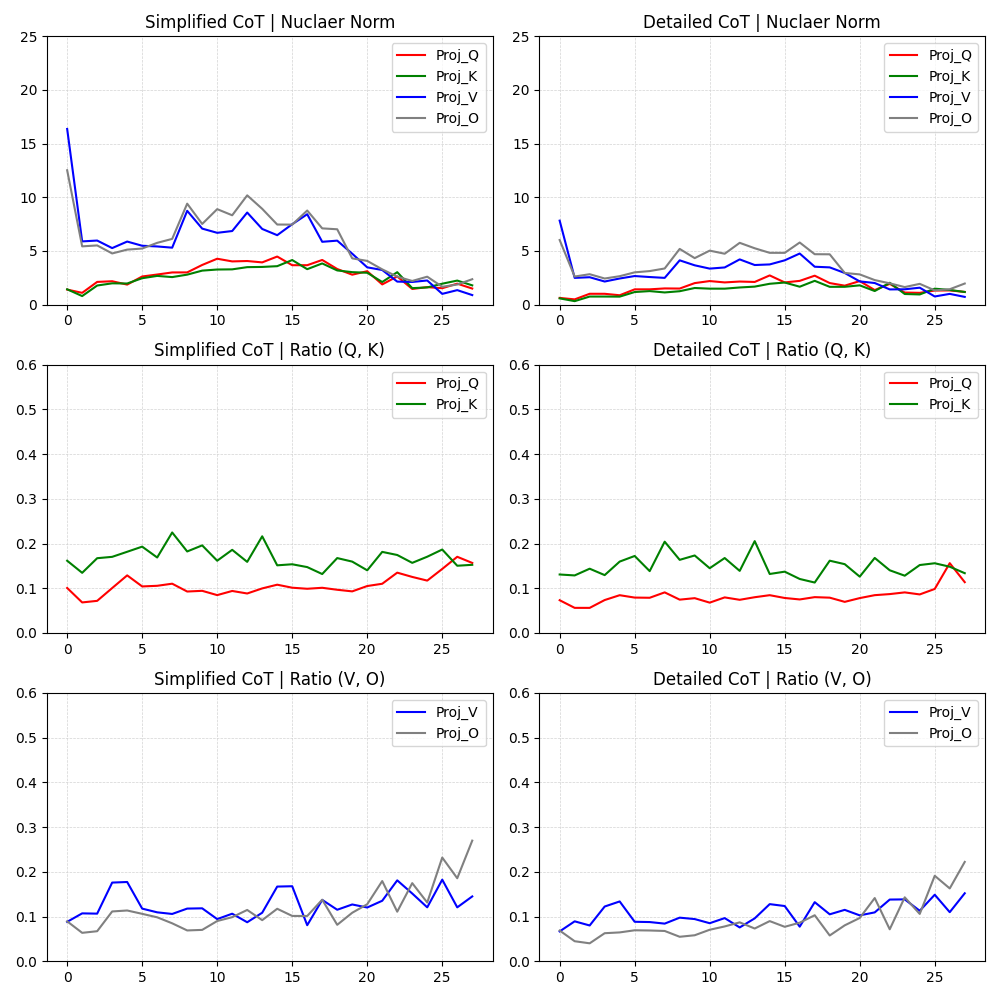}
\caption{Visualization for MATH-Counting using Qwen2-1.5B-Instruct on correct responses.}
\label{fig:math_correct_counting_it}
\vspace{-2mm}
\end{figure}

\begin{table}[!tbh]
\centering
\small
\scalebox{0.75}{
\begin{tabular}{l|l|l|cccc}
\toprule
\multirow{2}{*}{\textbf{Dataset}} & \multirow{2}{*}{\textbf{Curve}} & \multirow{2}{*}{\textbf{Cot}} &  \multicolumn{4}{c}{\textbf{Mean Absolute Difference (MAD)}} \\
 & &  & Early & Middle & Last & All\\
\midrule
\multirow{16}{*}{Geometry} & \multirow{3}{*}{$s_Q$} 
  & Simplified & 0.40 & 0.44 & 0.57 & 0.47 \\
& & Detailed & 0.26 & 0.36 & 0.40 & 0.36 \\
\cmidrule{2-7}
& \multirow{3}{*}{$s_K$} 
  & Simplified & 0.35 & 0.28 & 0.72 & 0.40 \\
& & Detailed & 0.23 & 0.24 & 0.51 & 0.30 \\
\cmidrule{2-7}
& \multirow{3}{*}{$s_V$} 
  & Simplified & 1.89 & 1.07 & 0.54 & 1.06 \\
& & Detailed & 1.28 & 0.62 & 0.39 & 0.68 \\
\cmidrule{2-7}
& \multirow{3}{*}{$s_O$} 
  & Simplified & 1.52 & 1.35 & 0.42 & 1.11 \\
& & Detailed & 0.97 & 0.86 & 0.36 & 0.73 \\
\cmidrule{2-7}
 & \multirow{3}{*}{$r_Q$} 
  & Simplified & 0.02 & 0.01 & 0.01 & 0.01 \\
& & Detailed & 0.01 & 0.01 & 0.03 & 0.01 \\
\cmidrule{2-7}
& \multirow{3}{*}{$r_K$} 
  & Simplified & 0.01 & 0.03 & 0.01 & 0.02 \\
& & Detailed & 0.02 & 0.03 & 0.02 & 0.03 \\
\cmidrule{2-7}
 & \multirow{3}{*}{$r_V$} 
  & Simplified & 0.03 & 0.02 & 0.03 & 0.02 \\
& & Detailed & 0.02 & 0.02 & 0.03 & 0.02 \\
\cmidrule{2-7}
& \multirow{3}{*}{$r_O$} 
  & Simplified & 0.01 & 0.01 & 0.06 & 0.02 \\
& & Detailed & 0.01 & 0.01 & 0.06 & 0.02 \\
\bottomrule
\end{tabular}
}
\caption{
Statistical results for MATH-Geometry using Qwen2-1.5B-Instruct on correct responses.
}
\label{tbl:math_lighteval_math_qwen2_1_5b_it_grads_geometry}
\end{table}

\begin{figure}[!h]
\centering
\includegraphics[width=\linewidth]{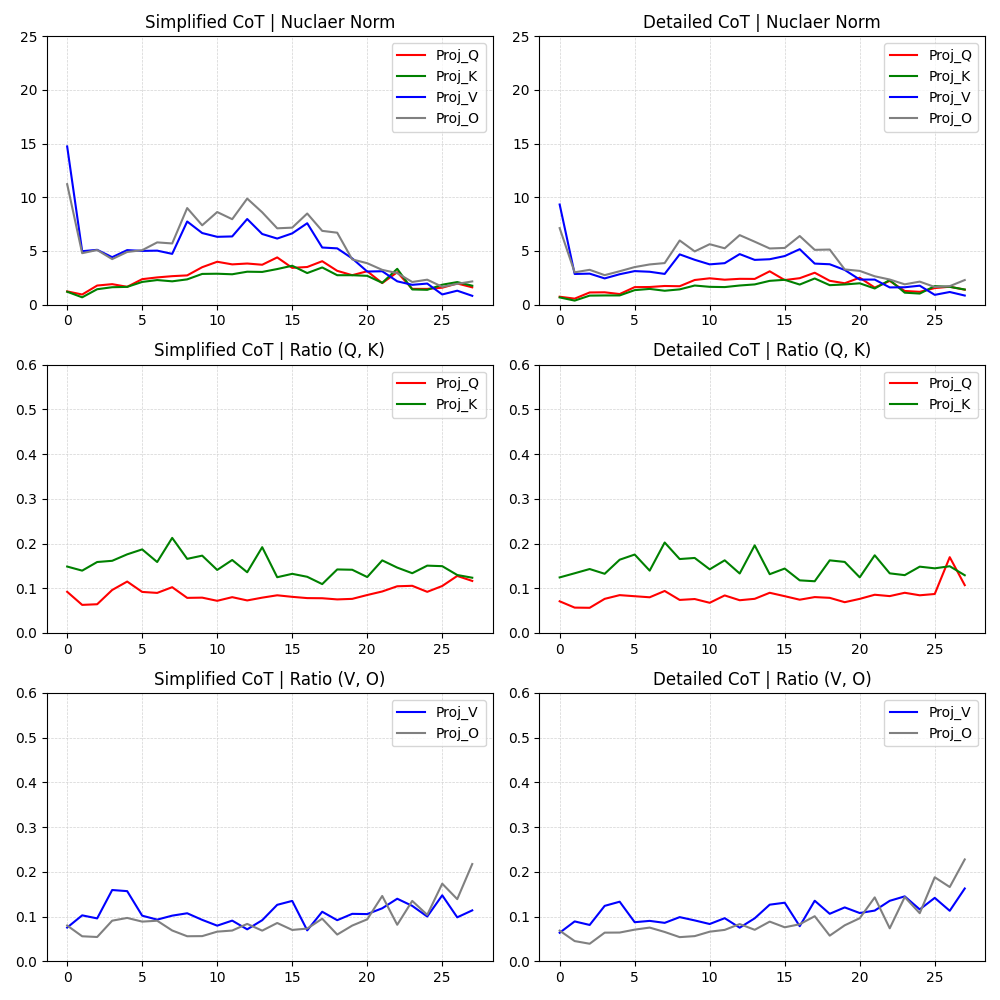}
\caption{Visualization for MATH-Geometry using Qwen2-1.5B-Instruct on correct responses.}
\label{fig:math_correct_geometry_it}
\vspace{-2mm}
\end{figure}

The visualizations and statistical results on other reasoning tasks: AQuA (Figure \ref{fig:reasoning_correct_aqua_it}, Table \ref{tbl:reasoning_cot_500_qwen2_1_5b_it_grads_aqua_train}), GSM8K (Figure \ref{fig:reasoning_correct_gsm8k_it}, Table \ref{tbl:reasoning_cot_500_qwen2_1_5b_it_grads_gsm8k_train}), StrategyQA (Figure \ref{fig:reasoning_correct_strategyqa_it}, Table \ref{tbl:reasoning_cot_500_qwen2_1_5b_it_grads_strategyqa_train}), ECQA (Figure \ref{fig:reasoning_correct_ecqa_it}, Table \ref{tbl:reasoning_cot_500_qwen2_1_5b_it_grads_ecqa_train}), CREAK (Figure \ref{fig:reasoning_correct_creak_it}, Table \ref{tbl:reasoning_cot_500_qwen2_1_5b_it_grads_creak_train}), Sensemaking (Figure \ref{fig:reasoning_correct_sensemaking_it}, Table \ref{tbl:reasoning_cot_500_qwen2_1_5b_it_grads_sensemaking_train}).

\begin{table}[!tbh]
\centering
\small
\scalebox{0.75}{
\begin{tabular}{l|l|l|cccc}
\toprule
\multirow{2}{*}{\textbf{Dataset}} & \multirow{2}{*}{\textbf{Curve}} & \multirow{2}{*}{\textbf{Cot}} &  \multicolumn{4}{c}{\textbf{Mean Absolute Difference (MAD)}} \\
 & &  & Early & Middle & Last & All\\
\midrule
\multirow{24}{*}{AQuA} & \multirow{3}{*}{$s_Q$} 
  & None & 12.15 & 6.73 & 3.11 & 7.40 \\
& & Simplified & 0.82 & 0.72 & 1.04 & 0.84 \\
& & Detailed & 0.21 & 0.30 & 0.23 & 0.28 \\
\cmidrule{2-7}
& \multirow{3}{*}{$s_K$} 
  & None & 15.61 & 11.96 & 6.78 & 12.92 \\
& & Simplified & 1.09 & 0.50 & 1.46 & 0.90 \\
& & Detailed & 0.20 & 0.20 & 0.39 & 0.25 \\
\cmidrule{2-7}
& \multirow{3}{*}{$s_V$} 
  & None & 64.92 & 30.95 & 4.82 & 31.89 \\
& & Simplified & 4.49 & 2.32 & 0.98 & 2.39 \\
& & Detailed & 1.04 & 0.56 & 0.35 & 0.59 \\
\cmidrule{2-7}
& \multirow{3}{*}{$s_O$} 
  & None & 44.30 & 24.03 & 3.70 & 22.69 \\
& & Simplified & 3.22 & 2.45 & 0.71 & 2.10 \\
& & Detailed & 0.76 & 0.69 & 0.32 & 0.60 \\
\cmidrule{2-7}
 & \multirow{3}{*}{$r_Q$} 
  & None & 0.04 & 0.08 & 0.18 & 0.10 \\
& & Simplified & 0.04 & 0.01 & 0.02 & 0.02 \\
& & Detailed & 0.01 & 0.01 & 0.01 & 0.01 \\
\cmidrule{2-7}
& \multirow{3}{*}{$r_K$} 
  & None & 0.05 & 0.05 & 0.13 & 0.07 \\
& & Simplified & 0.03 & 0.04 & 0.01 & 0.03 \\
& & Detailed & 0.02 & 0.03 & 0.01 & 0.02 \\
\cmidrule{2-7}
 & \multirow{3}{*}{$r_V$} 
  & None & 0.04 & 0.05 & 0.09 & 0.05 \\
& & Simplified & 0.03 & 0.03 & 0.03 & 0.03 \\
& & Detailed & 0.03 & 0.02 & 0.03 & 0.02 \\
\cmidrule{2-7}
& \multirow{3}{*}{$r_O$} 
  & None & 0.03 & 0.06 & 0.11 & 0.06 \\
& & Simplified & 0.02 & 0.03 & 0.08 & 0.04 \\
& & Detailed & 0.01 & 0.02 & 0.06 & 0.03 \\
\bottomrule
\end{tabular}
}
\caption{
Statistical results for AQuA using Qwen2-1.5B-Instruct on correct responses.
}
\label{tbl:reasoning_cot_500_qwen2_1_5b_it_grads_aqua_train}
\end{table}

\begin{figure*}[!h]
\centering
\includegraphics[width=\linewidth]{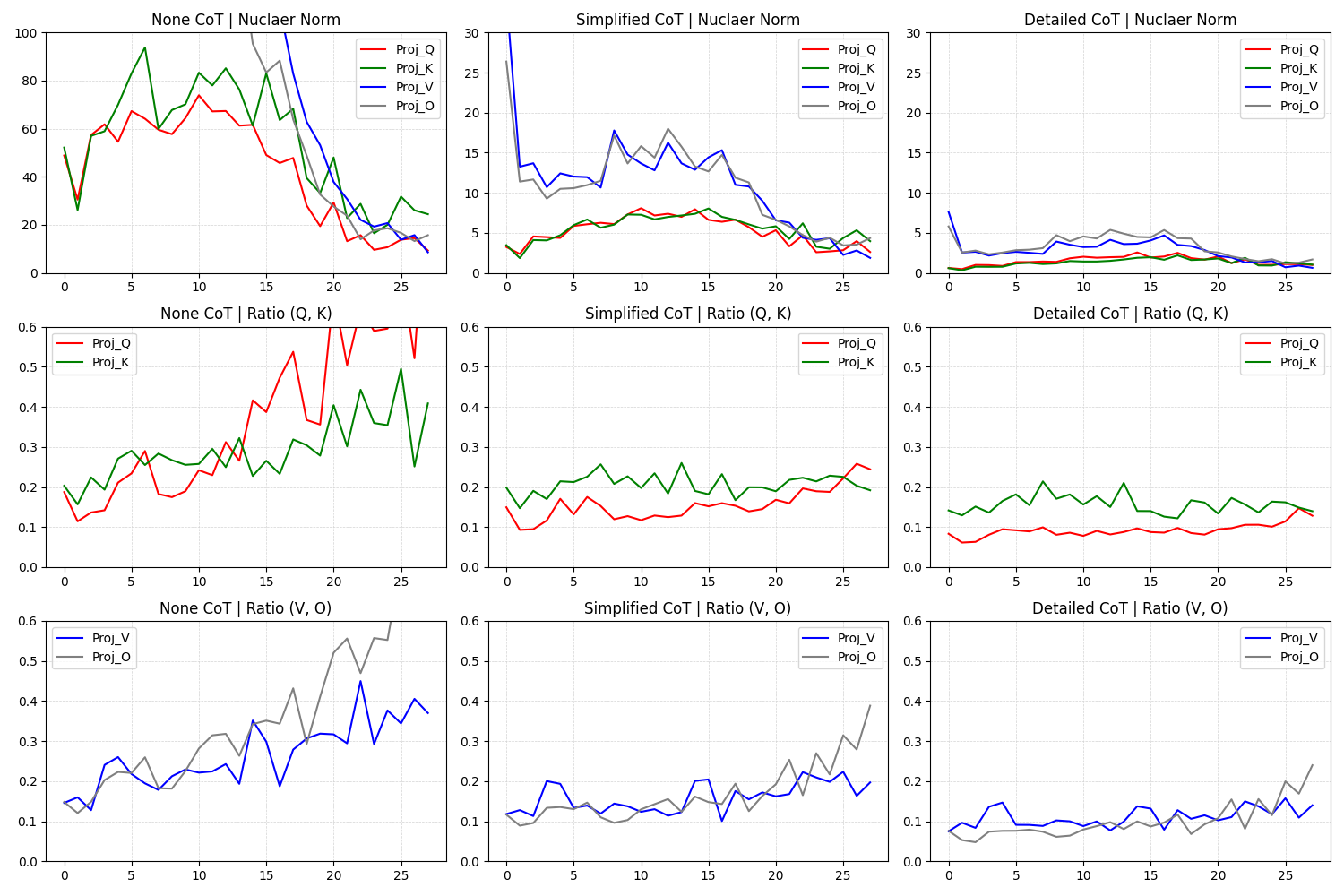}
\caption{Visualization for AQuA using Qwen2-1.5B-Instruct on correct responses.}
\label{fig:reasoning_correct_aqua_it}
\vspace{-2mm}
\end{figure*}

\begin{table}[!tbh]
\centering
\small
\scalebox{0.75}{
\begin{tabular}{l|l|l|cccc}
\toprule
\multirow{2}{*}{\textbf{Dataset}} & \multirow{2}{*}{\textbf{Curve}} & \multirow{2}{*}{\textbf{Cot}} &  \multicolumn{4}{c}{\textbf{Mean Absolute Difference (MAD)}} \\
 & &  & Early & Middle & Last & All\\
\midrule
\multirow{24}{*}{GSM8K} & \multirow{3}{*}{$s_Q$} 
  & None & 7.44 & 4.82 & 10.79 & 6.77 \\
& & Simplified & 0.68 & 0.55 & 1.04 & 0.74 \\
& & Detailed & 0.21 & 0.26 & 0.20 & 0.25 \\
\cmidrule{2-7}
& \multirow{3}{*}{$s_K$} 
  & None & 10.73 & 8.05 & 12.32 & 9.89 \\
& & Simplified & 0.81 & 0.73 & 0.98 & 0.83 \\
& & Detailed & 0.22 & 0.22 & 0.40 & 0.28 \\
\cmidrule{2-7}
& \multirow{3}{*}{$s_V$} 
  & None & 39.35 & 23.28 & 7.91 & 22.41 \\
& & Simplified & 3.20 & 2.42 & 1.23 & 2.19 \\
& & Detailed & 1.05 & 0.65 & 0.33 & 0.64 \\
\cmidrule{2-7}
& \multirow{3}{*}{$s_O$} 
  & None & 27.77 & 22.56 & 4.89 & 18.65 \\
& & Simplified & 2.25 & 2.44 & 0.70 & 1.92 \\
& & Detailed & 0.73 & 0.75 & 0.30 & 0.62 \\
\cmidrule{2-7}
 & \multirow{3}{*}{$r_Q$} 
  & None & 0.03 & 0.03 & 0.12 & 0.05 \\
& & Simplified & 0.02 & 0.01 & 0.04 & 0.02 \\
& & Detailed & 0.01 & 0.01 & 0.01 & 0.01 \\
\cmidrule{2-7}
& \multirow{3}{*}{$r_K$} 
  & None & 0.03 & 0.04 & 0.04 & 0.04 \\
& & Simplified & 0.02 & 0.04 & 0.05 & 0.04 \\
& & Detailed & 0.03 & 0.03 & 0.02 & 0.03 \\
\cmidrule{2-7}
 & \multirow{3}{*}{$r_V$} 
  & None & 0.03 & 0.05 & 0.05 & 0.04 \\
& & Simplified & 0.02 & 0.03 & 0.04 & 0.03 \\
& & Detailed & 0.02 & 0.02 & 0.03 & 0.02 \\
\cmidrule{2-7}
& \multirow{3}{*}{$r_O$} 
  & None & 0.02 & 0.04 & 0.08 & 0.05 \\
& & Simplified & 0.01 & 0.02 & 0.08 & 0.03 \\
& & Detailed & 0.01 & 0.02 & 0.06 & 0.03 \\
\bottomrule
\end{tabular}
}
\caption{
Statistical results for GSM8K using Qwen2-1.5B-Instruct on correct responses.
}
\label{tbl:reasoning_cot_500_qwen2_1_5b_it_grads_gsm8k_train}
\end{table}

\begin{figure*}[!h]
\centering
\includegraphics[width=\linewidth]{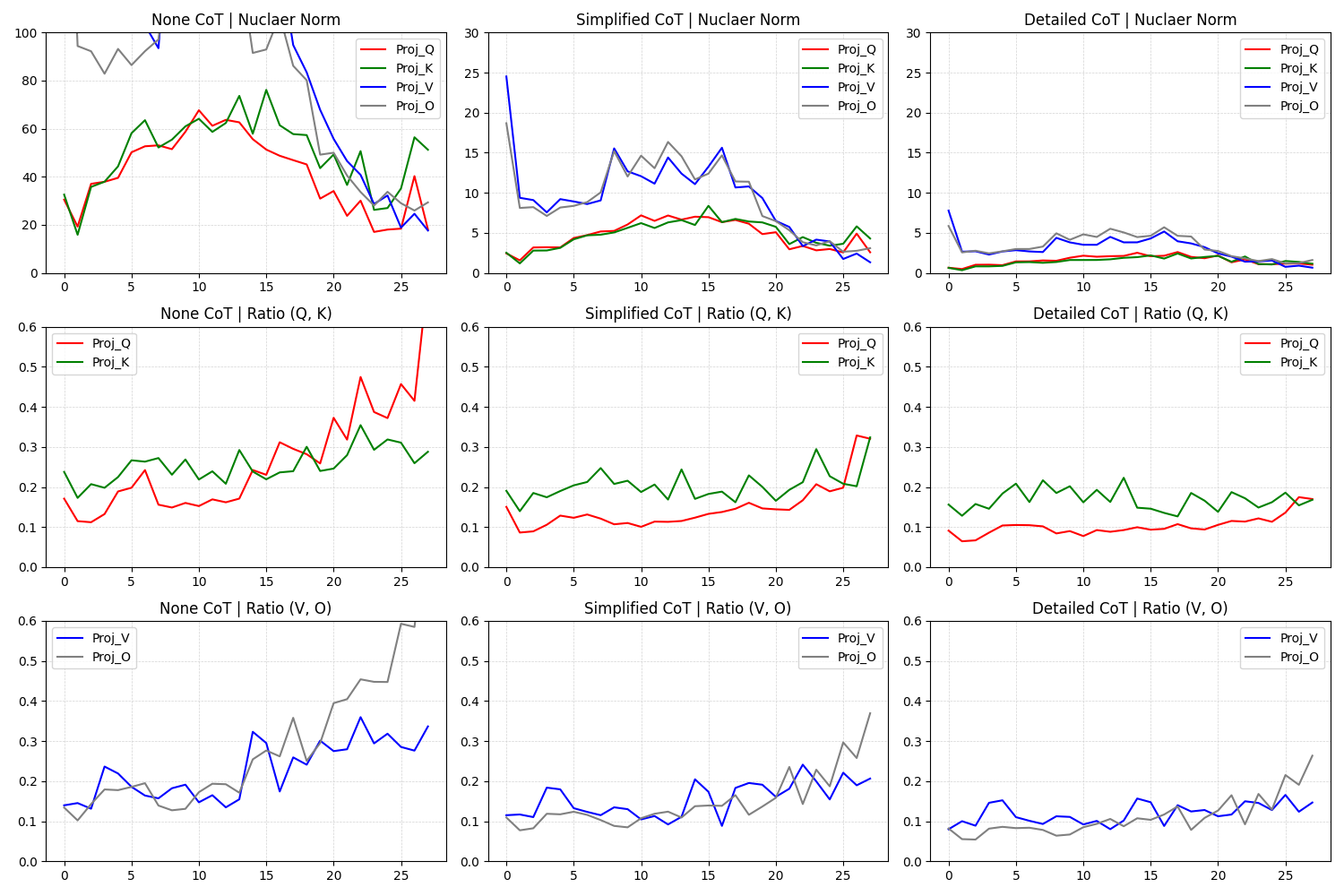}
\caption{Visualization for GSM8K using Qwen2-1.5B-Instruct on correct responses.}
\label{fig:reasoning_correct_gsm8k_it}
\vspace{-2mm}
\end{figure*}

\begin{table}[!tbh]
\centering
\small
\scalebox{0.75}{
\begin{tabular}{l|l|l|cccc}
\toprule
\multirow{2}{*}{\textbf{Dataset}} & \multirow{2}{*}{\textbf{Curve}} & \multirow{2}{*}{\textbf{Cot}} &  \multicolumn{4}{c}{\textbf{Mean Absolute Difference (MAD)}} \\
 & &  & Early & Middle & Last & All\\
\midrule
\multirow{24}{*}{StrategyQA} & \multirow{3}{*}{$s_Q$} 
  & None & 3.76 & 2.78 & 0.71 & 2.57 \\
& & Simplified & 0.74 & 0.51 & 0.32 & 0.56 \\
& & Detailed & 0.28 & 0.35 & 0.22 & 0.32 \\
\cmidrule{2-7}
& \multirow{3}{*}{$s_K$} 
  & None & 8.61 & 6.27 & 2.09 & 5.67 \\
& & Simplified & 1.18 & 0.70 & 0.47 & 0.74 \\
& & Detailed & 0.27 & 0.20 & 0.42 & 0.27 \\
\cmidrule{2-7}
& \multirow{3}{*}{$s_V$} 
  & None & 31.00 & 16.26 & 3.75 & 15.97 \\
& & Simplified & 5.58 & 2.28 & 1.44 & 2.71 \\
& & Detailed & 1.59 & 0.69 & 0.56 & 0.82 \\
\cmidrule{2-7}
& \multirow{3}{*}{$s_O$} 
  & None & 22.21 & 9.87 & 3.95 & 10.76 \\
& & Simplified & 3.85 & 2.11 & 1.15 & 2.18 \\
& & Detailed & 1.03 & 0.97 & 0.60 & 0.86 \\
\cmidrule{2-7}
 & \multirow{3}{*}{$r_Q$} 
  & None & 0.05 & 0.06 & 0.09 & 0.07 \\
& & Simplified & 0.02 & 0.01 & 0.02 & 0.02 \\
& & Detailed & 0.01 & 0.01 & 0.01 & 0.01 \\
\cmidrule{2-7}
& \multirow{3}{*}{$r_K$} 
  & None & 0.06 & 0.04 & 0.06 & 0.05 \\
& & Simplified & 0.05 & 0.03 & 0.03 & 0.04 \\
& & Detailed & 0.03 & 0.03 & 0.01 & 0.02 \\
\cmidrule{2-7}
 & \multirow{3}{*}{$r_V$} 
  & None & 0.04 & 0.08 & 0.07 & 0.06 \\
& & Simplified & 0.03 & 0.06 & 0.08 & 0.05 \\
& & Detailed & 0.02 & 0.03 & 0.04 & 0.03 \\
\cmidrule{2-7}
& \multirow{3}{*}{$r_O$} 
  & None & 0.03 & 0.06 & 0.05 & 0.05 \\
& & Simplified & 0.01 & 0.04 & 0.08 & 0.04 \\
& & Detailed & 0.01 & 0.02 & 0.05 & 0.02 \\
\bottomrule
\end{tabular}
}
\caption{
Statistical results for StrategyQA using Qwen2-1.5B-Instruct on correct responses.
}
\label{tbl:reasoning_cot_500_qwen2_1_5b_it_grads_strategyqa_train}
\end{table}

\begin{figure*}[!h]
\centering
\includegraphics[width=\linewidth]{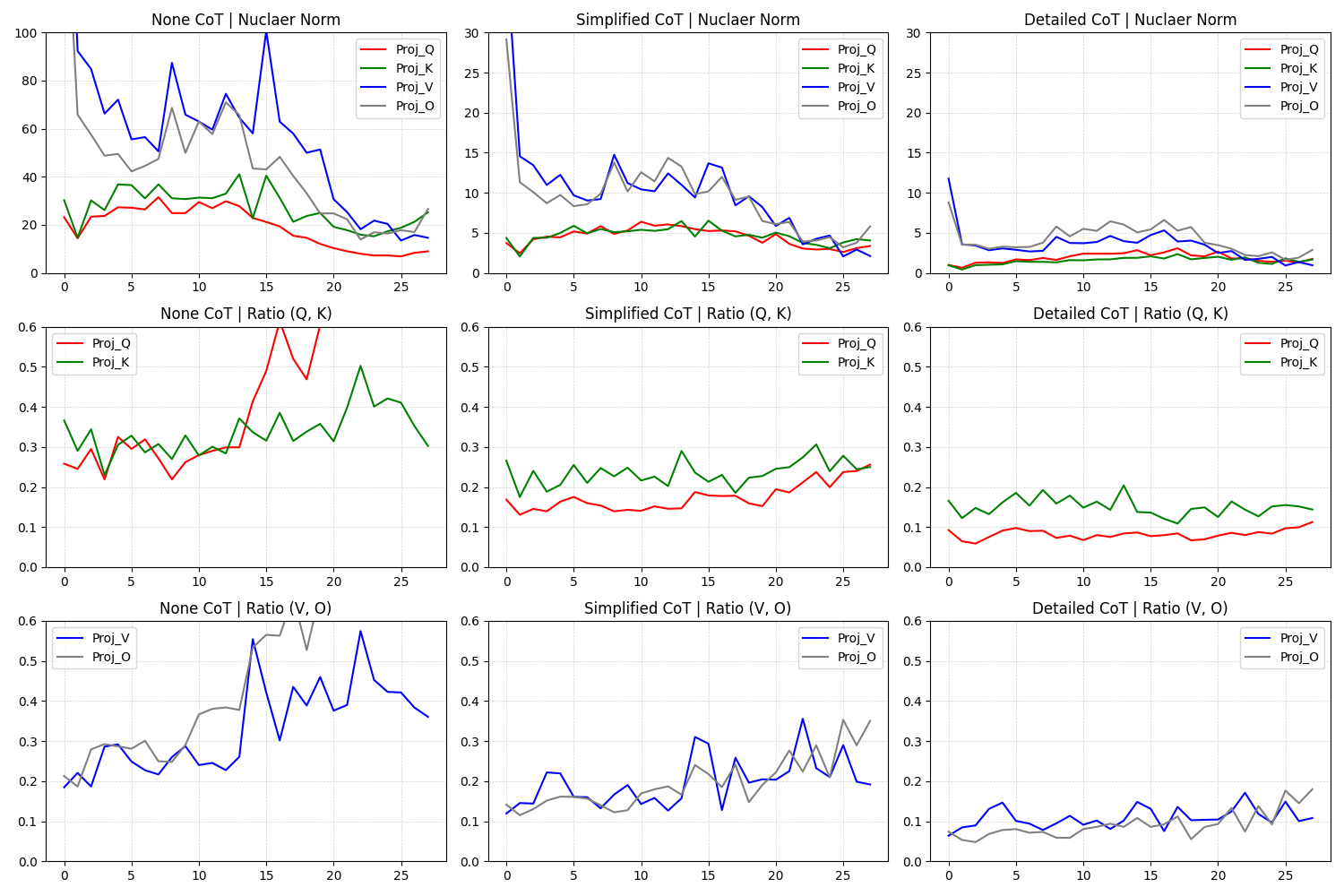}
\caption{Visualization for StrategyQA using Qwen2-1.5B-Instruct on correct responses.}
\label{fig:reasoning_correct_strategyqa_it}
\vspace{-2mm}
\end{figure*}

\begin{table}[!tbh]
\centering
\small
\scalebox{0.75}{
\begin{tabular}{l|l|l|cccc}
\toprule
\multirow{2}{*}{\textbf{Dataset}} & \multirow{2}{*}{\textbf{Curve}} & \multirow{2}{*}{\textbf{Cot}} &  \multicolumn{4}{c}{\textbf{Mean Absolute Difference (MAD)}} \\
 & &  & Early & Middle & Last & All\\
\midrule
\multirow{24}{*}{ECQA} & \multirow{3}{*}{$s_Q$} 
  & None & 7.09 & 4.87 & 3.16 & 5.15 \\
& & Simplified & 1.34 & 0.85 & 1.01 & 1.07 \\
& & Detailed & 0.33 & 0.41 & 0.25 & 0.38 \\
\cmidrule{2-7}
& \multirow{3}{*}{$s_K$} 
  & None & 9.03 & 6.21 & 12.09 & 7.84 \\
& & Simplified & 1.75 & 1.19 & 1.75 & 1.42 \\
& & Detailed & 0.30 & 0.27 & 0.51 & 0.33 \\
\cmidrule{2-7}
& \multirow{3}{*}{$s_V$} 
  & None & 42.20 & 21.30 & 4.66 & 21.30 \\
& & Simplified & 8.69 & 3.82 & 1.97 & 4.27 \\
& & Detailed & 1.65 & 0.81 & 0.58 & 0.90 \\
\cmidrule{2-7}
& \multirow{3}{*}{$s_O$} 
  & None & 30.30 & 19.50 & 3.53 & 17.11 \\
& & Simplified & 6.03 & 3.86 & 1.23 & 3.59 \\
& & Detailed & 1.09 & 1.07 & 0.60 & 0.94 \\
\cmidrule{2-7}
 & \multirow{3}{*}{$r_Q$} 
  & None & 0.02 & 0.07 & 0.13 & 0.07 \\
& & Simplified & 0.02 & 0.01 & 0.04 & 0.02 \\
& & Detailed & 0.01 & 0.01 & 0.01 & 0.01 \\
\cmidrule{2-7}
& \multirow{3}{*}{$r_K$} 
  & None & 0.04 & 0.06 & 0.08 & 0.06 \\
& & Simplified & 0.03 & 0.05 & 0.02 & 0.04 \\
& & Detailed & 0.02 & 0.03 & 0.01 & 0.02 \\
\cmidrule{2-7}
 & \multirow{3}{*}{$r_V$} 
  & None & 0.04 & 0.04 & 0.07 & 0.05 \\
& & Simplified & 0.03 & 0.04 & 0.03 & 0.03 \\
& & Detailed & 0.03 & 0.02 & 0.03 & 0.02 \\
\cmidrule{2-7}
& \multirow{3}{*}{$r_O$} 
  & None & 0.02 & 0.07 & 0.12 & 0.07 \\
& & Simplified & 0.02 & 0.03 & 0.06 & 0.04 \\
& & Detailed & 0.01 & 0.02 & 0.05 & 0.03 \\
\bottomrule
\end{tabular}
}
\caption{
Statistical results for ECQA using Qwen2-1.5B-Instruct on correct responses.
}
\label{tbl:reasoning_cot_500_qwen2_1_5b_it_grads_ecqa_train}
\end{table}

\begin{figure*}[!h]
\centering
\includegraphics[width=\linewidth]{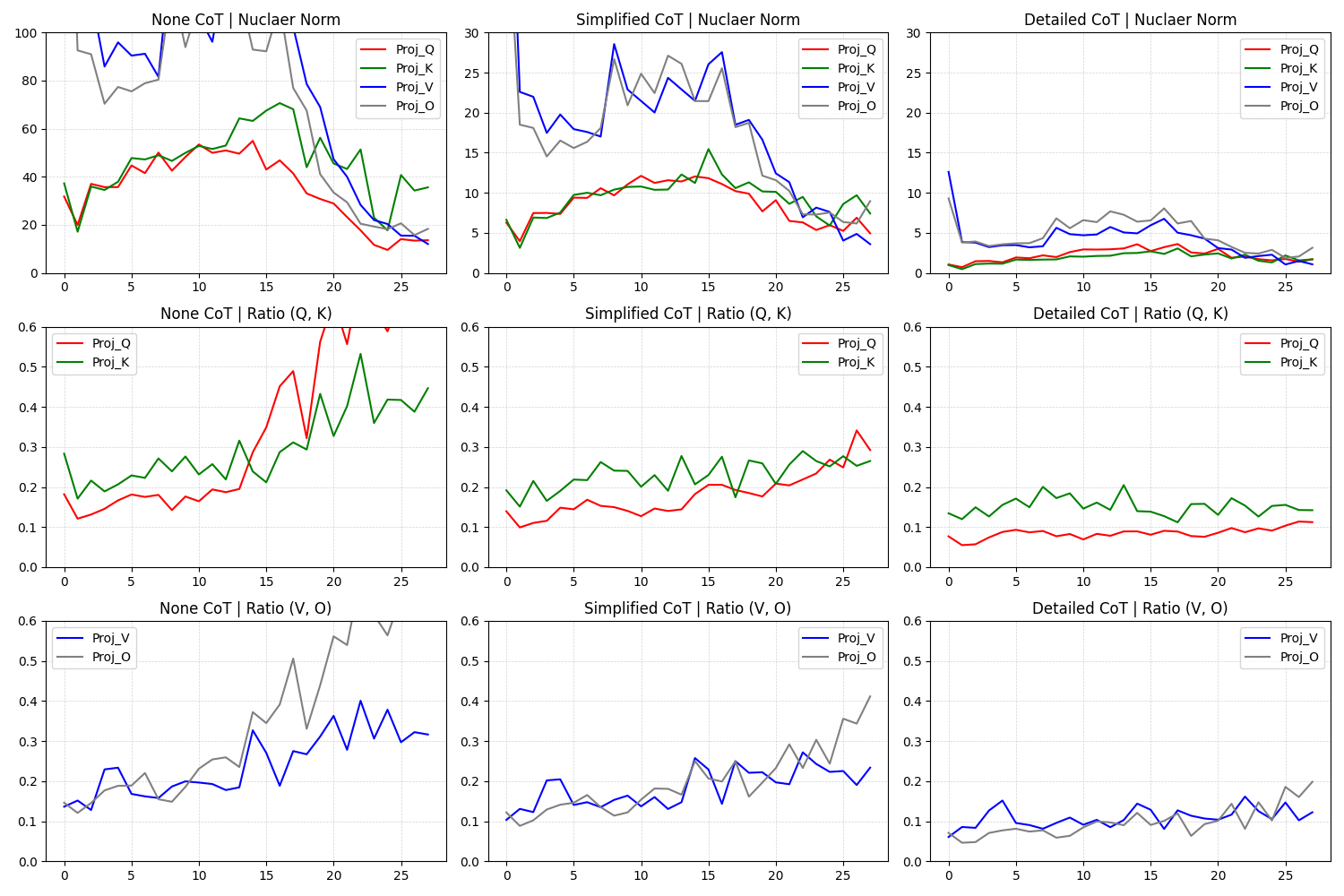}
\caption{Visualization for ECQA using Qwen2-1.5B-Instruct on correct responses.}
\label{fig:reasoning_correct_ecqa_it}
\vspace{-2mm}
\end{figure*}

\begin{table}[!tbh]
\centering
\small
\scalebox{0.75}{
\begin{tabular}{l|l|l|cccc}
\toprule
\multirow{2}{*}{\textbf{Dataset}} & \multirow{2}{*}{\textbf{Curve}} & \multirow{2}{*}{\textbf{Cot}} &  \multicolumn{4}{c}{\textbf{Mean Absolute Difference (MAD)}} \\
 & &  & Early & Middle & Last & All\\
\midrule
\multirow{24}{*}{CREAK} & \multirow{3}{*}{$s_Q$} 
  & None & 9.33 & 8.89 & 5.25 & 7.94 \\
& & Simplified & 1.79 & 1.14 & 1.94 & 1.57 \\
& & Detailed & 0.32 & 0.39 & 0.22 & 0.36 \\
\cmidrule{2-7}
& \multirow{3}{*}{$s_K$} 
  & None & 12.71 & 19.58 & 15.92 & 15.92 \\
& & Simplified & 2.27 & 1.97 & 2.13 & 2.07 \\
& & Detailed & 0.31 & 0.25 & 0.47 & 0.31 \\
\cmidrule{2-7}
& \multirow{3}{*}{$s_V$} 
  & None & 64.10 & 28.13 & 7.38 & 30.21 \\
& & Simplified & 11.90 & 5.69 & 2.98 & 6.11 \\
& & Detailed & 1.82 & 0.85 & 0.60 & 0.95 \\
\cmidrule{2-7}
& \multirow{3}{*}{$s_O$} 
  & None & 42.64 & 26.70 & 3.97 & 23.80 \\
& & Simplified & 8.02 & 5.24 & 1.66 & 4.80 \\
& & Detailed & 1.17 & 1.09 & 0.59 & 0.96 \\
\cmidrule{2-7}
 & \multirow{3}{*}{$r_Q$} 
  & None & 0.03 & 0.08 & 0.15 & 0.08 \\
& & Simplified & 0.02 & 0.02 & 0.06 & 0.03 \\
& & Detailed & 0.01 & 0.01 & 0.01 & 0.01 \\
\cmidrule{2-7}
& \multirow{3}{*}{$r_K$} 
  & None & 0.03 & 0.07 & 0.08 & 0.06 \\
& & Simplified & 0.03 & 0.04 & 0.03 & 0.03 \\
& & Detailed & 0.03 & 0.02 & 0.01 & 0.02 \\
\cmidrule{2-7}
 & \multirow{3}{*}{$r_V$} 
  & None & 0.04 & 0.06 & 0.08 & 0.06 \\
& & Simplified & 0.03 & 0.05 & 0.05 & 0.04 \\
& & Detailed & 0.02 & 0.03 & 0.04 & 0.03 \\
\cmidrule{2-7}
& \multirow{3}{*}{$r_O$} 
  & None & 0.02 & 0.06 & 0.10 & 0.06 \\
& & Simplified & 0.02 & 0.03 & 0.07 & 0.04 \\
& & Detailed & 0.01 & 0.02 & 0.05 & 0.02 \\
\bottomrule
\end{tabular}
}
\caption{
Statistical results for CREAK using Qwen2-1.5B-Instruct on correct responses.
}
\label{tbl:reasoning_cot_500_qwen2_1_5b_it_grads_creak_train}
\end{table}

\begin{figure*}[!h]
\centering
\includegraphics[width=\linewidth]{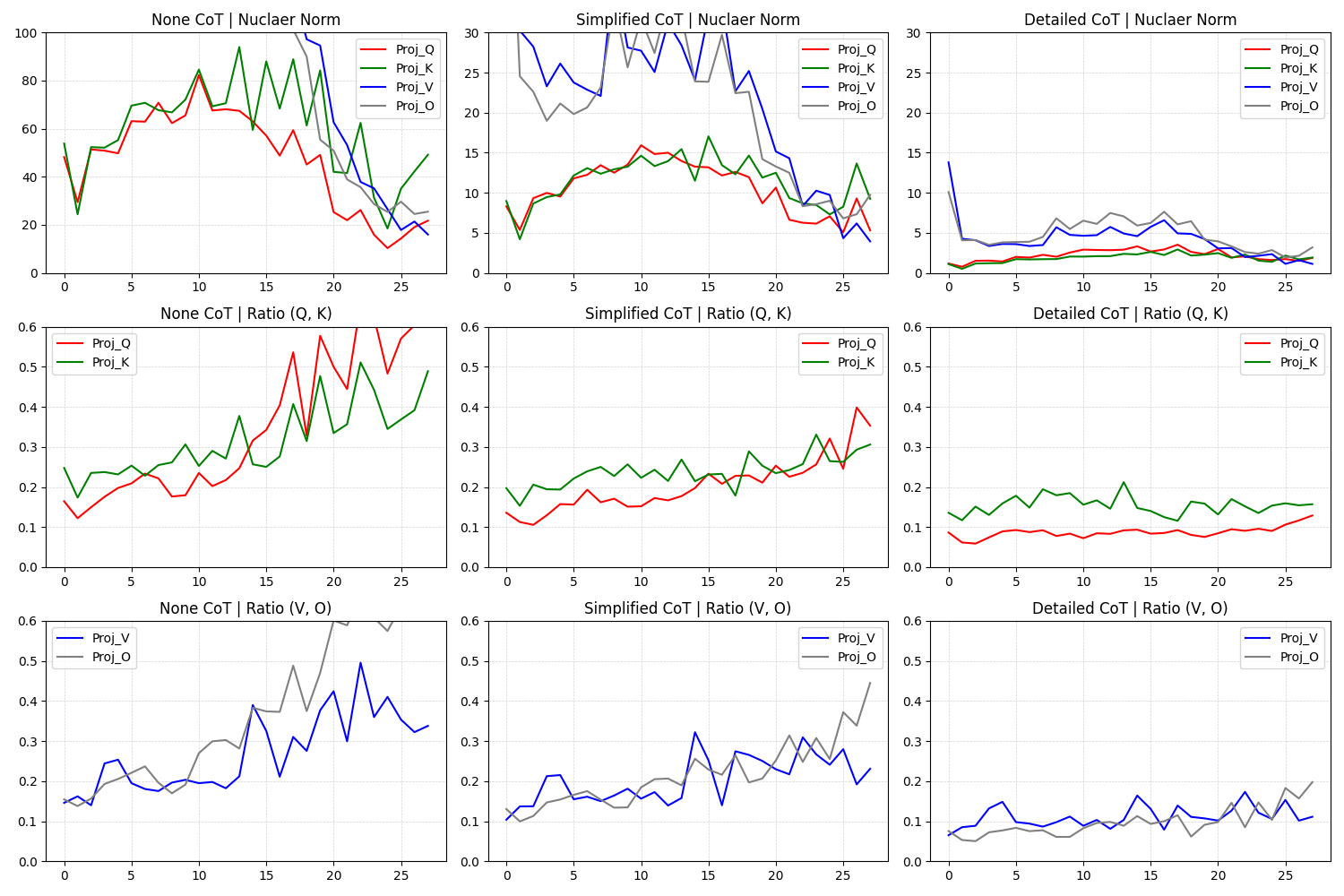}
\caption{Visualization for CREAK using Qwen2-1.5B-Instruct on correct responses.}
\label{fig:reasoning_correct_creak_it}
\vspace{-2mm}
\end{figure*}

\begin{table}[!tbh]
\centering
\small
\scalebox{0.75}{
\begin{tabular}{l|l|l|cccc}
\toprule
\multirow{2}{*}{\textbf{Dataset}} & \multirow{2}{*}{\textbf{Curve}} & \multirow{2}{*}{\textbf{Cot}} &  \multicolumn{4}{c}{\textbf{Mean Absolute Difference (MAD)}} \\
 & &  & Early & Middle & Last & All\\
\midrule
\multirow{24}{*}{Sensemaking} & \multirow{3}{*}{$s_Q$} 
  & None & 0.70 & 0.72 & 0.40 & 0.66 \\
& & Simplified & 2.08 & 1.35 & 2.43 & 1.86 \\
& & Detailed & 0.31 & 0.39 & 0.34 & 0.38 \\
\cmidrule{2-7}
& \multirow{3}{*}{$s_K$} 
  & None & 0.91 & 0.51 & 0.83 & 0.71 \\
& & Simplified & 2.65 & 2.32 & 2.27 & 2.38 \\
& & Detailed & 0.27 & 0.31 & 0.60 & 0.37 \\
\cmidrule{2-7}
& \multirow{3}{*}{$s_V$} 
  & None & 3.89 & 1.78 & 0.32 & 1.82 \\
& & Simplified & 12.02 & 6.20 & 3.13 & 6.45 \\
& & Detailed & 1.54 & 0.77 & 0.48 & 0.83 \\
\cmidrule{2-7}
& \multirow{3}{*}{$s_O$} 
  & None & 2.88 & 1.62 & 0.38 & 1.56 \\
& & Simplified & 8.04 & 5.68 & 1.98 & 5.11 \\
& & Detailed & 1.08 & 0.91 & 0.49 & 0.83 \\
\cmidrule{2-7}
 & \multirow{3}{*}{$r_Q$} 
  & None & 0.03 & 0.05 & 0.18 & 0.08 \\
& & Simplified & 0.02 & 0.02 & 0.08 & 0.03 \\
& & Detailed & 0.01 & 0.01 & 0.01 & 0.01 \\
\cmidrule{2-7}
& \multirow{3}{*}{$r_K$} 
  & None & 0.03 & 0.05 & 0.13 & 0.06 \\
& & Simplified & 0.03 & 0.05 & 0.03 & 0.04 \\
& & Detailed & 0.02 & 0.03 & 0.02 & 0.03 \\
\cmidrule{2-7}
 & \multirow{3}{*}{$r_V$} 
  & None & 0.03 & 0.04 & 0.05 & 0.04 \\
& & Simplified & 0.03 & 0.04 & 0.04 & 0.04 \\
& & Detailed & 0.03 & 0.02 & 0.03 & 0.02 \\
\cmidrule{2-7}
& \multirow{3}{*}{$r_O$} 
  & None & 0.02 & 0.06 & 0.10 & 0.06 \\
& & Simplified & 0.02 & 0.03 & 0.07 & 0.04 \\
& & Detailed & 0.01 & 0.02 & 0.05 & 0.02 \\
\bottomrule
\end{tabular}
}
\caption{
Statistical results for Sensemaking using Qwen2-1.5B-Instruct on correct responses.
}
\label{tbl:reasoning_cot_500_qwen2_1_5b_it_grads_sensemaking_train}
\end{table}

\begin{figure*}[!h]
\centering
\includegraphics[width=\linewidth]{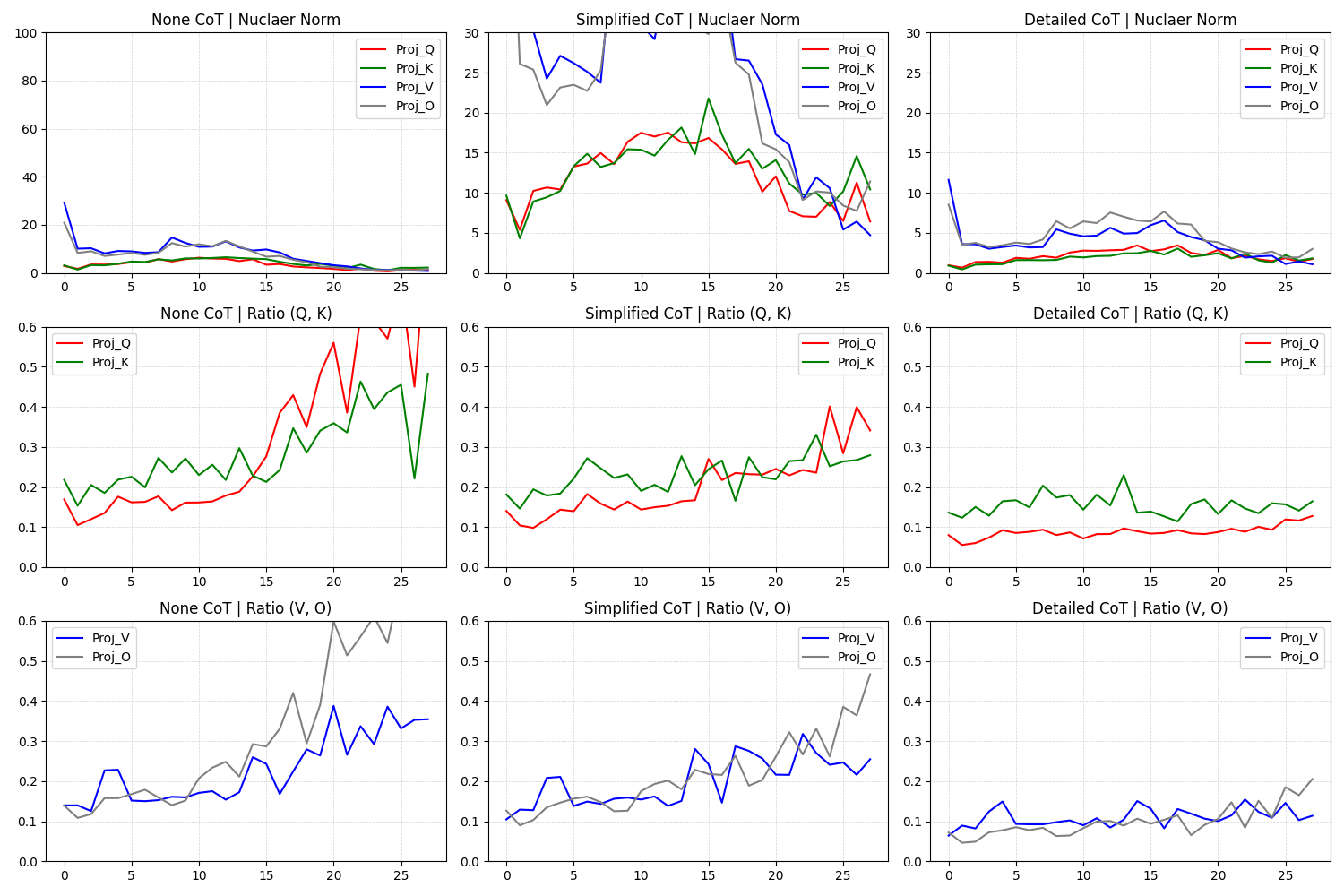}
\caption{Visualization for Sensemaking using Qwen2-1.5B-Instruct on correct responses.}
\label{fig:reasoning_correct_sensemaking_it}
\vspace{-2mm}
\end{figure*}

\subsubsection{Wiki Tasks}

The visualizations and statistical results on Wiki tasks are shown in Figure \ref{fig:wiki_correct_it} and Table \ref{tbl:wiki_wiki_popularity_qwen2_1_5b_it_grads}.

\begin{table}[!tbh]
\centering
\small
\scalebox{0.75}{
\begin{tabular}{l|l|l|cccc}
\toprule
\multirow{2}{*}{\textbf{Dataset}} & \multirow{2}{*}{\textbf{Curve}} & \multirow{2}{*}{\textbf{Cot}} &  \multicolumn{4}{c}{\textbf{Mean Absolute Difference (MAD)}} \\
 & &  & Early & Middle & Last & All\\
\midrule
\multirow{32}{*}{Wiki} & \multirow{3}{*}{$s_Q$} 
  & Len 100 & 0.54 & 0.37 & 0.34 & 0.41 \\
& & Len 500 & 0.37 & 0.33 & 0.24 & 0.32 \\
& & Len 1000 & 0.32 & 0.31 & 0.22 & 0.30 \\
& & Unpopular & 0.58 & 0.47 & 0.24 & 0.43 \\
\cmidrule{2-7}
& \multirow{3}{*}{$s_K$} 
  & Len 100 & 0.66 & 0.28 & 0.37 & 0.40 \\
& & Len 500 & 0.36 & 0.13 & 0.29 & 0.22 \\
& & Len 1000 & 0.27 & 0.12 & 0.32 & 0.19 \\
& & Unpopular & 0.74 & 0.60 & 0.34 & 0.54 \\
\cmidrule{2-7}
& \multirow{3}{*}{$s_V$} 
  & Len 100 & 3.94 & 1.70 & 1.13 & 2.03 \\
& & Len 500 & 2.45 & 1.01 & 0.78 & 1.26 \\
& & Len 1000 & 1.91 & 0.74 & 0.59 & 0.95 \\
& & Unpopular & 6.00 & 1.94 & 1.03 & 2.57 \\
\cmidrule{2-7}
& \multirow{3}{*}{$s_O$} 
  & Len 100 & 2.68 & 1.66 & 0.86 & 1.66 \\
& & Len 500 & 1.76 & 1.24 & 0.68 & 1.18 \\
& & Len 1000 & 1.44 & 1.04 & 0.57 & 0.98 \\
& & Unpopular & 4.09 & 1.76 & 0.80 & 1.98 \\
\cmidrule{2-7}
 & \multirow{3}{*}{$r_Q$} 
  & Len 100 & 0.02 & 0.01 & 0.03 & 0.02 \\
& & Len 500 & 0.01 & 0.01 & 0.01 & 0.01 \\
& & Len 1000 & 0.01 & 0.01 & 0.01 & 0.01 \\
& & Unpopular & 0.02 & 0.02 & 0.03 & 0.02 \\
\cmidrule{2-7}
& \multirow{3}{*}{$r_K$} 
  & Len 100 & 0.03 & 0.02 & 0.02 & 0.03 \\
& & Len 500 & 0.02 & 0.02 & 0.02 & 0.02 \\
& & Len 1000 & 0.02 & 0.02 & 0.02 & 0.02 \\
& & Unpopular & 0.05 & 0.03 & 0.04 & 0.04 \\
\cmidrule{2-7}
 & \multirow{3}{*}{$r_V$} 
  & Len 100 & 0.03 & 0.06 & 0.09 & 0.06 \\
& & Len 500 & 0.02 & 0.03 & 0.05 & 0.03 \\
& & Len 1000 & 0.02 & 0.03 & 0.04 & 0.03 \\
& & Unpopular & 0.04 & 0.06 & 0.07 & 0.05 \\
\cmidrule{2-7}
& \multirow{3}{*}{$r_O$} 
  & Len 100 & 0.02 & 0.03 & 0.07 & 0.04 \\
& & Len 500 & 0.01 & 0.02 & 0.05 & 0.03 \\
& & Len 1000 & 0.01 & 0.01 & 0.05 & 0.02 \\
& & Unpopular & 0.01 & 0.04 & 0.09 & 0.05 \\
\bottomrule
\end{tabular}
}
\caption{
Statistical results for Wiki using Qwen2-1.5B-Instruct on correct responses.
}
\label{tbl:wiki_wiki_popularity_qwen2_1_5b_it_new_grads}
\end{table}

\begin{figure*}[!h]
\centering
\includegraphics[width=\linewidth]{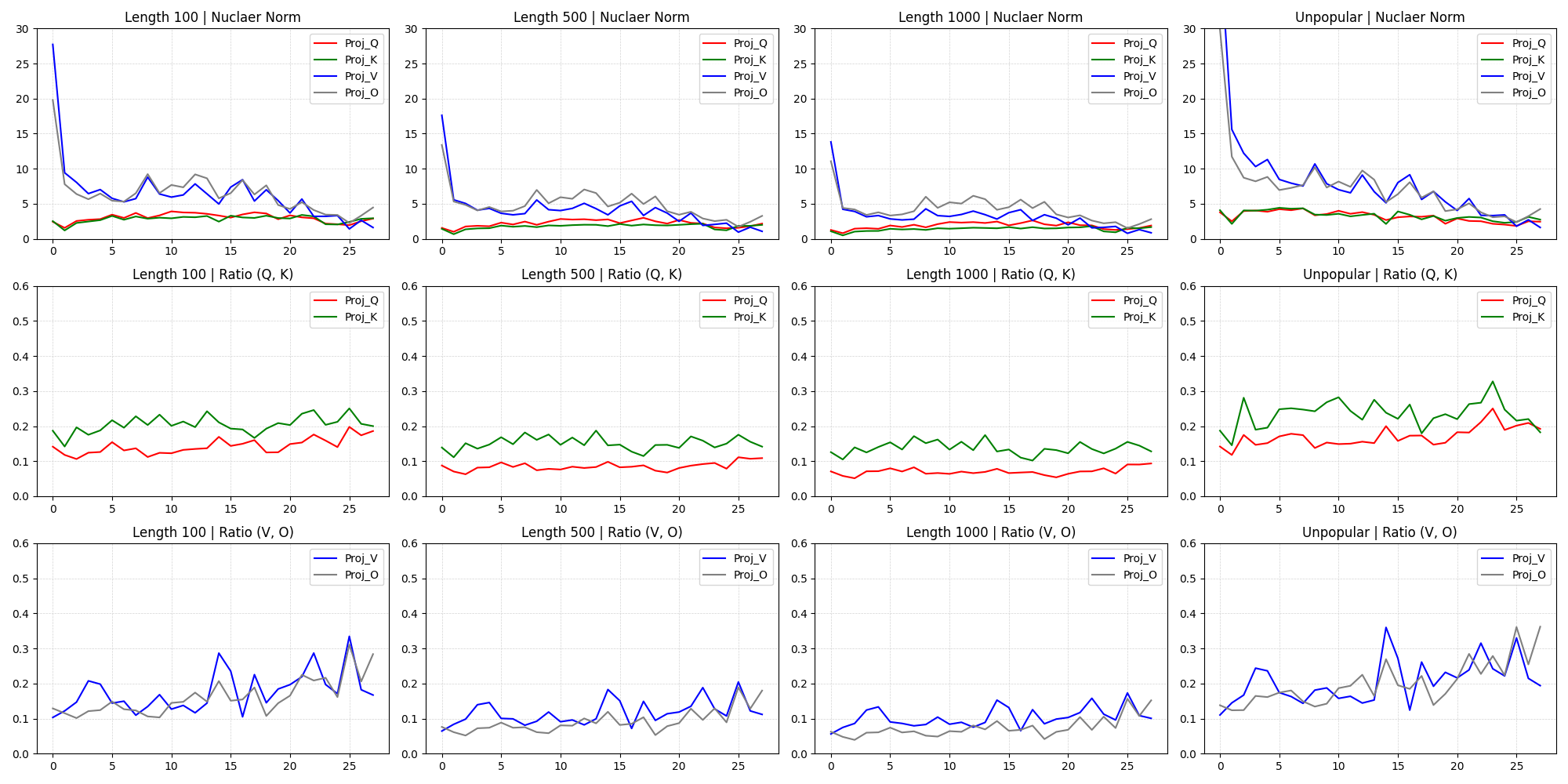}
\caption{Visualization for Wiki tasks using Qwen2-1.5B-Instruct on correct responses.}
\label{fig:wiki_correct_it}
\vspace{-2mm}
\end{figure*}

\subsection{Instructed LLM on Irrelevant Responses}

\subsubsection{Reasoning Tasks}

The visualizations and statistical results on MATH tasks: MATH-Algebra (Figure \ref{fig:math_nonsense_algebra_it}, Table \ref{tbl:math_lighteval_math_wrong_answer_shuffle_qwen2_1_5b_it_grads_algebra}), MATH-Counting (Figure \ref{fig:math_nonsense_counting_it}, Table \ref{tbl:math_lighteval_math_wrong_answer_shuffle_qwen2_1_5b_it_grads_counting_probability}), MATH-Geometry (Figure \ref{fig:math_nonsense_geometry_it}, Table \ref{tbl:math_lighteval_math_wrong_answer_shuffle_qwen2_1_5b_it_grads_geometry}).

\begin{table}[!tbh]
\centering
\small
\scalebox{0.75}{
\begin{tabular}{l|l|l|cccc}
\toprule
\multirow{2}{*}{\textbf{Dataset}} & \multirow{2}{*}{\textbf{Curve}} & \multirow{2}{*}{\textbf{Cot}} &  \multicolumn{4}{c}{\textbf{Mean Absolute Difference (MAD)}} \\
 & &  & Early & Middle & Last & All\\
\midrule
\multirow{16}{*}{Algebra} & \multirow{3}{*}{$s_Q$} 
  & Simplified & 0.71 & 0.78 & 1.27 & 0.90 \\
& & Detailed & 0.42 & 0.51 & 0.59 & 0.51 \\
\cmidrule{2-7}
& \multirow{3}{*}{$s_K$} 
  & Simplified & 0.71 & 0.52 & 1.27 & 0.71 \\
& & Detailed & 0.40 & 0.30 & 0.64 & 0.40 \\
\cmidrule{2-7}
& \multirow{3}{*}{$s_V$} 
  & Simplified & 3.35 & 2.29 & 1.01 & 2.10 \\
& & Detailed & 1.74 & 1.02 & 0.54 & 1.01 \\
\cmidrule{2-7}
& \multirow{3}{*}{$s_O$} 
  & Simplified & 2.48 & 2.53 & 0.72 & 1.97 \\
& & Detailed & 1.34 & 1.27 & 0.42 & 1.02 \\
\cmidrule{2-7}
 & \multirow{3}{*}{$r_Q$} 
  & Simplified & 0.01 & 0.02 & 0.05 & 0.03 \\
& & Detailed & 0.01 & 0.02 & 0.04 & 0.02 \\
\cmidrule{2-7}
& \multirow{3}{*}{$r_K$} 
  & Simplified & 0.02 & 0.03 & 0.03 & 0.03 \\
& & Detailed & 0.02 & 0.04 & 0.03 & 0.03 \\
\cmidrule{2-7}
 & \multirow{3}{*}{$r_V$} 
  & Simplified & 0.03 & 0.04 & 0.04 & 0.03 \\
& & Detailed & 0.03 & 0.03 & 0.03 & 0.03 \\
\cmidrule{2-7}
& \multirow{3}{*}{$r_O$} 
  & Simplified & 0.01 & 0.02 & 0.06 & 0.03 \\
& & Detailed & 0.01 & 0.01 & 0.06 & 0.02 \\
\bottomrule
\end{tabular}
}
\caption{
Statistical results for MATH-Algebra using Qwen2-1.5B-Instruct on irrelevant responses.
}
\label{tbl:math_lighteval_math_wrong_answer_shuffle_qwen2_1_5b_it_grads_algebra}
\end{table}

\begin{figure}[!h]
\centering
\includegraphics[width=\linewidth]{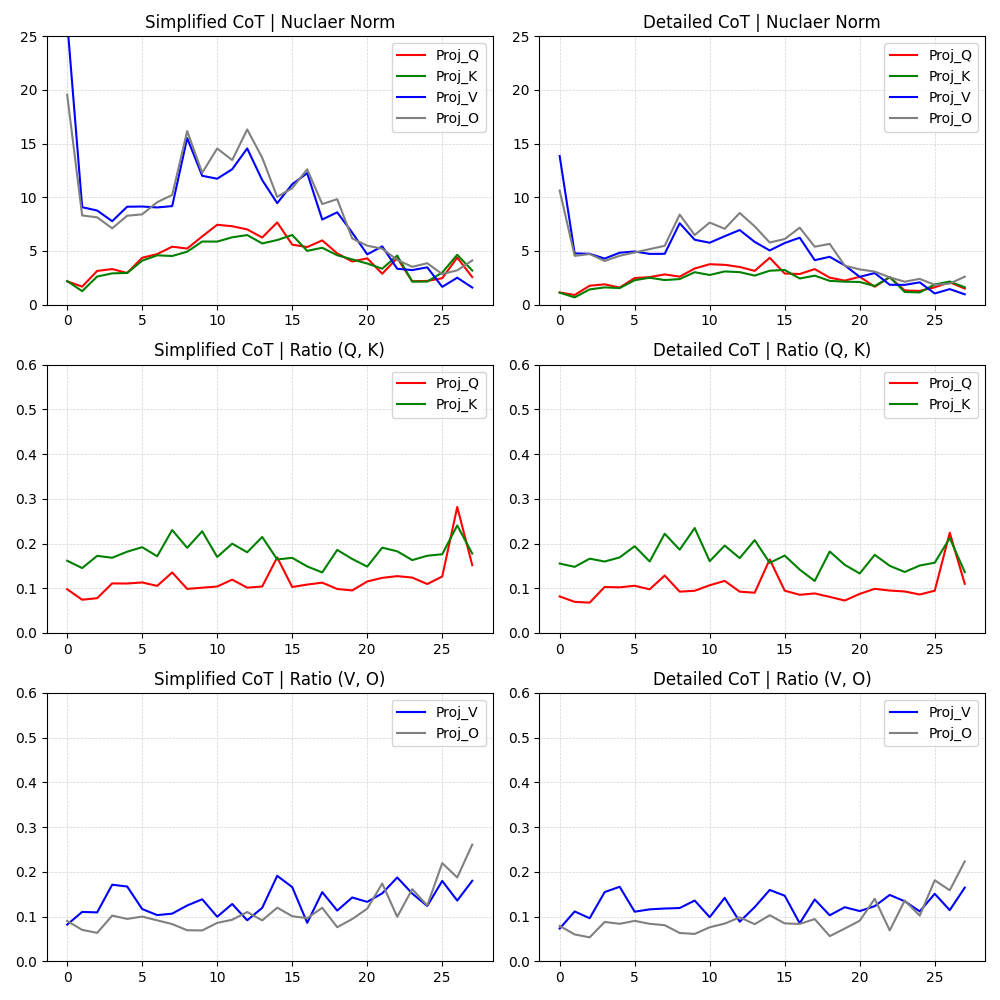}
\caption{Visualization for MATH-Algebra using Qwen2-1.5B-Instruct on irrelevant responses.}
\label{fig:math_nonsense_algebra_it}
\vspace{-2mm}
\end{figure}

\begin{table}[!tbh]
\centering
\small
\scalebox{0.75}{
\begin{tabular}{l|l|l|cccc}
\toprule
\multirow{2}{*}{\textbf{Dataset}} & \multirow{2}{*}{\textbf{Curve}} & \multirow{2}{*}{\textbf{Cot}} &  \multicolumn{4}{c}{\textbf{Mean Absolute Difference (MAD)}} \\
 & &  & Early & Middle & Last & All\\
\midrule
\multirow{16}{*}{Counting} & \multirow{3}{*}{$s_Q$} 
  & Simplified & 0.73 & 0.79 & 1.05 & 0.85 \\
& & Detailed & 0.47 & 0.52 & 0.47 & 0.50 \\
\cmidrule{2-7}
& \multirow{3}{*}{$s_K$} 
  & Simplified & 0.72 & 0.43 & 1.06 & 0.61 \\
& & Detailed & 0.43 & 0.29 & 0.55 & 0.38 \\
\cmidrule{2-7}
& \multirow{3}{*}{$s_V$} 
  & Simplified & 3.50 & 2.25 & 1.08 & 2.14 \\
& & Detailed & 1.96 & 0.96 & 0.58 & 1.04 \\
\cmidrule{2-7}
& \multirow{3}{*}{$s_O$} 
  & Simplified & 2.66 & 2.43 & 0.84 & 1.98 \\
& & Detailed & 1.48 & 1.24 & 0.50 & 1.06 \\
\cmidrule{2-7}
 & \multirow{3}{*}{$r_Q$} 
  & Simplified & 0.01 & 0.02 & 0.05 & 0.02 \\
& & Detailed & 0.01 & 0.02 & 0.04 & 0.02 \\
\cmidrule{2-7}
& \multirow{3}{*}{$r_K$} 
  & Simplified & 0.02 & 0.03 & 0.02 & 0.03 \\
& & Detailed & 0.02 & 0.04 & 0.03 & 0.03 \\
\cmidrule{2-7}
 & \multirow{3}{*}{$r_V$} 
  & Simplified & 0.02 & 0.04 & 0.04 & 0.04 \\
& & Detailed & 0.03 & 0.03 & 0.03 & 0.03 \\
\cmidrule{2-7}
& \multirow{3}{*}{$r_O$} 
  & Simplified & 0.01 & 0.02 & 0.06 & 0.03 \\
& & Detailed & 0.01 & 0.01 & 0.05 & 0.02 \\
\bottomrule
\end{tabular}
}
\caption{
Statistical results for MATH-Counting using Qwen2-1.5B-Instruct on irrelevant responses.
}
\label{tbl:math_lighteval_math_wrong_answer_shuffle_qwen2_1_5b_it_grads_counting_probability}
\end{table}

\begin{figure}[!h]
\centering
\includegraphics[width=\linewidth]{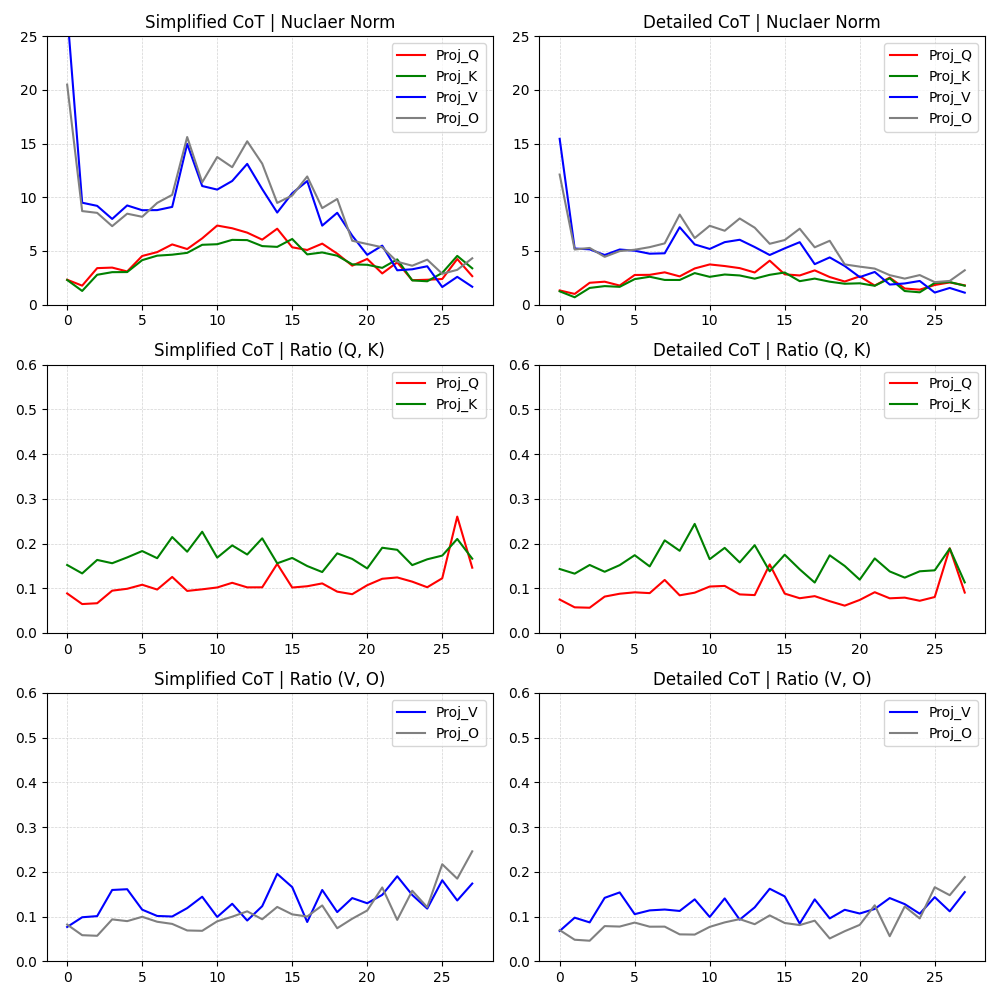}
\caption{Visualization for MATH-Counting using Qwen2-1.5B-Instruct on irrelevant responses.}
\label{fig:math_nonsense_counting_it}
\vspace{-2mm}
\end{figure}

\begin{table}[!tbh]
\centering
\small
\scalebox{0.75}{
\begin{tabular}{l|l|l|cccc}
\toprule
\multirow{2}{*}{\textbf{Dataset}} & \multirow{2}{*}{\textbf{Curve}} & \multirow{2}{*}{\textbf{Cot}} &  \multicolumn{4}{c}{\textbf{Mean Absolute Difference (MAD)}} \\
 & &  & Early & Middle & Last & All\\
\midrule
\multirow{16}{*}{Geometry} & \multirow{3}{*}{$s_Q$} 
  & Simplified & 0.62 & 0.69 & 1.20 & 0.80 \\
& & Detailed & 0.57 & 0.48 & 0.66 & 0.54 \\
\cmidrule{2-7}
& \multirow{3}{*}{$s_K$} 
  & Simplified & 0.58 & 0.41 & 1.26 & 0.63 \\
& & Detailed & 0.49 & 0.25 & 0.85 & 0.45 \\
\cmidrule{2-7}
& \multirow{3}{*}{$s_V$} 
  & Simplified & 3.09 & 1.82 & 0.94 & 1.81 \\
& & Detailed & 2.41 & 1.00 & 0.60 & 1.19 \\
\cmidrule{2-7}
& \multirow{3}{*}{$s_O$} 
  & Simplified & 2.43 & 2.26 & 0.64 & 1.79 \\
& & Detailed & 1.89 & 1.48 & 0.50 & 1.26 \\
\cmidrule{2-7}
 & \multirow{3}{*}{$r_Q$} 
  & Simplified & 0.01 & 0.01 & 0.04 & 0.02 \\
& & Detailed & 0.01 & 0.02 & 0.02 & 0.02 \\
\cmidrule{2-7}
& \multirow{3}{*}{$r_K$} 
  & Simplified & 0.01 & 0.03 & 0.02 & 0.03 \\
& & Detailed & 0.02 & 0.03 & 0.05 & 0.03 \\
\cmidrule{2-7}
 & \multirow{3}{*}{$r_V$} 
  & Simplified & 0.02 & 0.03 & 0.03 & 0.03 \\
& & Detailed & 0.03 & 0.03 & 0.03 & 0.03 \\
\cmidrule{2-7}
& \multirow{3}{*}{$r_O$} 
  & Simplified & 0.01 & 0.01 & 0.05 & 0.02 \\
& & Detailed & 0.01 & 0.01 & 0.05 & 0.02 \\
\bottomrule
\end{tabular}
}
\caption{
Statistical results for MATH-Geometry using Qwen2-1.5B-Instruct on irrelevant responses.
}
\label{tbl:math_lighteval_math_wrong_answer_shuffle_qwen2_1_5b_it_grads_geometry}
\end{table}

\begin{figure}[!h]
\centering
\includegraphics[width=\linewidth]{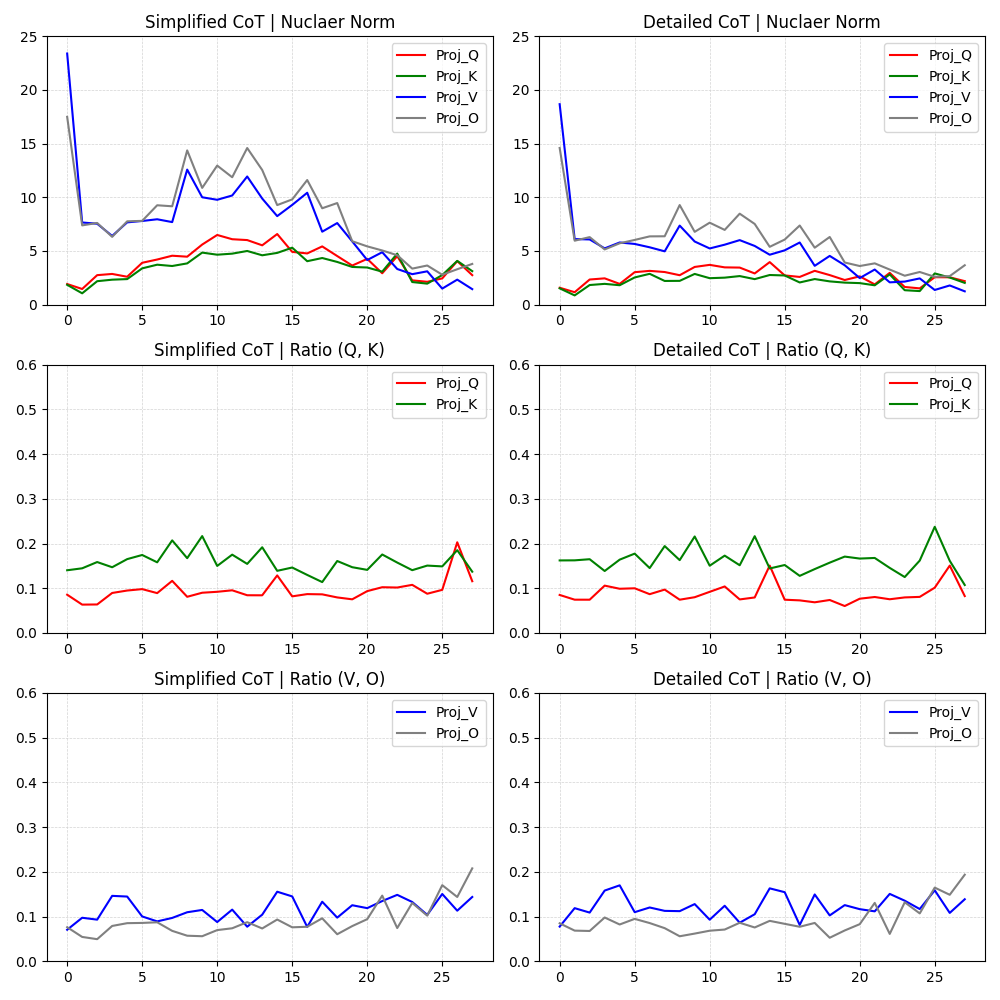}
\caption{Visualization for MATH-Geometry using Qwen2-1.5B-Instruct on irrelevant responses.}
\label{fig:math_nonsense_geometry_it}
\vspace{-2mm}
\end{figure}

The visualizations and statistical results on other reasoning tasks: AQuA (Figure \ref{fig:reasoning_nonsense_aqua_it}, Table \ref{tbl:reasoning_cot_500_wrong_answer_shuffle_qwen2_1_5b_it_grads_aqua_train}), GSM8K (Figure \ref{fig:reasoning_nonsense_gsm8k_it}, Table \ref{tbl:reasoning_cot_500_wrong_answer_shuffle_qwen2_1_5b_it_grads_gsm8k_train}), StrategyQA (Figure \ref{fig:reasoning_nonsense_strategyqa_it}, Table \ref{tbl:reasoning_cot_500_wrong_answer_shuffle_qwen2_1_5b_it_grads_strategyqa_train}), ECQA (Figure \ref{fig:reasoning_nonsense_ecqa_it}, Table \ref{tbl:reasoning_cot_500_wrong_answer_shuffle_qwen2_1_5b_it_grads_ecqa_train}), CREAK (Figure \ref{fig:reasoning_nonsense_creak_it}, Table \ref{tbl:reasoning_cot_500_wrong_answer_shuffle_qwen2_1_5b_it_grads_creak_train}), Sensemaking (Figure \ref{fig:reasoning_nonsense_sensemaking_it}, Table \ref{tbl:reasoning_cot_500_wrong_answer_shuffle_qwen2_1_5b_it_grads_sensemaking_train}).

\begin{table}[!tbh]
\centering
\small
\scalebox{0.75}{
\begin{tabular}{l|l|l|cccc}
\toprule
\multirow{2}{*}{\textbf{Dataset}} & \multirow{2}{*}{\textbf{Curve}} & \multirow{2}{*}{\textbf{Cot}} &  \multicolumn{4}{c}{\textbf{Mean Absolute Difference (MAD)}} \\
 & &  & Early & Middle & Last & All\\
\midrule
\multirow{24}{*}{AQuA} & \multirow{3}{*}{$s_Q$} 
  & None & 12.33 & 6.91 & 3.27 & 7.57 \\
& & Simplified & 1.13 & 1.01 & 1.55 & 1.19 \\
& & Detailed & 0.54 & 0.53 & 0.47 & 0.53 \\
\cmidrule{2-7}
& \multirow{3}{*}{$s_K$} 
  & None & 15.89 & 12.20 & 7.05 & 13.20 \\
& & Simplified & 1.37 & 0.72 & 1.82 & 1.10 \\
& & Detailed & 0.54 & 0.34 & 0.59 & 0.43 \\
\cmidrule{2-7}
& \multirow{3}{*}{$s_V$} 
  & None & 66.05 & 31.50 & 4.98 & 32.48 \\
& & Simplified & 6.27 & 3.21 & 1.60 & 3.35 \\
& & Detailed & 2.42 & 1.21 & 0.76 & 1.32 \\
\cmidrule{2-7}
& \multirow{3}{*}{$s_O$} 
  & None & 45.04 & 24.42 & 3.74 & 23.06 \\
& & Simplified & 4.58 & 3.30 & 1.09 & 2.90 \\
& & Detailed & 1.77 & 1.44 & 0.58 & 1.24 \\
\cmidrule{2-7}
 & \multirow{3}{*}{$r_Q$} 
  & None & 0.04 & 0.08 & 0.18 & 0.10 \\
& & Simplified & 0.02 & 0.01 & 0.04 & 0.02 \\
& & Detailed & 0.01 & 0.02 & 0.03 & 0.02 \\
\cmidrule{2-7}
& \multirow{3}{*}{$r_K$} 
  & None & 0.05 & 0.05 & 0.13 & 0.07 \\
& & Simplified & 0.03 & 0.04 & 0.03 & 0.03 \\
& & Detailed & 0.02 & 0.04 & 0.03 & 0.03 \\
\cmidrule{2-7}
 & \multirow{3}{*}{$r_V$} 
  & None & 0.04 & 0.04 & 0.09 & 0.05 \\
& & Simplified & 0.03 & 0.05 & 0.05 & 0.04 \\
& & Detailed & 0.03 & 0.04 & 0.04 & 0.03 \\
\cmidrule{2-7}
& \multirow{3}{*}{$r_O$} 
  & None & 0.03 & 0.06 & 0.11 & 0.06 \\
& & Simplified & 0.01 & 0.02 & 0.08 & 0.03 \\
& & Detailed & 0.01 & 0.01 & 0.06 & 0.02 \\
\bottomrule
\end{tabular}
}
\caption{
Statistical results for AQuA using Qwen2-1.5B-Instruct on irrelevant responses.
}
\label{tbl:reasoning_cot_500_wrong_answer_shuffle_qwen2_1_5b_it_grads_aqua_train}
\end{table}

\begin{figure*}[!h]
\centering
\includegraphics[width=\linewidth]{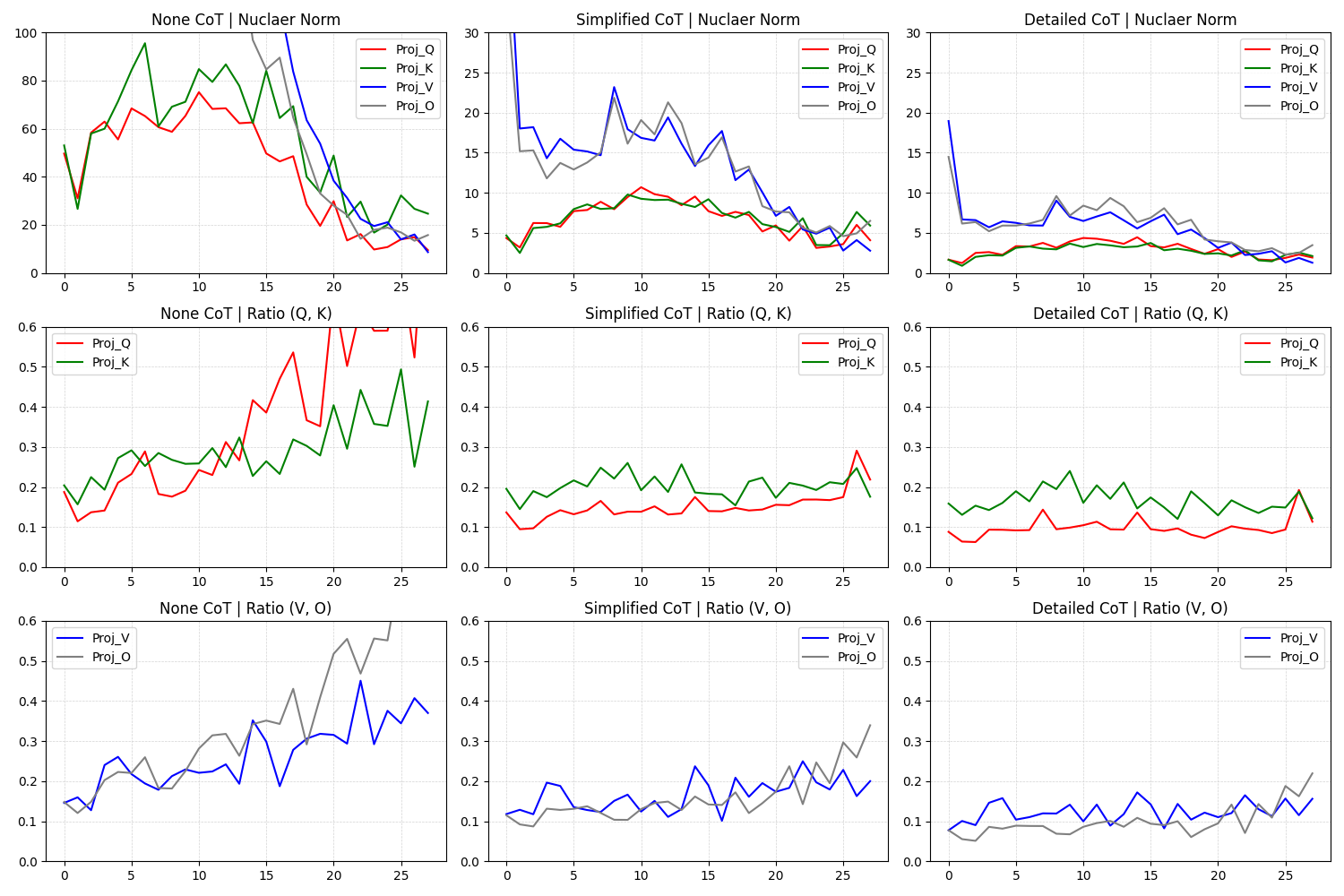}
\caption{Visualization for AQuA using Qwen2-1.5B-Instruct on irrelevant responses.}
\label{fig:reasoning_nonsense_aqua_it}
\vspace{-2mm}
\end{figure*}

\begin{table}[!tbh]
\centering
\small
\scalebox{0.75}{
\begin{tabular}{l|l|l|cccc}
\toprule
\multirow{2}{*}{\textbf{Dataset}} & \multirow{2}{*}{\textbf{Curve}} & \multirow{2}{*}{\textbf{Cot}} &  \multicolumn{4}{c}{\textbf{Mean Absolute Difference (MAD)}} \\
 & &  & Early & Middle & Last & All\\
\midrule
\multirow{24}{*}{GSM8K} & \multirow{3}{*}{$s_Q$} 
  & None & 7.27 & 4.88 & 12.46 & 7.22 \\
& & Simplified & 1.29 & 1.03 & 2.17 & 1.42 \\
& & Detailed & 0.52 & 0.49 & 0.47 & 0.51 \\
\cmidrule{2-7}
& \multirow{3}{*}{$s_K$} 
  & None & 10.22 & 8.58 & 12.97 & 10.05 \\
& & Simplified & 1.44 & 1.26 & 2.16 & 1.48 \\
& & Detailed & 0.51 & 0.36 & 0.62 & 0.44 \\
\cmidrule{2-7}
& \multirow{3}{*}{$s_V$} 
  & None & 39.15 & 24.33 & 8.64 & 22.97 \\
& & Simplified & 6.56 & 4.21 & 1.80 & 3.98 \\
& & Detailed & 2.38 & 1.30 & 0.72 & 1.35 \\
\cmidrule{2-7}
& \multirow{3}{*}{$s_O$} 
  & None & 27.74 & 23.41 & 5.31 & 19.20 \\
& & Simplified & 4.80 & 4.48 & 1.20 & 3.61 \\
& & Detailed & 1.65 & 1.49 & 0.55 & 1.24 \\
\cmidrule{2-7}
 & \multirow{3}{*}{$r_Q$} 
  & None & 0.03 & 0.03 & 0.11 & 0.05 \\
& & Simplified & 0.01 & 0.01 & 0.07 & 0.03 \\
& & Detailed & 0.01 & 0.02 & 0.04 & 0.02 \\
\cmidrule{2-7}
& \multirow{3}{*}{$r_K$} 
  & None & 0.03 & 0.04 & 0.04 & 0.03 \\
& & Simplified & 0.03 & 0.04 & 0.01 & 0.03 \\
& & Detailed & 0.02 & 0.05 & 0.02 & 0.04 \\
\cmidrule{2-7}
 & \multirow{3}{*}{$r_V$} 
  & None & 0.03 & 0.05 & 0.05 & 0.04 \\
& & Simplified & 0.02 & 0.05 & 0.05 & 0.04 \\
& & Detailed & 0.02 & 0.03 & 0.03 & 0.03 \\
\cmidrule{2-7}
& \multirow{3}{*}{$r_O$} 
  & None & 0.02 & 0.04 & 0.07 & 0.04 \\
& & Simplified & 0.02 & 0.02 & 0.07 & 0.03 \\
& & Detailed & 0.01 & 0.01 & 0.05 & 0.02 \\
\bottomrule
\end{tabular}
}
\caption{
Statistical results for GSM8K using Qwen2-1.5B-Instruct on irrelevant responses.
}
\label{tbl:reasoning_cot_500_wrong_answer_shuffle_qwen2_1_5b_it_grads_gsm8k_train}
\end{table}

\begin{figure*}[!h]
\centering
\includegraphics[width=\linewidth]{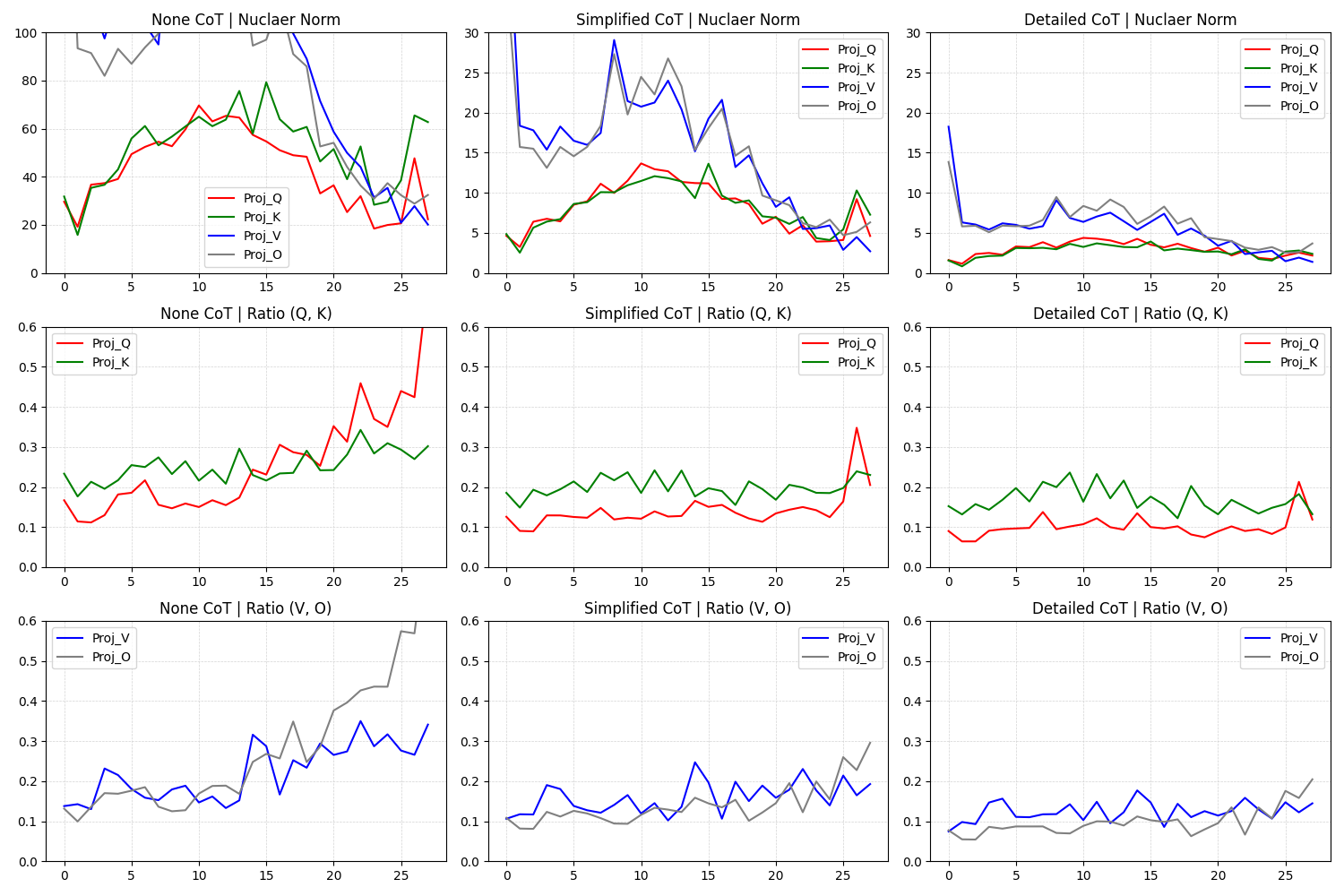}
\caption{Visualization for GSM8K using Qwen2-1.5B-Instruct on irrelevant responses.}
\label{fig:reasoning_nonsense_gsm8k_it}
\vspace{-2mm}
\end{figure*}

\begin{table}[!tbh]
\centering
\small
\scalebox{0.75}{
\begin{tabular}{l|l|l|cccc}
\toprule
\multirow{2}{*}{\textbf{Dataset}} & \multirow{2}{*}{\textbf{Curve}} & \multirow{2}{*}{\textbf{Cot}} &  \multicolumn{4}{c}{\textbf{Mean Absolute Difference (MAD)}} \\
 & &  & Early & Middle & Last & All\\
\midrule
\multirow{24}{*}{StrategyQA} & \multirow{3}{*}{$s_Q$} 
  & None & 3.80 & 2.86 & 0.75 & 2.64 \\
& & Simplified & 0.97 & 0.70 & 0.42 & 0.72 \\
& & Detailed & 0.66 & 0.46 & 0.27 & 0.47 \\
\cmidrule{2-7}
& \multirow{3}{*}{$s_K$} 
  & None & 8.60 & 6.48 & 2.08 & 5.77 \\
& & Simplified & 1.39 & 0.79 & 0.73 & 0.94 \\
& & Detailed & 0.52 & 0.35 & 0.42 & 0.39 \\
\cmidrule{2-7}
& \multirow{3}{*}{$s_V$} 
  & None & 31.41 & 16.63 & 3.81 & 16.24 \\
& & Simplified & 7.36 & 2.47 & 1.64 & 3.31 \\
& & Detailed & 3.06 & 1.03 & 0.88 & 1.43 \\
\cmidrule{2-7}
& \multirow{3}{*}{$s_O$} 
  & None & 22.50 & 10.10 & 3.98 & 10.96 \\
& & Simplified & 5.21 & 2.66 & 1.21 & 2.82 \\
& & Detailed & 2.24 & 1.56 & 0.79 & 1.47 \\
\cmidrule{2-7}
 & \multirow{3}{*}{$r_Q$} 
  & None & 0.05 & 0.06 & 0.09 & 0.07 \\
& & Simplified & 0.01 & 0.02 & 0.02 & 0.02 \\
& & Detailed & 0.02 & 0.02 & 0.01 & 0.02 \\
\cmidrule{2-7}
& \multirow{3}{*}{$r_K$} 
  & None & 0.06 & 0.04 & 0.06 & 0.05 \\
& & Simplified & 0.04 & 0.03 & 0.02 & 0.03 \\
& & Detailed & 0.02 & 0.03 & 0.02 & 0.03 \\
\cmidrule{2-7}
 & \multirow{3}{*}{$r_V$} 
  & None & 0.04 & 0.08 & 0.07 & 0.06 \\
& & Simplified & 0.03 & 0.07 & 0.10 & 0.06 \\
& & Detailed & 0.02 & 0.04 & 0.05 & 0.03 \\
\cmidrule{2-7}
& \multirow{3}{*}{$r_O$} 
  & None & 0.03 & 0.06 & 0.05 & 0.05 \\
& & Simplified & 0.01 & 0.03 & 0.07 & 0.03 \\
& & Detailed & 0.01 & 0.01 & 0.05 & 0.02 \\
\bottomrule
\end{tabular}
}
\caption{
Statistical results for StrategyQA using Qwen2-1.5B-Instruct on irrelevant responses.
}
\label{tbl:reasoning_cot_500_wrong_answer_shuffle_qwen2_1_5b_it_grads_strategyqa_train}
\end{table}

\begin{figure*}[!h]
\centering
\includegraphics[width=\linewidth]{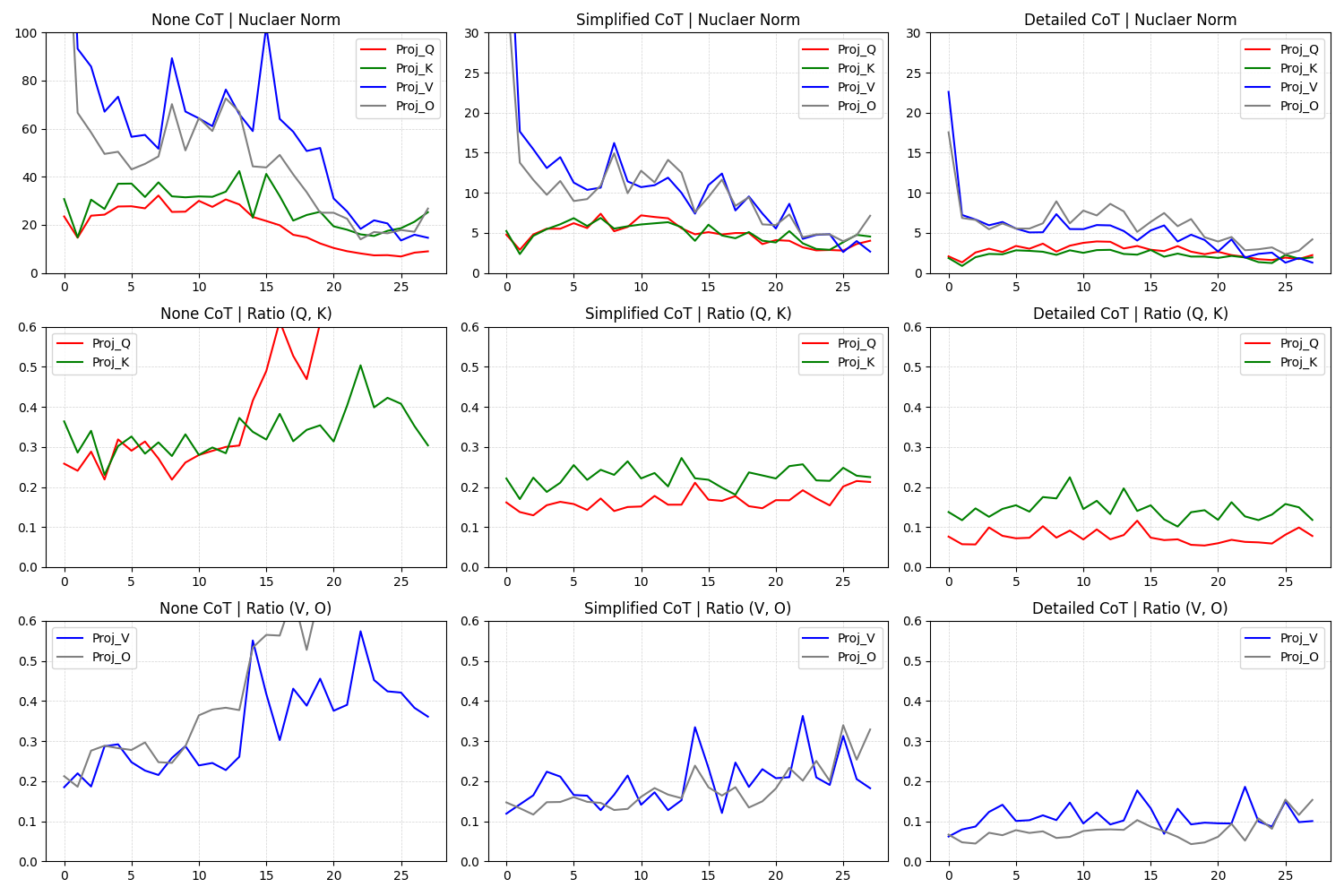}
\caption{Visualization for StrategyQA using Qwen2-1.5B-Instruct on irrelevant responses.}
\label{fig:reasoning_nonsense_strategyqa_it}
\vspace{-2mm}
\end{figure*}

\begin{table}[!tbh]
\centering
\small
\scalebox{0.75}{
\begin{tabular}{l|l|l|cccc}
\toprule
\multirow{2}{*}{\textbf{Dataset}} & \multirow{2}{*}{\textbf{Curve}} & \multirow{2}{*}{\textbf{Cot}} &  \multicolumn{4}{c}{\textbf{Mean Absolute Difference (MAD)}} \\
 & &  & Early & Middle & Last & All\\
\midrule
\multirow{24}{*}{ECQA} & \multirow{3}{*}{$s_Q$} 
  & None & 5.61 & 3.42 & 3.16 & 3.93 \\
& & Simplified & 1.70 & 1.11 & 2.04 & 1.51 \\
& & Detailed & 0.71 & 0.65 & 0.37 & 0.61 \\
\cmidrule{2-7}
& \multirow{3}{*}{$s_K$} 
  & None & 7.41 & 4.11 & 9.45 & 5.85 \\
& & Simplified & 1.96 & 1.64 & 3.10 & 1.93 \\
& & Detailed & 0.65 & 0.48 & 0.74 & 0.56 \\
\cmidrule{2-7}
& \multirow{3}{*}{$s_V$} 
  & None & 37.75 & 13.94 & 5.35 & 16.54 \\
& & Simplified & 11.34 & 4.99 & 2.53 & 5.60 \\
& & Detailed & 3.13 & 1.24 & 0.84 & 1.52 \\
\cmidrule{2-7}
& \multirow{3}{*}{$s_O$} 
  & None & 25.53 & 14.51 & 4.79 & 13.90 \\
& & Simplified & 8.06 & 5.22 & 2.11 & 4.86 \\
& & Detailed & 2.21 & 1.73 & 0.80 & 1.53 \\
\cmidrule{2-7}
 & \multirow{3}{*}{$r_Q$} 
  & None & 0.02 & 0.05 & 0.08 & 0.05 \\
& & Simplified & 0.02 & 0.01 & 0.12 & 0.04 \\
& & Detailed & 0.01 & 0.02 & 0.03 & 0.02 \\
\cmidrule{2-7}
& \multirow{3}{*}{$r_K$} 
  & None & 0.04 & 0.05 & 0.04 & 0.05 \\
& & Simplified & 0.03 & 0.04 & 0.06 & 0.04 \\
& & Detailed & 0.02 & 0.04 & 0.04 & 0.04 \\
\cmidrule{2-7}
 & \multirow{3}{*}{$r_V$} 
  & None & 0.03 & 0.05 & 0.06 & 0.04 \\
& & Simplified & 0.03 & 0.07 & 0.06 & 0.05 \\
& & Detailed & 0.02 & 0.03 & 0.04 & 0.03 \\
\cmidrule{2-7}
& \multirow{3}{*}{$r_O$} 
  & None & 0.02 & 0.06 & 0.07 & 0.05 \\
& & Simplified & 0.02 & 0.02 & 0.07 & 0.03 \\
& & Detailed & 0.02 & 0.02 & 0.04 & 0.02 \\
\bottomrule
\end{tabular}
}
\caption{
Statistical results for ECQA using Qwen2-1.5B-Instruct on irrelevant responses.
}
\label{tbl:reasoning_cot_500_wrong_answer_shuffle_qwen2_1_5b_it_grads_ecqa_train}
\end{table}

\begin{figure*}[!h]
\centering
\includegraphics[width=\linewidth]{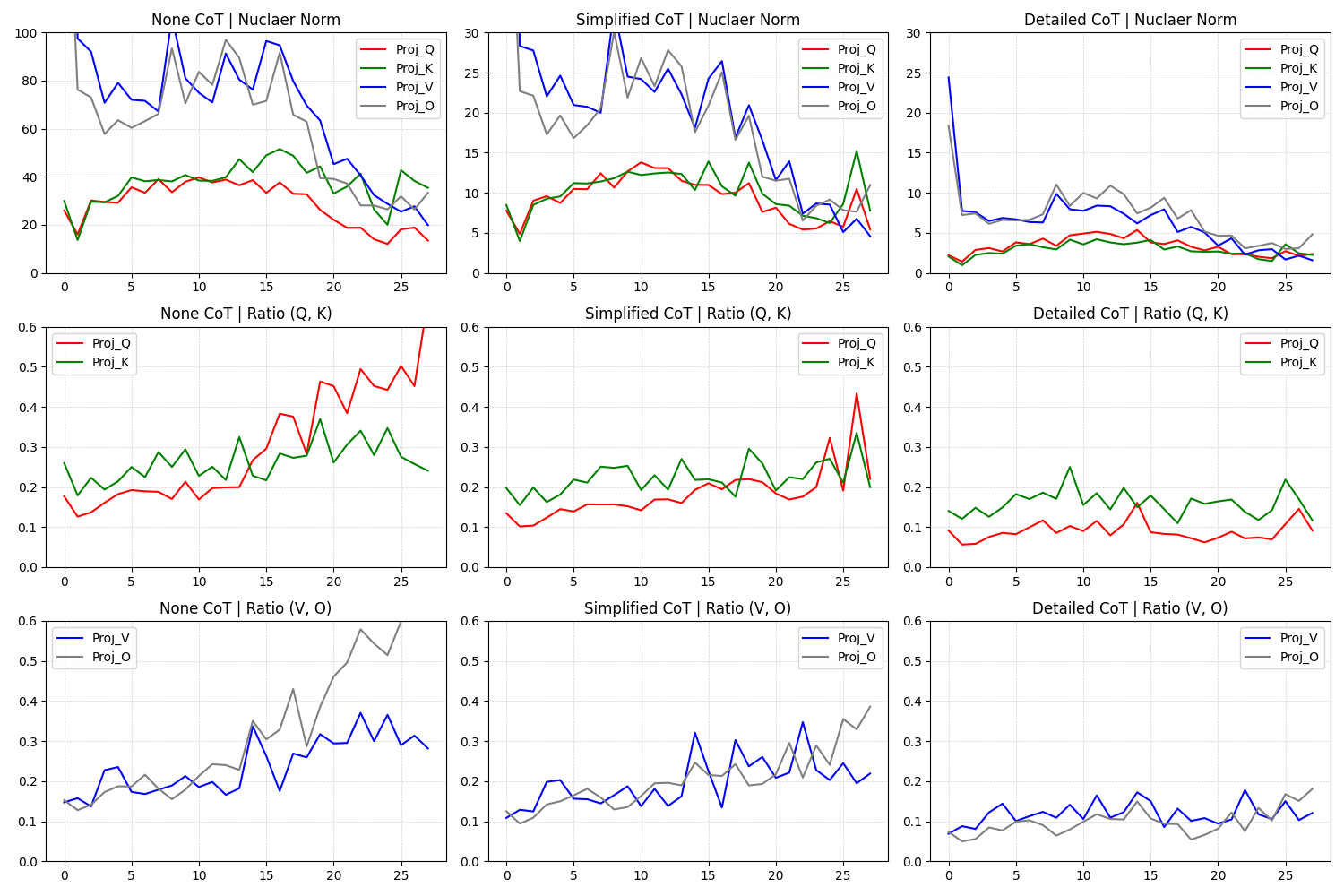}
\caption{Visualization for ECQA using Qwen2-1.5B-Instruct on irrelevant responses.}
\label{fig:reasoning_nonsense_ecqa_it}
\vspace{-2mm}
\end{figure*}

\begin{table}[!tbh]
\centering
\small
\scalebox{0.75}{
\begin{tabular}{l|l|l|cccc}
\toprule
\multirow{2}{*}{\textbf{Dataset}} & \multirow{2}{*}{\textbf{Curve}} & \multirow{2}{*}{\textbf{Cot}} &  \multicolumn{4}{c}{\textbf{Mean Absolute Difference (MAD)}} \\
 & &  & Early & Middle & Last & All\\
\midrule
\multirow{24}{*}{CREAK} & \multirow{3}{*}{$s_Q$} 
  & None & 10.37 & 10.21 & 6.14 & 9.12 \\
& & Simplified & 2.27 & 1.89 & 4.72 & 2.70 \\
& & Detailed & 0.68 & 0.53 & 0.33 & 0.53 \\
\cmidrule{2-7}
& \multirow{3}{*}{$s_K$} 
  & None & 14.60 & 18.67 & 16.06 & 15.90 \\
& & Simplified & 2.58 & 3.23 & 5.85 & 3.54 \\
& & Detailed & 0.58 & 0.39 & 0.69 & 0.48 \\
\cmidrule{2-7}
& \multirow{3}{*}{$s_V$} 
  & None & 74.28 & 33.01 & 8.01 & 34.90 \\
& & Simplified & 16.19 & 7.40 & 3.77 & 8.12 \\
& & Detailed & 3.52 & 1.35 & 0.89 & 1.68 \\
\cmidrule{2-7}
& \multirow{3}{*}{$s_O$} 
  & None & 48.66 & 30.51 & 4.55 & 27.25 \\
& & Simplified & 11.07 & 7.13 & 2.50 & 6.58 \\
& & Detailed & 2.40 & 1.79 & 0.74 & 1.60 \\
\cmidrule{2-7}
 & \multirow{3}{*}{$r_Q$} 
  & None & 0.02 & 0.07 & 0.14 & 0.08 \\
& & Simplified & 0.01 & 0.03 & 0.16 & 0.06 \\
& & Detailed & 0.01 & 0.01 & 0.03 & 0.02 \\
\cmidrule{2-7}
& \multirow{3}{*}{$r_K$} 
  & None & 0.03 & 0.06 & 0.07 & 0.05 \\
& & Simplified & 0.03 & 0.05 & 0.08 & 0.05 \\
& & Detailed & 0.02 & 0.03 & 0.04 & 0.03 \\
\cmidrule{2-7}
 & \multirow{3}{*}{$r_V$} 
  & None & 0.04 & 0.05 & 0.08 & 0.06 \\
& & Simplified & 0.03 & 0.07 & 0.09 & 0.06 \\
& & Detailed & 0.02 & 0.04 & 0.04 & 0.03 \\
\cmidrule{2-7}
& \multirow{3}{*}{$r_O$} 
  & None & 0.02 & 0.06 & 0.10 & 0.05 \\
& & Simplified & 0.02 & 0.03 & 0.08 & 0.04 \\
& & Detailed & 0.02 & 0.02 & 0.04 & 0.02 \\
\bottomrule
\end{tabular}
}
\caption{
Statistical results for CREAK using Qwen2-1.5B-Instruct on irrelevant responses.
}
\label{tbl:reasoning_cot_500_wrong_answer_shuffle_qwen2_1_5b_it_grads_creak_train}
\end{table}

\begin{figure*}[!h]
\centering
\includegraphics[width=\linewidth]{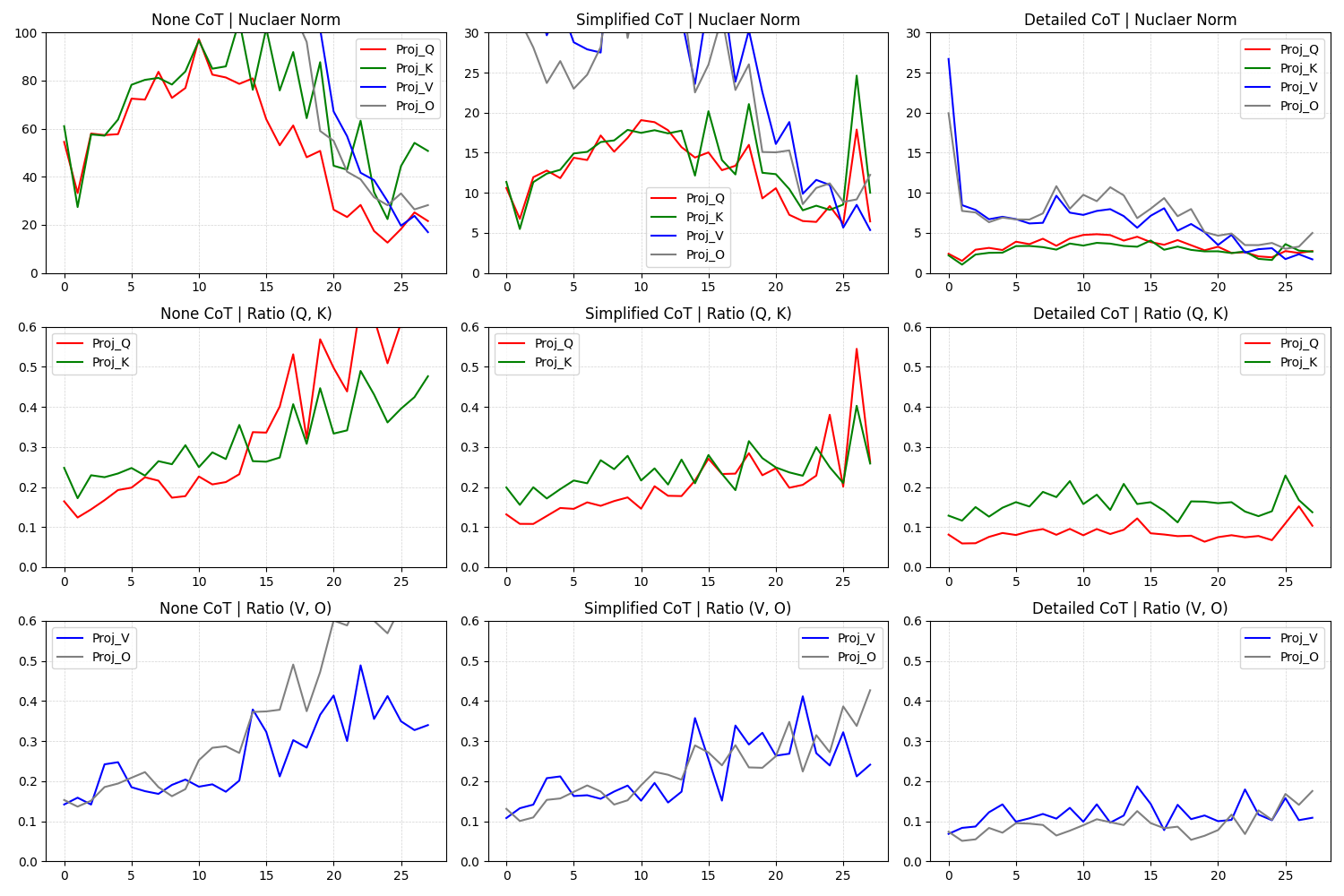}
\caption{Visualization for CREAK using Qwen2-1.5B-Instruct on irrelevant responses.}
\label{fig:reasoning_nonsense_creak_it}
\vspace{-2mm}
\end{figure*}

\begin{table}[!tbh]
\centering
\small
\scalebox{0.75}{
\begin{tabular}{l|l|l|cccc}
\toprule
\multirow{2}{*}{\textbf{Dataset}} & \multirow{2}{*}{\textbf{Curve}} & \multirow{2}{*}{\textbf{Cot}} &  \multicolumn{4}{c}{\textbf{Mean Absolute Difference (MAD)}} \\
 & &  & Early & Middle & Last & All\\
\midrule
\multirow{24}{*}{Sensemaking} & \multirow{3}{*}{$s_Q$} 
  & None & 3.04 & 3.16 & 4.11 & 3.41 \\
& & Simplified & 1.58 & 1.31 & 1.26 & 1.43 \\
& & Detailed & 0.71 & 0.66 & 0.42 & 0.62 \\
\cmidrule{2-7}
& \multirow{3}{*}{$s_K$} 
  & None & 3.70 & 2.41 & 6.81 & 3.76 \\
& & Simplified & 2.08 & 1.45 & 2.24 & 1.67 \\
& & Detailed & 0.59 & 0.50 & 0.78 & 0.56 \\
\cmidrule{2-7}
& \multirow{3}{*}{$s_V$} 
  & None & 17.71 & 7.85 & 2.32 & 8.34 \\
& & Simplified & 10.88 & 5.10 & 2.28 & 5.45 \\
& & Detailed & 2.88 & 1.18 & 0.74 & 1.41 \\
\cmidrule{2-7}
& \multirow{3}{*}{$s_O$} 
  & None & 13.38 & 7.51 & 1.74 & 7.23 \\
& & Simplified & 7.24 & 4.89 & 1.73 & 4.47 \\
& & Detailed & 2.18 & 1.53 & 0.62 & 1.40 \\
\cmidrule{2-7}
 & \multirow{3}{*}{$r_Q$} 
  & None & 0.03 & 0.05 & 0.15 & 0.07 \\
& & Simplified & 0.02 & 0.02 & 0.05 & 0.03 \\
& & Detailed & 0.01 & 0.02 & 0.04 & 0.02 \\
\cmidrule{2-7}
& \multirow{3}{*}{$r_K$} 
  & None & 0.03 & 0.03 & 0.11 & 0.05 \\
& & Simplified & 0.03 & 0.04 & 0.01 & 0.03 \\
& & Detailed & 0.02 & 0.05 & 0.02 & 0.04 \\
\cmidrule{2-7}
 & \multirow{3}{*}{$r_V$} 
  & None & 0.04 & 0.03 & 0.07 & 0.04 \\
& & Simplified & 0.03 & 0.06 & 0.05 & 0.05 \\
& & Detailed & 0.03 & 0.03 & 0.03 & 0.03 \\
\cmidrule{2-7}
& \multirow{3}{*}{$r_O$} 
  & None & 0.02 & 0.05 & 0.09 & 0.05 \\
& & Simplified & 0.02 & 0.03 & 0.07 & 0.04 \\
& & Detailed & 0.01 & 0.02 & 0.04 & 0.02 \\
\bottomrule
\end{tabular}
}
\caption{
Statistical results for Sensemaking using Qwen2-1.5B-Instruct on irrelevant responses.
}
\label{tbl:reasoning_cot_500_wrong_answer_shuffle_qwen2_1_5b_it_grads_sensemaking_train}
\end{table}

\begin{figure*}[!h]
\centering
\includegraphics[width=\linewidth]{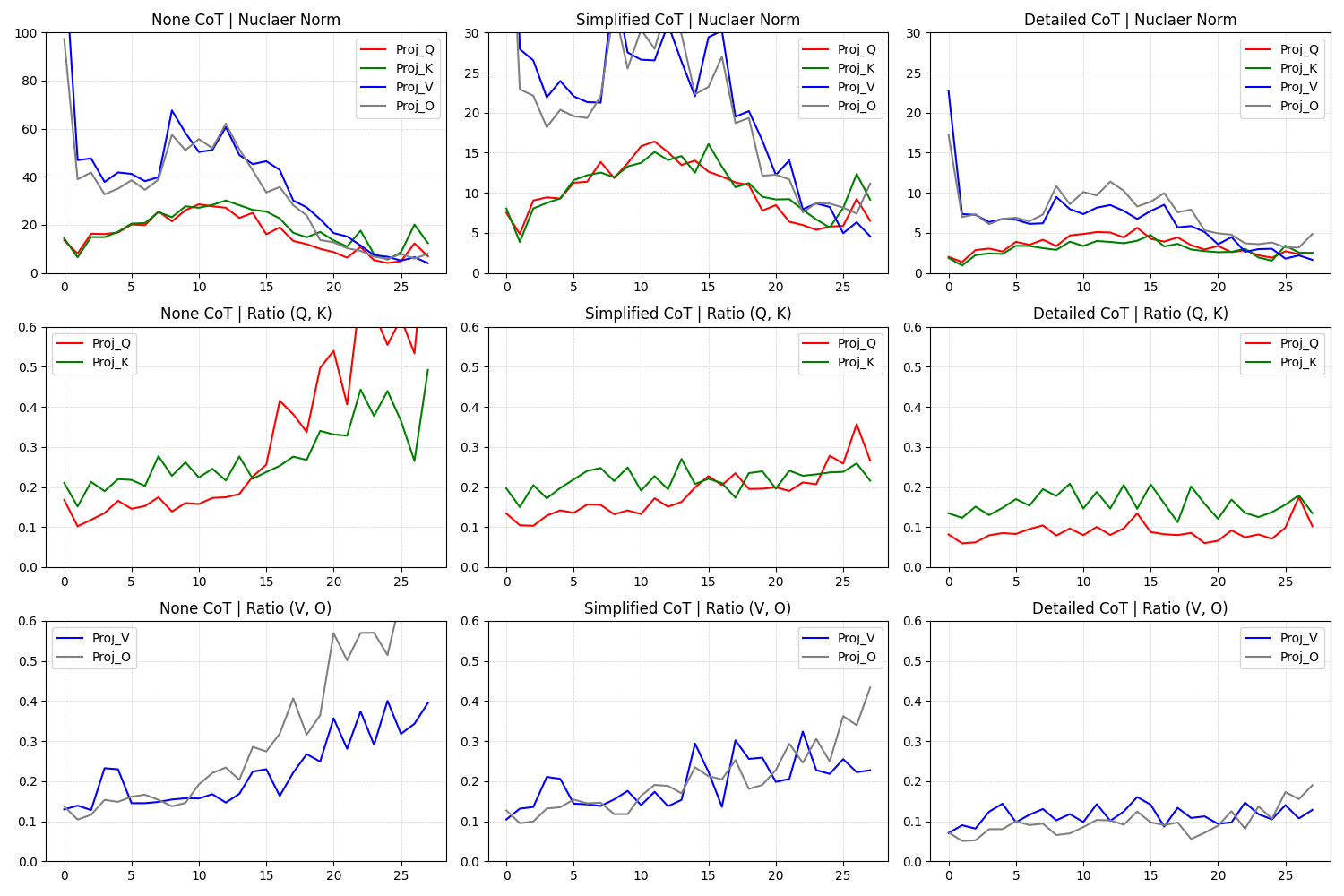}
\caption{Visualization for Sensemaking using Qwen2-1.5B-Instruct on irrelevant responses.}
\label{fig:reasoning_nonsense_sensemaking_it}
\vspace{-2mm}
\end{figure*}

\subsubsection{Wiki Tasks}

The visualizations and statistical results on Wiki tasks are shown in Figure \ref{fig:wiki_nonsense_it} and Table \ref{tbl:wiki_wiki_popularity_wrong_answer_shuffle_qwen2_1_5b_it_grads}.

\begin{table}[!tbh]
\centering
\small
\scalebox{0.75}{
\begin{tabular}{l|l|l|cccc}
\toprule
\multirow{2}{*}{\textbf{Dataset}} & \multirow{2}{*}{\textbf{Curve}} & \multirow{2}{*}{\textbf{Cot}} &  \multicolumn{4}{c}{\textbf{Mean Absolute Difference (MAD)}} \\
 & &  & Early & Middle & Last & All\\
\midrule
\multirow{32}{*}{Wiki} & \multirow{3}{*}{$s_Q$} 
  & Len 100 & 0.83 & 0.51 & 0.40 & 0.58 \\
& & Len 500 & 0.73 & 0.33 & 0.29 & 0.42 \\
& & Len 1000 & 0.65 & 0.28 & 0.26 & 0.36 \\
& & Unpopular & 0.86 & 0.60 & 0.45 & 0.61 \\
\cmidrule{2-7}
& \multirow{3}{*}{$s_K$} 
  & Len 100 & 0.98 & 0.53 & 0.54 & 0.67 \\
& & Len 500 & 0.60 & 0.27 & 0.27 & 0.35 \\
& & Len 1000 & 0.49 & 0.20 & 0.21 & 0.27 \\
& & Unpopular & 1.22 & 0.65 & 0.72 & 0.79 \\
\cmidrule{2-7}
& \multirow{3}{*}{$s_V$} 
  & Len 100 & 5.79 & 2.01 & 1.42 & 2.70 \\
& & Len 500 & 3.77 & 1.20 & 1.07 & 1.74 \\
& & Len 1000 & 3.07 & 0.91 & 0.89 & 1.39 \\
& & Unpopular & 8.64 & 2.27 & 1.33 & 3.42 \\
\cmidrule{2-7}
& \multirow{3}{*}{$s_O$} 
  & Len 100 & 4.06 & 2.07 & 0.98 & 2.22 \\
& & Len 500 & 2.84 & 1.56 & 0.82 & 1.63 \\
& & Len 1000 & 2.45 & 1.35 & 0.76 & 1.42 \\
& & Unpopular & 5.89 & 2.10 & 0.99 & 2.65 \\
\cmidrule{2-7}
 & \multirow{3}{*}{$r_Q$} 
  & Len 100 & 0.02 & 0.02 & 0.02 & 0.02 \\
& & Len 500 & 0.02 & 0.02 & 0.01 & 0.02 \\
& & Len 1000 & 0.03 & 0.01 & 0.01 & 0.02 \\
& & Unpopular & 0.01 & 0.02 & 0.02 & 0.02 \\
\cmidrule{2-7}
& \multirow{3}{*}{$r_K$} 
  & Len 100 & 0.03 & 0.02 & 0.02 & 0.03 \\
& & Len 500 & 0.02 & 0.03 & 0.02 & 0.02 \\
& & Len 1000 & 0.02 & 0.03 & 0.01 & 0.02 \\
& & Unpopular & 0.04 & 0.03 & 0.02 & 0.03 \\
\cmidrule{2-7}
 & \multirow{3}{*}{$r_V$} 
  & Len 100 & 0.03 & 0.06 & 0.09 & 0.06 \\
& & Len 500 & 0.02 & 0.05 & 0.08 & 0.05 \\
& & Len 1000 & 0.02 & 0.04 & 0.07 & 0.04 \\
& & Unpopular & 0.03 & 0.07 & 0.09 & 0.06 \\
\cmidrule{2-7}
& \multirow{3}{*}{$r_O$} 
  & Len 100 & 0.02 & 0.02 & 0.07 & 0.03 \\
& & Len 500 & 0.01 & 0.01 & 0.06 & 0.02 \\
& & Len 1000 & 0.01 & 0.01 & 0.05 & 0.02 \\
& & Unpopular & 0.01 & 0.03 & 0.08 & 0.04 \\
\bottomrule
\end{tabular}
}
\caption{
Statistical results for Wiki using Qwen2-1.5B-Instruct on irrelevant responses.
}
\label{tbl:wiki_wiki_popularity_wrong_answer_shuffle_qwen2_1_5b_it_grads}
\end{table}

\begin{figure*}[!h]
\centering
\includegraphics[width=\linewidth]{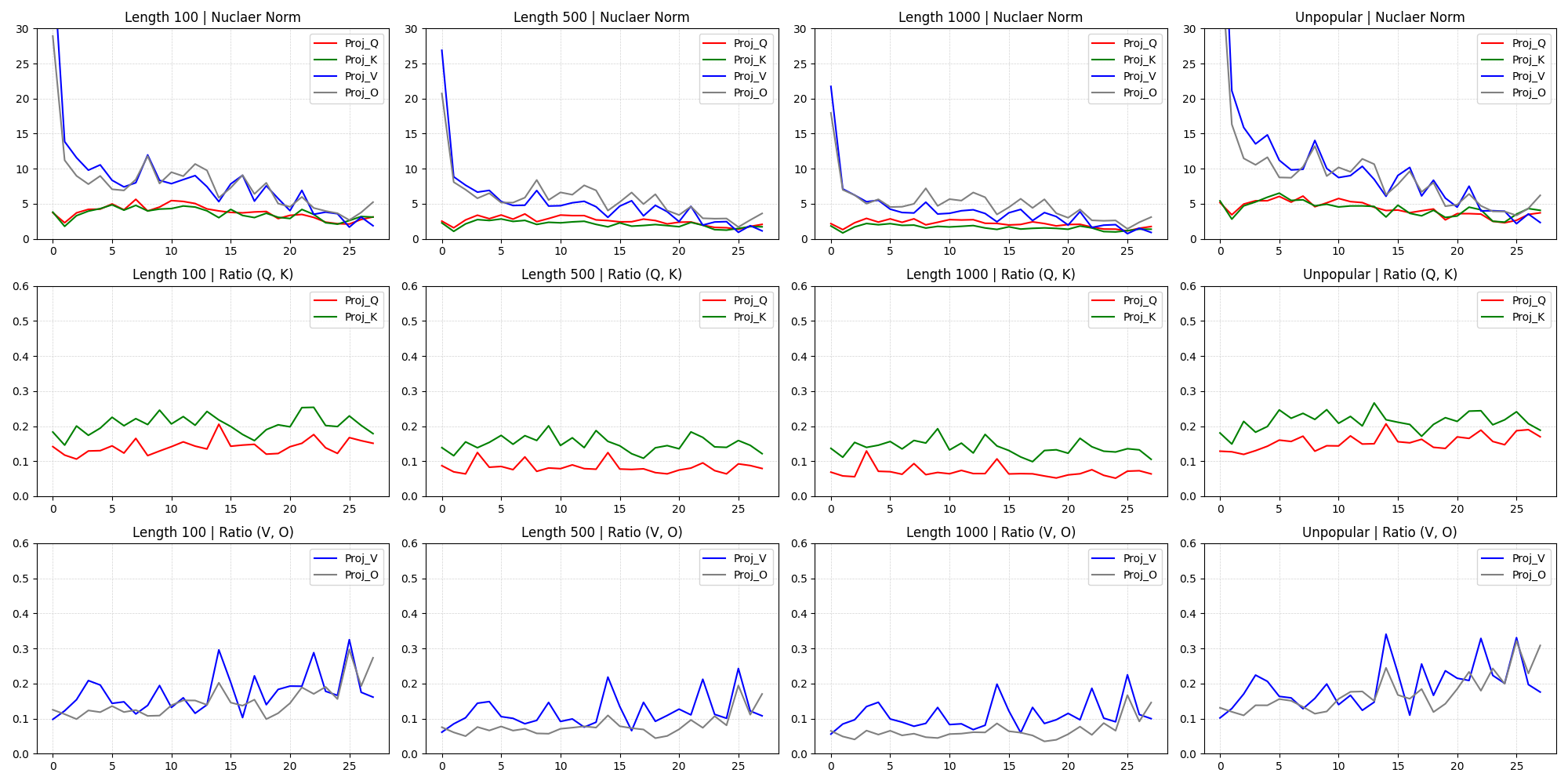}
\caption{Visualization for Wiki tasks using Qwen2-1.5B-Instruct on irrelevant responses.}
\label{fig:wiki_nonsense_it}
\vspace{-2mm}
\end{figure*}

\clearpage
\section{Results on Llama-2-7B-hf}
\subsection{Pre-trained LLM on Correct Responses}

\subsubsection{Reasoning Tasks}

The visualizations and statistical results on MATH tasks: MATH-Algebra (Figure \ref{fig:math_correct_algebra}, Table \ref{tbl:math_lighteval_math_llama2_7b_grads_algebra}), MATH-Counting (Figure \ref{fig:math_correct_counting}, Table \ref{tbl:math_lighteval_math_llama2_7b_grads_counting_probability}), MATH-Geometry (Figure \ref{fig:math_correct_geometry}, Table \ref{tbl:math_lighteval_math_llama2_7b_grads_geometry}).

\begin{table}[!tbh]
\centering
\small
\scalebox{0.75}{
\begin{tabular}{l|l|l|cccc}
\toprule
\multirow{2}{*}{\textbf{Dataset}} & \multirow{2}{*}{\textbf{Curve}} & \multirow{2}{*}{\textbf{Cot}} &  \multicolumn{4}{c}{\textbf{Mean Absolute Difference (MAD)}} \\
 & &  & Early & Middle & Last & All\\
\midrule
\multirow{16}{*}{Algebra} & \multirow{3}{*}{$s_Q$} 
  & Simplified & 0.27 & 0.19 & 0.08 & 0.18 \\
& & Detailed & 0.17 & 0.11 & 0.05 & 0.11 \\
\cmidrule{2-7}
& \multirow{3}{*}{$s_K$} 
  & Simplified & 0.26 & 0.19 & 0.12 & 0.19 \\
& & Detailed & 0.17 & 0.13 & 0.08 & 0.12 \\
\cmidrule{2-7}
& \multirow{3}{*}{$s_V$} 
  & Simplified & 1.24 & 0.62 & 0.18 & 0.65 \\
& & Detailed & 0.72 & 0.40 & 0.09 & 0.39 \\
\cmidrule{2-7}
& \multirow{3}{*}{$s_O$} 
  & Simplified & 0.66 & 0.59 & 0.17 & 0.45 \\
& & Detailed & 0.44 & 0.41 & 0.10 & 0.30 \\
\cmidrule{2-7}
 & \multirow{3}{*}{$r_Q$} 
  & Simplified & 0.03 & 0.01 & 0.01 & 0.01 \\
& & Detailed & 0.04 & 0.00 & 0.01 & 0.02 \\
\cmidrule{2-7}
& \multirow{3}{*}{$r_K$} 
  & Simplified & 0.03 & 0.01 & 0.01 & 0.02 \\
& & Detailed & 0.03 & 0.01 & 0.01 & 0.01 \\
\cmidrule{2-7}
 & \multirow{3}{*}{$r_V$} 
  & Simplified & 0.04 & 0.01 & 0.01 & 0.02 \\
& & Detailed & 0.04 & 0.01 & 0.01 & 0.02 \\
\cmidrule{2-7}
& \multirow{3}{*}{$r_O$} 
  & Simplified & 0.02 & 0.00 & 0.01 & 0.01 \\
& & Detailed & 0.02 & 0.00 & 0.01 & 0.01 \\
\bottomrule
\end{tabular}
}
\caption{
Statistical results for MATH-Algebra using Llama-2-7b-hf on correct responses.
}
\label{tbl:math_lighteval_math_llama2_7b_grads_algebra}
\end{table}

\begin{figure}[!h]
\centering
\includegraphics[width=\linewidth]{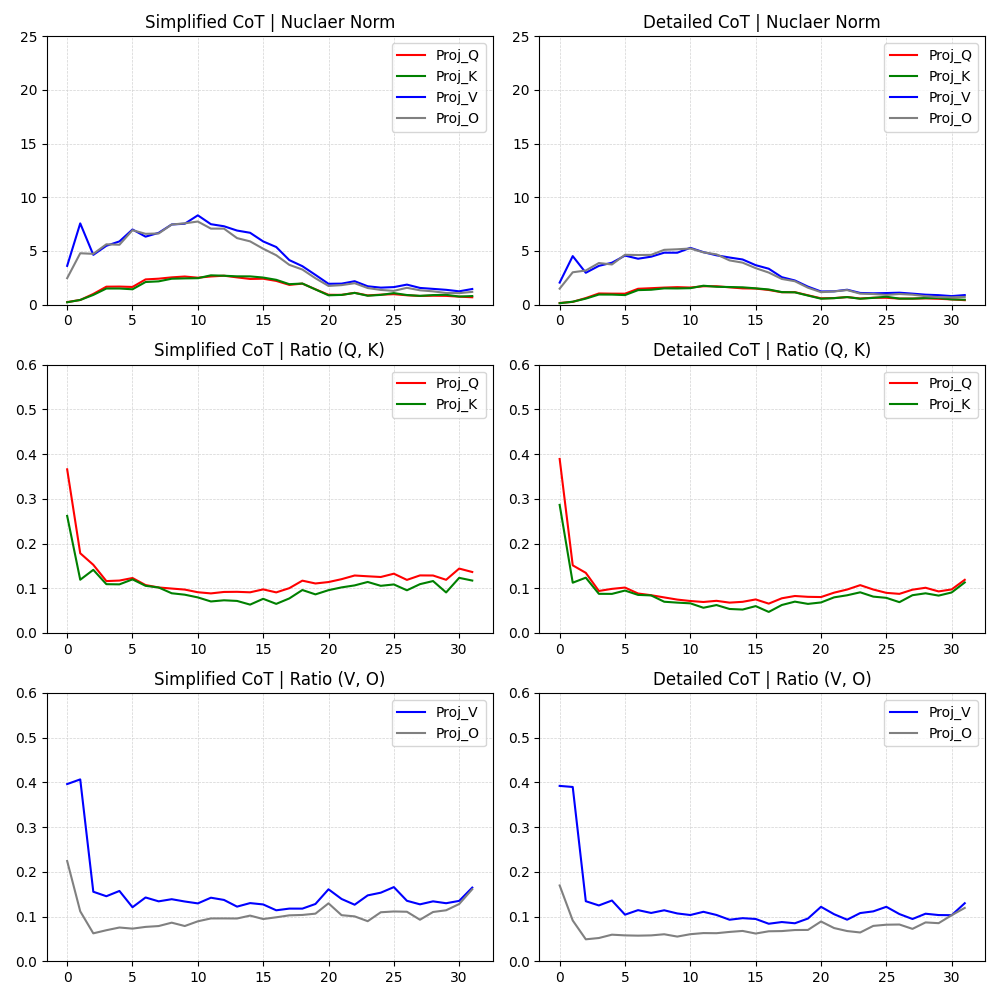}
\caption{Visualization for MATH-Algebra using Llama-2-7b-hf on correct responses.}
\label{fig:math_correct_algebra}
\vspace{-2mm}
\end{figure}

\begin{table}[!tbh]
\centering
\small
\scalebox{0.75}{
\begin{tabular}{l|l|l|cccc}
\toprule
\multirow{2}{*}{\textbf{Dataset}} & \multirow{2}{*}{\textbf{Curve}} & \multirow{2}{*}{\textbf{Cot}} &  \multicolumn{4}{c}{\textbf{Mean Absolute Difference (MAD)}} \\
 & &  & Early & Middle & Last & All\\
\midrule
\multirow{16}{*}{Counting} & \multirow{3}{*}{$s_Q$} 
  & Simplified & 0.26 & 0.16 & 0.11 & 0.18 \\
& & Detailed & 0.17 & 0.11 & 0.07 & 0.11 \\
\cmidrule{2-7}
& \multirow{3}{*}{$s_K$} 
  & Simplified & 0.25 & 0.17 & 0.14 & 0.19 \\
& & Detailed & 0.16 & 0.12 & 0.09 & 0.12 \\
\cmidrule{2-7}
& \multirow{3}{*}{$s_V$} 
  & Simplified & 1.25 & 0.56 & 0.20 & 0.64 \\
& & Detailed & 0.74 & 0.37 & 0.11 & 0.39 \\
\cmidrule{2-7}
& \multirow{3}{*}{$s_O$} 
  & Simplified & 0.73 & 0.59 & 0.18 & 0.47 \\
& & Detailed & 0.50 & 0.39 & 0.11 & 0.31 \\
\cmidrule{2-7}
 & \multirow{3}{*}{$r_Q$} 
  & Simplified & 0.04 & 0.00 & 0.01 & 0.01 \\
& & Detailed & 0.04 & 0.01 & 0.01 & 0.02 \\
\cmidrule{2-7}
& \multirow{3}{*}{$r_K$} 
  & Simplified & 0.03 & 0.01 & 0.01 & 0.01 \\
& & Detailed & 0.03 & 0.01 & 0.01 & 0.01 \\
\cmidrule{2-7}
 & \multirow{3}{*}{$r_V$} 
  & Simplified & 0.04 & 0.01 & 0.01 & 0.02 \\
& & Detailed & 0.04 & 0.01 & 0.01 & 0.02 \\
\cmidrule{2-7}
& \multirow{3}{*}{$r_O$} 
  & Simplified & 0.02 & 0.00 & 0.01 & 0.01 \\
& & Detailed & 0.01 & 0.00 & 0.01 & 0.01 \\
\bottomrule
\end{tabular}
}
\caption{
Statistical results for MATH-Counting using Llama-2-7b-hf on correct responses.
}
\label{tbl:math_lighteval_math_llama2_7b_grads_counting_probability}
\end{table}

\begin{figure}[!h]
\centering
\includegraphics[width=\linewidth]{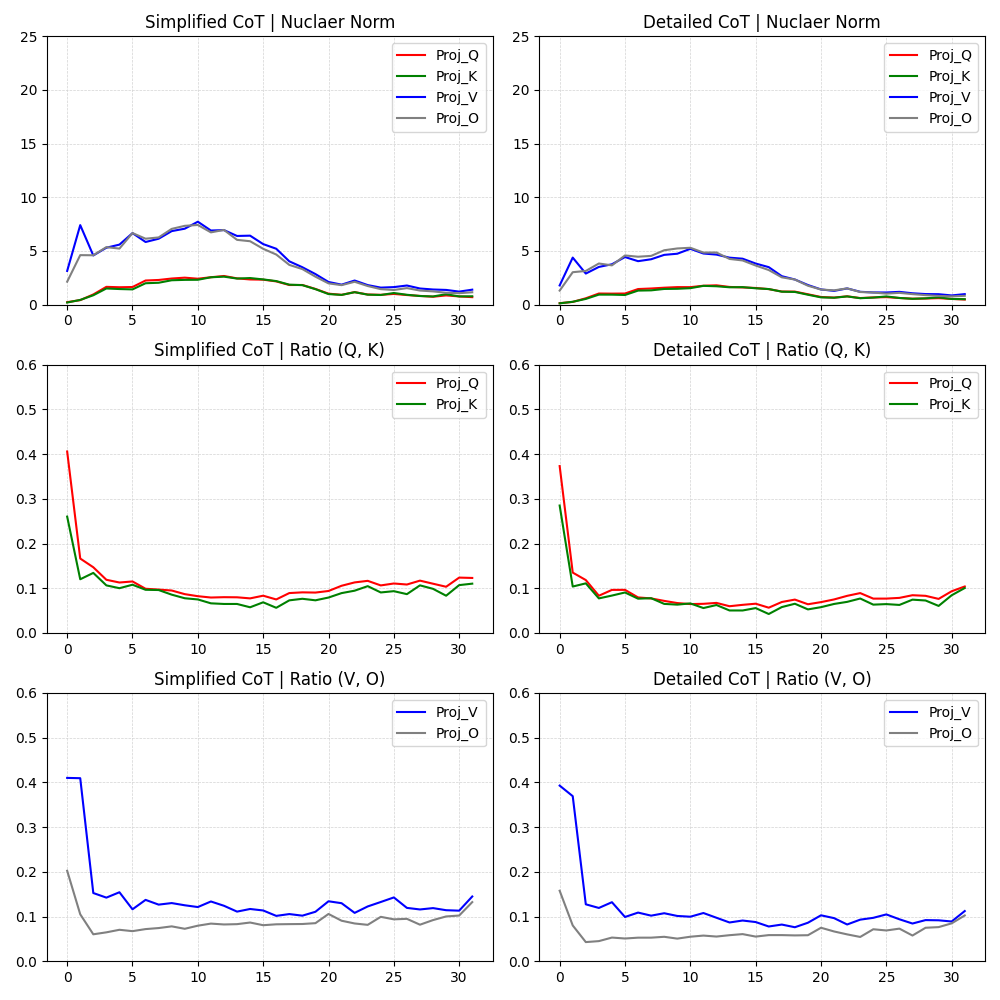}
\caption{Visualization for MATH-Counting using Llama-2-7b-hf on correct responses.}
\label{fig:math_correct_counting}
\vspace{-2mm}
\end{figure}

\begin{table}[!tbh]
\centering
\small
\scalebox{0.75}{
\begin{tabular}{l|l|l|cccc}
\toprule
\multirow{2}{*}{\textbf{Dataset}} & \multirow{2}{*}{\textbf{Curve}} & \multirow{2}{*}{\textbf{Cot}} &  \multicolumn{4}{c}{\textbf{Mean Absolute Difference (MAD)}} \\
 & &  & Early & Middle & Last & All\\
\midrule
\multirow{16}{*}{Geometry} & \multirow{3}{*}{$s_Q$} 
  & Simplified & 0.25 & 0.14 & 0.09 & 0.17 \\
& & Detailed & 0.18 & 0.11 & 0.07 & 0.12 \\
\cmidrule{2-7}
& \multirow{3}{*}{$s_K$} 
  & Simplified & 0.25 & 0.15 & 0.12 & 0.18 \\
& & Detailed & 0.17 & 0.12 & 0.08 & 0.13 \\
\cmidrule{2-7}
& \multirow{3}{*}{$s_V$} 
  & Simplified & 1.16 & 0.50 & 0.16 & 0.58 \\
& & Detailed & 0.79 & 0.40 & 0.12 & 0.42 \\
\cmidrule{2-7}
& \multirow{3}{*}{$s_O$} 
  & Simplified & 0.66 & 0.52 & 0.15 & 0.42 \\
& & Detailed & 0.53 & 0.41 & 0.12 & 0.33 \\
\cmidrule{2-7}
 & \multirow{3}{*}{$r_Q$} 
  & Simplified & 0.04 & 0.01 & 0.01 & 0.02 \\
& & Detailed & 0.04 & 0.01 & 0.01 & 0.02 \\
\cmidrule{2-7}
& \multirow{3}{*}{$r_K$} 
  & Simplified & 0.02 & 0.01 & 0.01 & 0.01 \\
& & Detailed & 0.03 & 0.01 & 0.01 & 0.01 \\
\cmidrule{2-7}
 & \multirow{3}{*}{$r_V$} 
  & Simplified & 0.04 & 0.01 & 0.01 & 0.02 \\
& & Detailed & 0.04 & 0.01 & 0.01 & 0.02 \\
\cmidrule{2-7}
& \multirow{3}{*}{$r_O$} 
  & Simplified & 0.02 & 0.00 & 0.01 & 0.01 \\
& & Detailed & 0.01 & 0.00 & 0.01 & 0.01 \\
\bottomrule
\end{tabular}
}
\caption{
Statistical results for MATH-Geometry using Llama-2-7b-hf on correct responses.
}
\label{tbl:math_lighteval_math_llama2_7b_grads_geometry}
\end{table}

\begin{figure}[!h]
\centering
\includegraphics[width=\linewidth]{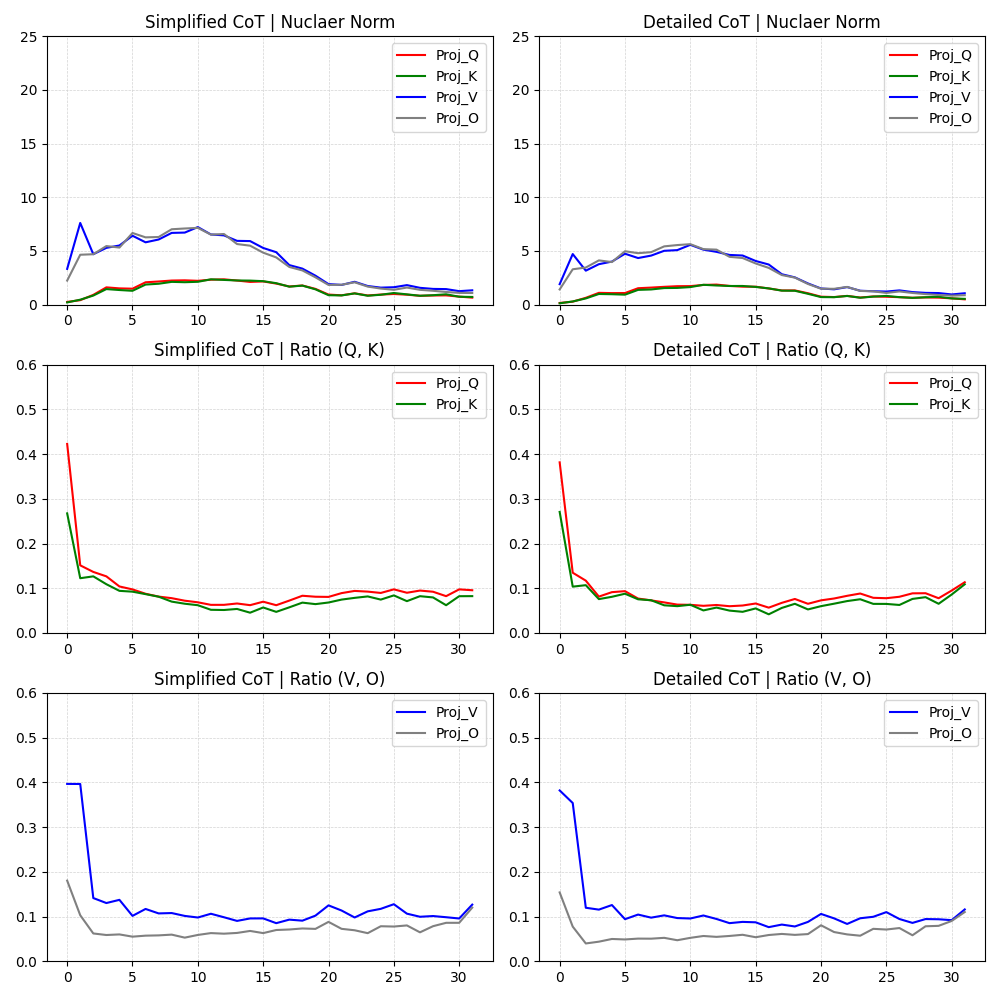}
\caption{Visualization for MATH-Geometry using Llama-2-7b-hf on correct responses.}
\label{fig:math_correct_geometry}
\vspace{-2mm}
\end{figure}

The visualizations and statistical results on other reasoning tasks: AQuA (Figure \ref{fig:reasoning_correct_aqua}, Table \ref{tbl:reasoning_cot_500_llama2_7b_grads_aqua_train}), GSM8K (Figure \ref{fig:reasoning_correct_gsm8k}, Table \ref{tbl:reasoning_cot_500_llama2_7b_grads_gsm8k_train}), StrategyQA (Figure \ref{fig:reasoning_correct_strategyqa}, Table \ref{tbl:reasoning_cot_500_llama2_7b_grads_strategyqa_train}), ECQA (Figure \ref{fig:reasoning_correct_ecqa}, Table \ref{tbl:reasoning_cot_500_llama2_7b_grads_ecqa_train}), CREAK (Figure \ref{fig:reasoning_correct_creak}, Table \ref{tbl:reasoning_cot_500_llama2_7b_grads_creak_train}), Sensemaking (Figure \ref{fig:reasoning_correct_sensemaking}, Table \ref{tbl:reasoning_cot_500_llama2_7b_grads_sensemaking_train}).

\begin{table}[!tbh]
\centering
\small
\scalebox{0.75}{
\begin{tabular}{l|l|l|cccc}
\toprule
\multirow{2}{*}{\textbf{Dataset}} & \multirow{2}{*}{\textbf{Curve}} & \multirow{2}{*}{\textbf{Cot}} &  \multicolumn{4}{c}{\textbf{Mean Absolute Difference (MAD)}} \\
 & &  & Early & Middle & Last & All\\
\midrule
\multirow{24}{*}{AQuA} & \multirow{3}{*}{$s_Q$} 
  & None & 0.86 & 0.77 & 1.00 & 0.88 \\
& & Simplified & 0.48 & 0.33 & 0.18 & 0.33 \\
& & Detailed & 0.17 & 0.12 & 0.06 & 0.11 \\
\cmidrule{2-7}
& \multirow{3}{*}{$s_K$} 
  & None & 1.16 & 0.95 & 1.77 & 1.28 \\
& & Simplified & 0.45 & 0.33 & 0.28 & 0.35 \\
& & Detailed & 0.16 & 0.12 & 0.08 & 0.12 \\
\cmidrule{2-7}
& \multirow{3}{*}{$s_V$} 
  & None & 13.84 & 1.70 & 0.64 & 4.81 \\
& & Simplified & 2.43 & 0.91 & 0.27 & 1.11 \\
& & Detailed & 0.75 & 0.36 & 0.09 & 0.38 \\
\cmidrule{2-7}
& \multirow{3}{*}{$s_O$} 
  & None & 4.38 & 1.26 & 0.42 & 1.89 \\
& & Simplified & 1.37 & 0.78 & 0.21 & 0.73 \\
& & Detailed & 0.51 & 0.37 & 0.09 & 0.30 \\
\cmidrule{2-7}
 & \multirow{3}{*}{$r_Q$} 
  & None & 0.07 & 0.05 & 0.13 & 0.08 \\
& & Simplified & 0.03 & 0.01 & 0.01 & 0.02 \\
& & Detailed & 0.03 & 0.00 & 0.01 & 0.02 \\
\cmidrule{2-7}
& \multirow{3}{*}{$r_K$} 
  & None & 0.04 & 0.02 & 0.08 & 0.04 \\
& & Simplified & 0.03 & 0.01 & 0.01 & 0.02 \\
& & Detailed & 0.03 & 0.01 & 0.01 & 0.01 \\
\cmidrule{2-7}
 & \multirow{3}{*}{$r_V$} 
  & None & 0.07 & 0.02 & 0.03 & 0.04 \\
& & Simplified & 0.06 & 0.01 & 0.01 & 0.02 \\
& & Detailed & 0.04 & 0.01 & 0.01 & 0.02 \\
\cmidrule{2-7}
& \multirow{3}{*}{$r_O$} 
  & None & 0.04 & 0.03 & 0.08 & 0.05 \\
& & Simplified & 0.02 & 0.01 & 0.02 & 0.02 \\
& & Detailed & 0.02 & 0.00 & 0.01 & 0.01 \\
\bottomrule
\end{tabular}
}
\caption{
Statistical results for AQuA using Llama-2-7b-hf on correct responses.
}
\label{tbl:reasoning_cot_500_llama2_7b_grads_aqua_train}
\end{table}

\begin{figure*}[!h]
\centering
\includegraphics[width=\linewidth]{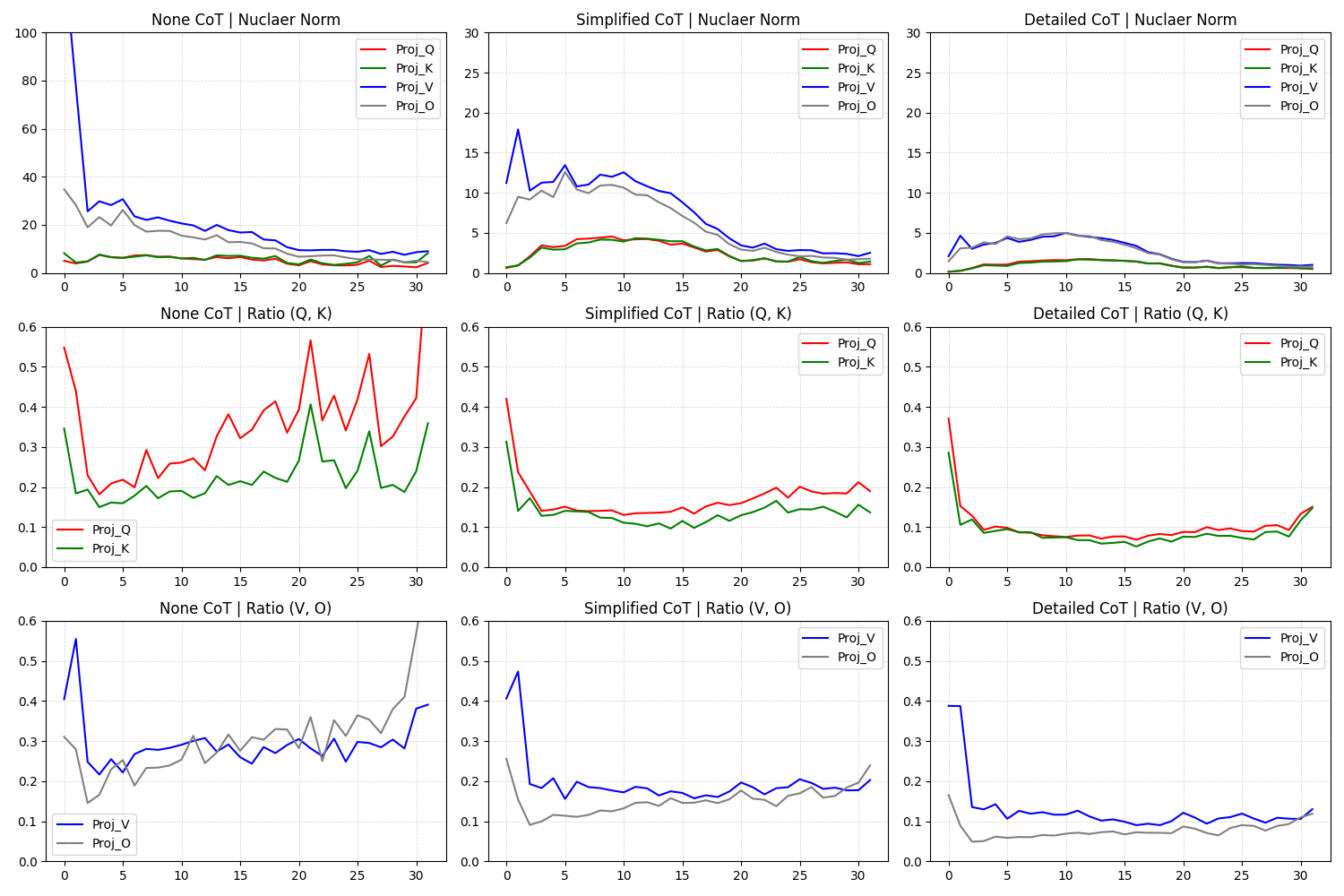}
\caption{Visualization for AQuA using Llama-2-7b-hf on correct responses.}
\label{fig:reasoning_correct_aqua}
\vspace{-2mm}
\end{figure*}

\begin{table}[!tbh]
\centering
\small
\scalebox{0.75}{
\begin{tabular}{l|l|l|cccc}
\toprule
\multirow{2}{*}{\textbf{Dataset}} & \multirow{2}{*}{\textbf{Curve}} & \multirow{2}{*}{\textbf{Cot}} &  \multicolumn{4}{c}{\textbf{Mean Absolute Difference (MAD)}} \\
 & &  & Early & Middle & Last & All\\
\midrule
\multirow{24}{*}{GSM8K} & \multirow{3}{*}{$s_Q$} 
  & None & 1.17 & 0.57 & 0.69 & 0.81 \\
& & Simplified & 0.29 & 0.19 & 0.10 & 0.19 \\
& & Detailed & 0.14 & 0.09 & 0.06 & 0.10 \\
\cmidrule{2-7}
& \multirow{3}{*}{$s_K$} 
  & None & 0.97 & 0.65 & 1.09 & 0.92 \\
& & Simplified & 0.28 & 0.18 & 0.14 & 0.20 \\
& & Detailed & 0.13 & 0.10 & 0.08 & 0.11 \\
\cmidrule{2-7}
& \multirow{3}{*}{$s_V$} 
  & None & 7.00 & 1.77 & 1.23 & 3.04 \\
& & Simplified & 1.38 & 0.59 & 0.16 & 0.67 \\
& & Detailed & 0.74 & 0.31 & 0.08 & 0.36 \\
\cmidrule{2-7}
& \multirow{3}{*}{$s_O$} 
  & None & 2.52 & 1.27 & 0.87 & 1.50 \\
& & Simplified & 0.66 & 0.54 & 0.13 & 0.42 \\
& & Detailed & 0.43 & 0.32 & 0.09 & 0.26 \\
\cmidrule{2-7}
 & \multirow{3}{*}{$r_Q$} 
  & None & 0.03 & 0.03 & 0.06 & 0.04 \\
& & Simplified & 0.03 & 0.01 & 0.01 & 0.02 \\
& & Detailed & 0.04 & 0.01 & 0.02 & 0.02 \\
\cmidrule{2-7}
& \multirow{3}{*}{$r_K$} 
  & None & 0.04 & 0.02 & 0.05 & 0.04 \\
& & Simplified & 0.03 & 0.01 & 0.01 & 0.02 \\
& & Detailed & 0.04 & 0.01 & 0.02 & 0.02 \\
\cmidrule{2-7}
 & \multirow{3}{*}{$r_V$} 
  & None & 0.06 & 0.01 & 0.03 & 0.03 \\
& & Simplified & 0.05 & 0.01 & 0.01 & 0.02 \\
& & Detailed & 0.04 & 0.01 & 0.01 & 0.02 \\
\cmidrule{2-7}
& \multirow{3}{*}{$r_O$} 
  & None & 0.05 & 0.02 & 0.06 & 0.04 \\
& & Simplified & 0.03 & 0.01 & 0.02 & 0.02 \\
& & Detailed & 0.02 & 0.00 & 0.01 & 0.01 \\
\bottomrule
\end{tabular}
}
\caption{
Statistical results for GSM8K using Llama-2-7b-hf on correct responses.
}
\label{tbl:reasoning_cot_500_llama2_7b_grads_gsm8k_train}
\end{table}

\begin{figure*}[!h]
\centering
\includegraphics[width=\linewidth]{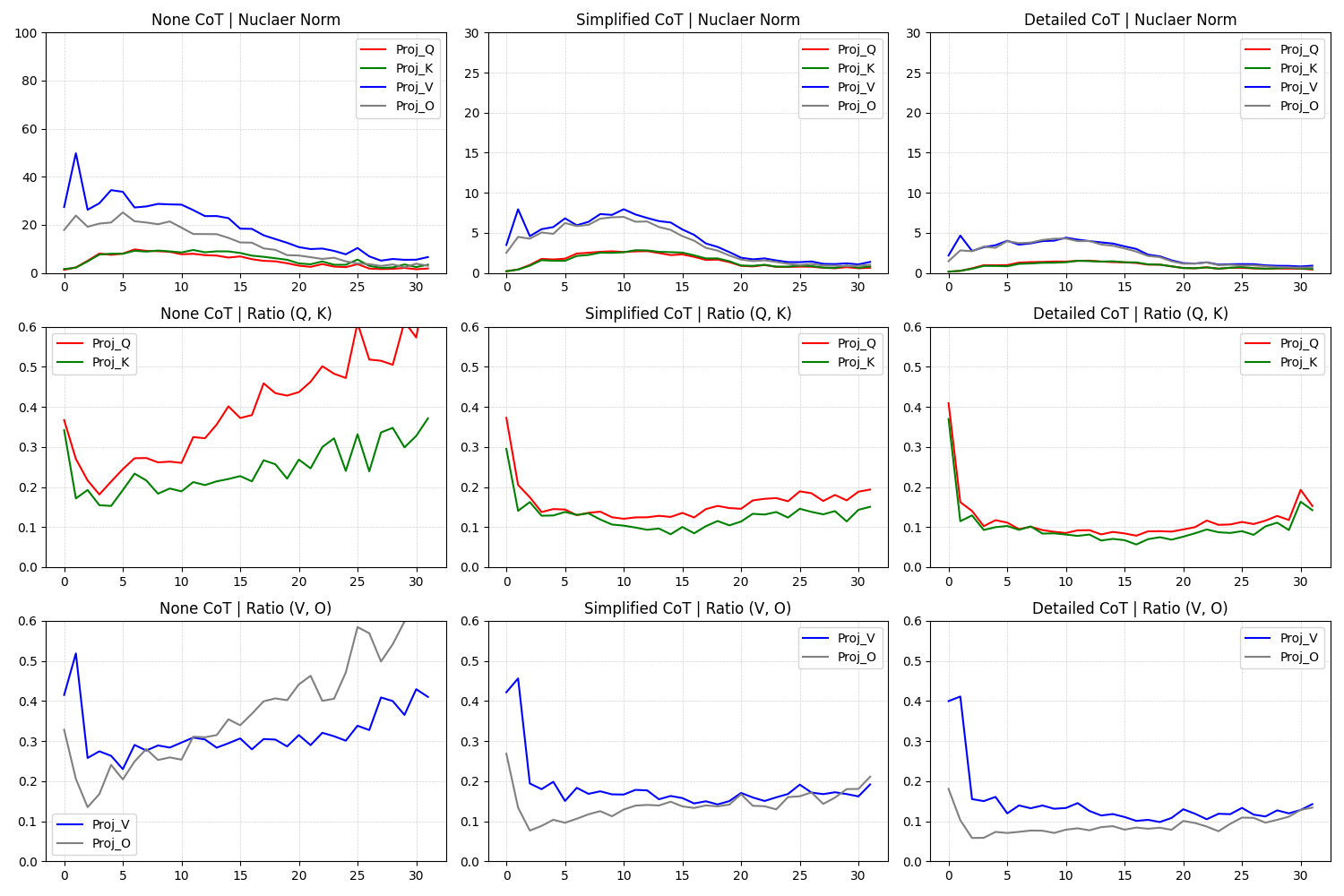}
\caption{Visualization for GSM8K using Llama-2-7b-hf on correct responses.}
\label{fig:reasoning_correct_gsm8k}
\vspace{-2mm}
\end{figure*}

\begin{table}[!tbh]
\centering
\small
\scalebox{0.75}{
\begin{tabular}{l|l|l|cccc}
\toprule
\multirow{2}{*}{\textbf{Dataset}} & \multirow{2}{*}{\textbf{Curve}} & \multirow{2}{*}{\textbf{Cot}} &  \multicolumn{4}{c}{\textbf{Mean Absolute Difference (MAD)}} \\
 & &  & Early & Middle & Last & All\\
\midrule
\multirow{24}{*}{StrategyQA} & \multirow{3}{*}{$s_Q$} 
  & None & 1.84 & 1.74 & 0.82 & 1.39 \\
& & Simplified & 0.34 & 0.17 & 0.13 & 0.22 \\
& & Detailed & 0.20 & 0.10 & 0.08 & 0.12 \\
\cmidrule{2-7}
& \multirow{3}{*}{$s_K$} 
  & None & 1.85 & 2.14 & 2.49 & 2.05 \\
& & Simplified & 0.33 & 0.14 & 0.14 & 0.20 \\
& & Detailed & 0.19 & 0.10 & 0.07 & 0.12 \\
\cmidrule{2-7}
& \multirow{3}{*}{$s_V$} 
  & None & 15.66 & 3.58 & 3.69 & 7.06 \\
& & Simplified & 2.19 & 0.52 & 0.31 & 0.94 \\
& & Detailed & 0.90 & 0.36 & 0.19 & 0.46 \\
\cmidrule{2-7}
& \multirow{3}{*}{$s_O$} 
  & None & 5.49 & 1.91 & 4.27 & 3.86 \\
& & Simplified & 0.98 & 0.38 & 0.29 & 0.52 \\
& & Detailed & 0.58 & 0.39 & 0.19 & 0.37 \\
\cmidrule{2-7}
 & \multirow{3}{*}{$r_Q$} 
  & None & 0.04 & 0.08 & 0.08 & 0.07 \\
& & Simplified & 0.02 & 0.00 & 0.01 & 0.01 \\
& & Detailed & 0.03 & 0.01 & 0.01 & 0.01 \\
\cmidrule{2-7}
& \multirow{3}{*}{$r_K$} 
  & None & 0.02 & 0.07 & 0.06 & 0.05 \\
& & Simplified & 0.03 & 0.00 & 0.01 & 0.01 \\
& & Detailed & 0.03 & 0.01 & 0.01 & 0.02 \\
\cmidrule{2-7}
 & \multirow{3}{*}{$r_V$} 
  & None & 0.06 & 0.04 & 0.07 & 0.06 \\
& & Simplified & 0.05 & 0.01 & 0.02 & 0.03 \\
& & Detailed & 0.04 & 0.01 & 0.01 & 0.02 \\
\cmidrule{2-7}
& \multirow{3}{*}{$r_O$} 
  & None & 0.05 & 0.04 & 0.07 & 0.05 \\
& & Simplified & 0.03 & 0.01 & 0.01 & 0.02 \\
& & Detailed & 0.02 & 0.01 & 0.01 & 0.01 \\
\bottomrule
\end{tabular}
}
\caption{
Statistical results for StrategyQA using Llama-2-7b-hf on correct responses.
}
\label{tbl:reasoning_cot_500_llama2_7b_grads_strategyqa_train}
\end{table}

\begin{figure*}[!h]
\centering
\includegraphics[width=\linewidth]{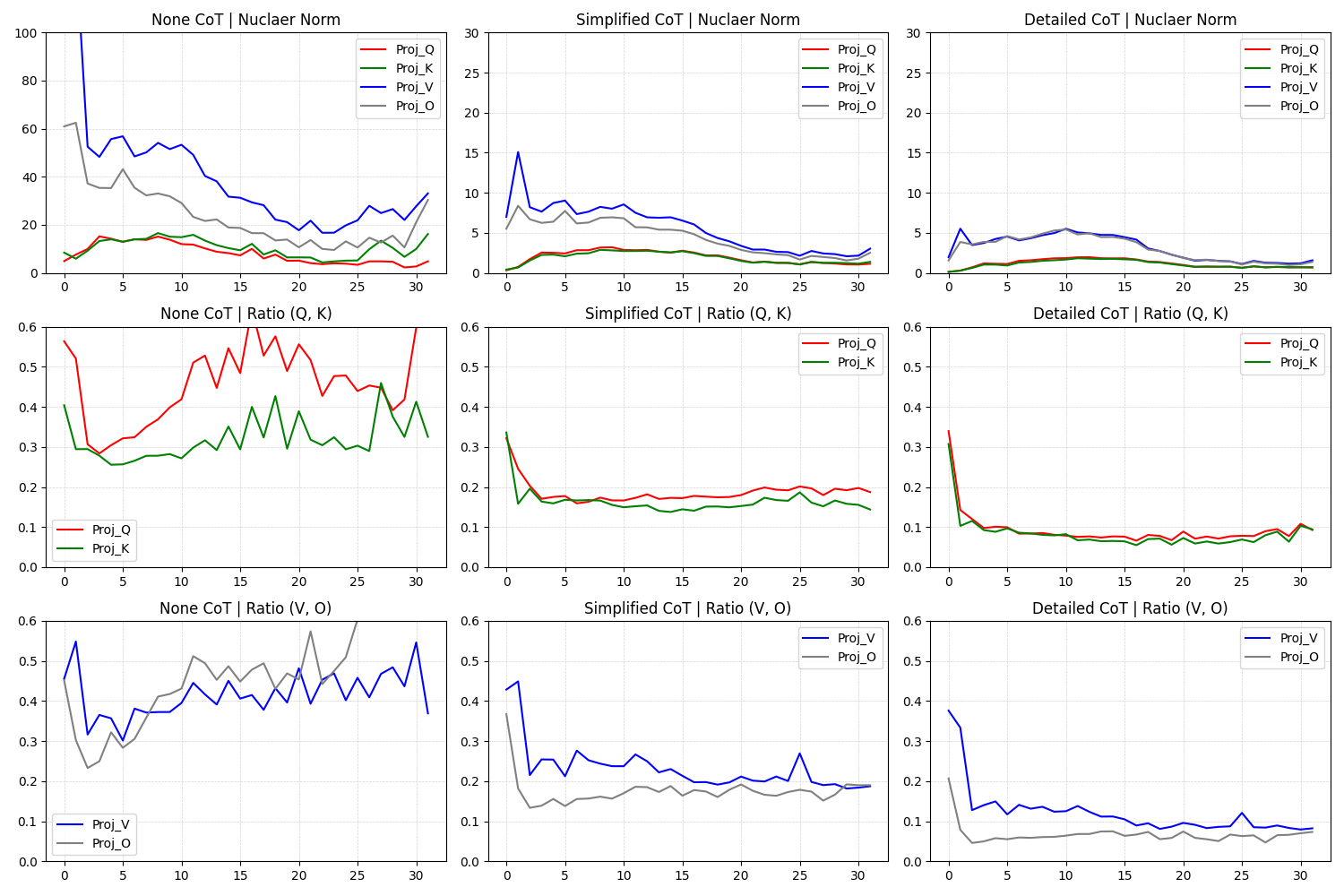}
\caption{Visualization for StrategyQA using Llama-2-7b-hf on correct responses.}
\label{fig:reasoning_correct_strategyqa}
\vspace{-2mm}
\end{figure*}

\begin{table}[!tbh]
\centering
\small
\scalebox{0.75}{
\begin{tabular}{l|l|l|cccc}
\toprule
\multirow{2}{*}{\textbf{Dataset}} & \multirow{2}{*}{\textbf{Curve}} & \multirow{2}{*}{\textbf{Cot}} &  \multicolumn{4}{c}{\textbf{Mean Absolute Difference (MAD)}} \\
 & &  & Early & Middle & Last & All\\
\midrule
\multirow{24}{*}{ECQA} & \multirow{3}{*}{$s_Q$} 
  & None & 1.03 & 0.70 & 0.84 & 0.86 \\
& & Simplified & 0.66 & 0.31 & 0.20 & 0.39 \\
& & Detailed & 0.27 & 0.14 & 0.11 & 0.17 \\
\cmidrule{2-7}
& \multirow{3}{*}{$s_K$} 
  & None & 1.34 & 0.70 & 2.12 & 1.39 \\
& & Simplified & 0.60 & 0.29 & 0.27 & 0.37 \\
& & Detailed & 0.25 & 0.14 & 0.11 & 0.16 \\
\cmidrule{2-7}
& \multirow{3}{*}{$s_V$} 
  & None & 11.00 & 1.60 & 1.88 & 4.40 \\
& & Simplified & 3.86 & 1.04 & 0.49 & 1.64 \\
& & Detailed & 1.18 & 0.51 & 0.26 & 0.62 \\
\cmidrule{2-7}
& \multirow{3}{*}{$s_O$} 
  & None & 4.11 & 1.24 & 1.45 & 2.17 \\
& & Simplified & 1.89 & 0.76 & 0.40 & 0.93 \\
& & Detailed & 0.76 & 0.49 & 0.25 & 0.47 \\
\cmidrule{2-7}
 & \multirow{3}{*}{$r_Q$} 
  & None & 0.05 & 0.01 & 0.06 & 0.04 \\
& & Simplified & 0.03 & 0.01 & 0.02 & 0.02 \\
& & Detailed & 0.03 & 0.01 & 0.01 & 0.02 \\
\cmidrule{2-7}
& \multirow{3}{*}{$r_K$} 
  & None & 0.03 & 0.02 & 0.03 & 0.03 \\
& & Simplified & 0.03 & 0.01 & 0.02 & 0.02 \\
& & Detailed & 0.03 & 0.01 & 0.02 & 0.02 \\
\cmidrule{2-7}
 & \multirow{3}{*}{$r_V$} 
  & None & 0.07 & 0.01 & 0.03 & 0.04 \\
& & Simplified & 0.05 & 0.01 & 0.02 & 0.02 \\
& & Detailed & 0.04 & 0.01 & 0.01 & 0.02 \\
\cmidrule{2-7}
& \multirow{3}{*}{$r_O$} 
  & None & 0.03 & 0.02 & 0.07 & 0.04 \\
& & Simplified & 0.04 & 0.01 & 0.02 & 0.02 \\
& & Detailed & 0.02 & 0.01 & 0.01 & 0.01 \\
\bottomrule
\end{tabular}
}
\caption{
Statistical results for ECQA using Llama-2-7b-hf on correct responses.
}
\label{tbl:reasoning_cot_500_llama2_7b_grads_ecqa_train}
\end{table}

\begin{figure*}[!h]
\centering
\includegraphics[width=\linewidth]{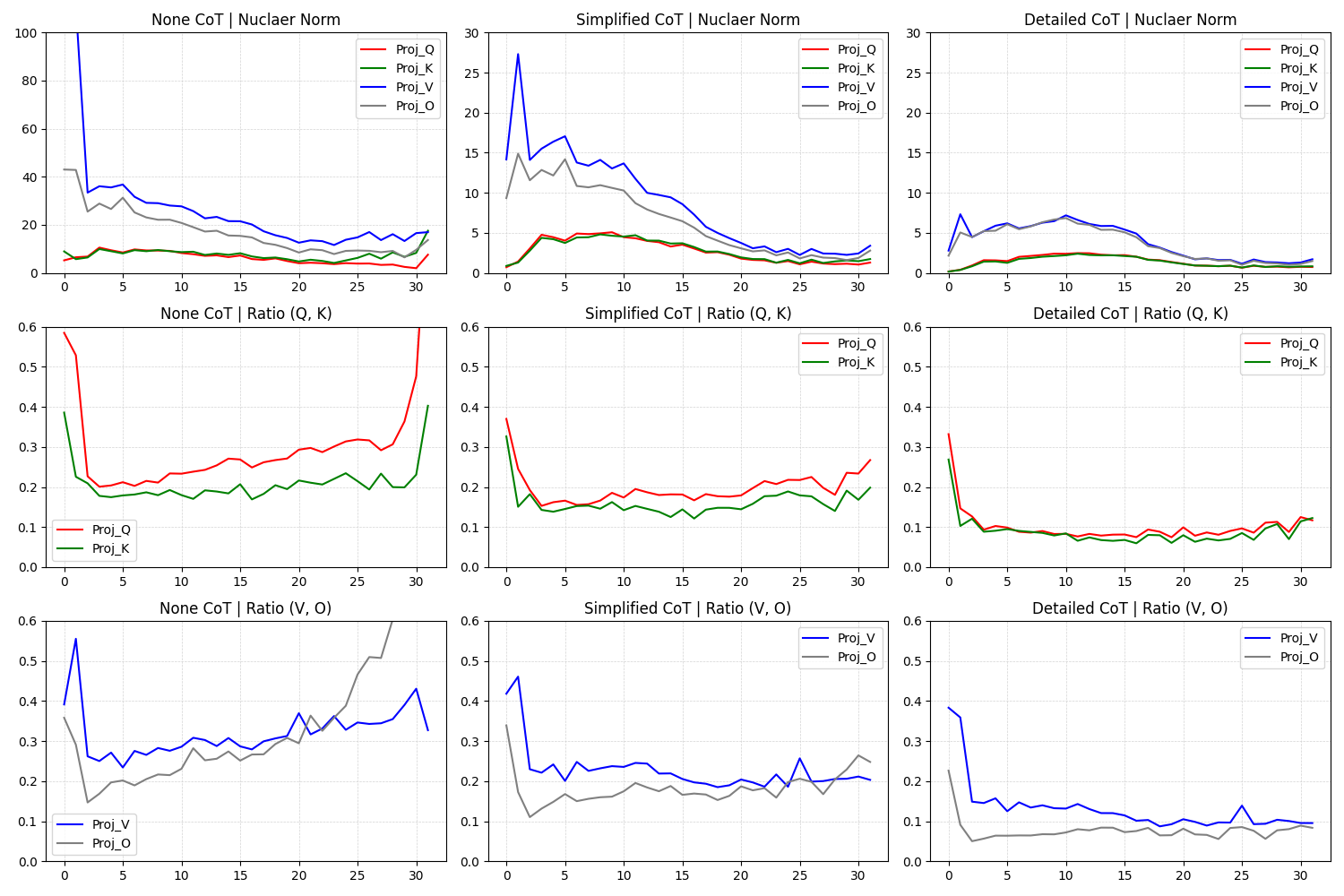}
\caption{Visualization for ECQA using Llama-2-7b-hf on correct responses.}
\label{fig:reasoning_correct_ecqa}
\vspace{-2mm}
\end{figure*}

\begin{table}[!tbh]
\centering
\small
\scalebox{0.75}{
\begin{tabular}{l|l|l|cccc}
\toprule
\multirow{2}{*}{\textbf{Dataset}} & \multirow{2}{*}{\textbf{Curve}} & \multirow{2}{*}{\textbf{Cot}} &  \multicolumn{4}{c}{\textbf{Mean Absolute Difference (MAD)}} \\
 & &  & Early & Middle & Last & All\\
\midrule
\multirow{24}{*}{CREAK} & \multirow{3}{*}{$s_Q$} 
  & None & 1.45 & 1.02 & 1.87 & 1.48 \\
& & Simplified & 0.51 & 0.26 & 0.19 & 0.32 \\
& & Detailed & 0.26 & 0.13 & 0.09 & 0.15 \\
\cmidrule{2-7}
& \multirow{3}{*}{$s_K$} 
  & None & 2.07 & 1.27 & 3.70 & 2.37 \\
& & Simplified & 0.45 & 0.23 & 0.25 & 0.31 \\
& & Detailed & 0.25 & 0.14 & 0.08 & 0.15 \\
\cmidrule{2-7}
& \multirow{3}{*}{$s_V$} 
  & None & 19.35 & 2.92 & 3.63 & 7.88 \\
& & Simplified & 3.00 & 0.82 & 0.49 & 1.32 \\
& & Detailed & 1.20 & 0.47 & 0.20 & 0.59 \\
\cmidrule{2-7}
& \multirow{3}{*}{$s_O$} 
  & None & 6.51 & 2.33 & 3.32 & 3.97 \\
& & Simplified & 1.48 & 0.57 & 0.40 & 0.76 \\
& & Detailed & 0.71 & 0.43 & 0.20 & 0.42 \\
\cmidrule{2-7}
 & \multirow{3}{*}{$r_Q$} 
  & None & 0.05 & 0.03 & 0.12 & 0.07 \\
& & Simplified & 0.03 & 0.01 & 0.01 & 0.01 \\
& & Detailed & 0.03 & 0.01 & 0.01 & 0.02 \\
\cmidrule{2-7}
& \multirow{3}{*}{$r_K$} 
  & None & 0.03 & 0.02 & 0.07 & 0.04 \\
& & Simplified & 0.03 & 0.01 & 0.01 & 0.01 \\
& & Detailed & 0.03 & 0.01 & 0.01 & 0.02 \\
\cmidrule{2-7}
 & \multirow{3}{*}{$r_V$} 
  & None & 0.07 & 0.02 & 0.05 & 0.05 \\
& & Simplified & 0.05 & 0.01 & 0.02 & 0.03 \\
& & Detailed & 0.04 & 0.01 & 0.01 & 0.02 \\
\cmidrule{2-7}
& \multirow{3}{*}{$r_O$} 
  & None & 0.04 & 0.02 & 0.07 & 0.04 \\
& & Simplified & 0.03 & 0.01 & 0.02 & 0.02 \\
& & Detailed & 0.02 & 0.01 & 0.01 & 0.01 \\
\bottomrule
\end{tabular}
}
\caption{
Statistical results for CREAK using Llama-2-7b-hf on correct responses.
}
\label{tbl:reasoning_cot_500_llama2_7b_grads_creak_train}
\end{table}

\begin{figure*}[!h]
\centering
\includegraphics[width=\linewidth]{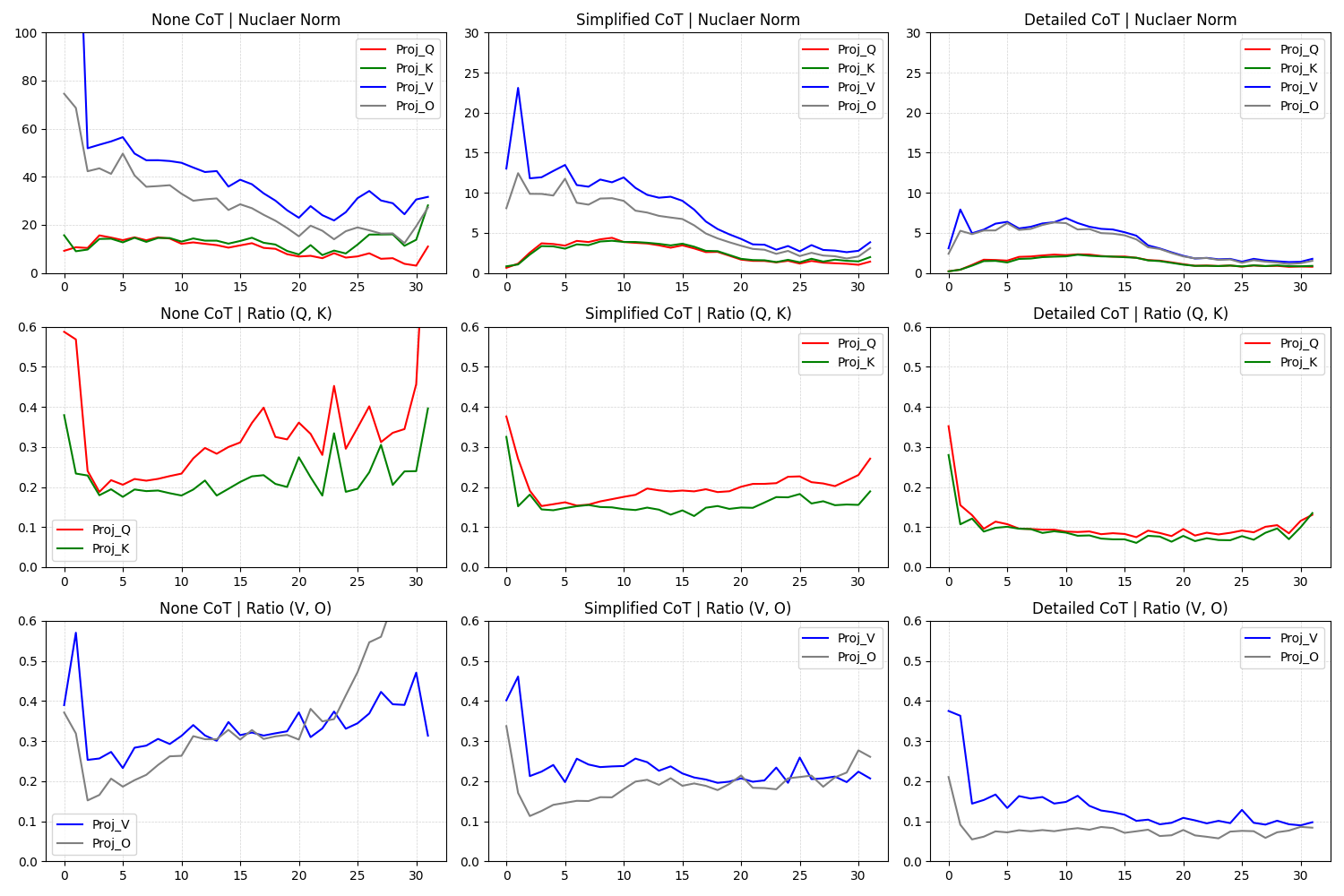}
\caption{Visualization for CREAK using Llama-2-7b-hf on correct responses.}
\label{fig:reasoning_correct_creak}
\vspace{-2mm}
\end{figure*}

\begin{table}[!tbh]
\centering
\small
\scalebox{0.75}{
\begin{tabular}{l|l|l|cccc}
\toprule
\multirow{2}{*}{\textbf{Dataset}} & \multirow{2}{*}{\textbf{Curve}} & \multirow{2}{*}{\textbf{Cot}} &  \multicolumn{4}{c}{\textbf{Mean Absolute Difference (MAD)}} \\
 & &  & Early & Middle & Last & All\\
\midrule
\multirow{24}{*}{Sensemaking} & \multirow{3}{*}{$s_Q$} 
  & None & 0.42 & 0.44 & 0.55 & 0.48 \\
& & Simplified & 0.54 & 0.34 & 0.30 & 0.40 \\
& & Detailed & 0.22 & 0.12 & 0.07 & 0.13 \\
\cmidrule{2-7}
& \multirow{3}{*}{$s_K$} 
  & None & 0.54 & 0.44 & 0.96 & 0.65 \\
& & Simplified & 0.49 & 0.27 & 0.46 & 0.41 \\
& & Detailed & 0.21 & 0.13 & 0.07 & 0.13 \\
\cmidrule{2-7}
& \multirow{3}{*}{$s_V$} 
  & None & 5.18 & 0.90 & 0.50 & 1.99 \\
& & Simplified & 2.87 & 1.01 & 0.63 & 1.39 \\
& & Detailed & 0.99 & 0.45 & 0.17 & 0.51 \\
\cmidrule{2-7}
& \multirow{3}{*}{$s_O$} 
  & None & 1.86 & 0.70 & 0.60 & 1.02 \\
& & Simplified & 1.76 & 0.77 & 0.46 & 0.93 \\
& & Detailed & 0.69 & 0.43 & 0.18 & 0.40 \\
\cmidrule{2-7}
 & \multirow{3}{*}{$r_Q$} 
  & None & 0.05 & 0.04 & 0.12 & 0.07 \\
& & Simplified & 0.03 & 0.01 & 0.04 & 0.03 \\
& & Detailed & 0.04 & 0.00 & 0.01 & 0.02 \\
\cmidrule{2-7}
& \multirow{3}{*}{$r_K$} 
  & None & 0.04 & 0.02 & 0.07 & 0.04 \\
& & Simplified & 0.03 & 0.01 & 0.03 & 0.02 \\
& & Detailed & 0.03 & 0.01 & 0.01 & 0.02 \\
\cmidrule{2-7}
 & \multirow{3}{*}{$r_V$} 
  & None & 0.06 & 0.02 & 0.04 & 0.04 \\
& & Simplified & 0.05 & 0.01 & 0.03 & 0.03 \\
& & Detailed & 0.04 & 0.01 & 0.01 & 0.02 \\
\cmidrule{2-7}
& \multirow{3}{*}{$r_O$} 
  & None & 0.03 & 0.02 & 0.06 & 0.04 \\
& & Simplified & 0.03 & 0.01 & 0.03 & 0.02 \\
& & Detailed & 0.02 & 0.00 & 0.01 & 0.01 \\
\bottomrule
\end{tabular}
}
\caption{
Statistical results for Sensemaking using Llama-2-7b-hf on correct responses.
}
\label{tbl:reasoning_cot_500_llama2_7b_grads_sensemaking_train}
\end{table}

\begin{figure*}[!h]
\centering
\includegraphics[width=\linewidth]{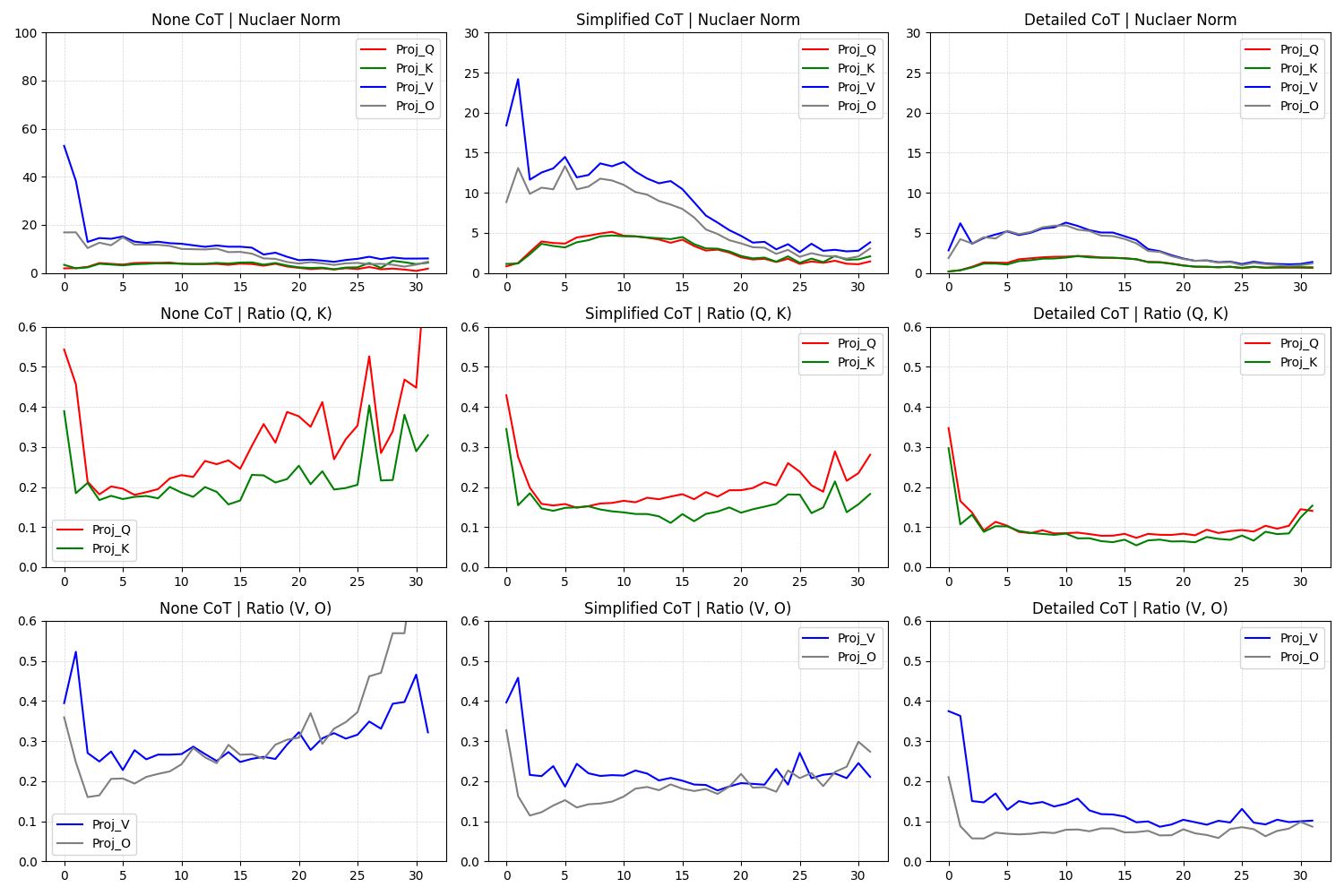}
\caption{Visualization for Sensemaking using Llama-2-7b-hf on correct responses.}
\label{fig:reasoning_correct_sensemaking}
\vspace{-2mm}
\end{figure*}

\subsubsection{Wiki Tasks}

The visualizations and statistical results on Wiki tasks are shown in Figure \ref{fig:wiki_correct} and Table \ref{tbl:wiki_wiki_popularity_llama2_7b_grads}.

\begin{table}[!tbh]
\centering
\small
\scalebox{0.75}{
\begin{tabular}{l|l|l|cccc}
\toprule
\multirow{2}{*}{\textbf{Dataset}} & \multirow{2}{*}{\textbf{Curve}} & \multirow{2}{*}{\textbf{Cot}} &  \multicolumn{4}{c}{\textbf{Mean Absolute Difference (MAD)}} \\
 & &  & Early & Middle & Last & All\\
\midrule
\multirow{32}{*}{Wiki} & \multirow{3}{*}{$s_Q$} 
  & Len 100 & 0.36 & 0.17 & 0.24 & 0.25 \\
& & Len 500 & 0.27 & 0.13 & 0.19 & 0.19 \\
& & Len 1000 & 0.23 & 0.12 & 0.16 & 0.16 \\
& & Unpopular & 0.36 & 0.21 & 0.29 & 0.30 \\
\cmidrule{2-7}
& \multirow{3}{*}{$s_K$} 
  & Len 100 & 0.35 & 0.17 & 0.27 & 0.26 \\
& & Len 500 & 0.25 & 0.12 & 0.22 & 0.19 \\
& & Len 1000 & 0.22 & 0.10 & 0.17 & 0.16 \\
& & Unpopular & 0.29 & 0.25 & 0.32 & 0.30 \\
\cmidrule{2-7}
& \multirow{3}{*}{$s_V$} 
  & Len 100 & 2.19 & 0.49 & 0.29 & 0.91 \\
& & Len 500 & 1.35 & 0.42 & 0.29 & 0.64 \\
& & Len 1000 & 1.05 & 0.40 & 0.28 & 0.55 \\
& & Unpopular & 3.83 & 0.44 & 0.45 & 1.42 \\
\cmidrule{2-7}
& \multirow{3}{*}{$s_O$} 
  & Len 100 & 1.15 & 0.44 & 0.28 & 0.58 \\
& & Len 500 & 0.86 & 0.46 & 0.28 & 0.50 \\
& & Len 1000 & 0.74 & 0.44 & 0.27 & 0.46 \\
& & Unpopular & 1.89 & 0.43 & 0.44 & 0.88 \\
\cmidrule{2-7}
 & \multirow{3}{*}{$r_Q$} 
  & Len 100 & 0.02 & 0.01 & 0.01 & 0.01 \\
& & Len 500 & 0.03 & 0.01 & 0.01 & 0.02 \\
& & Len 1000 & 0.04 & 0.01 & 0.01 & 0.02 \\
& & Unpopular & 0.03 & 0.02 & 0.02 & 0.02 \\
\cmidrule{2-7}
& \multirow{3}{*}{$r_K$} 
  & Len 100 & 0.03 & 0.02 & 0.01 & 0.02 \\
& & Len 500 & 0.02 & 0.01 & 0.01 & 0.01 \\
& & Len 1000 & 0.02 & 0.01 & 0.01 & 0.01 \\
& & Unpopular & 0.04 & 0.02 & 0.01 & 0.02 \\
\cmidrule{2-7}
 & \multirow{3}{*}{$r_V$} 
  & Len 100 & 0.05 & 0.02 & 0.01 & 0.02 \\
& & Len 500 & 0.04 & 0.01 & 0.01 & 0.02 \\
& & Len 1000 & 0.04 & 0.01 & 0.01 & 0.02 \\
& & Unpopular & 0.06 & 0.01 & 0.02 & 0.03 \\
\cmidrule{2-7}
& \multirow{3}{*}{$r_O$} 
  & Len 100 & 0.03 & 0.01 & 0.01 & 0.02 \\
& & Len 500 & 0.02 & 0.01 & 0.01 & 0.01 \\
& & Len 1000 & 0.02 & 0.00 & 0.01 & 0.01 \\
& & Unpopular & 0.02 & 0.02 & 0.02 & 0.02 \\
\bottomrule
\end{tabular}
}
\caption{
Statistical results for Wiki using Llama-2-7b-hf on correct responses.
}
\label{tbl:wiki_wiki_popularity_llama2_7b_grads}
\end{table}

\begin{figure*}[!h]
\centering
\includegraphics[width=\linewidth]{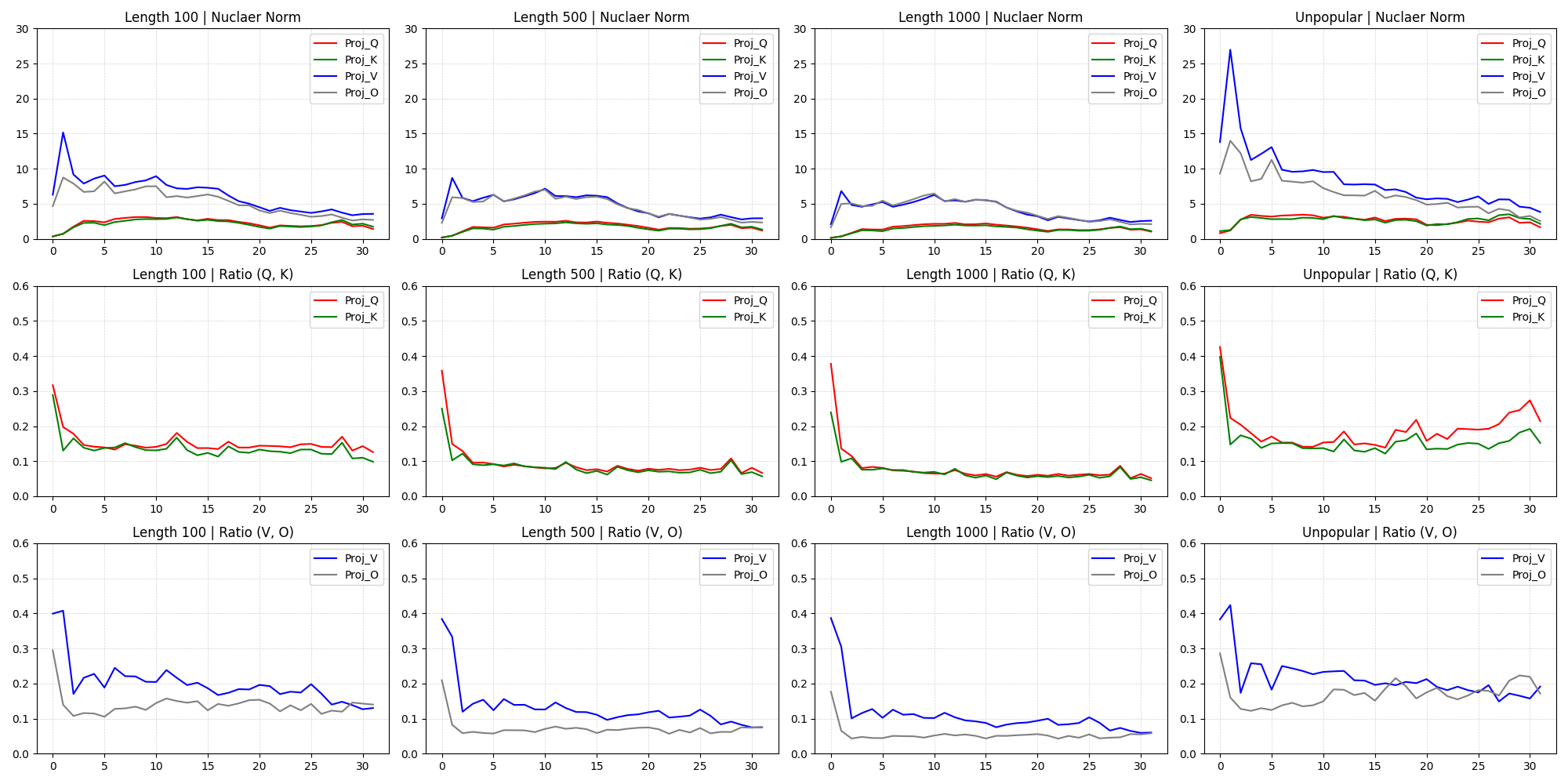}
\caption{Visualization for Wiki tasks using Llama-2-7b-hf on correct responses.}
\label{fig:wiki_correct}
\vspace{-2mm}
\end{figure*}

\subsection{Pre-trained LLM on Irrelevant Responses}

\subsubsection{Reasoning Tasks}

The visualizations and statistical results on MATH tasks: MATH-Algebra (Figure \ref{fig:math_nonsense_algebra}, Table \ref{tbl:math_lighteval_math_wrong_answer_shuffle_llama2_7b_grads_algebra}), MATH-Counting (Figure \ref{fig:math_nonsense_counting}, Table \ref{tbl:math_lighteval_math_wrong_answer_shuffle_llama2_7b_grads_counting_probability}), MATH-Geometry (Figure \ref{fig:math_nonsense_geometry}, Table \ref{tbl:math_lighteval_math_wrong_answer_shuffle_llama2_7b_grads_geometry}).

\begin{table}[!tbh]
\centering
\small
\scalebox{0.75}{
\begin{tabular}{l|l|l|cccc}
\toprule
\multirow{2}{*}{\textbf{Dataset}} & \multirow{2}{*}{\textbf{Curve}} & \multirow{2}{*}{\textbf{Cot}} &  \multicolumn{4}{c}{\textbf{Mean Absolute Difference (MAD)}} \\
 & &  & Early & Middle & Last & All\\
\midrule
\multirow{16}{*}{Algebra} & \multirow{3}{*}{$s_Q$} 
  & Simplified & 0.38 & 0.24 & 0.14 & 0.26 \\
& & Detailed & 0.23 & 0.14 & 0.09 & 0.16 \\
\cmidrule{2-7}
& \multirow{3}{*}{$s_K$} 
  & Simplified & 0.39 & 0.24 & 0.17 & 0.27 \\
& & Detailed & 0.24 & 0.16 & 0.12 & 0.17 \\
\cmidrule{2-7}
& \multirow{3}{*}{$s_V$} 
  & Simplified & 1.83 & 0.72 & 0.27 & 0.89 \\
& & Detailed & 1.01 & 0.49 & 0.16 & 0.52 \\
\cmidrule{2-7}
& \multirow{3}{*}{$s_O$} 
  & Simplified & 0.97 & 0.70 & 0.25 & 0.61 \\
& & Detailed & 0.62 & 0.47 & 0.16 & 0.39 \\
\cmidrule{2-7}
 & \multirow{3}{*}{$r_Q$} 
  & Simplified & 0.03 & 0.01 & 0.01 & 0.02 \\
& & Detailed & 0.04 & 0.01 & 0.01 & 0.02 \\
\cmidrule{2-7}
& \multirow{3}{*}{$r_K$} 
  & Simplified & 0.02 & 0.01 & 0.01 & 0.02 \\
& & Detailed & 0.03 & 0.01 & 0.01 & 0.01 \\
\cmidrule{2-7}
 & \multirow{3}{*}{$r_V$} 
  & Simplified & 0.04 & 0.01 & 0.01 & 0.02 \\
& & Detailed & 0.04 & 0.01 & 0.01 & 0.02 \\
\cmidrule{2-7}
& \multirow{3}{*}{$r_O$} 
  & Simplified & 0.02 & 0.00 & 0.01 & 0.01 \\
& & Detailed & 0.02 & 0.00 & 0.01 & 0.01 \\
\bottomrule
\end{tabular}
}
\caption{
Statistical results for MATH-Algebra using Llama-2-7b-hf on irrelevant responses.
}
\label{tbl:math_lighteval_math_wrong_answer_shuffle_llama2_7b_grads_algebra}
\end{table}

\begin{figure}[!h]
\centering
\includegraphics[width=\linewidth]{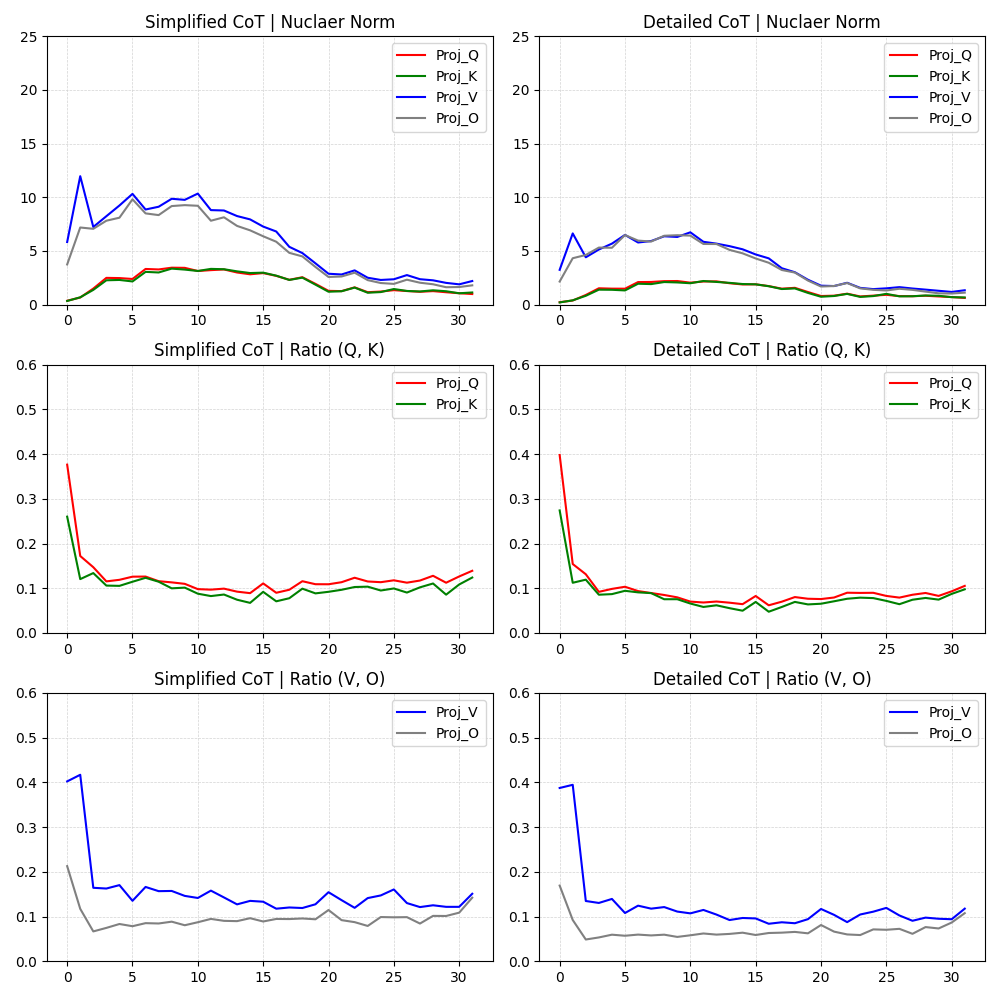}
\caption{Visualization for MATH-Algebra using Llama-2-7b-hf on irrelevant responses.}
\label{fig:math_nonsense_algebra}
\vspace{-2mm}
\end{figure}

\begin{table}[!tbh]
\centering
\small
\scalebox{0.75}{
\begin{tabular}{l|l|l|cccc}
\toprule
\multirow{2}{*}{\textbf{Dataset}} & \multirow{2}{*}{\textbf{Curve}} & \multirow{2}{*}{\textbf{Cot}} &  \multicolumn{4}{c}{\textbf{Mean Absolute Difference (MAD)}} \\
 & &  & Early & Middle & Last & All\\
\midrule
\multirow{16}{*}{Counting} & \multirow{3}{*}{$s_Q$} 
  & Simplified & 0.32 & 0.20 & 0.14 & 0.23 \\
& & Detailed & 0.25 & 0.13 & 0.10 & 0.16 \\
\cmidrule{2-7}
& \multirow{3}{*}{$s_K$} 
  & Simplified & 0.33 & 0.20 & 0.16 & 0.23 \\
& & Detailed & 0.26 & 0.14 & 0.12 & 0.17 \\
\cmidrule{2-7}
& \multirow{3}{*}{$s_V$} 
  & Simplified & 1.62 & 0.62 & 0.32 & 0.80 \\
& & Detailed & 1.13 & 0.46 & 0.20 & 0.56 \\
\cmidrule{2-7}
& \multirow{3}{*}{$s_O$} 
  & Simplified & 0.91 & 0.63 & 0.29 & 0.57 \\
& & Detailed & 0.69 & 0.46 & 0.18 & 0.42 \\
\cmidrule{2-7}
 & \multirow{3}{*}{$r_Q$} 
  & Simplified & 0.03 & 0.01 & 0.00 & 0.01 \\
& & Detailed & 0.04 & 0.01 & 0.01 & 0.02 \\
\cmidrule{2-7}
& \multirow{3}{*}{$r_K$} 
  & Simplified & 0.02 & 0.01 & 0.01 & 0.01 \\
& & Detailed & 0.03 & 0.01 & 0.01 & 0.01 \\
\cmidrule{2-7}
 & \multirow{3}{*}{$r_V$} 
  & Simplified & 0.04 & 0.01 & 0.01 & 0.02 \\
& & Detailed & 0.04 & 0.01 & 0.01 & 0.02 \\
\cmidrule{2-7}
& \multirow{3}{*}{$r_O$} 
  & Simplified & 0.02 & 0.00 & 0.01 & 0.01 \\
& & Detailed & 0.02 & 0.00 & 0.01 & 0.01 \\
\bottomrule
\end{tabular}
}
\caption{
Statistical results for MATH-Counting using Llama-2-7b-hf on irrelevant responses.
}
\label{tbl:math_lighteval_math_wrong_answer_shuffle_llama2_7b_grads_counting_probability}
\end{table}

\begin{figure}[!h]
\centering
\includegraphics[width=\linewidth]{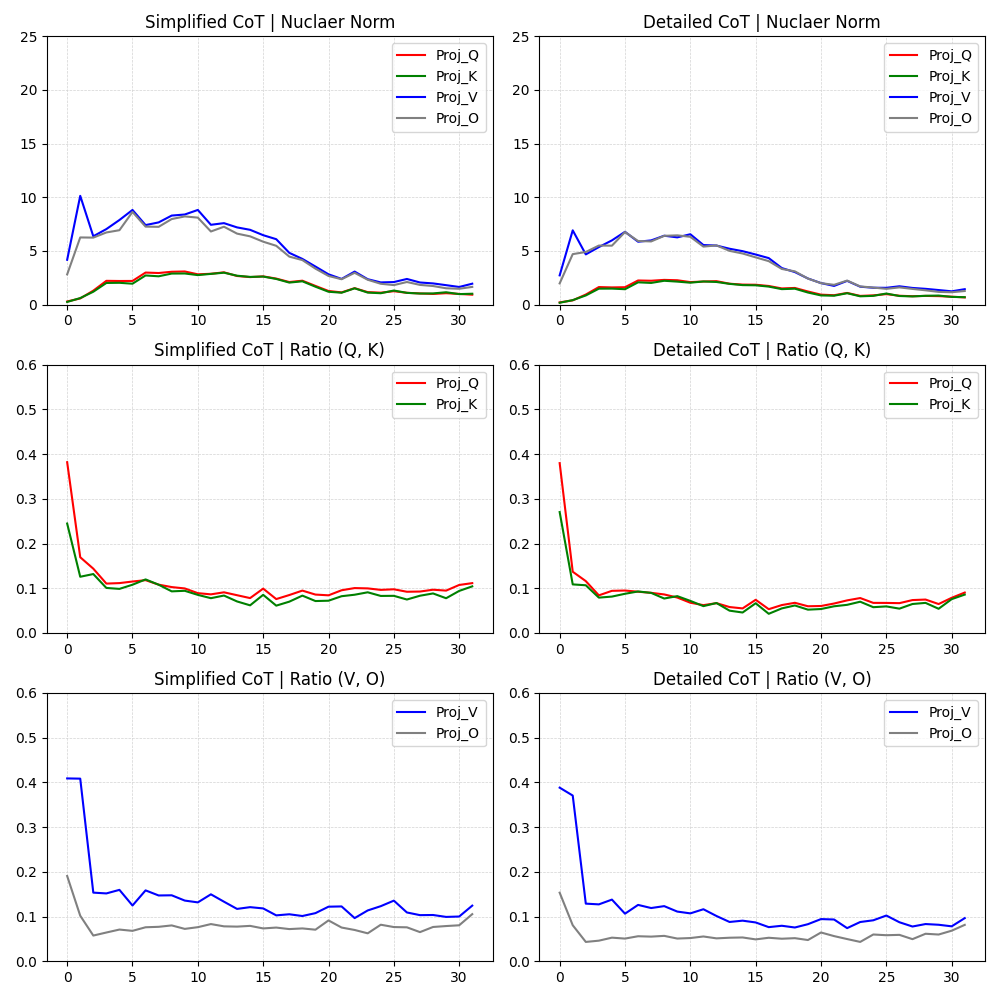}
\caption{Visualization for MATH-Counting using Llama-2-7b-hf on irrelevant responses.}
\label{fig:math_nonsense_counting}
\vspace{-2mm}
\end{figure}

\begin{table}[!tbh]
\centering
\small
\scalebox{0.75}{
\begin{tabular}{l|l|l|cccc}
\toprule
\multirow{2}{*}{\textbf{Dataset}} & \multirow{2}{*}{\textbf{Curve}} & \multirow{2}{*}{\textbf{Cot}} &  \multicolumn{4}{c}{\textbf{Mean Absolute Difference (MAD)}} \\
 & &  & Early & Middle & Last & All\\
\midrule
\multirow{16}{*}{Geometry} & \multirow{3}{*}{$s_Q$} 
  & Simplified & 0.32 & 0.16 & 0.14 & 0.21 \\
& & Detailed & 0.29 & 0.11 & 0.14 & 0.18 \\
\cmidrule{2-7}
& \multirow{3}{*}{$s_K$} 
  & Simplified & 0.33 & 0.17 & 0.15 & 0.22 \\
& & Detailed & 0.29 & 0.12 & 0.18 & 0.20 \\
\cmidrule{2-7}
& \multirow{3}{*}{$s_V$} 
  & Simplified & 1.50 & 0.57 & 0.25 & 0.73 \\
& & Detailed & 1.18 & 0.44 & 0.24 & 0.57 \\
\cmidrule{2-7}
& \multirow{3}{*}{$s_O$} 
  & Simplified & 0.84 & 0.59 & 0.23 & 0.52 \\
& & Detailed & 0.84 & 0.47 & 0.23 & 0.48 \\
\cmidrule{2-7}
 & \multirow{3}{*}{$r_Q$} 
  & Simplified & 0.04 & 0.01 & 0.01 & 0.02 \\
& & Detailed & 0.04 & 0.01 & 0.01 & 0.02 \\
\cmidrule{2-7}
& \multirow{3}{*}{$r_K$} 
  & Simplified & 0.02 & 0.01 & 0.01 & 0.01 \\
& & Detailed & 0.02 & 0.01 & 0.01 & 0.01 \\
\cmidrule{2-7}
 & \multirow{3}{*}{$r_V$} 
  & Simplified & 0.04 & 0.01 & 0.01 & 0.02 \\
& & Detailed & 0.04 & 0.01 & 0.01 & 0.02 \\
\cmidrule{2-7}
& \multirow{3}{*}{$r_O$} 
  & Simplified & 0.02 & 0.00 & 0.01 & 0.01 \\
& & Detailed & 0.02 & 0.00 & 0.01 & 0.01 \\
\bottomrule
\end{tabular}
}
\caption{
Statistical results for MATH-Geometry using Llama-2-7b-hf on irrelevant responses.
}
\label{tbl:math_lighteval_math_wrong_answer_shuffle_llama2_7b_grads_geometry}
\end{table}

\begin{figure}[!h]
\centering
\includegraphics[width=\linewidth]{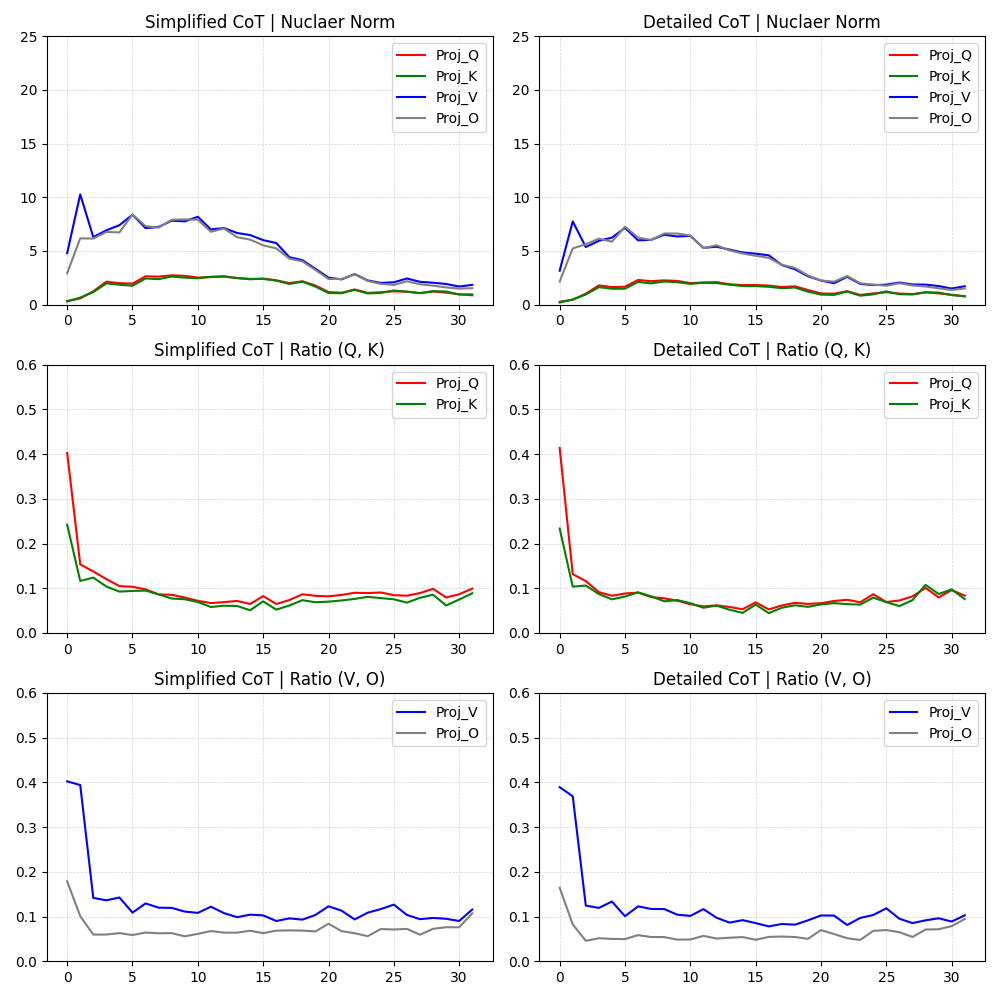}
\caption{Visualization for MATH-Geometry using Llama-2-7b-hf on irrelevant responses.}
\label{fig:math_nonsense_geometry}
\vspace{-2mm}
\end{figure}

The visualizations and statistical results on other reasoning tasks: AQuA (Figure \ref{fig:reasoning_nonsense_aqua}, Table \ref{tbl:reasoning_cot_500_wrong_answer_shuffle_llama2_7b_grads_aqua_train}), GSM8K (Figure \ref{fig:reasoning_nonsense_gsm8k}, Table \ref{tbl:reasoning_cot_500_wrong_answer_shuffle_llama2_7b_grads_gsm8k_train}), StrategyQA (Figure \ref{fig:reasoning_nonsense_strategyqa}, Table \ref{tbl:reasoning_cot_500_wrong_answer_shuffle_llama2_7b_grads_strategyqa_train}), ECQA (Figure \ref{fig:reasoning_nonsense_ecqa}, Table \ref{tbl:reasoning_cot_500_wrong_answer_shuffle_llama2_7b_grads_ecqa_train}), CREAK (Figure \ref{fig:reasoning_nonsense_creak}, Table \ref{tbl:reasoning_cot_500_wrong_answer_shuffle_llama2_7b_grads_creak_train}), Sensemaking (Figure \ref{fig:reasoning_nonsense_sensemaking}, Table \ref{tbl:reasoning_cot_500_wrong_answer_shuffle_llama2_7b_grads_sensemaking_train}).

\begin{table}[!tbh]
\centering
\small
\scalebox{0.75}{
\begin{tabular}{l|l|l|cccc}
\toprule
\multirow{2}{*}{\textbf{Dataset}} & \multirow{2}{*}{\textbf{Curve}} & \multirow{2}{*}{\textbf{Cot}} &  \multicolumn{4}{c}{\textbf{Mean Absolute Difference (MAD)}} \\
 & &  & Early & Middle & Last & All\\
\midrule
\multirow{24}{*}{AQuA} & \multirow{3}{*}{$s_Q$} 
  & None & 0.86 & 0.78 & 1.01 & 0.88 \\
& & Simplified & 0.54 & 0.35 & 0.23 & 0.38 \\
& & Detailed & 0.28 & 0.16 & 0.10 & 0.18 \\
\cmidrule{2-7}
& \multirow{3}{*}{$s_K$} 
  & None & 1.16 & 0.95 & 1.78 & 1.29 \\
& & Simplified & 0.50 & 0.34 & 0.31 & 0.39 \\
& & Detailed & 0.29 & 0.18 & 0.13 & 0.19 \\
\cmidrule{2-7}
& \multirow{3}{*}{$s_V$} 
  & None & 13.76 & 1.71 & 0.64 & 4.79 \\
& & Simplified & 2.93 & 0.89 & 0.43 & 1.28 \\
& & Detailed & 1.35 & 0.52 & 0.22 & 0.64 \\
\cmidrule{2-7}
& \multirow{3}{*}{$s_O$} 
  & None & 4.38 & 1.28 & 0.43 & 1.90 \\
& & Simplified & 1.58 & 0.82 & 0.34 & 0.86 \\
& & Detailed & 0.79 & 0.48 & 0.21 & 0.47 \\
\cmidrule{2-7}
 & \multirow{3}{*}{$r_Q$} 
  & None & 0.07 & 0.05 & 0.13 & 0.08 \\
& & Simplified & 0.04 & 0.01 & 0.01 & 0.02 \\
& & Detailed & 0.04 & 0.01 & 0.01 & 0.02 \\
\cmidrule{2-7}
& \multirow{3}{*}{$r_K$} 
  & None & 0.04 & 0.02 & 0.08 & 0.04 \\
& & Simplified & 0.03 & 0.01 & 0.01 & 0.02 \\
& & Detailed & 0.03 & 0.01 & 0.01 & 0.01 \\
\cmidrule{2-7}
 & \multirow{3}{*}{$r_V$} 
  & None & 0.07 & 0.02 & 0.03 & 0.04 \\
& & Simplified & 0.06 & 0.01 & 0.01 & 0.03 \\
& & Detailed & 0.04 & 0.01 & 0.01 & 0.02 \\
\cmidrule{2-7}
& \multirow{3}{*}{$r_O$} 
  & None & 0.04 & 0.03 & 0.08 & 0.05 \\
& & Simplified & 0.02 & 0.01 & 0.02 & 0.02 \\
& & Detailed & 0.02 & 0.00 & 0.01 & 0.01 \\
\bottomrule
\end{tabular}
}
\caption{
Statistical results for AQuA using Llama-2-7b-hf on irrelevant responses.
}
\label{tbl:reasoning_cot_500_wrong_answer_shuffle_llama2_7b_grads_aqua_train}
\end{table}

\begin{figure*}[!h]
\centering
\includegraphics[width=\linewidth]{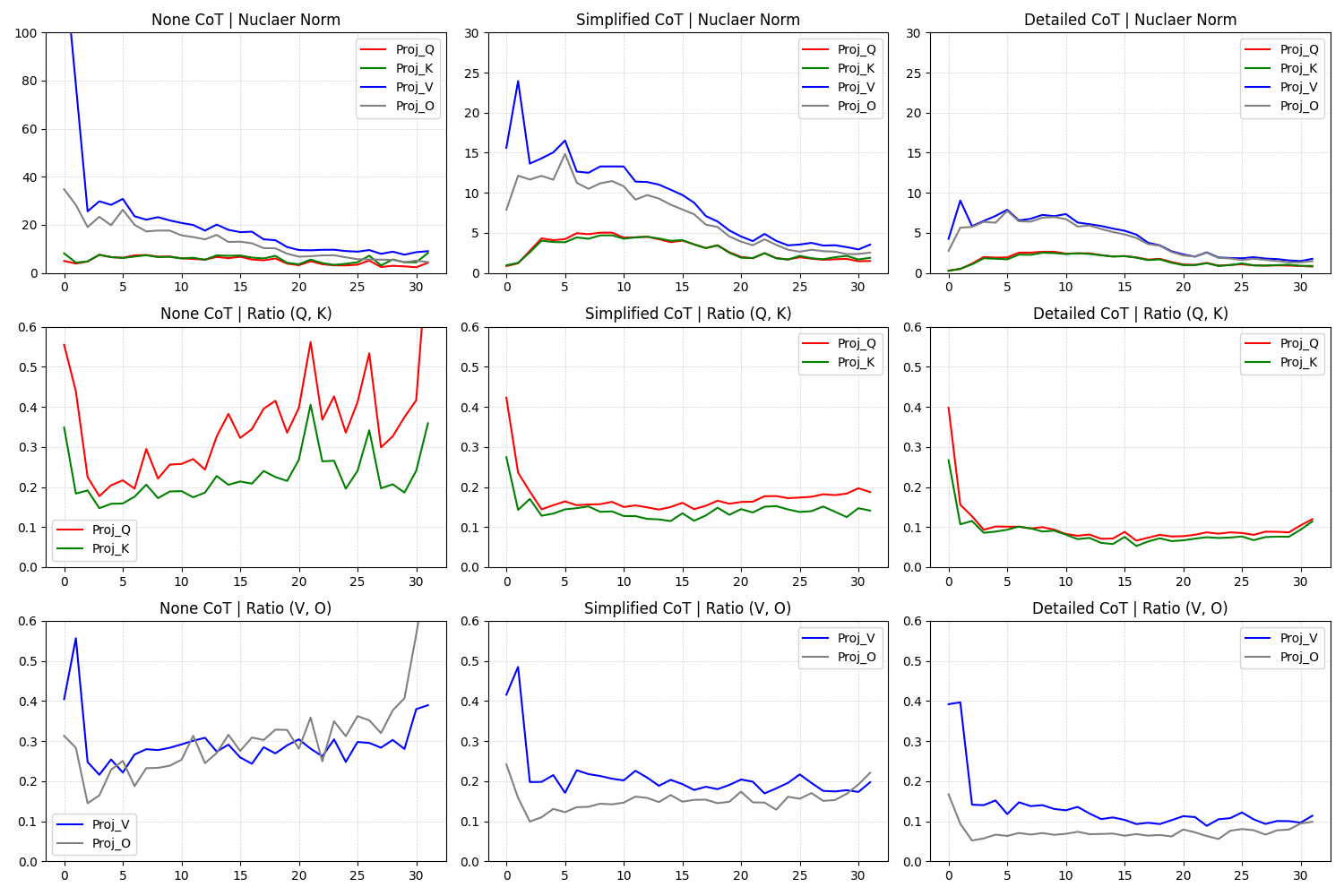}
\caption{Visualization for AQuA using Llama-2-7b-hf on irrelevant responses.}
\label{fig:reasoning_nonsense_aqua}
\vspace{-2mm}
\end{figure*}

\begin{table}[!tbh]
\centering
\small
\scalebox{0.75}{
\begin{tabular}{l|l|l|cccc}
\toprule
\multirow{2}{*}{\textbf{Dataset}} & \multirow{2}{*}{\textbf{Curve}} & \multirow{2}{*}{\textbf{Cot}} &  \multicolumn{4}{c}{\textbf{Mean Absolute Difference (MAD)}} \\
 & &  & Early & Middle & Last & All\\
\midrule
\multirow{24}{*}{GSM8K} & \multirow{3}{*}{$s_Q$} 
  & None & 1.29 & 0.60 & 0.80 & 0.90 \\
& & Simplified & 0.41 & 0.28 & 0.17 & 0.29 \\
& & Detailed & 0.26 & 0.15 & 0.12 & 0.18 \\
\cmidrule{2-7}
& \multirow{3}{*}{$s_K$} 
  & None & 1.06 & 0.64 & 1.26 & 1.00 \\
& & Simplified & 0.39 & 0.27 & 0.21 & 0.29 \\
& & Detailed & 0.25 & 0.16 & 0.14 & 0.18 \\
\cmidrule{2-7}
& \multirow{3}{*}{$s_V$} 
  & None & 7.23 & 1.79 & 1.27 & 3.13 \\
& & Simplified & 2.24 & 0.71 & 0.38 & 1.01 \\
& & Detailed & 1.35 & 0.47 & 0.23 & 0.63 \\
\cmidrule{2-7}
& \multirow{3}{*}{$s_O$} 
  & None & 2.64 & 1.31 & 0.89 & 1.55 \\
& & Simplified & 1.02 & 0.65 & 0.35 & 0.64 \\
& & Detailed & 0.74 & 0.45 & 0.23 & 0.44 \\
\cmidrule{2-7}
 & \multirow{3}{*}{$r_Q$} 
  & None & 0.03 & 0.03 & 0.07 & 0.04 \\
& & Simplified & 0.03 & 0.01 & 0.01 & 0.02 \\
& & Detailed & 0.04 & 0.01 & 0.01 & 0.02 \\
\cmidrule{2-7}
& \multirow{3}{*}{$r_K$} 
  & None & 0.04 & 0.01 & 0.06 & 0.04 \\
& & Simplified & 0.03 & 0.01 & 0.01 & 0.02 \\
& & Detailed & 0.03 & 0.01 & 0.01 & 0.02 \\
\cmidrule{2-7}
 & \multirow{3}{*}{$r_V$} 
  & None & 0.06 & 0.01 & 0.03 & 0.03 \\
& & Simplified & 0.05 & 0.01 & 0.02 & 0.02 \\
& & Detailed & 0.04 & 0.01 & 0.01 & 0.02 \\
\cmidrule{2-7}
& \multirow{3}{*}{$r_O$} 
  & None & 0.05 & 0.02 & 0.06 & 0.04 \\
& & Simplified & 0.03 & 0.01 & 0.01 & 0.02 \\
& & Detailed & 0.02 & 0.00 & 0.01 & 0.01 \\
\bottomrule
\end{tabular}
}
\caption{
Statistical results for GSM8K using Llama-2-7b-hf on irrelevant responses.
}
\label{tbl:reasoning_cot_500_wrong_answer_shuffle_llama2_7b_grads_gsm8k_train}
\end{table}

\begin{figure*}[!h]
\centering
\includegraphics[width=\linewidth]{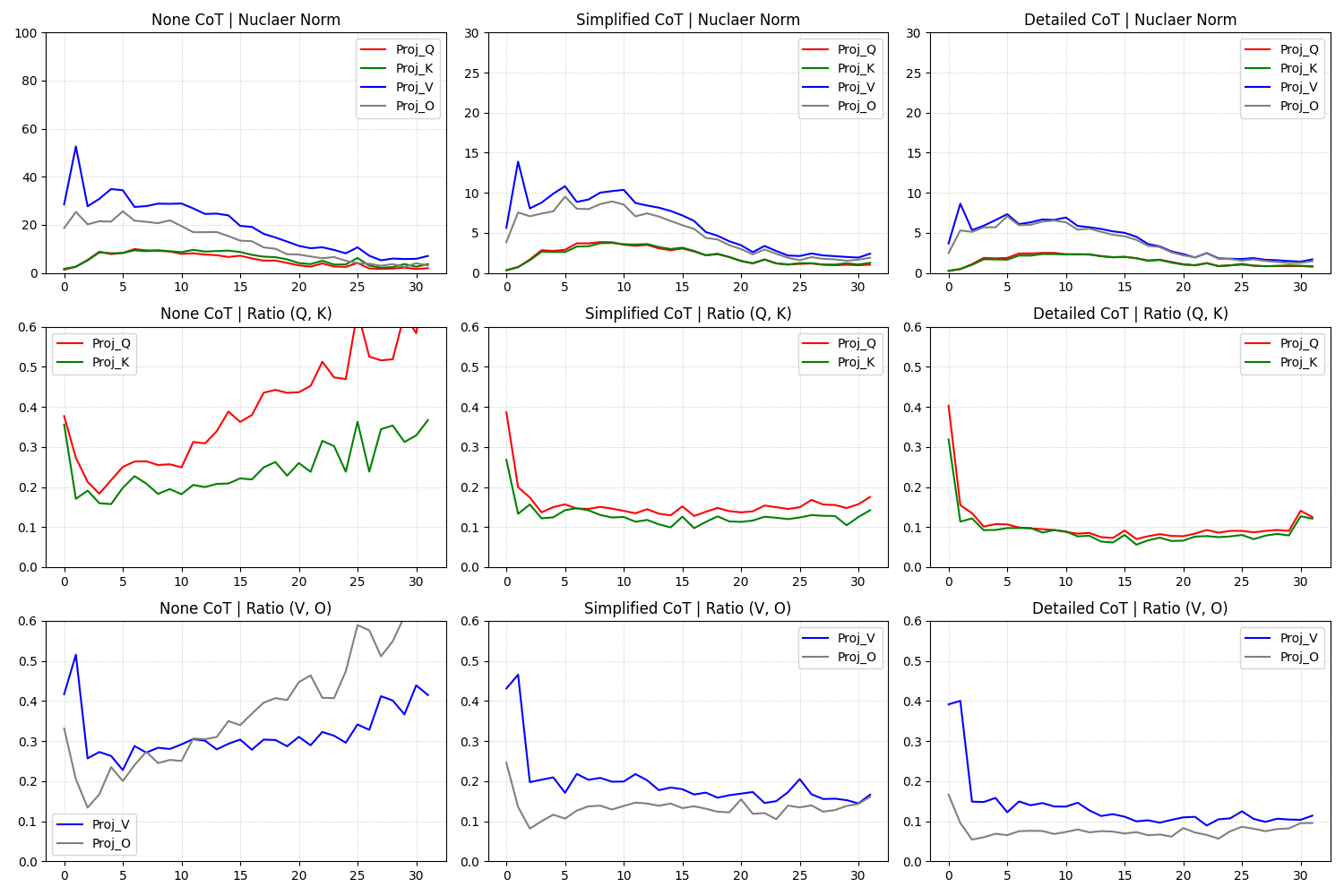}
\caption{Visualization for GSM8K using Llama-2-7b-hf on irrelevant responses.}
\label{fig:reasoning_nonsense_gsm8k}
\vspace{-2mm}
\end{figure*}

\begin{table}[!tbh]
\centering
\small
\scalebox{0.75}{
\begin{tabular}{l|l|l|cccc}
\toprule
\multirow{2}{*}{\textbf{Dataset}} & \multirow{2}{*}{\textbf{Curve}} & \multirow{2}{*}{\textbf{Cot}} &  \multicolumn{4}{c}{\textbf{Mean Absolute Difference (MAD)}} \\
 & &  & Early & Middle & Last & All\\
\midrule
\multirow{24}{*}{StrategyQA} & \multirow{3}{*}{$s_Q$} 
  & None & 1.74 & 1.59 & 0.85 & 1.35 \\
& & Simplified & 0.52 & 0.25 & 0.17 & 0.31 \\
& & Detailed & 0.29 & 0.14 & 0.09 & 0.17 \\
\cmidrule{2-7}
& \multirow{3}{*}{$s_K$} 
  & None & 1.93 & 2.01 & 2.16 & 1.94 \\
& & Simplified & 0.53 & 0.23 & 0.18 & 0.30 \\
& & Detailed & 0.29 & 0.14 & 0.08 & 0.16 \\
\cmidrule{2-7}
& \multirow{3}{*}{$s_V$} 
  & None & 16.91 & 3.74 & 3.63 & 7.44 \\
& & Simplified & 3.33 & 0.65 & 0.52 & 1.35 \\
& & Detailed & 1.44 & 0.46 & 0.32 & 0.68 \\
\cmidrule{2-7}
& \multirow{3}{*}{$s_O$} 
  & None & 5.70 & 1.92 & 4.24 & 3.92 \\
& & Simplified & 1.60 & 0.42 & 0.44 & 0.78 \\
& & Detailed & 0.91 & 0.39 & 0.33 & 0.51 \\
\cmidrule{2-7}
 & \multirow{3}{*}{$r_Q$} 
  & None & 0.05 & 0.08 & 0.08 & 0.07 \\
& & Simplified & 0.02 & 0.01 & 0.01 & 0.01 \\
& & Detailed & 0.03 & 0.01 & 0.01 & 0.01 \\
\cmidrule{2-7}
& \multirow{3}{*}{$r_K$} 
  & None & 0.02 & 0.06 & 0.05 & 0.04 \\
& & Simplified & 0.03 & 0.01 & 0.01 & 0.02 \\
& & Detailed & 0.02 & 0.01 & 0.01 & 0.01 \\
\cmidrule{2-7}
 & \multirow{3}{*}{$r_V$} 
  & None & 0.06 & 0.04 & 0.07 & 0.06 \\
& & Simplified & 0.05 & 0.01 & 0.02 & 0.03 \\
& & Detailed & 0.04 & 0.01 & 0.01 & 0.02 \\
\cmidrule{2-7}
& \multirow{3}{*}{$r_O$} 
  & None & 0.05 & 0.04 & 0.07 & 0.05 \\
& & Simplified & 0.03 & 0.01 & 0.01 & 0.02 \\
& & Detailed & 0.02 & 0.00 & 0.01 & 0.01 \\
\bottomrule
\end{tabular}
}
\caption{
Statistical results for StrategyQA using Llama-2-7b-hf on irrelevant responses.
}
\label{tbl:reasoning_cot_500_wrong_answer_shuffle_llama2_7b_grads_strategyqa_train}
\end{table}

\begin{figure*}[!h]
\centering
\includegraphics[width=\linewidth]{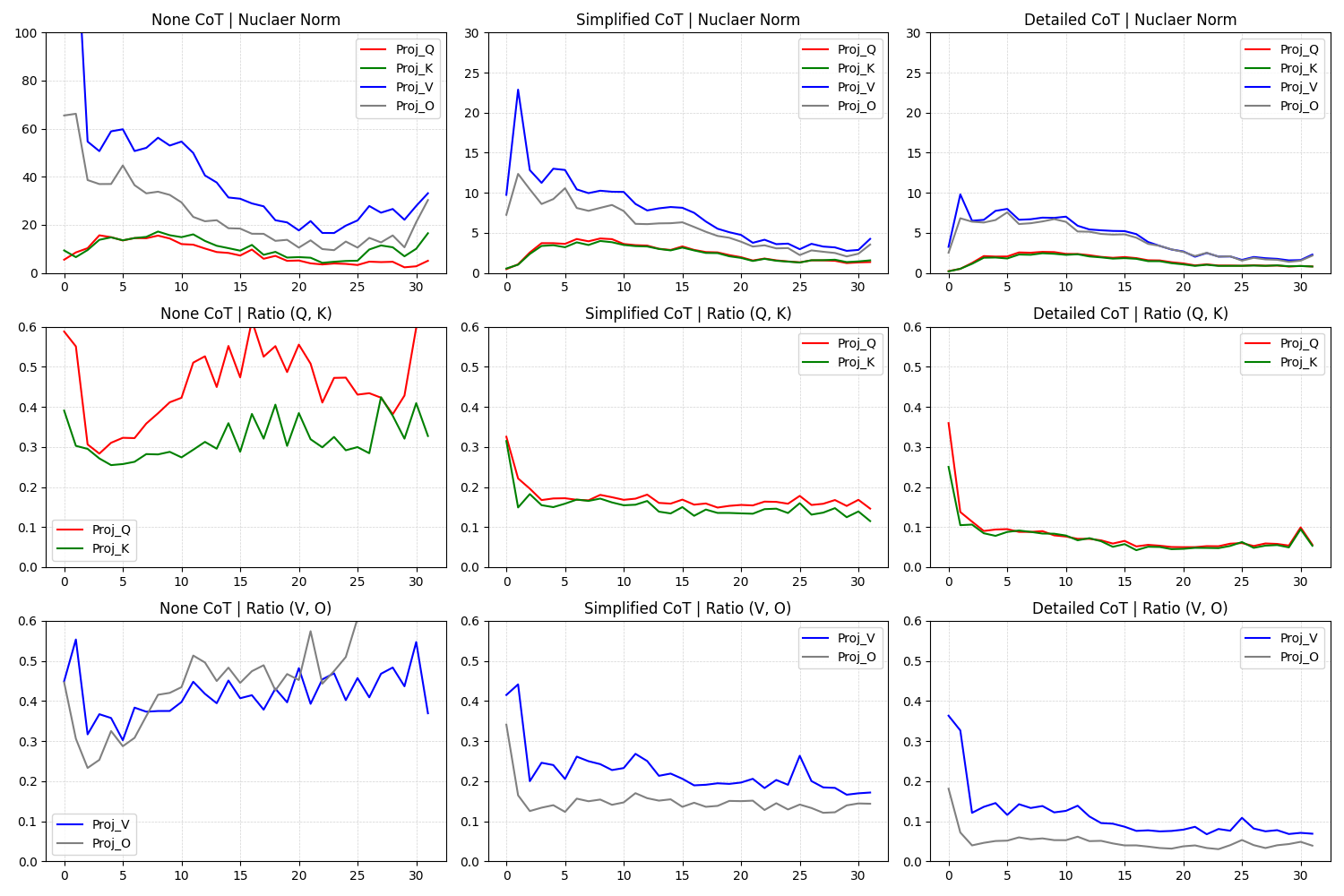}
\caption{Visualization for StrategyQA using Llama-2-7b-hf on irrelevant responses.}
\label{fig:reasoning_nonsense_strategyqa}
\vspace{-2mm}
\end{figure*}

\begin{table}[!tbh]
\centering
\small
\scalebox{0.75}{
\begin{tabular}{l|l|l|cccc}
\toprule
\multirow{2}{*}{\textbf{Dataset}} & \multirow{2}{*}{\textbf{Curve}} & \multirow{2}{*}{\textbf{Cot}} &  \multicolumn{4}{c}{\textbf{Mean Absolute Difference (MAD)}} \\
 & &  & Early & Middle & Last & All\\
\midrule
\multirow{24}{*}{ECQA} & \multirow{3}{*}{$s_Q$} 
  & None & 1.45 & 0.71 & 0.83 & 0.98 \\
& & Simplified & 0.64 & 0.29 & 0.20 & 0.37 \\
& & Detailed & 0.35 & 0.18 & 0.12 & 0.21 \\
\cmidrule{2-7}
& \multirow{3}{*}{$s_K$} 
  & None & 1.58 & 0.70 & 1.97 & 1.39 \\
& & Simplified & 0.57 & 0.26 & 0.26 & 0.34 \\
& & Detailed & 0.34 & 0.18 & 0.11 & 0.20 \\
\cmidrule{2-7}
& \multirow{3}{*}{$s_V$} 
  & None & 12.22 & 1.92 & 2.34 & 5.00 \\
& & Simplified & 3.70 & 0.85 & 0.63 & 1.56 \\
& & Detailed & 1.60 & 0.59 & 0.39 & 0.79 \\
\cmidrule{2-7}
& \multirow{3}{*}{$s_O$} 
  & None & 6.02 & 1.47 & 1.81 & 2.95 \\
& & Simplified & 1.80 & 0.59 & 0.54 & 0.91 \\
& & Detailed & 0.98 & 0.49 & 0.38 & 0.58 \\
\cmidrule{2-7}
 & \multirow{3}{*}{$r_Q$} 
  & None & 0.05 & 0.01 & 0.06 & 0.04 \\
& & Simplified & 0.03 & 0.01 & 0.01 & 0.01 \\
& & Detailed & 0.03 & 0.01 & 0.01 & 0.02 \\
\cmidrule{2-7}
& \multirow{3}{*}{$r_K$} 
  & None & 0.03 & 0.02 & 0.04 & 0.03 \\
& & Simplified & 0.03 & 0.01 & 0.01 & 0.02 \\
& & Detailed & 0.02 & 0.01 & 0.02 & 0.02 \\
\cmidrule{2-7}
 & \multirow{3}{*}{$r_V$} 
  & None & 0.06 & 0.01 & 0.03 & 0.03 \\
& & Simplified & 0.05 & 0.01 & 0.02 & 0.02 \\
& & Detailed & 0.04 & 0.01 & 0.01 & 0.02 \\
\cmidrule{2-7}
& \multirow{3}{*}{$r_O$} 
  & None & 0.04 & 0.02 & 0.04 & 0.04 \\
& & Simplified & 0.03 & 0.01 & 0.03 & 0.02 \\
& & Detailed & 0.02 & 0.01 & 0.01 & 0.01 \\
\bottomrule
\end{tabular}
}
\caption{
Statistical results for ECQA using Llama-2-7b-hf on irrelevant responses.
}
\label{tbl:reasoning_cot_500_wrong_answer_shuffle_llama2_7b_grads_ecqa_train}
\end{table}

\begin{figure*}[!h]
\centering
\includegraphics[width=\linewidth]{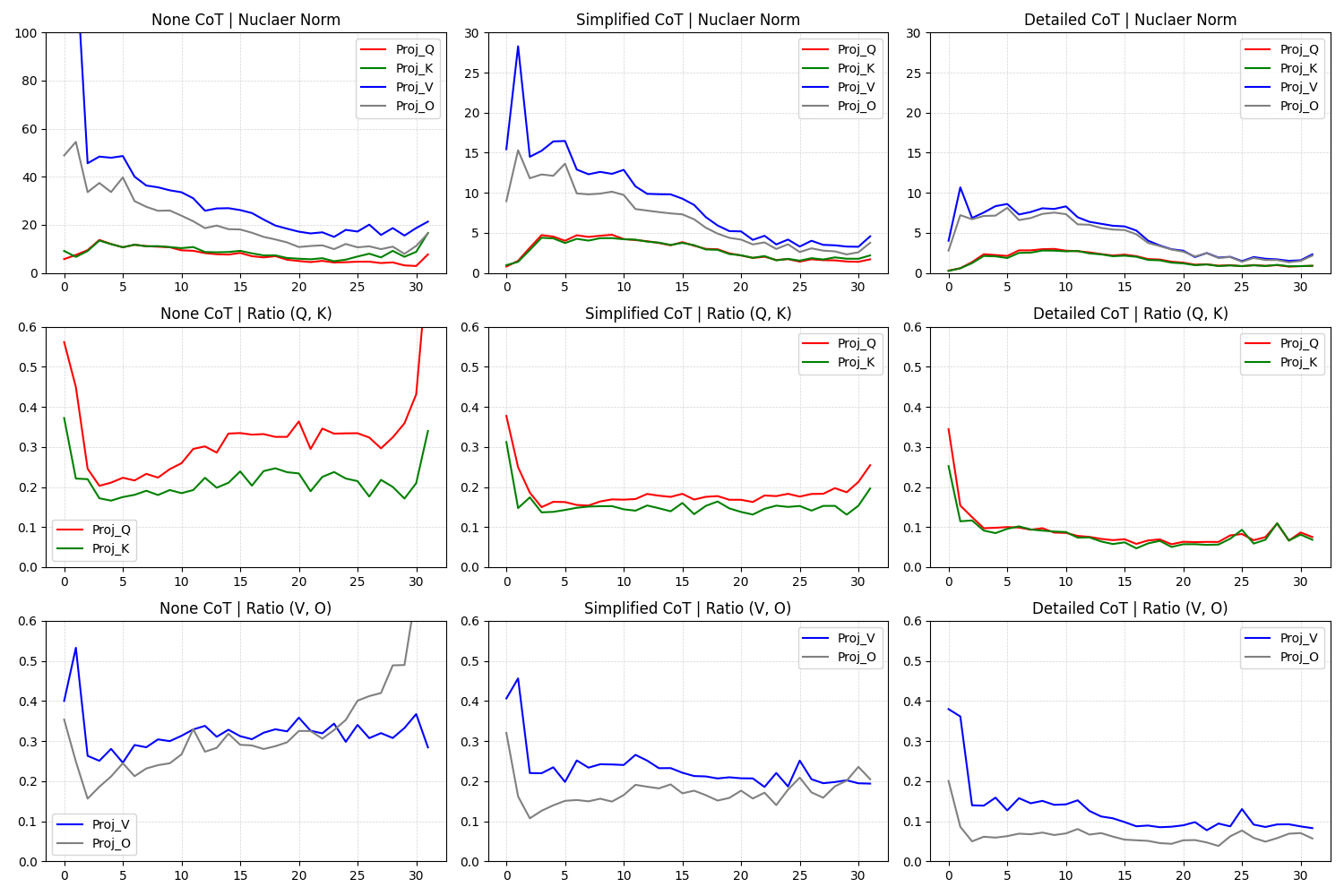}
\caption{Visualization for ECQA using Llama-2-7b-hf on irrelevant responses.}
\label{fig:reasoning_nonsense_ecqa}
\vspace{-2mm}
\end{figure*}

\begin{table}[!tbh]
\centering
\small
\scalebox{0.75}{
\begin{tabular}{l|l|l|cccc}
\toprule
\multirow{2}{*}{\textbf{Dataset}} & \multirow{2}{*}{\textbf{Curve}} & \multirow{2}{*}{\textbf{Cot}} &  \multicolumn{4}{c}{\textbf{Mean Absolute Difference (MAD)}} \\
 & &  & Early & Middle & Last & All\\
\midrule
\multirow{24}{*}{CREAK} & \multirow{3}{*}{$s_Q$} 
  & None & 1.54 & 1.12 & 1.87 & 1.54 \\
& & Simplified & 0.62 & 0.37 & 0.23 & 0.41 \\
& & Detailed & 0.34 & 0.15 & 0.11 & 0.20 \\
\cmidrule{2-7}
& \multirow{3}{*}{$s_K$} 
  & None & 2.14 & 1.39 & 3.62 & 2.42 \\
& & Simplified & 0.57 & 0.36 & 0.30 & 0.40 \\
& & Detailed & 0.34 & 0.16 & 0.11 & 0.19 \\
\cmidrule{2-7}
& \multirow{3}{*}{$s_V$} 
  & None & 18.53 & 3.19 & 3.71 & 7.75 \\
& & Simplified & 3.81 & 0.96 & 0.65 & 1.63 \\
& & Detailed & 1.53 & 0.50 & 0.33 & 0.72 \\
\cmidrule{2-7}
& \multirow{3}{*}{$s_O$} 
  & None & 6.50 & 2.43 & 3.36 & 4.02 \\
& & Simplified & 2.01 & 0.61 & 0.54 & 0.99 \\
& & Detailed & 0.91 & 0.43 & 0.32 & 0.53 \\
\cmidrule{2-7}
 & \multirow{3}{*}{$r_Q$} 
  & None & 0.05 & 0.03 & 0.12 & 0.07 \\
& & Simplified & 0.03 & 0.01 & 0.01 & 0.01 \\
& & Detailed & 0.03 & 0.01 & 0.01 & 0.01 \\
\cmidrule{2-7}
& \multirow{3}{*}{$r_K$} 
  & None & 0.03 & 0.02 & 0.08 & 0.04 \\
& & Simplified & 0.03 & 0.01 & 0.01 & 0.02 \\
& & Detailed & 0.02 & 0.01 & 0.01 & 0.02 \\
\cmidrule{2-7}
 & \multirow{3}{*}{$r_V$} 
  & None & 0.07 & 0.02 & 0.05 & 0.04 \\
& & Simplified & 0.05 & 0.01 & 0.02 & 0.03 \\
& & Detailed & 0.04 & 0.01 & 0.01 & 0.02 \\
\cmidrule{2-7}
& \multirow{3}{*}{$r_O$} 
  & None & 0.04 & 0.02 & 0.07 & 0.04 \\
& & Simplified & 0.03 & 0.01 & 0.02 & 0.02 \\
& & Detailed & 0.02 & 0.01 & 0.01 & 0.01 \\
\bottomrule
\end{tabular}
}
\caption{
Statistical results for CREAK using Llama-2-7b-hf on irrelevant responses.
}
\label{tbl:reasoning_cot_500_wrong_answer_shuffle_llama2_7b_grads_creak_train}
\end{table}

\begin{figure*}[!h]
\centering
\includegraphics[width=\linewidth]{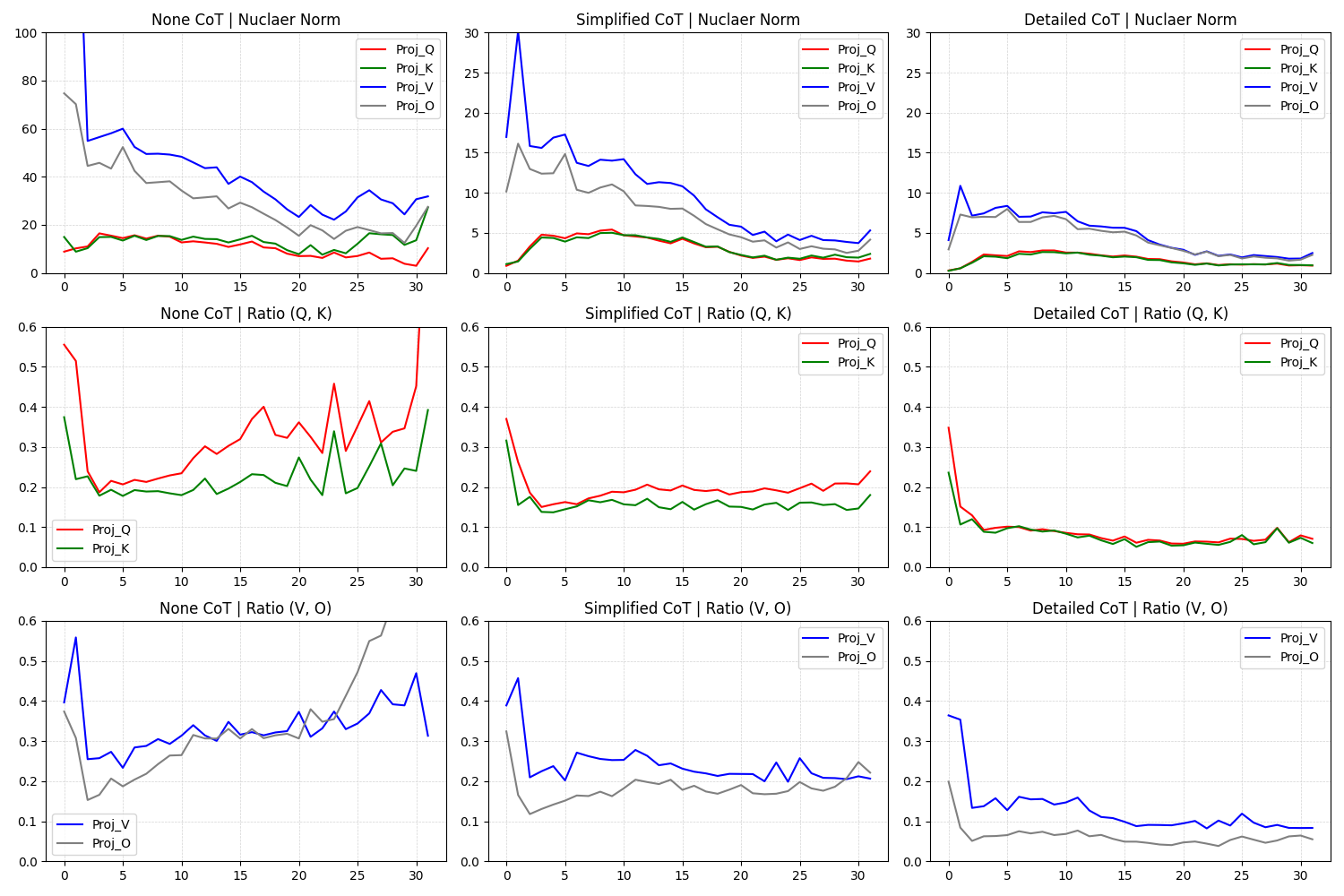}
\caption{Visualization for CREAK using Llama-2-7b-hf on irrelevant responses.}
\label{fig:reasoning_nonsense_creak}
\vspace{-2mm}
\end{figure*}

\begin{table}[!tbh]
\centering
\small
\scalebox{0.75}{
\begin{tabular}{l|l|l|cccc}
\toprule
\multirow{2}{*}{\textbf{Dataset}} & \multirow{2}{*}{\textbf{Curve}} & \multirow{2}{*}{\textbf{Cot}} &  \multicolumn{4}{c}{\textbf{Mean Absolute Difference (MAD)}} \\
 & &  & Early & Middle & Last & All\\
\midrule
\multirow{24}{*}{Sensemaking} & \multirow{3}{*}{$s_Q$} 
  & None & 0.44 & 0.46 & 0.56 & 0.50 \\
& & Simplified & 0.69 & 0.51 & 0.33 & 0.52 \\
& & Detailed & 0.32 & 0.17 & 0.10 & 0.19 \\
\cmidrule{2-7}
& \multirow{3}{*}{$s_K$} 
  & None & 0.56 & 0.45 & 0.97 & 0.67 \\
& & Simplified & 0.66 & 0.55 & 0.48 & 0.56 \\
& & Detailed & 0.34 & 0.20 & 0.09 & 0.20 \\
\cmidrule{2-7}
& \multirow{3}{*}{$s_V$} 
  & None & 5.22 & 0.95 & 0.50 & 2.02 \\
& & Simplified & 3.68 & 1.17 & 0.81 & 1.71 \\
& & Detailed & 1.59 & 0.57 & 0.32 & 0.76 \\
\cmidrule{2-7}
& \multirow{3}{*}{$s_O$} 
  & None & 1.94 & 0.72 & 0.60 & 1.05 \\
& & Simplified & 2.12 & 0.82 & 0.58 & 1.11 \\
& & Detailed & 0.99 & 0.50 & 0.32 & 0.57 \\
\cmidrule{2-7}
 & \multirow{3}{*}{$r_Q$} 
  & None & 0.05 & 0.03 & 0.12 & 0.07 \\
& & Simplified & 0.04 & 0.01 & 0.04 & 0.03 \\
& & Detailed & 0.03 & 0.01 & 0.02 & 0.02 \\
\cmidrule{2-7}
& \multirow{3}{*}{$r_K$} 
  & None & 0.04 & 0.02 & 0.07 & 0.04 \\
& & Simplified & 0.03 & 0.01 & 0.03 & 0.02 \\
& & Detailed & 0.02 & 0.01 & 0.02 & 0.02 \\
\cmidrule{2-7}
 & \multirow{3}{*}{$r_V$} 
  & None & 0.06 & 0.02 & 0.04 & 0.04 \\
& & Simplified & 0.05 & 0.01 & 0.03 & 0.03 \\
& & Detailed & 0.04 & 0.01 & 0.01 & 0.02 \\
\cmidrule{2-7}
& \multirow{3}{*}{$r_O$} 
  & None & 0.03 & 0.02 & 0.06 & 0.04 \\
& & Simplified & 0.03 & 0.01 & 0.02 & 0.02 \\
& & Detailed & 0.02 & 0.00 & 0.01 & 0.01 \\
\bottomrule
\end{tabular}
}
\caption{
Statistical results for Sensemaking using Llama-2-7b-hf on irrelevant responses.
}
\label{tbl:reasoning_cot_500_wrong_answer_shuffle_llama2_7b_grads_sensemaking_train}
\end{table}

\begin{figure*}[!h]
\centering
\includegraphics[width=\linewidth]{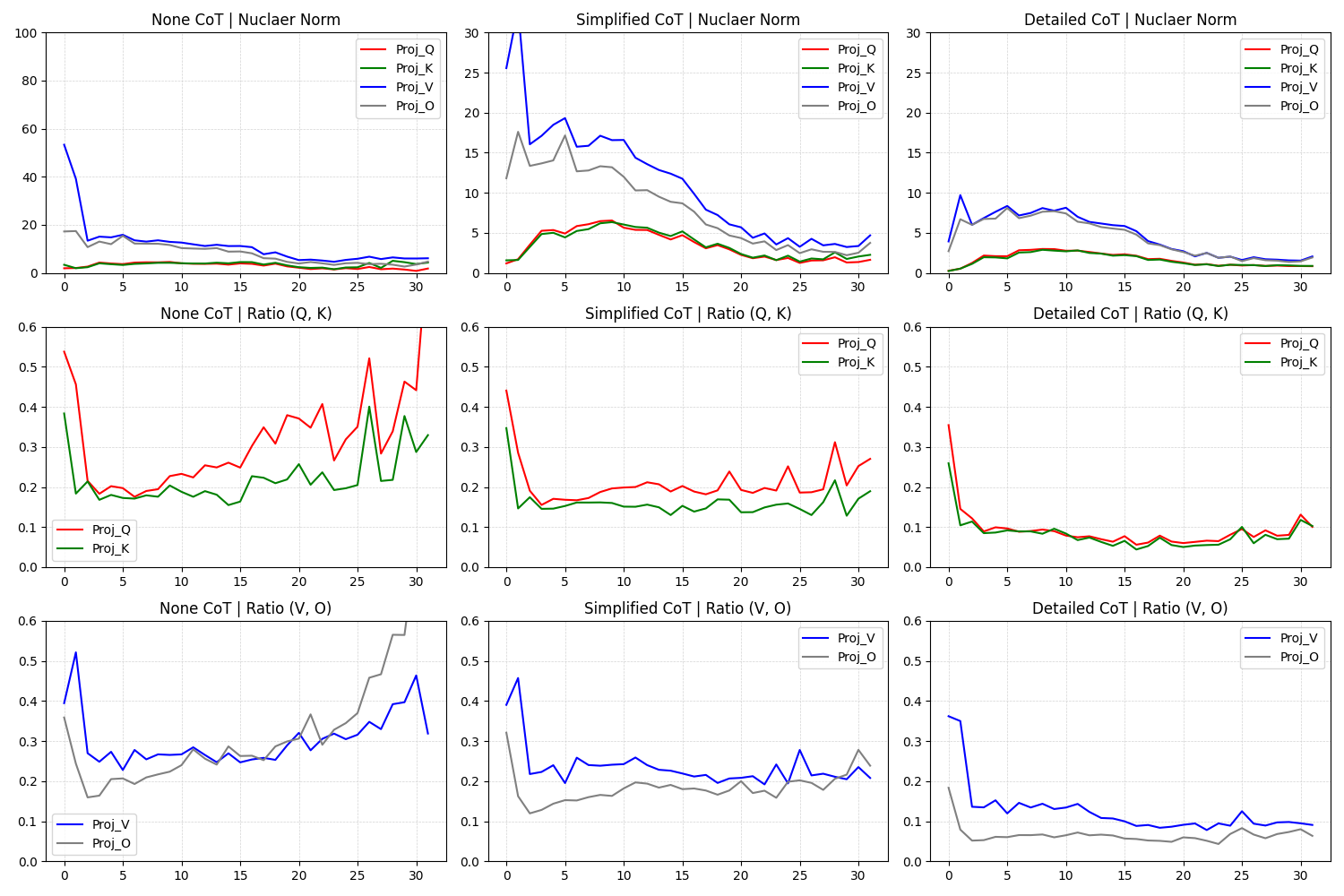}
\caption{Visualization for Sensemaking using Llama-2-7b-hf on irrelevant responses.}
\label{fig:reasoning_nonsense_sensemaking}
\vspace{-2mm}
\end{figure*}

\subsubsection{Wiki Tasks}

The visualizations and statistical results on Wiki tasks are shown in Figure \ref{fig:wiki_nonsense} and Table \ref{tbl:wiki_wiki_popularity_wrong_answer_shuffle_llama2_7b_grads}.

\begin{table}[!tbh]
\centering
\small
\scalebox{0.75}{
\begin{tabular}{l|l|l|cccc}
\toprule
\multirow{2}{*}{\textbf{Dataset}} & \multirow{2}{*}{\textbf{Curve}} & \multirow{2}{*}{\textbf{Cot}} &  \multicolumn{4}{c}{\textbf{Mean Absolute Difference (MAD)}} \\
 & &  & Early & Middle & Last & All\\
\midrule
\multirow{32}{*}{Wiki} & \multirow{3}{*}{$s_Q$} 
  & Len 100 & 0.51 & 0.22 & 0.23 & 0.31 \\
& & Len 500 & 0.31 & 0.14 & 0.18 & 0.21 \\
& & Len 1000 & 0.29 & 0.12 & 0.18 & 0.19 \\
& & Unpopular & 0.54 & 0.25 & 0.25 & 0.35 \\
\cmidrule{2-7}
& \multirow{3}{*}{$s_K$} 
  & Len 100 & 0.50 & 0.22 & 0.26 & 0.31 \\
& & Len 500 & 0.30 & 0.12 & 0.20 & 0.20 \\
& & Len 1000 & 0.28 & 0.11 & 0.20 & 0.19 \\
& & Unpopular & 0.53 & 0.26 & 0.27 & 0.35 \\
\cmidrule{2-7}
& \multirow{3}{*}{$s_V$} 
  & Len 100 & 3.20 & 0.64 & 0.37 & 1.26 \\
& & Len 500 & 2.01 & 0.47 & 0.37 & 0.86 \\
& & Len 1000 & 1.72 & 0.45 & 0.37 & 0.77 \\
& & Unpopular & 5.84 & 0.80 & 0.47 & 2.13 \\
\cmidrule{2-7}
& \multirow{3}{*}{$s_O$} 
  & Len 100 & 1.62 & 0.49 & 0.33 & 0.76 \\
& & Len 500 & 1.29 & 0.43 & 0.35 & 0.64 \\
& & Len 1000 & 1.17 & 0.42 & 0.35 & 0.60 \\
& & Unpopular & 2.86 & 0.67 & 0.39 & 1.22 \\
\cmidrule{2-7}
 & \multirow{3}{*}{$r_Q$} 
  & Len 100 & 0.02 & 0.01 & 0.01 & 0.01 \\
& & Len 500 & 0.03 & 0.01 & 0.01 & 0.02 \\
& & Len 1000 & 0.04 & 0.01 & 0.01 & 0.02 \\
& & Unpopular & 0.02 & 0.01 & 0.01 & 0.01 \\
\cmidrule{2-7}
& \multirow{3}{*}{$r_K$} 
  & Len 100 & 0.03 & 0.02 & 0.01 & 0.02 \\
& & Len 500 & 0.02 & 0.01 & 0.01 & 0.01 \\
& & Len 1000 & 0.02 & 0.01 & 0.01 & 0.01 \\
& & Unpopular & 0.03 & 0.02 & 0.01 & 0.02 \\
\cmidrule{2-7}
 & \multirow{3}{*}{$r_V$} 
  & Len 100 & 0.05 & 0.02 & 0.02 & 0.03 \\
& & Len 500 & 0.04 & 0.01 & 0.01 & 0.02 \\
& & Len 1000 & 0.04 & 0.01 & 0.01 & 0.02 \\
& & Unpopular & 0.05 & 0.02 & 0.01 & 0.03 \\
\cmidrule{2-7}
& \multirow{3}{*}{$r_O$} 
  & Len 100 & 0.03 & 0.01 & 0.01 & 0.02 \\
& & Len 500 & 0.02 & 0.01 & 0.01 & 0.01 \\
& & Len 1000 & 0.02 & 0.00 & 0.00 & 0.01 \\
& & Unpopular & 0.03 & 0.01 & 0.01 & 0.01 \\
\bottomrule
\end{tabular}
}
\caption{
Statistical results for Wiki using Llama-2-7b-hf on irrelevant responses.
}
\label{tbl:wiki_wiki_popularity_wrong_answer_shuffle_llama2_7b_grads}
\end{table}

\begin{figure*}[!h]
\centering
\includegraphics[width=\linewidth]{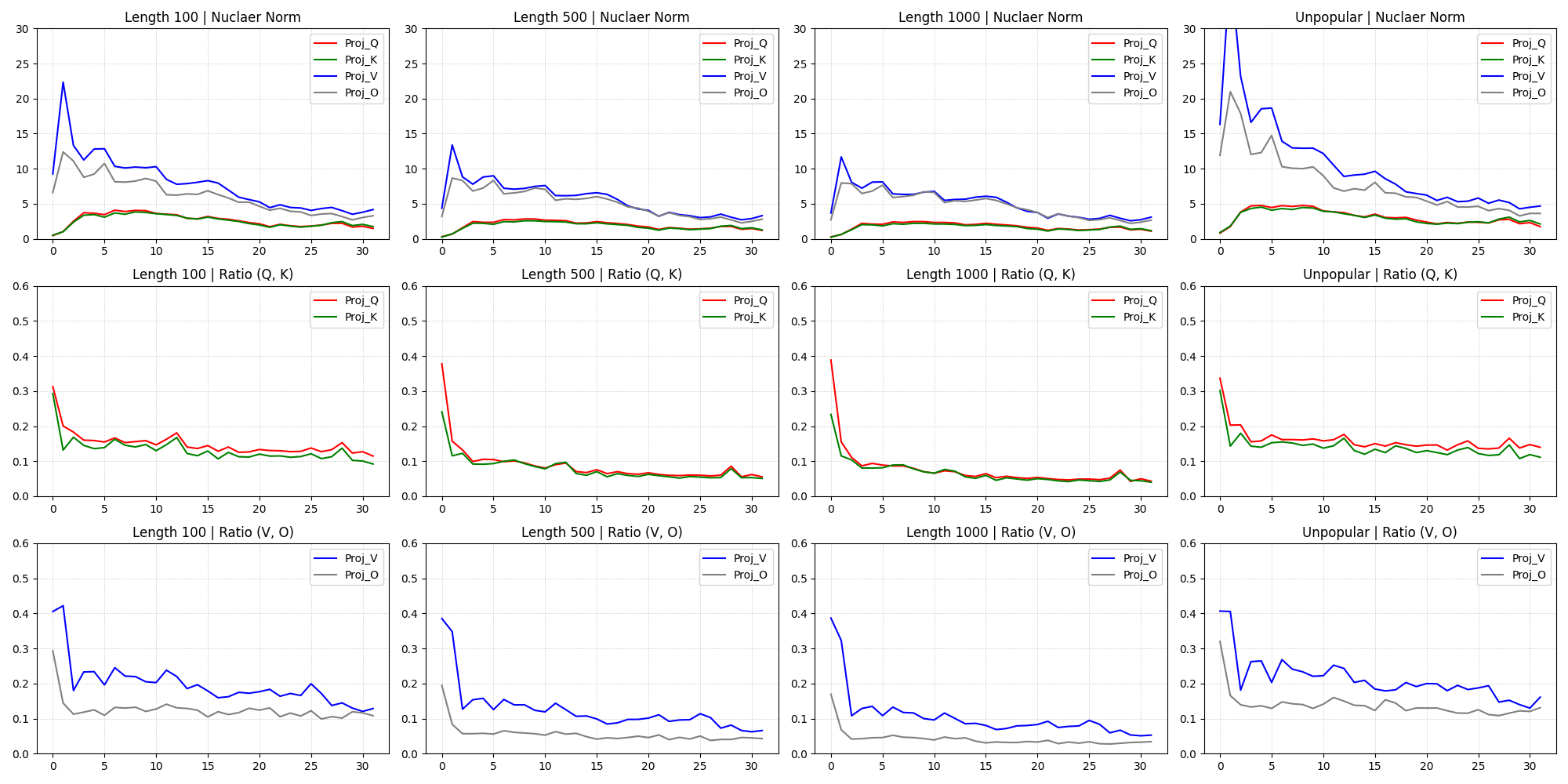}
\caption{Visualization for Wiki tasks using Llama-2-7b-hf on irrelevant responses.}
\label{fig:wiki_nonsense}
\vspace{-2mm}
\end{figure*}

\clearpage

\subsection{Instructed LLM on Correct Responses}

\subsubsection{Reasoning Tasks}

The visualizations and statistical results on MATH tasks: MATH-Algebra (Figure \ref{fig:math_correct_algebra_it}, Table \ref{tbl:math_lighteval_math_llama2_7b_it_new_grads_algebra}), MATH-Counting (Figure \ref{fig:math_correct_counting_it}, Table \ref{tbl:math_lighteval_math_llama2_7b_it_new_grads_counting_probability}), MATH-Geometry (Figure \ref{fig:math_correct_geometry_it}, Table \ref{tbl:math_lighteval_math_llama2_7b_it_new_grads_geometry}).

\begin{table}[!tbh]
\centering
\small
\scalebox{0.75}{
\begin{tabular}{l|l|l|cccc}
\toprule
\multirow{2}{*}{\textbf{Dataset}} & \multirow{2}{*}{\textbf{Curve}} & \multirow{2}{*}{\textbf{Cot}} &  \multicolumn{4}{c}{\textbf{Mean Absolute Difference (MAD)}} \\
 & &  & Early & Middle & Last & All\\
\midrule
\multirow{16}{*}{Algebra} & \multirow{3}{*}{$s_Q$} 
  & Simplified & 0.85 & 0.74 & 0.53 & 0.70 \\
& & Detailed & 0.45 & 0.38 & 0.34 & 0.38 \\
\cmidrule{2-7}
& \multirow{3}{*}{$s_K$} 
  & Simplified & 0.83 & 0.78 & 0.63 & 0.73 \\
& & Detailed & 0.43 & 0.40 & 0.38 & 0.40 \\
\cmidrule{2-7}
& \multirow{3}{*}{$s_V$} 
  & Simplified & 4.11 & 1.93 & 0.60 & 2.10 \\
& & Detailed & 1.98 & 1.04 & 0.33 & 1.07 \\
\cmidrule{2-7}
& \multirow{3}{*}{$s_O$} 
  & Simplified & 2.66 & 1.84 & 0.53 & 1.56 \\
& & Detailed & 1.41 & 1.08 & 0.30 & 0.87 \\
\cmidrule{2-7}
 & \multirow{3}{*}{$r_Q$} 
  & Simplified & 0.03 & 0.02 & 0.04 & 0.03 \\
& & Detailed & 0.04 & 0.02 & 0.04 & 0.03 \\
\cmidrule{2-7}
& \multirow{3}{*}{$r_K$} 
  & Simplified & 0.02 & 0.02 & 0.04 & 0.03 \\
& & Detailed & 0.02 & 0.02 & 0.05 & 0.03 \\
\cmidrule{2-7}
 & \multirow{3}{*}{$r_V$} 
  & Simplified & 0.05 & 0.01 & 0.03 & 0.03 \\
& & Detailed & 0.05 & 0.01 & 0.03 & 0.03 \\
\cmidrule{2-7}
& \multirow{3}{*}{$r_O$} 
  & Simplified & 0.02 & 0.01 & 0.02 & 0.01 \\
& & Detailed & 0.02 & 0.00 & 0.02 & 0.01 \\
\bottomrule
\end{tabular}
}
\caption{
Statistical results for MATH-Algebra using Llama-2-7b-chat-hf on correct responses.
}
\label{tbl:math_lighteval_math_llama2_7b_it_new_grads_algebra}
\end{table}

\begin{figure}[!h]
\centering
\includegraphics[width=\linewidth]{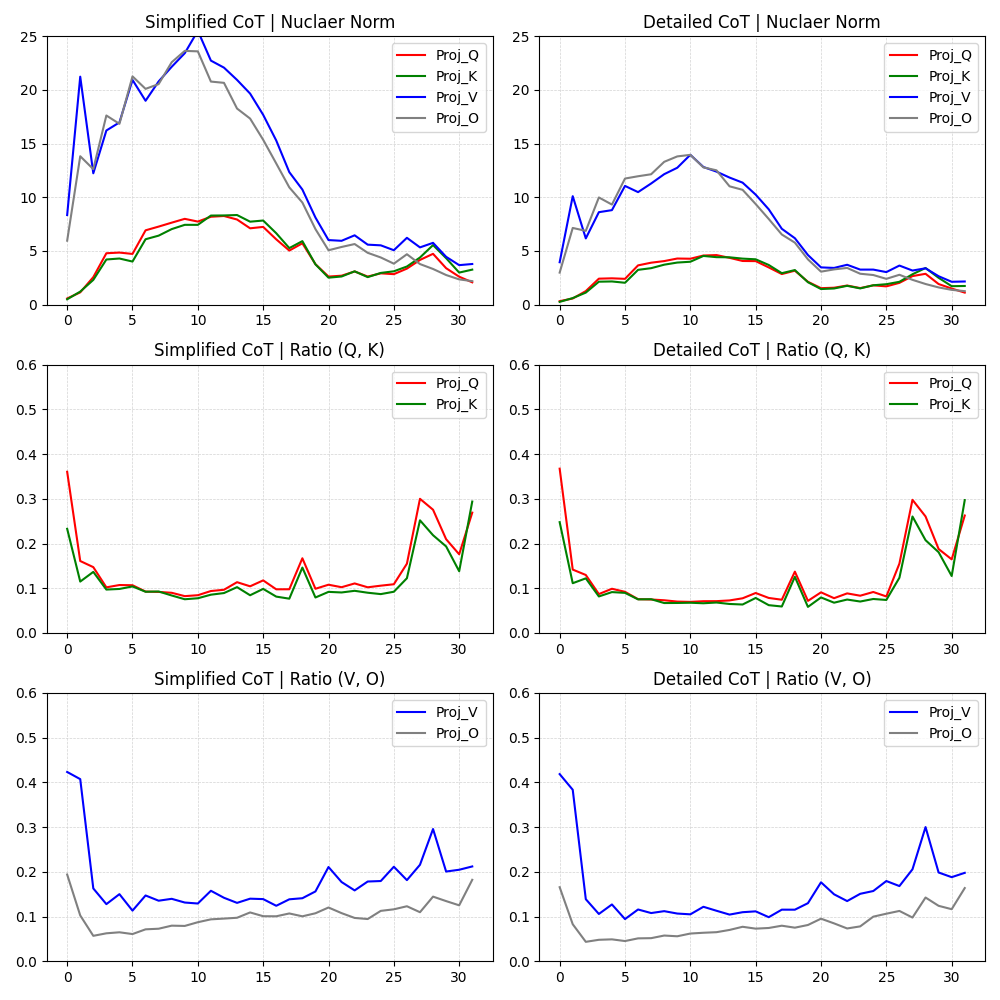}
\caption{Visualization for MATH-Algebra using Llama-2-7b-chat-hf on correct responses.}
\label{fig:math_correct_algebra_it}
\vspace{-2mm}
\end{figure}

\begin{table}[!tbh]
\centering
\small
\scalebox{0.75}{
\begin{tabular}{l|l|l|cccc}
\toprule
\multirow{2}{*}{\textbf{Dataset}} & \multirow{2}{*}{\textbf{Curve}} & \multirow{2}{*}{\textbf{Cot}} &  \multicolumn{4}{c}{\textbf{Mean Absolute Difference (MAD)}} \\
 & &  & Early & Middle & Last & All\\
\midrule
\multirow{16}{*}{Counting} & \multirow{3}{*}{$s_Q$} 
  & Simplified & 0.80 & 0.65 & 0.44 & 0.62 \\
& & Detailed & 0.47 & 0.39 & 0.36 & 0.40 \\
\cmidrule{2-7}
& \multirow{3}{*}{$s_K$} 
  & Simplified & 0.78 & 0.72 & 0.53 & 0.65 \\
& & Detailed & 0.45 & 0.40 & 0.43 & 0.42 \\
\cmidrule{2-7}
& \multirow{3}{*}{$s_V$} 
  & Simplified & 4.05 & 1.82 & 0.65 & 2.05 \\
& & Detailed & 2.17 & 1.03 & 0.44 & 1.16 \\
\cmidrule{2-7}
& \multirow{3}{*}{$s_O$} 
  & Simplified & 2.80 & 1.83 & 0.51 & 1.58 \\
& & Detailed & 1.62 & 1.13 & 0.33 & 0.95 \\
\cmidrule{2-7}
 & \multirow{3}{*}{$r_Q$} 
  & Simplified & 0.03 & 0.02 & 0.04 & 0.03 \\
& & Detailed & 0.04 & 0.02 & 0.04 & 0.03 \\
\cmidrule{2-7}
& \multirow{3}{*}{$r_K$} 
  & Simplified & 0.02 & 0.02 & 0.04 & 0.03 \\
& & Detailed & 0.02 & 0.02 & 0.05 & 0.03 \\
\cmidrule{2-7}
 & \multirow{3}{*}{$r_V$} 
  & Simplified & 0.05 & 0.01 & 0.03 & 0.03 \\
& & Detailed & 0.05 & 0.01 & 0.03 & 0.03 \\
\cmidrule{2-7}
& \multirow{3}{*}{$r_O$} 
  & Simplified & 0.02 & 0.01 & 0.01 & 0.01 \\
& & Detailed & 0.02 & 0.00 & 0.01 & 0.01 \\
\bottomrule
\end{tabular}
}
\caption{
Statistical results for MATH-Counting using Llama-2-7b-chat-hf on correct responses.
}
\label{tbl:math_lighteval_math_llama2_7b_it_new_grads_counting_probability}
\end{table}

\begin{figure}[!h]
\centering
\includegraphics[width=\linewidth]{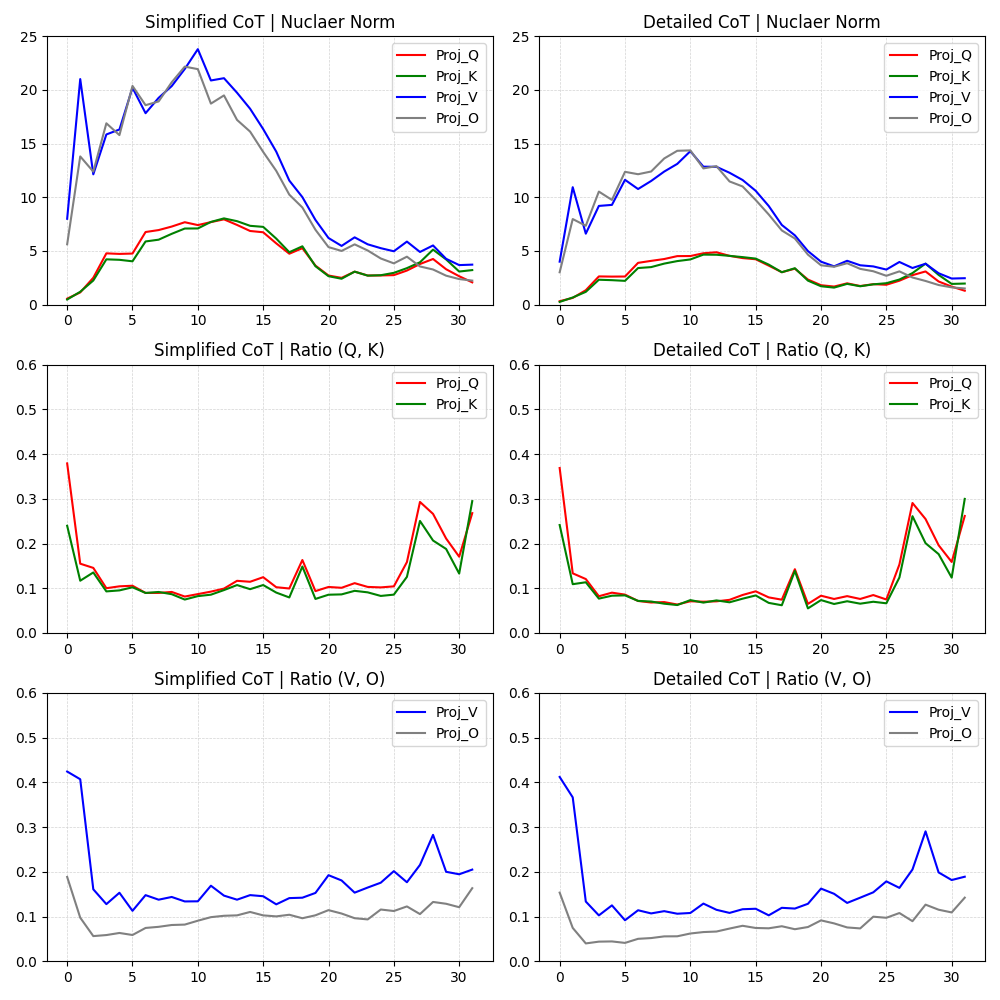}
\caption{Visualization for MATH-Counting using Llama-2-7b-chat-hf on correct responses.}
\label{fig:math_correct_counting_it}
\vspace{-2mm}
\end{figure}

\begin{table}[!tbh]
\centering
\small
\scalebox{0.75}{
\begin{tabular}{l|l|l|cccc}
\toprule
\multirow{2}{*}{\textbf{Dataset}} & \multirow{2}{*}{\textbf{Curve}} & \multirow{2}{*}{\textbf{Cot}} &  \multicolumn{4}{c}{\textbf{Mean Absolute Difference (MAD)}} \\
 & &  & Early & Middle & Last & All\\
\midrule
\multirow{16}{*}{Geometry} & \multirow{3}{*}{$s_Q$} 
  & Simplified & 0.71 & 0.57 & 0.52 & 0.60 \\
& & Detailed & 0.48 & 0.38 & 0.39 & 0.41 \\
\cmidrule{2-7}
& \multirow{3}{*}{$s_K$} 
  & Simplified & 0.69 & 0.65 & 0.62 & 0.65 \\
& & Detailed & 0.47 & 0.43 & 0.43 & 0.44 \\
\cmidrule{2-7}
& \multirow{3}{*}{$s_V$} 
  & Simplified & 3.51 & 1.58 & 0.59 & 1.78 \\
& & Detailed & 2.23 & 1.08 & 0.43 & 1.18 \\
\cmidrule{2-7}
& \multirow{3}{*}{$s_O$} 
  & Simplified & 2.39 & 1.58 & 0.51 & 1.39 \\
& & Detailed & 1.62 & 1.11 & 0.35 & 0.95 \\
\cmidrule{2-7}
 & \multirow{3}{*}{$r_Q$} 
  & Simplified & 0.04 & 0.02 & 0.04 & 0.03 \\
& & Detailed & 0.04 & 0.02 & 0.05 & 0.03 \\
\cmidrule{2-7}
& \multirow{3}{*}{$r_K$} 
  & Simplified & 0.02 & 0.02 & 0.04 & 0.03 \\
& & Detailed & 0.02 & 0.02 & 0.05 & 0.03 \\
\cmidrule{2-7}
 & \multirow{3}{*}{$r_V$} 
  & Simplified & 0.04 & 0.01 & 0.03 & 0.03 \\
& & Detailed & 0.04 & 0.01 & 0.03 & 0.03 \\
\cmidrule{2-7}
& \multirow{3}{*}{$r_O$} 
  & Simplified & 0.02 & 0.00 & 0.01 & 0.01 \\
& & Detailed & 0.02 & 0.00 & 0.02 & 0.01 \\
\bottomrule
\end{tabular}
}
\caption{
Statistical results for MATH-Geometry using Llama-2-7b-chat-hf on correct responses.
}
\label{tbl:math_lighteval_math_llama2_7b_it_new_grads_geometry}
\end{table}

\begin{figure}[!h]
\centering
\includegraphics[width=\linewidth]{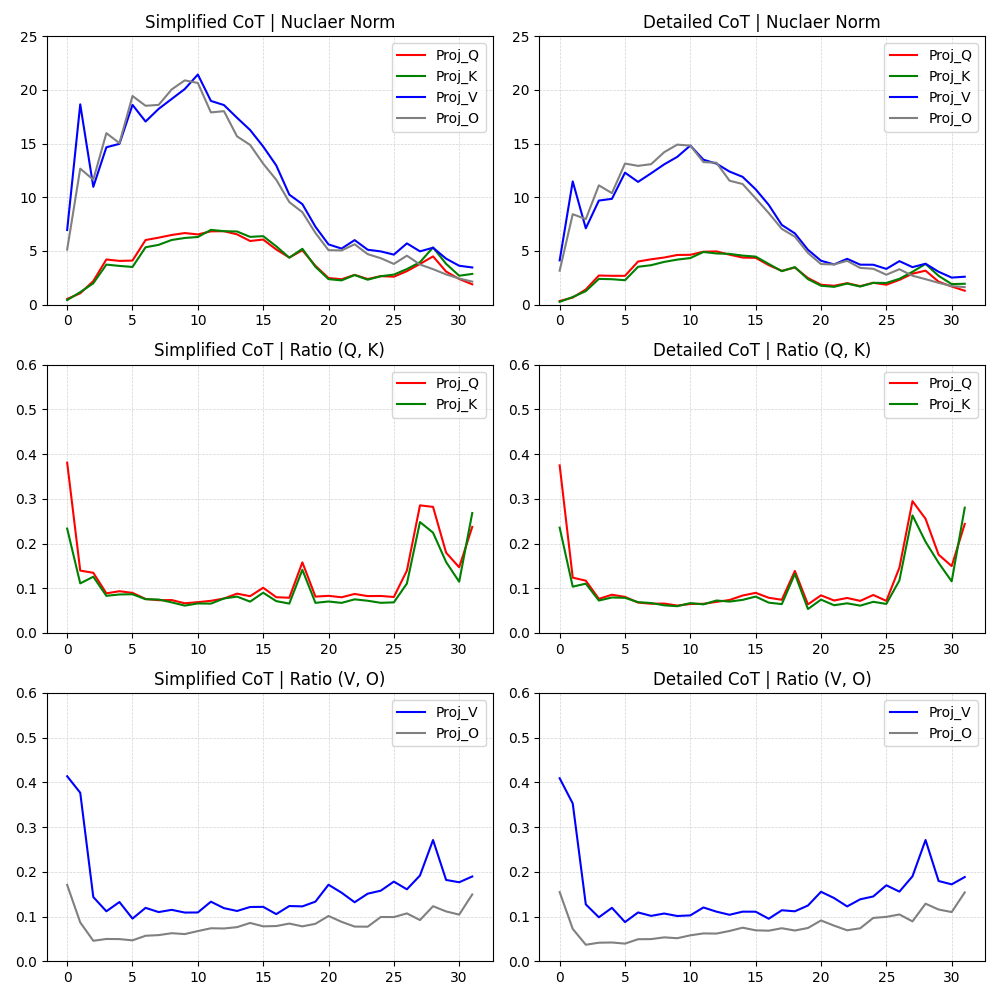}
\caption{Visualization for MATH-Geometry using Llama-2-7b-chat-hf on correct responses.}
\label{fig:math_correct_geometry_it}
\vspace{-2mm}
\end{figure}

The visualizations and statistical results on other reasoning tasks: AQuA (Figure \ref{fig:reasoning_correct_aqua_it}, Table \ref{tbl:reasoning_cot_500_llama2_7b_it_new_grads_aqua_train}), GSM8K (Figure \ref{fig:reasoning_correct_gsm8k_it}, Table \ref{tbl:reasoning_cot_500_llama2_7b_it_new_grads_gsm8k_train}), StrategyQA (Figure \ref{fig:reasoning_correct_strategyqa_it}, Table \ref{tbl:reasoning_cot_500_llama2_7b_it_new_grads_strategyqa_train}), ECQA (Figure \ref{fig:reasoning_correct_ecqa_it}, Table \ref{tbl:reasoning_cot_500_llama2_7b_it_new_grads_ecqa_train}), CREAK (Figure \ref{fig:reasoning_correct_creak_it}, Table \ref{tbl:reasoning_cot_500_llama2_7b_it_new_grads_creak_train}), Sensemaking (Figure \ref{fig:reasoning_correct_sensemaking_it}, Table \ref{tbl:reasoning_cot_500_llama2_7b_it_new_grads_sensemaking_train}).

\begin{table}[!tbh]
\centering
\small
\scalebox{0.75}{
\begin{tabular}{l|l|l|cccc}
\toprule
\multirow{2}{*}{\textbf{Dataset}} & \multirow{2}{*}{\textbf{Curve}} & \multirow{2}{*}{\textbf{Cot}} &  \multicolumn{4}{c}{\textbf{Mean Absolute Difference (MAD)}} \\
 & &  & Early & Middle & Last & All\\
\midrule
\multirow{24}{*}{AQuA} & \multirow{3}{*}{$s_Q$} 
  & None & 9.84 & 5.39 & 4.01 & 6.13 \\
& & Simplified & 1.38 & 1.00 & 0.63 & 0.98 \\
& & Detailed & 0.47 & 0.39 & 0.40 & 0.41 \\
\cmidrule{2-7}
& \multirow{3}{*}{$s_K$} 
  & None & 11.77 & 7.22 & 7.84 & 8.70 \\
& & Simplified & 1.37 & 1.17 & 0.77 & 1.07 \\
& & Detailed & 0.44 & 0.38 & 0.43 & 0.41 \\
\cmidrule{2-7}
& \multirow{3}{*}{$s_V$} 
  & None & 82.67 & 13.02 & 6.96 & 31.47 \\
& & Simplified & 7.31 & 2.77 & 0.98 & 3.42 \\
& & Detailed & 2.17 & 1.00 & 0.45 & 1.15 \\
\cmidrule{2-7}
& \multirow{3}{*}{$s_O$} 
  & None & 32.51 & 8.05 & 4.15 & 14.13 \\
& & Simplified & 4.50 & 2.62 & 0.72 & 2.42 \\
& & Detailed & 1.63 & 1.10 & 0.35 & 0.95 \\
\cmidrule{2-7}
 & \multirow{3}{*}{$r_Q$} 
  & None & 0.08 & 0.05 & 0.11 & 0.08 \\
& & Simplified & 0.03 & 0.02 & 0.04 & 0.03 \\
& & Detailed & 0.04 & 0.02 & 0.04 & 0.03 \\
\cmidrule{2-7}
& \multirow{3}{*}{$r_K$} 
  & None & 0.05 & 0.01 & 0.03 & 0.03 \\
& & Simplified & 0.03 & 0.02 & 0.04 & 0.03 \\
& & Detailed & 0.02 & 0.02 & 0.05 & 0.03 \\
\cmidrule{2-7}
 & \multirow{3}{*}{$r_V$} 
  & None & 0.09 & 0.03 & 0.04 & 0.05 \\
& & Simplified & 0.06 & 0.01 & 0.03 & 0.03 \\
& & Detailed & 0.05 & 0.01 & 0.03 & 0.03 \\
\cmidrule{2-7}
& \multirow{3}{*}{$r_O$} 
  & None & 0.05 & 0.03 & 0.06 & 0.05 \\
& & Simplified & 0.02 & 0.01 & 0.02 & 0.02 \\
& & Detailed & 0.02 & 0.01 & 0.01 & 0.01 \\
\bottomrule
\end{tabular}
}
\caption{
Statistical results for AQuA using Llama-2-7b-chat-hf on correct responses.
}
\label{tbl:reasoning_cot_500_llama2_7b_it_new_grads_aqua_train}
\end{table}

\begin{figure*}[!h]
\centering
\includegraphics[width=\linewidth]{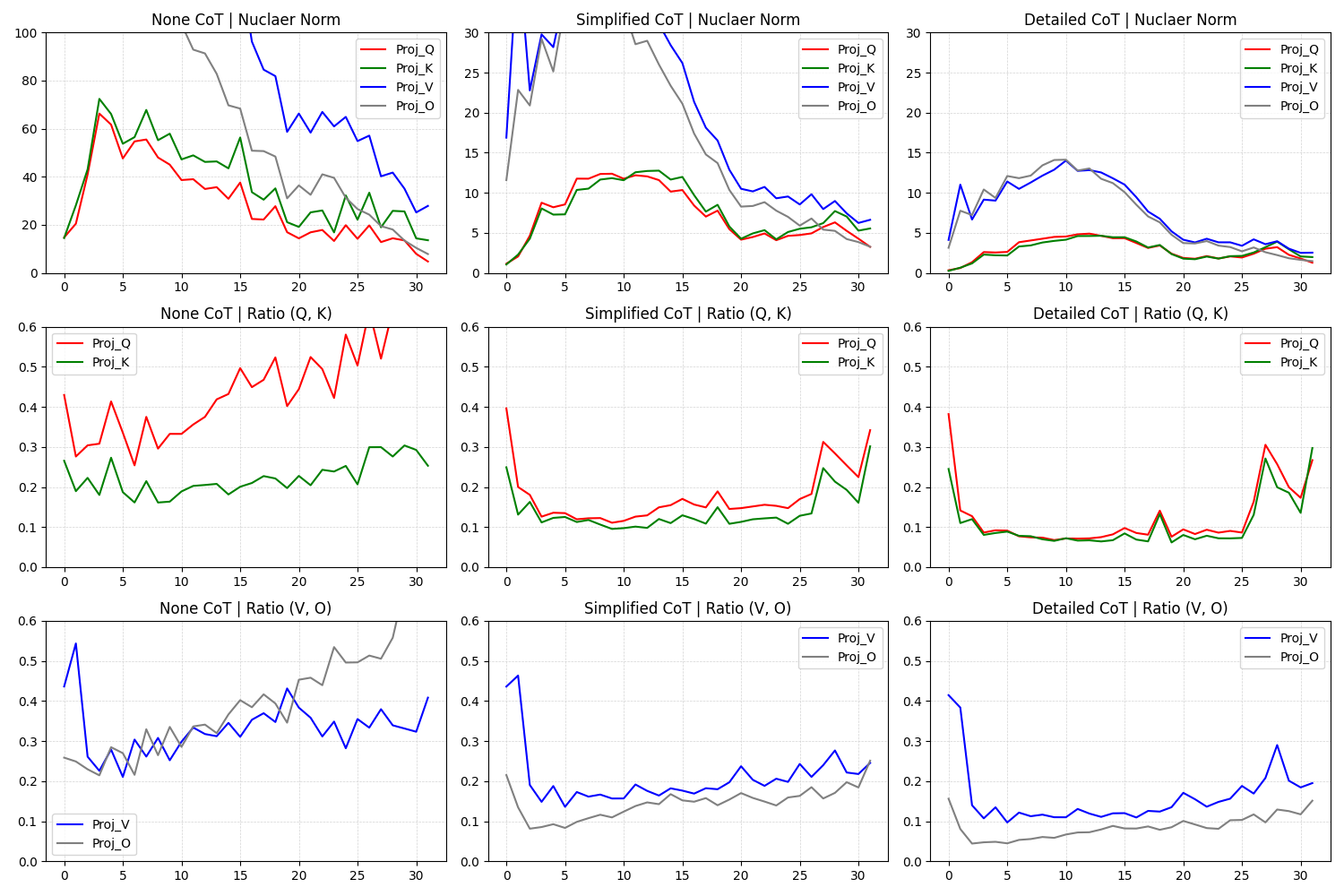}
\caption{Visualization for AQuA using Llama-2-7b-chat-hf on correct responses.}
\label{fig:reasoning_correct_aqua_it}
\vspace{-2mm}
\end{figure*}

\begin{table}[!tbh]
\centering
\small
\scalebox{0.75}{
\begin{tabular}{l|l|l|cccc}
\toprule
\multirow{2}{*}{\textbf{Dataset}} & \multirow{2}{*}{\textbf{Curve}} & \multirow{2}{*}{\textbf{Cot}} &  \multicolumn{4}{c}{\textbf{Mean Absolute Difference (MAD)}} \\
 & &  & Early & Middle & Last & All\\
\midrule
\multirow{24}{*}{GSM8K} & \multirow{3}{*}{$s_Q$} 
  & None & 4.79 & 3.74 & 2.69 & 3.51 \\
& & Simplified & 0.88 & 0.71 & 0.42 & 0.65 \\
& & Detailed & 0.45 & 0.38 & 0.39 & 0.40 \\
\cmidrule{2-7}
& \multirow{3}{*}{$s_K$} 
  & None & 5.31 & 5.10 & 3.77 & 4.42 \\
& & Simplified & 0.86 & 0.85 & 0.53 & 0.72 \\
& & Detailed & 0.41 & 0.36 & 0.43 & 0.40 \\
\cmidrule{2-7}
& \multirow{3}{*}{$s_V$} 
  & None & 31.34 & 11.51 & 7.86 & 15.28 \\
& & Simplified & 4.46 & 1.95 & 0.67 & 2.20 \\
& & Detailed & 2.18 & 0.94 & 0.48 & 1.14 \\
\cmidrule{2-7}
& \multirow{3}{*}{$s_O$} 
  & None & 19.29 & 9.66 & 3.91 & 10.13 \\
& & Simplified & 2.95 & 1.79 & 0.39 & 1.58 \\
& & Detailed & 1.65 & 1.03 & 0.32 & 0.93 \\
\cmidrule{2-7}
 & \multirow{3}{*}{$r_Q$} 
  & None & 0.03 & 0.04 & 0.07 & 0.04 \\
& & Simplified & 0.03 & 0.02 & 0.04 & 0.03 \\
& & Detailed & 0.04 & 0.02 & 0.04 & 0.03 \\
\cmidrule{2-7}
& \multirow{3}{*}{$r_K$} 
  & None & 0.03 & 0.03 & 0.06 & 0.04 \\
& & Simplified & 0.03 & 0.02 & 0.04 & 0.03 \\
& & Detailed & 0.02 & 0.02 & 0.05 & 0.03 \\
\cmidrule{2-7}
 & \multirow{3}{*}{$r_V$} 
  & None & 0.07 & 0.03 & 0.05 & 0.04 \\
& & Simplified & 0.06 & 0.01 & 0.03 & 0.03 \\
& & Detailed & 0.05 & 0.01 & 0.03 & 0.03 \\
\cmidrule{2-7}
& \multirow{3}{*}{$r_O$} 
  & None & 0.03 & 0.03 & 0.08 & 0.05 \\
& & Simplified & 0.02 & 0.01 & 0.02 & 0.02 \\
& & Detailed & 0.02 & 0.01 & 0.01 & 0.01 \\
\bottomrule
\end{tabular}
}
\caption{
Statistical results for GSM8K using Llama-2-7b-chat-hf on correct responses.
}
\label{tbl:reasoning_cot_500_llama2_7b_it_new_grads_gsm8k_train}
\end{table}

\begin{figure*}[!h]
\centering
\includegraphics[width=\linewidth]{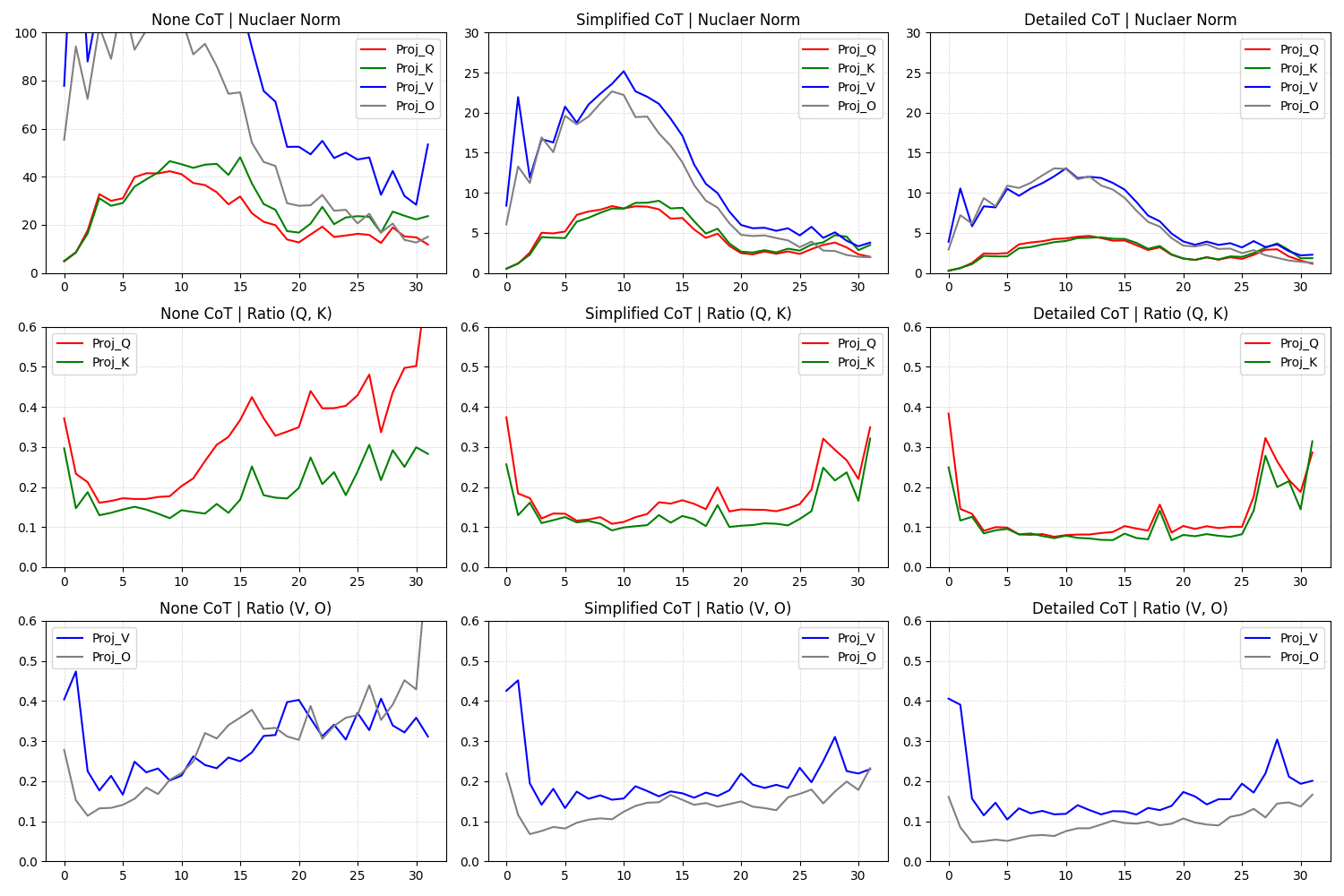}
\caption{Visualization for GSM8K using Llama-2-7b-chat-hf on correct responses.}
\label{fig:reasoning_correct_gsm8k_it}
\vspace{-2mm}
\end{figure*}

\begin{table}[!tbh]
\centering
\small
\scalebox{0.75}{
\begin{tabular}{l|l|l|cccc}
\toprule
\multirow{2}{*}{\textbf{Dataset}} & \multirow{2}{*}{\textbf{Curve}} & \multirow{2}{*}{\textbf{Cot}} &  \multicolumn{4}{c}{\textbf{Mean Absolute Difference (MAD)}} \\
 & &  & Early & Middle & Last & All\\
\midrule
\multirow{24}{*}{StrategyQA} & \multirow{3}{*}{$s_Q$} 
  & None & 6.24 & 5.38 & 6.40 & 5.71 \\
& & Simplified & 0.91 & 0.56 & 0.68 & 0.70 \\
& & Detailed & 0.43 & 0.29 & 0.41 & 0.37 \\
\cmidrule{2-7}
& \multirow{3}{*}{$s_K$} 
  & None & 6.17 & 7.25 & 8.80 & 7.08 \\
& & Simplified & 0.87 & 0.71 & 0.88 & 0.79 \\
& & Detailed & 0.40 & 0.30 & 0.53 & 0.41 \\
\cmidrule{2-7}
& \multirow{3}{*}{$s_V$} 
  & None & 47.27 & 17.66 & 17.60 & 25.66 \\
& & Simplified & 5.43 & 1.60 & 1.03 & 2.49 \\
& & Detailed & 2.10 & 0.79 & 0.56 & 1.10 \\
\cmidrule{2-7}
& \multirow{3}{*}{$s_O$} 
  & None & 28.77 & 14.81 & 9.00 & 16.46 \\
& & Simplified & 3.46 & 1.38 & 0.57 & 1.67 \\
& & Detailed & 1.58 & 0.90 & 0.40 & 0.89 \\
\cmidrule{2-7}
 & \multirow{3}{*}{$r_Q$} 
  & None & 0.03 & 0.05 & 0.09 & 0.06 \\
& & Simplified & 0.03 & 0.02 & 0.02 & 0.02 \\
& & Detailed & 0.04 & 0.02 & 0.04 & 0.03 \\
\cmidrule{2-7}
& \multirow{3}{*}{$r_K$} 
  & None & 0.03 & 0.01 & 0.04 & 0.03 \\
& & Simplified & 0.03 & 0.02 & 0.03 & 0.03 \\
& & Detailed & 0.02 & 0.02 & 0.05 & 0.03 \\
\cmidrule{2-7}
 & \multirow{3}{*}{$r_V$} 
  & None & 0.07 & 0.04 & 0.08 & 0.06 \\
& & Simplified & 0.05 & 0.02 & 0.04 & 0.03 \\
& & Detailed & 0.04 & 0.01 & 0.04 & 0.03 \\
\cmidrule{2-7}
& \multirow{3}{*}{$r_O$} 
  & None & 0.04 & 0.05 & 0.10 & 0.06 \\
& & Simplified & 0.03 & 0.01 & 0.01 & 0.02 \\
& & Detailed & 0.02 & 0.00 & 0.01 & 0.01 \\
\bottomrule
\end{tabular}
}
\caption{
Statistical results for StrategyQA using Llama-2-7b-chat-hf on correct responses.
}
\label{tbl:reasoning_cot_500_llama2_7b_it_new_grads_strategyqa_train}
\end{table}

\begin{figure*}[!h]
\centering
\includegraphics[width=\linewidth]{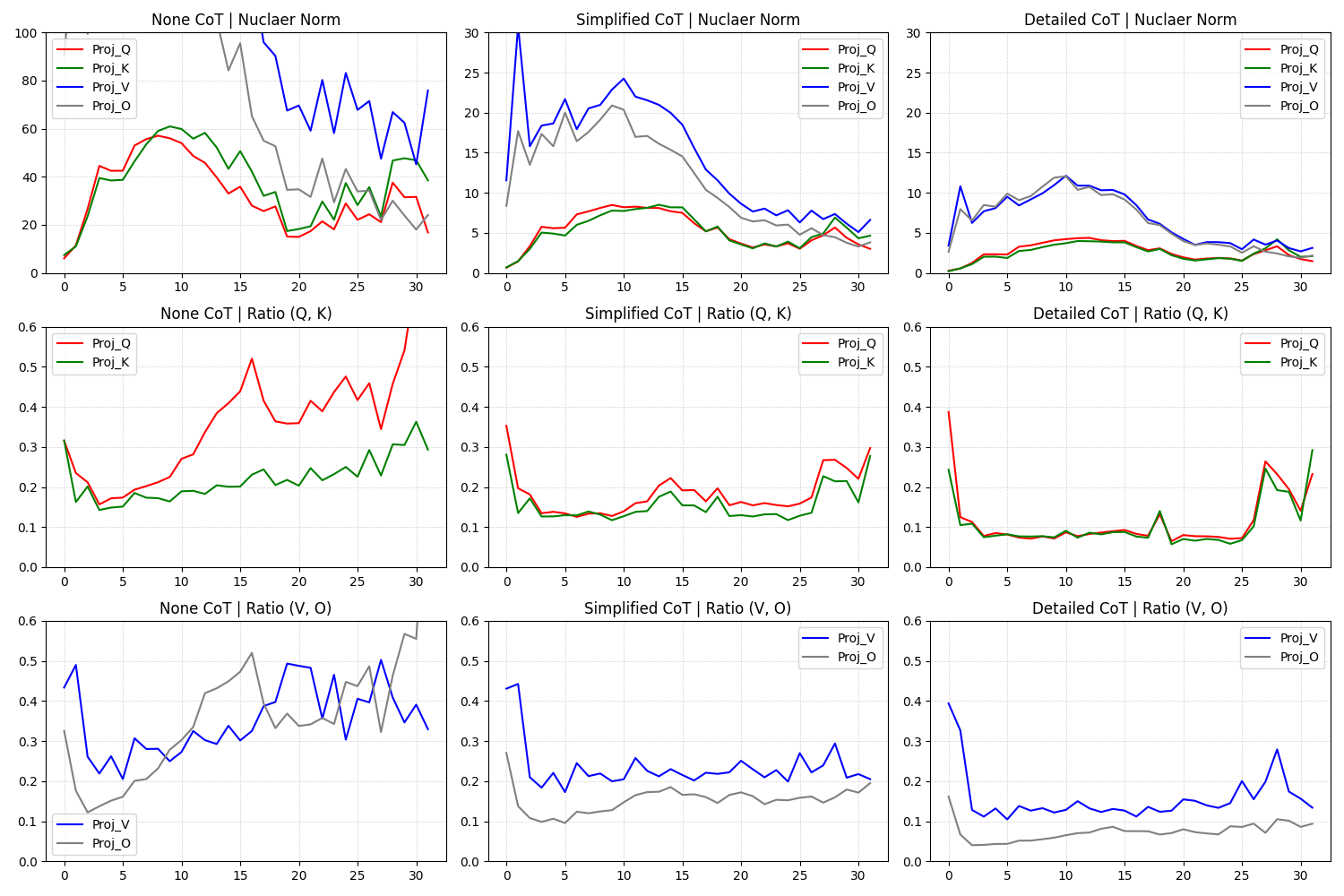}
\caption{Visualization for StrategyQA using Llama-2-7b-chat-hf on correct responses.}
\label{fig:reasoning_correct_strategyqa_it}
\vspace{-2mm}
\end{figure*}

\begin{table}[!tbh]
\centering
\small
\scalebox{0.75}{
\begin{tabular}{l|l|l|cccc}
\toprule
\multirow{2}{*}{\textbf{Dataset}} & \multirow{2}{*}{\textbf{Curve}} & \multirow{2}{*}{\textbf{Cot}} &  \multicolumn{4}{c}{\textbf{Mean Absolute Difference (MAD)}} \\
 & &  & Early & Middle & Last & All\\
\midrule
\multirow{24}{*}{ECQA} & \multirow{3}{*}{$s_Q$} 
  & None & 3.62 & 2.46 & 3.58 & 3.18 \\
& & Simplified & 1.41 & 0.86 & 0.83 & 1.01 \\
& & Detailed & 0.58 & 0.43 & 0.50 & 0.49 \\
\cmidrule{2-7}
& \multirow{3}{*}{$s_K$} 
  & None & 3.91 & 3.97 & 4.79 & 4.15 \\
& & Simplified & 1.42 & 1.04 & 1.24 & 1.19 \\
& & Detailed & 0.54 & 0.41 & 0.69 & 0.55 \\
\cmidrule{2-7}
& \multirow{3}{*}{$s_V$} 
  & None & 27.04 & 8.65 & 8.79 & 13.76 \\
& & Simplified & 8.06 & 2.94 & 1.71 & 3.93 \\
& & Detailed & 2.59 & 1.11 & 0.76 & 1.43 \\
\cmidrule{2-7}
& \multirow{3}{*}{$s_O$} 
  & None & 15.08 & 6.91 & 4.56 & 8.33 \\
& & Simplified & 5.14 & 2.53 & 1.03 & 2.67 \\
& & Detailed & 1.92 & 1.21 & 0.49 & 1.13 \\
\cmidrule{2-7}
 & \multirow{3}{*}{$r_Q$} 
  & None & 0.04 & 0.04 & 0.12 & 0.06 \\
& & Simplified & 0.03 & 0.02 & 0.03 & 0.03 \\
& & Detailed & 0.04 & 0.02 & 0.05 & 0.03 \\
\cmidrule{2-7}
& \multirow{3}{*}{$r_K$} 
  & None & 0.03 & 0.01 & 0.08 & 0.04 \\
& & Simplified & 0.03 & 0.02 & 0.03 & 0.02 \\
& & Detailed & 0.02 & 0.03 & 0.06 & 0.04 \\
\cmidrule{2-7}
 & \multirow{3}{*}{$r_V$} 
  & None & 0.07 & 0.03 & 0.09 & 0.06 \\
& & Simplified & 0.05 & 0.02 & 0.05 & 0.04 \\
& & Detailed & 0.04 & 0.01 & 0.04 & 0.03 \\
\cmidrule{2-7}
& \multirow{3}{*}{$r_O$} 
  & None & 0.04 & 0.05 & 0.11 & 0.07 \\
& & Simplified & 0.03 & 0.01 & 0.02 & 0.02 \\
& & Detailed & 0.02 & 0.01 & 0.01 & 0.01 \\
\bottomrule
\end{tabular}
}
\caption{
Statistical results for ECQA using Llama-2-7b-chat-hf on correct responses.
}
\label{tbl:reasoning_cot_500_llama2_7b_it_new_grads_ecqa_train}
\end{table}

\begin{figure*}[!h]
\centering
\includegraphics[width=\linewidth]{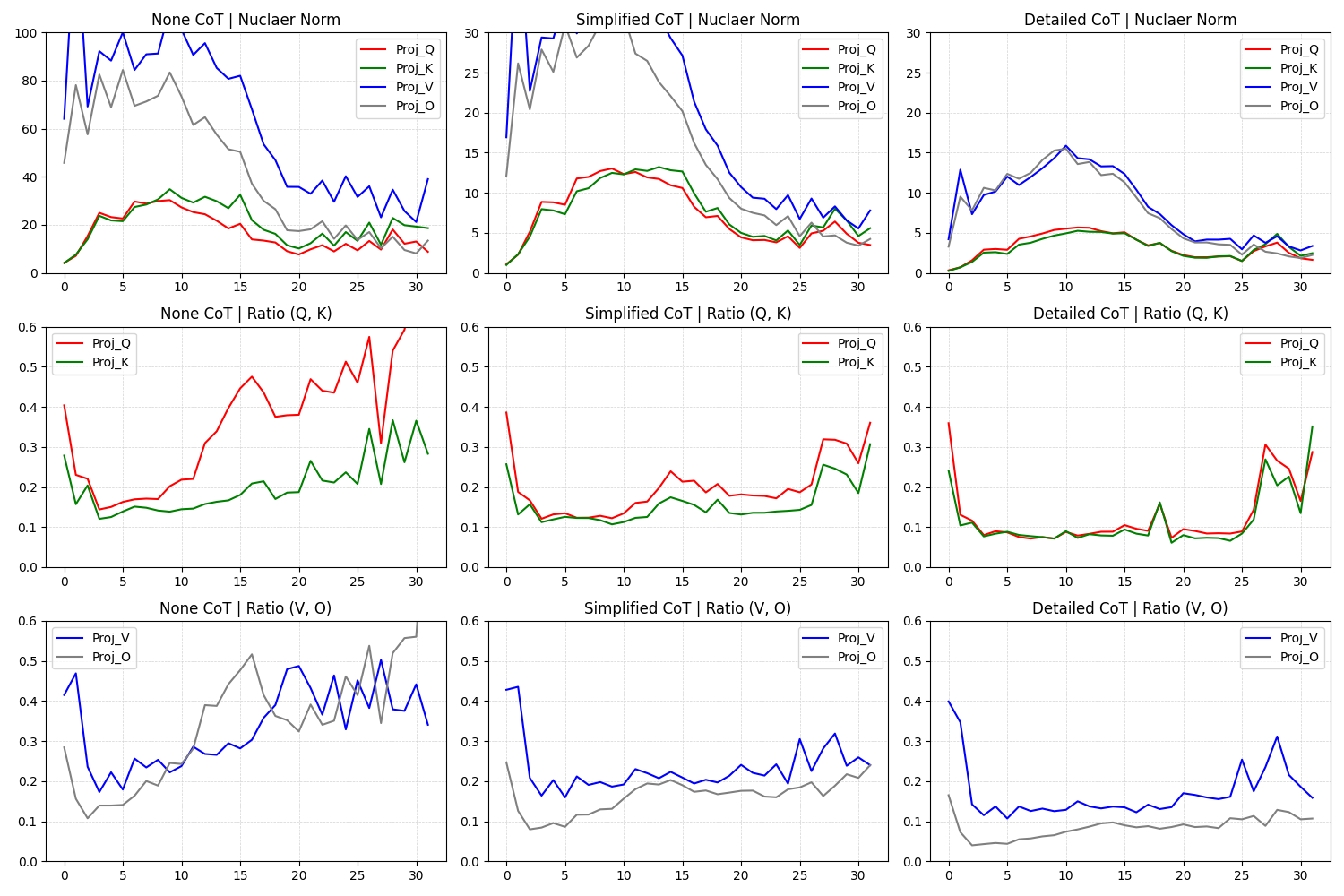}
\caption{Visualization for ECQA using Llama-2-7b-chat-hf on correct responses.}
\label{fig:reasoning_correct_ecqa_it}
\vspace{-2mm}
\end{figure*}

\begin{table}[!tbh]
\centering
\small
\scalebox{0.75}{
\begin{tabular}{l|l|l|cccc}
\toprule
\multirow{2}{*}{\textbf{Dataset}} & \multirow{2}{*}{\textbf{Curve}} & \multirow{2}{*}{\textbf{Cot}} &  \multicolumn{4}{c}{\textbf{Mean Absolute Difference (MAD)}} \\
 & &  & Early & Middle & Last & All\\
\midrule
\multirow{24}{*}{CREAK} & \multirow{3}{*}{$s_Q$} 
  & None & 7.65 & 5.59 & 6.68 & 6.42 \\
& & Simplified & 1.53 & 1.14 & 1.05 & 1.21 \\
& & Detailed & 0.53 & 0.38 & 0.44 & 0.44 \\
\cmidrule{2-7}
& \multirow{3}{*}{$s_K$} 
  & None & 8.34 & 7.09 & 9.21 & 8.02 \\
& & Simplified & 1.57 & 1.45 & 1.53 & 1.48 \\
& & Detailed & 0.50 & 0.40 & 0.57 & 0.48 \\
\cmidrule{2-7}
& \multirow{3}{*}{$s_V$} 
  & None & 55.83 & 19.94 & 16.12 & 28.78 \\
& & Simplified & 9.31 & 3.19 & 2.07 & 4.45 \\
& & Detailed & 2.58 & 0.98 & 0.63 & 1.33 \\
\cmidrule{2-7}
& \multirow{3}{*}{$s_O$} 
  & None & 33.10 & 15.36 & 8.03 & 18.01 \\
& & Simplified & 5.72 & 2.62 & 1.18 & 2.97 \\
& & Detailed & 1.86 & 1.02 & 0.42 & 1.03 \\
\cmidrule{2-7}
 & \multirow{3}{*}{$r_Q$} 
  & None & 0.03 & 0.05 & 0.11 & 0.07 \\
& & Simplified & 0.03 & 0.03 & 0.03 & 0.03 \\
& & Detailed & 0.04 & 0.02 & 0.04 & 0.03 \\
\cmidrule{2-7}
& \multirow{3}{*}{$r_K$} 
  & None & 0.03 & 0.02 & 0.07 & 0.04 \\
& & Simplified & 0.03 & 0.03 & 0.03 & 0.03 \\
& & Detailed & 0.02 & 0.02 & 0.05 & 0.03 \\
\cmidrule{2-7}
 & \multirow{3}{*}{$r_V$} 
  & None & 0.06 & 0.04 & 0.07 & 0.05 \\
& & Simplified & 0.05 & 0.02 & 0.05 & 0.04 \\
& & Detailed & 0.04 & 0.01 & 0.04 & 0.03 \\
\cmidrule{2-7}
& \multirow{3}{*}{$r_O$} 
  & None & 0.04 & 0.05 & 0.12 & 0.07 \\
& & Simplified & 0.03 & 0.01 & 0.02 & 0.02 \\
& & Detailed & 0.02 & 0.00 & 0.01 & 0.01 \\
\bottomrule
\end{tabular}
}
\caption{
Statistical results for CREAK using Llama-2-7b-chat-hf on correct responses.
}
\label{tbl:reasoning_cot_500_llama2_7b_it_new_grads_creak_train}
\end{table}

\begin{figure*}[!h]
\centering
\includegraphics[width=\linewidth]{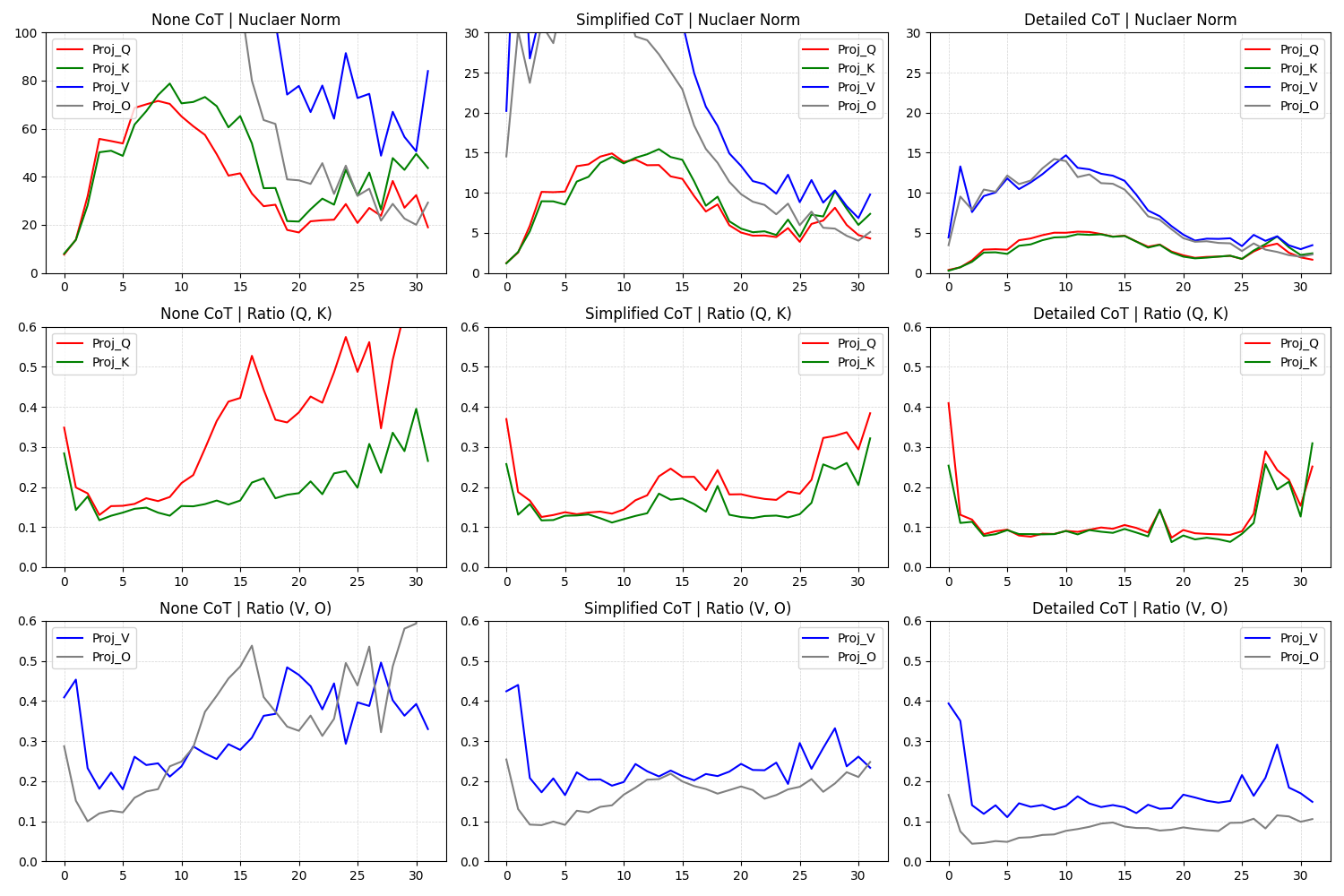}
\caption{Visualization for CREAK using Llama-2-7b-chat-hf on correct responses.}
\label{fig:reasoning_correct_creak_it}
\vspace{-2mm}
\end{figure*}

\begin{table}[!tbh]
\centering
\small
\scalebox{0.75}{
\begin{tabular}{l|l|l|cccc}
\toprule
\multirow{2}{*}{\textbf{Dataset}} & \multirow{2}{*}{\textbf{Curve}} & \multirow{2}{*}{\textbf{Cot}} &  \multicolumn{4}{c}{\textbf{Mean Absolute Difference (MAD)}} \\
 & &  & Early & Middle & Last & All\\
\midrule
\multirow{24}{*}{Sensemaking} & \multirow{3}{*}{$s_Q$} 
  & None & 3.53 & 2.69 & 3.06 & 3.07 \\
& & Simplified & 1.85 & 1.45 & 1.51 & 1.58 \\
& & Detailed & 0.47 & 0.34 & 0.38 & 0.39 \\
\cmidrule{2-7}
& \multirow{3}{*}{$s_K$} 
  & None & 3.85 & 3.34 & 4.61 & 3.93 \\
& & Simplified & 1.94 & 1.81 & 2.14 & 1.93 \\
& & Detailed & 0.45 & 0.35 & 0.51 & 0.43 \\
\cmidrule{2-7}
& \multirow{3}{*}{$s_V$} 
  & None & 23.30 & 7.57 & 6.10 & 11.41 \\
& & Simplified & 10.36 & 3.63 & 2.60 & 5.05 \\
& & Detailed & 2.17 & 0.92 & 0.58 & 1.17 \\
\cmidrule{2-7}
& \multirow{3}{*}{$s_O$} 
  & None & 14.40 & 6.56 & 3.24 & 7.64 \\
& & Simplified & 6.65 & 3.25 & 1.42 & 3.51 \\
& & Detailed & 1.71 & 1.01 & 0.38 & 0.96 \\
\cmidrule{2-7}
 & \multirow{3}{*}{$r_Q$} 
  & None & 0.03 & 0.03 & 0.08 & 0.05 \\
& & Simplified & 0.03 & 0.02 & 0.03 & 0.03 \\
& & Detailed & 0.04 & 0.02 & 0.04 & 0.03 \\
\cmidrule{2-7}
& \multirow{3}{*}{$r_K$} 
  & None & 0.04 & 0.01 & 0.06 & 0.04 \\
& & Simplified & 0.03 & 0.03 & 0.02 & 0.03 \\
& & Detailed & 0.02 & 0.02 & 0.05 & 0.03 \\
\cmidrule{2-7}
 & \multirow{3}{*}{$r_V$} 
  & None & 0.05 & 0.03 & 0.06 & 0.05 \\
& & Simplified & 0.05 & 0.02 & 0.05 & 0.04 \\
& & Detailed & 0.04 & 0.01 & 0.04 & 0.03 \\
\cmidrule{2-7}
& \multirow{3}{*}{$r_O$} 
  & None & 0.03 & 0.03 & 0.09 & 0.05 \\
& & Simplified & 0.03 & 0.02 & 0.03 & 0.02 \\
& & Detailed & 0.02 & 0.01 & 0.01 & 0.01 \\
\bottomrule
\end{tabular}
}
\caption{
Statistical results for Sensemaking using Llama-2-7b-chat-hf on correct responses.
}
\label{tbl:reasoning_cot_500_llama2_7b_it_new_grads_sensemaking_train}
\end{table}

\begin{figure*}[!h]
\centering
\includegraphics[width=\linewidth]{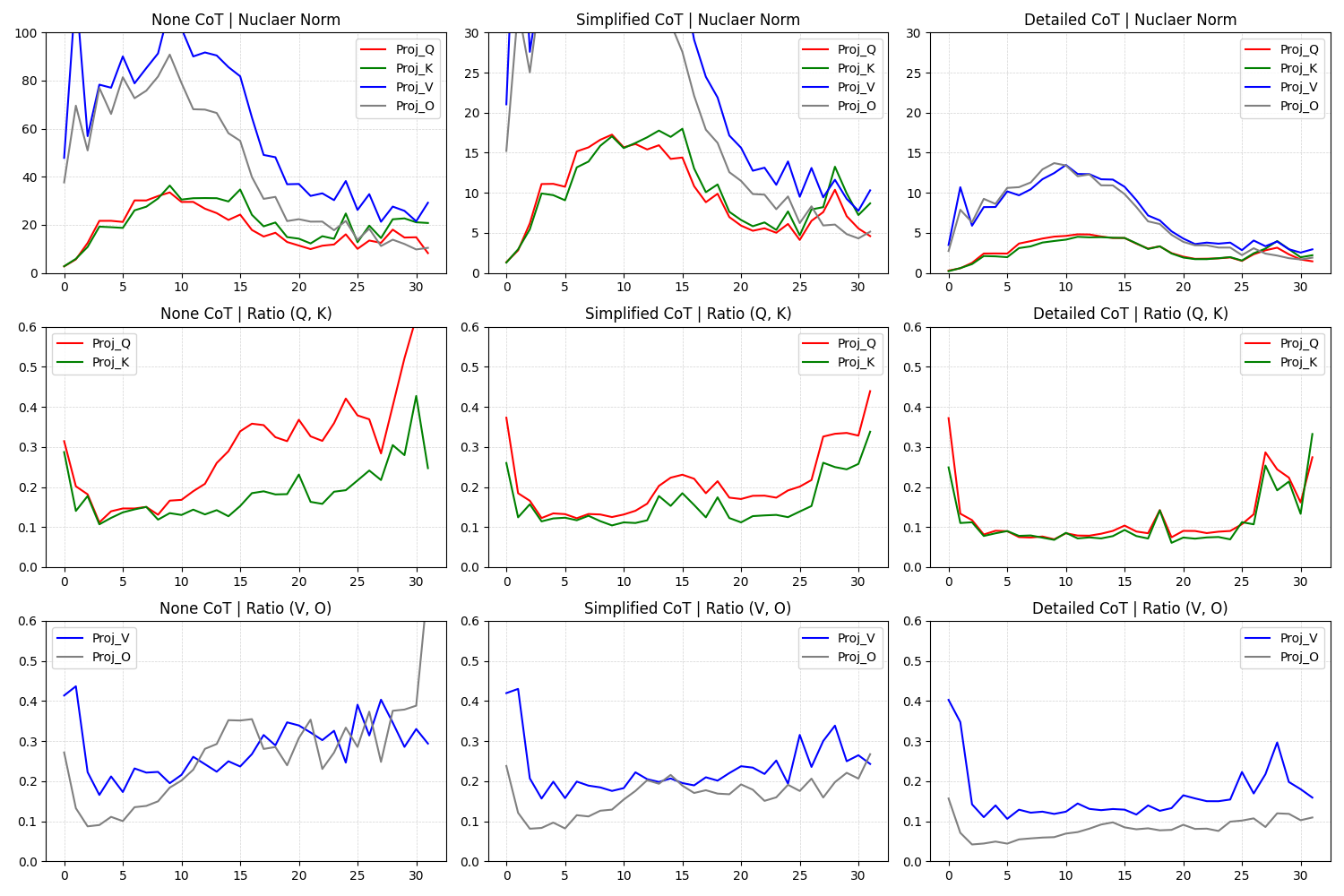}
\caption{Visualization for Sensemaking using Llama-2-7b-chat-hf on correct responses.}
\label{fig:reasoning_correct_sensemaking_it}
\vspace{-2mm}
\end{figure*}

\subsubsection{Wiki Tasks}

The visualizations and statistical results on Wiki tasks are shown in Figure \ref{fig:wiki_correct_it} and Table \ref{tbl:wiki_wiki_popularity_llama2_7b_it_new_grads}.

\begin{table}[!tbh]
\centering
\small
\scalebox{0.75}{
\begin{tabular}{l|l|l|cccc}
\toprule
\multirow{2}{*}{\textbf{Dataset}} & \multirow{2}{*}{\textbf{Curve}} & \multirow{2}{*}{\textbf{Cot}} &  \multicolumn{4}{c}{\textbf{Mean Absolute Difference (MAD)}} \\
 & &  & Early & Middle & Last & All\\
\midrule
\multirow{32}{*}{Wiki} & \multirow{3}{*}{$s_Q$} 
  & Len 100 & 0.67 & 0.28 & 0.58 & 0.50 \\
& & Len 500 & 0.48 & 0.23 & 0.43 & 0.38 \\
& & Len 1000 & 0.41 & 0.20 & 0.36 & 0.32 \\
& & Unpopular & 0.96 & 0.78 & 0.77 & 0.83 \\
\cmidrule{2-7}
& \multirow{3}{*}{$s_K$} 
  & Len 100 & 0.65 & 0.35 & 0.67 & 0.54 \\
& & Len 500 & 0.45 & 0.23 & 0.51 & 0.40 \\
& & Len 1000 & 0.38 & 0.20 & 0.43 & 0.34 \\
& & Unpopular & 0.97 & 0.92 & 0.87 & 0.90 \\
\cmidrule{2-7}
& \multirow{3}{*}{$s_V$} 
  & Len 100 & 3.75 & 0.81 & 0.47 & 1.56 \\
& & Len 500 & 2.33 & 0.73 & 0.40 & 1.09 \\
& & Len 1000 & 1.85 & 0.65 & 0.38 & 0.91 \\
& & Unpopular & 7.65 & 1.50 & 0.66 & 2.95 \\
\cmidrule{2-7}
& \multirow{3}{*}{$s_O$} 
  & Len 100 & 2.29 & 0.87 & 0.43 & 1.12 \\
& & Len 500 & 1.64 & 0.76 & 0.38 & 0.88 \\
& & Len 1000 & 1.37 & 0.70 & 0.35 & 0.77 \\
& & Unpopular & 3.88 & 1.49 & 0.61 & 1.93 \\
\cmidrule{2-7}
 & \multirow{3}{*}{$r_Q$} 
  & Len 100 & 0.03 & 0.01 & 0.02 & 0.02 \\
& & Len 500 & 0.03 & 0.01 & 0.02 & 0.02 \\
& & Len 1000 & 0.04 & 0.01 & 0.03 & 0.02 \\
& & Unpopular & 0.03 & 0.03 & 0.02 & 0.02 \\
\cmidrule{2-7}
& \multirow{3}{*}{$r_K$} 
  & Len 100 & 0.03 & 0.01 & 0.02 & 0.02 \\
& & Len 500 & 0.02 & 0.01 & 0.02 & 0.02 \\
& & Len 1000 & 0.02 & 0.02 & 0.03 & 0.02 \\
& & Unpopular & 0.04 & 0.02 & 0.02 & 0.02 \\
\cmidrule{2-7}
 & \multirow{3}{*}{$r_V$} 
  & Len 100 & 0.05 & 0.02 & 0.02 & 0.03 \\
& & Len 500 & 0.04 & 0.01 & 0.02 & 0.02 \\
& & Len 1000 & 0.04 & 0.01 & 0.02 & 0.02 \\
& & Unpopular & 0.05 & 0.02 & 0.02 & 0.03 \\
\cmidrule{2-7}
& \multirow{3}{*}{$r_O$} 
  & Len 100 & 0.02 & 0.01 & 0.02 & 0.02 \\
& & Len 500 & 0.02 & 0.01 & 0.01 & 0.01 \\
& & Len 1000 & 0.01 & 0.01 & 0.01 & 0.01 \\
& & Unpopular & 0.02 & 0.02 & 0.01 & 0.02 \\
\bottomrule
\end{tabular}
}
\caption{
Statistical results for Wiki using Llama-2-7b-chat-hf on correct responses.
}
\label{tbl:wiki_wiki_popularity_llama2_7b_it_new_grads}
\end{table}

\begin{figure*}[!h]
\centering
\includegraphics[width=\linewidth]{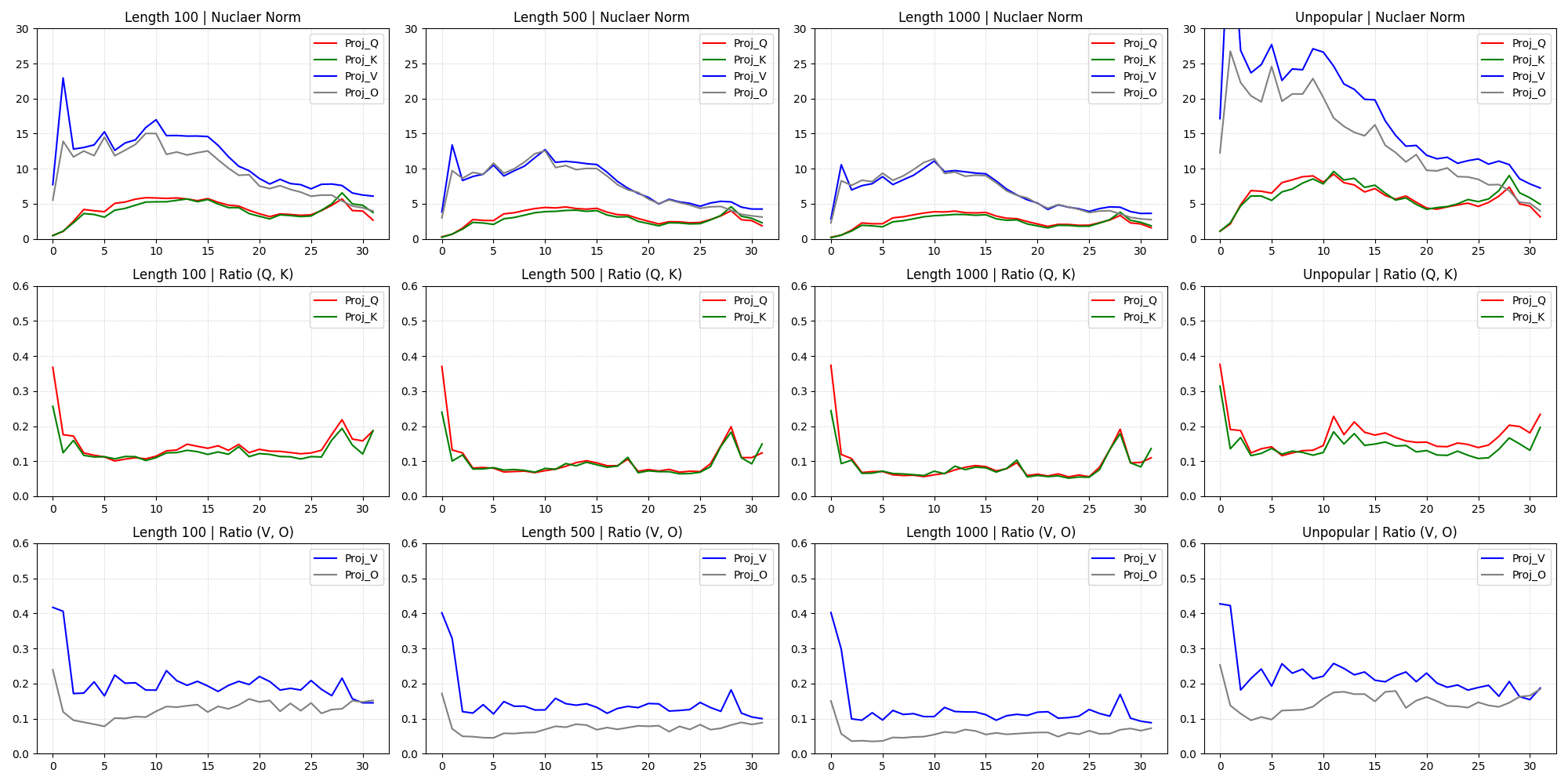}
\caption{Visualization for Wiki tasks using Llama-2-7b-chat-hf on correct responses.}
\label{fig:wiki_correct_it}
\vspace{-2mm}
\end{figure*}

\subsection{Instructed LLM on Irrelevant Responses}

\subsubsection{Reasoning Tasks}

The visualizations and statistical results on MATH tasks: MATH-Algebra (Figure \ref{fig:math_nonsense_algebra_it}, Table \ref{tbl:math_lighteval_math_wrong_answer_shuffle_llama2_7b_it_new_grads_algebra}), MATH-Counting (Figure \ref{fig:math_nonsense_counting_it}, Table \ref{tbl:math_lighteval_math_wrong_answer_shuffle_llama2_7b_it_new_grads_counting_probability}), MATH-Geometry (Figure \ref{fig:math_nonsense_geometry_it}, Table \ref{tbl:math_lighteval_math_wrong_answer_shuffle_llama2_7b_it_new_grads_geometry}).

\begin{table}[!tbh]
\centering
\small
\scalebox{0.75}{
\begin{tabular}{l|l|l|cccc}
\toprule
\multirow{2}{*}{\textbf{Dataset}} & \multirow{2}{*}{\textbf{Curve}} & \multirow{2}{*}{\textbf{Cot}} &  \multicolumn{4}{c}{\textbf{Mean Absolute Difference (MAD)}} \\
 & &  & Early & Middle & Last & All\\
\midrule
\multirow{16}{*}{Algebra} & \multirow{3}{*}{$s_Q$} 
  & Simplified & 0.88 & 0.77 & 0.62 & 0.75 \\
& & Detailed & 0.52 & 0.43 & 0.37 & 0.43 \\
\cmidrule{2-7}
& \multirow{3}{*}{$s_K$} 
  & Simplified & 0.86 & 0.91 & 0.78 & 0.83 \\
& & Detailed & 0.50 & 0.46 & 0.43 & 0.46 \\
\cmidrule{2-7}
& \multirow{3}{*}{$s_V$} 
  & Simplified & 4.54 & 1.94 & 0.72 & 2.26 \\
& & Detailed & 2.39 & 1.17 & 0.41 & 1.26 \\
\cmidrule{2-7}
& \multirow{3}{*}{$s_O$} 
  & Simplified & 2.73 & 1.88 & 0.59 & 1.62 \\
& & Detailed & 1.51 & 1.15 & 0.36 & 0.94 \\
\cmidrule{2-7}
 & \multirow{3}{*}{$r_Q$} 
  & Simplified & 0.03 & 0.03 & 0.04 & 0.03 \\
& & Detailed & 0.03 & 0.02 & 0.04 & 0.03 \\
\cmidrule{2-7}
& \multirow{3}{*}{$r_K$} 
  & Simplified & 0.02 & 0.02 & 0.04 & 0.03 \\
& & Detailed & 0.02 & 0.02 & 0.05 & 0.03 \\
\cmidrule{2-7}
 & \multirow{3}{*}{$r_V$} 
  & Simplified & 0.04 & 0.02 & 0.03 & 0.03 \\
& & Detailed & 0.04 & 0.01 & 0.03 & 0.03 \\
\cmidrule{2-7}
& \multirow{3}{*}{$r_O$} 
  & Simplified & 0.02 & 0.01 & 0.02 & 0.01 \\
& & Detailed & 0.02 & 0.00 & 0.02 & 0.01 \\
\bottomrule
\end{tabular}
}
\caption{
Statistical results for MATH-Algebra using Llama-2-7b-chat-hf on irrelevant responses.
}
\label{tbl:math_lighteval_math_wrong_answer_shuffle_llama2_7b_it_new_grads_algebra}
\end{table}

\begin{figure}[!h]
\centering
\includegraphics[width=\linewidth]{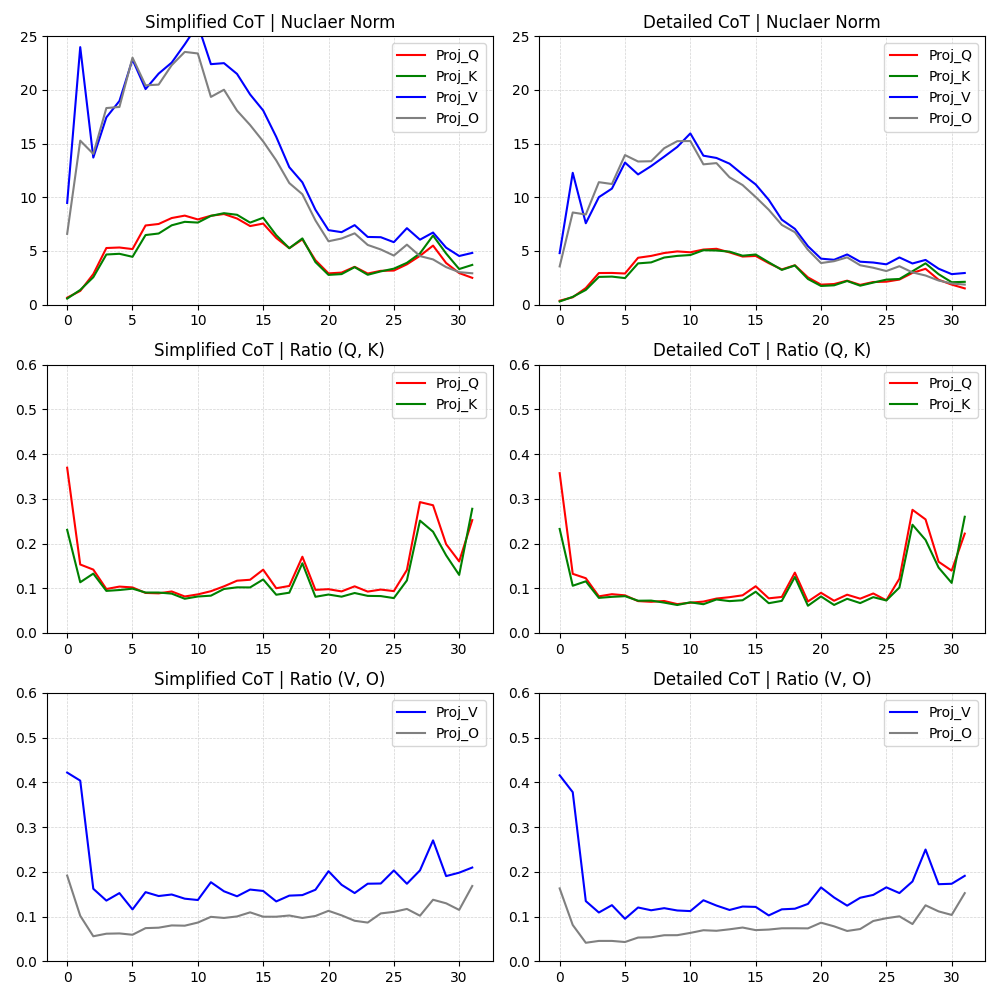}
\caption{Visualization for MATH-Algebra using Llama-2-7b-chat-hf on irrelevant responses.}
\label{fig:math_nonsense_algebra_it}
\vspace{-2mm}
\end{figure}

\begin{table}[!tbh]
\centering
\small
\scalebox{0.75}{
\begin{tabular}{l|l|l|cccc}
\toprule
\multirow{2}{*}{\textbf{Dataset}} & \multirow{2}{*}{\textbf{Curve}} & \multirow{2}{*}{\textbf{Cot}} &  \multicolumn{4}{c}{\textbf{Mean Absolute Difference (MAD)}} \\
 & &  & Early & Middle & Last & All\\
\midrule
\multirow{16}{*}{Counting} & \multirow{3}{*}{$s_Q$} 
  & Simplified & 0.82 & 0.64 & 0.50 & 0.64 \\
& & Detailed & 0.51 & 0.37 & 0.34 & 0.40 \\
\cmidrule{2-7}
& \multirow{3}{*}{$s_K$} 
  & Simplified & 0.78 & 0.76 & 0.63 & 0.70 \\
& & Detailed & 0.48 & 0.40 & 0.43 & 0.42 \\
\cmidrule{2-7}
& \multirow{3}{*}{$s_V$} 
  & Simplified & 4.36 & 1.82 & 0.79 & 2.17 \\
& & Detailed & 2.40 & 1.08 & 0.48 & 1.24 \\
\cmidrule{2-7}
& \multirow{3}{*}{$s_O$} 
  & Simplified & 2.79 & 1.77 & 0.61 & 1.60 \\
& & Detailed & 1.64 & 1.10 & 0.39 & 0.97 \\
\cmidrule{2-7}
 & \multirow{3}{*}{$r_Q$} 
  & Simplified & 0.04 & 0.02 & 0.04 & 0.03 \\
& & Detailed & 0.03 & 0.02 & 0.04 & 0.03 \\
\cmidrule{2-7}
& \multirow{3}{*}{$r_K$} 
  & Simplified & 0.02 & 0.02 & 0.04 & 0.03 \\
& & Detailed & 0.02 & 0.02 & 0.04 & 0.03 \\
\cmidrule{2-7}
 & \multirow{3}{*}{$r_V$} 
  & Simplified & 0.05 & 0.02 & 0.03 & 0.03 \\
& & Detailed & 0.04 & 0.01 & 0.02 & 0.02 \\
\cmidrule{2-7}
& \multirow{3}{*}{$r_O$} 
  & Simplified & 0.02 & 0.00 & 0.01 & 0.01 \\
& & Detailed & 0.02 & 0.00 & 0.01 & 0.01 \\
\bottomrule
\end{tabular}
}
\caption{
Statistical results for MATH-Counting using Llama-2-7b-chat-hf on irrelevant responses.
}
\label{tbl:math_lighteval_math_wrong_answer_shuffle_llama2_7b_it_new_grads_counting_probability}
\end{table}

\begin{figure}[!h]
\centering
\includegraphics[width=\linewidth]{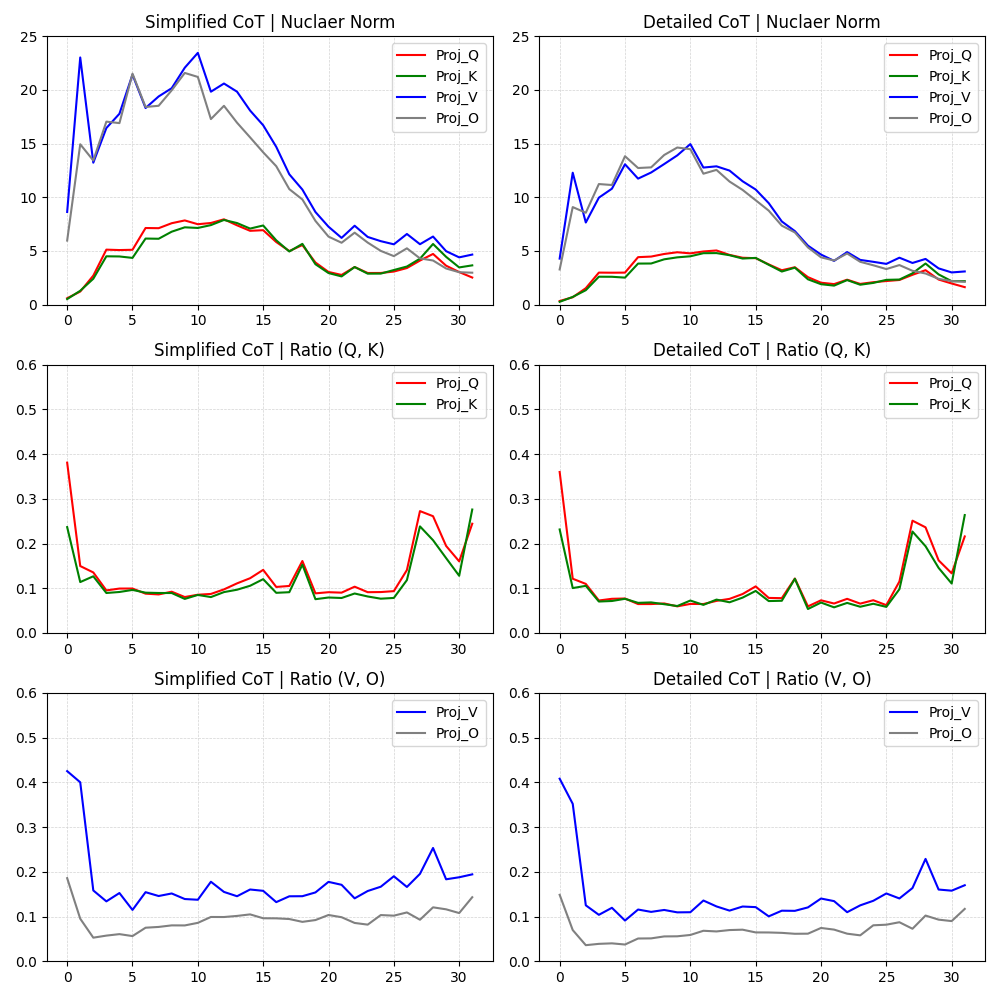}
\caption{Visualization for MATH-Counting using Llama-2-7b-chat-hf on irrelevant responses.}
\label{fig:math_nonsense_counting_it}
\vspace{-2mm}
\end{figure}

\begin{table}[!tbh]
\centering
\small
\scalebox{0.75}{
\begin{tabular}{l|l|l|cccc}
\toprule
\multirow{2}{*}{\textbf{Dataset}} & \multirow{2}{*}{\textbf{Curve}} & \multirow{2}{*}{\textbf{Cot}} &  \multicolumn{4}{c}{\textbf{Mean Absolute Difference (MAD)}} \\
 & &  & Early & Middle & Last & All\\
\midrule
\multirow{16}{*}{Geometry} & \multirow{3}{*}{$s_Q$} 
  & Simplified & 0.73 & 0.54 & 0.61 & 0.63 \\
& & Detailed & 0.54 & 0.33 & 0.41 & 0.42 \\
\cmidrule{2-7}
& \multirow{3}{*}{$s_K$} 
  & Simplified & 0.70 & 0.64 & 0.77 & 0.70 \\
& & Detailed & 0.52 & 0.40 & 0.52 & 0.47 \\
\cmidrule{2-7}
& \multirow{3}{*}{$s_V$} 
  & Simplified & 3.80 & 1.55 & 0.69 & 1.88 \\
& & Detailed & 2.48 & 1.06 & 0.55 & 1.26 \\
\cmidrule{2-7}
& \multirow{3}{*}{$s_O$} 
  & Simplified & 2.50 & 1.57 & 0.61 & 1.46 \\
& & Detailed & 1.75 & 1.11 & 0.48 & 1.04 \\
\cmidrule{2-7}
 & \multirow{3}{*}{$r_Q$} 
  & Simplified & 0.04 & 0.02 & 0.04 & 0.03 \\
& & Detailed & 0.04 & 0.02 & 0.04 & 0.03 \\
\cmidrule{2-7}
& \multirow{3}{*}{$r_K$} 
  & Simplified & 0.02 & 0.02 & 0.04 & 0.03 \\
& & Detailed & 0.02 & 0.02 & 0.04 & 0.03 \\
\cmidrule{2-7}
 & \multirow{3}{*}{$r_V$} 
  & Simplified & 0.04 & 0.01 & 0.03 & 0.03 \\
& & Detailed & 0.04 & 0.01 & 0.02 & 0.02 \\
\cmidrule{2-7}
& \multirow{3}{*}{$r_O$} 
  & Simplified & 0.02 & 0.01 & 0.01 & 0.01 \\
& & Detailed & 0.02 & 0.00 & 0.01 & 0.01 \\
\bottomrule
\end{tabular}
}
\caption{
Statistical results for MATH-Geometry using Llama-2-7b-chat-hf on irrelevant responses.
}
\label{tbl:math_lighteval_math_wrong_answer_shuffle_llama2_7b_it_new_grads_geometry}
\end{table}

\begin{figure}[!h]
\centering
\includegraphics[width=\linewidth]{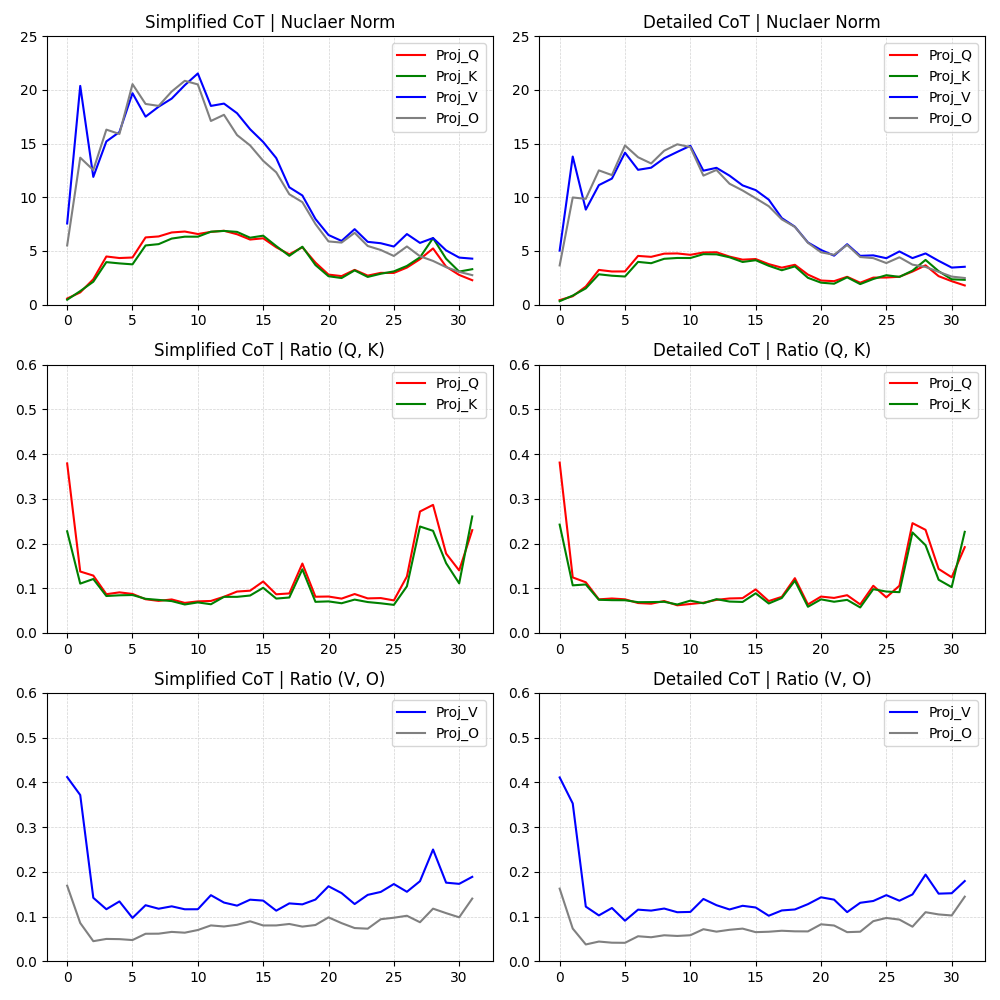}
\caption{Visualization for MATH-Geometry using Llama-2-7b-chat-hf on irrelevant responses.}
\label{fig:math_nonsense_geometry_it}
\vspace{-2mm}
\end{figure}

The visualizations and statistical results on other reasoning tasks: AQuA (Figure \ref{fig:reasoning_nonsense_aqua_it}, Table \ref{tbl:reasoning_cot_500_wrong_answer_shuffle_llama2_7b_it_new_grads_aqua_train}), GSM8K (Figure \ref{fig:reasoning_nonsense_gsm8k_it}, Table \ref{tbl:reasoning_cot_500_wrong_answer_shuffle_llama2_7b_it_new_grads_gsm8k_train}), StrategyQA (Figure \ref{fig:reasoning_nonsense_strategyqa_it}, Table \ref{tbl:reasoning_cot_500_wrong_answer_shuffle_llama2_7b_it_new_grads_strategyqa_train}), ECQA (Figure \ref{fig:reasoning_nonsense_ecqa_it}, Table \ref{tbl:reasoning_cot_500_wrong_answer_shuffle_llama2_7b_it_new_grads_ecqa_train}), CREAK (Figure \ref{fig:reasoning_nonsense_creak_it}, Table \ref{tbl:reasoning_cot_500_wrong_answer_shuffle_llama2_7b_it_new_grads_creak_train}), Sensemaking (Figure \ref{fig:reasoning_nonsense_sensemaking_it}, Table \ref{tbl:reasoning_cot_500_wrong_answer_shuffle_llama2_7b_it_new_grads_sensemaking_train}).

\begin{table}[!tbh]
\centering
\small
\scalebox{0.75}{
\begin{tabular}{l|l|l|cccc}
\toprule
\multirow{2}{*}{\textbf{Dataset}} & \multirow{2}{*}{\textbf{Curve}} & \multirow{2}{*}{\textbf{Cot}} &  \multicolumn{4}{c}{\textbf{Mean Absolute Difference (MAD)}} \\
 & &  & Early & Middle & Last & All\\
\midrule
\multirow{24}{*}{AQuA} & \multirow{3}{*}{$s_Q$} 
  & None & 9.67 & 5.34 & 3.97 & 6.05 \\
& & Simplified & 1.46 & 1.10 & 0.62 & 1.03 \\
& & Detailed & 0.66 & 0.53 & 0.47 & 0.54 \\
\cmidrule{2-7}
& \multirow{3}{*}{$s_K$} 
  & None & 11.48 & 7.26 & 7.76 & 8.60 \\
& & Simplified & 1.40 & 1.44 & 0.79 & 1.16 \\
& & Detailed & 0.62 & 0.61 & 0.54 & 0.57 \\
\cmidrule{2-7}
& \multirow{3}{*}{$s_V$} 
  & None & 81.42 & 12.90 & 6.94 & 31.05 \\
& & Simplified & 7.57 & 2.92 & 1.19 & 3.57 \\
& & Detailed & 3.17 & 1.31 & 0.65 & 1.60 \\
\cmidrule{2-7}
& \multirow{3}{*}{$s_O$} 
  & None & 32.04 & 8.00 & 4.11 & 13.96 \\
& & Simplified & 4.53 & 2.70 & 0.83 & 2.48 \\
& & Detailed & 2.11 & 1.30 & 0.51 & 1.22 \\
\cmidrule{2-7}
 & \multirow{3}{*}{$r_Q$} 
  & None & 0.07 & 0.05 & 0.11 & 0.08 \\
& & Simplified & 0.04 & 0.02 & 0.03 & 0.03 \\
& & Detailed & 0.04 & 0.02 & 0.05 & 0.03 \\
\cmidrule{2-7}
& \multirow{3}{*}{$r_K$} 
  & None & 0.05 & 0.01 & 0.03 & 0.03 \\
& & Simplified & 0.03 & 0.02 & 0.04 & 0.03 \\
& & Detailed & 0.02 & 0.02 & 0.05 & 0.03 \\
\cmidrule{2-7}
 & \multirow{3}{*}{$r_V$} 
  & None & 0.09 & 0.03 & 0.04 & 0.05 \\
& & Simplified & 0.06 & 0.02 & 0.02 & 0.03 \\
& & Detailed & 0.05 & 0.01 & 0.03 & 0.03 \\
\cmidrule{2-7}
& \multirow{3}{*}{$r_O$} 
  & None & 0.05 & 0.03 & 0.06 & 0.05 \\
& & Simplified & 0.02 & 0.01 & 0.02 & 0.02 \\
& & Detailed & 0.02 & 0.00 & 0.01 & 0.01 \\
\bottomrule
\end{tabular}
}
\caption{
Statistical results for AQuA using Llama-2-7b-chat-hf on irrelevant responses.
}
\label{tbl:reasoning_cot_500_wrong_answer_shuffle_llama2_7b_it_new_grads_aqua_train}
\end{table}

\begin{figure*}[!h]
\centering
\includegraphics[width=\linewidth]{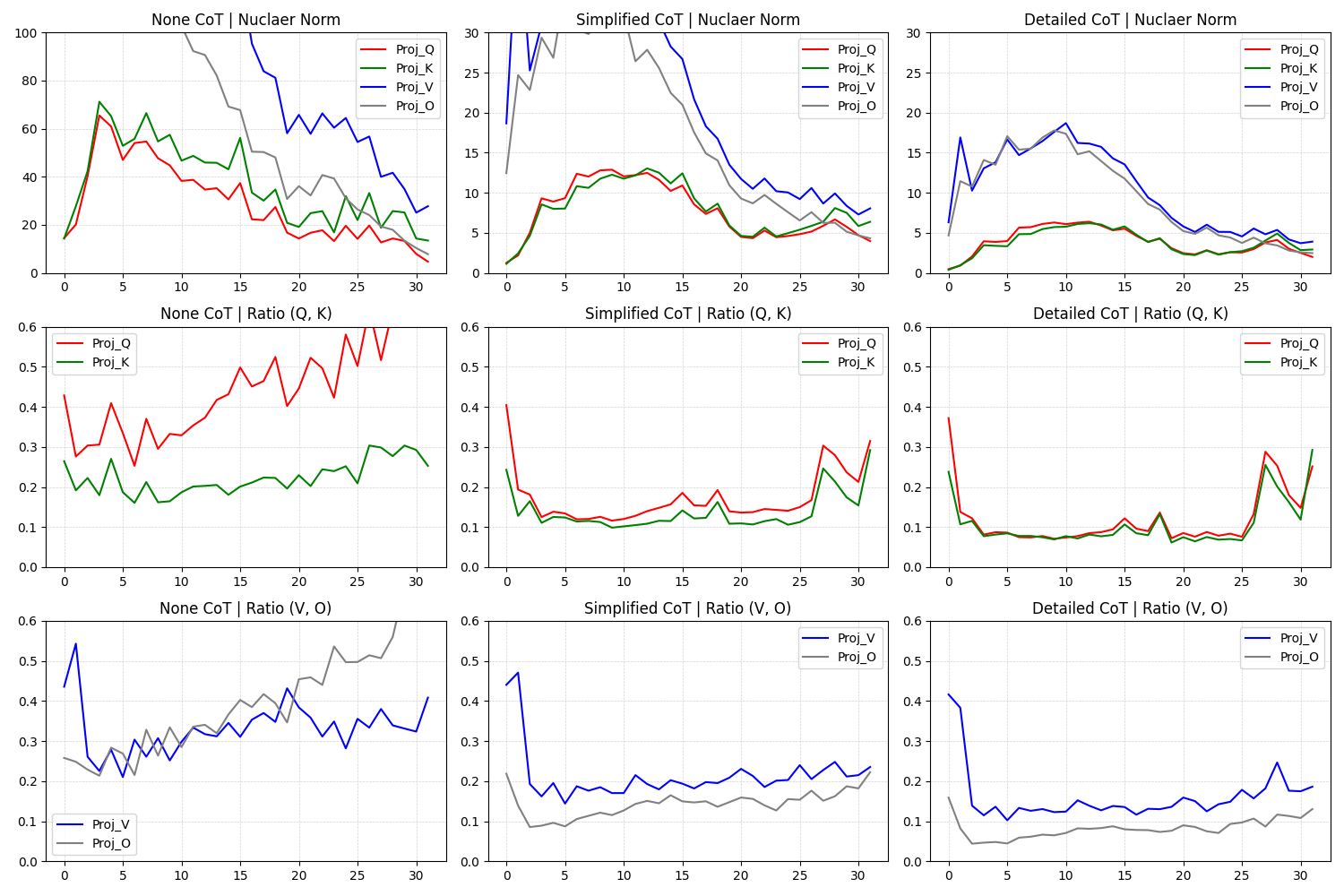}
\caption{Visualization for AQuA using Llama-2-7b-chat-hf on irrelevant responses.}
\label{fig:reasoning_nonsense_aqua_it}
\vspace{-2mm}
\end{figure*}

\begin{table}[!tbh]
\centering
\small
\scalebox{0.75}{
\begin{tabular}{l|l|l|cccc}
\toprule
\multirow{2}{*}{\textbf{Dataset}} & \multirow{2}{*}{\textbf{Curve}} & \multirow{2}{*}{\textbf{Cot}} &  \multicolumn{4}{c}{\textbf{Mean Absolute Difference (MAD)}} \\
 & &  & Early & Middle & Last & All\\
\midrule
\multirow{24}{*}{GSM8K} & \multirow{3}{*}{$s_Q$} 
  & None & 4.90 & 3.83 & 2.67 & 3.56 \\
& & Simplified & 1.02 & 0.86 & 0.53 & 0.78 \\
& & Detailed & 0.59 & 0.48 & 0.45 & 0.49 \\
\cmidrule{2-7}
& \multirow{3}{*}{$s_K$} 
  & None & 5.42 & 5.16 & 3.73 & 4.45 \\
& & Simplified & 0.97 & 1.11 & 0.69 & 0.88 \\
& & Detailed & 0.55 & 0.55 & 0.58 & 0.55 \\
\cmidrule{2-7}
& \multirow{3}{*}{$s_V$} 
  & None & 31.91 & 11.68 & 7.97 & 15.55 \\
& & Simplified & 5.40 & 1.99 & 1.00 & 2.58 \\
& & Detailed & 2.86 & 1.09 & 0.66 & 1.45 \\
\cmidrule{2-7}
& \multirow{3}{*}{$s_O$} 
  & None & 19.67 & 9.77 & 3.96 & 10.29 \\
& & Simplified & 3.29 & 1.88 & 0.70 & 1.81 \\
& & Detailed & 2.05 & 1.13 & 0.47 & 1.13 \\
\cmidrule{2-7}
 & \multirow{3}{*}{$r_Q$} 
  & None & 0.03 & 0.03 & 0.06 & 0.04 \\
& & Simplified & 0.03 & 0.02 & 0.03 & 0.03 \\
& & Detailed & 0.04 & 0.02 & 0.05 & 0.03 \\
\cmidrule{2-7}
& \multirow{3}{*}{$r_K$} 
  & None & 0.03 & 0.03 & 0.05 & 0.04 \\
& & Simplified & 0.02 & 0.02 & 0.03 & 0.03 \\
& & Detailed & 0.02 & 0.03 & 0.05 & 0.03 \\
\cmidrule{2-7}
 & \multirow{3}{*}{$r_V$} 
  & None & 0.07 & 0.03 & 0.05 & 0.04 \\
& & Simplified & 0.05 & 0.01 & 0.03 & 0.03 \\
& & Detailed & 0.05 & 0.01 & 0.03 & 0.03 \\
\cmidrule{2-7}
& \multirow{3}{*}{$r_O$} 
  & None & 0.03 & 0.03 & 0.07 & 0.04 \\
& & Simplified & 0.02 & 0.01 & 0.02 & 0.01 \\
& & Detailed & 0.02 & 0.00 & 0.01 & 0.01 \\
\bottomrule
\end{tabular}
}
\caption{
Statistical results for GSM8K using Llama-2-7b-chat-hf on irrelevant responses.
}
\label{tbl:reasoning_cot_500_wrong_answer_shuffle_llama2_7b_it_new_grads_gsm8k_train}
\end{table}

\begin{figure*}[!h]
\centering
\includegraphics[width=\linewidth]{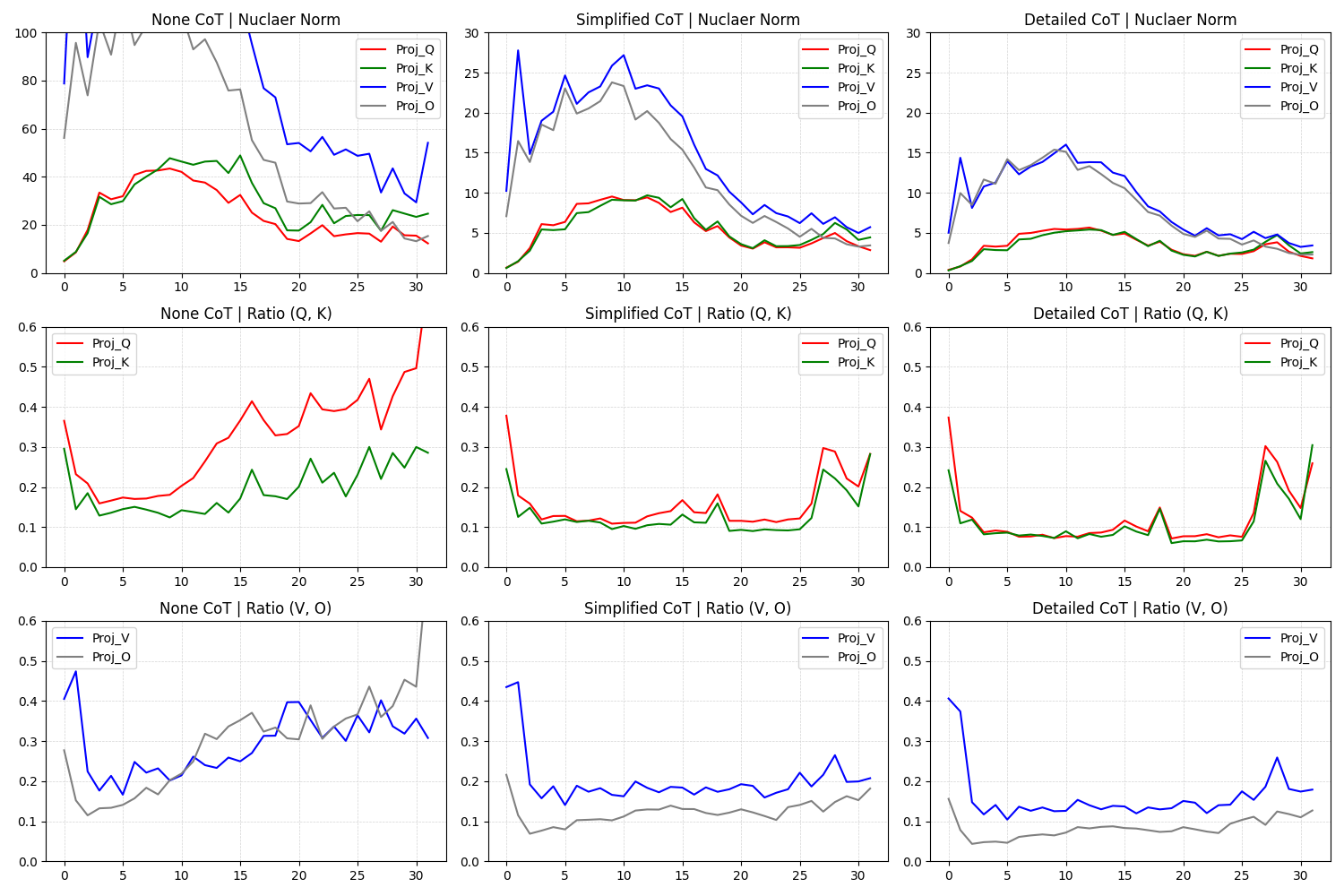}
\caption{Visualization for GSM8K using Llama-2-7b-chat-hf on irrelevant responses.}
\label{fig:reasoning_nonsense_gsm8k_it}
\vspace{-2mm}
\end{figure*}

\begin{table}[!tbh]
\centering
\small
\scalebox{0.75}{
\begin{tabular}{l|l|l|cccc}
\toprule
\multirow{2}{*}{\textbf{Dataset}} & \multirow{2}{*}{\textbf{Curve}} & \multirow{2}{*}{\textbf{Cot}} &  \multicolumn{4}{c}{\textbf{Mean Absolute Difference (MAD)}} \\
 & &  & Early & Middle & Last & All\\
\midrule
\multirow{24}{*}{StrategyQA} & \multirow{3}{*}{$s_Q$} 
  & None & 6.22 & 5.35 & 6.41 & 5.71 \\
& & Simplified & 1.05 & 0.59 & 0.69 & 0.76 \\
& & Detailed & 0.50 & 0.33 & 0.38 & 0.39 \\
\cmidrule{2-7}
& \multirow{3}{*}{$s_K$} 
  & None & 6.13 & 7.26 & 8.85 & 7.09 \\
& & Simplified & 0.95 & 0.69 & 0.88 & 0.81 \\
& & Detailed & 0.47 & 0.37 & 0.52 & 0.45 \\
\cmidrule{2-7}
& \multirow{3}{*}{$s_V$} 
  & None & 46.97 & 17.60 & 17.59 & 25.55 \\
& & Simplified & 6.61 & 1.55 & 1.37 & 2.90 \\
& & Detailed & 2.58 & 0.85 & 0.66 & 1.28 \\
\cmidrule{2-7}
& \multirow{3}{*}{$s_O$} 
  & None & 28.58 & 14.76 & 9.00 & 16.38 \\
& & Simplified & 3.79 & 1.32 & 0.80 & 1.84 \\
& & Detailed & 1.88 & 0.80 & 0.51 & 0.99 \\
\cmidrule{2-7}
 & \multirow{3}{*}{$r_Q$} 
  & None & 0.03 & 0.05 & 0.09 & 0.06 \\
& & Simplified & 0.03 & 0.01 & 0.02 & 0.02 \\
& & Detailed & 0.04 & 0.01 & 0.03 & 0.03 \\
\cmidrule{2-7}
& \multirow{3}{*}{$r_K$} 
  & None & 0.03 & 0.01 & 0.04 & 0.03 \\
& & Simplified & 0.03 & 0.01 & 0.03 & 0.02 \\
& & Detailed & 0.02 & 0.02 & 0.03 & 0.02 \\
\cmidrule{2-7}
 & \multirow{3}{*}{$r_V$} 
  & None & 0.07 & 0.04 & 0.08 & 0.06 \\
& & Simplified & 0.05 & 0.02 & 0.03 & 0.03 \\
& & Detailed & 0.04 & 0.01 & 0.03 & 0.03 \\
\cmidrule{2-7}
& \multirow{3}{*}{$r_O$} 
  & None & 0.04 & 0.05 & 0.10 & 0.06 \\
& & Simplified & 0.03 & 0.01 & 0.01 & 0.01 \\
& & Detailed & 0.02 & 0.01 & 0.01 & 0.01 \\
\bottomrule
\end{tabular}
}
\caption{
Statistical results for StrategyQA using Llama-2-7b-chat-hf on irrelevant responses.
}
\label{tbl:reasoning_cot_500_wrong_answer_shuffle_llama2_7b_it_new_grads_strategyqa_train}
\end{table}

\begin{figure*}[!h]
\centering
\includegraphics[width=\linewidth]{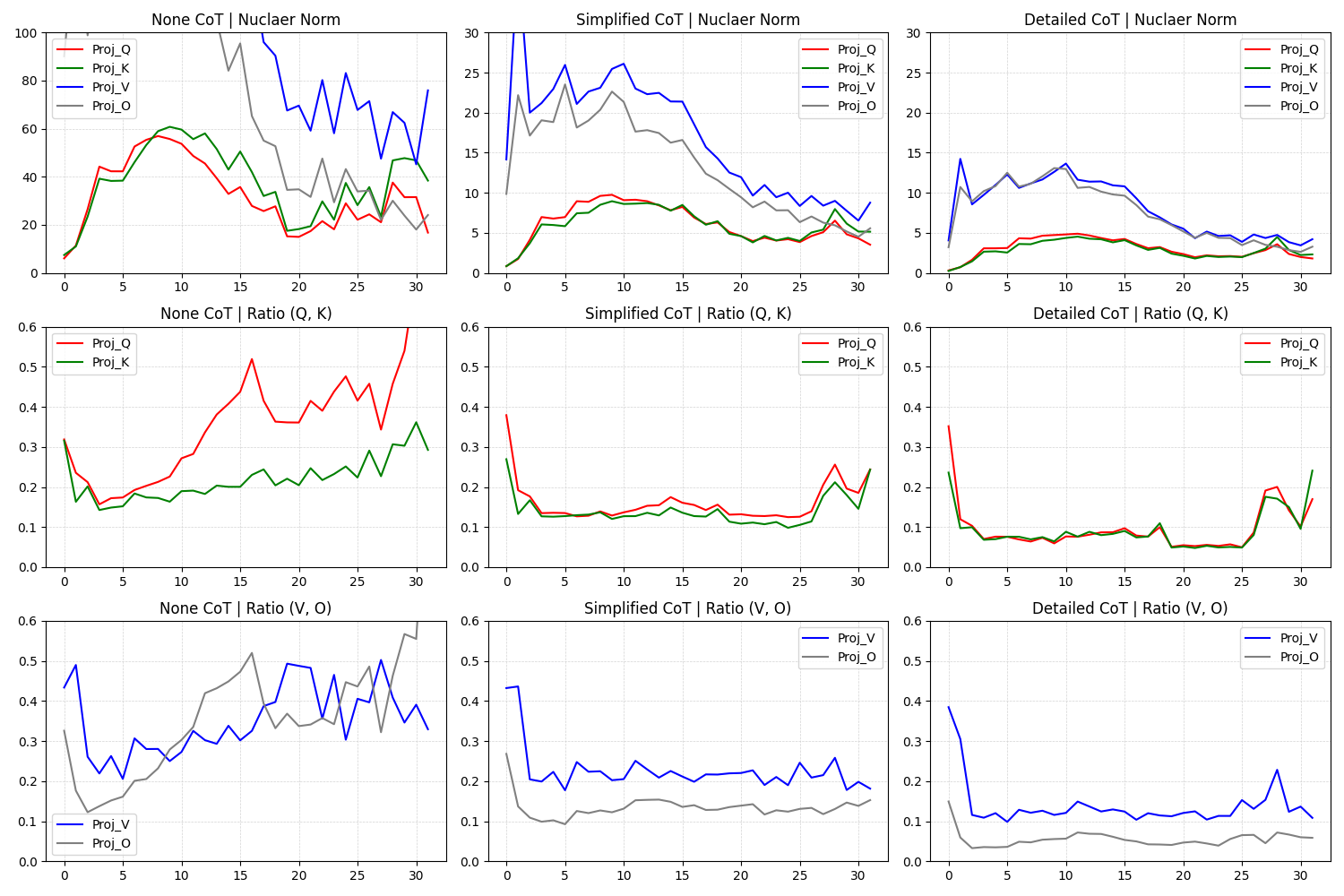}
\caption{Visualization for StrategyQA using Llama-2-7b-chat-hf on irrelevant responses.}
\label{fig:reasoning_nonsense_strategyqa_it}
\vspace{-2mm}
\end{figure*}

\begin{table}[!tbh]
\centering
\small
\scalebox{0.75}{
\begin{tabular}{l|l|l|cccc}
\toprule
\multirow{2}{*}{\textbf{Dataset}} & \multirow{2}{*}{\textbf{Curve}} & \multirow{2}{*}{\textbf{Cot}} &  \multicolumn{4}{c}{\textbf{Mean Absolute Difference (MAD)}} \\
 & &  & Early & Middle & Last & All\\
\midrule
\multirow{24}{*}{ECQA} & \multirow{3}{*}{$s_Q$} 
  & None & 4.99 & 3.02 & 3.68 & 3.78 \\
& & Simplified & 1.45 & 0.88 & 0.83 & 1.03 \\
& & Detailed & 0.63 & 0.44 & 0.44 & 0.48 \\
\cmidrule{2-7}
& \multirow{3}{*}{$s_K$} 
  & None & 5.30 & 4.70 & 5.35 & 4.95 \\
& & Simplified & 1.42 & 1.24 & 1.26 & 1.25 \\
& & Detailed & 0.60 & 0.51 & 0.64 & 0.56 \\
\cmidrule{2-7}
& \multirow{3}{*}{$s_V$} 
  & None & 35.81 & 10.43 & 10.68 & 17.50 \\
& & Simplified & 8.51 & 2.51 & 1.99 & 3.95 \\
& & Detailed & 2.86 & 1.17 & 0.89 & 1.54 \\
\cmidrule{2-7}
& \multirow{3}{*}{$s_O$} 
  & None & 19.70 & 8.54 & 5.41 & 10.47 \\
& & Simplified & 5.10 & 2.19 & 1.18 & 2.61 \\
& & Detailed & 2.00 & 1.15 & 0.62 & 1.17 \\
\cmidrule{2-7}
 & \multirow{3}{*}{$r_Q$} 
  & None & 0.04 & 0.02 & 0.06 & 0.04 \\
& & Simplified & 0.04 & 0.02 & 0.03 & 0.03 \\
& & Detailed & 0.03 & 0.02 & 0.04 & 0.03 \\
\cmidrule{2-7}
& \multirow{3}{*}{$r_K$} 
  & None & 0.02 & 0.01 & 0.05 & 0.03 \\
& & Simplified & 0.03 & 0.02 & 0.03 & 0.03 \\
& & Detailed & 0.02 & 0.03 & 0.04 & 0.03 \\
\cmidrule{2-7}
 & \multirow{3}{*}{$r_V$} 
  & None & 0.06 & 0.03 & 0.07 & 0.05 \\
& & Simplified & 0.05 & 0.02 & 0.04 & 0.04 \\
& & Detailed & 0.04 & 0.01 & 0.03 & 0.03 \\
\cmidrule{2-7}
& \multirow{3}{*}{$r_O$} 
  & None & 0.03 & 0.03 & 0.06 & 0.04 \\
& & Simplified & 0.02 & 0.01 & 0.01 & 0.01 \\
& & Detailed & 0.02 & 0.01 & 0.01 & 0.01 \\
\bottomrule
\end{tabular}
}
\caption{
Statistical results for ECQA using Llama-2-7b-chat-hf on irrelevant responses.
}
\label{tbl:reasoning_cot_500_wrong_answer_shuffle_llama2_7b_it_new_grads_ecqa_train}
\end{table}

\begin{figure*}[!h]
\centering
\includegraphics[width=\linewidth]{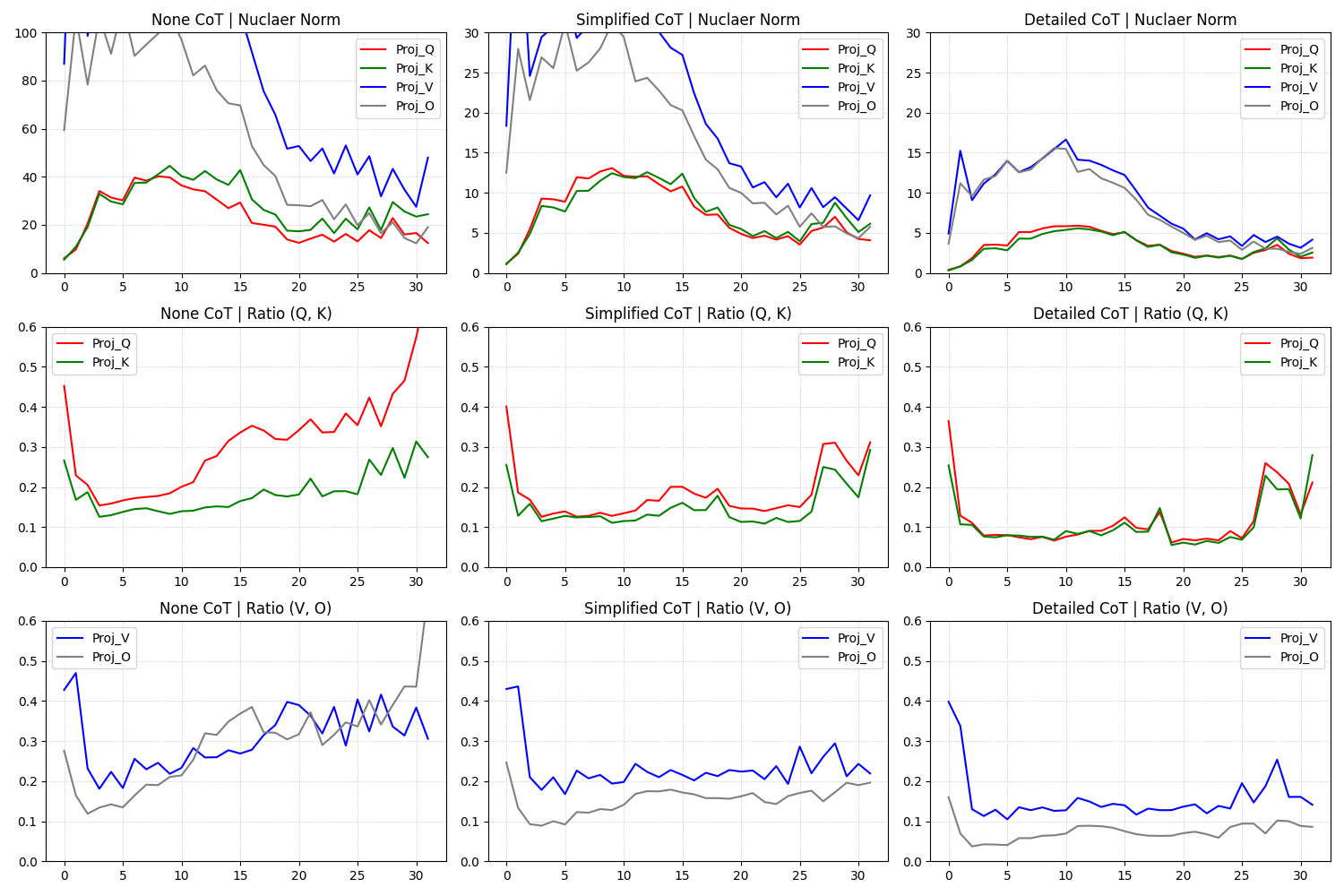}
\caption{Visualization for ECQA using Llama-2-7b-chat-hf on irrelevant responses.}
\label{fig:reasoning_nonsense_ecqa_it}
\vspace{-2mm}
\end{figure*}

\begin{table}[!tbh]
\centering
\small
\scalebox{0.75}{
\begin{tabular}{l|l|l|cccc}
\toprule
\multirow{2}{*}{\textbf{Dataset}} & \multirow{2}{*}{\textbf{Curve}} & \multirow{2}{*}{\textbf{Cot}} &  \multicolumn{4}{c}{\textbf{Mean Absolute Difference (MAD)}} \\
 & &  & Early & Middle & Last & All\\
\midrule
\multirow{24}{*}{CREAK} & \multirow{3}{*}{$s_Q$} 
  & None & 7.78 & 5.74 & 6.80 & 6.54 \\
& & Simplified & 1.60 & 1.16 & 1.03 & 1.23 \\
& & Detailed & 0.61 & 0.43 & 0.48 & 0.49 \\
\cmidrule{2-7}
& \multirow{3}{*}{$s_K$} 
  & None & 8.44 & 7.41 & 9.30 & 8.17 \\
& & Simplified & 1.61 & 1.52 & 1.55 & 1.50 \\
& & Detailed & 0.59 & 0.48 & 0.67 & 0.56 \\
\cmidrule{2-7}
& \multirow{3}{*}{$s_V$} 
  & None & 56.30 & 20.20 & 16.14 & 28.98 \\
& & Simplified & 10.10 & 2.93 & 2.23 & 4.61 \\
& & Detailed & 2.95 & 1.02 & 0.82 & 1.49 \\
\cmidrule{2-7}
& \multirow{3}{*}{$s_O$} 
  & None & 33.37 & 15.48 & 8.03 & 18.13 \\
& & Simplified & 5.86 & 2.53 & 1.24 & 3.00 \\
& & Detailed & 2.02 & 1.01 & 0.58 & 1.13 \\
\cmidrule{2-7}
 & \multirow{3}{*}{$r_Q$} 
  & None & 0.03 & 0.05 & 0.12 & 0.07 \\
& & Simplified & 0.03 & 0.02 & 0.03 & 0.03 \\
& & Detailed & 0.04 & 0.02 & 0.04 & 0.03 \\
\cmidrule{2-7}
& \multirow{3}{*}{$r_K$} 
  & None & 0.03 & 0.02 & 0.07 & 0.04 \\
& & Simplified & 0.02 & 0.02 & 0.03 & 0.02 \\
& & Detailed & 0.02 & 0.02 & 0.04 & 0.03 \\
\cmidrule{2-7}
 & \multirow{3}{*}{$r_V$} 
  & None & 0.06 & 0.04 & 0.07 & 0.05 \\
& & Simplified & 0.05 & 0.02 & 0.04 & 0.03 \\
& & Detailed & 0.04 & 0.01 & 0.03 & 0.03 \\
\cmidrule{2-7}
& \multirow{3}{*}{$r_O$} 
  & None & 0.04 & 0.05 & 0.12 & 0.07 \\
& & Simplified & 0.03 & 0.01 & 0.02 & 0.02 \\
& & Detailed & 0.02 & 0.01 & 0.01 & 0.01 \\
\bottomrule
\end{tabular}
}
\caption{
Statistical results for CREAK using Llama-2-7b-chat-hf on irrelevant responses.
}
\label{tbl:reasoning_cot_500_wrong_answer_shuffle_llama2_7b_it_new_grads_creak_train}
\end{table}

\begin{figure*}[!h]
\centering
\includegraphics[width=\linewidth]{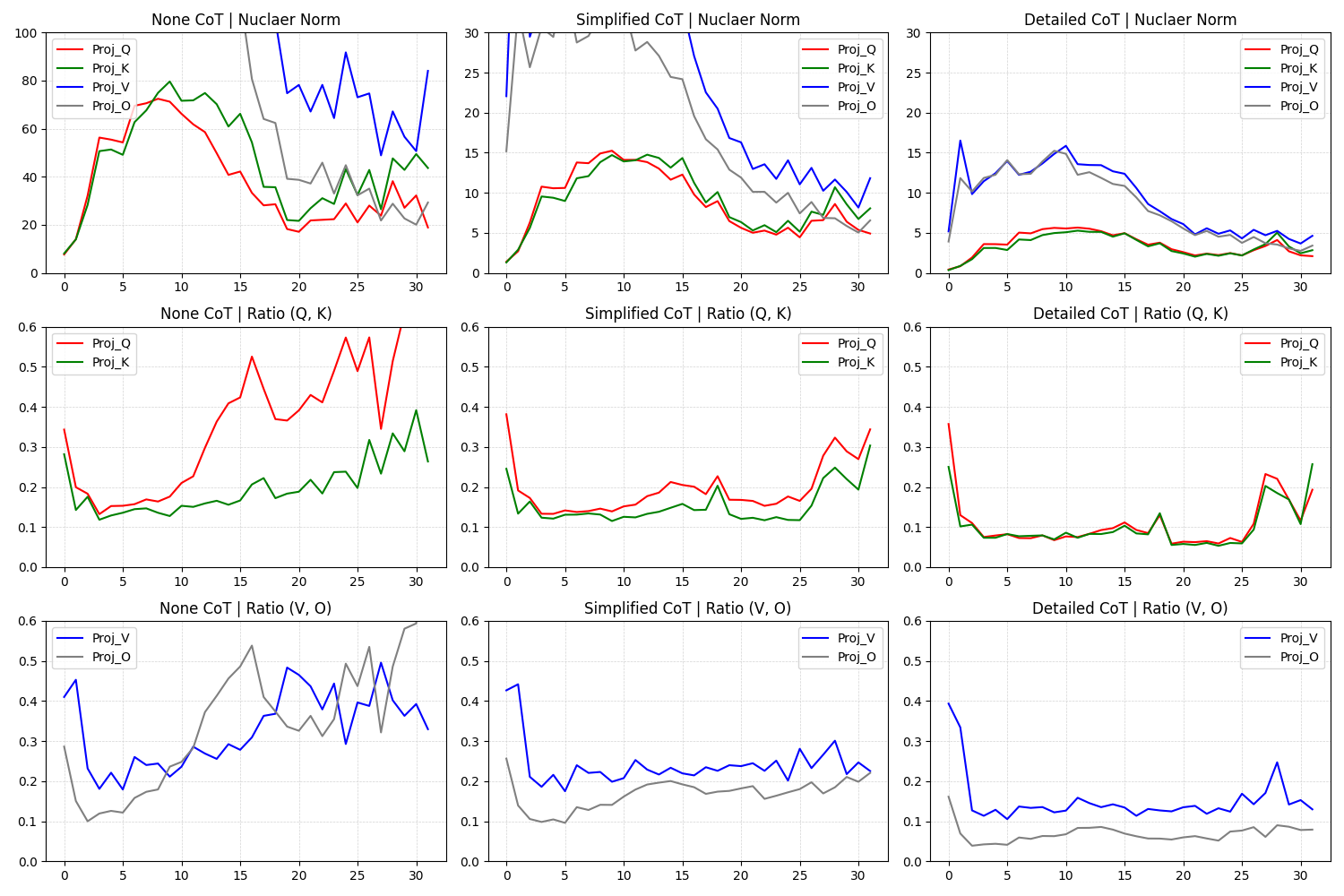}
\caption{Visualization for CREAK using Llama-2-7b-chat-hf on irrelevant responses.}
\label{fig:reasoning_nonsense_creak_it}
\vspace{-2mm}
\end{figure*}

\begin{table}[!tbh]
\centering
\small
\scalebox{0.75}{
\begin{tabular}{l|l|l|cccc}
\toprule
\multirow{2}{*}{\textbf{Dataset}} & \multirow{2}{*}{\textbf{Curve}} & \multirow{2}{*}{\textbf{Cot}} &  \multicolumn{4}{c}{\textbf{Mean Absolute Difference (MAD)}} \\
 & &  & Early & Middle & Last & All\\
\midrule
\multirow{24}{*}{Sensemaking} & \multirow{3}{*}{$s_Q$} 
  & None & 3.98 & 3.08 & 3.24 & 3.39 \\
& & Simplified & 1.69 & 1.21 & 1.31 & 1.39 \\
& & Detailed & 0.61 & 0.46 & 0.43 & 0.48 \\
\cmidrule{2-7}
& \multirow{3}{*}{$s_K$} 
  & None & 4.30 & 3.92 & 4.94 & 4.38 \\
& & Simplified & 1.78 & 1.67 & 1.86 & 1.73 \\
& & Detailed & 0.59 & 0.55 & 0.59 & 0.55 \\
\cmidrule{2-7}
& \multirow{3}{*}{$s_V$} 
  & None & 25.58 & 8.46 & 6.55 & 12.49 \\
& & Simplified & 10.10 & 3.08 & 2.73 & 4.82 \\
& & Detailed & 2.82 & 1.15 & 0.77 & 1.48 \\
\cmidrule{2-7}
& \multirow{3}{*}{$s_O$} 
  & None & 15.93 & 7.37 & 3.43 & 8.40 \\
& & Simplified & 6.08 & 2.75 & 1.54 & 3.23 \\
& & Detailed & 2.00 & 1.10 & 0.56 & 1.14 \\
\cmidrule{2-7}
 & \multirow{3}{*}{$r_Q$} 
  & None & 0.03 & 0.02 & 0.07 & 0.04 \\
& & Simplified & 0.03 & 0.02 & 0.03 & 0.03 \\
& & Detailed & 0.03 & 0.02 & 0.04 & 0.03 \\
\cmidrule{2-7}
& \multirow{3}{*}{$r_K$} 
  & None & 0.04 & 0.01 & 0.05 & 0.03 \\
& & Simplified & 0.03 & 0.02 & 0.03 & 0.02 \\
& & Detailed & 0.02 & 0.02 & 0.04 & 0.03 \\
\cmidrule{2-7}
 & \multirow{3}{*}{$r_V$} 
  & None & 0.05 & 0.03 & 0.06 & 0.04 \\
& & Simplified & 0.05 & 0.02 & 0.05 & 0.04 \\
& & Detailed & 0.04 & 0.01 & 0.03 & 0.03 \\
\cmidrule{2-7}
& \multirow{3}{*}{$r_O$} 
  & None & 0.03 & 0.03 & 0.09 & 0.05 \\
& & Simplified & 0.03 & 0.01 & 0.02 & 0.02 \\
& & Detailed & 0.02 & 0.01 & 0.01 & 0.01 \\
\bottomrule
\end{tabular}
}
\caption{
Statistical results for Sensemaking using Llama-2-7b-chat-hf on irrelevant responses.
}
\label{tbl:reasoning_cot_500_wrong_answer_shuffle_llama2_7b_it_new_grads_sensemaking_train}
\end{table}

\begin{figure*}[!h]
\centering
\includegraphics[width=\linewidth]{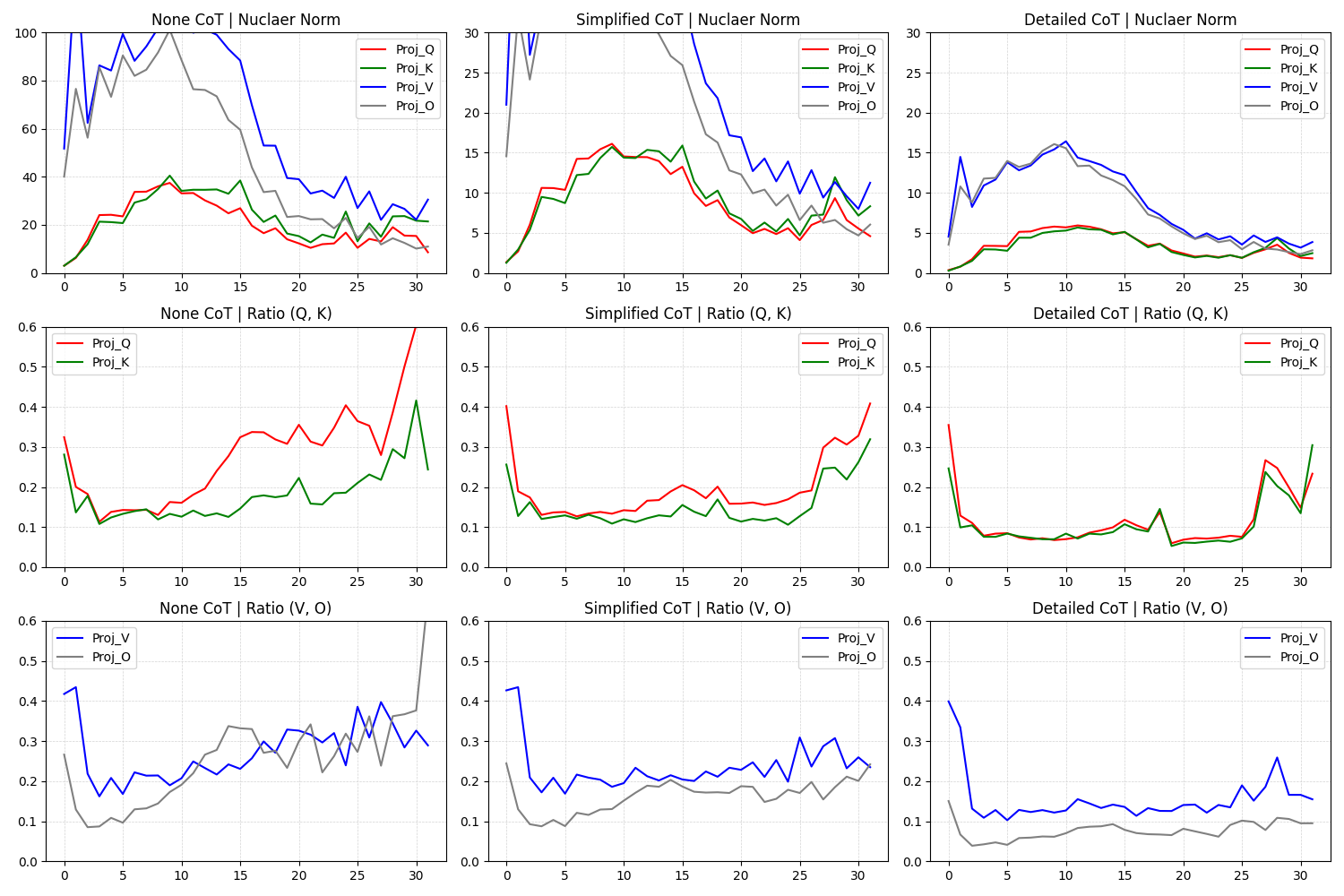}
\caption{Visualization for Sensemaking using Llama-2-7b-chat-hf on irrelevant responses.}
\label{fig:reasoning_nonsense_sensemaking_it}
\vspace{-2mm}
\end{figure*}

\subsubsection{Wiki Tasks}

The visualizations and statistical results on Wiki tasks are shown in Figure \ref{fig:wiki_nonsense_it} and Table \ref{tbl:wiki_wiki_popularity_wrong_answer_shuffle_llama2_7b_it_new_grads}.

\begin{table}[!tbh]
\centering
\small
\scalebox{0.75}{
\begin{tabular}{l|l|l|cccc}
\toprule
\multirow{2}{*}{\textbf{Dataset}} & \multirow{2}{*}{\textbf{Curve}} & \multirow{2}{*}{\textbf{Cot}} &  \multicolumn{4}{c}{\textbf{Mean Absolute Difference (MAD)}} \\
 & &  & Early & Middle & Last & All\\
\midrule
\multirow{32}{*}{Wiki} & \multirow{3}{*}{$s_Q$} 
  & Len 100 & 0.87 & 0.50 & 0.61 & 0.64 \\
& & Len 500 & 0.58 & 0.30 & 0.51 & 0.44 \\
& & Len 1000 & 0.48 & 0.25 & 0.46 & 0.38 \\
& & Unpopular & 1.00 & 0.59 & 0.66 & 0.74 \\
\cmidrule{2-7}
& \multirow{3}{*}{$s_K$} 
  & Len 100 & 0.84 & 0.51 & 0.74 & 0.67 \\
& & Len 500 & 0.53 & 0.30 & 0.64 & 0.48 \\
& & Len 1000 & 0.45 & 0.25 & 0.59 & 0.42 \\
& & Unpopular & 0.98 & 0.60 & 0.77 & 0.76 \\
\cmidrule{2-7}
& \multirow{3}{*}{$s_V$} 
  & Len 100 & 5.33 & 1.19 & 0.55 & 2.14 \\
& & Len 500 & 3.13 & 0.87 & 0.51 & 1.38 \\
& & Len 1000 & 2.55 & 0.75 & 0.49 & 1.16 \\
& & Unpopular & 9.83 & 1.51 & 0.79 & 3.59 \\
\cmidrule{2-7}
& \multirow{3}{*}{$s_O$} 
  & Len 100 & 2.93 & 0.98 & 0.45 & 1.36 \\
& & Len 500 & 2.08 & 0.76 & 0.44 & 1.01 \\
& & Len 1000 & 1.81 & 0.69 & 0.43 & 0.91 \\
& & Unpopular & 5.34 & 1.37 & 0.69 & 2.28 \\
\cmidrule{2-7}
 & \multirow{3}{*}{$r_Q$} 
  & Len 100 & 0.03 & 0.01 & 0.02 & 0.02 \\
& & Len 500 & 0.03 & 0.01 & 0.03 & 0.02 \\
& & Len 1000 & 0.04 & 0.01 & 0.03 & 0.02 \\
& & Unpopular & 0.03 & 0.01 & 0.02 & 0.02 \\
\cmidrule{2-7}
& \multirow{3}{*}{$r_K$} 
  & Len 100 & 0.03 & 0.01 & 0.02 & 0.02 \\
& & Len 500 & 0.02 & 0.01 & 0.03 & 0.02 \\
& & Len 1000 & 0.02 & 0.01 & 0.03 & 0.02 \\
& & Unpopular & 0.03 & 0.01 & 0.02 & 0.02 \\
\cmidrule{2-7}
 & \multirow{3}{*}{$r_V$} 
  & Len 100 & 0.05 & 0.02 & 0.02 & 0.03 \\
& & Len 500 & 0.04 & 0.01 & 0.02 & 0.02 \\
& & Len 1000 & 0.04 & 0.01 & 0.02 & 0.02 \\
& & Unpopular & 0.06 & 0.02 & 0.02 & 0.03 \\
\cmidrule{2-7}
& \multirow{3}{*}{$r_O$} 
  & Len 100 & 0.02 & 0.01 & 0.01 & 0.01 \\
& & Len 500 & 0.02 & 0.01 & 0.00 & 0.01 \\
& & Len 1000 & 0.02 & 0.01 & 0.00 & 0.01 \\
& & Unpopular & 0.02 & 0.01 & 0.01 & 0.01 \\
\bottomrule
\end{tabular}
}
\caption{
Statistical results for Wiki using Llama-2-7b-chat-hf on irrelevant responses.
}
\label{tbl:wiki_wiki_popularity_wrong_answer_shuffle_llama2_7b_it_new_grads}
\end{table}

\begin{figure*}[!h]
\centering
\includegraphics[width=\linewidth]{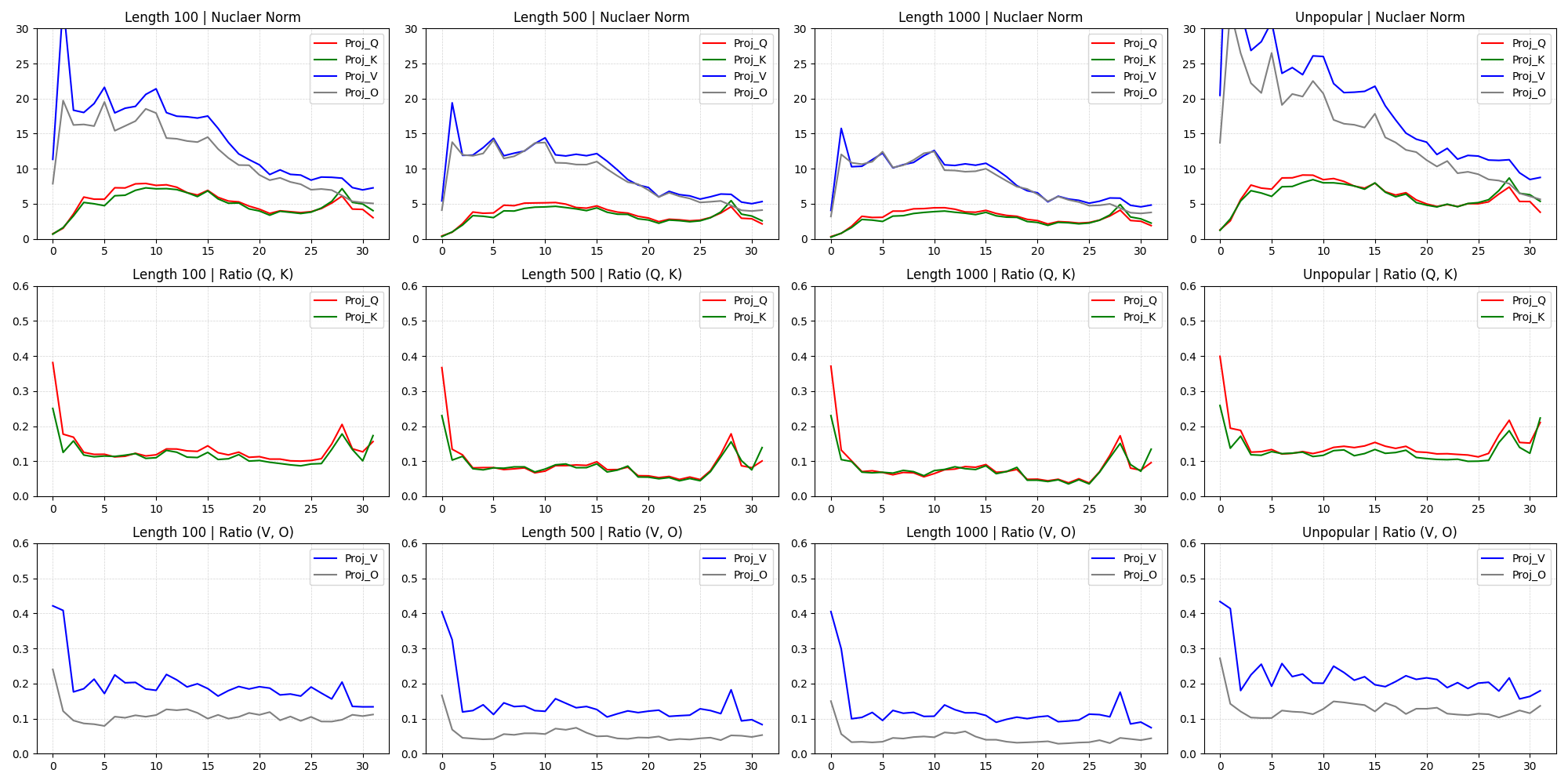}
\caption{Visualization for Wiki tasks using Llama-2-7b-chat-hf on irrelevant responses.}
\label{fig:wiki_nonsense_it}
\vspace{-2mm}
\end{figure*}


\end{document}